\documentclass[letterpaper, 10 pt, conference]{ieeeconf}  % Comment this line out if you need a4paper

\IEEEoverridecommandlockouts
\let\labelindent\relax

\usepackage{array}
\usepackage{graphicx}
\usepackage{amsmath,amssymb} % define this before the line numbering.
\usepackage{amssymb}
\usepackage{multirow}
\usepackage{enumitem}
\usepackage{amsthm}
\usepackage{hhline}
\usepackage{pifont}
\usepackage{xcolor}
\usepackage[11pt]{moresize}
\usepackage{setspace}

\newtheorem{claim}{Claim}

\usepackage{gensymb}
\usepackage{hyperref}

\newcommand{\bX}{\mathbf{X}}
\newcommand{\bY}{\mathbf{Y}}

\newcommand{\bx}{\mathbf{x}}
\newcommand{\by}{\mathbf{y}}

\newcommand{\bw}{\mathbf{w}}

\newcommand{\bv}{\mathbf{v}}

\newcommand{\bp}{\mathbf{p}}

\newcommand{\bZ}{\mathbf{Z}}

\newcommand{\be}{\mathbf{e}}

\newcommand{\argmax}[1]{\underset{#1}{\text{argmax}} \:}

\newcommand{\minarg}[1]{\underset{#1}{\text{min}} \:}

\def\be {\begin{equation}}
\def\ee {\end{equation}}
\def\beas {\begin{eqnarray*}}
	\def\eeas {\end{eqnarray*}}
\def\bea {\begin{eqnarray}}
\def\eea {\end{eqnarray}}

\makeatletter
\usepackage{xspace}
\def\@onedot{\ifx\@let@token.\else.\null\fi\xspace}
\DeclareRobustCommand\onedot{\futurelet\@let@token\@onedot}

\newcommand{\figref}[1]{Fig\onedot~\ref{#1}}

\newcommand{\secref}[1]{Sec\onedot~\ref{#1}}
\newcommand{\tabref}[1]{Tab\onedot~\ref{#1}}

\newcommand{\clmref}[1]{Claim~\ref{#1}}

\newcolumntype{C}[1]{>{\centering\let\newline\\\arraybackslash\hspace{0pt}}m{#1}}

\def\ie{\emph{i.e}\onedot} 
 
\def\etc{\emph{etc}\onedot}

\title{\LARGE \bf Pose Estimation for Objects with Rotational Symmetry}

\author{Enric Corona, Kaustav Kundu, Sanja Fidler% <-this % stops a space
 \thanks{Enric Corona, Kaustav Kundu and Sanja Fidler are with Department of Computer Science,
        University of Toronto, and the Vector Institute, S.F. is also with NVIDIA. {\tt\footnotesize \{ecorona,kkundu,fidler\}@cs.toronto.edu} }
        }%

\begin{document}

\maketitle
\thispagestyle{empty}
\pagestyle{empty}

%%%%%%%%% ABSTRACT
\begin{abstract}
Pose estimation is a widely explored problem, enabling many robotic tasks such as grasping and manipulation. In this paper, we tackle the problem of pose estimation for objects that exhibit rotational symmetry, which are common in man-made and industrial environments. 
In particular, our aim is to infer poses for objects not seen at training time, but for which their 3D CAD models are available at test time. 
Previous work has tackled this problem by learning to compare captured views of real objects with the rendered views of their 3D CAD models, by embedding them in a joint latent space using neural networks. We show that sidestepping the issue of symmetry in this scenario during training leads to poor performance at test time.  
We propose a model that reasons about rotational symmetry during training by having access to only a small set of symmetry-labeled objects, whereby exploiting a large collection of unlabeled CAD models. 
We demonstrate that our approach significantly outperforms a naively trained neural network on a new pose dataset containing images of tools and hardware.
\end{abstract}

%%%%%%%%% BODY TEXT
\section{Introduction}
\label{sec:intro}

In the past few years, we have seen significant advances in domains such as autonomous driving~\cite{kitti_dataset}, control for flying vehicles~\cite{airsim}, warehouse automation popularized by the Amazon Picking Challenge~\cite{zeng2016multi}, and navigation in complex environments~\cite{gupta17}. %However, there is still a path to be paved towards reaching full autonomy.
%Object picking has been a well-explored problem and popularized with the recent Amazon Picking Challenge. In order to plan its grasp, an automated method needs to be able to accurately predict 
%the 3D pose of the object from visual observation. 
Most of these domains rely on accurate estimation of 3D object pose. For example, in driving, understanding object pose helps us to perceive the traffic flow, while in the object picking challenge knowing the pose helps us grasp the object better. 

The typical approach to pose estimation has been to train a neural network to directly regress to object pose from the RGB or RGB-D input~\cite{zeng2016multi,Gupta2015}. However, this line of work requires a reference coordinate system for each object to be given in training, and thus cannot handle novel objects at test time. %While this may be viable for domains with a smaller set of classes of interest such as in road scenarios, we might need a more scalable solution in general. 
In many domains such as for example automated assembly where robots are to be deployed to different warehouses or industrial sites, handling novel objects is crucial. In our work, we tackle the problem of pose estimation for objects both, seen or unseen in training time.

%, and estimate 6DoF pose by aligning a CAD model using the predicted initial pose via ICP. 
%However, these approaches typically completely sidestep the issues of rotationally symmetric objects. 

In such a scenario, one typically assumes to be given a reference 3D model for each object at test time, and the goal is to estimate the object's pose from visual input with reference to this model~\cite{linemod}.  Most methods tackle this problem by comparing the view from the scene with a set of rendered viewpoints via either hand designed similarity metrics~\cite{linemod}, or learned embeddings~\cite{wohlhart2015learning,air}. %The main idea is to learn to embed an image and a rendered view into a latent representation, such that the similarity between the embeddings of the correct viewpoint is the highest.  
The main idea is to embed both a real image and a rendered CAD view into a joint embedding space, such that the true viewpoint pair scores the highest similarity among all alternatives. Note that this is not a trivial task, as the rendered views may look very different from objects in real images, both because of different background, lighting, and possible occlusion that arise in real scenes. Furthermore, CAD models are typically not textured/colored and thus only capture the geometry of the objects but not their appearance.
%Here, we follow this line of work. %, by learning joint latent embeddings of real visual input and rendered viewpoints of 3D CAD models for pose estimation. 

In man-made environments, most objects such as tools/hardware have simple shapes with diverse symmetries (Fig.~\ref{fig:concept}). One common symmetry is a rotational symmetry which occurs when an object is equivalent under certain 3D rotations. Such objects are problematic for the embedding-based approaches,  since multiple rendered views may look exactly the same (or very similar), leading to ambiguities in  the standard loss functions that rely on negative examples. 
Most existing work has sidestepped the issue of symmetry, which we show has a huge impact on performance. In this paper, we tackle the problem of training embeddings for pose estimation by reasoning about rotational symmetries. %We assume a weakly labeled setting, in which a only small set of symmetry-labeled objects needs to be provided. 

%Training such embeddings requires access to positive and (difficult) negative examples for objects seen during training. This, however, is problematic for objects exhibiting rotational symmetries, since multiple object's views look exactly the same.  

\setlength\tabcolsep{1.5pt}
\vspace{-0mm}
\begin{figure}[t!]
    \includegraphics[trim={0 0mm 0 2mm}, clip = true,height=1.4cm]{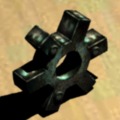}\hspace{0.7mm}
    \includegraphics[trim={0 17mm 0 17mm}, clip = true,height=1.4cm]{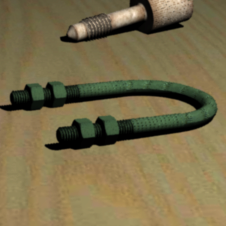}\hspace{0.7mm}
    \includegraphics[trim={0 9mm 0 9mm}, clip = true,height=1.4cm]{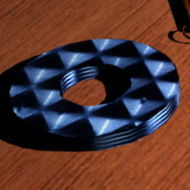}\hspace{0.7mm}
    \includegraphics[trim={0 15mm 0 15mm}, clip = true,height=1.4cm]{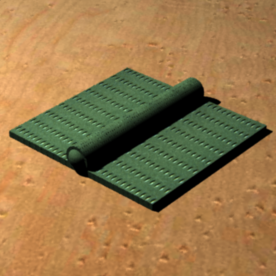}\hspace{0.7mm}
\\[0.1mm]
\addtolength{\tabcolsep}{3pt}
\begin{scriptsize}
	\begin{tabular}{ ccccc }
	\hspace{-0mm}\includegraphics[trim={0 0 0cm 0}, clip = true,width=0.9cm]{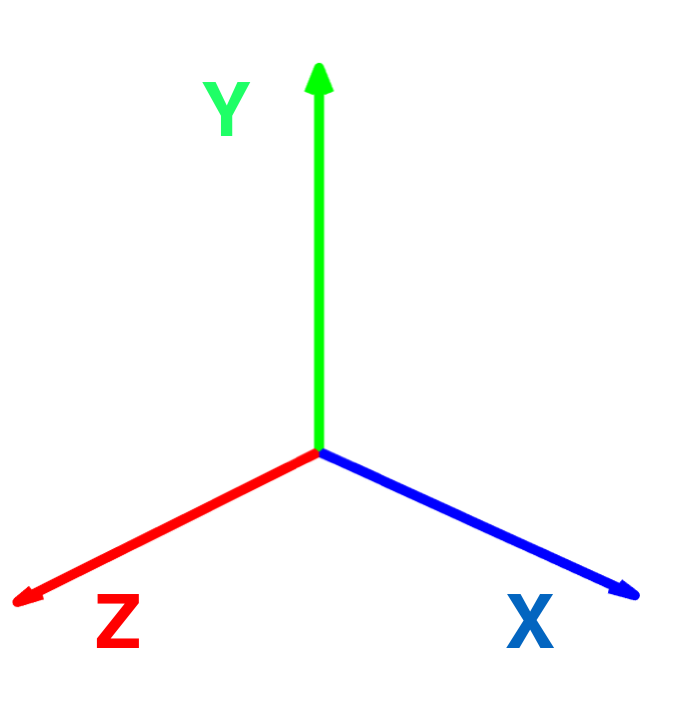}&\hspace{-0mm}
	\includegraphics[trim={6cm 10mm 6cm 20mm}, clip = true,height=1.1cm]{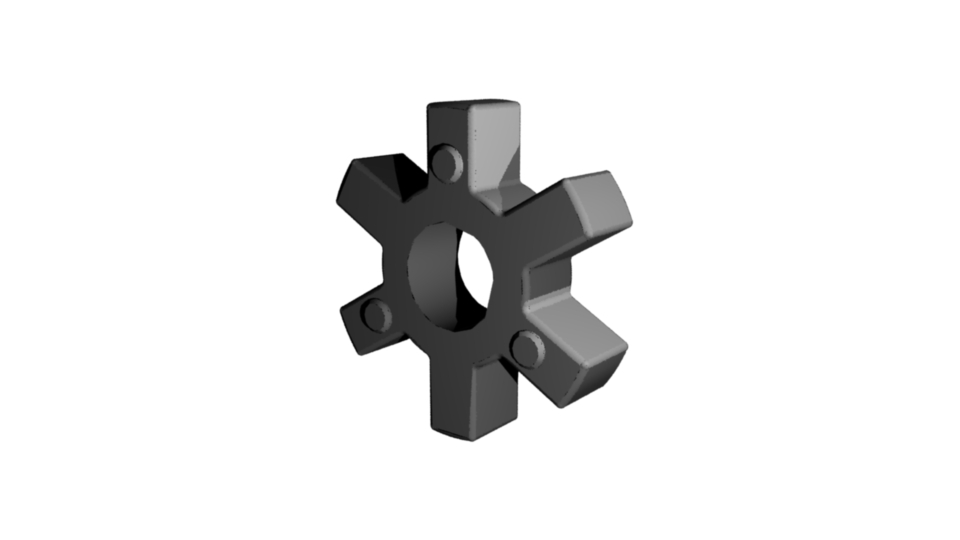}&
	\includegraphics[trim={10cm 0 6cm 1mm}, clip = true,height=1.1cm]{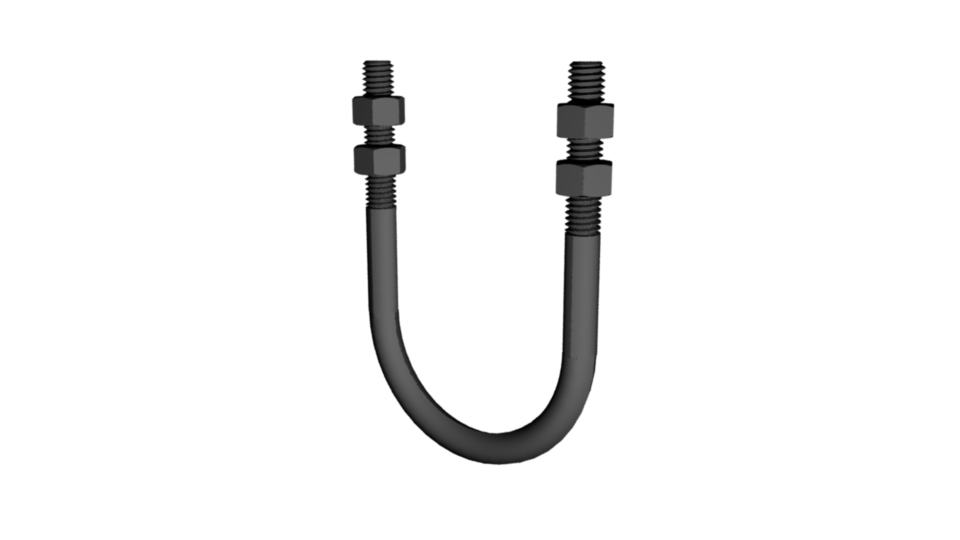}&\hspace{0mm}
	\includegraphics[trim={6cm 10mm 6cm 10mm}, clip = true,height=1.1cm]{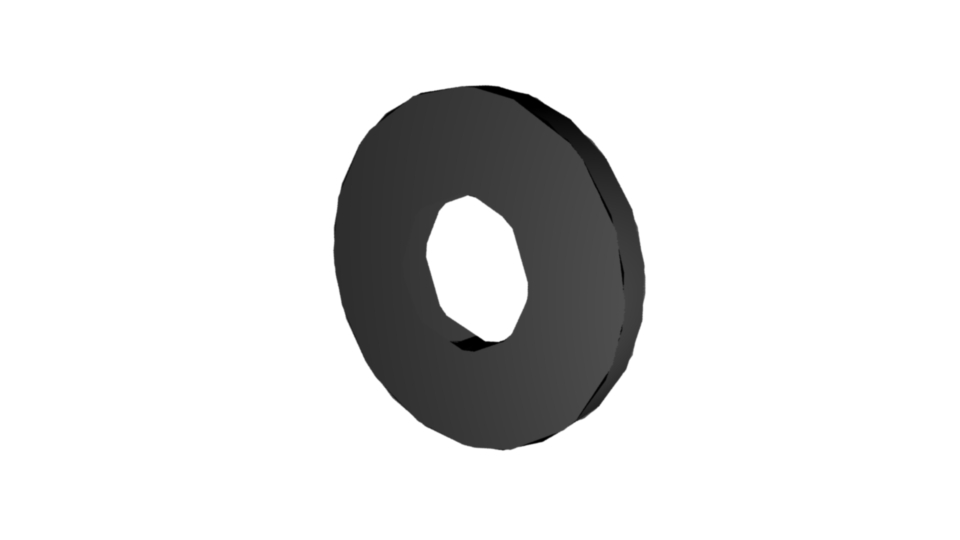}&
	\includegraphics[trim={6cm 25mm 6cm 35mm}, clip = true,height=1.1cm]{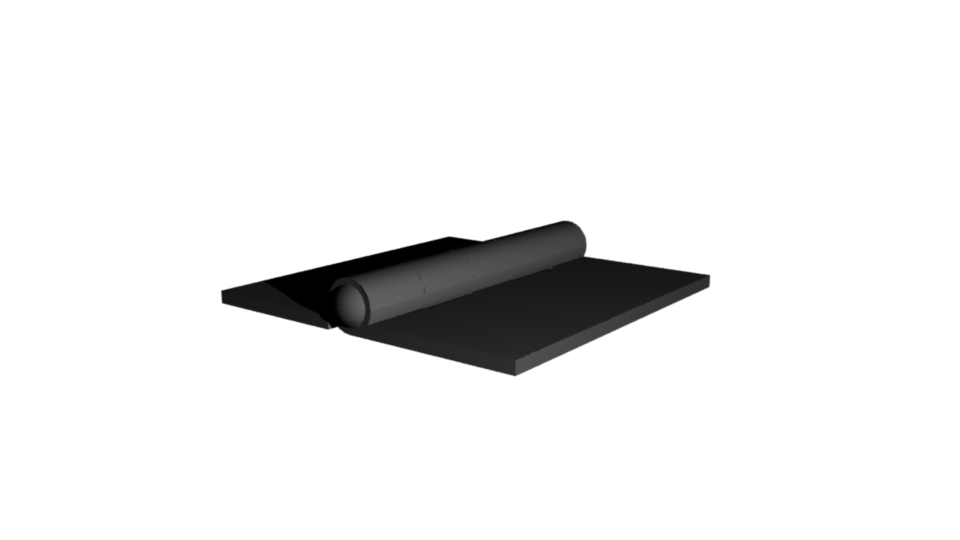}\\[-1mm]
	 & \hspace{-0mm}	$X \sim 2$ & $X \sim 1$ &	$X \sim 2$ & $X \sim 1$ \\
	 & \hspace{-0mm}	$Y \sim 2$ & $Y \sim 2$ &	$Y \sim 2$ & $Y \sim 2$ \\
	& \hspace{-0mm}	$Z \sim 6$ & $Z \sim 1$ &	$Z \sim \infty$ & $Z \sim 1$ \\[-1.2mm]
	\end{tabular}
	\end{scriptsize}
	\centering
	\vspace{0.1mm}
	\caption{Many industrial objects such as various tools exhibit rotational symmetries. In our work, we address pose estimation for such objects.}
	\label{fig:concept}
	\vspace{-2.6mm}
\end{figure}
\setlength\tabcolsep{6pt}

%In man-made environments, most tools/hardware have simple shapes with diverse symmetries (Fig.~\ref{fig:concept}). However, most existing works have sidestepped the issue of symmetry, which we show has a huge impact on performance. In this paper, we tackle the problem of training embeddings for pose estimation by reasoning about rotational symmetries. 

%Since accurate pose for real objects is time-consuming to collect, such datasets are typically very small, and thus training powerful neural models is challenging. Here, we aim to leverage a large collection of 3D CAD models in order to augment real datasets. However, while intricate, rotational symmetries are rarely provided with 3D models.

We propose a neural model to pose estimation by learning to compare real views of objects to the viewpoints rendered from their CAD models. 
Our model reasons about rotational symmetry during training by having access to only a small set of symmetry-labeled objects, whereby exploiting a large collection of unlabeled CAD models crawled from the web. 
%We show how to jointly infer rotational symmetries on a large dataset of 3D CAD models and exploit them to train better embeddings for pose estimation. 
We evaluate our approach on a new dataset for pose estimation, that allows us to carefully evaluate the effect of symmetry on performance. We show that our approach, which infers symmetries, significantly outperforms a naively trained neural network. Our code and data are online: {\color{magenta}{{\footnotesize \url{http://www.cs.utoronto.ca/~ecorona/symmetry_pose_estimation/index.html}}}}.

\section{Related Work}
\label{sec:related}

While many pose estimation methods exist, we restrict our review to work most related to ours. 

%\vspace{-4mm}
\paragraph{Pose Estimation} Pose estimation has been treated as either a classification task, \ie, predicting a coarse viewpoint~\cite{Gupta2015,tulsiani2015pose}, or as a  regression problem~\cite{doumanoglou2016siamese,3dop,kehl2017ssd, xiang2016objectnet3d}.
%~\cite{brachmann2014learning,doumanoglou2016siamese,3dop,kehl2017ssd}. %However, it is challenging for such approaches to generalize to unseen objects and accurately predict the pose of arbitrarily complex objects found in warehouses.
However, such methods inherently assume consistent viewpoint annotation across objects in training, and cannot infer poses for objects belonging to novel classes at test time.

Alternatively, one of the traditional approaches for pose estimation is that of matching a given 3D model to the image. Typical matching methods for pose estimation involve computing multiple local feature descriptors or a global descriptor from the input, followed by a matching procedure with either a 3D model or a coarse set of examplar viewpoints. %Aligning an input image with a 3D model has been one of the first approaches explored in the community. 
%Since the statistics of an image and the 3D model differ vastly, various (hand-crafted) descriptors were used%
%~\cite{ulrich2009cad} 
%\cite{payet2011contours,strzodka2003graphics,ulrich2009cad} 
%to get a coarse pose prediction. 
Precise alignment to a CAD model was then posed as an optimization problem using RANSAC, Iterative Closest Point (ICP)~\cite{besl1992method}, Particle Swarm Optimization (PSO)~\cite{eberhart1995new}, or variants~\cite{huttenlocher1990recognizing,lpt2013ikea}. % {\color{red}{Do we really need all these refs?}}%One of the main challenges of aligning an input image with a 3D model is that the pose parameters belong in the six dimensional continuous space.  

\iffalse
For matching the input image to a set of 2D templates, L2 distance or cosine similarity is typically used. Matching functions such as hashing~\cite{kehl2016hashmod} or edge based distance functions~\cite{david2003simultaneous,linemod,liu2010pose} have also been explored. To improve the robustness of the matching procedures, local level patches descriptors are refined through geometric filtering~\cite{michel2016global,zach2015dynamic} 
%~\cite{hao2013efficient,michel2016global,zach2015dynamic} 
and matched by a voting based procedure~\cite{Aubry14,sun2010depth,tejani2014latent}, 
%~\cite{Aubry14,rodrigues20126d,sun2010depth,tejani2014latent}, 
geometry constrained matching~\cite{zhu2014single} or posing it as an optimization problem~\cite{collet2009object,lpt2013ikea,pavlakos20176,payet2011contours,wong2017segicp,zhang2013joint}. One of the challenges in such approaches have been the fact that handcrafted feature descriptors are not robust to the small pose changes or occlusions. 
\fi

%\vspace{-4mm}
\paragraph{Learning Embeddings for Pose Estimation} Following the recent developments of CNN-based Siamese networks~\cite{han2015matchnet,luo2016efficient} for matching, CNNs have also been used for pose  estimation~\cite{wohlhart2015learning,kehl2016deep,krull2015learning, xiang2017posecnn}. CNN extracts image/template representation, and uses L2 distance or cosine similarity for matching. Typically such networks are trained in an end to end fashion to minimize and maximize the L2 distance between the pairs of matches and non matches, respectively~\cite{wohlhart2015learning}. \cite{krull2015learning} sample more views around the top predictions and iteratively refine the matches. Training such networks require positive and negative examples. Due to rotational symmetric objects found in industrial settings, it is not trivial to determine the negative examples.

\paragraph{Symmetry in 3D Objects} Symmetry is a well studied property. There have been various works on detecting reflectional/bilateral symmetry~\cite{li_symmetry,symmetrization,summary_symmetry,tulsiani2015}, 
%~\cite{li_symmetry,mitra2006,symmetrization,summary_symmetry,tulsiani2015}, 
medial axes~\cite{tsogkas2017amat}, symmetric parts~\cite{sie2013detecting}. %Approaches such as~\cite{carreira2016lifting, gao2016symmetric, murthy2017reconstructing} use bilateral symmetry annotations~\cite{shapenet} to reconstruct more accurately. 
Please refer to~\cite{hermann1952symmetry} for a detailed review of different types of symmetry. For pose estimation, handling rotational symmetry is very important~\cite{air}, a problem that we address here.

The problem of detecting rotational symmetries has been explored extensively~\cite{cho2009bilateral, flynn19943, labonte1993perceptually, Symmetry17}. 
%~\cite{cho2009bilateral, flynn19943, kondra2013multi,  labonte1993perceptually, prasad2005detecting, Symmetry17}. 
These approaches identify similar local patches via handcrafted features. Such patches are then grouped to predict the rotational symmetry orders along different axes. In comparison, our approach works in an end to end manner and is trained jointly with the pose estimation task. \cite{martinet2006accurate} proposes to detect symmetries by computing the extrema of the generalized moments of the 3D CAD model. Since this results in a number of false positives, a post-processing step is used to prune them. However, since their code is not public, a head-to-head comparison is hard. Recently~\cite{cohen2016group, dieleman2016exploiting} 
%~\cite{cohen2016group, dieleman2016exploiting, gens2014deep} 
introduced 2D rotation invariance in CNNs. However extending these approaches to 3D rotation is not trivial due to computational and memory overhead.

%{\color{blue} Approaches such as~\cite{hi} which detect rotational symmetry in images cannot be scaled to 3D models due to the additional dimension.}
\cite{tless} introduced a dataset for pose estimation where objects with one axis of rotational symmetry have been annotated. However, most objects in industrial settings have multiple axes of rotational symmetry. Approaches such as~\cite{bb8,bregier2017symmetry}  use these symmetry labels to modify the output space at test time. Since annotating rotational symmetries is hard, building large scale datasets with symmetry labels is expensive and time consuming. We show that with a small set of symmetry labels, our approach can be extended to predict rotational symmetries about multiple axes, which in turn can help to learn better embeddings for pose estimation. %Moreover, we also show that predicting the rotational symmetry about multiple axes is not independent and how we can explore the constraints on the orders of rotational symmetry on pairs and triplets of axes to improve the symmetry prediction.

\section{Our Approach}
\label{sec:method}

\begin{figure}[t!]
\vspace{2mm}
\scalebox{0.65}[0.65]{
\addtolength{\tabcolsep}{-4.5pt}
	\begin{tabular}{cccccc}
		\includegraphics[trim={0 0 0 0}, clip = true,width=2cm]{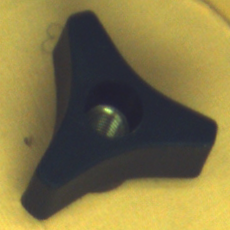}  &
		\includegraphics[trim={14cm 6.5cm 13cm 6cm}, clip = true, width=2cm]{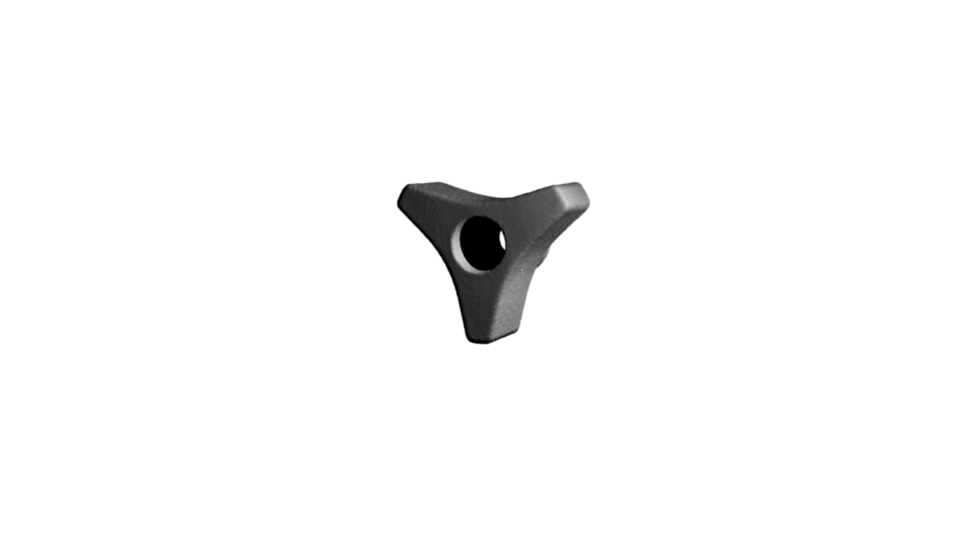} &
		\includegraphics[trim={3.5cm  1.5cm 3.5cm 1.5cm }, clip = true,width=2cm]{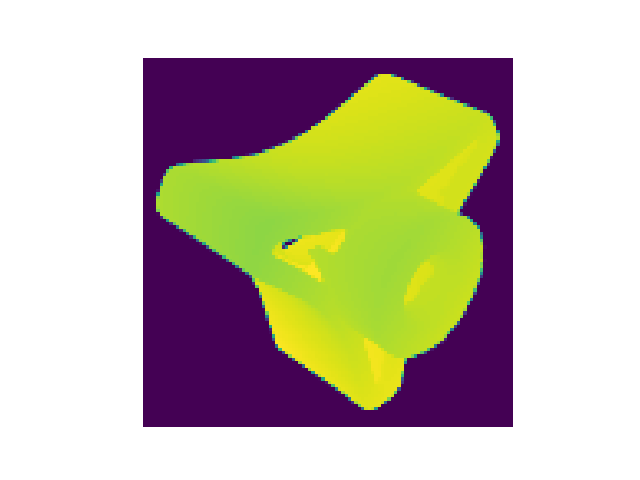}&
		\includegraphics[trim={3.5cm  1.5cm 3.5cm 1.5cm }, clip = true,width=2cm]{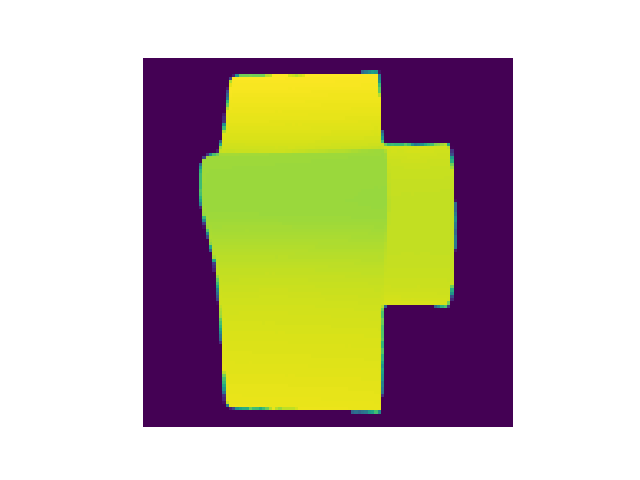} &
		\includegraphics[trim={3.5cm  1.5cm 3.5cm 1.5cm }, clip = true,width=2cm]{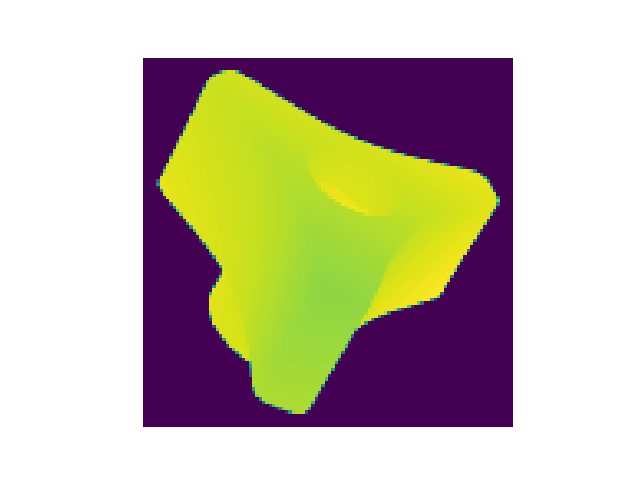} &
		\includegraphics[trim={3.5cm  1.5cm 3.5cm 1.5cm }, clip = true,width=2cm]{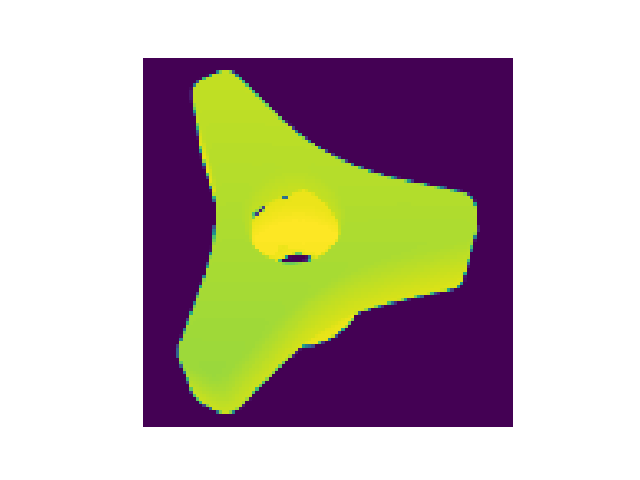} \\
		(image) & (CAD model) & \multicolumn{4}{c}{(rendered CAD views (depth maps))}\\
	\end{tabular}
	}
	\centering
	\vspace{-0.5mm}
	\caption{\small Our problem entails estimating the pose of the object in the image (left) given its CAD model. We exploit rendered depth images of the CAD model in order to determine the pose.}
	\label{fig:problem}
	\vspace{-4mm}
\end{figure}

We tackle the problem of pose estimation in the presence of rotational symmetry. In particular, we assume we are given a test RGB image of an object unseen at training time, as well as the object's 3D CAD model. Our goal is to compute the pose of the object in the image by matching it to the rendered views of the CAD model. To be robust to mismatches in appearance between the real image and the textureless CAD model, we exploit rendered depth maps instead of RGB views. 
Fig.~\ref{fig:problem} visualizes an example image of an object, the corresponding 3D model, and rendered views. 

Our approach follows~\cite{wohlhart2015learning} in learning  a neural network that embeds a real image and a synthetic view in the joint semantic space. In order to perform pose estimation, we then find the view closest to the image in this embedding space. As typical in such scenarios, neural networks are trained with e.g. a triplet loss, which aims to embed the matching views closer than any of the non-matching views. However, in order for this loss function to work well, ambiguity with respect to rotational symmetry needs to be resolved. That is, due to equivalence of shape under certain 3D rotations for rotationally symmetric objects, certain rendered viewpoints look exactly the same. If such  views are used as negative examples during training, we may introduce an ambiguity that prevents us in learning a better model. Note that this is not true for other symmetries such as a reflective symmetry, since the object does not necessarily have equivalent shape in different 3D poses. 
 We propose to deal with this issue by inferring reflectional symmetries of objects, and exploiting them in designing a more robust loss function. 

We first provide basic concepts of rotational symmetry in~\secref{sec:rot_sym}. In Sec.~\ref{sec:method}, we propose our neural network for joint pose and symmetry estimation, and introduce a loss function that takes into account equivalence of certain views. %In Subsec.~\ref{sec:rot_infer}, we propose a network that infers orders of symmetry for an object using only a small number of training examples. 
We show how to train this network by requiring only a small set of symmetry-labeled objects, and by exploiting a large collection of unlabeled CAD models. 

\vspace{-2mm}
\subsection*{Rotational Symmetry}
\label{sec:rot_sym}

We start by introducing notation and basic concepts. 

%\vspace{-4mm}
\paragraph{Rotation Matrix} We denote a rotation of an angle $\phi $ around an axis $ \theta $ using a matrix $ \mathbf{R}_{\theta} (\phi) $. For example, if the axis of rotation is the X-axis, then
\vspace{-1mm}
\[
\mathbf{R}_{X} (\phi) = \left[ \begin{array}{ccc}
1 & 0 & 0 \\
0 & \cos{\phi} & - \sin{\phi}   \\
0 & \sin{\phi}  & \cos{\phi}  \\
\end{array} \right]
\]
\paragraph{Order of Rotational Symmetry} We say that an object  has an $n$  order of rotational symmetry around the axis $ \theta $, \ie, $ \mathcal{O}(\theta) = n $, when its 3D shape is equivalent to its shape rotated by $ \mathbf{R}_{\theta} \left( \tfrac{2 \pi i}{n} \right), \forall i \in \{0, \ldots, n - 1\} $. % Thus, $ \mathcal{O}_{\theta} (x) = \mathcal{O}_{\theta} (x) =  $

\begin{figure}[t!]
\vspace{-0mm}
	\scalebox{0.415}[0.415]{
	\addtolength{\tabcolsep}{18pt}
		\begin{tabular}{ccc}
			\raisebox{0.2\height}{\includegraphics[trim={0 0 0 0}, clip = true,width=3.6cm]{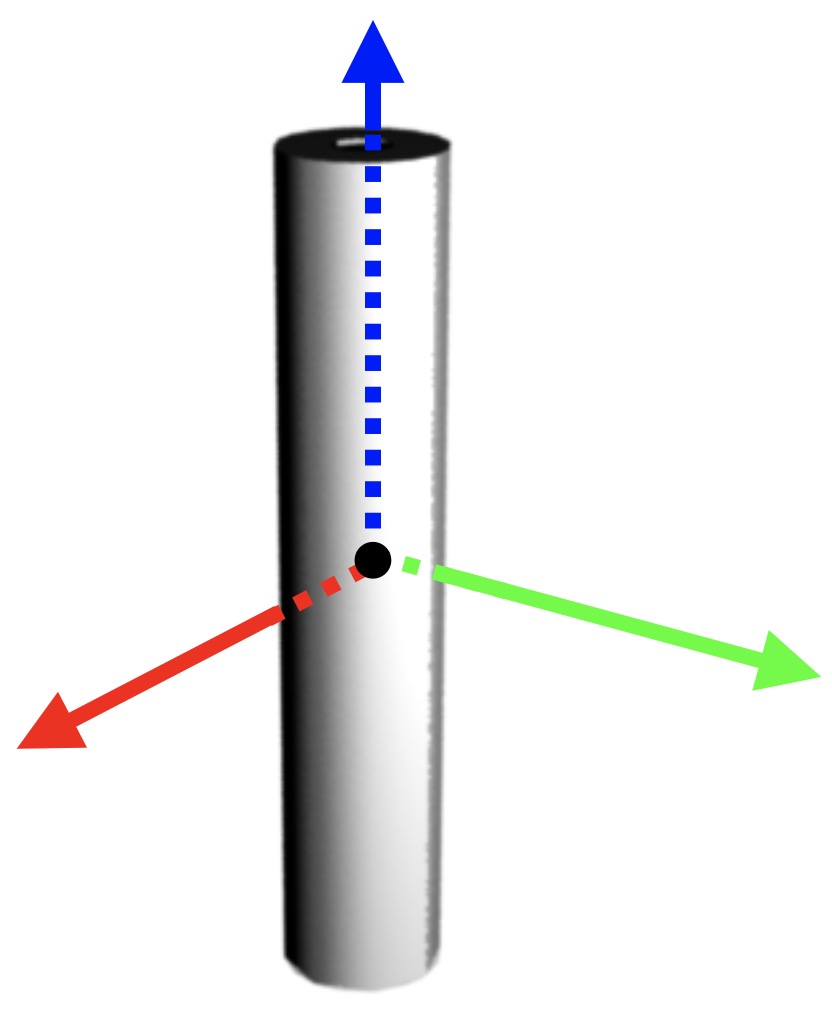}}
			\put(-110,44){\Large $ \bX $} 
			\put(-9,50){\Large  $ \bY $}
			\put(-62,149){\Large $ \bZ $} &
			\includegraphics[trim={0 0 0 0}, clip = true,width=5.7cm]{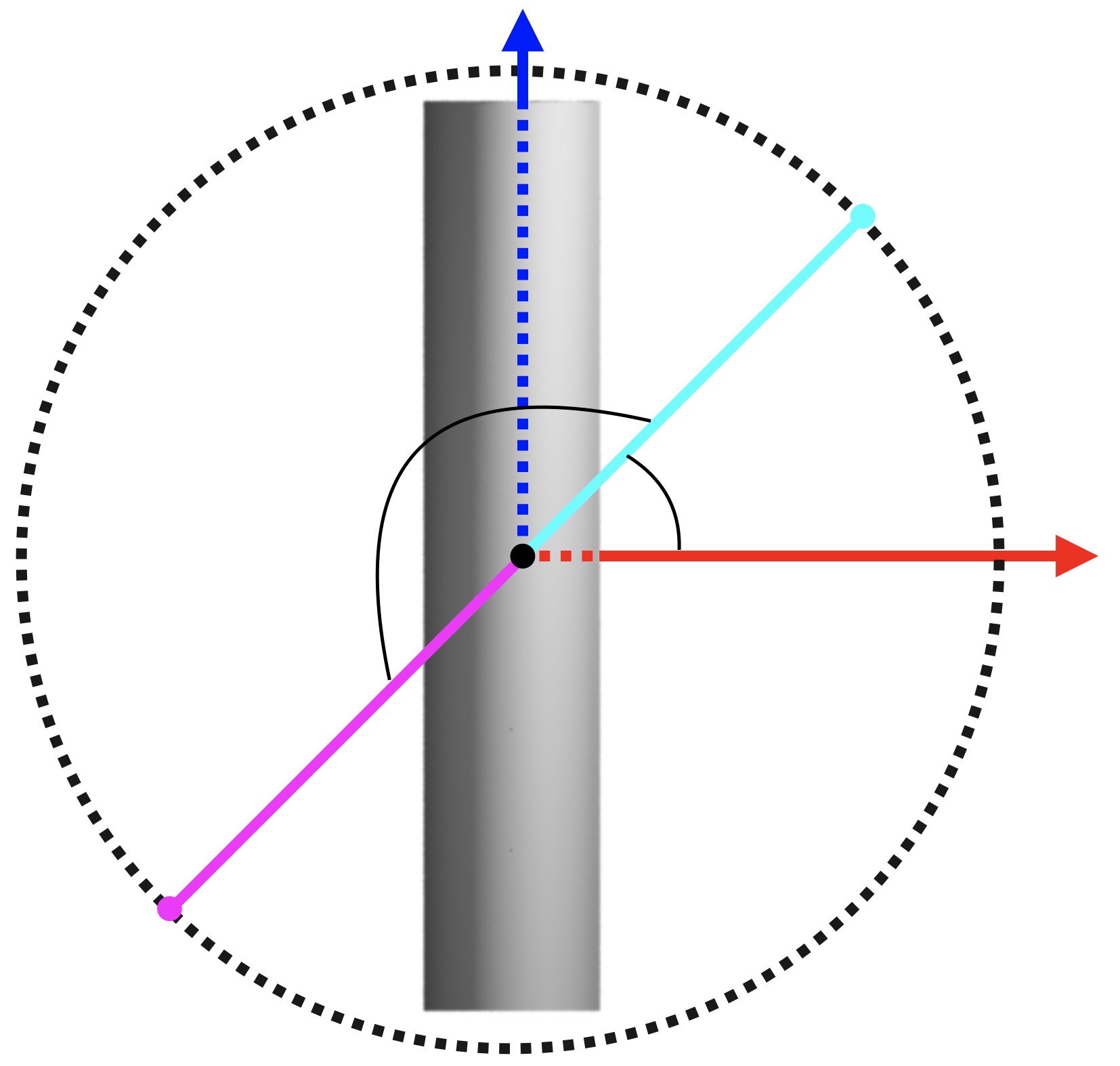}
			\put(-12, 60){\Large $ \bX $}
			\put(-105, 155){\Large $ \bZ $}
			\put(-64, 84){\Large $ \theta $}
			\put(-33, 128){\LARGE $ v $}
			\put(-160, 10){\LARGE $ v_\pi $}
			\put(-124, 88){\huge $ \pi $} &
			\includegraphics[trim={0 0 0 0}, clip = true,width=5.8cm]{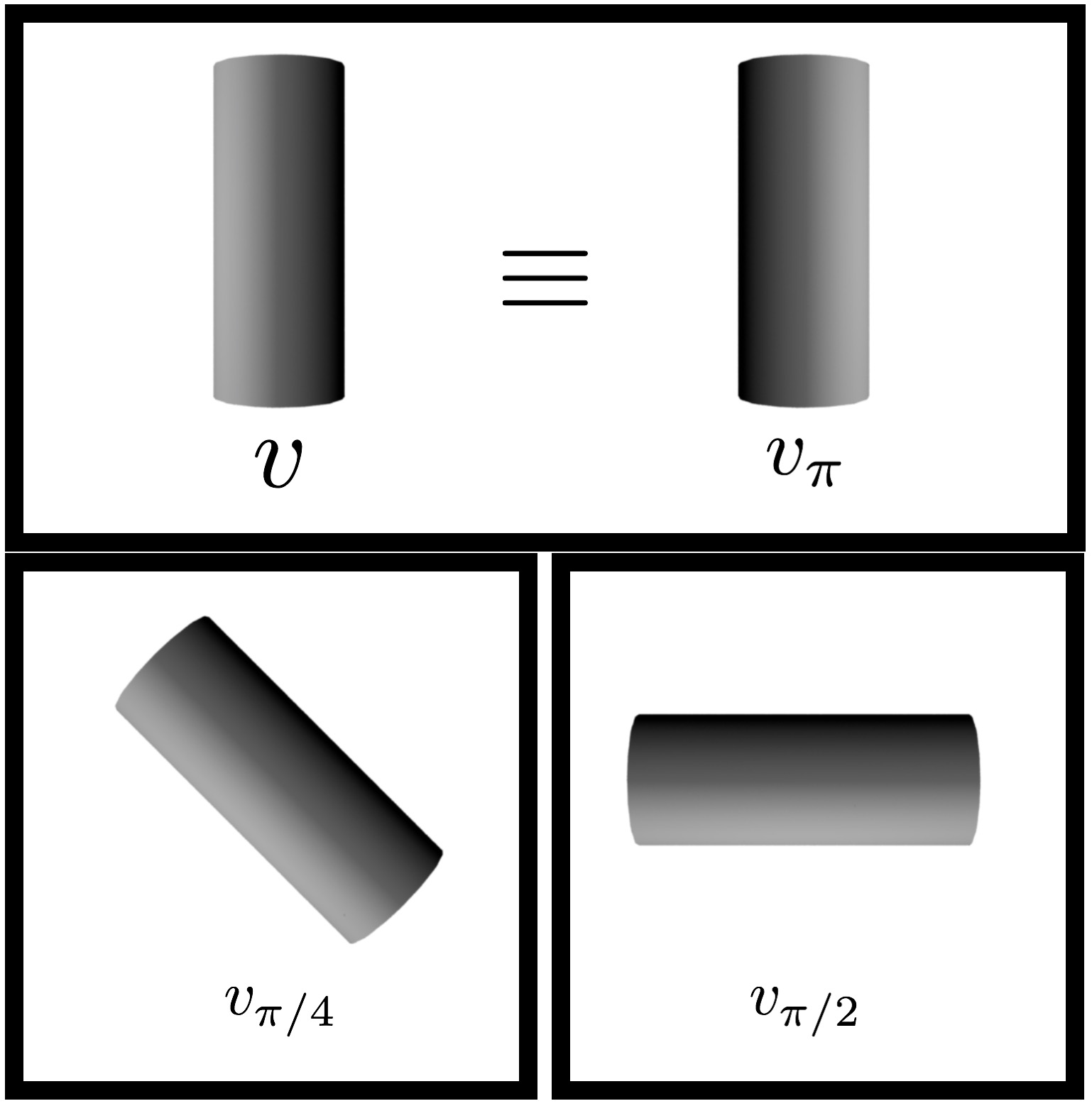}  \\
			{\LARGE (a) 3D Model} & {\LARGE (b) XZ plane}  
		\end{tabular}
	}
	\vspace{-0.5mm}
	\caption{Order of Rotational Symmetry. An object has $n$th order of rotational symmetry wrt an axis when its 3D shape is equivalent to its rotated versions $ \left( \tfrac{2 \pi i}{n} \right), \forall i \in \{0, \ldots, n - 1\} $ across this axis. For example, the cylinder in (a) has rotational symmetry wrt axes X, Y and Z. In (b), we show its second order of symmetry wrt Y, as the shape repeats every $\pi$.}
	\label{fig:rot_sym}
	\vspace{-2mm}
\end{figure}
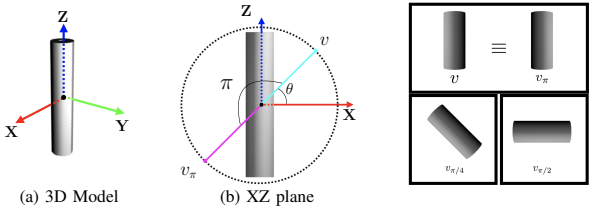

The min value of $ \mathcal{O}(\theta) $ is 1, and holds for objects non-symmetric around axis $\theta$. The max value is $ \infty $, which indicates that the 3D shape is equivalent when rotated by any angle around its axis of symmetry. This symmetry is also referred to as the revolution symmetry~\cite{bregier2017symmetry}. In ~\figref{fig:rot_sym}, we can see an example of our rotational order definition. For a 3D model shown in~\figref{fig:rot_sym} (a), the rotational order about the $ \bY $ axis is 2, \ie, $ \mathcal{O} (\bY) = 2 $. Thus for any viewpoint $ v $ (cyan) in~\figref{fig:rot_sym} (b), if we rotate it by $ \pi $ about the Y-axis to form, $ v_\pi = \mathbf{R}_{\bY} (\pi) v $, the 3D shapes will be equivalent (\figref{fig:rot_sym} (right)). The 3D shape in any other viewpoint (such as, $ v_{\pi/4} $ or $ v_{\pi/2} $) will not be equivalent to that of $ \bv $. Similarly, we have $ \mathcal{O} (\bZ) = \infty $. 
In our paper, we only consider the values of rotational order to be one of $ \{ 1, 2, 4, \infty \} $, however, our method will not depend on this choice.

\paragraph{Equivalent Viewpoint Sets} %For the axis, $ \bY $ in~\figref{fig:rot_sym} (middle), we have $ \mathcal{O}_{x} (\theta) = 2 $. 
Let us define the set of all pairs of equivalent viewpoints as $ E_o ( \bY ) = \{ (i, j) | v_j = \mathcal{R}_\theta ( \pi ) v_i \} $, with symmetry order $o \in \{ 2, 3, \infty \} $.  %Thus, we can similarly define, $ E_o ( \theta ), \forall o \in \{ 2, 3, \infty \} $. 
Note that $ E_1 (\theta) $ is a null set (object is asymmetric). In our case, we have $ E_2 (\theta)\subset E_4(\theta) \subset E_\infty (\theta) $ and $ E_3(\theta) \subset E_\infty (\theta) $.

%\subsection{Geometrical Constraints on Orders of Rotational Symmetry about Multiple Axes}
%\vspace{-3mm}
\paragraph{Geometric Constraints} We note that the orders of symmetries across multiple axes are not independent. We derive the following claim~\footnote{We give the proof in suppl. material: \url{http://www.cs.utoronto.ca/~ecorona/symmetry_pose_estimation/supplementary.pdf}.}:
%We use the following claim to constrain the output space of the predicted rotational symmetry in our approach. 
\vspace{-1mm}
\begin{claim}
	\label{clm:order_constraint}
	If an object is not a sphere, then the following conditions must hold:\\[-4.5mm]
	\begin{enumerate}[noitemsep,label=(\alph*)]
		\item The object can have up to one axis with infinite order rotational symmetry.
		\item If an axis $ \theta $ has infinite order rotational symmetry, then the order of symmetry of any axis not orthogonal to $ \theta $ can only be one.%, \ie, other non-orthogonal axes can't have symmetric viewpoints. ({\color{red} re-phrase})
		\item If an axis $ \theta $ has infinite order rotational symmetry, then the order of symmetry of any axis orthogonal to $ \theta $ can be a maximum of two.
	\end{enumerate}
	\vspace{-1mm}
\end{claim}
Since in our experiments, none of the objects is a perfect sphere, we will use these constraints in Subsec.~\ref{sec:nw_arch} in order to improve the accuracy of our symmetry predicting network. 

%Since none of the objects in our dataset is a sphere, we use this to constrain the output space of our order predictions, \ie, order prediction of multiple axes is not an independent problem. In the next section we show how we incorporate these constraints for order prediction. The proof of this claim is given in the supplementary material.

\section{Pose Estimation}
\label{sec:method}

% Overview paragraph (

We assume we are given an image crop containing the object which lies on a horizontal surface. Our goal is to predict the object's coarse pose given its 3D CAD model. Thus, we focus on recovering only the three rotation parameters.
%We assume that the objects have been detected and in each input image, the center of the object is at the origin. Thus we are interested in recovering only the three rotation parameters of pose. We also assume that we are given the CAD model corresponding to the object in the input image. % Our approach matches the input color image with the depth renderings of the CAD model from a discrete set of viewpoints along a unit sphere. The rendering with the closest match is our coarse pose estimate. Moreover, we use the same matching network to predict orders of rotational symmetries about multiple axes. %, which is used to improve our coarse pose estimate. % and this is used to improve the training of the matching network for pose estimation.

We first describe our discretization of the viewing sphere of the 3D model in order to generate synthetic viewpoints for matching.  We then introduce the joint neural architecture for pose and symmetry estimation in~\secref{sec:nw_arch}. We introduce a loss function that takes symmetry into account in~\secref{sec:loss_func}. Finally,~\secref{sec:training_details} provides our training algorithm.

\begin{figure}[t!]
\vspace{-0mm}
\begin{minipage}{0.25\linewidth}
	\includegraphics[trim={7cm 4cm 9cm 4cm}, clip = true, width=1\linewidth]{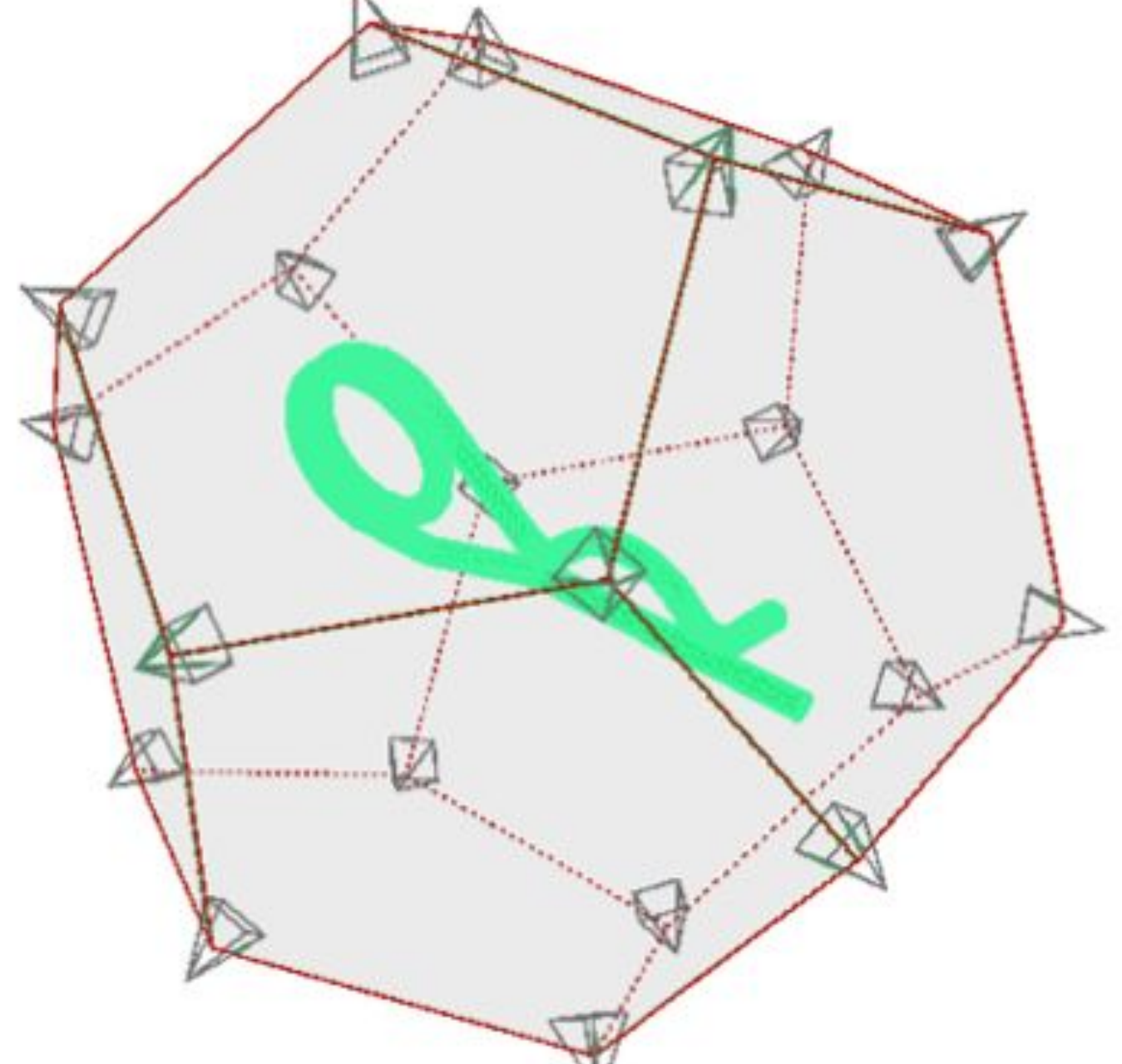} 
	\end{minipage}
	\hspace{5mm}
	\begin{minipage}{0.64\linewidth}
	\vspace{-2mm}
	\caption[Viewpoints in the sphere]{We place four cameras in each of the 20 vertices of a dodecahedron, yielding a total of 80 cameras, and place the CAD model in the origin. We render the CAD model in each viewpoint and use these for matching. We also exploit a finer discretization into $168$ views.} 
	\label{fig:physics_sim}
		\end{minipage}
		\vspace{-3mm}
\end{figure}

%\vspace{-3mm}
\paragraph*{Discretization of the viewing sphere}
Using the regular structure of a dodecahedron, we divide the surface of the viewing sphere into 20 equidistant points. This division corresponds to dividing the pitch and yaw angles. At each vertex, we have $ 4 $ roll angles, obtaining a total of $ 80 $ viewpoints. This is shown in~\figref{fig:physics_sim}. We also experiment with a finer discretization, where the triangular faces of an icosahedron are sub-divided into $ 4 $ triangles, giving an additional vertex for each edge. This results in a total of $ 42 $ vertices and $ 168 $ viewpoints.
%{\color{red} talk about the angle between the neighboring viewpoints. what it is and what it means?}

% The setup is depicted in Figure \ref{fig:physics_sim} when obtaining views from a CAD model of a screw.

\begin{figure*}[t!]
\vspace{-0mm}
	\begin{minipage}{0.55\linewidth}
	\includegraphics[trim={6.5cm 6cm 4cm 6.5cm}, clip = true, width=\linewidth]{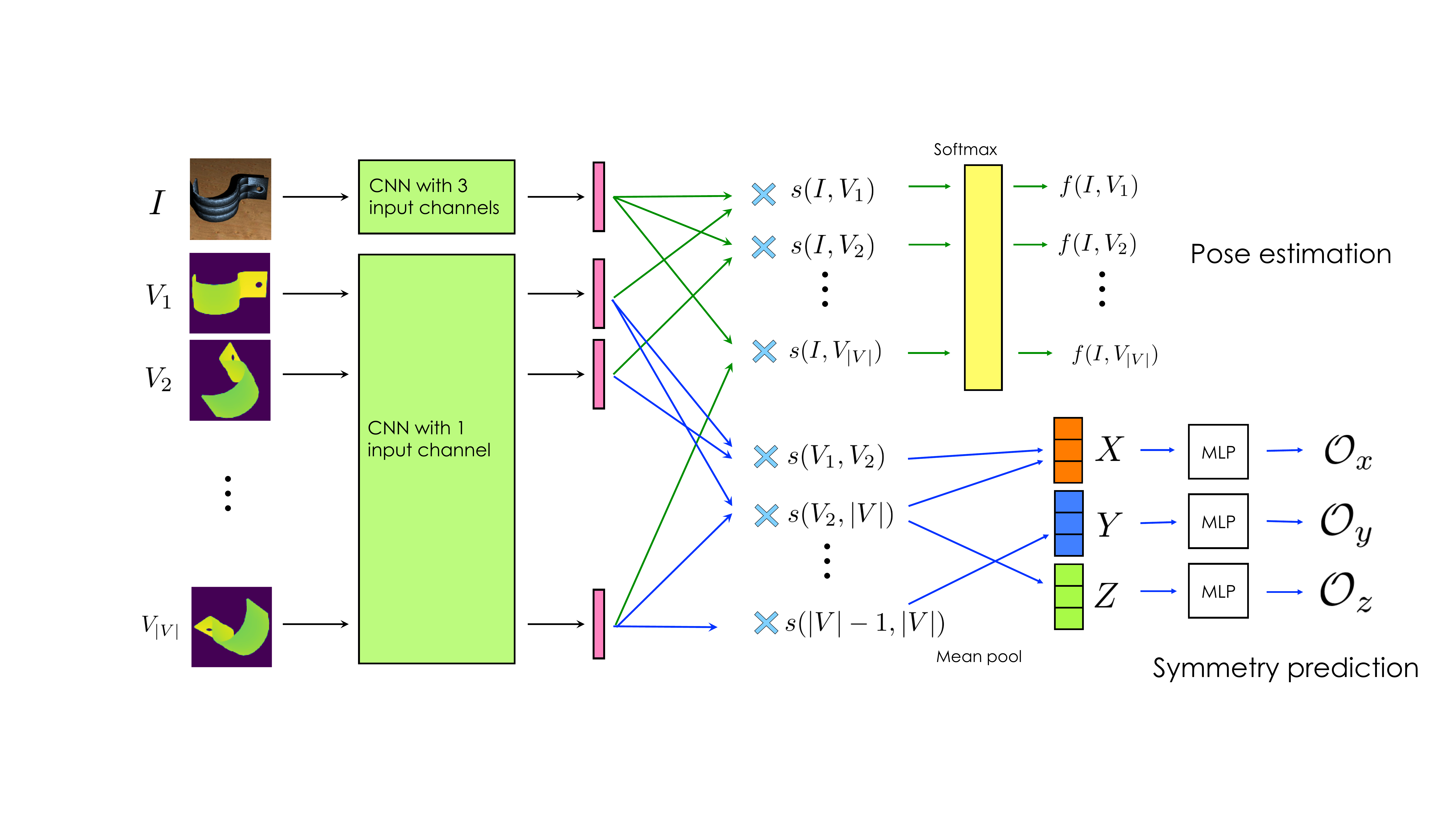} 
	\end{minipage}\hspace{7mm}
	\centering
	\begin{minipage}{0.34\linewidth}
		\vspace{-3mm}
	\caption[Method overview]{{\bf Overview of our model.} We use a convolutional neural network to embed the RGB image of an object in the scene and the rendered depth maps of the CAD model into a common embedding space. We then define two branches, one performing pose estimation by comparing the image embedding with the rendered depth embeddings, and another branch which performs classification of the order of symmetry of the CAD model. We show how to train this network with very few symmetry-labeled CAD models, by additionally exploiting a large collection of unlabeled CAD models crawled from the web. }
	\label{fig:squeme}
	\end{minipage}
\vspace{-5mm}	
	%Once the object image is cropped from the whole scene, it is compared to a set of depth views of the object with different rotations. These are obtained offline to discretize the viewing sphere so that each possible rotation in a scene is estimated by the nearest view. The RGB image of the object and the depth views are mapped by a 3-Channel-input and 1-Channel-input CNNs to a new common space where they can be compared through the dot product operation. This new space encodes rotation, which provides the similarity score a geometrical meaning. Ultimately, this is a classification task where the highest score should belong to the nearest discretized rotation. We then transform the scores to probabilities using a Softmax function, to train the network with a structured ranking loss. }

\end{figure*}

\subsection{Network Architecture}
\label{sec:nw_arch}

The input to our neural network is an RGB image $ \bx $, and depth maps corresponding to the renderings of the CAD model, one for each viewpoint $ \bv_i $. With a slight abuse of notation we refer to a depth map corresponding to the $i$-th viewpoint as $ \bv_i $. Our network embeds both, the RGB image and each depth map into feature vectors, $ g_\text{rgb}(\bx)$ and $ g_\text{depth} ( \bv_i )$, respectively, by sharing the network parameters across different viewpoints. We then form two branches, one to predict object pose, and another to predict the CAD model's orders of symmetry. The full architecture is shown in~ \figref{fig:squeme}. We discuss both branches next. 

%\vspace{-3mm}
\paragraph{Pose Estimation} Let $C(k, n, s)$ denote a convolutional layer with kernel size $k \times k$, $ n $ filters and a stride of $ s $. Let $ P(k, s)$ denote a max pooling layer of kernel size $ k \times k$ with a stride $ s $. The network $ g_\text{rgb} $ has the following architecture: $ C(8, 32, 2) - ReLU - P(2,1) - C(4, 64, 1) - ReLU - P(2,1) -  C(3,64, 1) - ReLU - P(2, 1) - FC(124) - ReLU - FC(64) - L2\_Norm $.   With slight abuse of notation, we denote our image embedding with $g_\text{rgb} ( \bx ) $, which we take to be the final layer of this network, i.e., a $ 64 $-dimensional unit vector. We define a similar network for $ g_\text{depth} $, where, however, the input has a single channel.

%\vspace{-3mm}
%\paragraph{Matching Function.} 
We follow the typical approach~\cite{luo2016efficient,gordo2017end} in computing the  similarity score $ f ( \bx, \bv_i ) $ in the joint semantic space:
\begin{align}
\label{eq:matching}
	%f ( \bx, \bv_i )  & =  \dfrac{\text{exp} \left( s ( \bx, \bv_i ) \right) }{ \sum_{j = 1}^{N} \text{exp} \left( s ( \bx, \bv_j ) \right)}  \\
	s ( \bx, \bv_i ) & =  g_\text{rgb} ( \bx  )^\top g_\text{depth} ( \bv_i )\\
	f ( \bx, \bv_i )  & =  \mathrm{softmax}_{i}\ s ( \bx, \bv_i )
\end{align}

To compute the object's pose, we thus take the viewpoint $\bv^*$ with the highest probability $v^*=\mathrm{argmax}_{i}f ( \bx, \bv_i)$.

%\vspace{-3mm}
\paragraph{Rotational Symmetry Classification}  Obtaining symmetry labels for CAD models is time-consuming to collect. The annotator needs to open the model in a 3D viewer, and carefully inspect all three major axes in order to decide on the orders of symmetry for each. In our work, we manually labeled a very small subset of 45 CAD models, which we make use of here. In the next section, we show how to exploit unlabeled large-scale CAD collections for our task. 

Note that symmetry classification is performed on the renderings of the CAD viewpoints, thus effectively estimating the order of symmetry of the 3D object. We add an additional branch on top of the depth features to perform classification of order of symmetry for all three orthogonal axes (each into $4$ symmetry classes).  In particular, we define a scoring function for predicting symmetry as follows:
\begin{align}
\label{eq:score}
	&S \left( \mathcal{O} ( \bX ),\mathcal{O} (\bY ),  \mathcal{O} (\bZ ) \right) \nonumber\\
	& = \sum_{\theta} S_\text{unary} \left( \mathcal{O} ( \theta ) \right) + \sum_{\theta_1 \neq \theta_2} S_\text{pair} \left( \mathcal{O} ( \theta_1 ) , \mathcal{O} ( \theta_2 ) \right)  \\
	& \qquad + S_\text{triplet} \left( \mathcal{O} ( \bX ), \mathcal{O} ( \bY ), \mathcal{O} ( \bZ ) \right)	
\end{align}
Note that our scoring function jointly reasons about rotational symmetry across the three axes. Here, the pairwise and triplet terms refer to the geometrically impossible order configurations based on~\clmref{clm:order_constraint}. We now define how we compute the unary term. 
%We now define each of the above terms.

\emph{Unary Scoring Term.} We first compute the similarity scores between pairs of (rendered) viewpoints. We then form simple features on top of these scores that take into account the geometry of the symmetry prediction problem. Finally, we use a simple  Multilayer Perceptron (MLP) on top of these features to predict the order of symmetry.

The similarity between pairs of rendered viewpoints measures whether two viewpoints are a match or not:
%We use the above network to compute the similarity scores between pairs of renderings. These scores are used to vote for the corresponding order class. Towards this goal, we first convert the similarity score, $ s ( \bv_i, \bv_j ) $ between all pairs of renderings, $ (\bv_i, \bv_j) $ into a match/non-match score, $ p_{i, j} $,
\begin{equation}
p_{i, j} = \sigma( w\cdot s ( \bv_i, \bv_j ) + b ) ,
\end{equation}
where $\sigma$, $w$ and $b$ are the activation function, weight and bias, of the model respectively.
One could use a MLP on top of $\bp$  to predict the order of symmetries as a classification task based on the similarities. 
However, due to the limited amount of training data for this branch, such an approach heavily overfits. Thus, we aim to exploit the geometric nature of our prediction task. In particular, we know that for symmetries of order $2$, every pair of opposite viewpoints (cyan and magenta in Fig.~\ref{fig:rot_sym}) corresponds to a pair of equivalent views. We have similar constraints for other orders of symmetry. 

We thus form a few simple features as follows. For, $ \theta \in \{ \bX, \bY, \bZ \}$, and $o \in \{2, 4, \infty\} $, we perform average pooling of $ p_{i,j} $ values for $ (i,j) \in  E_o (\theta) $. Intuitively, if the object has symmetry of order $o$, its corresponding pooled score should be high. However, since eg $E_2\subset E_{\infty}$, scores for higher orders will always be higher. 
We thus create a simple descriptor for each axis $ \theta $. More precisely, our descriptor $ m_o (\theta) $ is computed as follows:
\begin{align*}
	m_2 (\theta) & = \dfrac{1}{ \left| E_2 (\theta) \right| } \sum_{(i, j) \in  E_2 (\theta)} p_{i, j}  \\
	m_4 (\theta) & = \dfrac{1}{ \left| E_4 (\theta) - E_2 (\theta) \right| } \sum_{(i, j) \in  E_4 (\theta) - E_2 (\theta)} p_{i, j}  \\
	m_\infty (\theta) & = \dfrac{1}{ \left| E_\infty (\theta) - E_4 (\theta) \right| } \sum_{(i, j) \in  E_\infty (\theta) - E_4 (\theta)} p_{i, j}  
\end{align*}
Since $ E_2 (\theta) \subset E_4 (\theta) \subset E_\infty (\theta) $, we take the set differences. We then use a single layer MLP with ReLU non-linearity to get the unary scores, $ S_\text{unary} \left( \mathcal{O} ( \theta ) \right) $. These parameters are shared across all three axes.

%\noindent{\it Pairwise and Triplet Terms.} 
Since we have four order classes per axis, we have a total of 64 combinations. Taking only the possible configurations into account, the total number of combinations reduces to 21. We simply enumerate these  21 configurations and choose the highest scoring one as our symmetry order prediction.

\subsection{Loss Function}
\label{sec:loss_func}

Given $ B $ training pairs, $ X=\{ \bx^{(i)}, \bv^{(i)} \}_{i = 1, \ldots, B} $ in a batch, we define the loss function as the sum of the pose loss and rotational order classification loss:
\[
L(X, \bw) = \sum_{i = 1}^B L_{\text{pose}}^{(i)}(X, \bw) + \lambda L_{\text{order}}^{(i)}(X, \bw)
\]
We describe both loss functions next.

% We obtain a similarity score $s(\cdot,\cdot)$ between the pose of the object in the image and each view, as the cosine similarity between its representations. In this subsection we describe how the network was trained. 
%However, since we approach it as a classification problem of $\left\vert{V}\right\vert$ classes, we obtain better results imposing a Softmax function. Then, we obtain the probability $f(I_i, V_{i,j})$ for each view $V_{i,j}$ to be nearest to the object in the scene $I_i$ 

%\vspace{-3mm}
\paragraph{Pose Loss} We use the structured hinge loss: % function to impose a margin on the probabilities of the negative pairs of all the viewpoints, \wrt the nearest discretized ground truth viewpoint.
\begin{align*}
L_\text{pose}^{(i)} & = \sum_{j = 1}^{N} \max \left( 0, m_{j}^{(i)} + f ( \bx^{(i)}, \bv_{j}^{(i)} )- f ( \bx^{(i)}, \bar{\bv}^{(i)} ) \right)
\end{align*} 
where $\bv_{j}^{(i)}$ corresponds to the negative viewpoints, and $\bar{\bv}^{(i)}$ denotes the closest (discrete) viewpoint wrt to ${\bv}^{(i)}$ in our discretization of the sphere. 
%where, $  \bar{\bv}^{(i)} $ is the discretized version of $ \bv^{(i)} $. 
In order to provide the network with a knowledge of the rotational space, we impose a rotational similarity function as the margin $ m_{j}^{(i)} $. Intuitively, we want to impose a higher penalty for the mistakes in poses that are far away than those close together:
\[
 m_{j}^{(i)} = d_\text{rot} ( \bv^{(i)}, \bv_{j}^{(i)} )- d_\text{rot} ( \bv^{(i)}, \bar{\bv}^{(i)} )
\]
where $ d_\text{rot} $ is the spherical distance between the two viewpoints in the quaternion space. Other representations of viewpoints are Euler angles,  rotation matrices in the $ SO(3) $ space and quaternions~\cite{gimbal_lock}. While the Euler angles suffer from the gimbal lock~\cite{gimbal_lock} problem, measuring distances between two matrices in the $ SO(3) $ space is not trivial. The quaternion space is continuous and smooth, which makes it easy to compute the distances between two viewpoints. The quaternion representation, $ q_\bv $ of a viewpoint, $ \bv $ is a four-dimensional unit vector. Thus each 3D viewpoint is mapped to two points in the quaternion hypersphere, one on each hemisphere.  We measure the difference between rotations as the angle between the vectors defined by each pair of points, which is defined by their dot product. Since the quaternion hypersphere is unit normalized, this is equivalent to the spherical distance between the points. 

%However, computing $ d_\text{rot} \left( \bv_a, \bv_b \right) $ as $ cos^{-1} \left( q_{\bv_a}^{\top} q_{\bv_b} \right) $ is not enough. When the dot product is negative, the angle obtained is the angle between two vectors on opposite sides of the hypersphere, which is not the appropriate distance. 
To restrict the spherical distance to be always positive, we use the distance function defined as:
\[
d_\text{rot} \left( \bv_a, \bv_b \right) = \frac{1}{2} cos^{-1} \left( ( 2( q_{\bv_a}^{\top} q_{\bv_b} \right)^2 - 1) , 
\]
When the objects have rotational symmetries, multiple viewpoints could be considered ground truth. In this case, $ \bv^{(i)} $ corresponds to the set of equivalent ground-truth viewpoints. Thus the margin $ m_{j}^{(i)} $ takes the form of:
\[
m_{j}^{sym, (i)} = \minarg{\bv \in \bv^{(i)}} d_\text{rot} ( \bv, \bv_{j}^{(i)} )- d_\text{rot} ( \bv, \bar{\bv} )
\]
The modified pose loss which takes symmetry into account will be referred to as $ L_\text{match}^{sym, (i)} $.

\paragraph{Rotational Order Classification Loss} Considering the axis as $ \bX $, $ \bY $ and $ \bZ $, we use a weighted cross entropy:
\begin{equation}
L_{\text{order}}^{(i)} = - \sum_{\theta \in \{ \bX, \bY, \bZ  \}} \sum_{o \in \{1, 2, 4, \infty \} } \alpha_o\cdot y_{i,o,\theta}\cdot \log(p_\theta^i (o) )
\end{equation}
where $ \by_i(\cdot,\cdot,\theta)$ is the one-hot encoding of $i$-th ground-truth symmetry order around axis $\theta$, and $ \bp_\theta^i $ is the predicted probability for symmetry around axis $ \theta $. Here, $ \alpha_o $ is the inverse frequency for order class $ o $, and is used to balance the labels across the training set.

\begin{figure*}[t!]
\vspace{1mm}
\centering
\begin{tabular}{ll}
\hspace{-2mm}\begin{tabular}[c]{@{}l@{}}
\includegraphics[trim={0 10mm 0 5mm}, clip = true, width=4.07cm]{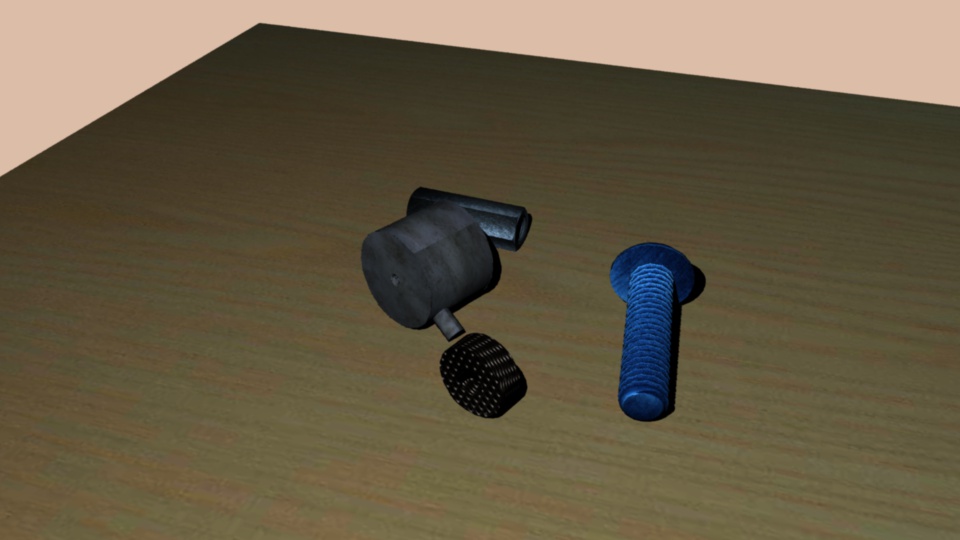}
\includegraphics[trim={0 15mm 0 0}, clip = true,width=4.07cm]{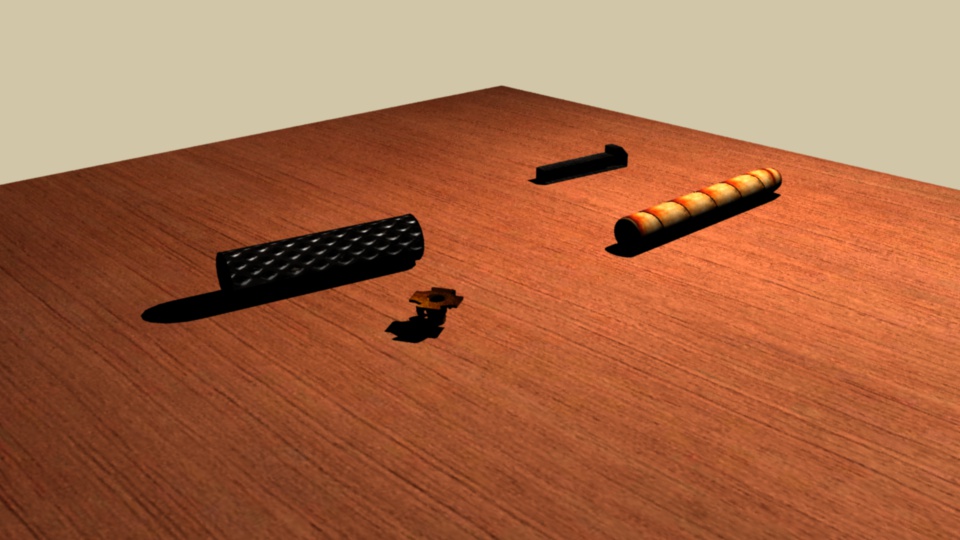}
\\
\includegraphics[trim={0 15mm 0 0}, clip = true,width=4.07cm]{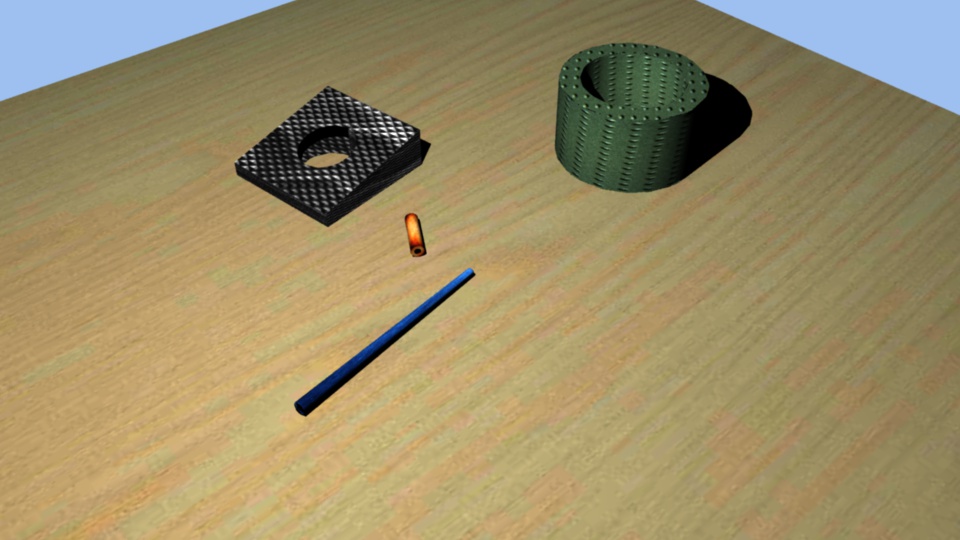}
\includegraphics[trim={0 15mm 0 0}, clip = true,width=4.07cm]{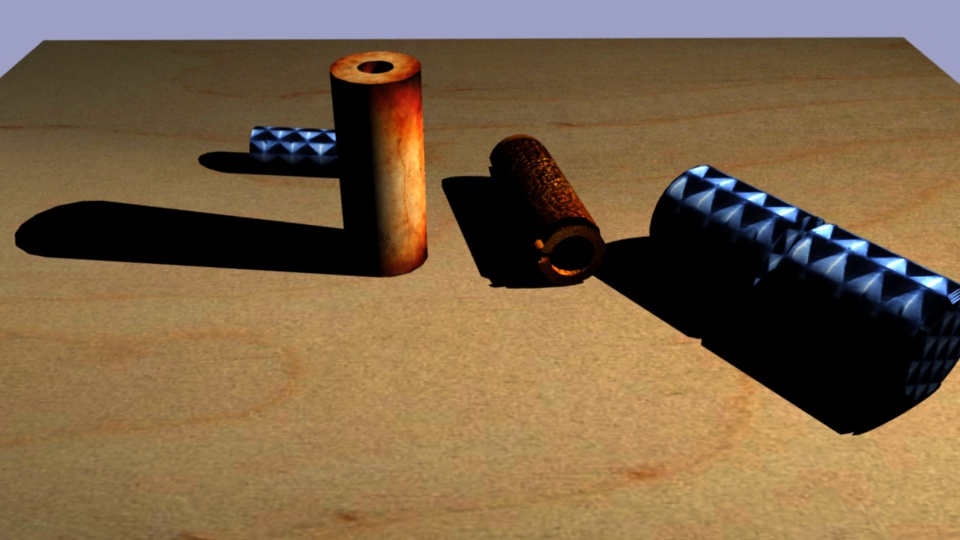}
\end{tabular} & \hspace{1mm}\begin{tabular}[c]{@{}l@{}}   		
		\includegraphics[trim={0 0 0 0}, clip = true,width=1.38cm]{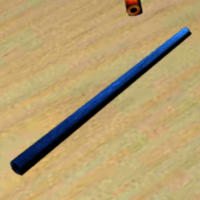}
		\includegraphics[trim={0 0 0 0}, clip = true,width=1.38cm]{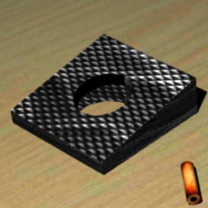}
		\includegraphics[trim={0 0 0 0}, clip = true,width=1.38cm]{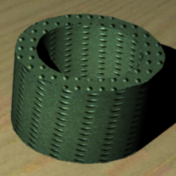}
		\includegraphics[trim={0 0 0 0}, clip = true,width=1.38cm]{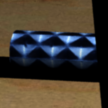}
		\includegraphics[trim={0 0 0 0}, clip = true,width=1.38cm]{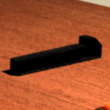}
		\includegraphics[trim={0 0 0 0}, clip = true,width=1.38cm]{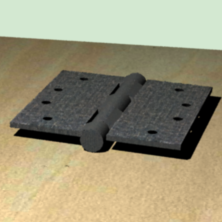}
		\\
		\includegraphics[trim={0 0 0 0}, clip = true,width=1.38cm]{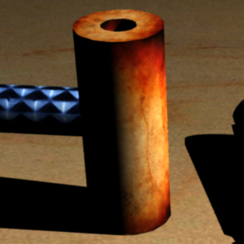}
		\includegraphics[trim={0 0 0 0}, clip = true,width=1.38cm]{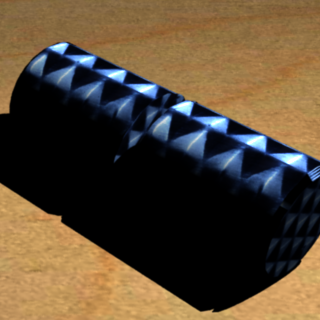}
		\includegraphics[trim={0 0 0 0}, clip = true,width=1.38cm]{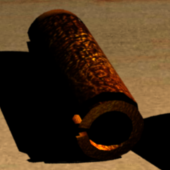}
		\includegraphics[trim={0 0 0 0}, clip = true,width=1.38cm]{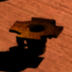}
		\includegraphics[trim={0 0 0 0}, clip = true,width=1.38cm]{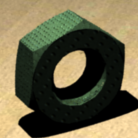}
		\includegraphics[trim={0 0 0 0}, clip = true,width=1.38cm]{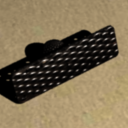}
		\\
		\includegraphics[trim={0 0 0 0}, clip = true,width=1.38cm]{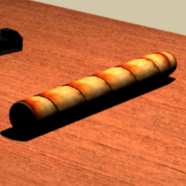}
		\includegraphics[trim={0 0 0 0}, clip = true,width=1.38cm]{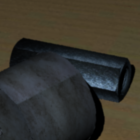}
		\includegraphics[trim={0 0 0 0}, clip = true,width=1.38cm]{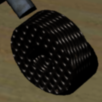}
		\includegraphics[trim={0 0 0 0}, clip = true,width=1.38cm]{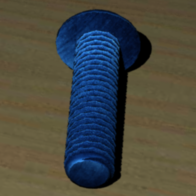}
		\includegraphics[trim={0 0 0 0}, clip = true,width=1.38cm]{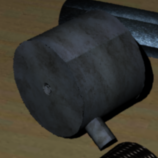}
		\includegraphics[trim={0 0 0 0}, clip = true,width=1.38cm]{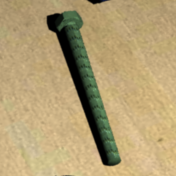}
		 \end{tabular} \\                       
\end{tabular}
\vspace{-3mm}
\caption[Rendered images of table-top scenarios]{ {\bf Left:} Rendered synthetic scenes, {\bf Right:} Object crops from the scene. We use these to train our model.}
\label{fig:synthetic_data}
\vspace{-2mm}
\end{figure*}

\subsection{Training Details}
\label{sec:training_details}

Here, we aim to exploit both real data as well as a large collection of CAD models in order to train our model. We assume we have a small subset of CAD models labeled with symmetry, while the remaining ones are unlabeled. For the unlabeled CAD models, we additionally render a dataset for pose estimation, referred to as the \emph{synthetic} dataset. The details of the dataset are given in Sec.~\ref{sec:datasets}. 
In particular, we use the following iterative training procedure:
\begin{enumerate}[noitemsep]
	\item Train on the synthetic dataset with the $ L_{\text{pose}} $ loss % + \lambda L_{\text{order}}$
	\item \label{item:sym_pred} Fine-tune on the labeled synthetic and real examples with the $ \lambda L_{\text{order}} $ loss function
	\item \label{item:sym_train} Infer symmetries of unlabeled CAD models via Eq.~\eqref{eq:score}
	%$S \left( \mathcal{O} ( \bX ), \mathcal{O} (\bY ),  \mathcal{O} (\bZ ) \right) $
	\item \label{item:order_pred} Fine-tune on the synthetic dataset with the $  L_{\text{pose}}^{sym} $ loss
	\item Fine-tune on the real data with the $  L_{\text{pose}}^{sym} $ loss function
\end{enumerate}
Note that in step~\ref{item:order_pred}, we use the predictions from the network in step~\ref{item:sym_train}  as our ground-truth labels.

%\vspace{-3mm}
\paragraph{Implementation details} The input depth map is normalized across the image to lie in the range, $ [0, 1] $, with the missing depth values being $0$. %The network weights are initialized using the Xavier initialization~\cite{xavier_initialization}. However for step~\ref{item:sym_train}, we carefully initialize the MLP weights as follows. All the MLP parameters are initialized to $ 0 $, except for the weights which provide positive scores for order prediction. For example, for the score, $ S_\text{unary} \left( \mathcal{O} ( \theta ) = 4 \right) $, the corresponding weights from $ m_2 (\theta), m_4 (\theta), m_\infty (\theta) $ were set to $ 10, 10, 0 $ respectively. 
The learning rate for the CNN were set to two orders of magnitude less than the weights for the MLP ($10^{-2}$ and $10^{-4}$).  We use the Adam optimizer. Training was stopped when there was no improvement in the validation performance for 50 iterations.

\iffalse
\paragraph{Data Augmentation via Deformations in the 3D model}

In the symmetry prediction step, the number of labels is significantly low. To handle this issue, we introduced small deformations to the 3D model to generate hard negative examples. A vertex of the mesh was randomly selected and moved by a small amount along the direction of its outward normal. Since the projected view of the deformed model is different from the rendering of the original model, this can be used to train a binary match vs not-match problem. In the results section, we show the usefulness of this approach.

{\color{red} Also we could have images of deformations, if there's space in the paper. }
\fi

\vspace{-1mm}
\section{Datasets}
\label{sec:datasets}

In an industrial setting, the objects can be arbitrary complex and can exhibit rotational symmetries (examples are shown in Fig.~\ref{fig:epson1034_result}). Current datasets with 3D models such as~\cite{kitti_dataset,beyond_pascal} have objects like cars, beds, \etc, which have much simpler shapes with few symmetries. Datasets such as~\cite{bregier2017symmetry,tless} contain industrial objects. However, these objects have only one axis with rotational symmetry. Here, we consider a more realistic scenario of object that can have symmetries for multiple axes. 
%The objects that are commonly used in industrial settings can have multiple axes of rotational symmetry.

We introduce two datasets, one containing real images of objects with accompanying CAD models, and a large-scale dataset of industrial CAD models which we crawl from the web. We describe both of these datasets next. 

\iffalse
Annotating rotational symmetry is challenging and time consuming in 3D models, specially the ones with complex shapes. It requires rotating each of the models from multiple viewpoints, which makes the process extremely time consuming. Thus hand-labeling rotational symmetry for arbitrary complex 3D models is not a scalable solution. The success of deep neural networks comes at a price of requiring large quantities of data.~\cite{renderforcnn} has shown that using synthetic datasets such as~\cite{shapenet} can enable a better pose estimation performance in natural image datasets such as~\cite{beyond_pascal}. However, unlike~\cite{renderforcnn, bb8}, our goal is that our approach should generalize to unseen objects having rotational symmetry along multiple axes.
% Learning robust representations from small scale datasets with arbitrary complexities for generalization to unseen objects is even more challenging.
We crawl a hardware website to obtain such objects that can be found in table top scenario. Then, we use a physics engine (Maya~\cite{maya}) to simulate scenes with multiple CAD models of the objects.
\fi

\subsection{Real Images} 
\label{sec:real_dataset} 

We obtain a dataset containing images of real 3D objects in a table-top scenario from the company Epson. %\footnote{https://epson.ca/}. 
This dataset has 27,458 images containing different viewpoints of 17 different types of objects. Each CAD model is labeled with the order of symmetry for each of the axes, while each image is labeled  with accurate 3D pose. %Examples of the dataset are shown in~\figref{fig:identifications}(b).

%If we split the data randomly into train, validation and test sets, the networks can overfit due to the images being similar. 
We propose two different splits: (a) \textit{timestamp}-based: divide images of each of the objects into training and testing, while making sure that the images were taken at times far apart (thus having varying appearance), (b) \textit{object}-based: the dataset is split such that the training, val and testing objects are disjunct. %In timestamp based splits, due to differenet lighting conditions, the images can tend to look different. In the object-based splitting option, we can see how well our network can generalize to unseen data. 
We divide 17 objects into 10 train, 3 val, and 4 test objects. Dataset statistics is reported in~\tabref{tab:data_stats}.

\begin{table}[t!]
\begin{scriptsize}
\vspace{0mm}
	\begin{center}
	{\small
	\setlength{\tabcolsep}{6.8pt}
		\begin{tabular}{|c||c||c|c|c|}
			\hline
			Data Type & Split Type & Train & Validation & Test \\
			\hhline{|=#=|=|=|=|}
			\multirow{3}{*}{Real} & Timestamp & 21,966 & 746 &  2,746  \\
			\cline{2-5}
			& 	\multirow{2}{*}{Object}  & 16,265 & 3,571 & 7,622   \\
			& & (10) & (3) & (4) \\
			\hline
			\multirow{2}{*}{Synthetic} &  \multirow{2}{*}{Object} & 52,763 &  5,863 & \multirow{2}{*}{-} \\
			& & (5,987) & (673) &  \\
			\hline
		\end{tabular}
		}
		\vspace{0.1mm}
		\begin{spacing}{0.8}
	\caption{Dataset Statistics. Numbers refer to images, while number in brackets correspond to CAD models.}
	\end{spacing}
	\label{tab:data_stats}
	\end{center}
	\vspace{-8mm}
	\end{scriptsize}
\end{table}

\subsection{Synthetic Dataset} 
\label{sec:synth_data}

To augment our dataset, we exploited 6,660 CAD models of very different objects from a hardware company~\footnote{\url{https://www.mcmaster.com/}}. %Since the contents of the web are dynamically loaded we used Chrome browser API for python instead of the default HTML crawling libraries. 
This varied set contains very simple 3D shapes such as tubes or nails to very complex forms like hydraulic bombs. %\figref{fig:identifications} shows some of the crawled 3D models. 
We labelled rotational symmetry for 28 objects from this dataset.  Dataset statistics is reported in~\tabref{tab:data_stats}. We now describe how we render the synthetic dataset for pose estimation.

% Loading and combining objects: 
%The CAD models downloaded are in IGES format, used in applications such as AutoCAD or Solidworks. Initially, we notice that these objects are designed in several parts. We combine them so that the geometrical transforms and the simulation will apply only to a joint body. We also label rotational symmetry for 28 objects from this dataset.  The statistics of the number of images can be seen in~\tabref{tab:data_stats} 

%\vspace{-4mm}
\paragraph{Scene Generation} We generate scenes of a table-top scenario using Maya, where each scene contains a subset of CAD models. In each scene, we import a set of objects, placing them on top of a squared plane with a side length of 1 meter that simulates a table. We simulate large variations in location, appearance, lighting conditions, viewpoint (as shown in~\figref{fig:synthetic_data}) as follows.

\begin{table*}[t!] 
\begin{footnotesize}
\vspace{0mm}
	\centering 
	\setlength{\tabcolsep}{11.2pt}
	%{\small
	\begin{tabular}{|C{0.68cm}||C{0.48cm}|C{1cm}|C{1cm}||c|c|c||c|c|c|} 
		\hline
		\multirow{3}{*}{\parbox{0.68cm}{\centering  Split Type }} & \multirow{3}{*}{\parbox{0.48cm}{\centering  Syn}} & \multirow{3}{*}{\parbox{1cm}{\centering  SynSym Pred}} & \multirow{3}{*}{\parbox{1cm}{\centering  RSym Sup}} & \multicolumn{3}{c||}{$ N = 80 $} & \multicolumn{3}{c|}{$ N = 168 $} \\
		\cline{5-10}
		 & &  &  & R@$ 20^\circ $ & R@$ 40^\circ $ & $ d_\text{rot, avg}^{sym} $ & R@$ 20^\circ $ & R@$ 40^\circ $ & $ d_\text{rot, avg}^{sym} $ \\
		& &  &  & (in \%) & (in \%) & (in $ {}^\circ $) & (in \%) & (in \%) & (in $ {}^\circ $) \\
		\hhline{|=#=|=|=#=|=|=#=|=|=|}
		\multirow{5}{*}{\parbox{0.68cm}{\centering Time stamp}} & & &  & 62.3 & 81.0 &   22.5    & 43.0 & 70.1 &  26.2 \\
		\cline{2-10}
		&  & & \checkmark & 70.2 & 86.3 &   19.2    & 72.4 & 88.2 &    17.5   \\
		\cline{2-10}
		& \checkmark & &  & 64.4 & 84.8 &    22.1    &  82.3 & 96.0 &    14.5   \\
		\cline{2-10}
		& \checkmark & & \checkmark  & 72.5 & 88.3 &   16.6     & {\bf 84.8} & {\bf 97.2} &  {\bf 13.3}   \\
		\cline{2-10}
		& \checkmark & \checkmark  & \checkmark  & {\bf 77.3} & {\bf 92.1} &   {\bf  12.0}   & 82.0  &  96.7  & 14.2  \\
		\hhline{|=#=|=|=#=|=|=#=|=|=|}
		\multirow{5}{*}{\parbox{0.68cm}{\centering Object }} & & &  & 23.6 & 45.4 &   37.9     &  24.5 & 42.2 &    33.6 \\
		\cline{2-10}
		&  & & \checkmark & 29.6 & 55.4 & 34.6       & 18.7 & 54.0 &    36.6    \\
		\cline{2-10}
		& \checkmark & &  & 24.9 & 58.2 &   35.5     &  31.7 & 62.3 &   33.2  \\
		\cline{2-10}
		& \checkmark & & \checkmark  & 33.6 & 67.0 &   31.4     & 31.9 & 75.2 &  29.9  \\
		\cline{2-10}
		& \checkmark & \checkmark  & \checkmark  & {\bf 35.7} & {\bf 68.7} &  {\bf 30.0}     &  {\bf 41.8} & {\bf  79.3 } &   {\bf  26.4}  \\
		\hline
	\end{tabular}
	%}
	\vspace{0.2mm}
	\caption{Pose Estimation Performance with an ablation study. } 
	\label{tab:results_pose} 
	\vspace{-3mm}
	\end{footnotesize}
\end{table*}

\paragraph{Location} The objects are set to a random translation and rotation, and scaled so that the diagonal of their 3D bounding box is smaller than $30$ centimetres. 
We then run a physical simulation that pushes the objects towards a stable equilibrium. If the system does not achieve equilibrium after a predefined amount of time we stop the simulation. In cases when objects intersect with one another, we restart the simulation to avoid these implausible situations.

%\vspace{-4mm}
\paragraph{Appearance} We collected a set of 45 high definition wooden textures for the table and 21 different materials (wood, leather and several metals) and used them for texturing the objects. The textures are randomly attached to the objects, mapping them to the whole 3D CAD model.

%\vspace{-4mm}
\paragraph{Lighting}  In each simulation, we randomly set a light point within a certain intensity range. 

%\vspace{-4mm}
\paragraph{Viewpoint} For each scene we set 15 cameras in different positions, pointing towards the origin. Their location is distributed on the surface of a sphere of radius $\mu = 75 cm$ as follows. The location along $Y$ axis follows a normal distribution $Y \sim \mathcal{N}(50,10) cm$. 
For the position over the $XZ$ plane, instead of Cartesian coordinates, we adopt Circular coordinates where the location is parametrized by $(d_{xz},\theta_{y})$. Here, $d_{xz}$ represents the distance from the origin to the point and is distributed as $ \mathcal{N}(\sqrt{\mu^2 - Y^2}, 10) cm$, where $\theta$ is the angle around the $Y$ axis.
%The angle in the $XZ$ plane follows a uniform distribution $\phi \in (-180,180]$. Finally, the distance from the camera on top of this plane is parametrized by $ \mathcal{N}(d, 10) cm$ where $d = \sqrt{\mu^2 - Y^2}$ and $\mu$ stands for the average distance from the cameras to the plane center, which we set to 75 centimetres. 
This procedure generates views of a table-top scene from varying oblique angles. We also add cameras directly above the table with  $X, Z \in (-5,5) cm$ to also include overhead views of the scene. 
%Algorithm \ref{algo-generate} defines the process we follow to obtain our dataset. 

For our task, we crop objects with respect to their bounding boxes, and use these for pose prediction. The complete scenes help us in creating context for the object crops that typically appear in real scenes.

\vspace{-1mm}
\section{Experimental Results}
\label{sec:results}
\vspace{-1mm}

We evaluate our approach on the real dataset, and ablate the use of the CAD model collection and the synthetic dataset.  %We evaluate with both discretization schemes ($ N = 80 $ and $ N = 168 $ viewpoints). 
We first describe our evaluation metrics in~\secref{sec:eval_metric} and show quantitative and qualitative results in~\secref{sec:res_quant} and~\secref{sec:res_qual}, respectively.

\begin{table}[t!]
	\begin{footnotesize}
\vspace{-0mm}
	\centering
	\setlength{\tabcolsep}{4.5pt}
	\begin{tabular}{|c||c|c|c||c|c|c|}
		\hline
		Using  & \multicolumn{3}{c||}{$ N = 80 $ } & \multicolumn{3}{c|}{$ N = 168 $}  \\
		\cline{2-7} 
		 constraints  & { Recall }      & { Prec.}       & {F1 }     & {Recall}       & {Prec.}       & { F1  }   \\
		\hhline{|=#=|=|=#=|=|=|}
		%\multicolumn{1}{|l|}{Handcrafted method}                                                                                                                                                              &              &                 &         &              &                 &          \\ \hline
		%\multicolumn{1}{|l|}{\begin{tabular}[c]{@{}l@{}}Binary CE Loss on match/not-match, \\ fixed conversion from ranking to match/not-match\end{tabular}}                                                  &              &                 &         &              &                 &          \\ \hline
		%\multicolumn{1}{|l|}{\begin{tabular}[c]{@{}l@{}}Binary CE Loss on match/not-match, \\ fixed conversion from ranking to match/not-match,\\ considering theoretical limits\end{tabular}}                &              &                 &         &              &                 &          \\ \hline
		Ours: \ding{55}                      &    97.4       &      96.3        &   96.8     &    91.2     &      90.6           &    90.9      \\ \hline
		Ours: \ding{51} &      {\bf 100.0}       &         {\bf 100.0}      &    {\bf 100.0}    &     {\bf 96.3}     &        {\bf 97.6}       &    {\bf 96.7}  \\ 
		\hhline{|=#=|=|=#=|=|=|}
		 Baselines &     \multicolumn{2}{c|}{Recall}      &     \multicolumn{2}{c|}{Prec.}        &   \multicolumn{2}{c|}{F1} \\ \hline
		 Baseline ICP &     \multicolumn{2}{c|}{77.8}      &     \multicolumn{2}{c|}{91.7}        &   \multicolumn{2}{c|}{84.2}   \\ \hline
		 ~\cite{Symmetry17}  &    \multicolumn{2}{c|}{58.3}      &     \multicolumn{2}{c|}{68.2}        &   \multicolumn{2}{c|}{62.9} \\ \hline
	
		\end{tabular} 
		\vspace{0.4mm}
		\begin{spacing}{0.8}
			\caption{{\bf Rotational Symmetry Performance.} For different choices of discretization, $ \bf N $, we report \textit{recall}, \textit{precision} and \textit{F1} measures, averaged across the $4$ symmetry classes. Numbers are in \% .}
			\end{spacing}
		%\caption{ {\color{red} Same table but results measured differently. Here, recall and prec where measured by doing mean of recall and precision of each class. In the other one, recall by summing all true positives dividing by all false positives, independently of the class. And similarly for precision. Which table? Similar results} }
		\label{tab:results_sym}
		\vspace{-2mm}
		 \end{footnotesize}
\end{table} 

\vspace{-1mm}
\subsection{Evaluation Metrics} 
\label{sec:eval_metric} 

\paragraph{Rotational Symmetry Classification} For rotational symmetry classification, we report the mean of \textit{precision}, \textit{recall} and the \textit{F1} scores of the order predictions across different rotational axes ($ \bX, \bY, \bZ $) and object models.

%\vspace{-3mm}
\paragraph{Pose Estimation} For pose estimation, we report the recall performance (R@$ d_\text{rot}^{sym} $) for our top-k predictions. Using the distance measure in the quaternion space, $ d_\text{rot}^{sym} $ (defined in~\secref{sec:loss_func}), we compute the minimum distance of the ground truth pose wrt our top-k predictions and report how many times this distance falls below $ 20^{\circ} $ or $ 40^{\circ} $. The choice of these values are based on the fact that the distance between two adjacent viewpoints for $ N = 80 $ and $ N = 168 $ discretization schemes are $ 21.2^\circ $ and $ 17.8^\circ $, respectively. We also report the average spherical distance, $ d_\text{rot, avg}^{sym} $ of the best match among the top-k predictions relative to the ground truth pose. In all our experiments, we choose top-5\% of the total possible viewpoints, \ie, for $ N = 80 $ and $ 160 $ discretization schemes, $k = 4$ and $8$, respectively.

\begin{figure*}[t!]
\vspace{-4mm}
\begin{minipage}{0.52\linewidth}
	\centering
	\setlength{\tabcolsep}{-5pt}
	\begin{tabular}{cc}
		 \includegraphics[trim={0 6cm 50 6cm}, clip = true,width=0.5\linewidth]{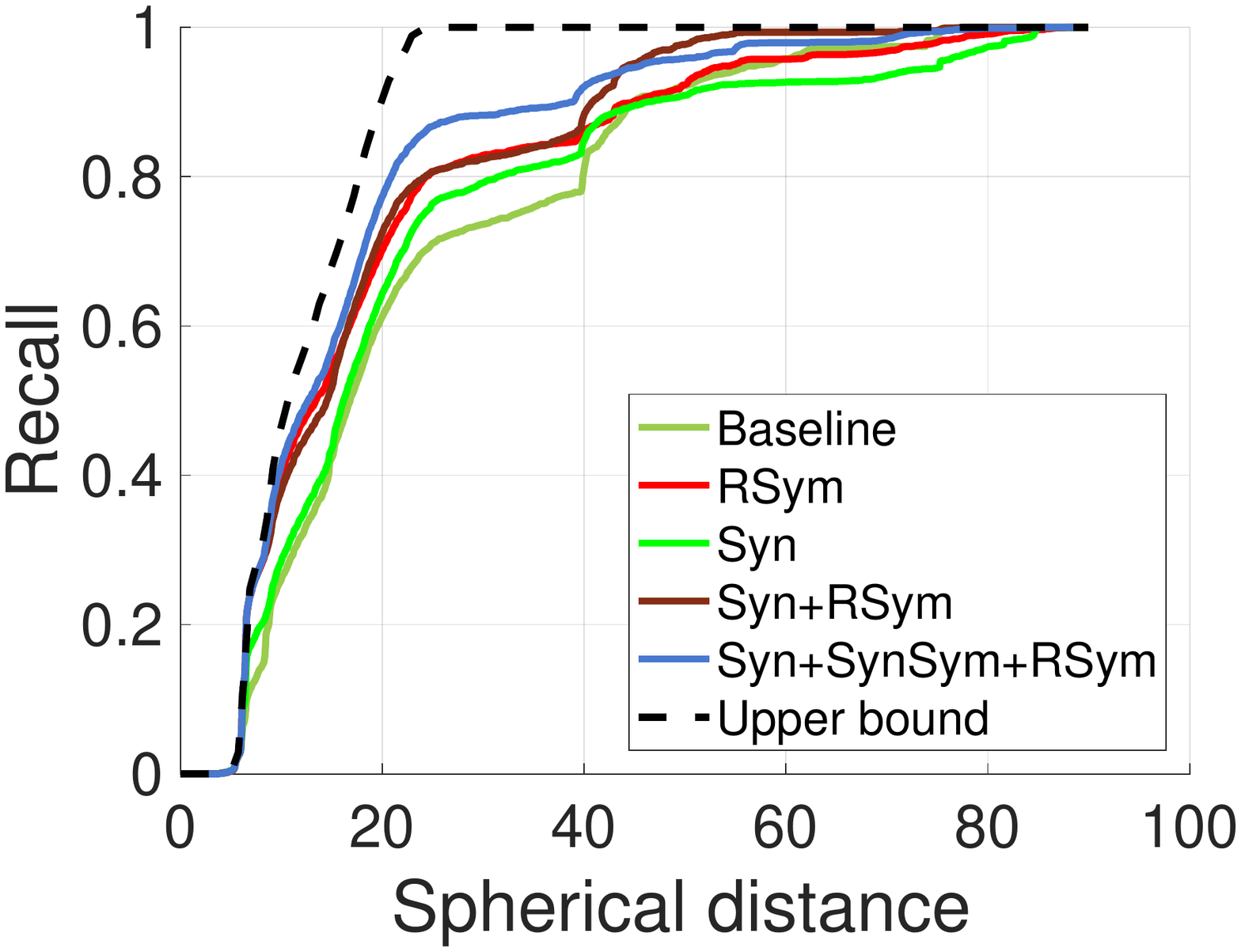} & \includegraphics[trim={0 6cm 50 6cm}, clip = true,width=0.5\linewidth]{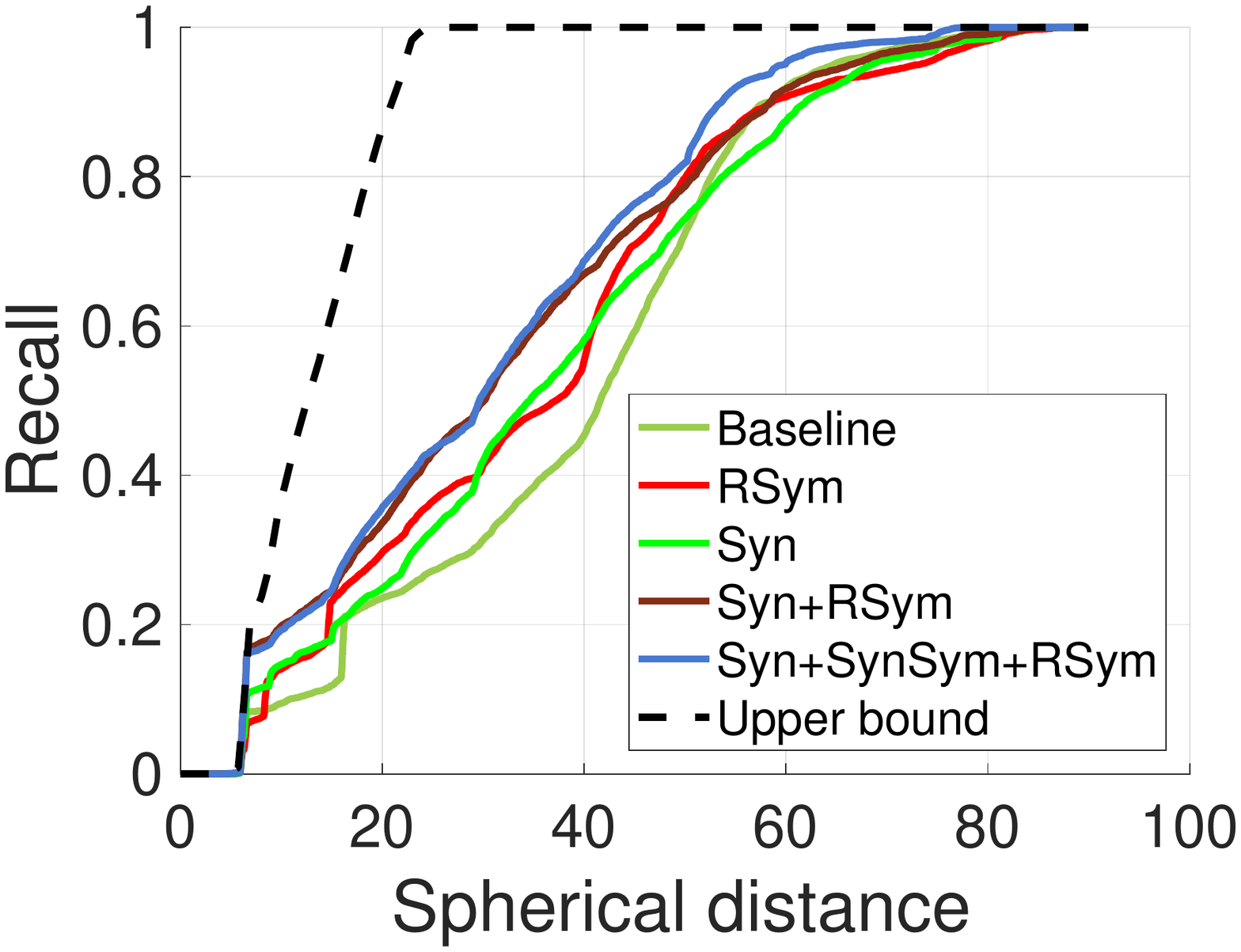} \\
		 (a) \textit{Timestamp}-based split & (b) \textit{Object}-based split
	\end{tabular} 
	\vspace{-1mm}
	\caption{{\bf Recall vs spherical distance}}
	\label{fig:pose_plots}
	\end{minipage}
	\hspace{-2mm}
	\setlength{\tabcolsep}{-0pt}
	\begin{minipage}{0.48\linewidth}
	\vspace{-3mm}
		\begin{tabular}{cc}
 		\includegraphics[trim={3cm 2.0cm 1.5cm 2.4cm}, clip = true,width=0.5\linewidth]{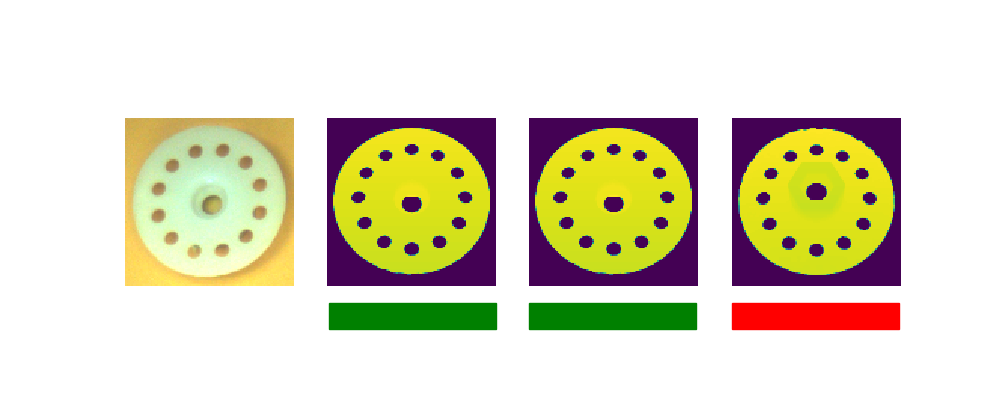}  &
		\includegraphics[trim={3cm 2.0cm 1.5cm 2.4cm}, clip = true,width=0.5\linewidth]{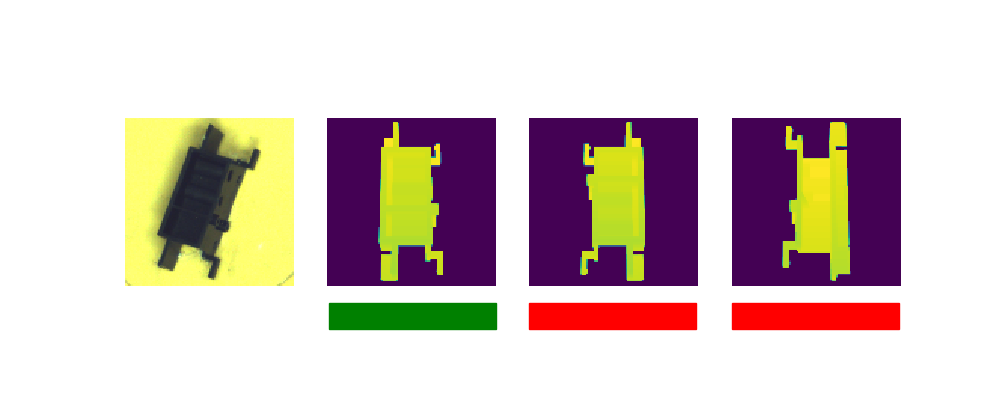}  \\
		\includegraphics[trim={3cm 2.0cm 1.5cm 2.0cm}, clip = true,width=0.5\linewidth]{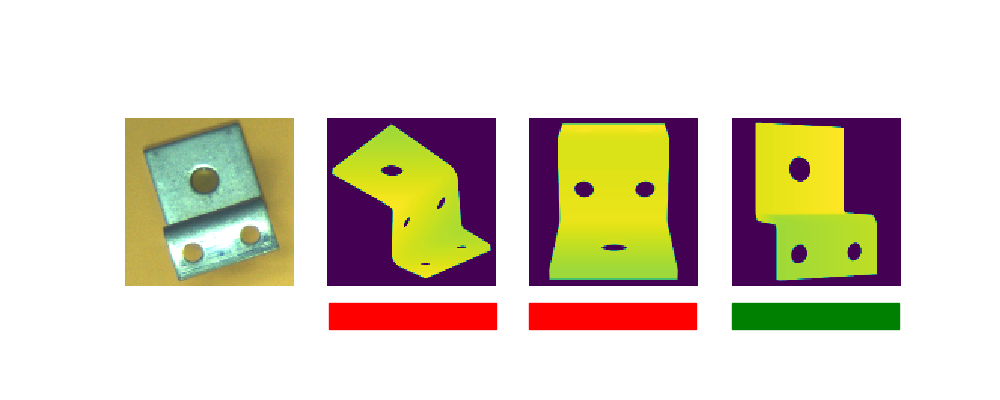}  &
		\includegraphics[trim={3cm 2.0cm 1.5cm 2.4cm}, clip = true,width=0.5\linewidth]{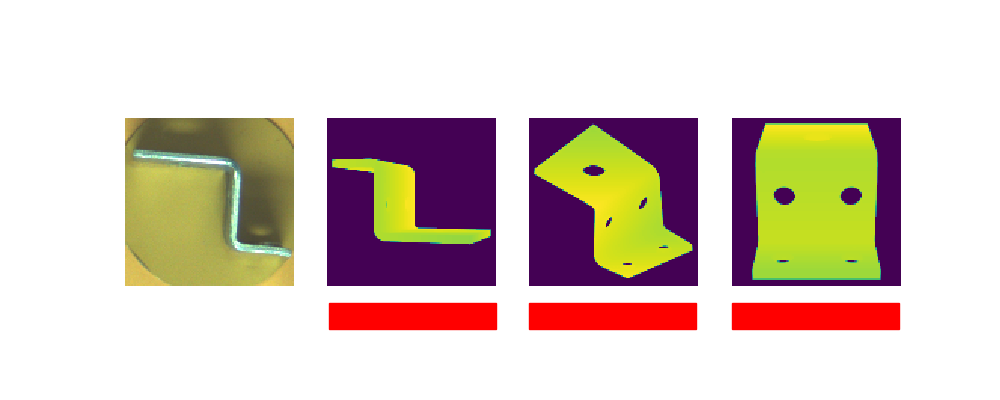}\\[-2mm]
	\end{tabular}
	\centering
	\caption{Qualitative results for pose estimation. Green box indicates correct viewpoint. The bottom-right  shows an error case. }
	\label{fig:epson1034_result}
	\end{minipage}
		\vspace{-6mm}
\end{figure*}

\begin{figure}[t!]%[!htbp]
\vspace{-5.5mm}
	\centering
	\resizebox{1.06\linewidth}{!}{
	\setlength{\tabcolsep}{1pt}
	\begin{tabular}{ccc}
		\hspace{-0.25cm}\includegraphics[trim={0 0 0cm 0}, clip = true,width=3.0cm]{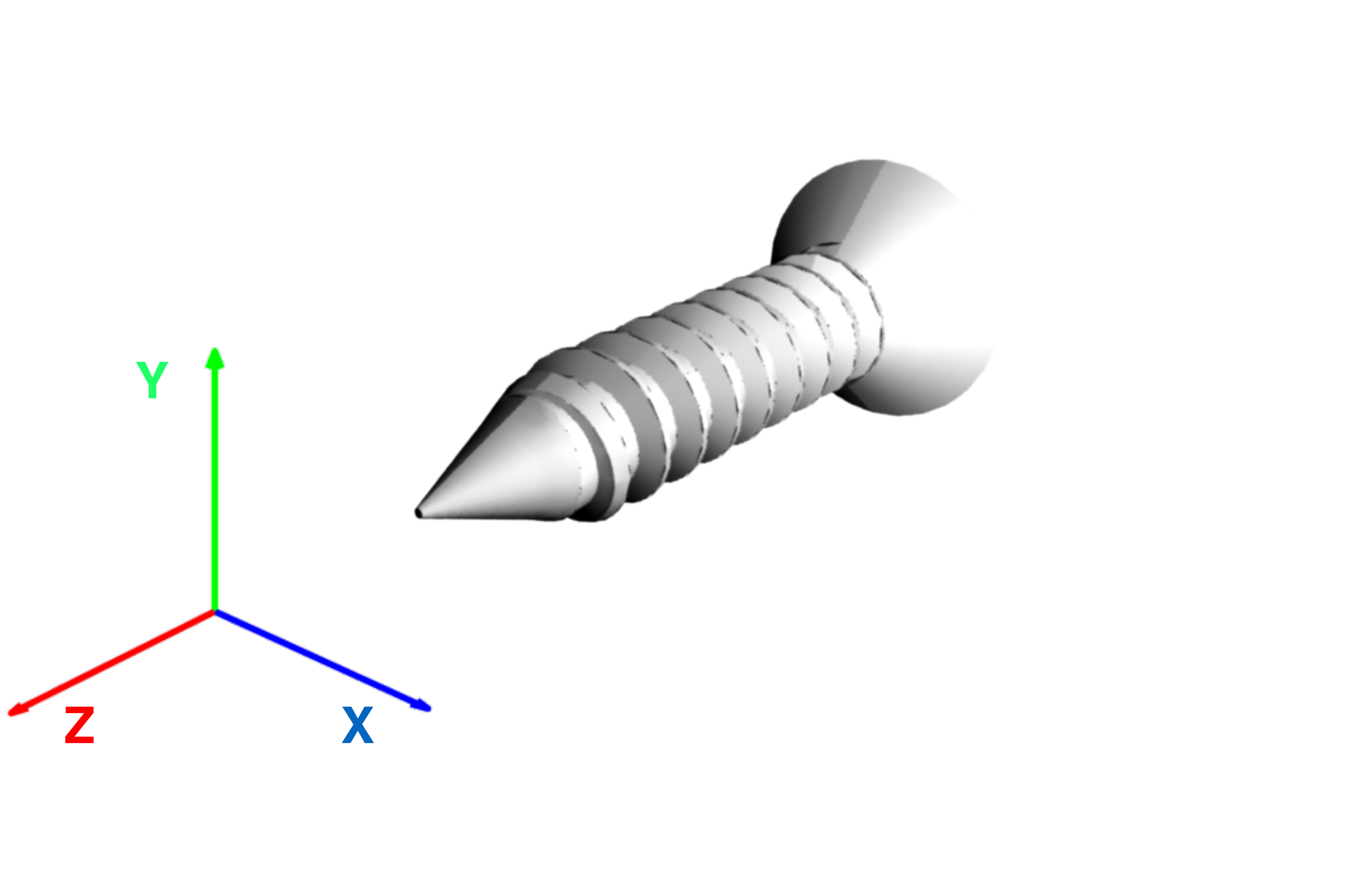} & 
		\hspace{-0.15cm}\includegraphics[trim={0cm 0 0 0}, clip = true,width=2.9cm]{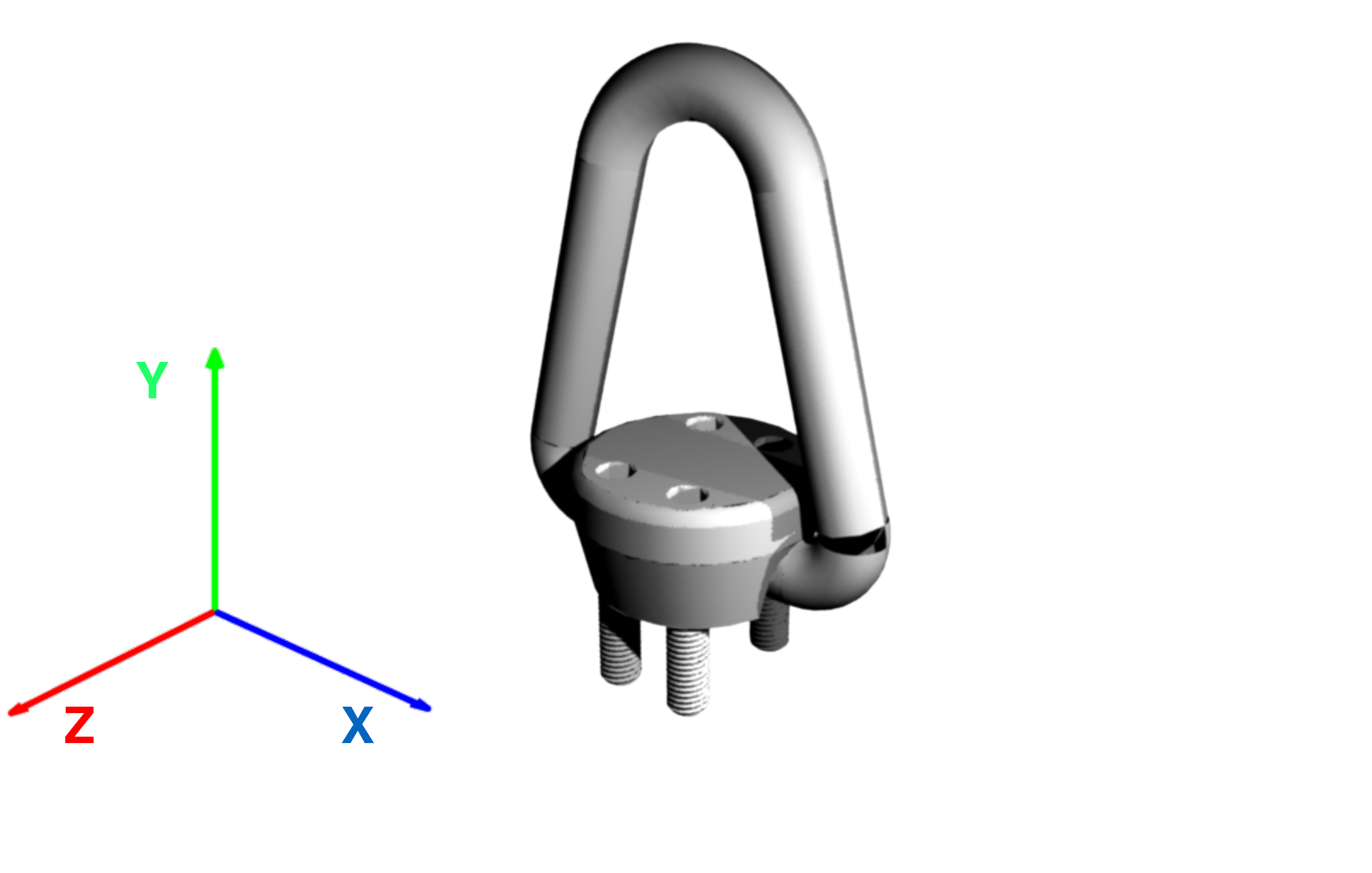} &
		\hspace{-0.15cm}\includegraphics[trim={0 0 0 0}, clip = true,width=3.0cm]{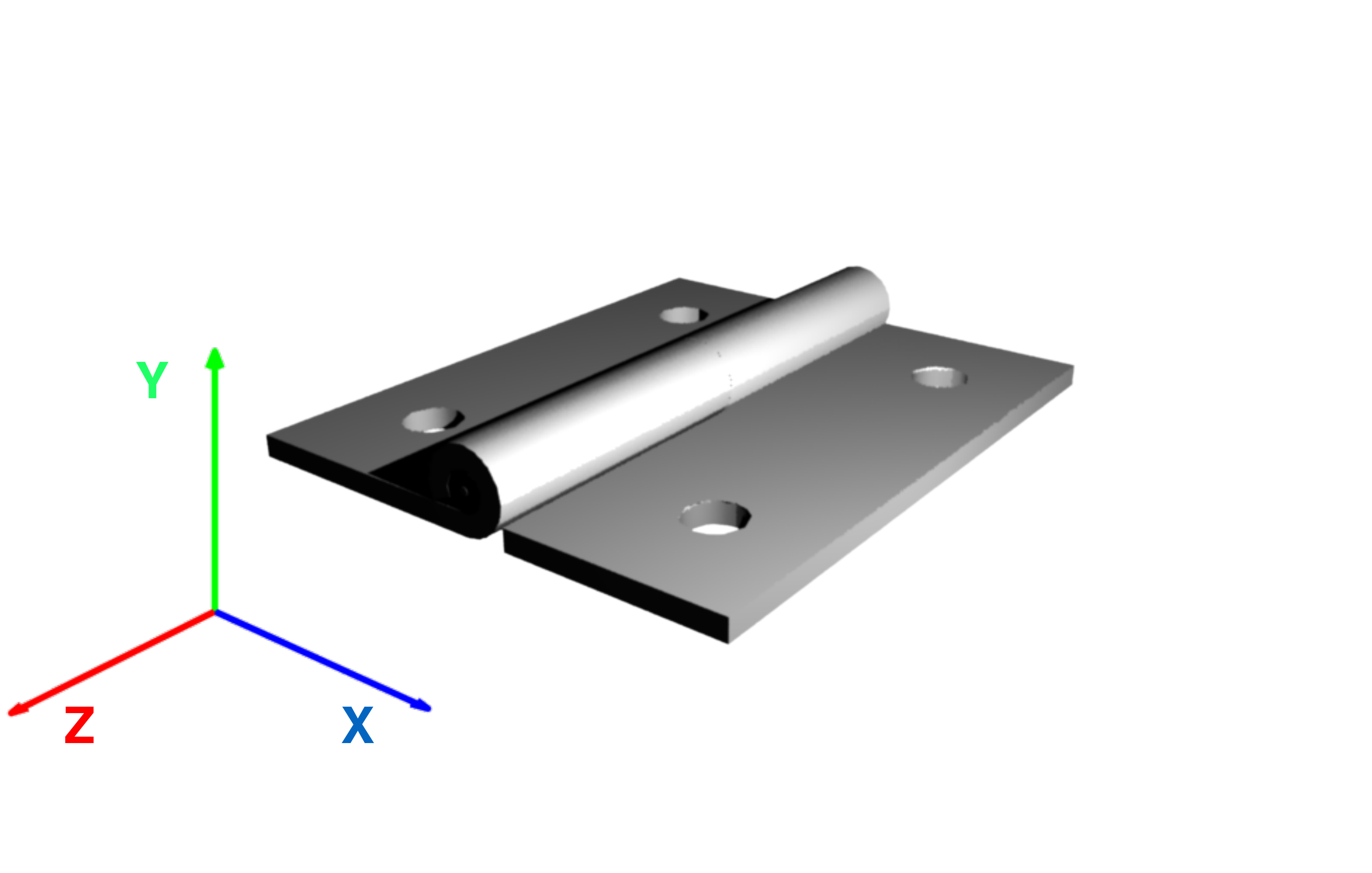} \\[-3mm]
		\hspace{-0.25cm}{\scriptsize $X \sim 1, Y \sim 1, Z \sim \infty.$ } & \hspace{-0.5cm}{\scriptsize $X \sim 1, Y \sim 2, Z \sim 1.$} & \hspace{-1cm}{\scriptsize $X \sim 2, Y \sim 1, Z \sim 1.$} \\[1mm]
		\hspace{-0.25cm}\includegraphics[trim={0 0 0 0}, clip = true,width=2.8cm]{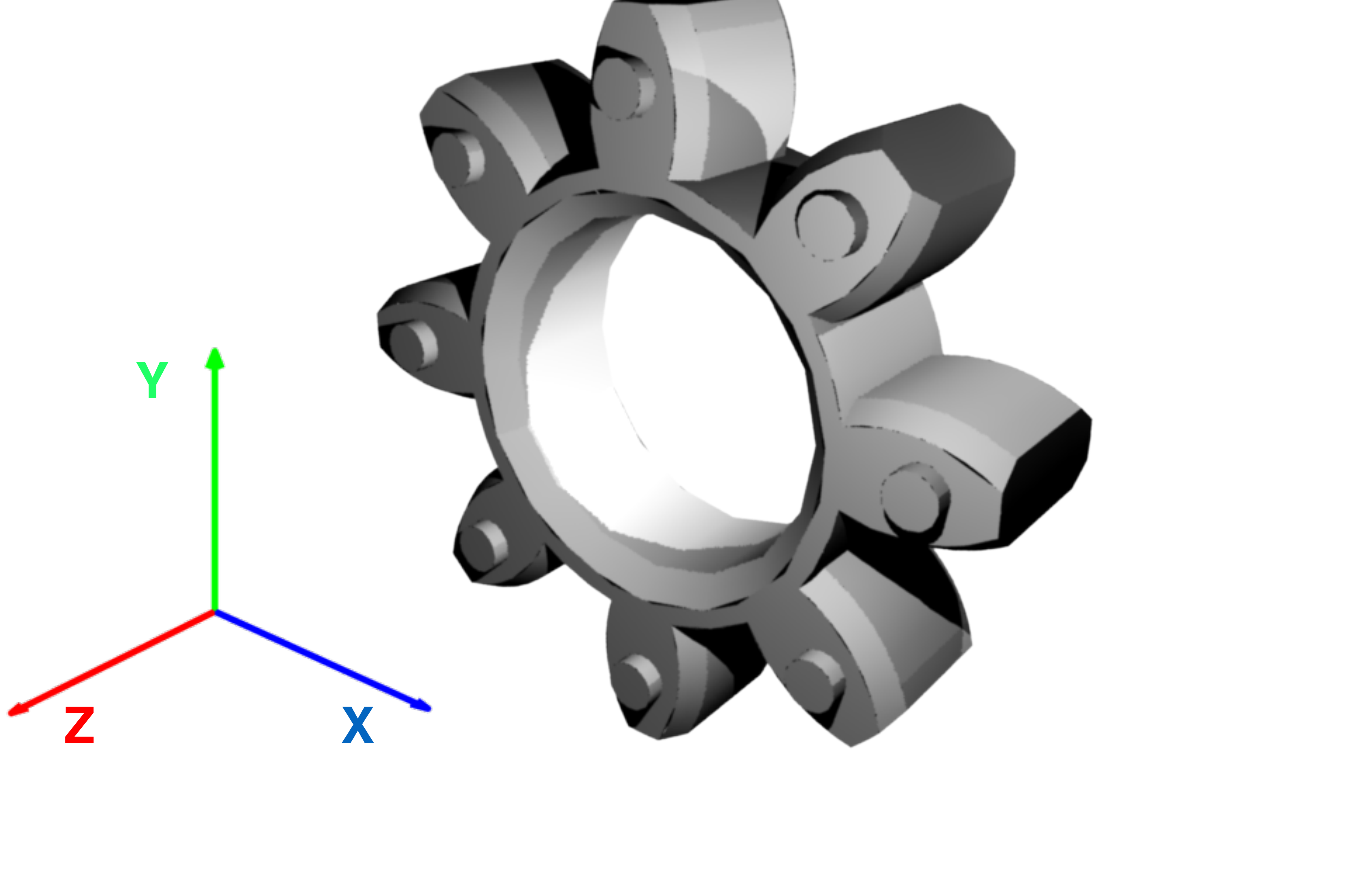} &
		\hspace{-0.15cm}\includegraphics[trim={0 0 0 0}, clip = true,width=3.0cm]{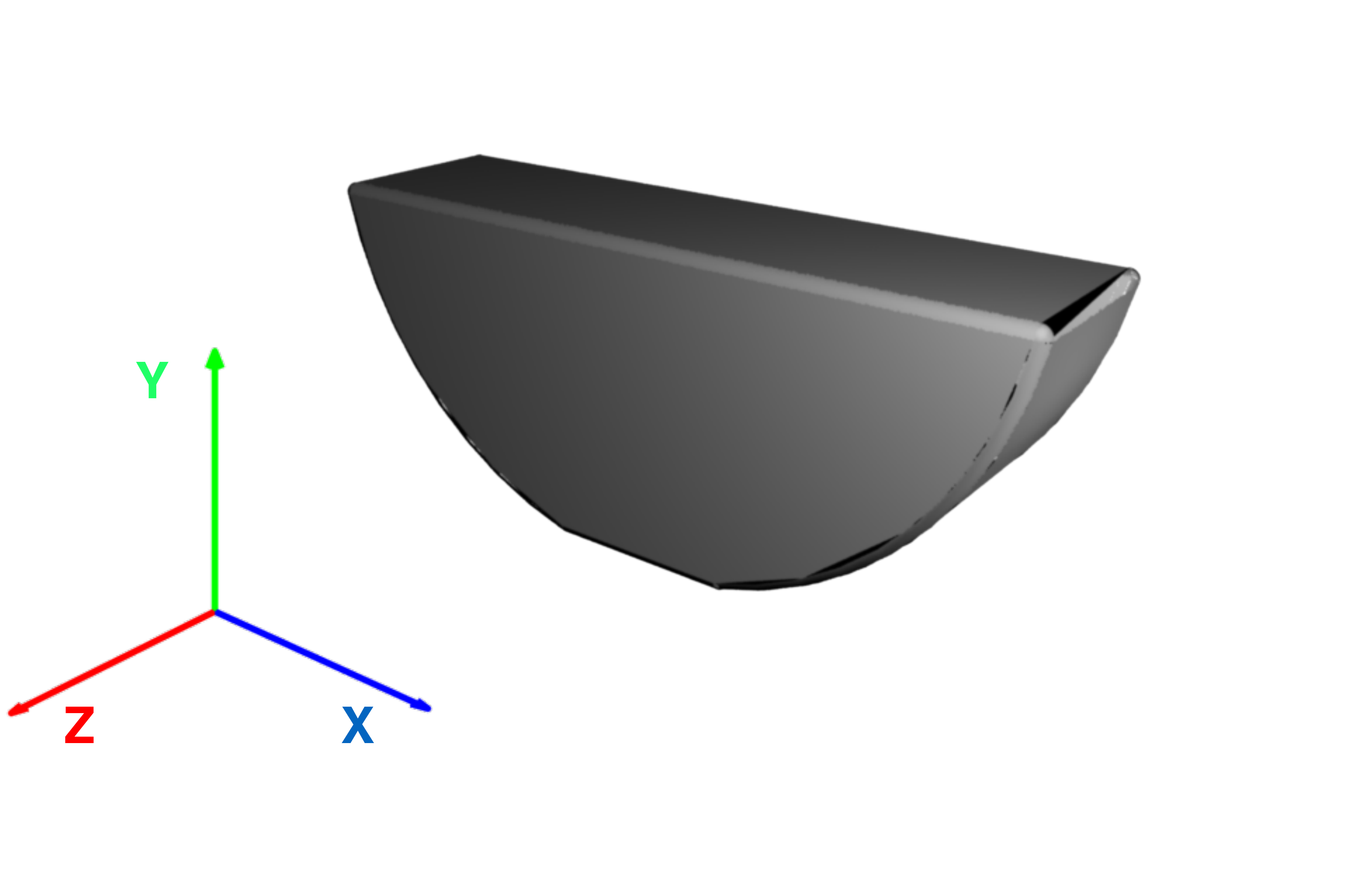} & 
		\hspace{-0.15cm}\includegraphics[trim={0 0 0 0}, clip = true,width=3.0cm]{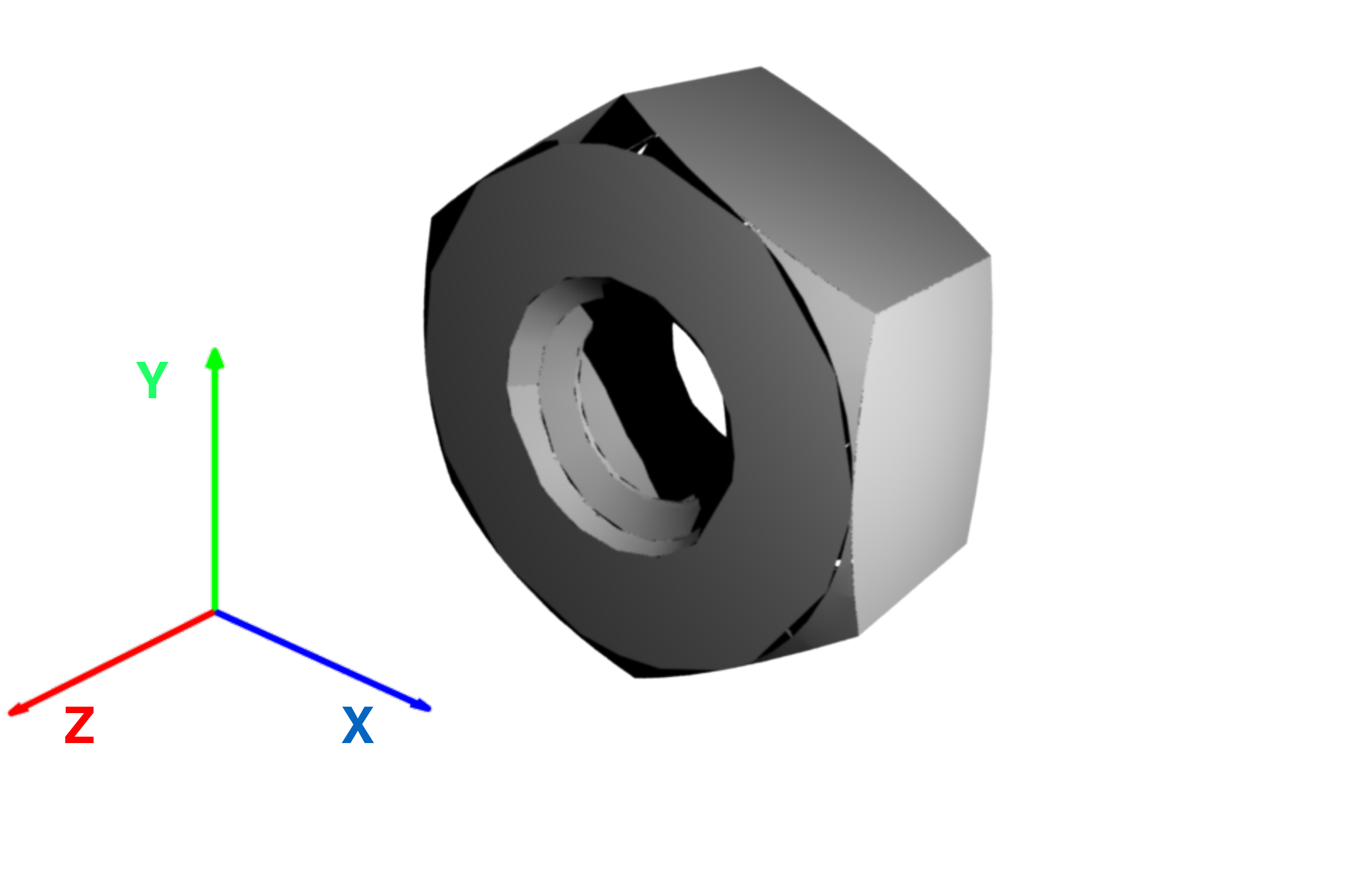} \\[-3mm]
		\hspace{-0.25cm}{\scriptsize $X \sim 2, Y \sim 2, Z \sim \infty.$} & \hspace{-0.5cm} {\scriptsize $X \sim 1, Y \sim 2, Z \sim 1.$} & \hspace{-1cm}{\scriptsize $X \sim 2, Y \sim 2, Z \sim 2.$ }
	\end{tabular}
	}
	\vspace{-3.5mm}
	\caption{Examples of predicted symmetry. Variability of rot. symmetry shows in bottom-left/-right objects. These objects have higher order of symmetry (8 and 6) than what we consider, for which our model predicts $\mathcal{O}_\infty$. }
	\label{fig:symmetries_inferred}
	\vspace{-4mm}
\end{figure}

\subsection{Quantitative Results} 
\label{sec:res_quant}

\paragraph{Rotational Symmetry Prediction} We have rotational symmetry annotations for all 17 objects in the real dataset and 28 objects in the synthetic dataset. We split these into 25 objects for training, 10 for validation, and 10 for test. 

% In this case we are interested in increasing the recall. Since the inferred symmetries are being used to train the pose estimation model in a structured ranking loss, assuming a higher level of symmetry for an object implies the network is not learning as much from it. However, ignoring symmetries implies imposing a margin on the Hinge Loss for views that are actually correct.

We show our quantitative results in~\tabref{tab:results_sym}. The first row shows the performance of our approach by considering order prediction for multiple axes to be independent. In the second row, we show that by reasoning about impossible order configurations,  the performance of our symmetry prediction improves. At a finer discretization, we are also able to predict $\mathcal{O}_3$, making the difference even more evident.

We compare our approach to two baselines. For our first baseline, we use one iteration of ICP to align a CAD model to its rotated version by angles $180^{\circ}$, $90^{\circ}$ and $45^{\circ}$, to detect orders 2, 4 and $\infty$, respectively. When the alignment error is smaller than a threshold (tuned on the training data), we say that the corresponding order is true. This process is done for each of the three axes considered independently.

For our second baseline, we use~\cite{Symmetry17} which finds equivalent points in the mesh. We obtain the amount of these points that are explained by every rotational order considered and, based on a threshold (tuned on the training data), we predict order of symmetry for each axis. This baseline works well when the object considered has only one axis of symmetry, but fails to explain symmetries in more than one axis.

% \multicolumn{1}{|l|}{\begin{tabular}[c]{@{}l@{}}End-to-end + CE loss on MLP + learnt sigmoid conversion\\ + trained on 6000 objs ignoring symmetry \\ + theoretical limits at inference\end{tabular}} 

%\vspace{-3mm}
\paragraph{Pose Estimation}~\tabref{tab:results_pose} reports results for pose estimation for different configurations. The first column corresponds to the the choice of the dataset split, \textit{timestamp} or \textit{object}-based. The second column indicates the usage of the large-scale CAD collection while training our network. Third column  indicates whether our symmetry prediction was used for the (unlabeled) CAD models during training. Fourth column indicates whether the symmetry annotations from the real dataset were used as a supervisory signal to adjust our training loss. For each discretization scheme, we report results for R@$  d_\text{rot}^{sym} $ ($ \phi \in  \{ 20^{\circ}, 40^{\circ} \} $) and $ d_\text{rot, avg}^{sym} $ metrics.

The first row for each split is a baseline which exploits embeddings, but does not reason about symmetry (a.k.a, previous work). We first notice that a model that uses symmetry labels in our loss function, significantly improves the results (first and second row for each dataset split) over the naively trained network. This showcases that reasoning about symmetry is important. Furthermore,  exploiting the additional large synthetic dataset outperforms the base model which only sees the real imagery (first and third rows). Finally, our full model that jointly reasons about symmetry and pose significantly outperforms the rest of the settings. 
%Furthermore, symmetry plays an important role. 
%Furthermore, we show how these factors become more prominent when we want to generalize to unseen objects.  

%The first row corresponds to the baseline approach, \ie, only using the small-sized real dataset without reasoning about rotational symmetry. We can see how using the symmetry labels for just the small set of real images (second row) can improve the quality of embeddings to disambiguate rotational symmetries(third row). While adding a large amount of data can help the network to train better (third row), this improvement is orthogonal to reasoning about rotational symmetries (fourth row). Thus we can see that the accuracy is improved by more than 1.5x when the predicted symmetry labels of the synthetic dataset can be used to scale rotational symmetries to large data. Thus we can see how inferring the symmetries on a large dataset of 3D CAD models and utilize them to train a better embedding function for pose matching. We can also see that using a finer discretization can help us achieve better results at the cost of more computations. 

In~\figref{fig:pose_plots}(a) and (b), we plot  recall vs the spherical distance between the predicted and the GT viewpoint for $ N = 80 $. Since objects are shared across splits in the timestamp based data, the overall results are better than the corresponding numbers for the object-based split. However, the improvement of using synthetic data and rotational symmetries has a roughly 1.7x improvement for object-based split compared to around 1.4x improvement for the timestamp-based split. This shows that for generalization, reasoning about rotational symmetry on a large dataset is essential.

% For the timestamp based split, we can see how using the symmetry labels for even small-scale datasets (second row) during our training step can improve the pose estimation results. Even without using the symmetry labels, using the large scale synthetic dataset (third row) can improve the performance significantly. The fourth row shows that using both a large scale dataset and just the small-scale symmetry labels can boost performance by {\color{red} xx\%}. However the best performance comes when we use supervision from both the predicted symmetry labels of the synthetic dataset along with the annotated symmetry labels of the real dataset. {\color{red} not true for 168 views.} 

Only using the synthetic objects (green plot) can be better than using the symmetry labels for the small real dataset (red and brown plots). However, combining rotational symmetries with large-scale synthetic data (blue plot) gives the best performance. %This shows that both large-scale datasets and reasoning about symmetry are essential for generalizing the pose estimation performance for unseen objects.
 Please refer to the supplementary material
%<<<<<<< HEAD
%=======
%\footnote{{\color{magenta} Available at \url{www.cs.utoronto.ca/~ecorona/supplementary_rotationalsymm.pdf}} }
%>>>>>>> 2b694cace4974f280b2eaeb32da87bb064dacd0d
 for the $ N = 160 $ discretization scheme as well.

\subsection{Qualitative Results}
\label{sec:res_qual} 

We show qualitative results for real and synthetic data.

%\vspace{-4mm}
\paragraph{Symmetry Prediction} Qualitative results for symmetry prediction are shown in \figref{fig:symmetries_inferred}. %generalizes to unseen objects at test time. 
One of the primary reasons for failure is the non-alignment of viewpoints due the discretization. %Since circular shapes (having infinite order of symmetry) can only be represented by a discrete set of vertices, which if not aligned with the coarse viewpoints can result in failure to detect infinite order rotational symmetry. 
Another example of failure are examples of certain order classes that are not present in training. For example, the object in the bottom left of~\figref{fig:symmetries_inferred}) has an order eight symmetry which was not present in the training set.

%Cases of failure can be mostly because of subtly differences in the shape of the network that do not weigh enough to the network to break symmetries. When predicting symmetry from a projected mesh, we lose information and some views can look perfectly symmetric because other symmetry-breaking parts are occluded.

\iffalse
\begin{figure}%[!htbp]
	\centering
	\begin{tabular}{cc}
		\includegraphics[trim={0 0 0 0}, clip = true,width=4cm]{figs/obj1.pdf} & 
		\includegraphics[trim={0 0 0 0}, clip = true,width=4cm]{figs/obj2.pdf} \\
		$X \sim 1, Y \sim 1, Z \sim \infty.$ & $X \sim 1, Y \sim 2, Z \sim 1.$ \\
		\includegraphics[trim={0 0 0 0}, clip = true,width=4cm]{figs/obj3.pdf} &
		\includegraphics[trim={0 0 0 0}, clip = true,width=4cm]{figs/obj4.pdf} \\
		$X \sim 2, Y \sim 1, Z \sim 1.$ & $X \sim 2, Y \sim 2, Z \sim \infty.$ \\
		\includegraphics[trim={0 0 0 0}, clip = true,width=4cm]{figs/obj5.pdf} & 
		\includegraphics[trim={0 0 0 0}, clip = true,width=4cm]{figs/obj6.pdf} \\
		$X \sim 1, Y \sim 2, Z \sim 1.$ & $X \sim 2, Y \sim 2, Z \sim 2.$ 
	\end{tabular}
	\caption{Sample models among the 6669 unlabelled objects. We test our method to predict symmetry on new objects to infer their symmetry. Yellow, black and blue stand for X, Y and Z axes respectively. As seen in the last example, which should be $X \sim 2, Y \sim 2, Z \sim 6.$, some special orders cannot be directly captured with our initial discretized views. }
	\label{fig:symmetries_inferred}
\end{figure}
\fi

%\vspace{-4mm}
\paragraph{Pose Estimation} Examples of results are shown in~\figref{fig:epson1034_result}. In particular, we show images of objects from the real dataset in the first column, followed by the top-3 viewpoint predictions. The views indicated with a green box correspond to the ground truth. 
Most of the errors are due to the coarse discretization. If the actual pose lies in between two neighboring viewpoints, some discriminative parts may not be visible from either of the coarse viewpoints. This can lead to confusion of the matching network.

\vspace{-1mm}
\section{Conclusion}
\label{sec:conc}

In this paper, we tackled the problem of pose estimation for objects that exhibit rotational symmetry. We designed a neural network that matches a real image of an object to rendered depth maps of the object's CAD model, while simultaneously reasoning about the rotational symmetry of the object. Our experiments showed that reasoning about symmetries is important, and that a careful exploitation of large unlabeled collections of CAD models leads to significant improvements for pose estimation.

%\vspace{-3mm}
\vspace{3.5mm}
\begin{spacing}{0.9}
\begin{footnotesize}
\textbf{Acknowledgements:} This work was supported by Epson. We thank NVIDIA for donating GPUs, and Relu Patrascu for infrastructure support.
\end{footnotesize}
\end{spacing}
%\renewcommand{\baselinestretch}{1}
%\blfootnote{\textbf{Acknowledgements:} This work was supported by Epson. We thank NVIDIA for  donating GPUs, and Relu Patrascu for infrastructure support.}

{\ssmall
%\clearpage
\bibliographystyle{amsplain}
\bibliography{references}
}
\end{document}

%%%%%%%%% TITLE
\title{Supplementary Material\\ Pose Estimation for Objects with Rotational Symmetry}
\date{\vspace{-5ex}} % This is to avoid showing current date

\author{Enric Corona$^{1}$, Kaustav Kundu$^{1}$, Sanja Fidler$^{2}$ % <-this % stops a space
 \thanks{$^{1}$ Enric Corona and Kaustav Kundu are with Department of Computer Science,
        University of Toronto
        {\tt\small ecorona@cs.toronto.edu};
        {\tt\small kkundu@cs.toronto.edu}}
\thanks{$^{2}$ Sanja Fidler is with Department of Computer Science,
        University of Toronto, and the Vector Institute
        {\tt\small fidler@cs.toronto.edu}}
        } %

\maketitle

In the supplementary material, we provide a more detailed discussion about rotational symmetry and provide proofs for our claims in~\secref{sec:rot_sym}. We further provide additional results of our approach in~\secref{sec:add_results}.

\section{Rotational Symmetry}
\label{sec:rot_sym}

We start by introducing the notation and definitions as per~\secref{sec:rot_def}, followed by proofs of claims in~\secref{sec:rot_proofs}. In~\secref{sec:rot_limitations}, we discuss two of the limitations in our approach of reasoning about the rotational symmetry.

\subsection{Notation and Definitions}
\label{sec:rot_def}

\paragraph{Rotation Matrix.} We denoted a rotation for an angle $\phi $ around an axis $ \theta $ using a matrix $ \mathbf{R}_{\theta} (\phi) $. For example, if the axis of rotation is the X-axis, then

\[
\mathbf{R}_{X} (\phi) = \left[ \begin{array}{ccc}
1 & 0 & 0 \\
0 & \cos{\phi} & - \sin{\phi}   \\
0 & \sin{\phi}  & \cos{\phi}  \\
\end{array} \right]
\]

\paragraph{Order of Rotational Symmetry.} We say that an object  has an $n$  order of rotational symmetry around the axis $ \theta $, \ie, $ \mathcal{O}(\theta) = n $, when its 3D shape is equivalent to its shape rotated by $ \mathbf{R}_{\theta} \left( \dfrac{2 \pi i}{n} \right), \forall i \in \{0, \ldots, n - 1\} $. % Thus, $ \mathcal{O}_{\theta} (x) = \mathcal{O}_{\theta} (x) =  $

\begin{figure}[t!]
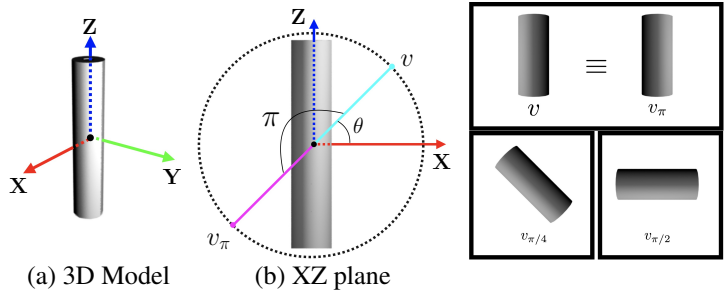

	\begin{center}
	\resizebox{0.8\columnwidth}{!}{
		\begin{tabular}{ccc}
			\raisebox{0.2\height}{\includegraphics[trim={0 0 0 0}, clip = true,width=3.6cm]{figs/rot_sym1.jpg}}
			\put(-110,44){\Large $ \bX $} 
			\put(-9,50){\Large  $ \bY $}
			\put(-62,149){\Large $ \bZ $} & 
			\includegraphics[trim={0 0 0 0}, clip = true,width=5.9cm]{figs/rot_sym2.jpg}
			\put(-12, 60){\Large $ \bX $}
			\put(-105, 155){\Large $ \bZ $}
			\put(-64, 84){\Large $ \theta $}
			\put(-33, 128){\LARGE $ v $}
			\put(-160, 10){\LARGE $ v_\pi $}
			\put(-124, 88){\huge $ \pi $} &
			\includegraphics[trim={0 0 0 0}, clip = true,width=5.9cm]{figs/rot_sym_v2.jpg}  \\
			{\LARGE (a) 3D Model} & {\LARGE (b) XZ plane}  
		\end{tabular}
		}
		\vspace{-1.0mm}
		\caption{Order of Rotational Symmetry}
		\label{fig:rot_sym}
	\end{center}
\vspace{-0.6cm}
\end{figure}

The minimum value of $ \mathcal{O}(\theta) $ is 1, and attained for objects non-symmetric around axis $\theta$. The maximum value is $ \infty $, which indicates that the 3D shape is equivalent when rotated by any angle around its axis of symmetry. This symmetry is also referred to as the revolution symmetry~\cite{bregier2017symmetry}. In ~\figref{fig:rot_sym}, we can see an example of our rotational order definition. For a 3D model shown in~\figref{fig:rot_sym} (a), the rotational order about the $ \bY $ axis is 2, \ie, $ \mathcal{O} (\bY) = 2 $. Thus for any viewpoint $ v $ (cyan) in~\figref{fig:rot_sym} (b), if we rotate it by $ \pi $ about the Y-axis to form, $ v_\pi = \mathbf{R}_{\bY} (\pi) v $, the 3D shapes will be equivalent (\figref{fig:rot_sym} (right)). The 3D shape in any other viewpoint (such as, $ v_{\pi/4} $ or $ v_{\pi/2} $) will not be equivalent to that of $ \bv $. Similarly, we have $ \mathcal{O} (\bZ) = \infty $. 
In our paper, we only consider the values of rotational order to be one of $ \{ 1, 2, 4, \infty \} $, however, our method will not depend on this choice.

\paragraph{Equivalent Viewpoint Sets.} %For the axis, $ \bY $ in~\figref{fig:rot_sym} (middle), we have $ \mathcal{O}_{x} (\theta) = 2 $. 
Let us define the set of all pairs of equivalent viewpoints as $ E_o ( \bY ) = \{ (i, j) | v_j = \mathcal{R}_\theta ( \pi ) v_i \} $, with an symmetry order $o \in \{ 2, 3, \infty \} $.  %Thus, we can similarly define, $ E_o ( \theta ), \forall o \in \{ 2, 3, \infty \} $. 
Note that $ E_1 (\theta) $ is a null set (object is asymmetric). In our case, we have $ E_2 (\theta)\subset E_4(\theta) \subset E_\infty (\theta) $ and $ E_3(\theta) \subset E_\infty (\theta) $.

\subsection{Geometrical Constraints on Orders of Rotational Symmetry about Multiple Axes}
\label{sec:rot_proofs}

%We use the following claim to constrain the output space of the predicted rotational symmetry in our approach. 

\begin{figure}[t!]
\vspace{0mm}
	\begin{center}
		\begin{tabular}{cc}
			\includegraphics[trim={0 0 0 0}, clip = true,width=5cm]{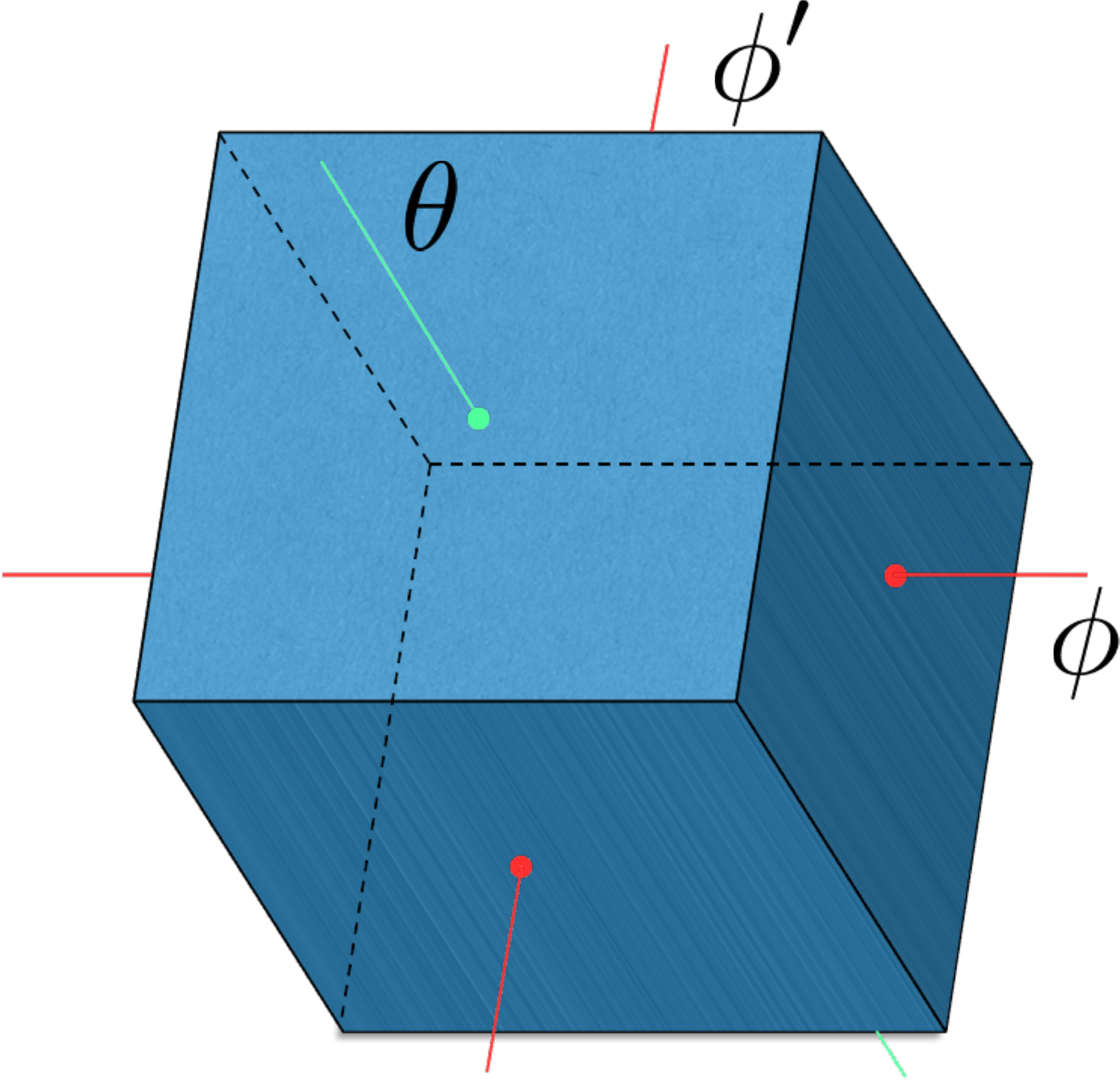} &
			\includegraphics[trim={0 0 0 0}, clip = true,width=5cm]{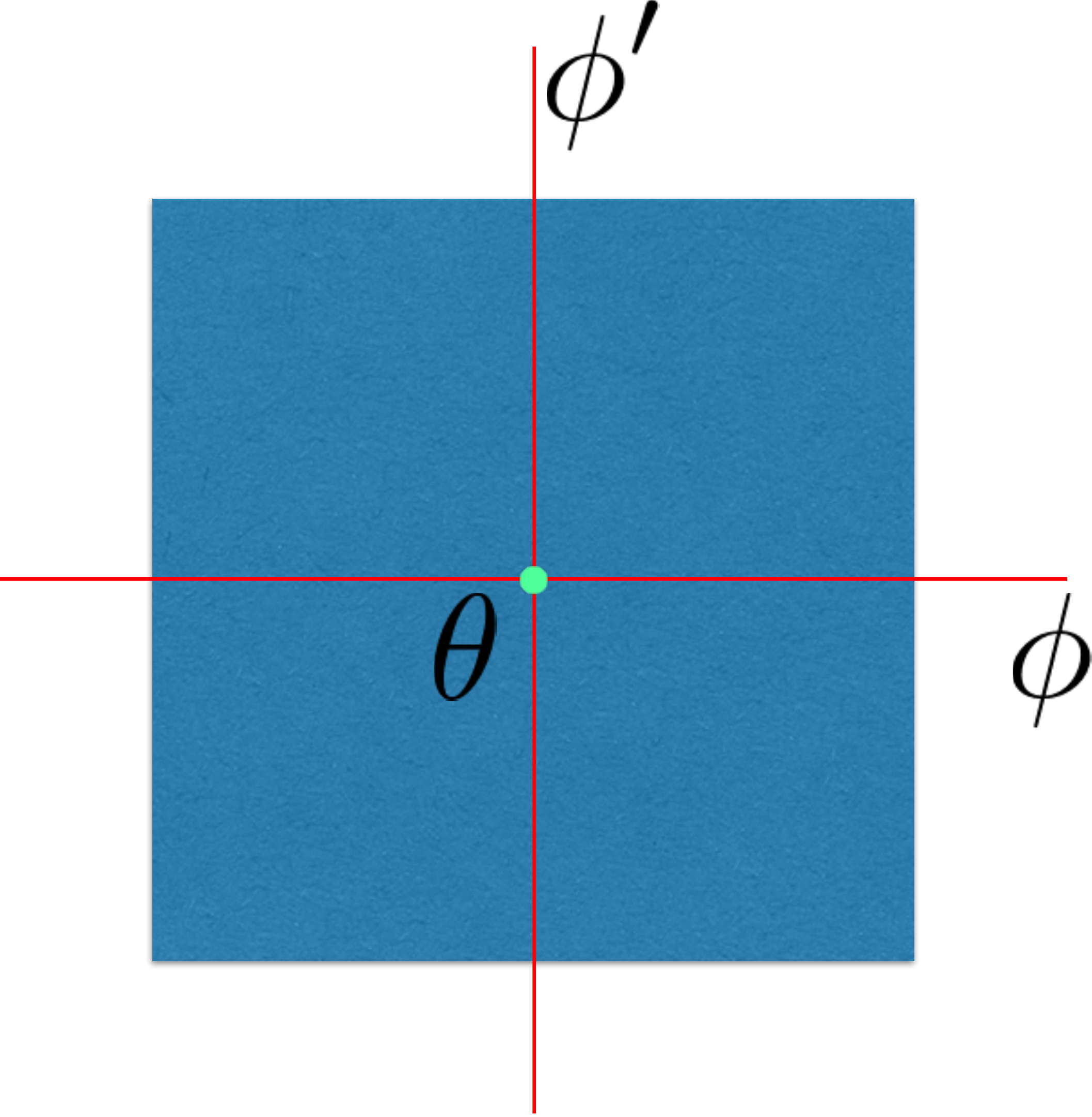}   \\
			(a) Oblique View & (b) Top-down view \\
		\end{tabular}
	
		\vspace{-1mm}
		\caption{Illustration for~\clmref{clm:order}. For the two axes shown, $\mathcal{O}_{cube}(\theta) = 4$ and $\mathcal{O}_{cube}(\phi) = 4$. Equivalent views repeat every $\frac{2\pi i}{n_{\theta}}$ when rotating around the axis $\theta$. From~\clmref{clm:order}, an axis $\phi' = \mathbf{R}_\theta \left( \frac{\pi}{2} \right) \phi$ (for $ i = 1 $) will have rotational order, $\mathcal{O}_{cube}(\phi') = 4$. } %Rotating $\phi$ again for values of $i = 2$ and $i = 3$ presents axes that are aligned with $\phi$ and $\phi'$, respectively. }
		\label{fig:claim1}
	\end{center}
%\vspace{1mm}
\end{figure}

\begin{claim}
	\label{clm:order}
	For an object $ x $ let two non-collinear axes $ \theta $ and $ \phi $ have orders of rotational symmetry, $ \mathcal{O}_x ( \theta ) = n_{\theta} $ and $ \mathcal{O}_x ( \phi ) = n_{\phi} $. Then, $ \forall i \in \{ 0, \ldots, n_{\theta}\} $, the order of symmetry around the axis $ \mathbf{R}_{\theta} (\dfrac{2 \pi i}{n_\theta}) \phi $ is also $ n_\phi $, \ie, $ \mathcal{O}_x \left( \mathbf{R}_\theta \left( \dfrac{2 \pi i}{n_\theta}  \right) \phi \right) = n_\phi $. Similarly $ \forall j \in \{ 0, \ldots, n_\phi \}, \ \mathcal{O}_x \left( \mathbf{R}_\theta \left( \dfrac{2 \pi j}{n_\phi}  \right) \theta \right) = n_\theta $.
	We provide an illustration of this statement with an example in~\figref{fig:claim1}.
	%	X, Y, Z axes have orders of symmetry as $ \mathcal{O}_{x} (X)  = n_X, \mathcal{O}_{x} (Y) = n_Y $ and $ \mathcal{O}_{x} (Z) = n_Z $, then the following holds:
	%	\[
	%	\begin{array}{ccccc}
	%		\mathcal{O}_x \left( \mathbf{R}_{X} \left( \dfrac{2 \pi i}{n_X} \right) Y \right) & = & \mathcal{O}_x \left( \mathbf{R}_{X} \left( \dfrac{2 \pi i}{n_X} \right) Z \right) & = & n_X \\
	%		\mathcal{O}_x \left( \mathbf{R}_{Y} \left( \dfrac{2 \pi j}{n_Y} \right) X \right) & = & \mathcal{O}_x \left( \mathbf{R}_{Y} \left( \dfrac{2 \pi j}{n_Y} \right) Z \right) & = & n_Y \\
	%		\mathcal{O}_x \left( \mathbf{R}_{Z} \left( \dfrac{2 \pi k}{n_Z} \right) X \right) & = & \mathcal{O}_x \left( \mathbf{R}_{Z} \left( \dfrac{2 \pi k}{n_Z} \right) Y \right) & = & n_Z \\
	% 	\end{array}
	%	\]
	%	$ \forall i \in \{0, \ldots, n_x - 1\}, \forall j \in \{0, \ldots, n_x - 1\}, \forall k \in \{0, \ldots, n_x - 1\}  $.
\end{claim}
\begin{proof}
	Since $ \mathcal{O}_x \left( \phi \right) = n_\phi $, for any viewpoint $ v $, the 3D shape of the object $ x $, is equivalent to that from the viewing direction, $ v_{\phi}^j = \mathbf{R}_\phi \left( \dfrac{2 \pi j}{n_\phi} \right) v, \ \forall j \in \{ 1, \ldots, n_\phi - 1 \} $. Also $ \mathcal{O}_x ( \theta ) = n_\theta $ implies that $ x $ is equivalent from the viewpoints, $ v $ and $ v_{\theta}^i = \mathbf{R}_\theta \left( \dfrac{2 \pi i}{n_\theta} \right) v $ for any $ i \in \{ 1, \ldots, n_\theta \} $. Similarly, $ x $ is equivalent from $ v_{\phi}^j $ and $ v_{\theta, \phi}^{i, j} = \mathbf{R}_\theta \left( \dfrac{2 \pi i}{n_\theta} \right) v_\phi^j $. If we can prove this claim for any arbitrary, $ i $ and $ j $, then we are done. For ease of notation, we replace the symbols, $ v_{\phi}^j, v_{\theta}^i, v_{\theta, \phi}^{i, j} $ by $ v_{\phi}, v_{\theta}, v_{\theta, \phi} $, respectively. 
	
	Now, let us consider a point $ p_\phi $ along the $ \phi $ axis, an unit distance away from the origin, $ O $. We rotate the polyhedron, $ P $ (from the vertices, $ v, v_\phi, p_\phi $ and $ O $) by an angle of $ \dfrac{2 \pi i}{n_\theta} $ around the axis, $ \theta $, \ie, each of the vertices are rotated by $ \mathbf{R}_\theta \left( \dfrac{2 \pi i}{n_\theta} \right) $. After the rotation, $ v $ and $ v_\phi $ would coincide with $ v_\theta $ and $ v_{\theta, \phi} $ respectively. The axis, $ \phi $ would become $ \phi' = \mathbf{R}_\theta \left( \dfrac{2 \pi i}{n_\theta} \right) \phi $ and $ p_\phi $ would be transformed to $ p_{\phi'} = \mathbf{R}_\theta \left( \dfrac{2 \pi i}{n_\theta} \right) p_\phi $.
	
	Since rotation is an isometry, the angles are preserved under the rotation transformation. Thus, the angle between $ v_\theta $ and $ v_{\theta,\phi} $ would remain as $ \dfrac{2 \pi j}{n_\phi} $, \ie, $ \angle v_\theta O v_{\theta, \phi} = \dfrac{2 \pi j}{n_\phi} $. Similarly, $ \angle v_\theta O p_{\phi'} = \angle v_{\theta, \phi} O p_{\phi'} = \pi/2 $. Thus, rotating $ v_\theta $ by $ \dfrac{2 \pi j}{n_\phi} $ around the axis, $ \phi' $ would give us $ v_{\theta, \phi} $. 
	
	Now since, the 3D shape is equivalent between the following pair of viewpoints, $ ( v, v_\phi),\ (v, v_\theta) $ and $ (v_\phi, v_{\theta, \phi}) $, thus it would be equivalent from $ v_\theta $ and $ v_{\theta, \phi} $. Thus if $ \mathcal{O}_x ( \phi ) = n_{\phi} $, then $ \mathcal{O}_x \left( \phi' \right) = n_\phi $.
	
	%	Also, since $ \mathcal{O}_x ( \theta ) = n_\theta $, then x is equivalent from $ v $ and $ \mathbf{R}_\theta \left( \dfrac{2 \pi i}{n_\theta} \right) v $ for any $ i \in \{ 0, \ldots, n_\theta \} $. Similarly x is equivalent from $ \mathbf{R}_\phi \left( \dfrac{2 \pi j}{n_\phi} \right) v $ and $ \mathbf{R}_\theta \left( \dfrac{2 \pi i}{n_\theta} \right) \mathbf{R}_\phi \left( \dfrac{2 \pi j}{n_\phi} \right) v, \ \forall j \in \{ 0, \ldots, n_\phi - 1 \} $. Since we have already established that $ x $ is equivalent from $ v $ and $ \mathbf{R}_\phi \left( \dfrac{2 \pi j}{n_\phi} \right) v $, thus $ x $ has to be equivalent from $ \mathbf{R}_\theta \left( \dfrac{2 \pi i}{n_\theta} \right) v $ and $ \mathbf{R}_\theta \left( \dfrac{2 \pi i}{n_\theta} \right) \mathbf{R}_\phi \left( \dfrac{2 \pi j}{n_\phi} \right) v $. Thus, $ \mathcal{O}_x \left(  \mathbf{R}_\theta \left( \dfrac{2 \pi i}{n_\theta}  \right) \phi \right) = n_\phi, \ \forall  i \in \{ 0, \ldots, n_\theta \}  $.
	
	Similarly, we can prove that, $ \forall j \in \{ 0, \ldots, n_\phi \}, \ \mathcal{O}_x \left( \mathbf{R}_\theta \left( \dfrac{2 \pi j}{n_\phi}  \right) \theta \right) = n_\theta $.
	\end{proof}
	
	Please note that here the equivalence corresponds to that of the entire 3D shape and not the projection of the 3d shape onto a plane at a particular viewpoint. Considering the latter is more difficult, we leave it for future work.

\begin{corollary}
	\label{cor:sphere_sym}
	An object is a sphere iff two non-collinear axes have infinite order rotational symmetry.
	%	If two different orthogonal axes have infinite-order rotational symmetry, all possible axes have infinite-order rotational symmetry and the object is an sphere.
\end{corollary}
\begin{proof}
	If an object $ x $ is a sphere, it is symmetric about all directions. Thus it has infinite order rotational symmetry in any pair of axes.
Let us now prove the opposite, i.e., if two non-collinear axes have 	infinite order rotational symmetry, then the object has to be a sphere.

	\paragraph{Case 1:} Let us first prove this corollary for the special case where the two axes are orthogonal to each other. That is, we aim to show that if any two orthogonal axes have infinite order rotational symmetry, then the object is a sphere. Let $ \mathbf{X} $ and $ \mathbf{Z} $ axes have infinite orders of rotational symmetry, \ie, $ \mathcal{O}_x ( \mathbf{X} ) = \mathcal{O}_x ( \mathbf{Z} ) = \infty $. Thus from~\clmref{clm:order}, $ \mathcal{O}_x \left( \mathcal{R}_{\mathbf{Z}} \left( \dfrac{\pi}{2} \right) \mathbf{X} \right) = \mathcal{O}_x (\mathbf{Y}) = \infty $. Since all orthogonal axes have infinite order of rotational symmetry, then the object's shape is equivalent when viewed from any direction. This is only possible for a sphere.
	
	\paragraph{Case 2:} Here we prove the corollary for the more general case where the two non-collinear axes are not orthogonal to each other. If we can show that there exist two orthogonal axes which have infinite order of rotational symmetry, then we can follow Case 1 and claim that the object is a sphere.
	
\begin{figure}[t!]
\vspace{-2mm}
	\begin{center}
		\resizebox{0.9\columnwidth}{!}{
		\begin{tabular}{ccc}
			\includegraphics[width=\linewidth]{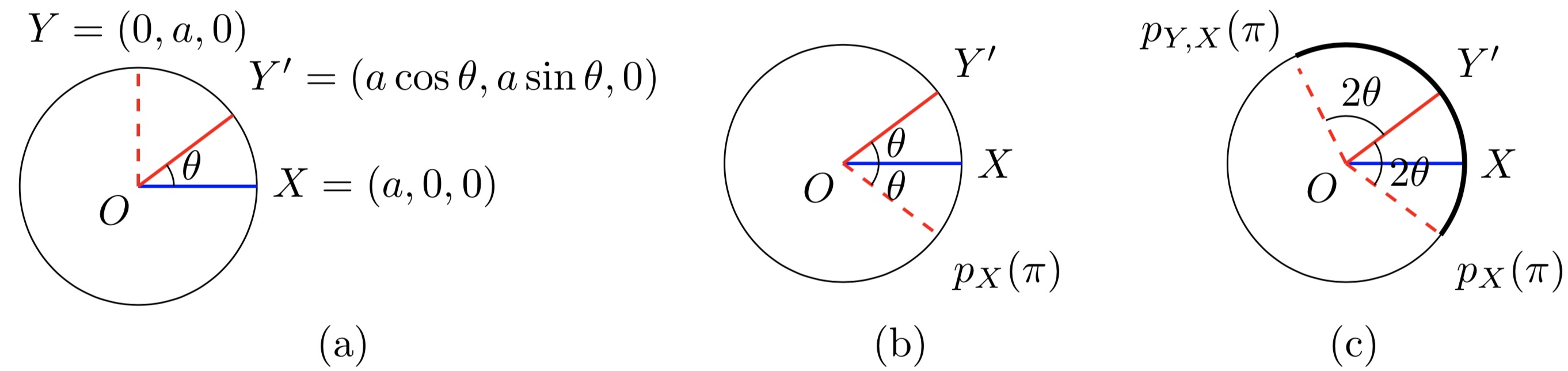}
		\end{tabular}
		}
	\end{center}
	\vspace{-5mm}
	\caption{Illustration for~\corref{cor:sphere_sym}. Please see text for details.}
	\label{fig:sphere_non_orth}
	\vspace{-0.0cm}
\end{figure}
	
	We let the two axes $ \mathbf{X} $ and $ \mathbf{Y'} $ lie on the $ \mathbf{X}-\mathbf{Y} $ plane with an angle of $ \theta $ between them and $ \mathcal{O}_x ( \bX ) = \mathcal{O}_x ( \bY ) = \infty $. Let, $ X = (a, 0, 0) $ and $ Y' = (a \cos{\theta}, a \sin{\theta}, 0) $, as shown in~\figref{fig:sphere_non_orth} (a). The rotation of $ Y' $ about $ X $ by an angle $ \alpha $ has the following form:
	\[
	p_{X} (\alpha) = \mathcal{R}_{\bf X} (\alpha) Y' = \left[ \begin{array}{ccc}
	1 & 0 & 0 \\
	0 & \cos{\alpha} & - \sin{\alpha}   \\
	0 & \sin{\alpha}  & \cos{\alpha}  \\
	\end{array} \right] \left[ \begin{array}{c}
	a \cos{\theta} \\
	a \sin{\theta} \\ 
	0
	\end{array} \right] =  \left[ \begin{array}{c}
	a \cos{\theta} \\
	a \sin{\theta} \cos{\alpha} \\
	a \sin{\theta} \sin{\alpha} \\
	\end{array} \right]
	\]
	
	Here, $ p_{X} (\alpha) $ is the parametric equation of a circle on the $ \mathbf{Y}-\mathbf{Z} $ plane with the center at $ (a \cos{\theta}, 0, 0) $ and radius $ a \sin{\theta} $. The line joining $ Y' $ and $ p_{X} (\pi) = (a \cos{\theta}, -a \sin{\theta}, 0) $ will form the diameter of this circle, as shown in~\figref{fig:sphere_non_orth} (b). Thus, $ \mathcal{O}_x \left( p_{X} (\alpha) \right)=\infty, \forall \alpha \in [ 0, 2 \pi ) $. 
	
	Similarly, for every $ p_{X} (\alpha) $ we can rotate it about the $ \mathbf{Y'} $ axis and those points will have rotational order as $ \infty $. 
%	In addition, we know $p_{Y'}(\alpha)$ and $p_{X}(\alpha)$ are continuous. Since the two rotations  maximum $ p_{Y',X} (\pi) = $ and minimum $ p_{X} (\pi)$. Therefore, the transformation $p_{Y',X}(\alpha)$ is also continuous in this region. Correspondingly,
Every point in the solid arc between $p_{Y',X}(\pi)$ and $p_X(\pi)$ can be obtained from a point $ p_{X} (\alpha) $ rotated on $Y'$. 
	Thus all the points along the solid arc (lying on the XY plane) shown in~\figref{fig:sphere_non_orth} (c) will have order of rotational symmetry $ \infty $. We can continue this process until the solid arc crosses $ Y $ (as shown in ~\figref{fig:sphere_non_orth} (a)) the axis orthogonal to $ \mathbf{X} $ (on the $ \bX-\bZ $ plane) and this will happen for any $ \theta > 0 $ as the arc keeps getting bigger. Thus, $ \mathcal{O} (Y) = \infty $. Since two orthogonal axes have rotational orders as $ \infty $, we can use the previous case to show that this object can only be a sphere.
\end{proof}

\begin{corollary}
	If an object $ x $ is not a sphere, then the following conditions must hold:
	\begin{enumerate}[noitemsep,label=(\alph*)]
		\item It can have up to one axis with infinite order rotational symmetry
		\item If an axis $ \theta $ has infinite order rotational symmetry, then the order of symmetry of any axis not orthogonal to $ \theta $ can only be one.%, \ie, other non-orthogonal axes can't have symmetric viewpoints. ({\color{red} re-phrase})
		\item If an axis $ \theta $ has infinite order rotational symmetry, then the order of symmetry of any axis orthogonal to $ \theta $ can be a maximum of two.
	\end{enumerate}
\end{corollary}
\begin{proof}
	$\ $\\[-5mm]
	\begin{enumerate}[noitemsep,label=(\alph*)]
		\item This follows directly from~\corref{cor:sphere_sym}.
		\item Let us assume that an axis $ \phi $ is not orthogonal to $ \theta $ and $ \mathcal{O}_x (\theta) = \infty, \mathcal{O}_x (\phi) = n > 1 $. Also, let us rotate $ \theta $ to $ \theta' = \mathcal{R}_\phi \left( \dfrac{2 \pi j}{n} \right) \theta $ for some, $ j \in \{ 1, \ldots, n - 1 \} $. Then, from~\clmref{clm:order}, $ \mathcal{O}_x (\theta')  = \infty $. But now two non-collinear axes, $ \theta $ and $ \theta' $ have infinite orders of symmetry. From~\corref{cor:sphere_sym}, this cannot be true for a non-spherical object. Thus we have a contradiction. Thus $ n = 1 $.
		\item It can be proved similarly to the previous part, except that with $ n = 2 $, the two axes are collinear. Thus $ n $ can be either $ 1 $ or $ 2 $.
	\end{enumerate}
	\vspace{-0.6cm}
\end{proof}

Since in our experiments none of the objects is a perfect sphere, we use these constraints to improve the accuracy of our symmetry predicting network. 

%Since none of the objects in our dataset is a sphere, we use this to constrain the output space of our order predictions, \ie, order prediction of multiple axes is not an independent problem. In the next section we show how we incorporate these constraints for order prediction. The proof of this claim is given in the supplementary material.

\subsection{Limitations of our approach}
\label{sec:rot_limitations}

To reason about rotational symmetry, we use the notion of equivalence between 3D shapes. However, in most practical settings and in our experiments the input is a 2D image, which is the projection of the 3D shape onto an image plane. Thus the occluded part of the 3D shape is not visible in the input. For the viewpoints from which the 3D shapes are equivalent, their corresponding projections will also be equivalent. However the opposite is not true, \ie, it is not necessary that two equivalent projections will have equivalent 3D shapes from their corresponding viewpoints. This problem arises because the back-projecting a 2D image into 3D can have infinite solutions. This cannot be handled by our derivations here.

Moreover, while considering rotational symmetry, we establish equivalence between pairs of viewpoints which are rotations about either $ \bX, \bY $ or $ \bZ $ axes. Since we do not reason about pairs of viewpoints which are rotated about any arbitrary axis, our approach does not avoid all the false negative examples in the training data. A trivial extension of our approach to compute rotational orders about any arbitrary axis would scale linearly in both computation and memory costs, which makes this unfeasible. 

% The rotational symmetry that we have handled in our case is from the ones which the roll angle is always the same. However there might be objects whose shape is equivalent once their roll angles are modified. {\color{red} Rephrase this and show its connection to using a single rotation, rather than multiple rotations. Explain with an example (illustration).}

\section{Additional Results}
\label{sec:add_results}

We provide additional quantitative results in~\secref{sec:add_quant} and qualitative results in~\secref{sec:add_qual}.

\begin{figure}[t!]
	\centering
	\setlength{\tabcolsep}{-5pt}
	\begin{tabular}{cc}
		(a) \textit{Timestamp}-based split & (b) \textit{Object}-based split \\
		\includegraphics[trim={0 0 50 0}, clip = true,width=0.48\linewidth]{figs/res_timestamp_80} &
		\includegraphics[trim={0 0 50 0}, clip = true,width=0.5\linewidth]{figs/res_1034_80} \\
		\multicolumn{2}{c}{$ N = 80 $ discretization scheme} \\
		\includegraphics[trim={0 0 0 0}, clip = true,width=0.48\linewidth]{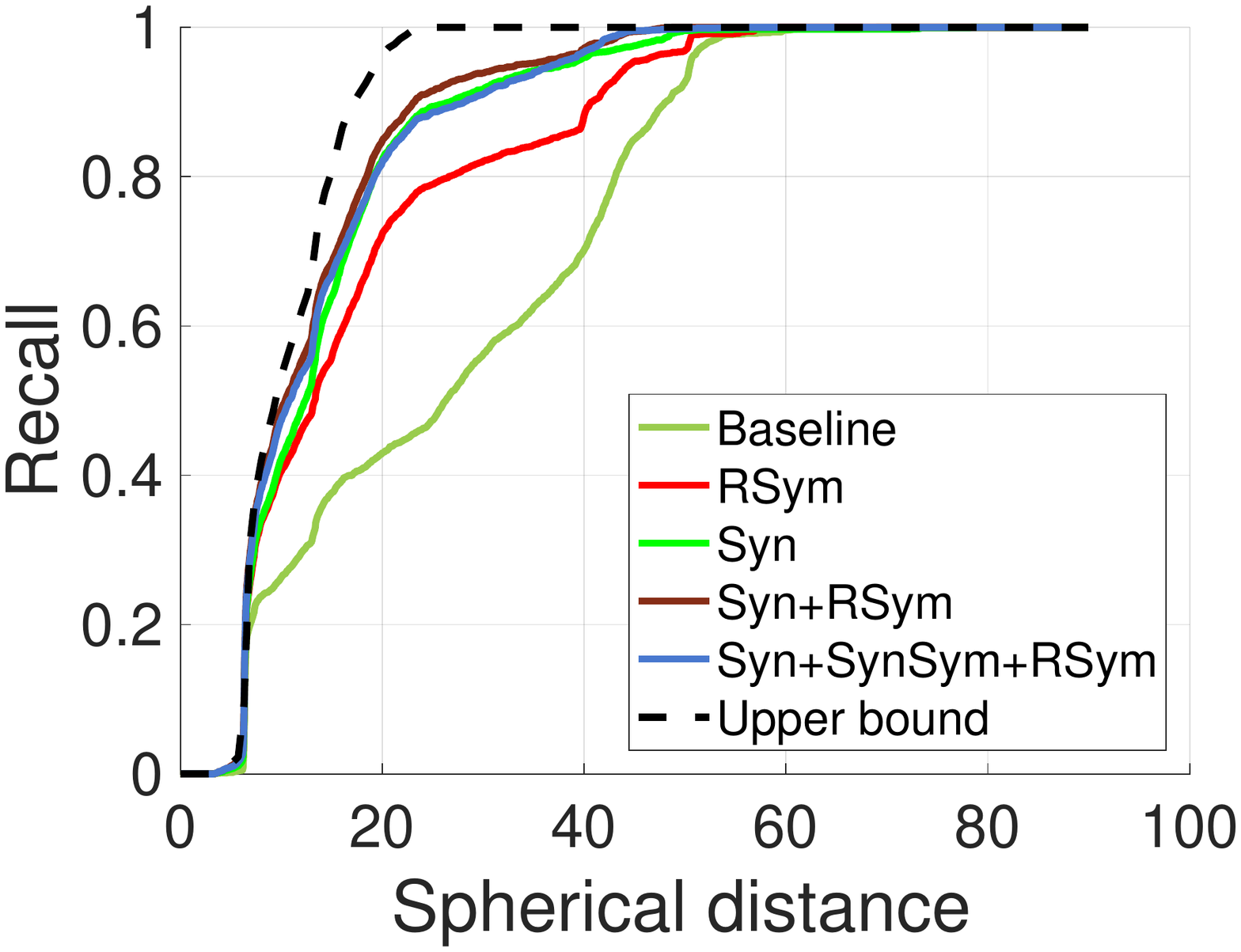} &
		\includegraphics[trim={0 0 0 0}, clip = true,width=0.5\linewidth]{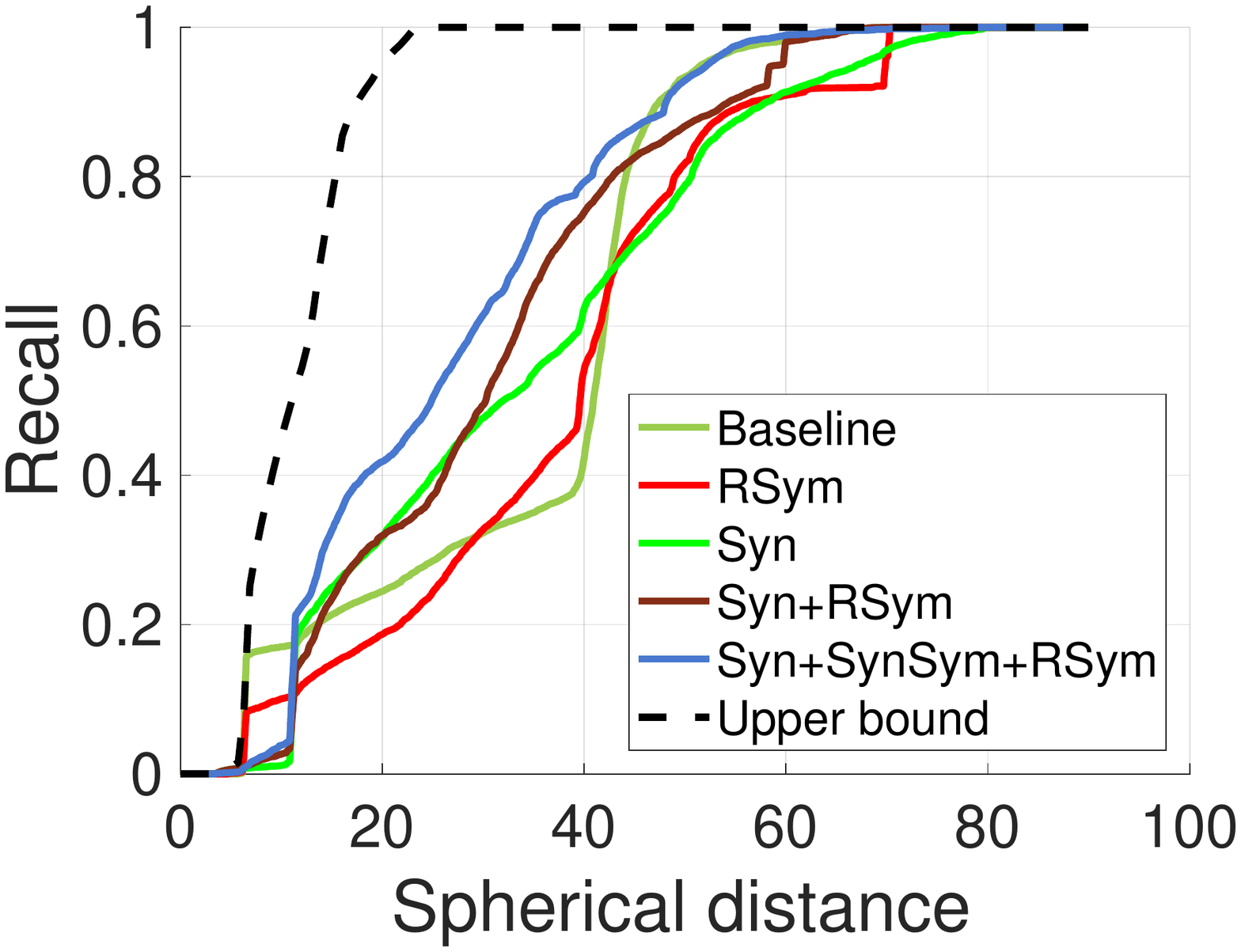} \\
		\multicolumn{2}{c}{$ N = 168 $ discretization scheme} \\
	\end{tabular} 
	\caption{{\bf Recall vs spherical distance}}
	\label{fig:pose_plots}
\end{figure}

\subsection{Pose Estimation}
\label{sec:add_quant}

In~\figref{fig:pose_plots}(a) and (b), we plot  recall vs the spherical distance between the predicted viewpoint and the GT viewpoint. The first and second rows depict the results from the $ N = 80 $  and $ N = 168 $ discretization schemes respectively. We can see that across different discretization schemes, reasoning about rotational symmetry on a large dataset is essential for achieving a good generalization performance.

\subsection{Qualitative Results}
\label{sec:add_qual}

\paragraph{Rotational Symmetry Prediction.}

\begin{figure}[t!]%[!htbp]
\vspace{-2mm}
	\centering
	\resizebox{1.06\linewidth}{!}{
	\setlength{\tabcolsep}{1pt}
	\begin{tabular}{cccc}
		\includegraphics[trim={0 0 0 0}, clip = true,width=3.3cm]{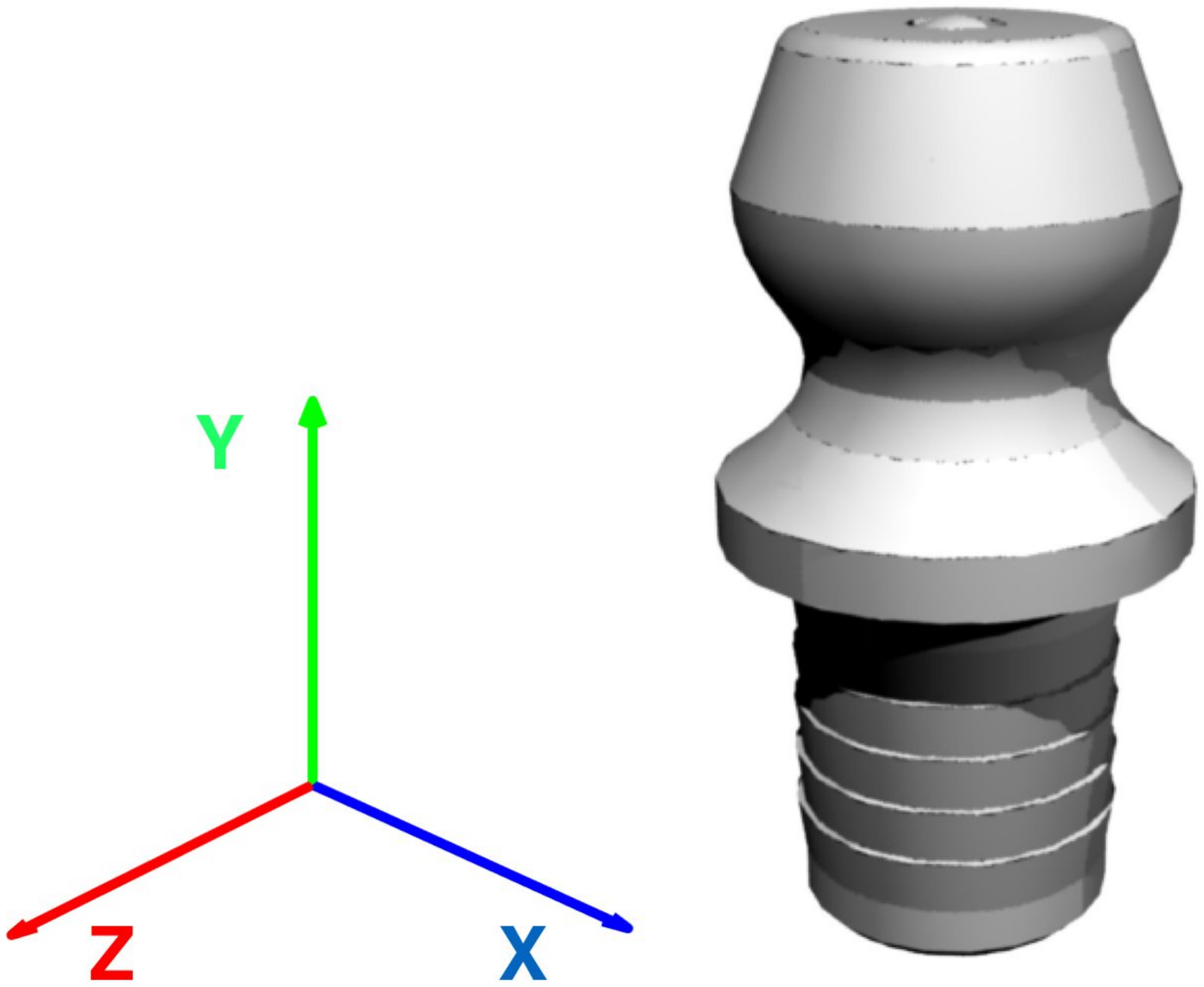} & 
		\includegraphics[trim={0 0 0 0}, clip = true,width=3.3cm]{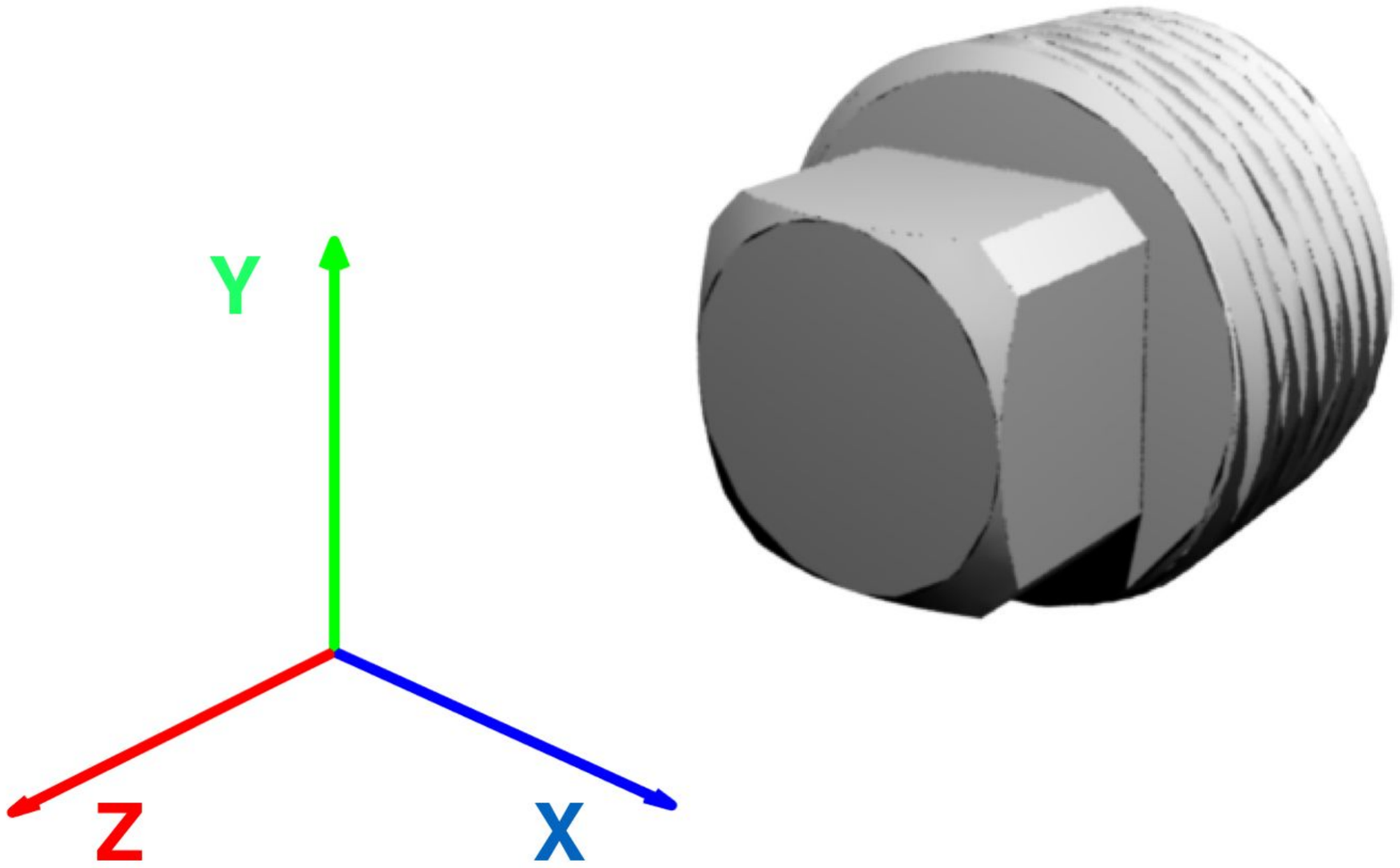} &	
		\includegraphics[trim={0 0 0 0}, clip = true,width=3.3cm]{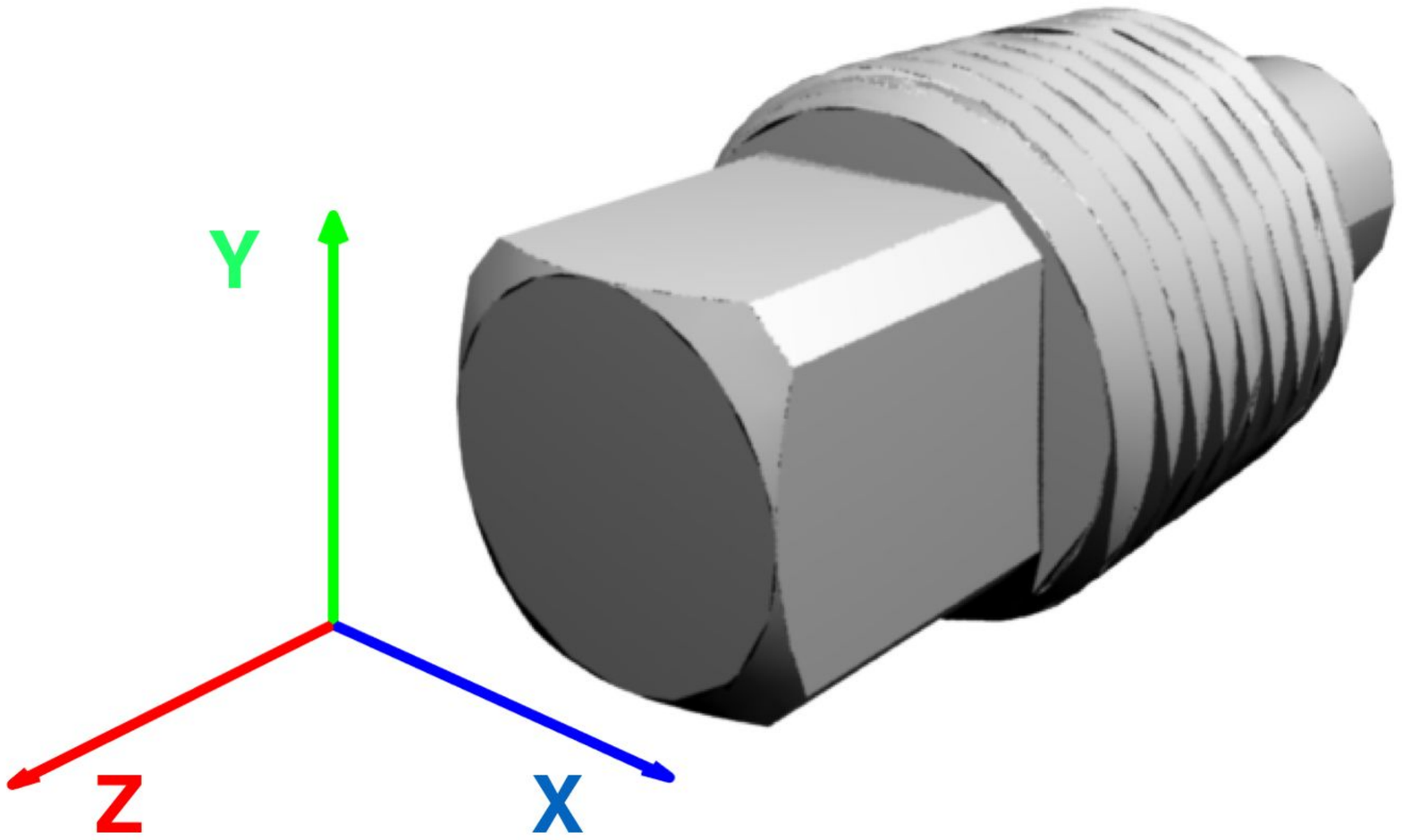} & 
		\includegraphics[trim={0 0 0 0}, clip = true,width=3.3cm]{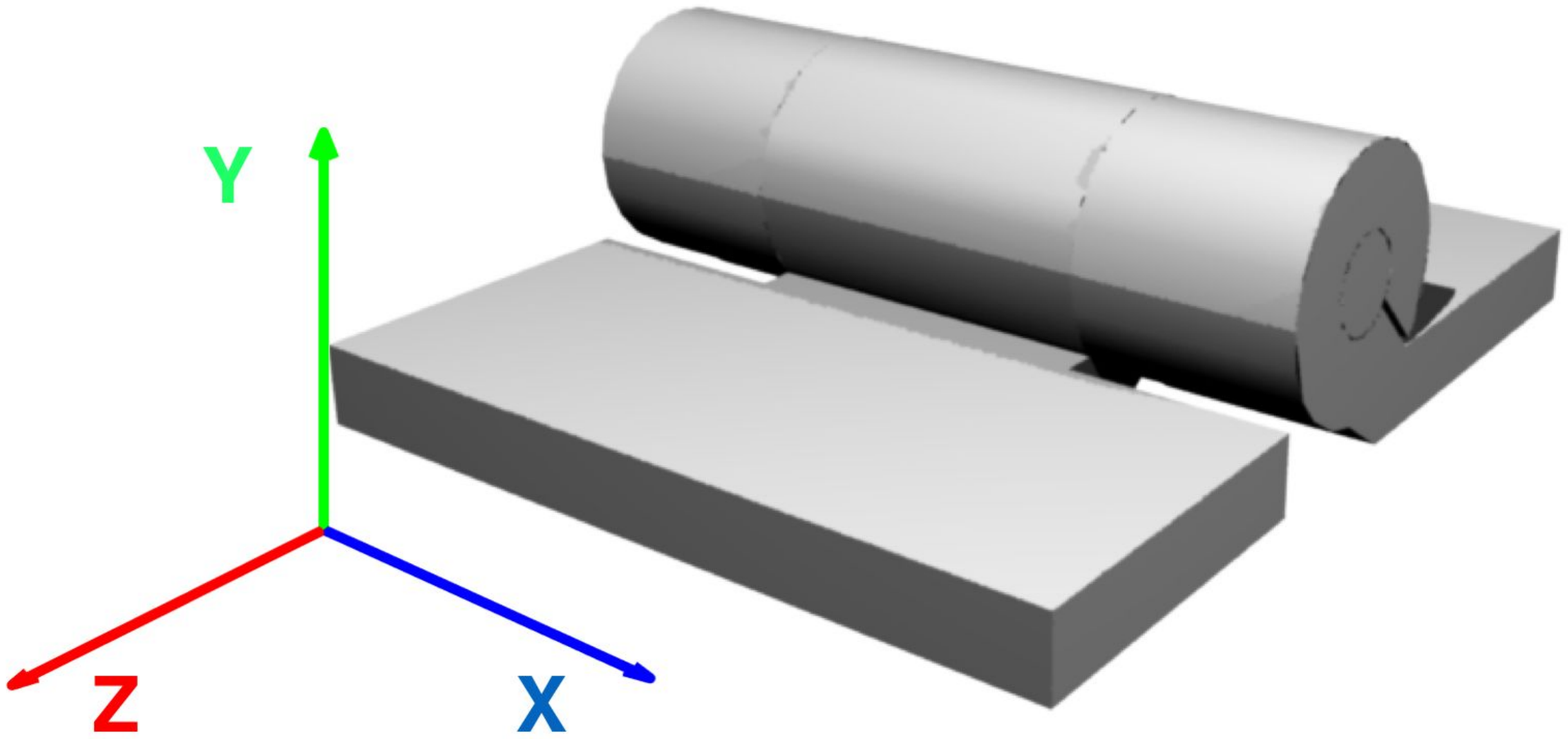}  \\
		{\scriptsize $X \sim 2, Y \sim \infty, Z \sim 2.$ } & \hspace{-0.5cm}{\scriptsize $X \sim 1, Y \sim 1, Z \sim \infty.$} & \hspace{-1cm}{\scriptsize $X \sim 2, Y \sim 2, Z \sim \infty.$} & \hspace{-1cm}{\scriptsize $X \sim 1, Y \sim 1, Z \sim 1.$}  \\	
		{\scriptsize (i)} & {\scriptsize (ii)} & {\scriptsize (iii)} & {\scriptsize (iv) } \\	
		\includegraphics[trim={0 0 0 0}, clip = true,width=3.3cm]{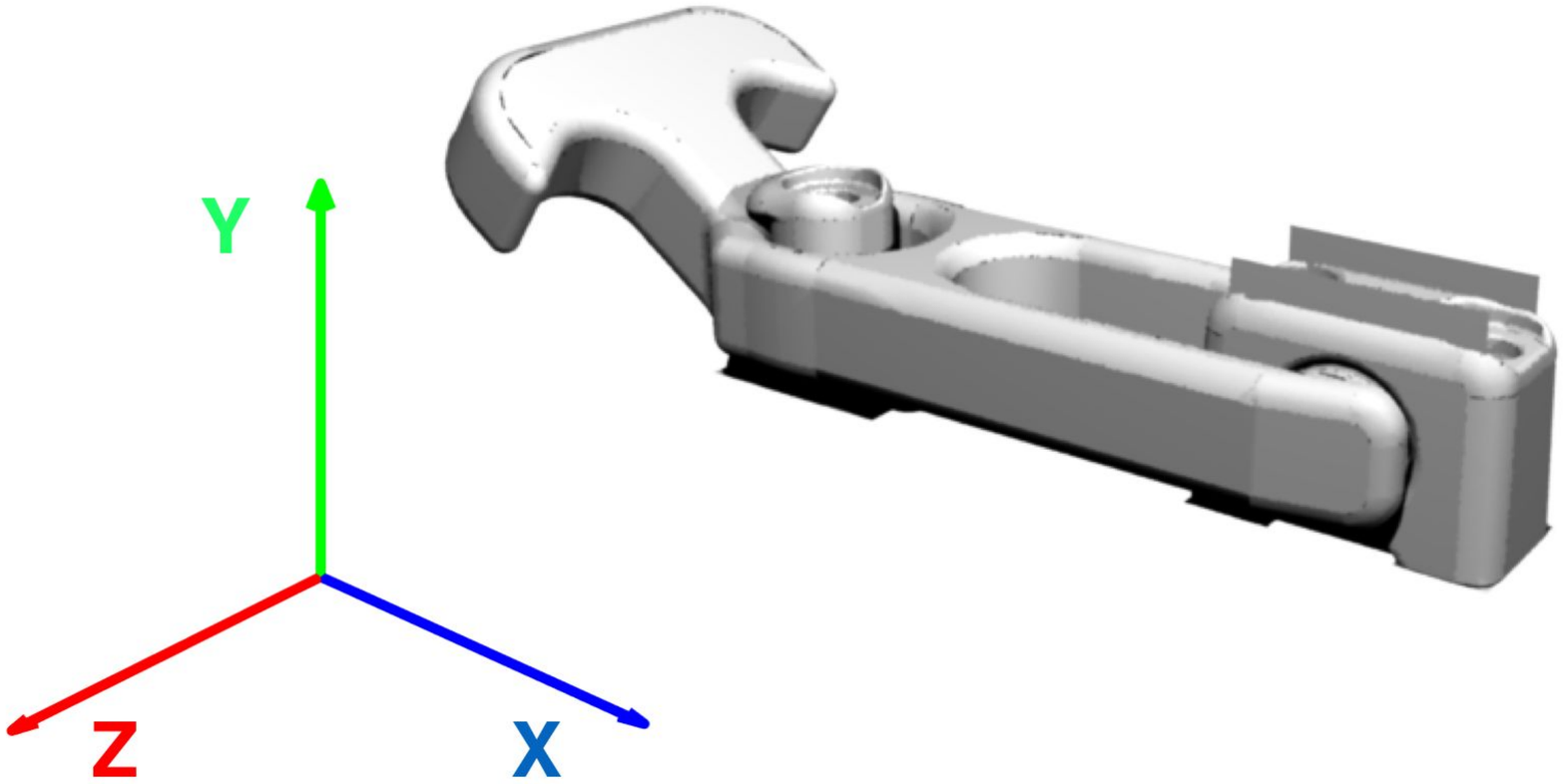} & 
		\includegraphics[trim={0 0 0 0}, clip = true,width=3.3cm]{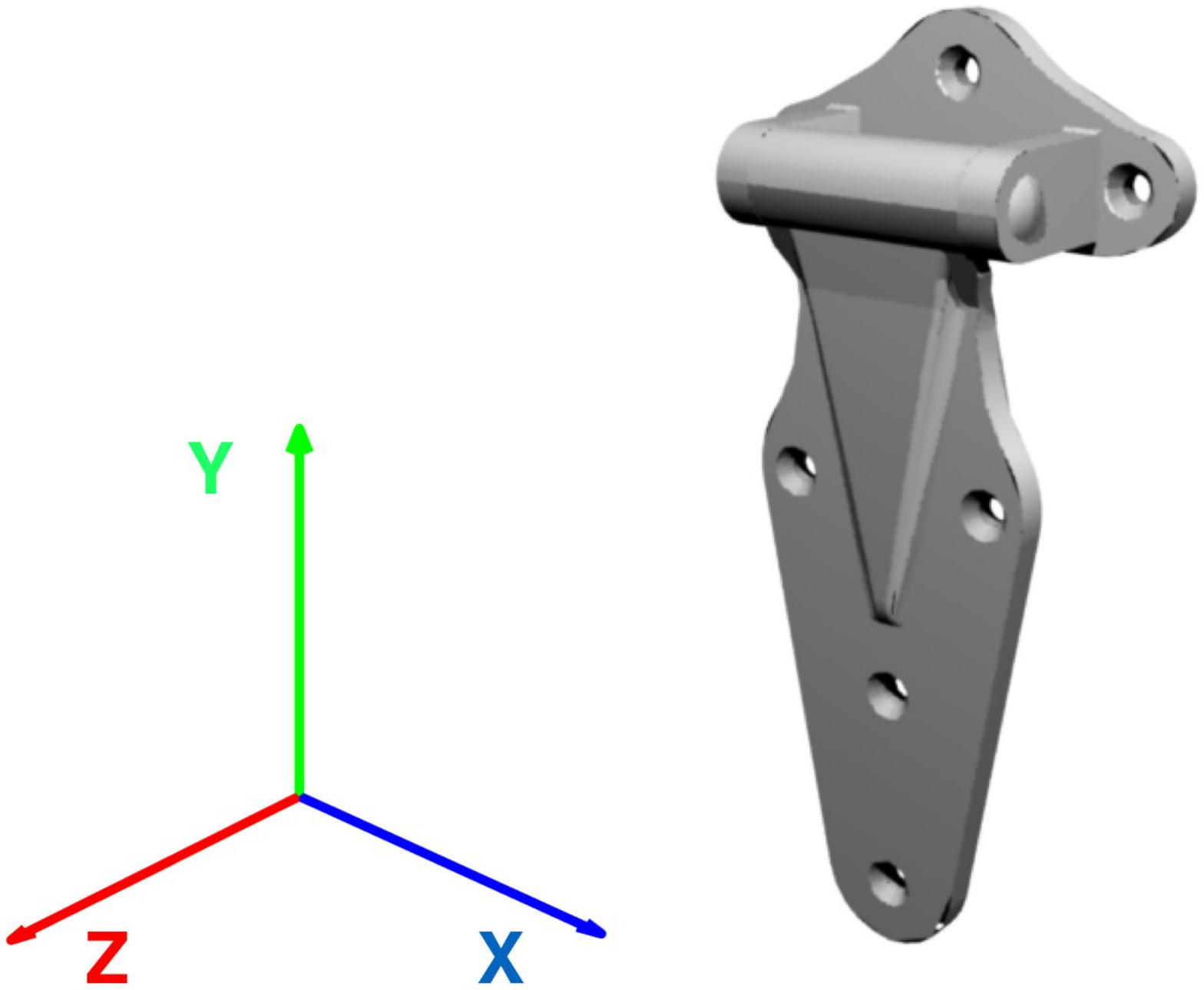} &	
		\includegraphics[trim={0 0 0 0}, clip = true,width=3.3cm]{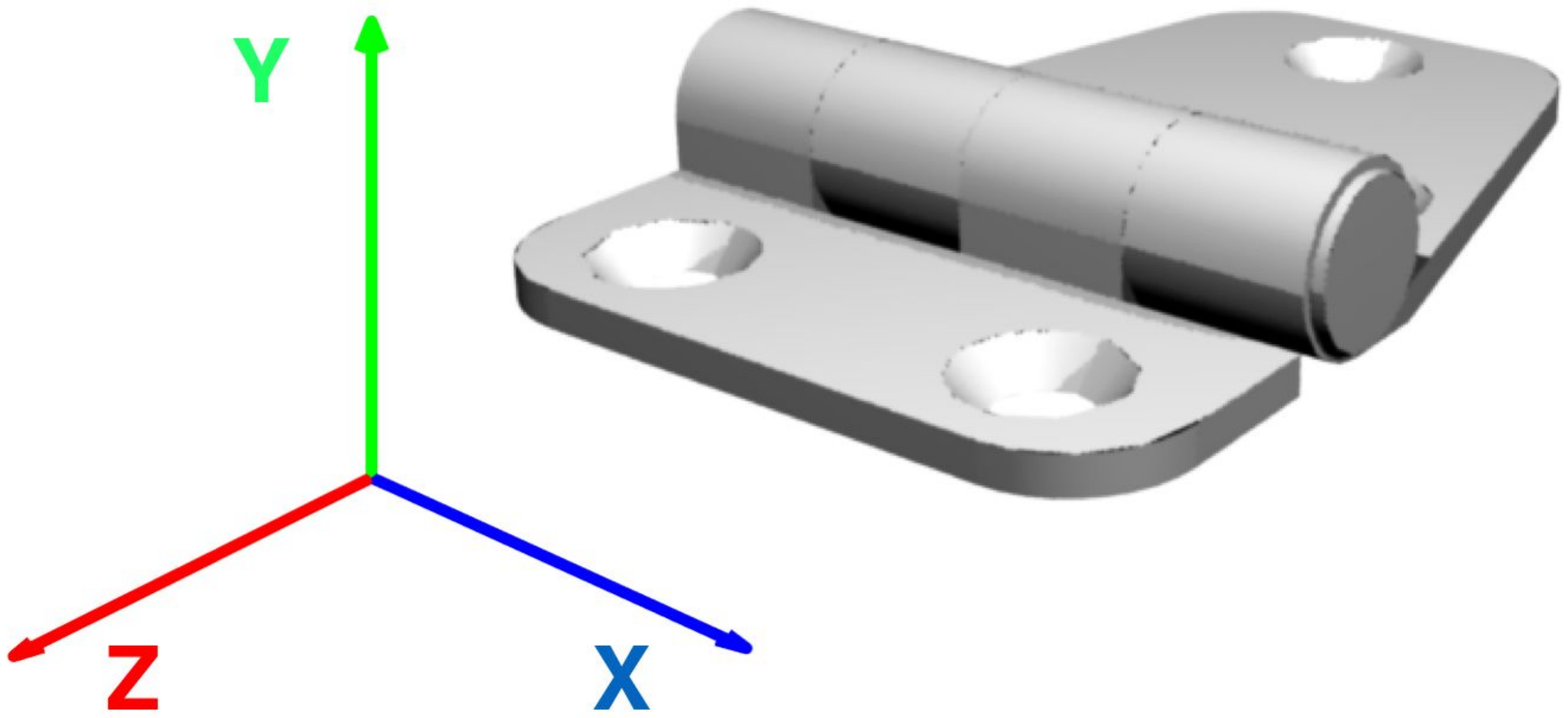} & 
		\includegraphics[trim={0 0 0 0}, clip = true,width=3.3cm]{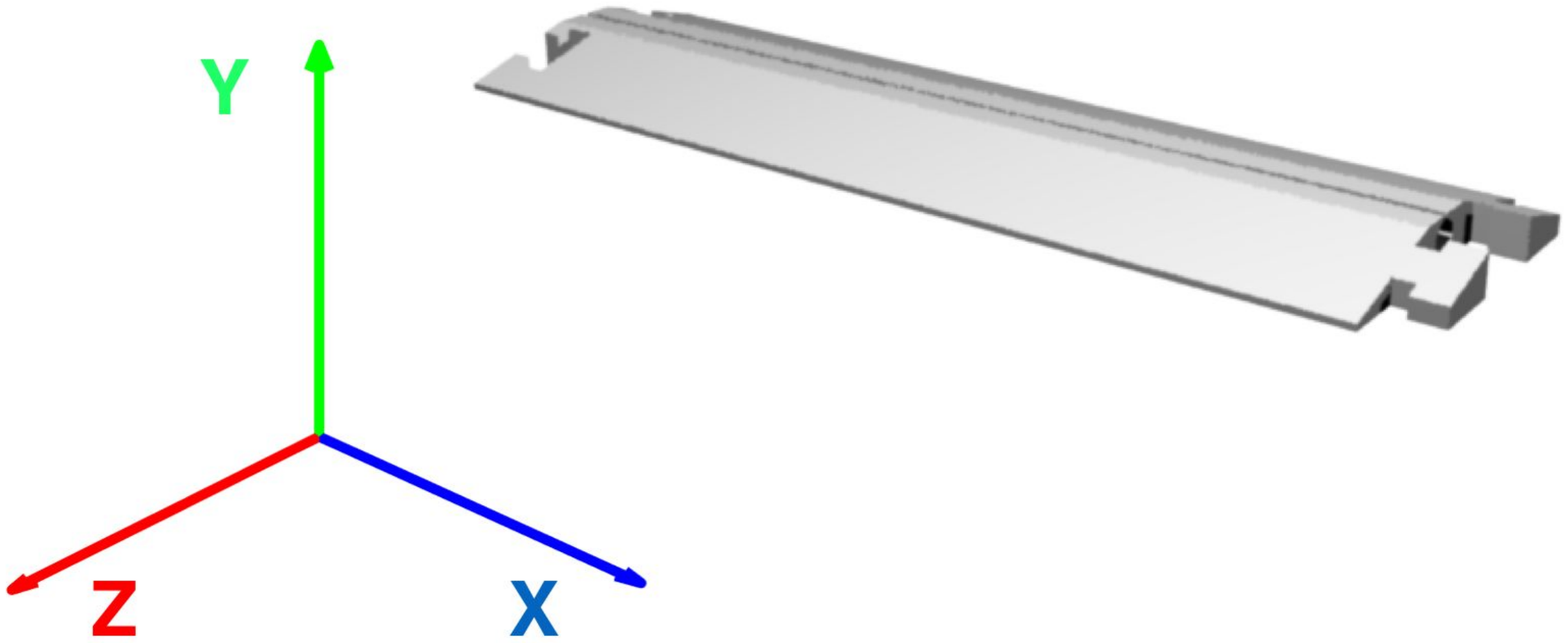} \\
		{\scriptsize $X \sim 1, Y \sim 1, Z \sim 1.$ } & \hspace{-0.5cm}{\scriptsize $X \sim 1, Y \sim 1, Z \sim 1.$} & \hspace{-1cm}{\scriptsize $X \sim 1, Y \sim 1, Z \sim 1.$} & \hspace{-1cm}{\scriptsize $X \sim 1, Y \sim \infty, Z \sim 1.$}   \\
		\scriptsize (v) & \scriptsize (vi) & \scriptsize (vii) & \scriptsize (viii) \\
		\includegraphics[trim={0 0 0 0}, clip = true,width=3.3cm]{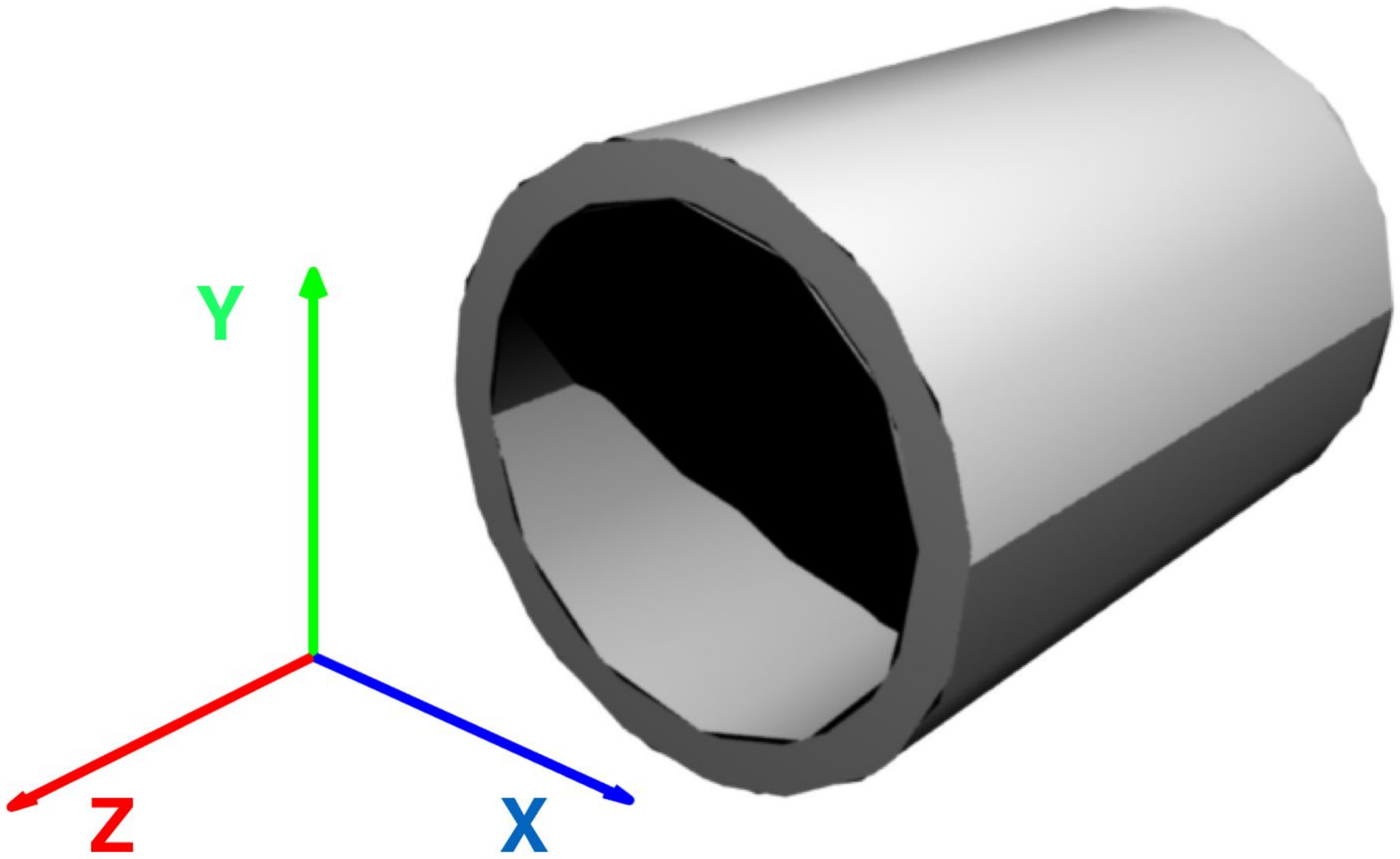} & 
		\includegraphics[trim={0 0 0 0}, clip = true,width=3.3cm]{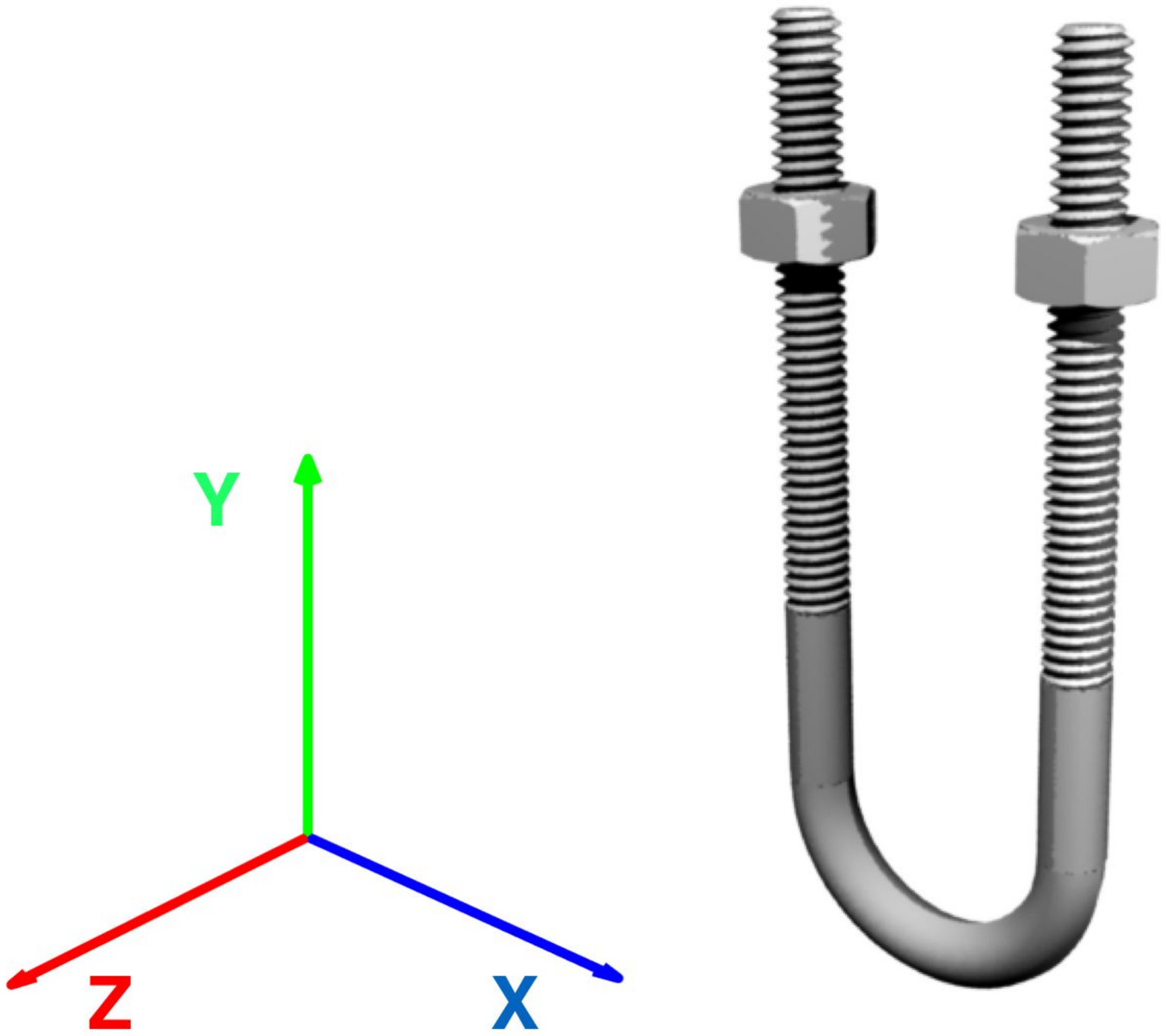} &	
		\includegraphics[trim={0 0 0 0}, clip = true,width=3.3cm]{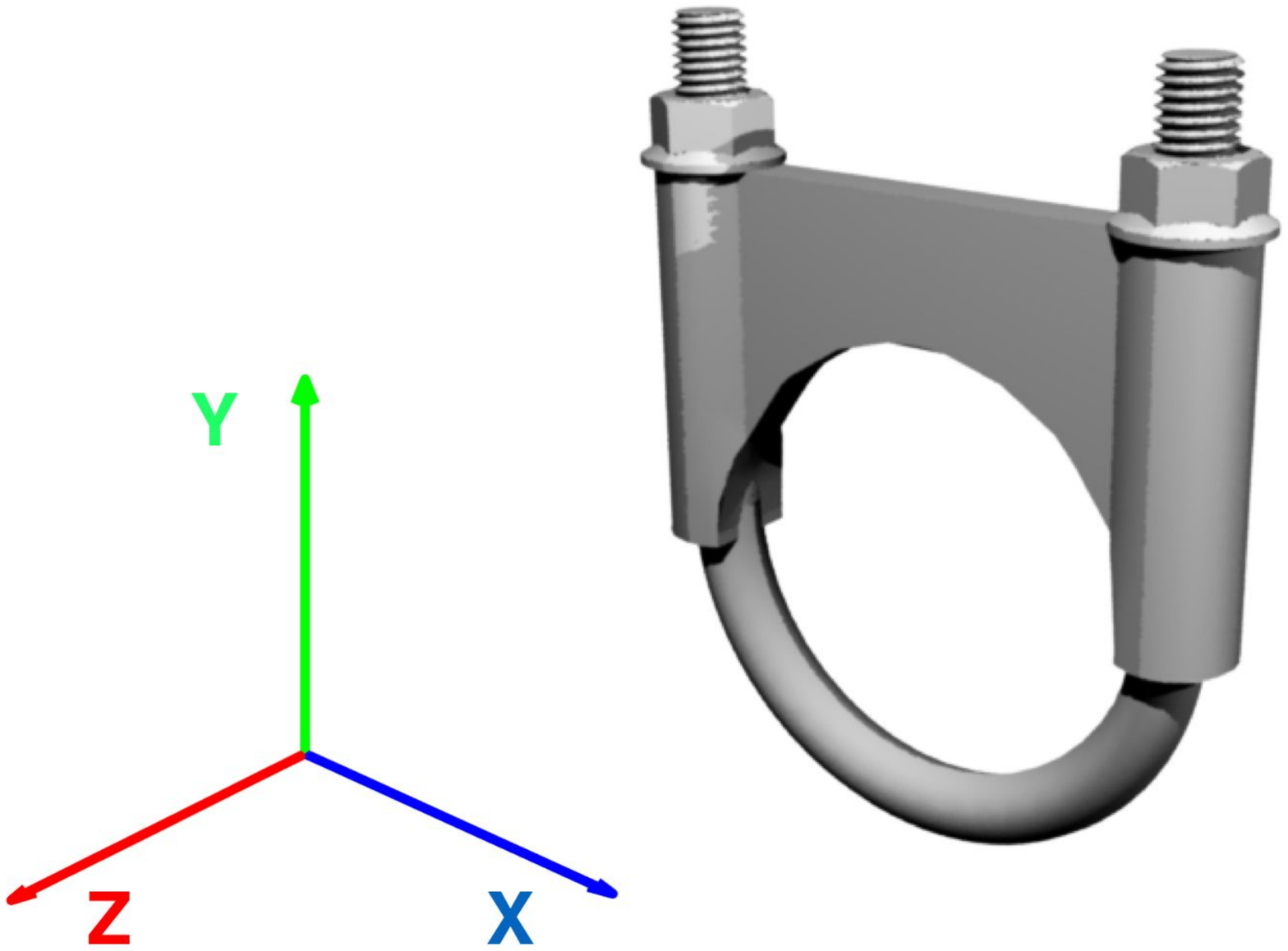} & 
		\includegraphics[trim={0 0 0 0}, clip = true,width=3.3cm]{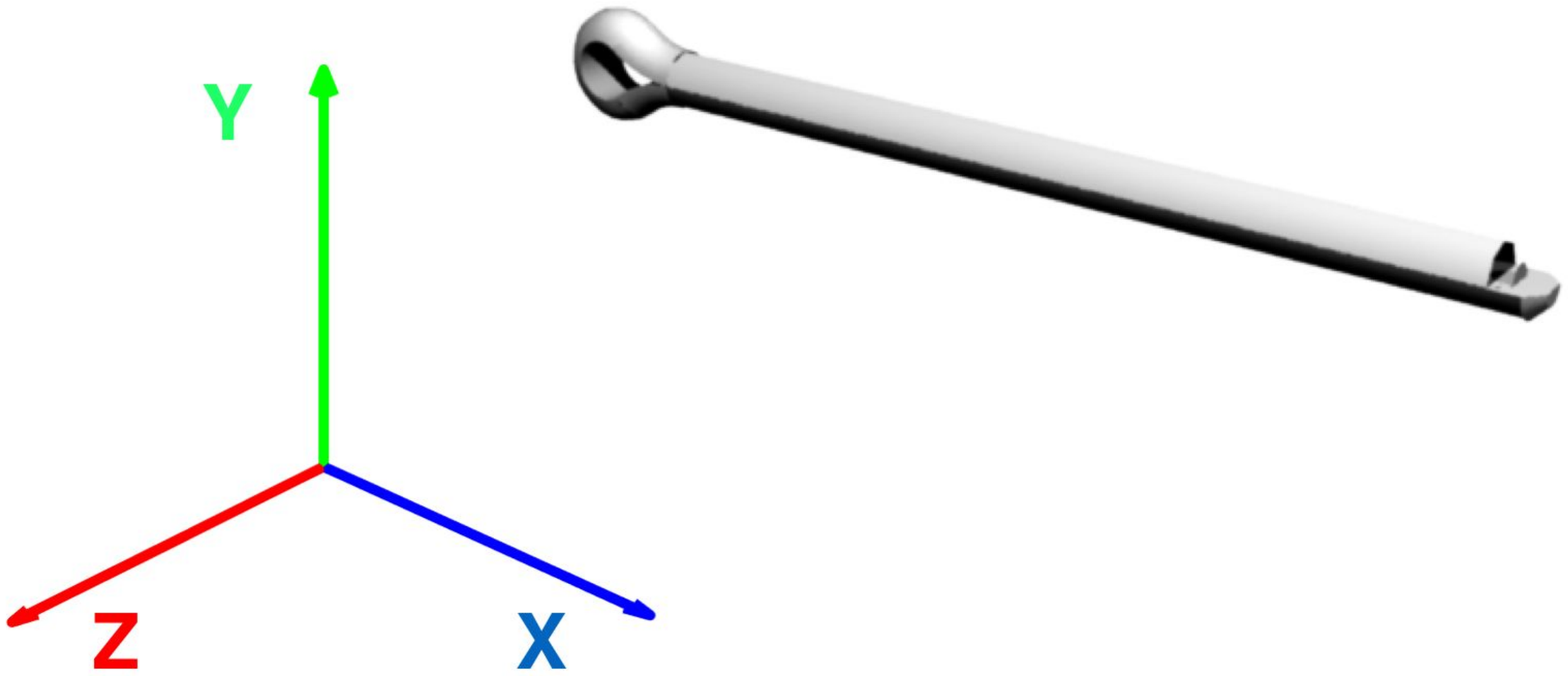}  \\
		{\scriptsize $X \sim 2, Y \sim 2, Z \sim \infty.$ } & \hspace{-0.5cm}{\scriptsize $X \sim 1, Y \sim \infty, Z \sim 1.$} & \hspace{-1cm}{\scriptsize $X \sim 1, Y \sim 1, Z \sim \infty.$} & \hspace{-1cm}{\scriptsize $X \sim \inf, Y \sim 2, Z \sim 2.$}  \\
		\scriptsize (ix) & \scriptsize (x) & \scriptsize (xi) & \scriptsize (xii) \\
 		\includegraphics[trim={0 0 0 0}, clip = true,width=3.3cm]{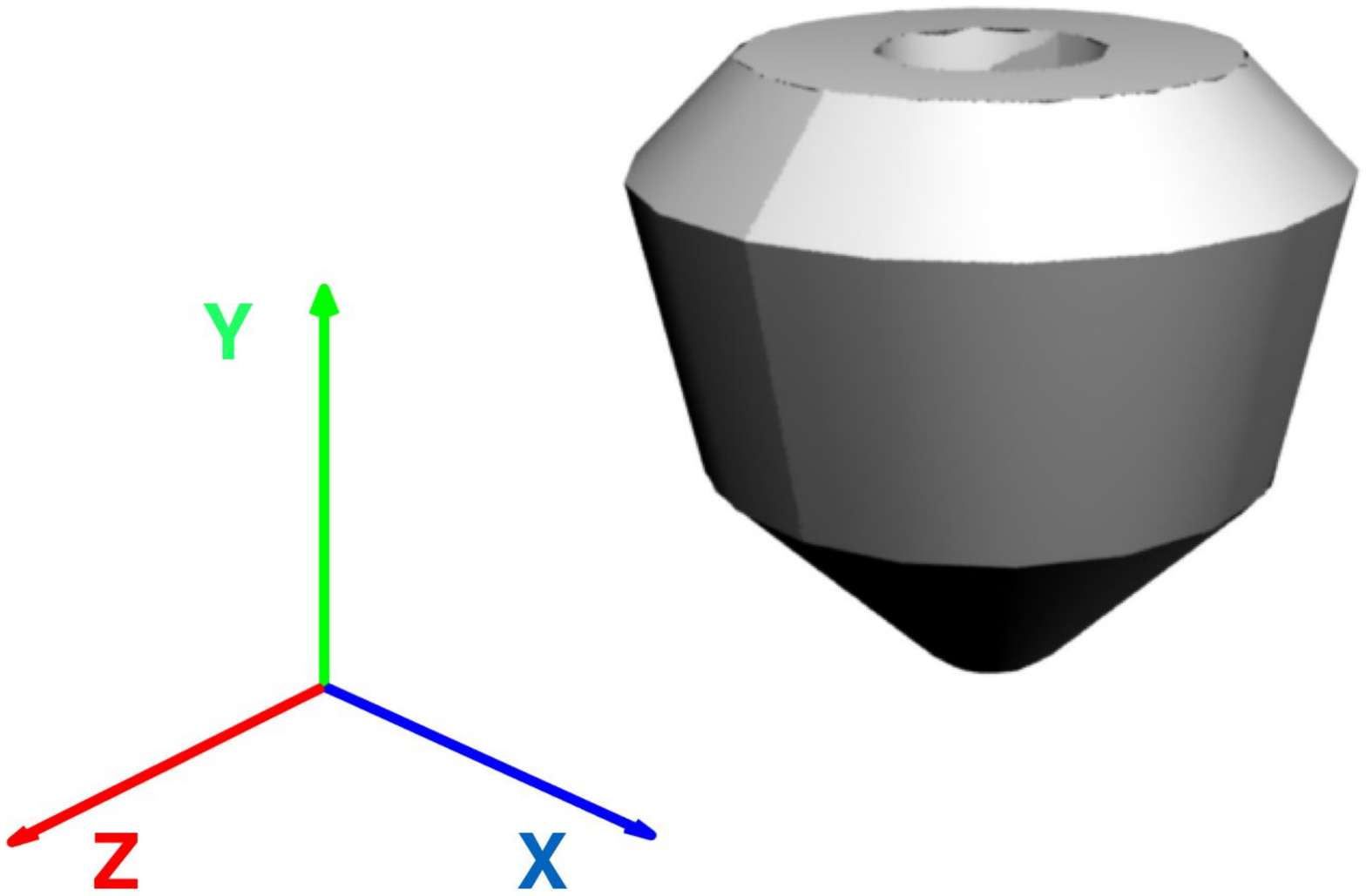} & 
		\includegraphics[trim={0 0 0 0}, clip = true,width=3.3cm]{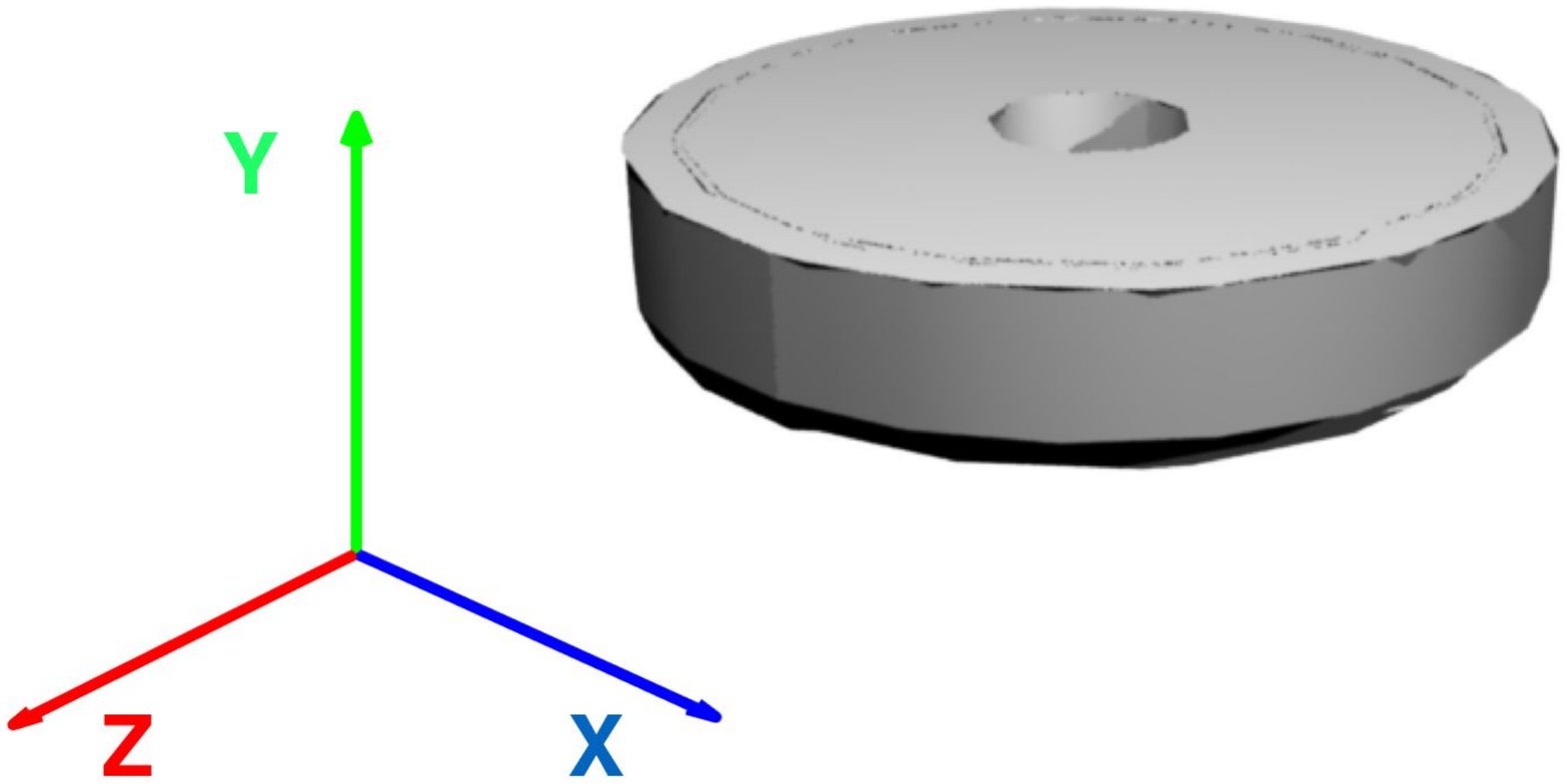} &	
		\includegraphics[trim={0 0 0 0}, clip = true,width=3.3cm]{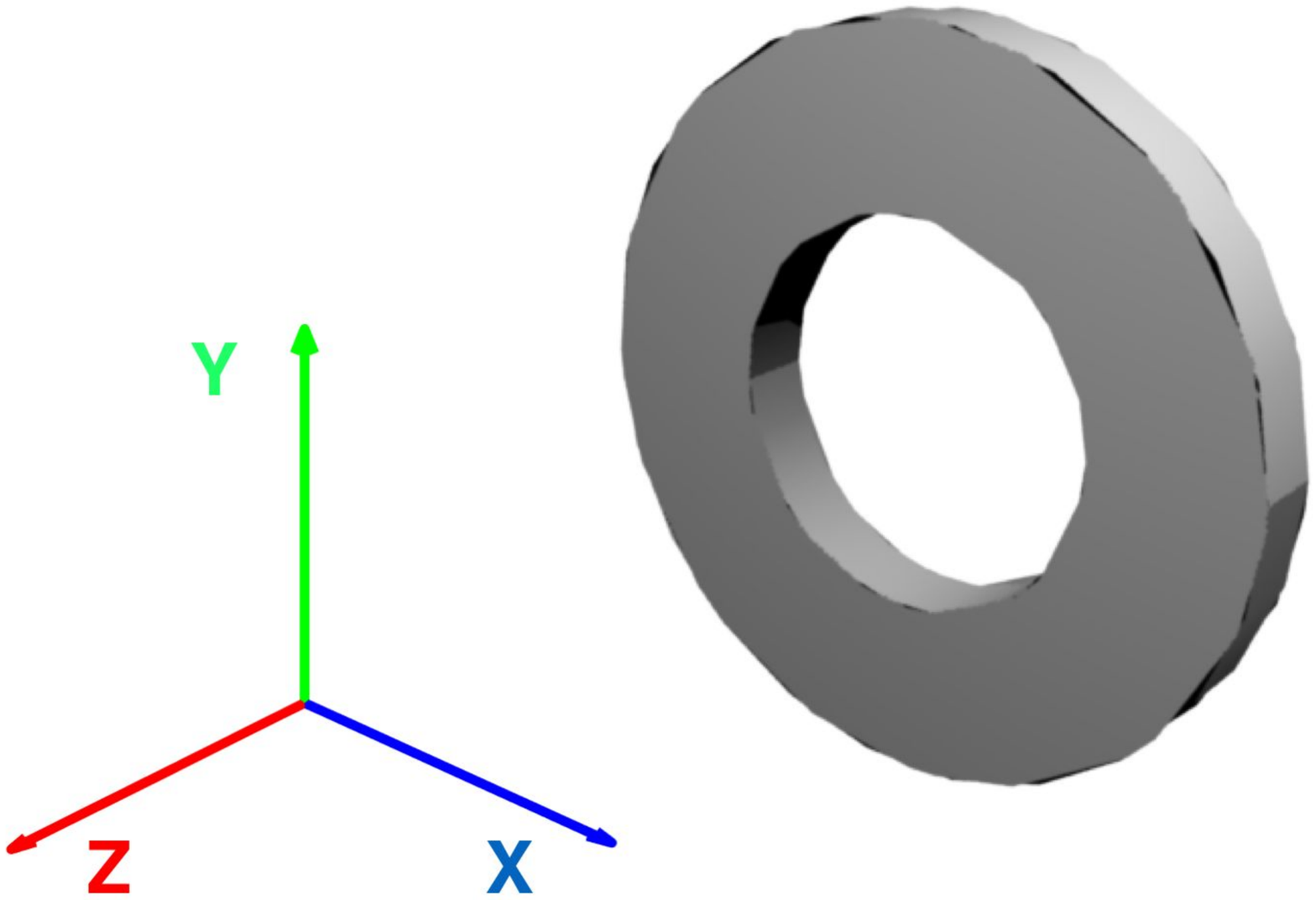} & 
		\includegraphics[trim={0 0 0 0}, clip = true,width=3.3cm]{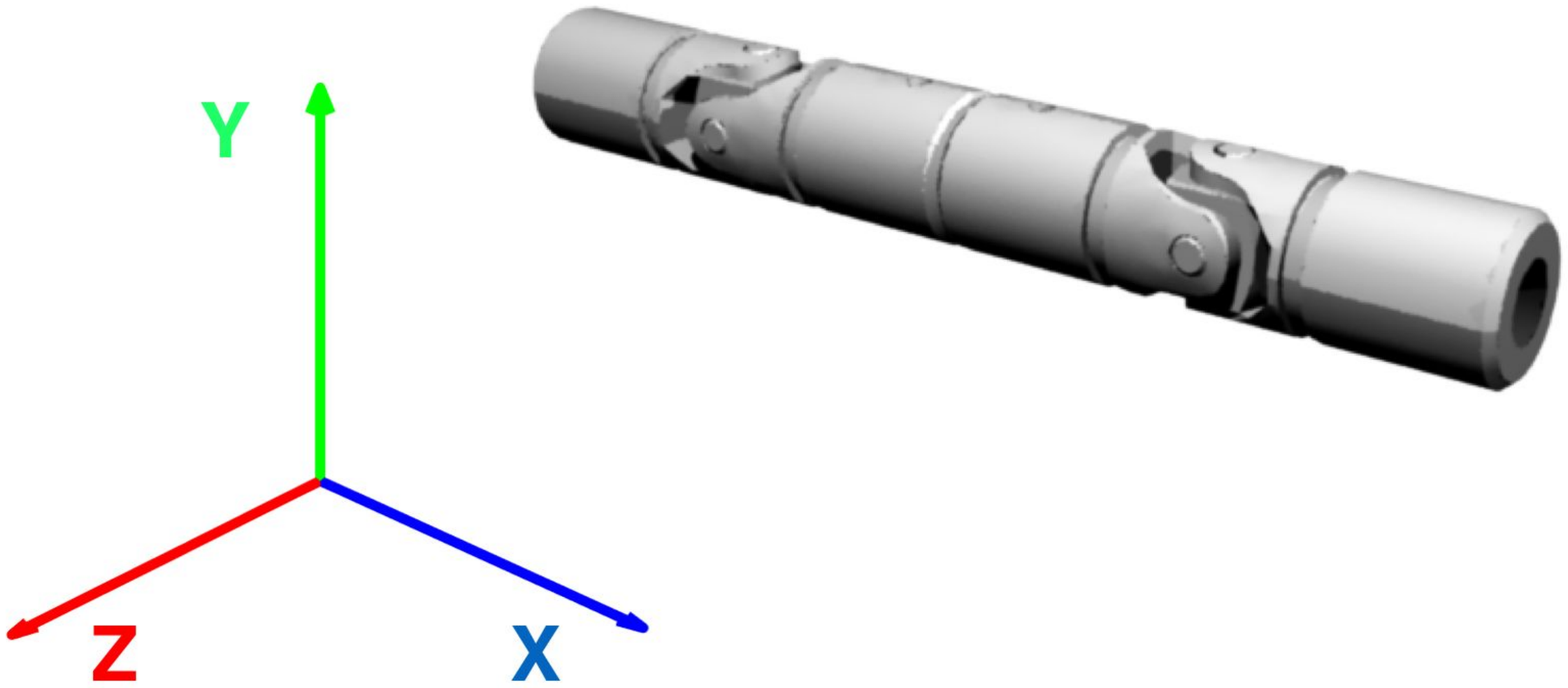}  \\
		{\scriptsize $X \sim 1, Y \sim \infty, Z \sim 1.$ } & \hspace{-0.5cm}{\scriptsize $X \sim 2, Y \sim \infty, Z \sim 2.$} & \hspace{-1cm}{\scriptsize $X \sim 2, Y \sim 2, Z \sim \infty.$} & \hspace{-1cm}{\scriptsize $X \sim \infty, Y \sim 2, Z \sim 2.$} \\
		\scriptsize  (xiii) & \scriptsize  (xiv) & \scriptsize  (xv) & \scriptsize  (xvi) \\
		\includegraphics[trim={0 0 0 0}, clip = true,width=3.3cm]{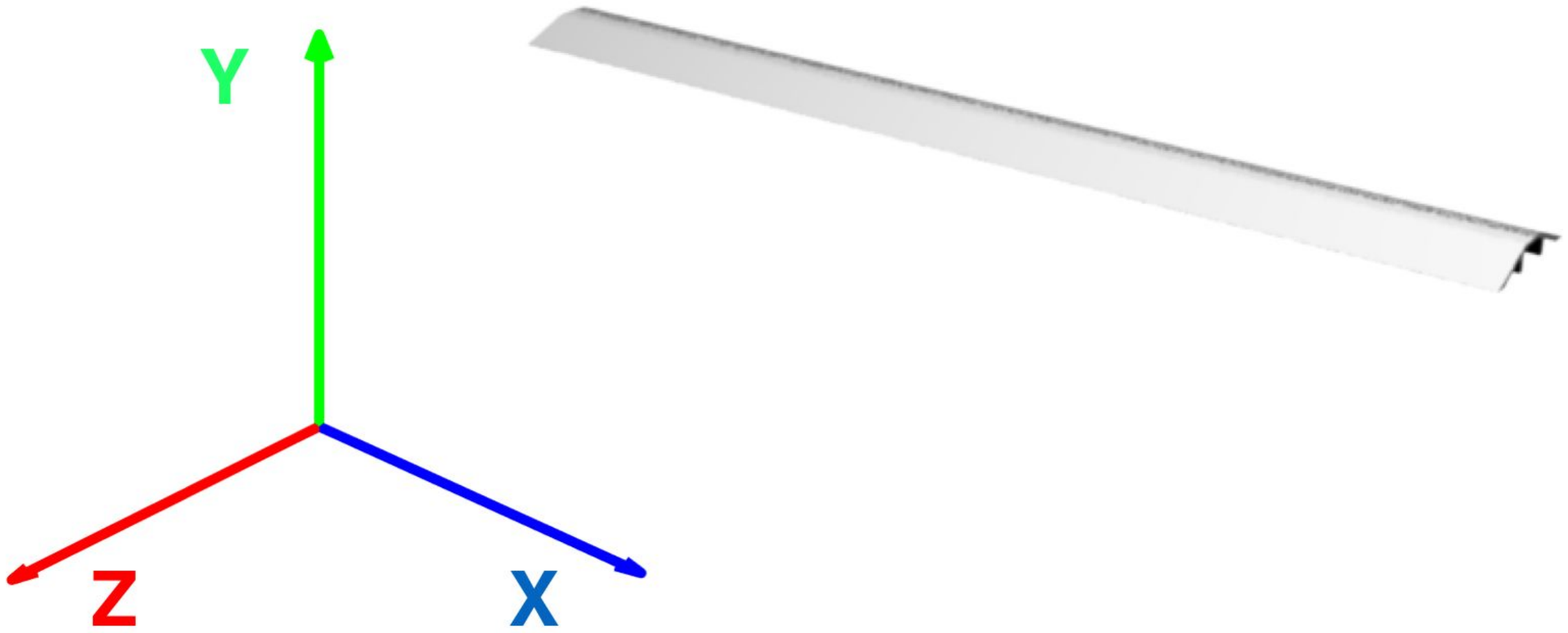} & 
		\includegraphics[trim={0 0 0 0}, clip = true,width=3.3cm]{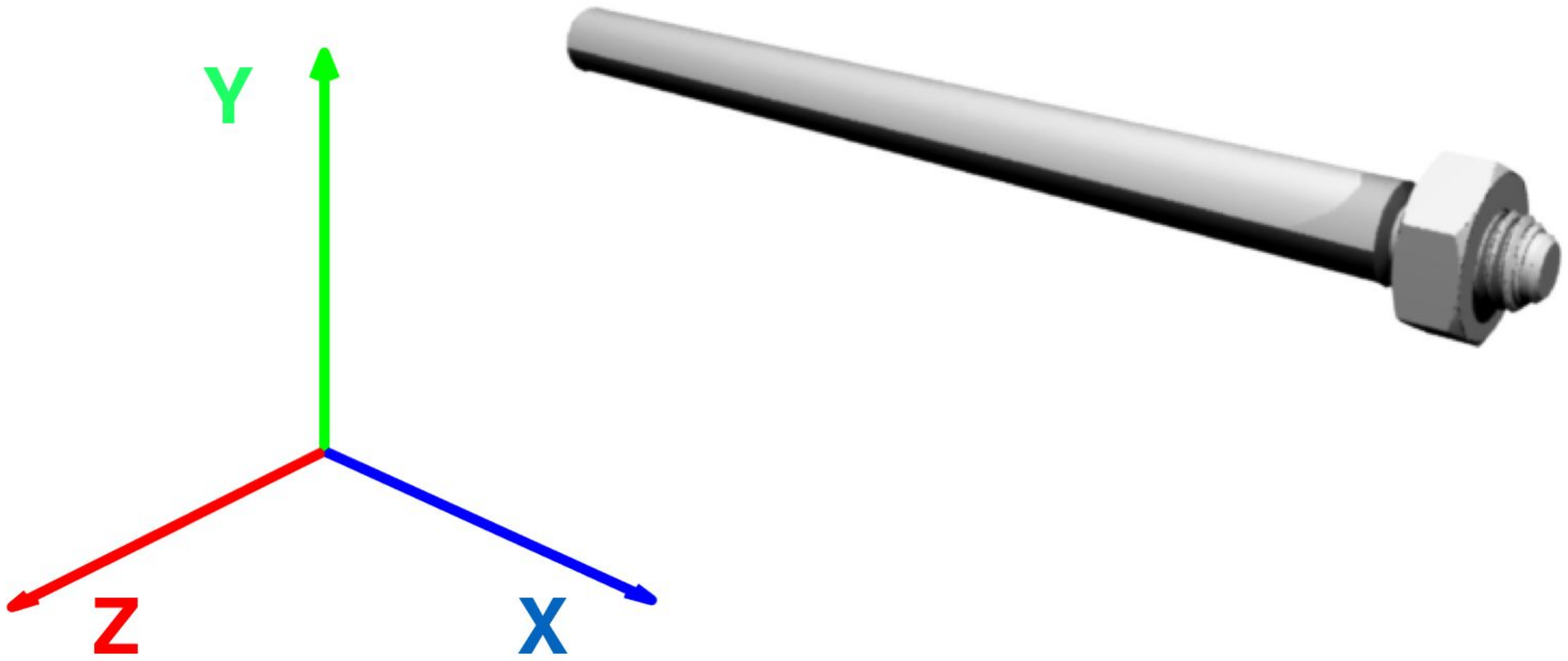} &	
		\includegraphics[trim={0 0 0 0}, clip = true,width=3.3cm]{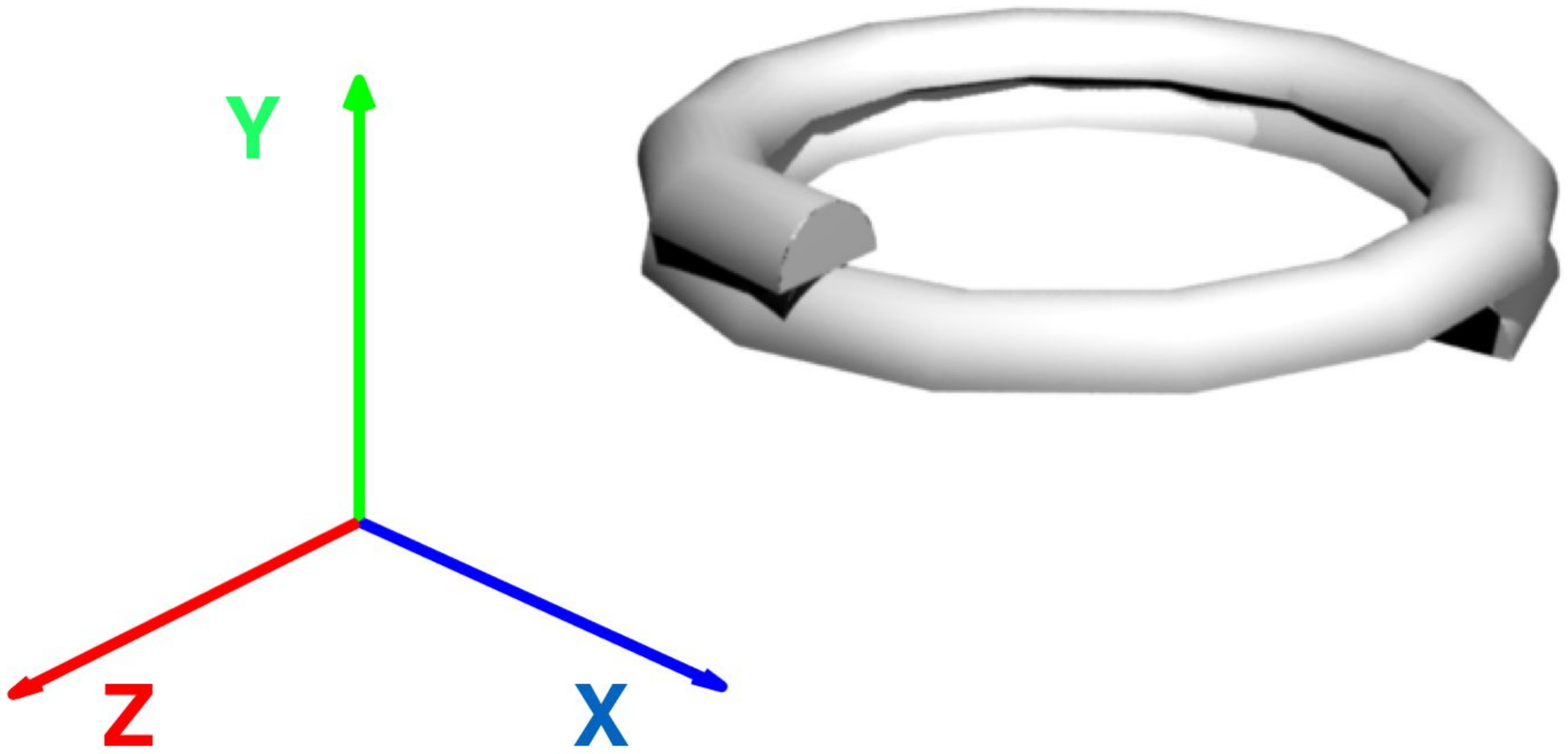} & 
		\includegraphics[trim={0 0 0 0}, clip = true,width=3.3cm]{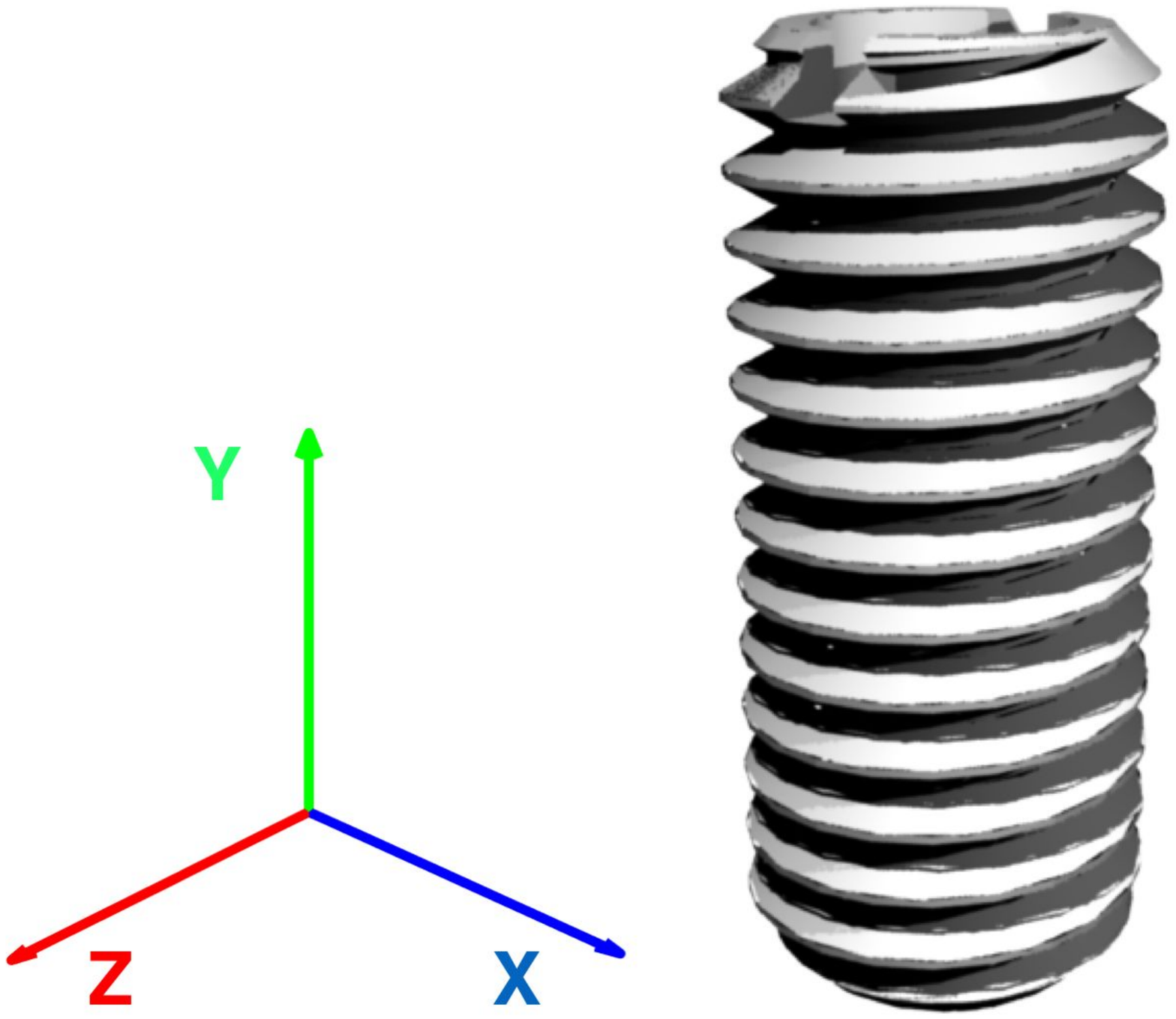} \\
		{\scriptsize $X \sim \infty, Y \sim 2, Z \sim 2.$ } & \hspace{-0.5cm}{\scriptsize $X \sim \infty, Y \sim 2, Z \sim 2.$} & \hspace{-1cm}{\scriptsize $X \sim 1, Y \sim \infty, Z \sim 1.$} & \hspace{-1cm}{\scriptsize $X \sim 2, Y \sim \infty, Z \sim 2.$} \\
		\scriptsize (xvii) & \scriptsize (xviii) &  \scriptsize  (xix) & \scriptsize (xx) \\
		\includegraphics[trim={0 0 0 0}, clip = true,width=3.3cm]{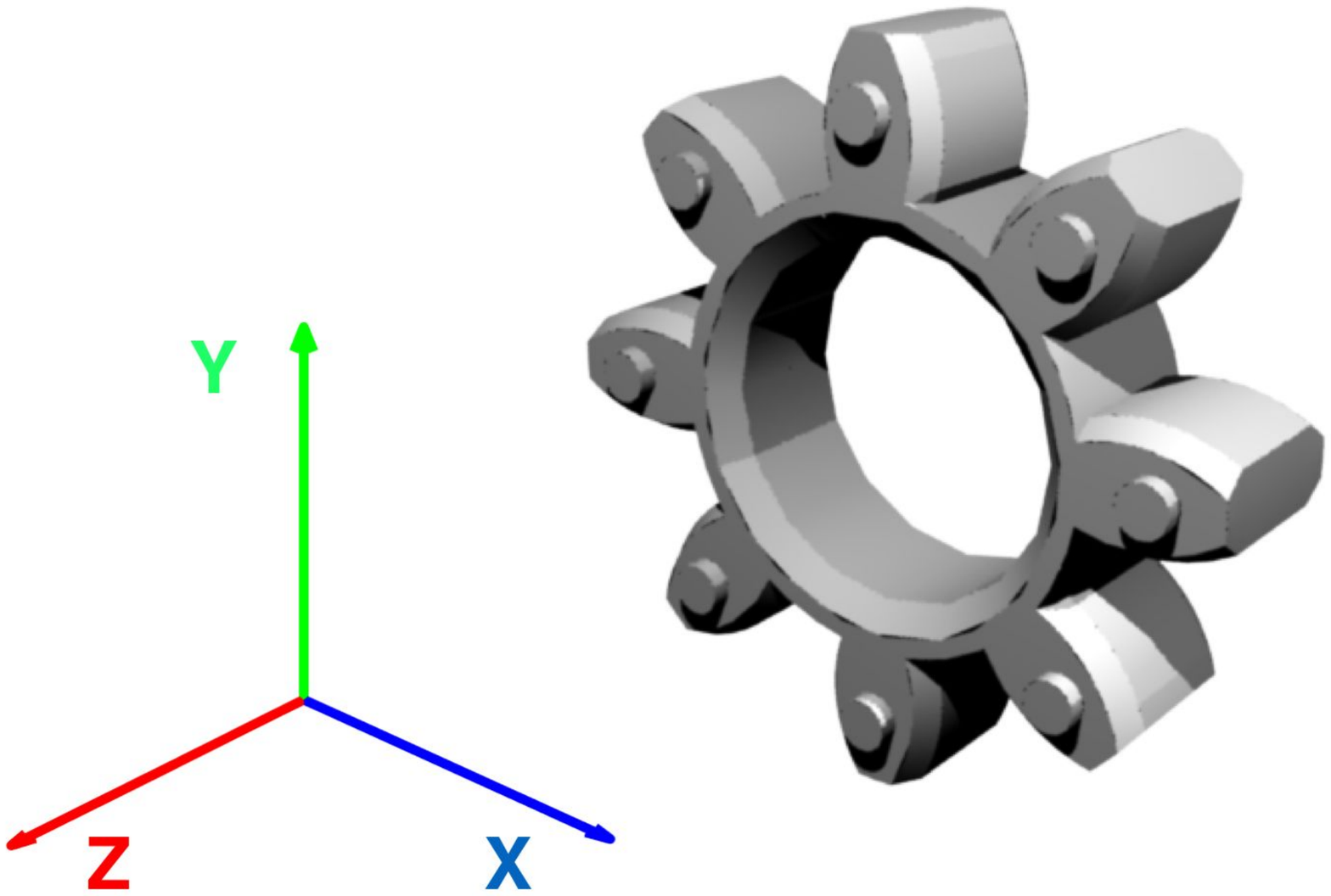} & 
		\includegraphics[trim={0 0 0 0}, clip = true,width=3.3cm]{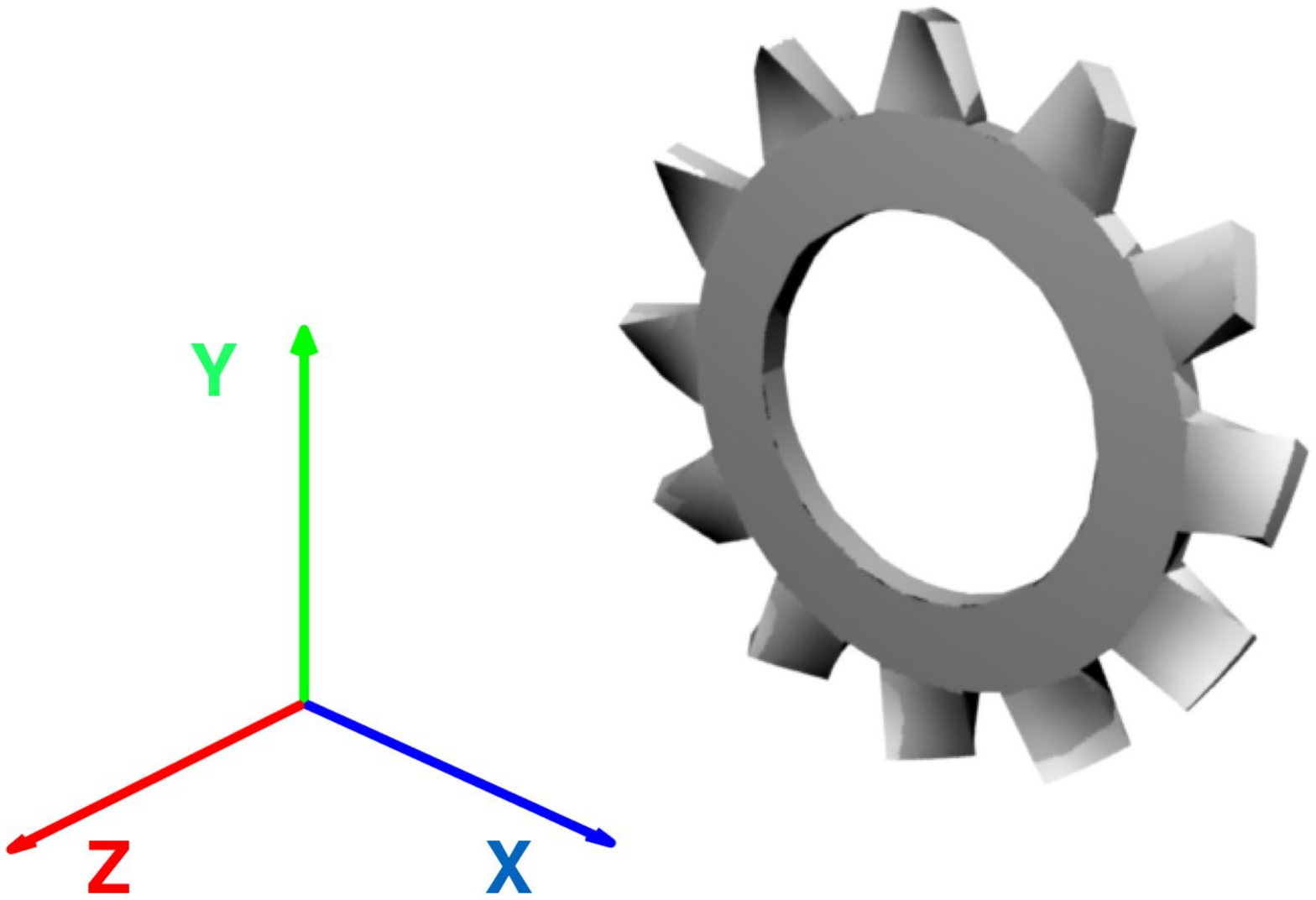} &
		\includegraphics[trim={0 0 0 0}, clip = true,width=3.3cm]{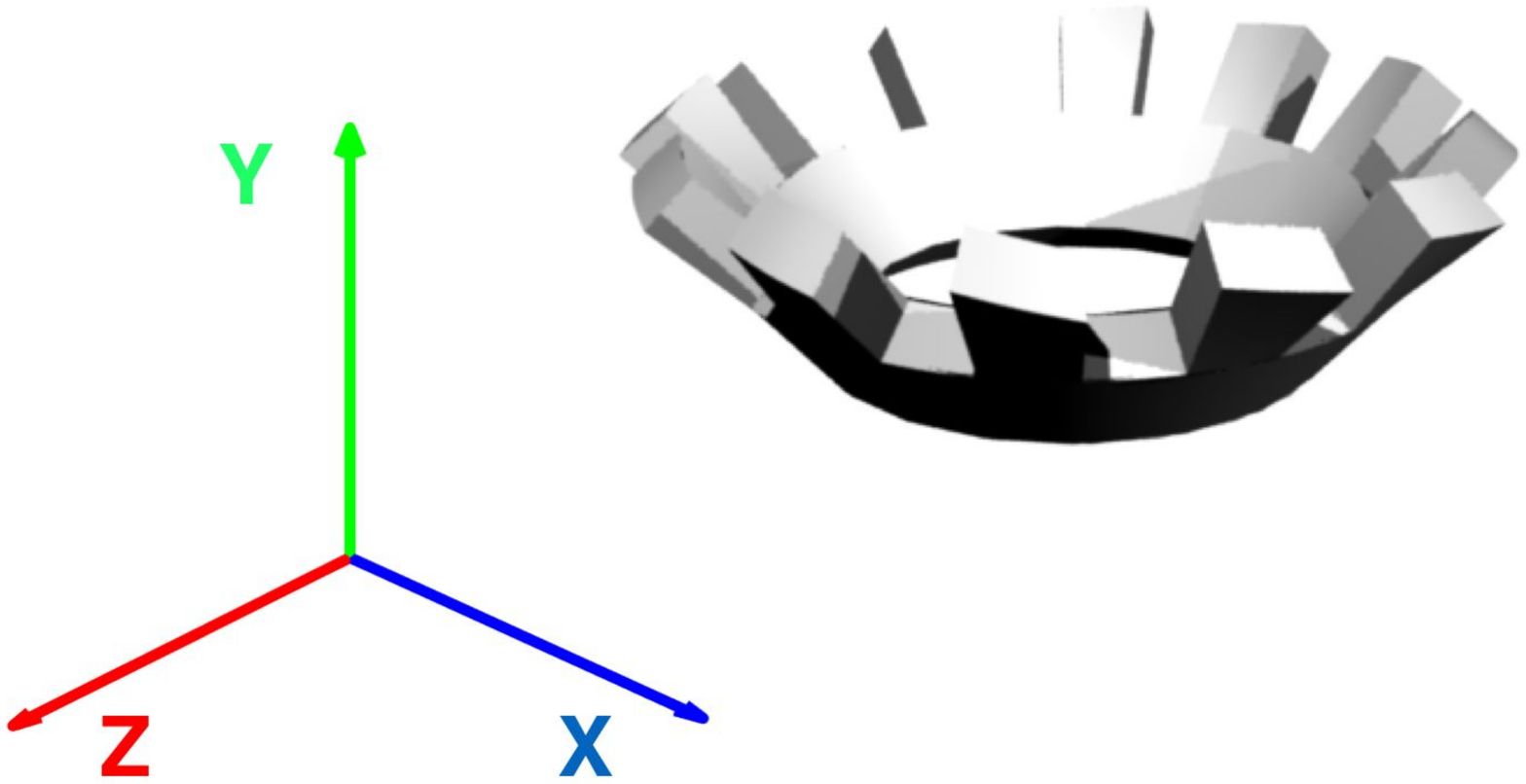} & 
		\includegraphics[trim={0 0 0 0}, clip = true,width=3.3cm]{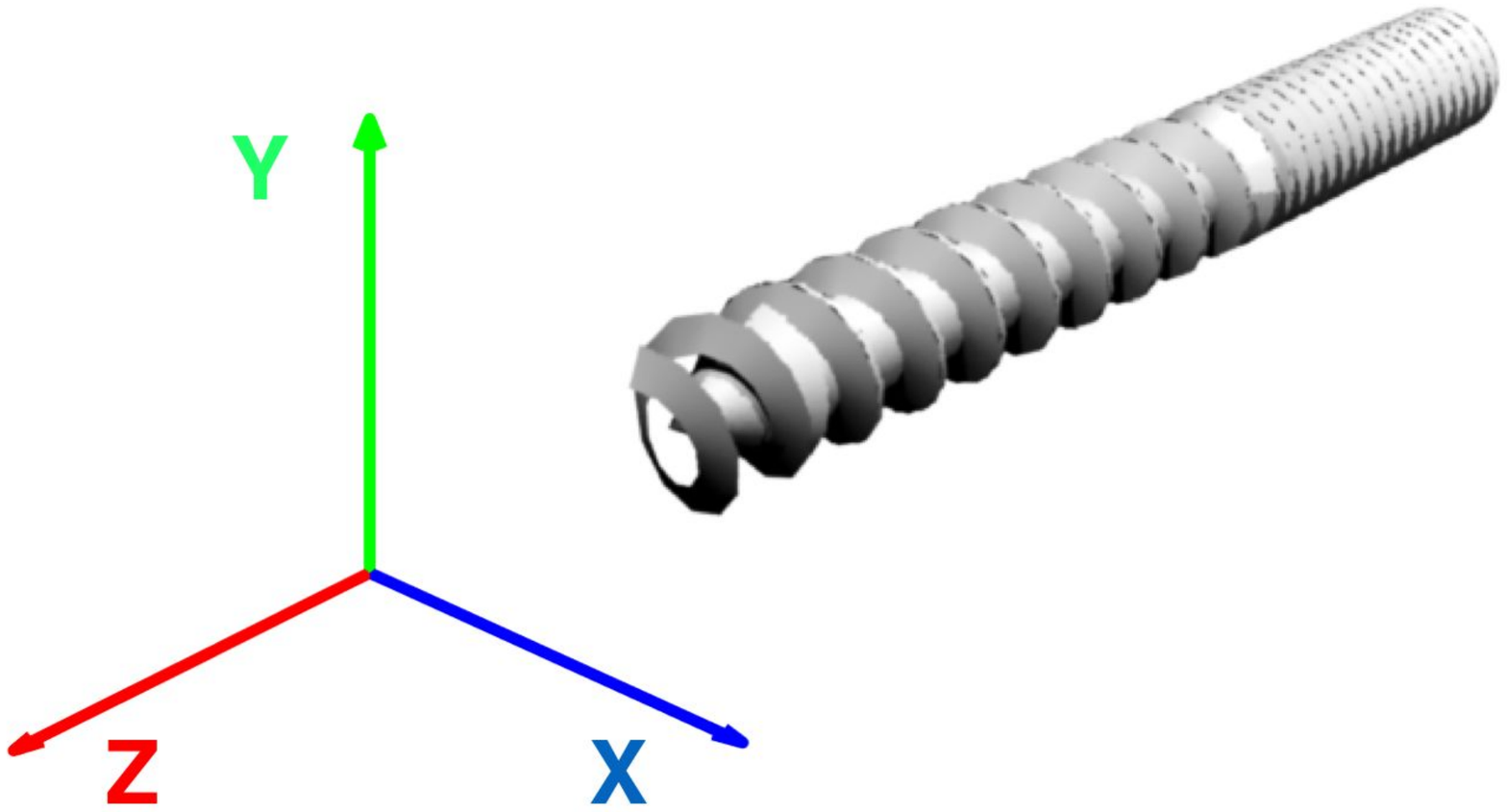} \\
		{\scriptsize $X \sim 2, Y \sim 2, Z \sim \infty.$ } & \hspace{-0.5cm}{\scriptsize $X \sim 2, Y \sim 2, Z \sim \infty.$ } & \hspace{-1cm}{\scriptsize $X \sim 1, Y \sim \infty, Z \sim 1.$} & \hspace{-1cm}{\scriptsize $X \sim 2, Y \sim 2, Z \sim \infty.$} \\
		\scriptsize (xxi) & \scriptsize (xxii) & \scriptsize (xxiii) & \scriptsize (xxiv) \\ 
	\end{tabular}
	}
	\vspace{-4mm}
	\caption{\bf Symmetry Prediction Results}
	\label{fig:symmetries_inferred_1}
	\vspace{-2mm}
\end{figure}

\begin{figure}[t!] % \ContinuedFloat
\vspace{-2mm}
	\centering
	\resizebox{1.06\linewidth}{!}{
	\setlength{\tabcolsep}{1pt}
	\begin{tabular}{cccc}
		\includegraphics[trim={0 0 0 0}, clip = true,width=3.3cm]{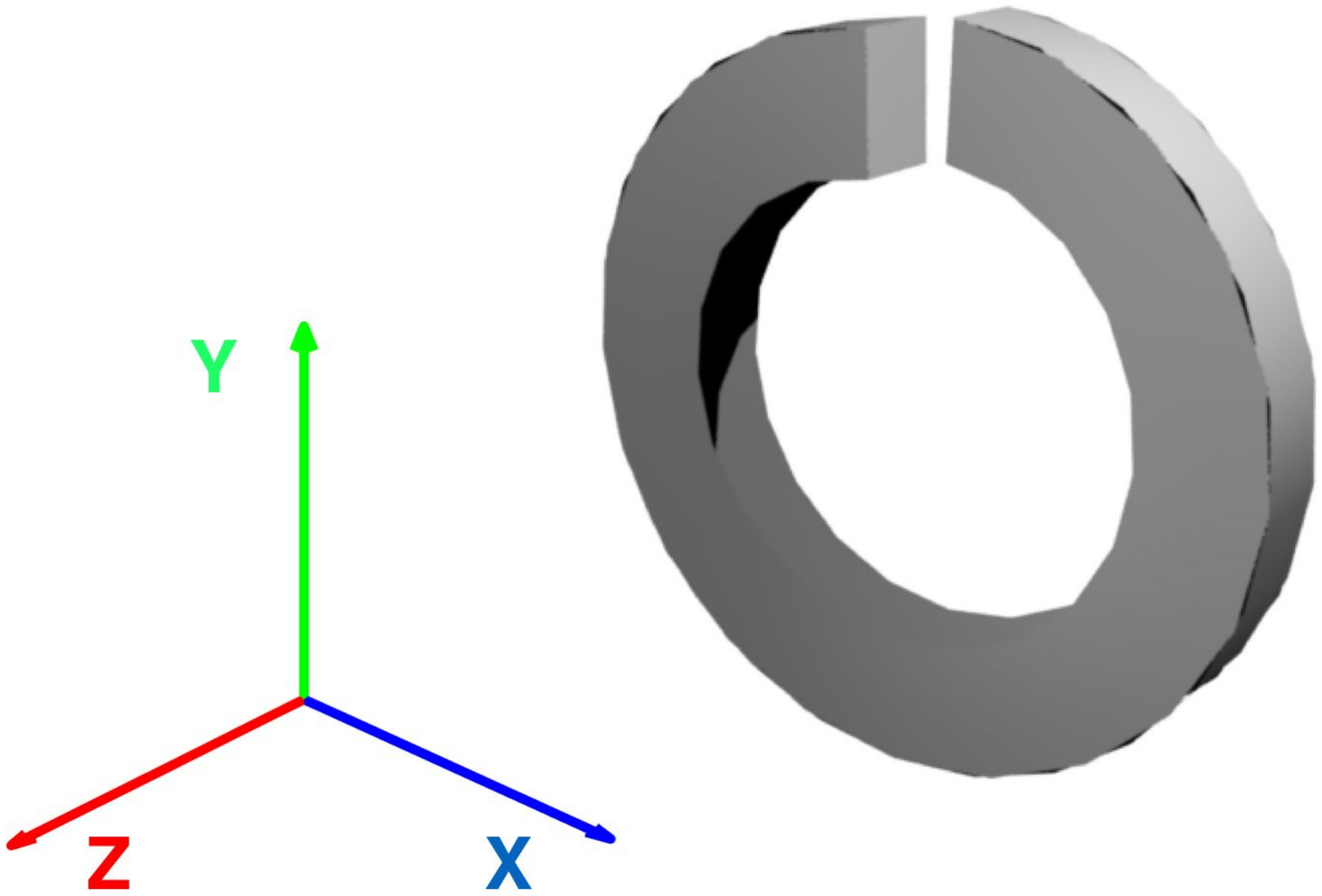} & 
		\includegraphics[trim={0 0 0 0}, clip = true,width=3.3cm]{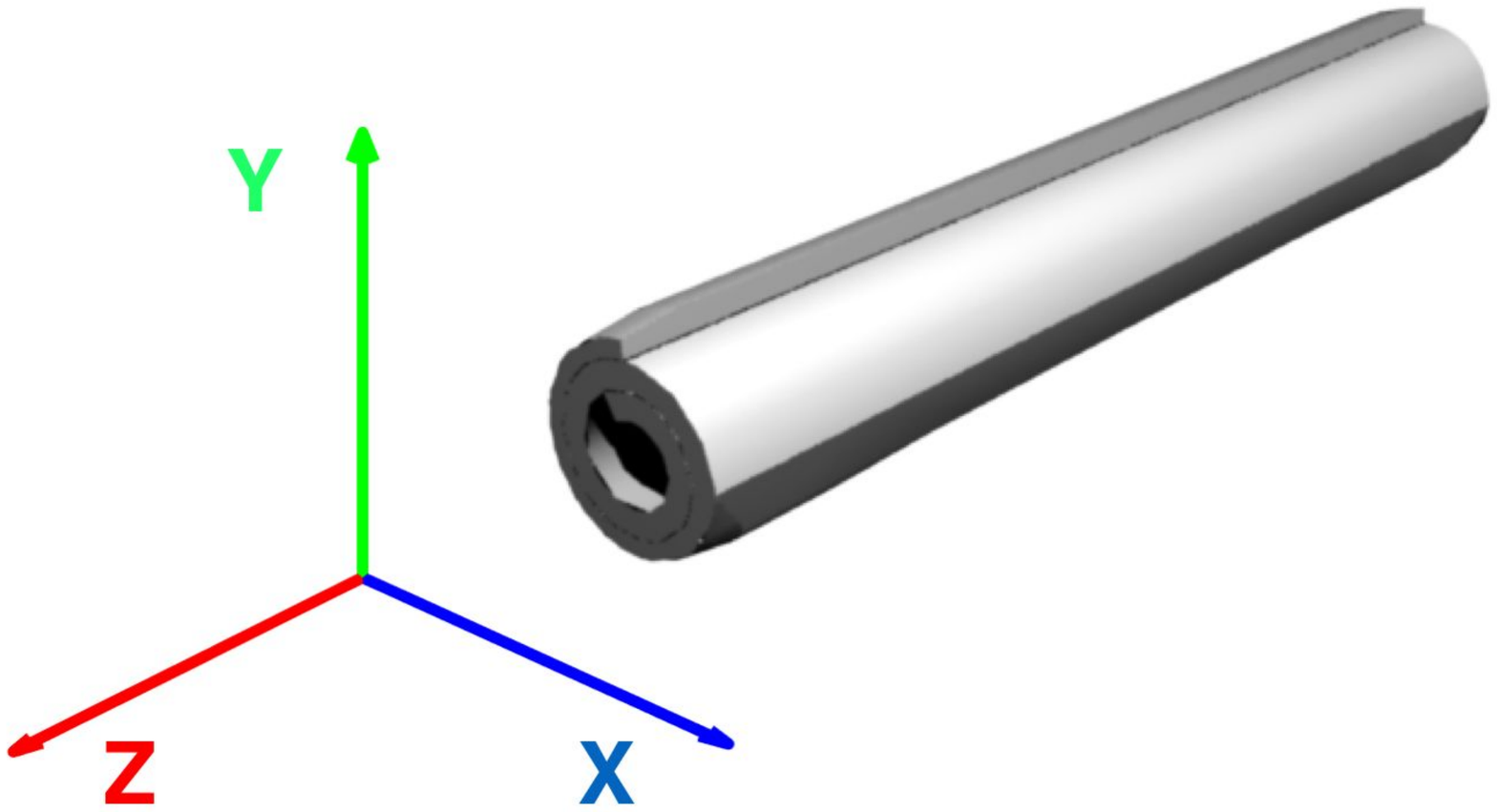} &	
		\includegraphics[trim={0 0 0 0}, clip = true,width=3.3cm]{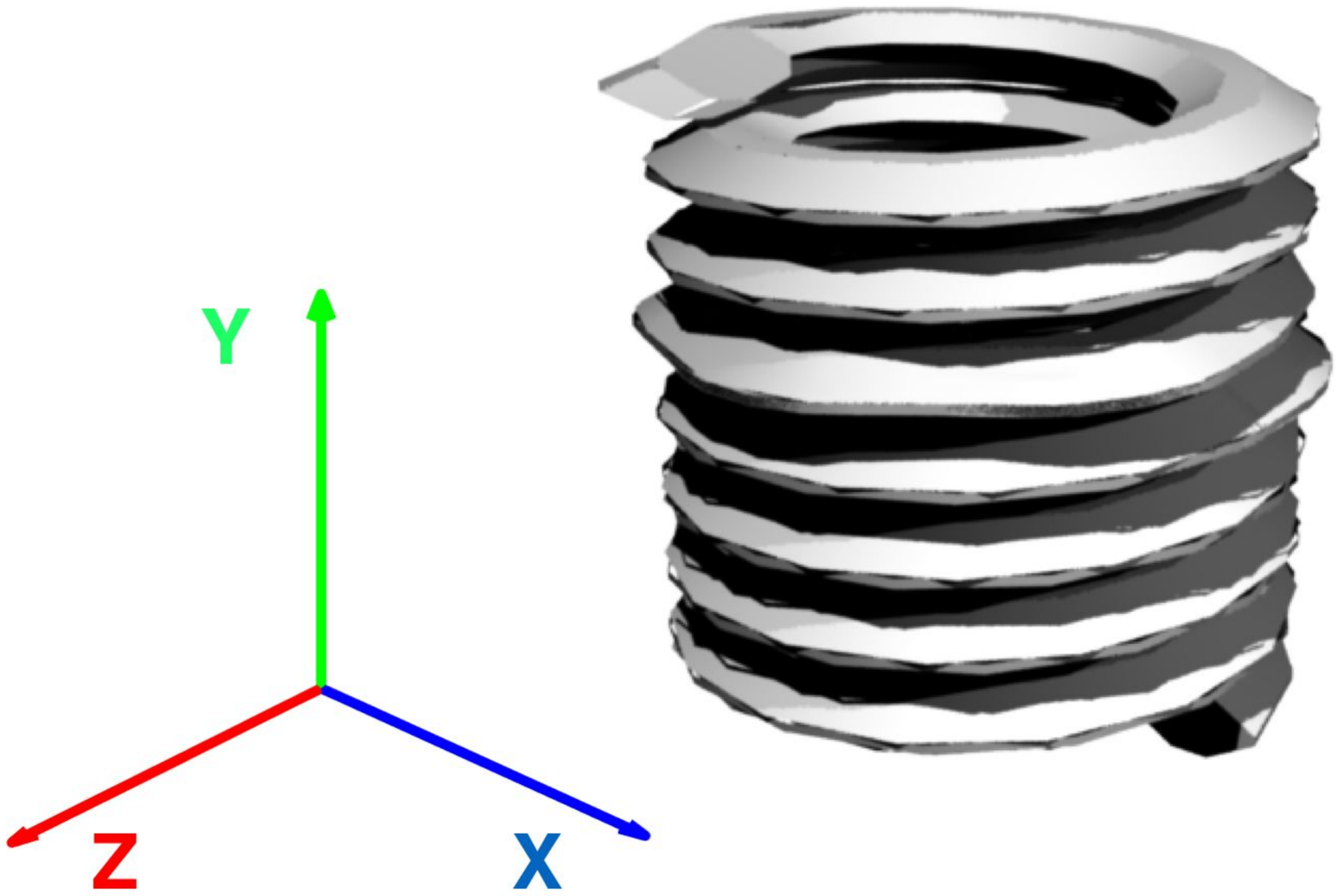} &
		\includegraphics[trim={0 0 0 0}, clip = true,width=3.3cm]{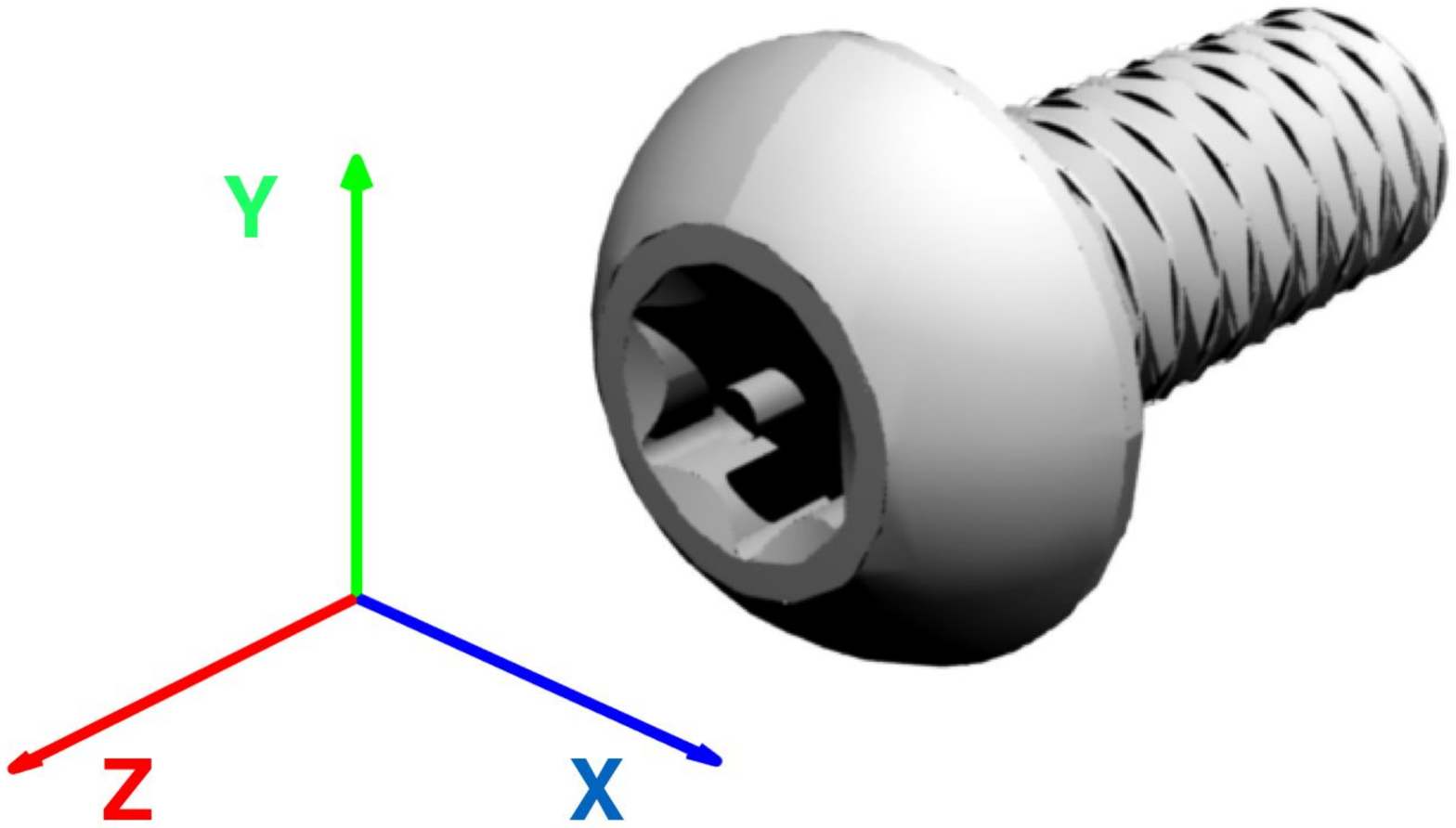} \\
		{\scriptsize $X \sim 1, Y \sim 2, Z \sim 1.$ } & \hspace{-0.5cm}{\scriptsize $X \sim 2, Y \sim 2, Z \sim \infty.$} & \hspace{-1cm}{\scriptsize $X \sim 1, Y \sim 1, Z \sim 1.$}  & \hspace{-1cm}{\scriptsize $X \sim 1, Y \sim 1, Z \sim \infty.$} \\
		{\scriptsize (i)} & {\scriptsize (ii)} & {\scriptsize (iii)} & {\scriptsize (iv) } \\	
		\includegraphics[trim={0 0 0 0}, clip = true,width=3.3cm]{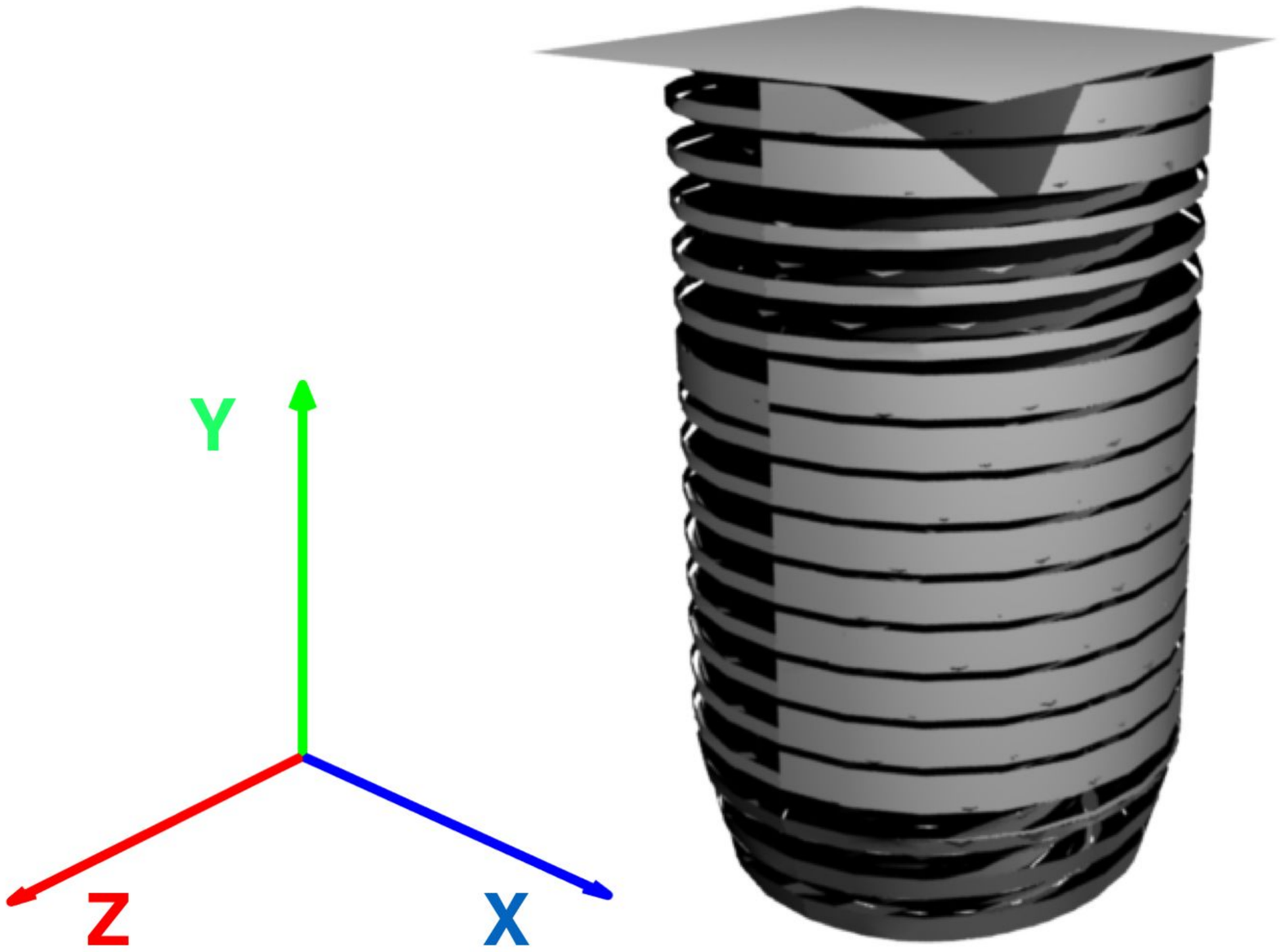} &	
		\includegraphics[trim={0 0 0 0}, clip = true,width=3.3cm]{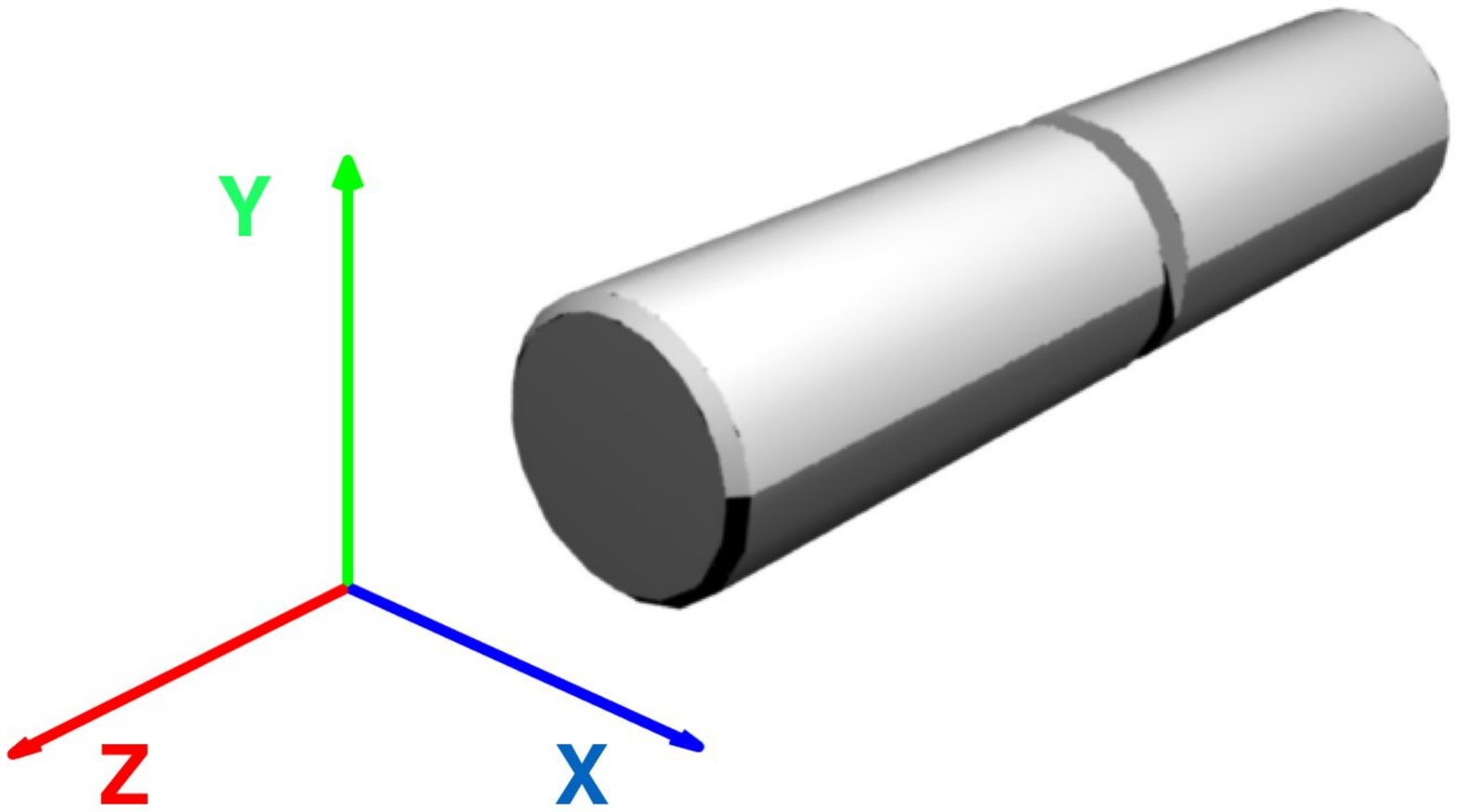} & 
		\includegraphics[trim={0 0 0 0}, clip = true,width=3.3cm]{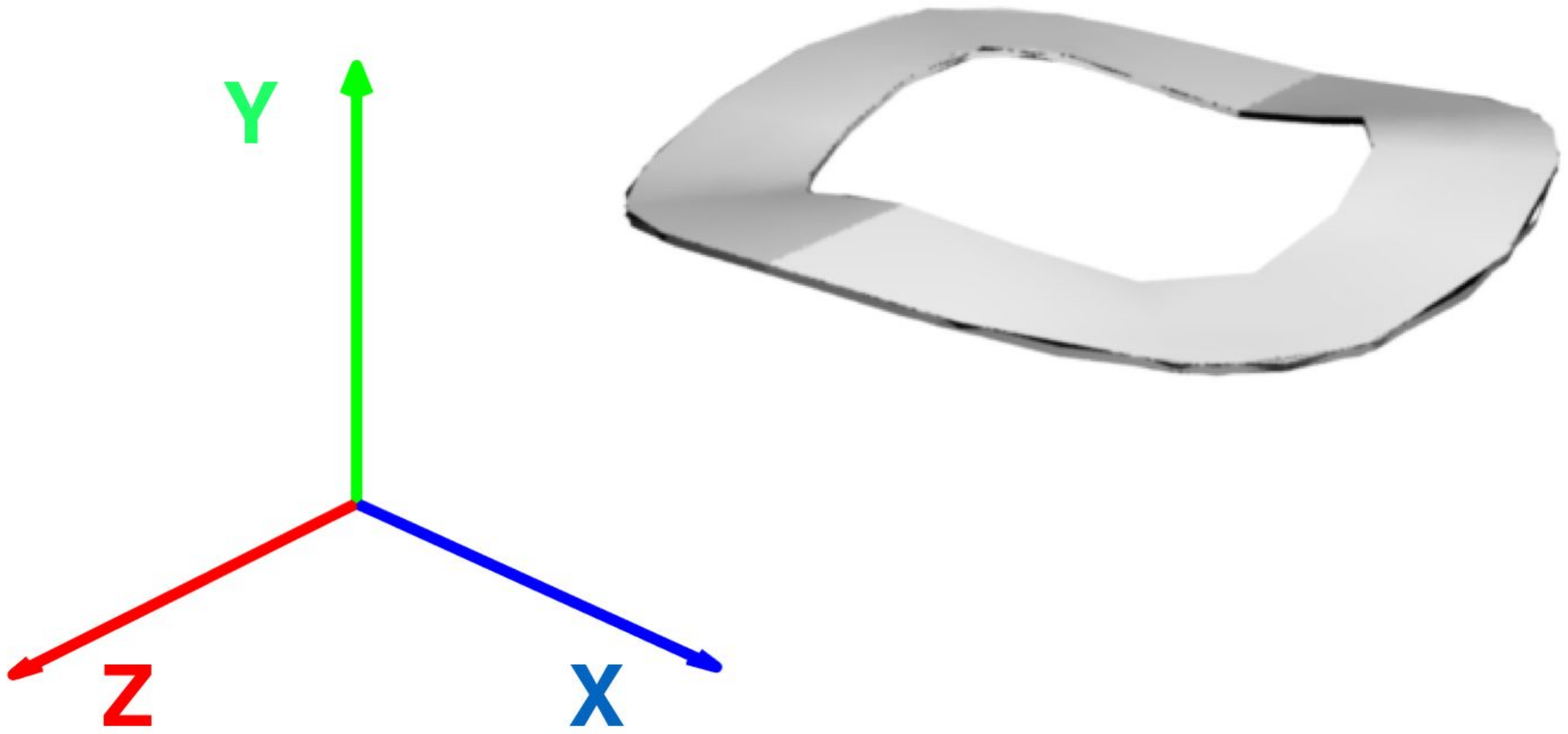} &
		\includegraphics[trim={0 0 0 0}, clip = true,width=3.3cm]{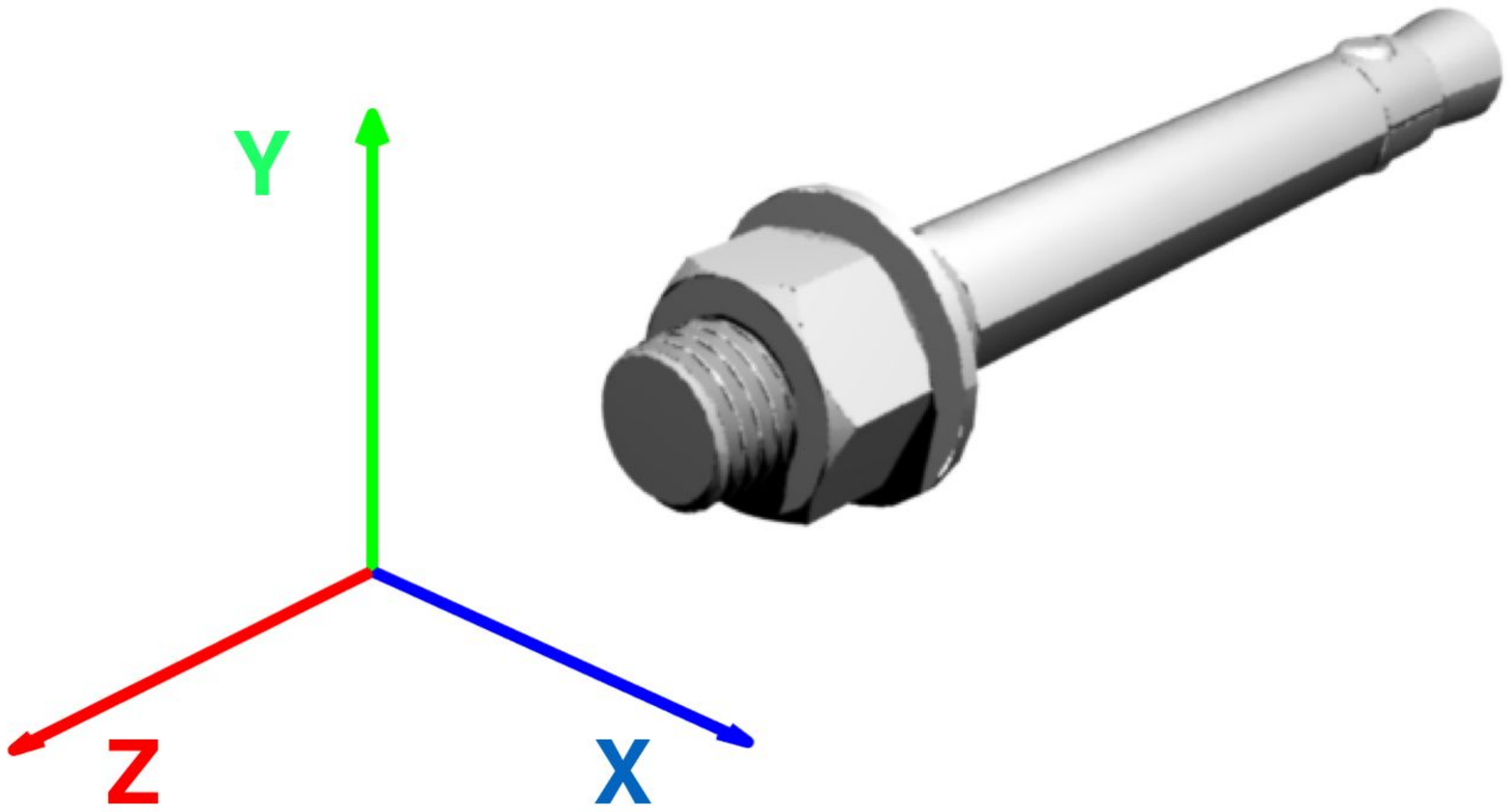} \\
		{\scriptsize $X \sim 1, Y \sim 1, Z \sim 1.$} & \hspace{-1cm}{\scriptsize $X \sim 2, Y \sim 2, Z \sim \infty.$} & \hspace{-1cm}{\scriptsize $X \sim 1, Y \sim \infty, Z \sim 1.$}  & \hspace{-1cm}{\scriptsize $X \sim 2, Y \sim 2, Z \sim \infty.$} \\
		\scriptsize (v) & \scriptsize (vi) & \scriptsize (vii) & \scriptsize (viii) \\
		\includegraphics[trim={0 0 0 0}, clip = true,width=3.3cm]{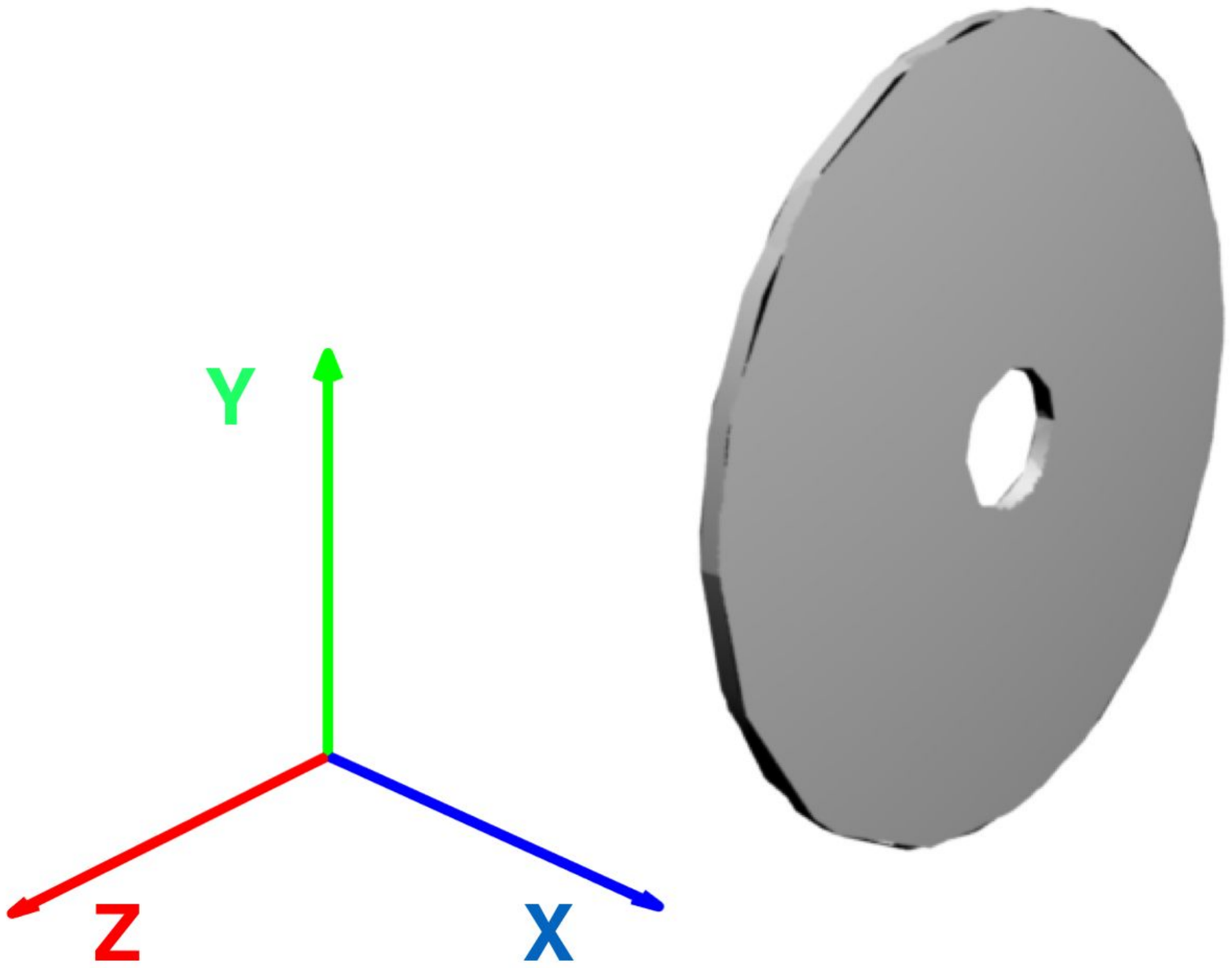} & 
		\includegraphics[trim={0 0 0 0}, clip = true,width=3.3cm]{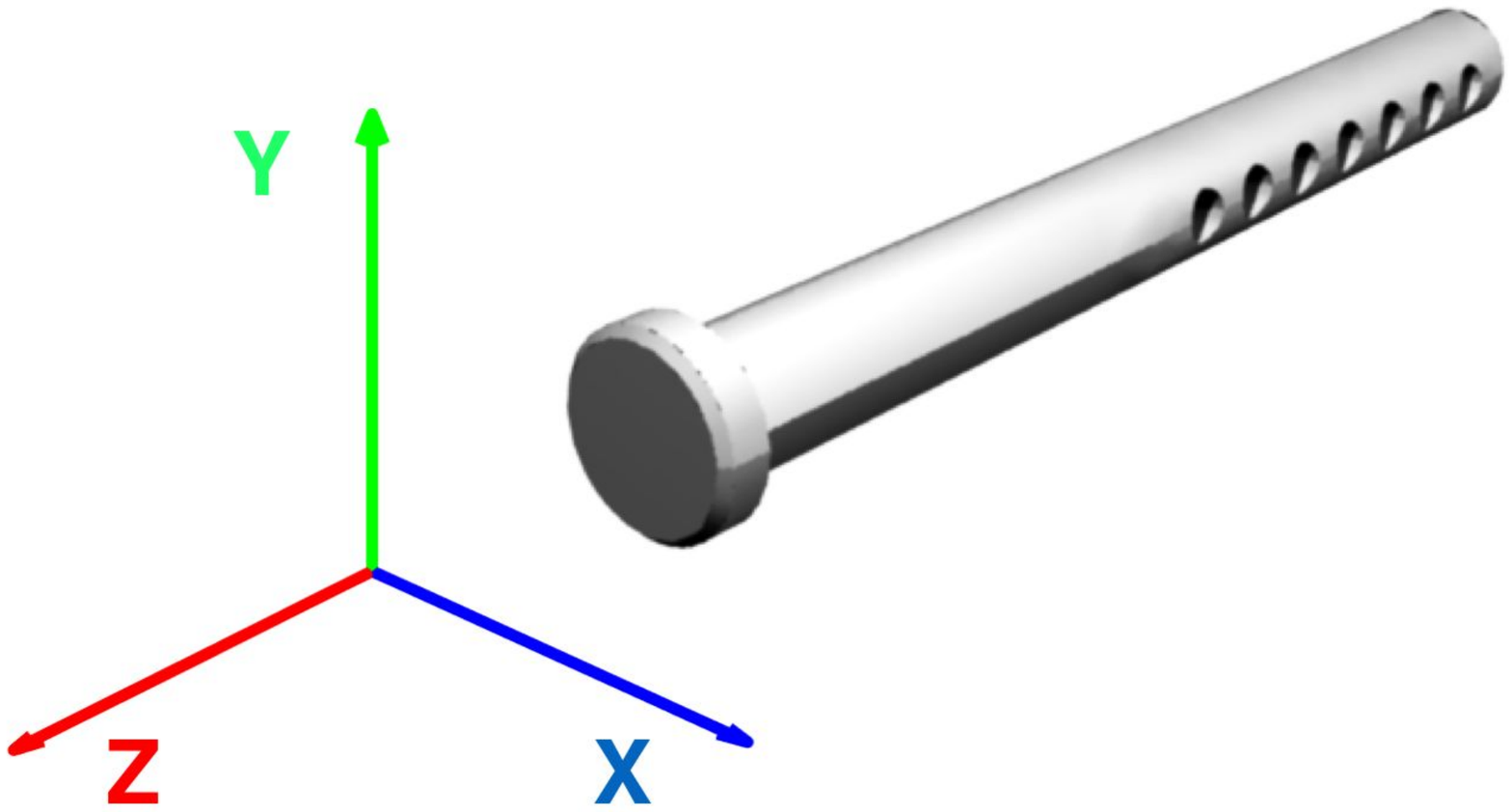} &	
		\includegraphics[trim={0 0 0 0}, clip = true,width=3.3cm]{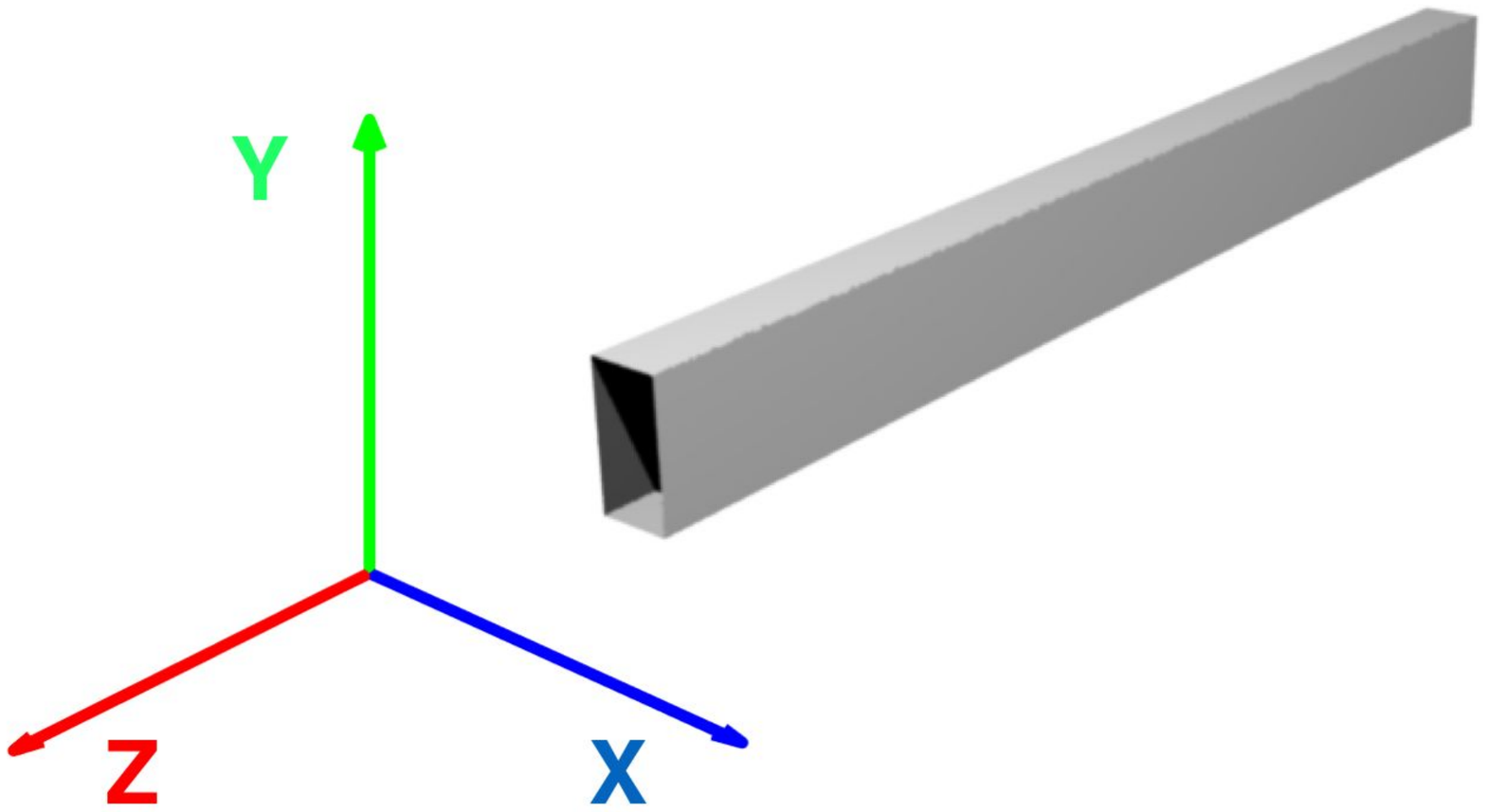} &
		\includegraphics[trim={0 0 0 0}, clip = true,width=3.3cm]{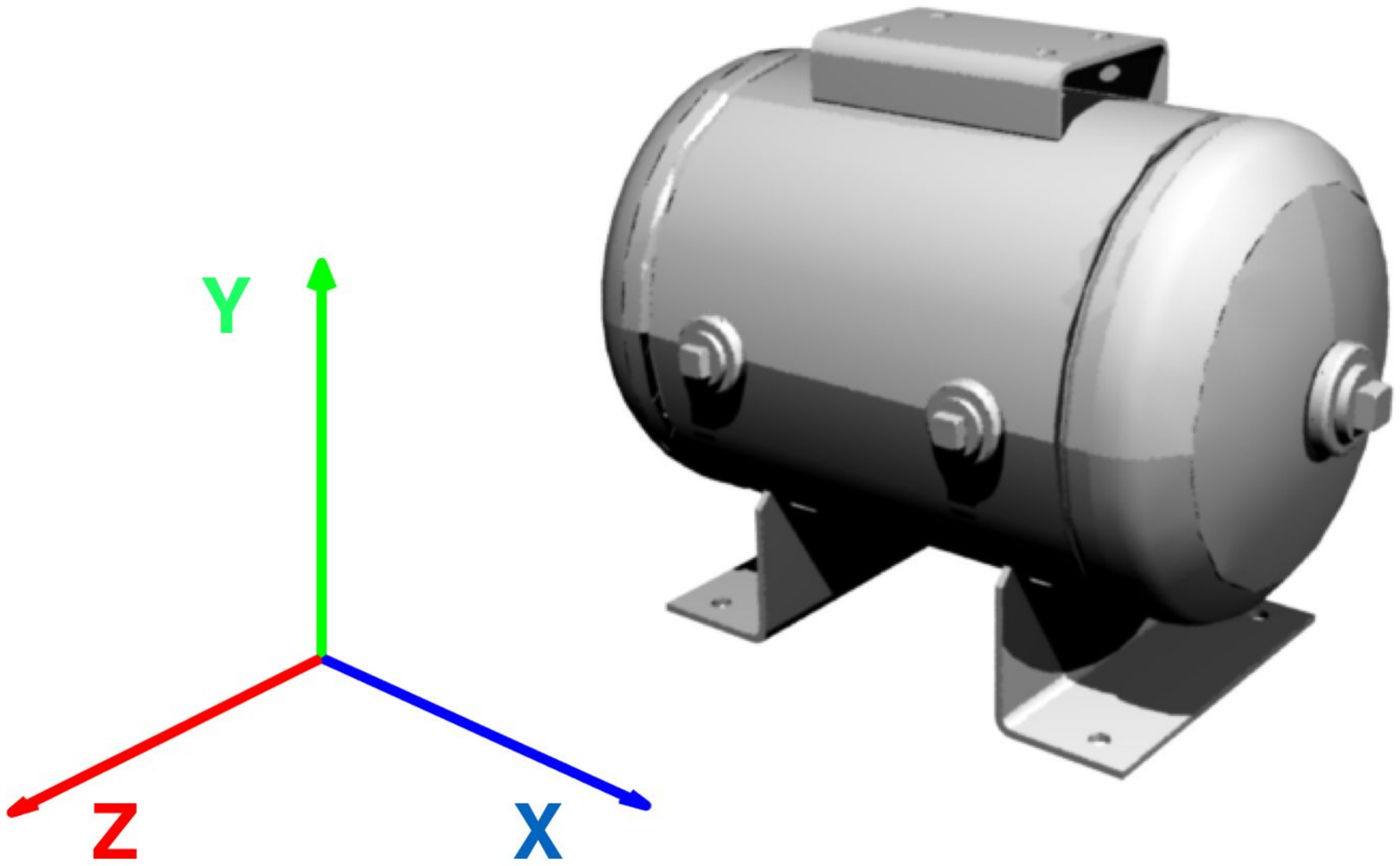} \\
		{\scriptsize $X \sim \infty, Y \sim 2, Z \sim 2.$ } & \hspace{-0.5cm}{\scriptsize $X \sim 2, Y \sim 2, Z \sim \infty.$} & \hspace{-1cm}{\scriptsize $X \sim 2, Y \sim 2, Z \sim 2.$}  & \hspace{-1cm}{\scriptsize $X \sim 1, Y \sim 2, Z \sim 1.$} \\
		\scriptsize (ix) & \scriptsize (x) & \scriptsize (xi) & \scriptsize (xii) \\	
		\includegraphics[trim={0 0 0 0}, clip = true,width=3.3cm]{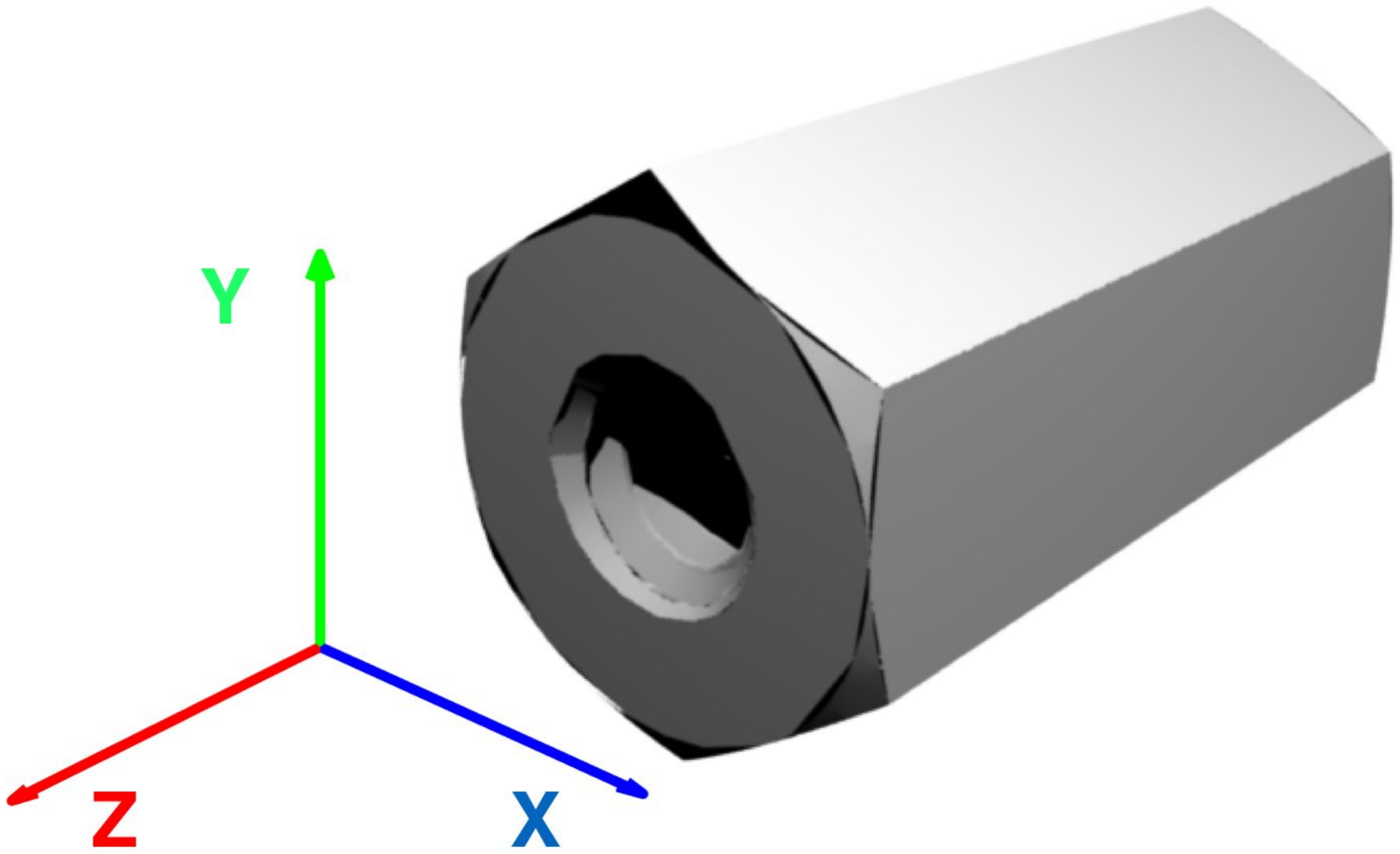} & 
		\includegraphics[trim={0 0 0 0}, clip = true,width=3.3cm]{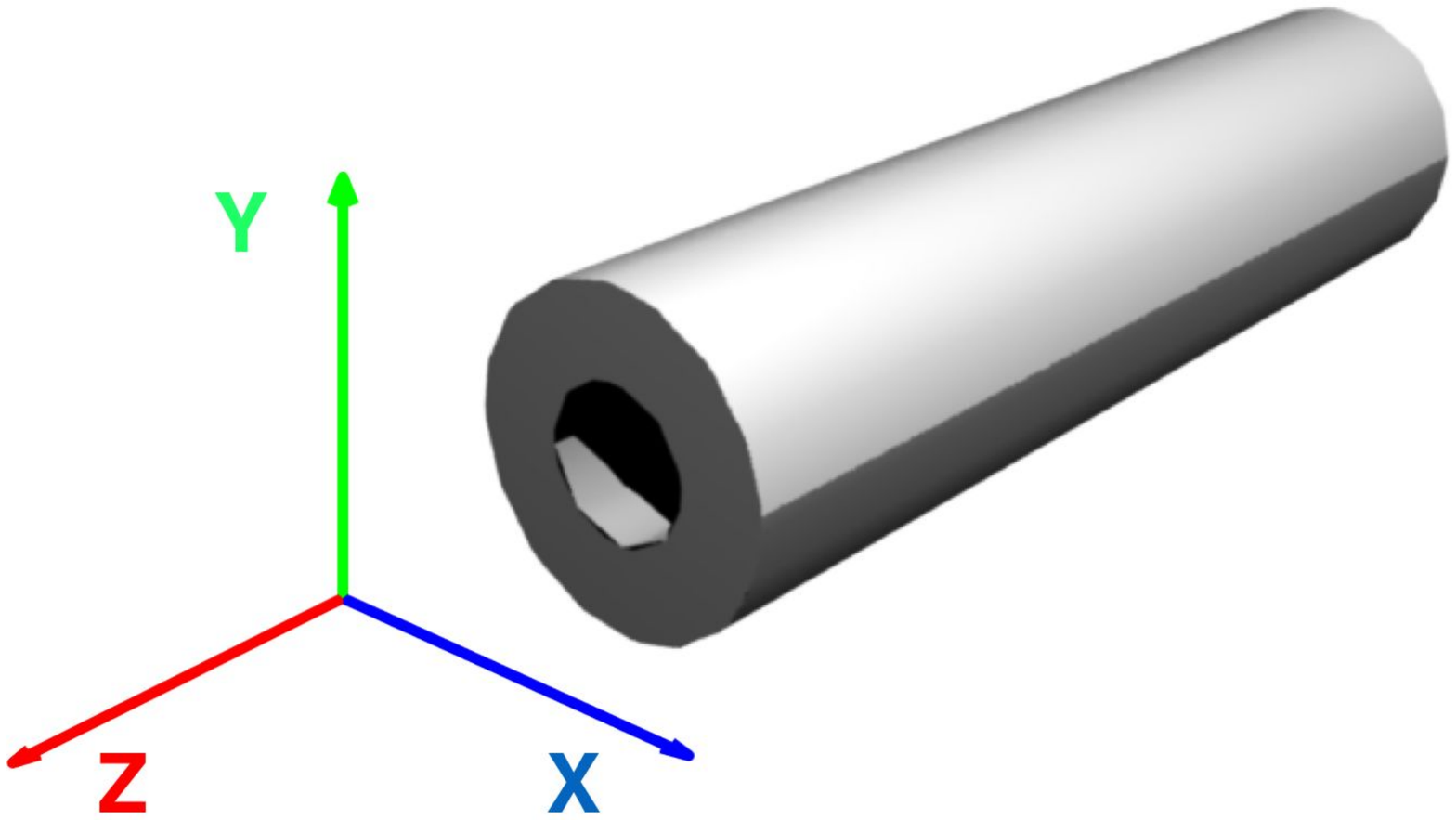} &	
		\includegraphics[trim={0 0 0 0}, clip = true,width=3.3cm]{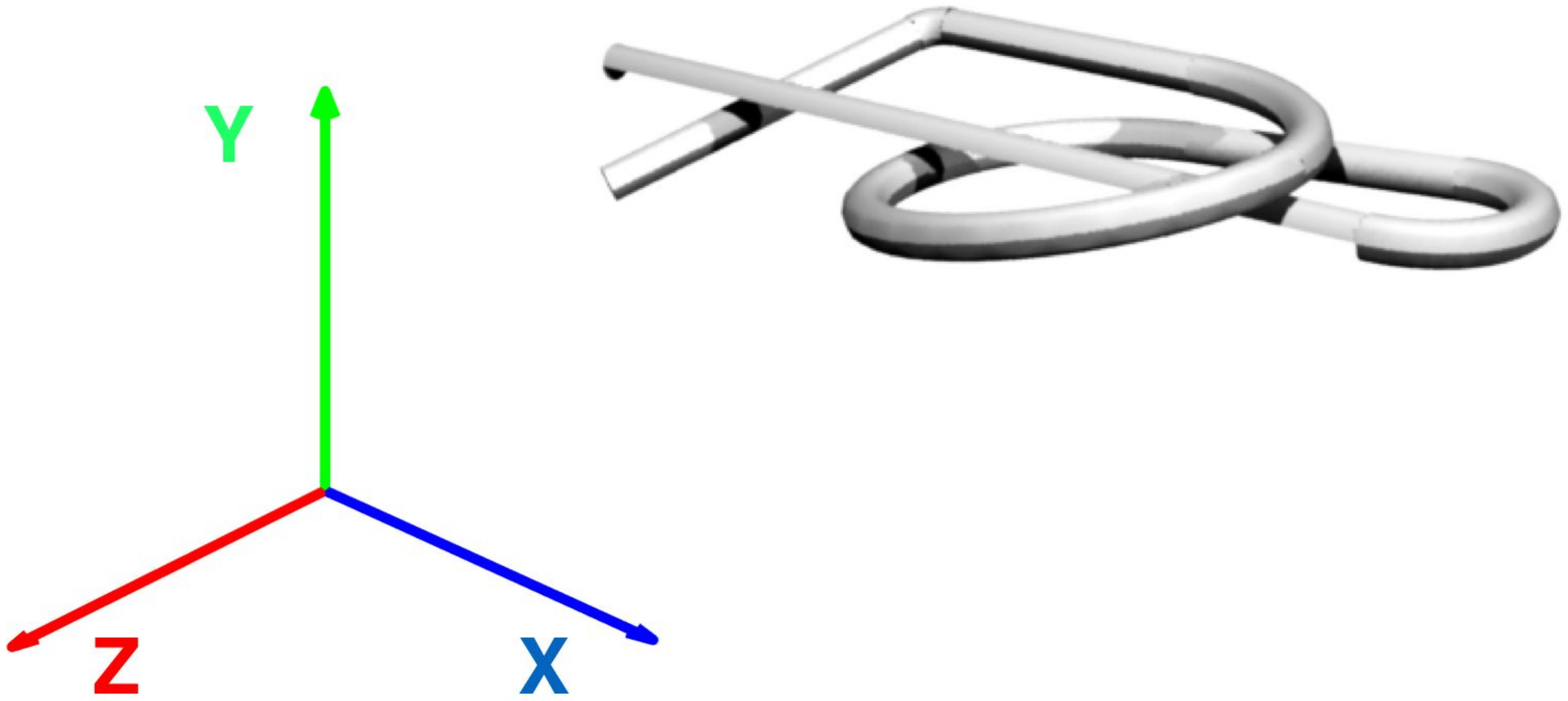} &
		\includegraphics[trim={0 0 0 0}, clip = true,width=3.3cm]{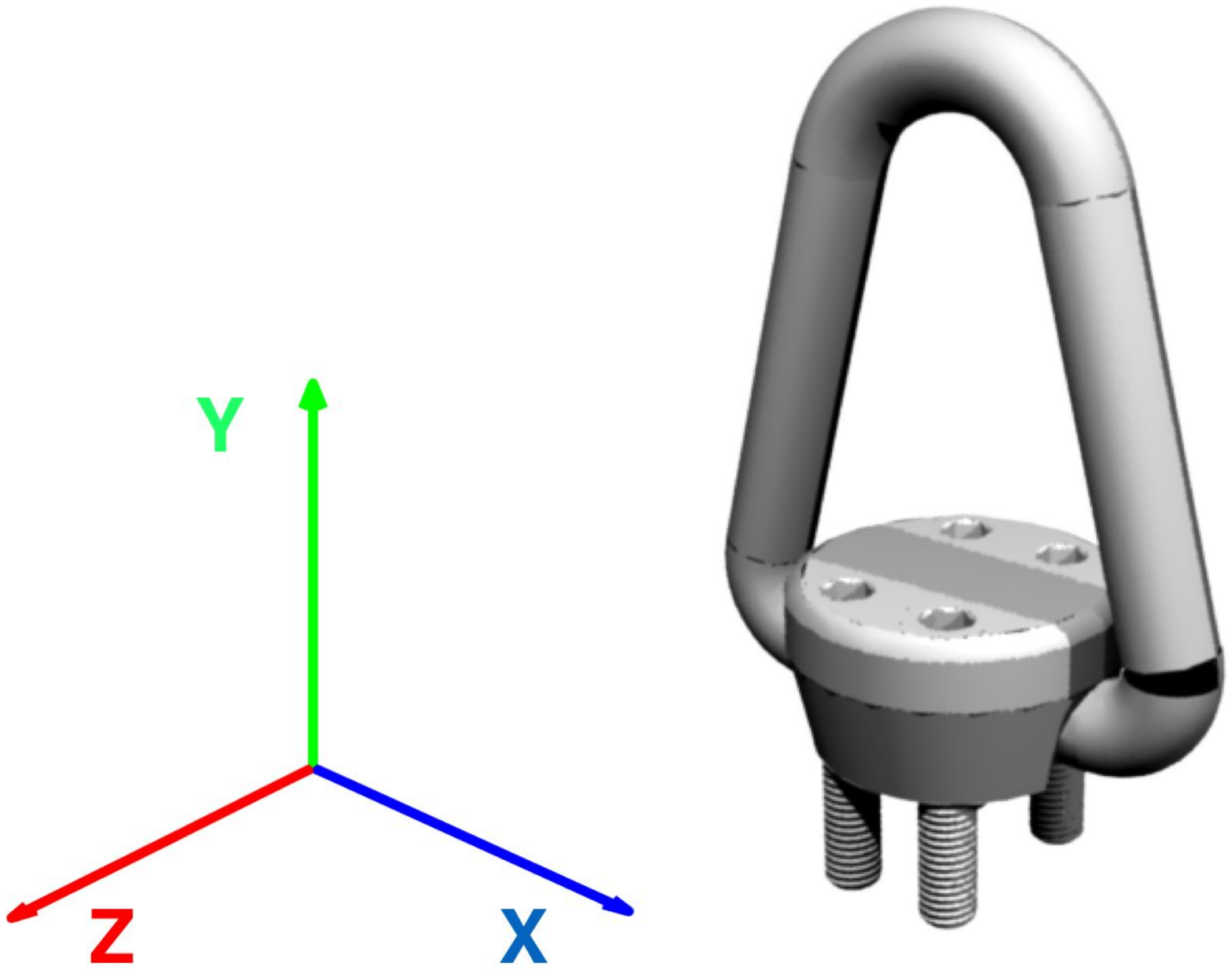} \\
		{\scriptsize $X \sim 2, Y \sim 2, Z \sim \infty.$ } & \hspace{-0.5cm}{\scriptsize $X \sim 2, Y \sim 2, Z \sim \infty.$}  & \hspace{-1cm}{\scriptsize $X \sim 1, Y \sim 1, Z \sim 1.$}  & \hspace{-1cm}{\scriptsize $X \sim 1, Y \sim 2, Z \sim 1.$} \\
		\scriptsize  (xiii) & \scriptsize  (xiv) & \scriptsize  (xv) & \scriptsize  (xvi) \\
		\includegraphics[trim={0 0 0 0}, clip = true,width=3.3cm]{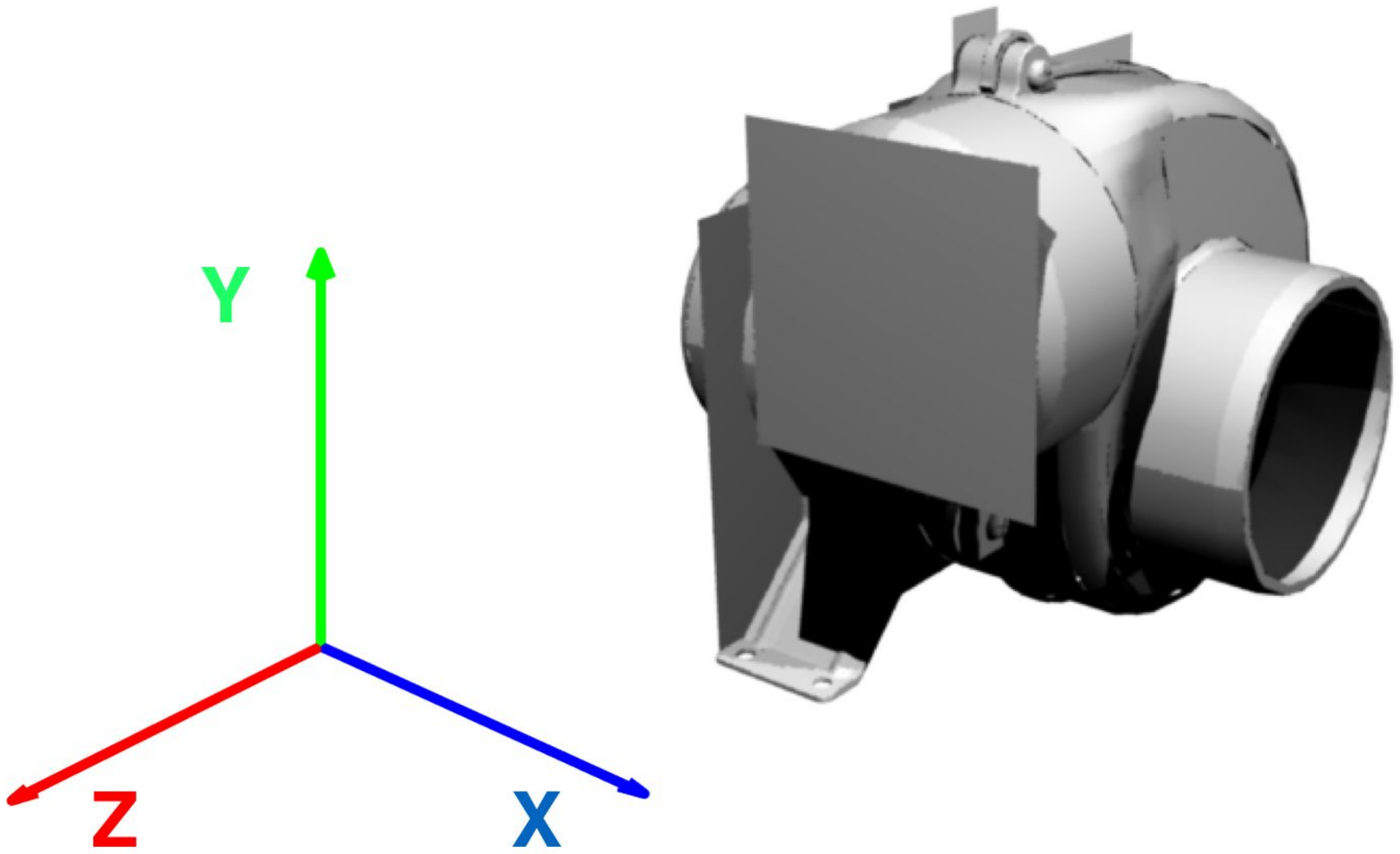} &
		\includegraphics[trim={0 0 0 0}, clip = true,width=3.3cm]{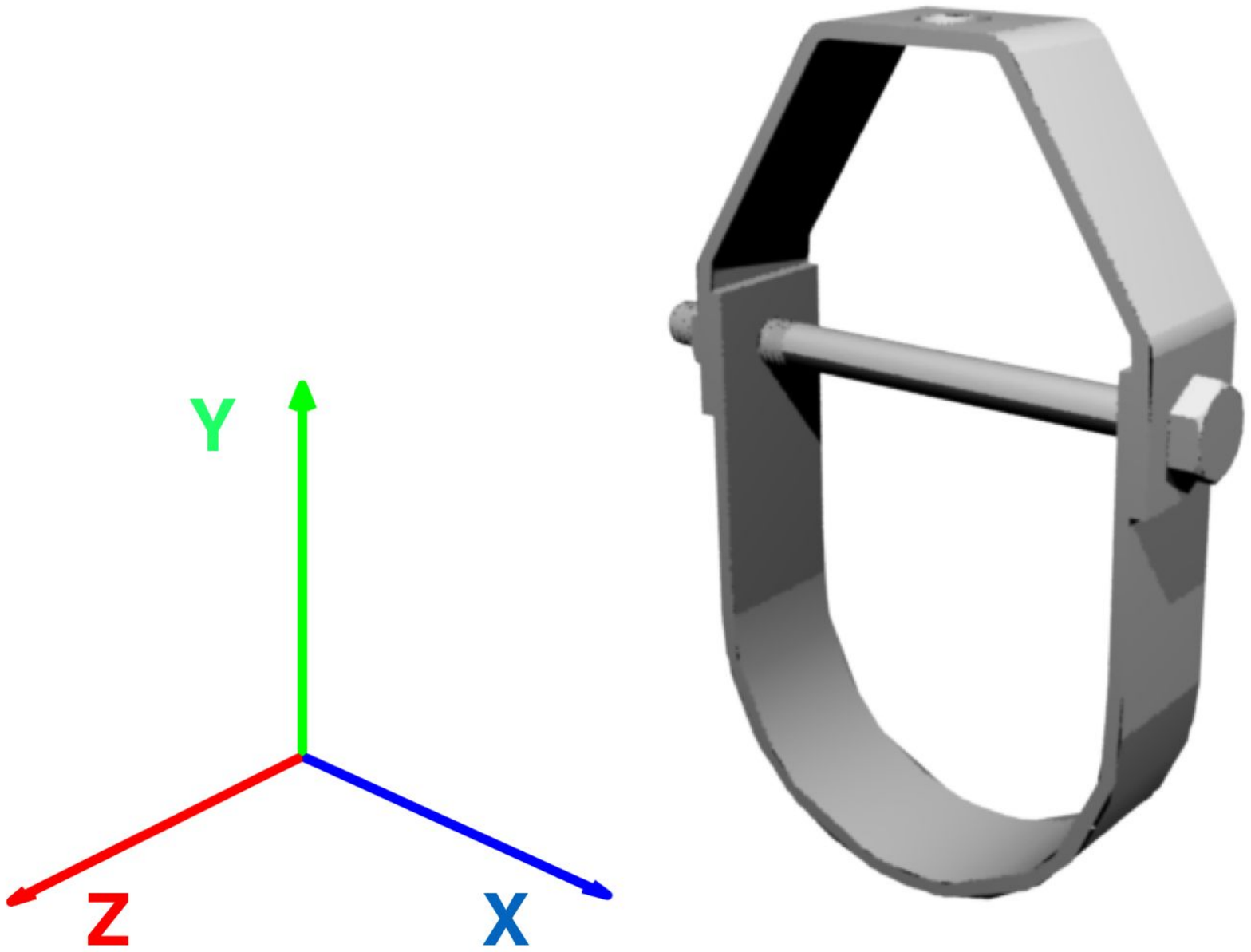} &
		\includegraphics[trim={0 0 0 0}, clip = true,width=3.3cm]{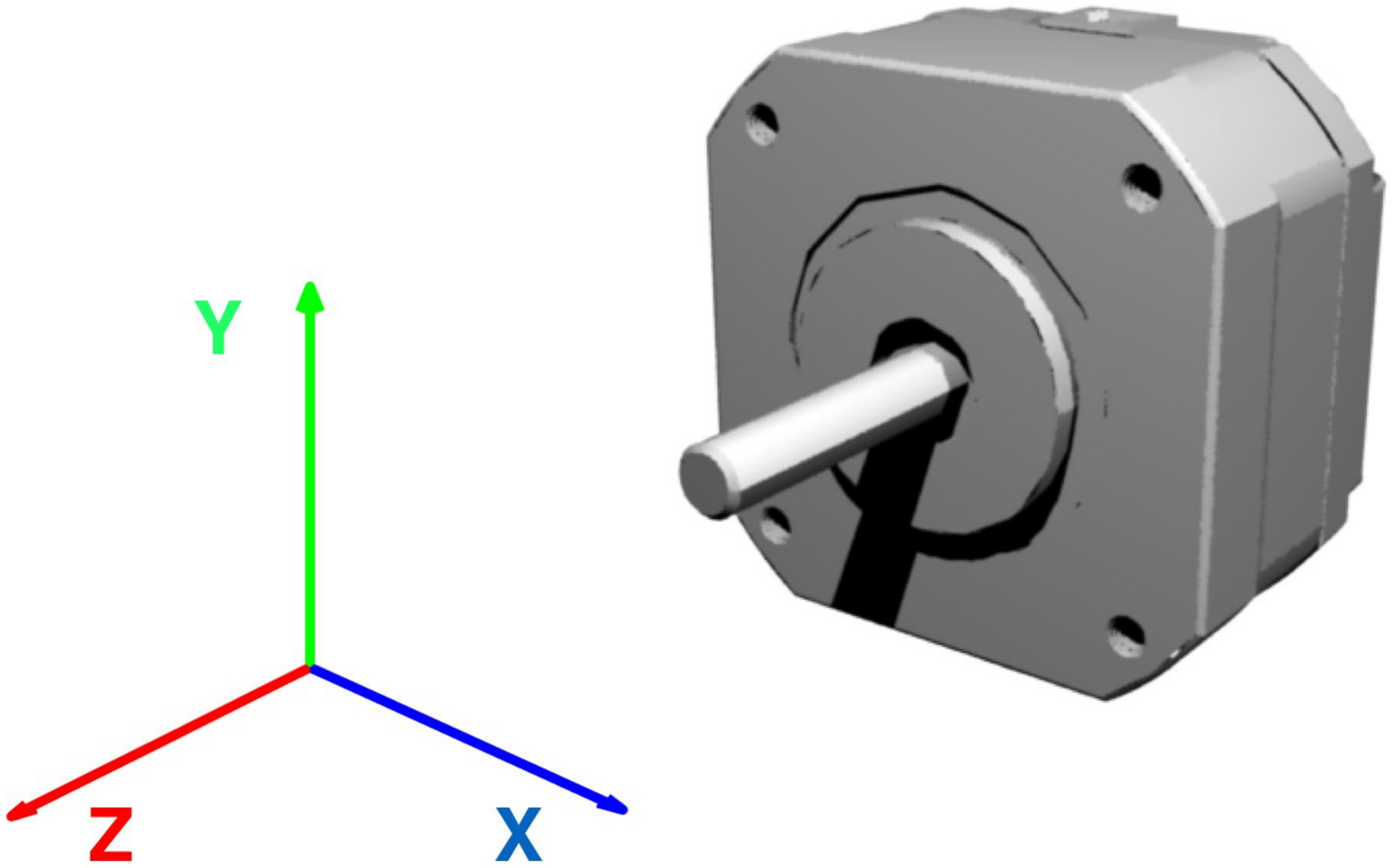} &
		\includegraphics[trim={0 0 0 0}, clip = true,width=3.3cm]{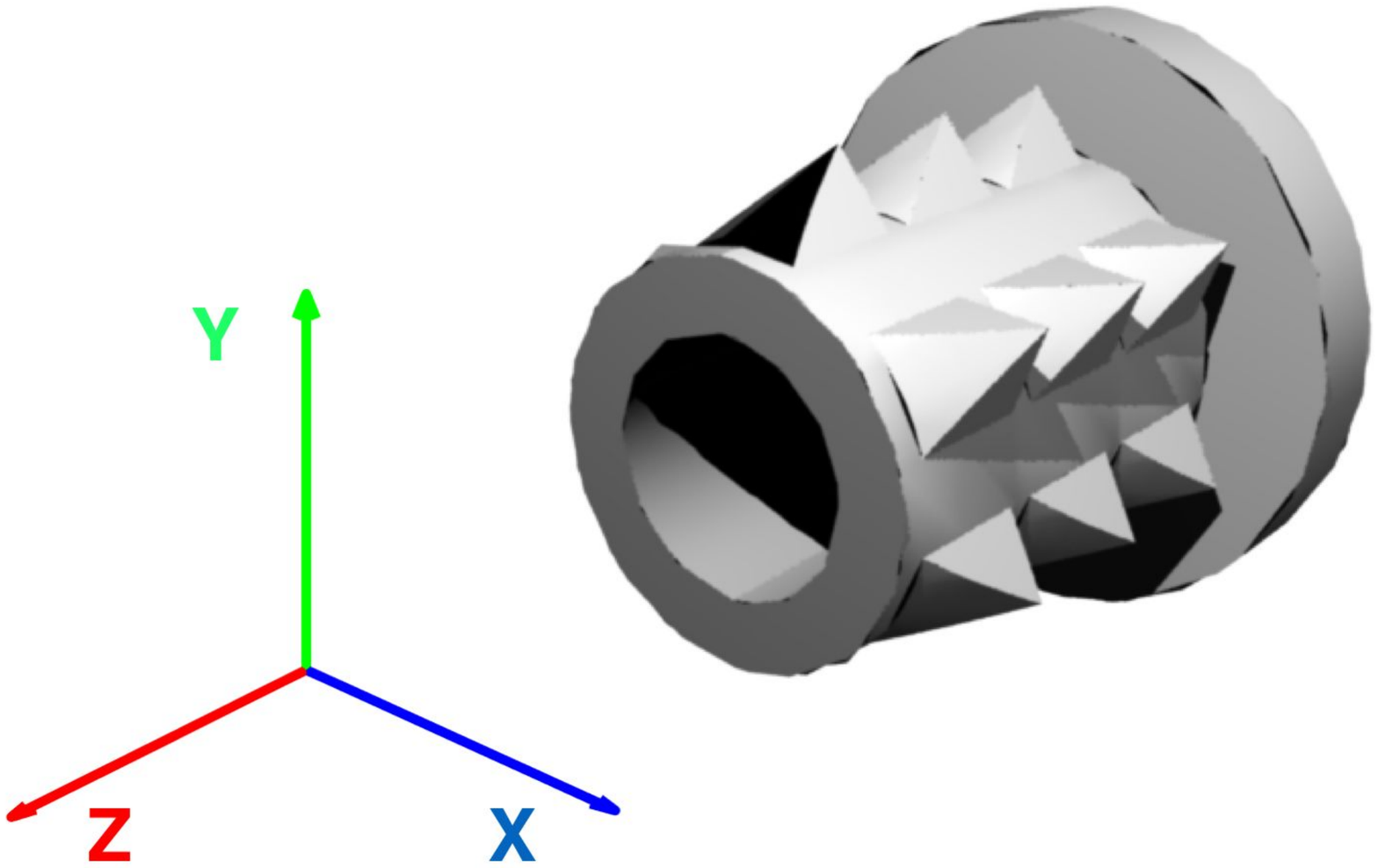} \\
		{\scriptsize $X \sim 1, Y \sim 1, Z \sim 1.$ } & \hspace{-0.5cm}{\scriptsize $X \sim 1, Y \sim 1, Z \sim 1.$} & \hspace{-1cm}{\scriptsize $X \sim 1, Y \sim 1, Z \sim \infty.$}  & \hspace{-1cm}{\scriptsize $X \sim 1, Y \sim 1, Z \sim \infty.$} \\
		\scriptsize (xvii) & \scriptsize (xviii) &  \scriptsize  (xix) & \scriptsize (xx) \\
		\includegraphics[trim={0 0 0 0}, clip = true,width=3.3cm]{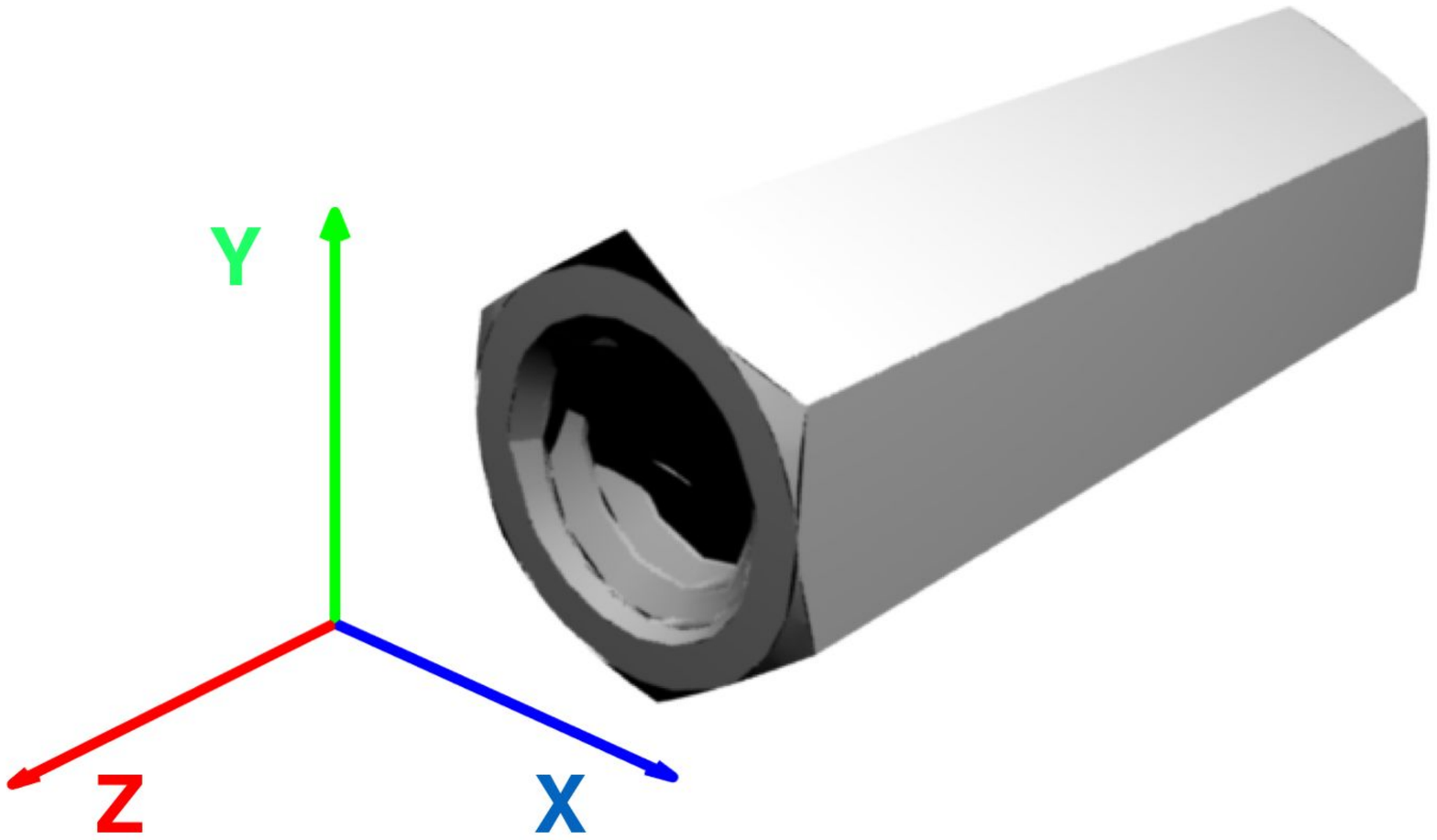} &
		\includegraphics[trim={0 0 0 0}, clip = true,width=3.3cm]{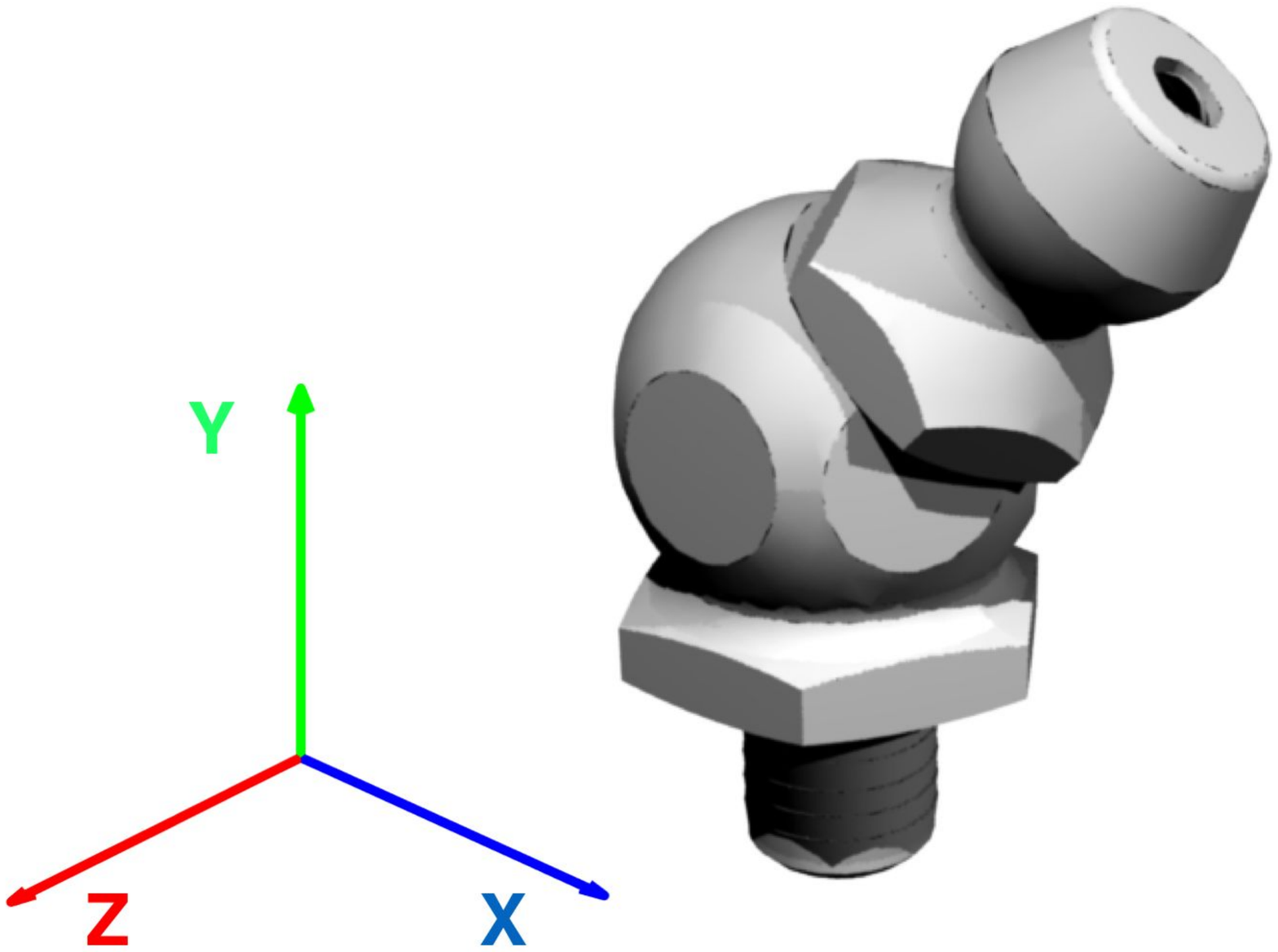} &	
		\includegraphics[trim={0 0 0 0}, clip = true,width=3.3cm]{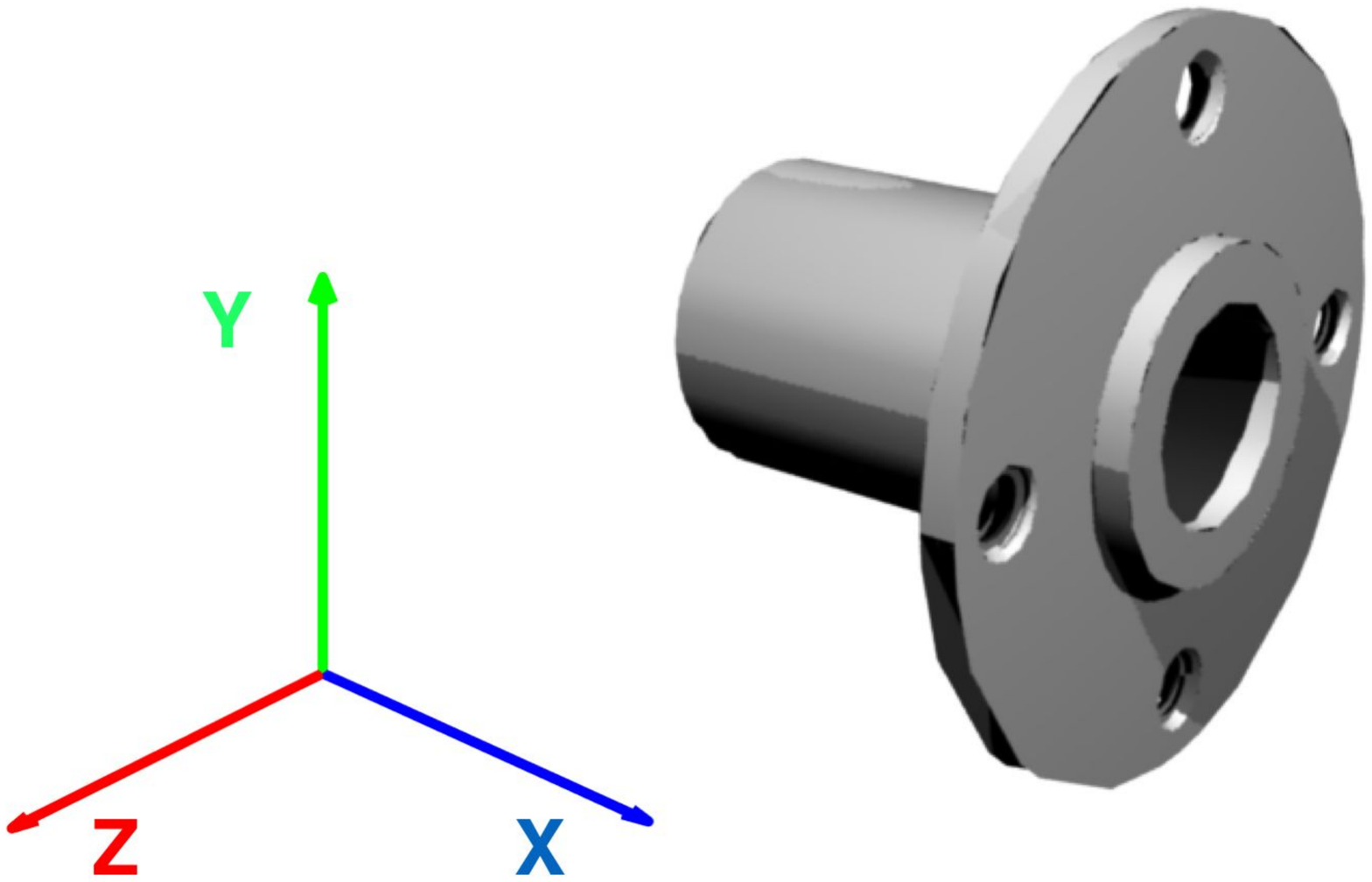} &
		\includegraphics[trim={0 0 0 0}, clip = true,width=3.3cm]{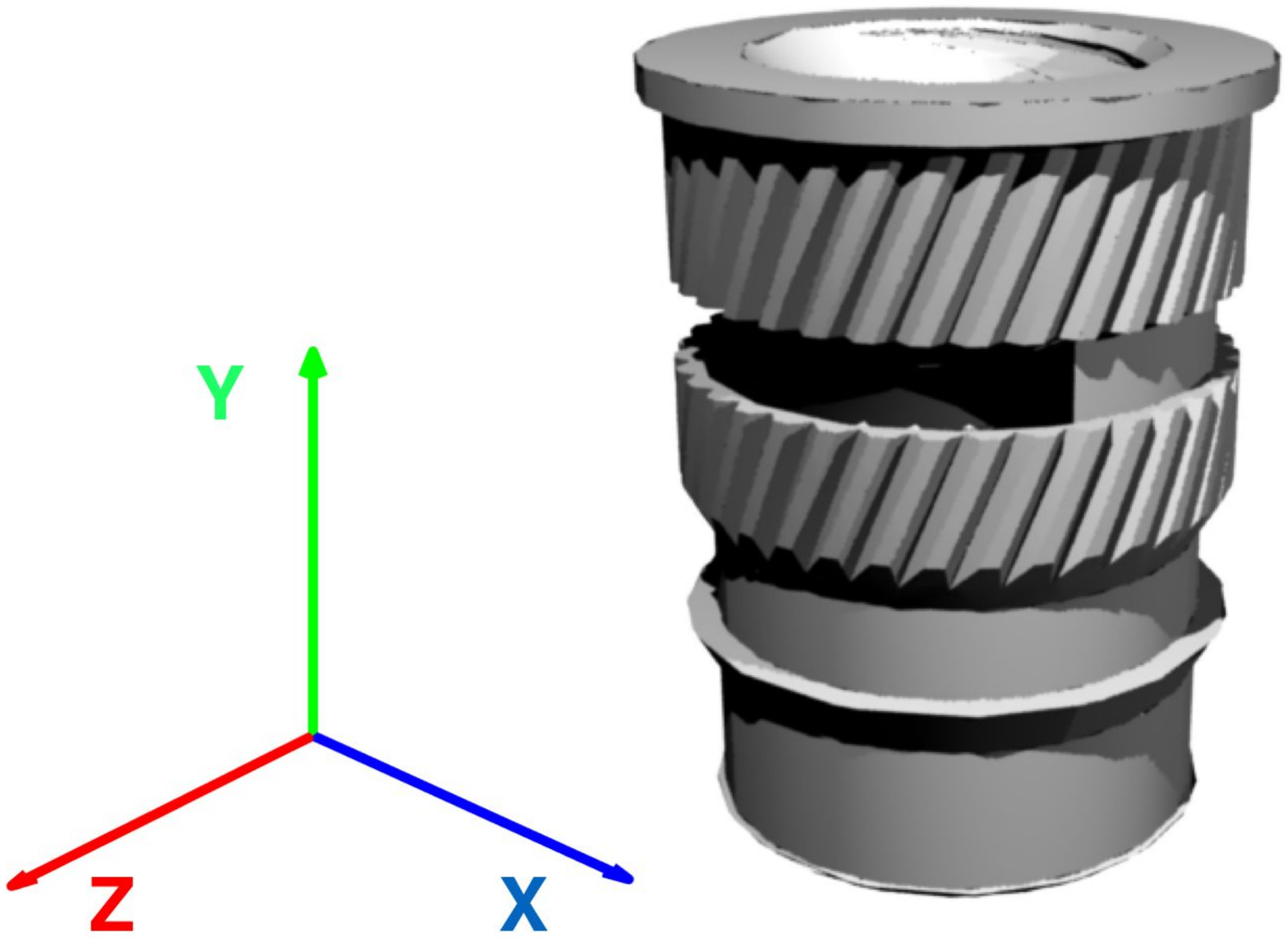} \\
		{\scriptsize $X \sim 2, Y \sim 2, Z \sim \infty.$ } & \hspace{-0.5cm}{\scriptsize $X \sim 1, Y \sim 1, Z \sim 1.$} & \hspace{-1cm}{\scriptsize $X \sim \infty, Y \sim 1, Z \sim 1.$}  & \hspace{-1cm}{\scriptsize $X \sim 1, Y \sim 1, Z \sim 1.$} \\
		\scriptsize (xxi) & \scriptsize (xxii) & \scriptsize (xxiii) & \scriptsize (xxiv) \\ 
%		\includegraphics[trim={0 0 0 0}, clip = true,width=3.3cm]{figs/sym_obj6553.png} & 
%		\includegraphics[trim={0 0 0 0}, clip = true,width=3.3cm]{figs/sym_obj3555.png} &	
%		\includegraphics[trim={0 0 0 0}, clip = true,width=3.3cm]{figs/sym_obj3597.png} & 
%		\includegraphics[trim={0 0 0 0}, clip = true,width=3.3cm]{figs/sym_obj3622.png}  \\
%		{\scriptsize $X \sim 2, Y \sim 2, Z \sim \infty.$ } & \hspace{-0.5cm}{\scriptsize $X \sim \infty, Y \sim 1, Z \sim 1.$} & \hspace{-1cm}{\scriptsize $X \sim \infty, Y \sim 2, Z \sim 2.$} & \hspace{-1cm}{\scriptsize $X \sim 2, Y \sim 2, Z \sim \infty.$}  \\
% 		\includegraphics[trim={0 0 0 0}, clip = true,width=3.3cm]{figs/sym_obj4219.png} & 
%		\includegraphics[trim={0 0 0 0}, clip = true,width=3.3cm]{figs/sym_obj4312.png} &	
%		\includegraphics[trim={0 0 0 0}, clip = true,width=3.3cm]{figs/sym_obj4365.png} & 
%		\includegraphics[trim={0 0 0 0}, clip = true,width=3.3cm]{figs/sym_obj4619.png}  \\
%		{\scriptsize $X \sim 1, Y \sim \infty, Z \sim 1.$ } & \hspace{-0.5cm}{\scriptsize $X \sim 1, Y \sim 1, Z \sim 1.$} & \hspace{-1cm}{\scriptsize $X \sim 1, Y \sim \infty, Z \sim 1.$} & \hspace{-1cm}{\scriptsize $X \sim 1, Y \sim 1, Z \sim 1.$}  \\
	\end{tabular}
	}
	\vspace{-4mm}
	\caption{\bf Symmetry Prediction Results}
	\label{fig:symmetries_inferred_2}
	\vspace{-2mm}
\end{figure}

We show examples of the CAD models obtained along with their inferred symmetry in~\figref{fig:symmetries_inferred_1} and~\figref{fig:symmetries_inferred_2}. One of the primary reasons for failure is the non-alignment of viewpoints due the discretization. Another reason of failure is that examples of certain order classes are not present in training. For example, the objects in the bottom left of~\figref{fig:symmetries_inferred_1} (xxi, xxii, xxiii) has $ \mathcal{O(\bZ)} $ as $ 8, 12 $ and $ 13 $ respectively, which was not present in the training set. The orders inferred in all these cases were $ \infty $. This leads to false positive examples when training for pose estimation.

Since our input is a 2D image, but we reason about the equivalence of 3D shapes, it can confuse the network. An example of this can be seen for the order prediction along the $ \bZ $ axis in~\figref{fig:symmetries_inferred_1} (ii). A viewpoint from the positive $ \bZ $ direction would indicate the order to be $ 4 $, but from the negative $ \bZ $ direction it should be inferred as $ \infty $. However the network in our approach predicts a single prediction of $ \infty $.

% Qualitatively, the cases where the model most often fails to predict rotational symmetries is in the boundaries of the limitation explained in Subsection \ref{sec:rot_limitations}. When an object has a slight section that breaks a higher order of symmetry, i.e an object with circular parts with a handle. 

% Also, we notice that different orders than shown in training tend to be predicted as infinite-order symmetries, given the lack of training samples with i.e four and sixth order symmetries. 

\paragraph{Pose Estimation.}

\begin{figure*}[] 
	\hspace{-1.3cm}\parbox{\linewidth}{
		\begin{tabular}{|c|c|cccc|c|}
			\hline
			{\bf Input } & {\bf GT } & \multicolumn{4}{c|}{\bf Top-4 Predictions} & {\bf $  d_\text{rot, best}^{sym} $} \\
			\hline 
			\includegraphics[trim={9cm 4cm 9cm 4cm}, clip = true,width=0.12\linewidth]{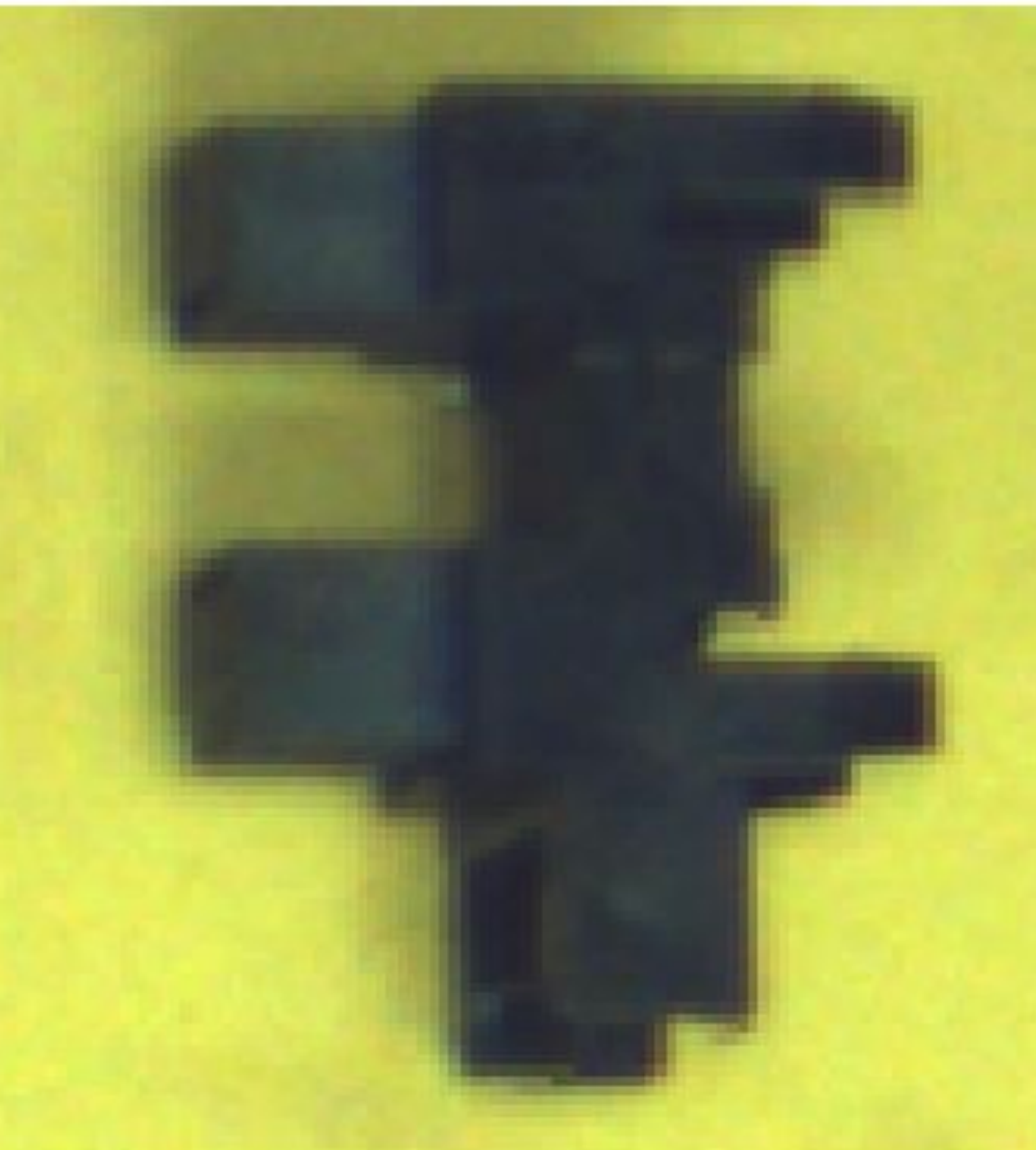} &
			\includegraphics[trim={9cm 4cm 9cm 4cm}, clip = true,width=0.12\linewidth]{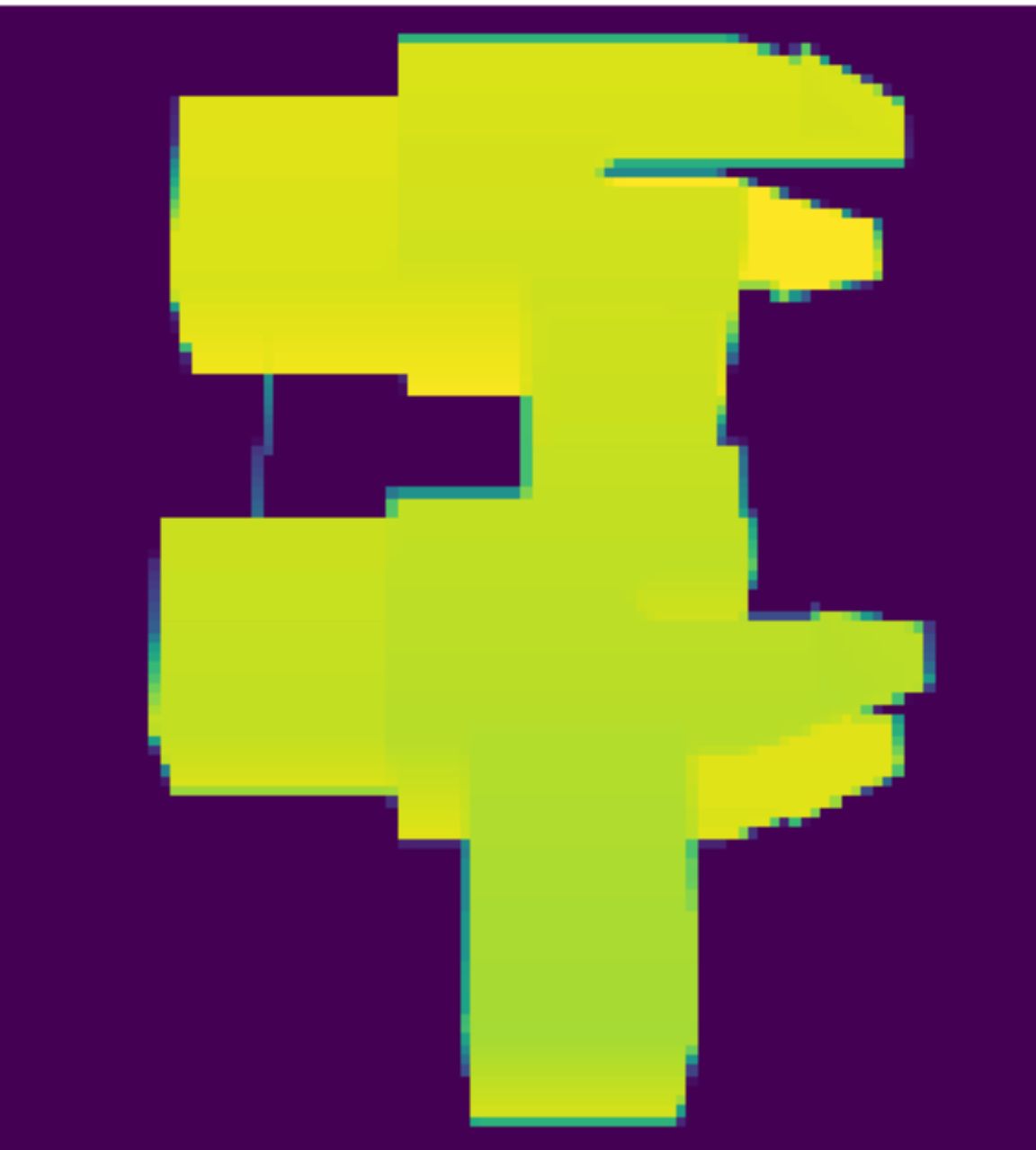}  &
			\fcolorbox{green}{white}{\includegraphics[trim={9cm 4cm 9cm 4cm}, clip = true,width=0.12\linewidth]{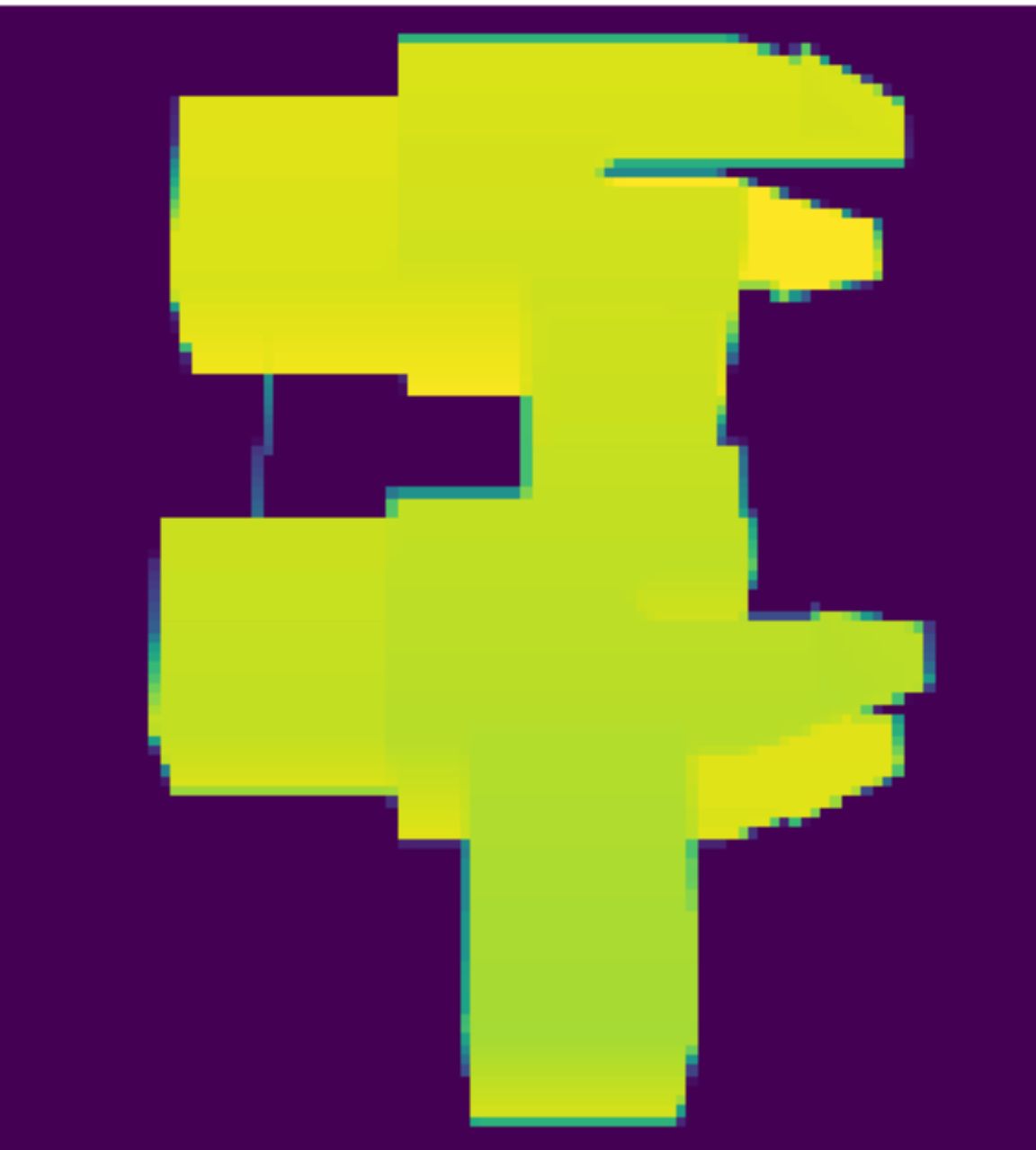} } &
			\includegraphics[trim={9cm 4cm 9cm 4cm}, clip = true,width=0.12\linewidth]{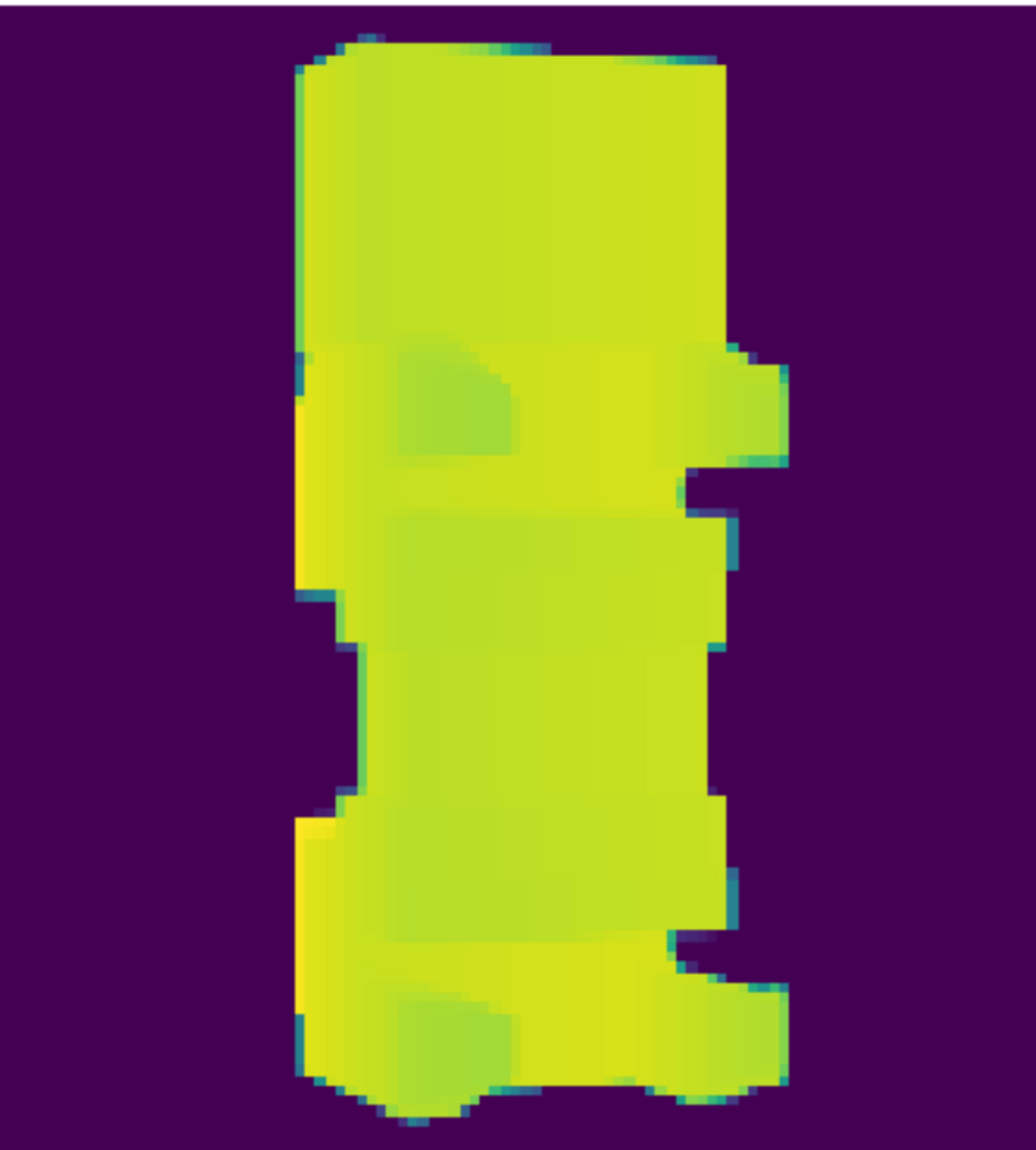}  &
			\includegraphics[trim={9cm 4cm 9cm 4cm}, clip = true,width=0.12\linewidth]{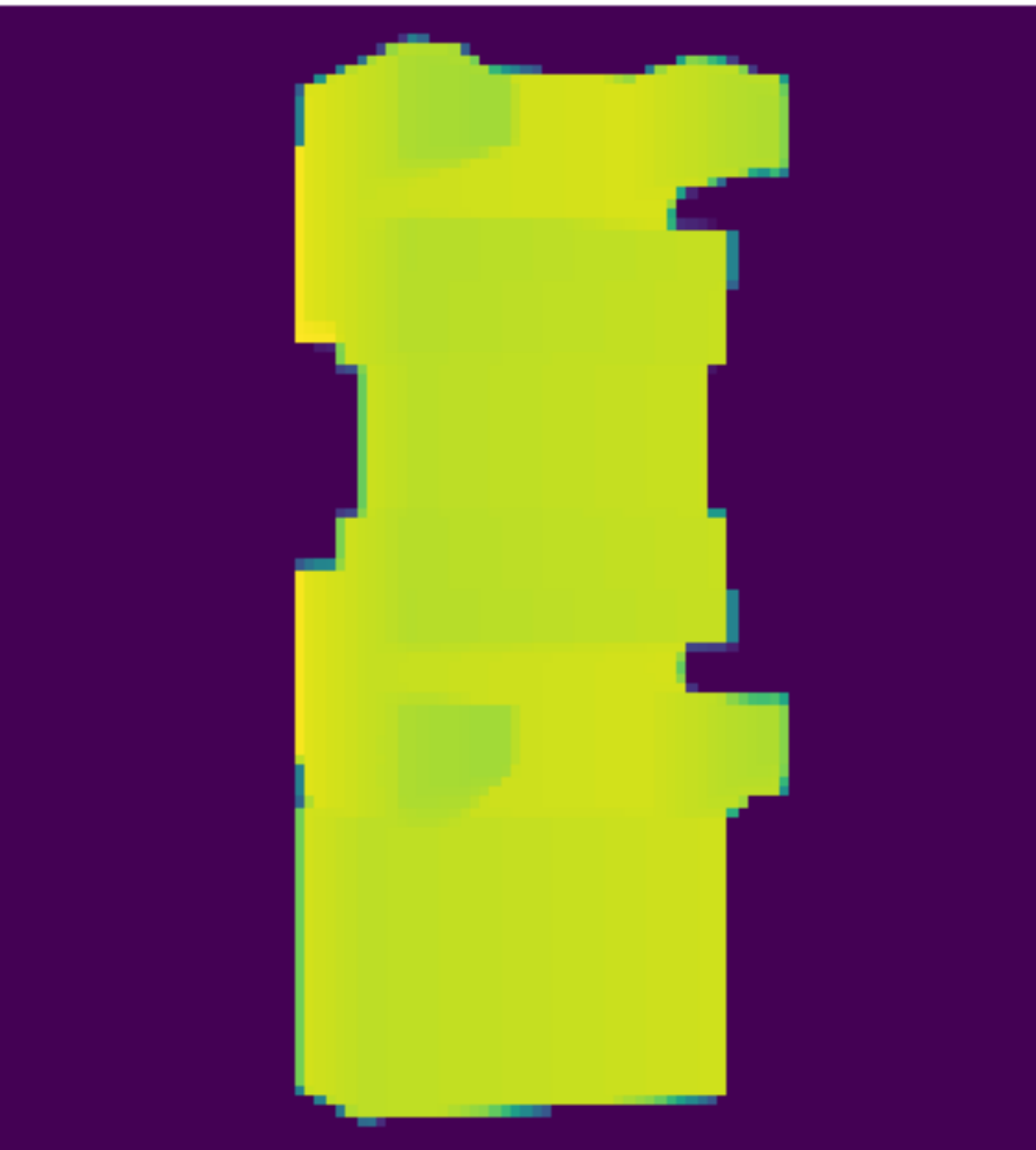}  &
			\includegraphics[trim={9cm 4cm 9cm 4cm}, clip = true,width=0.12\linewidth]{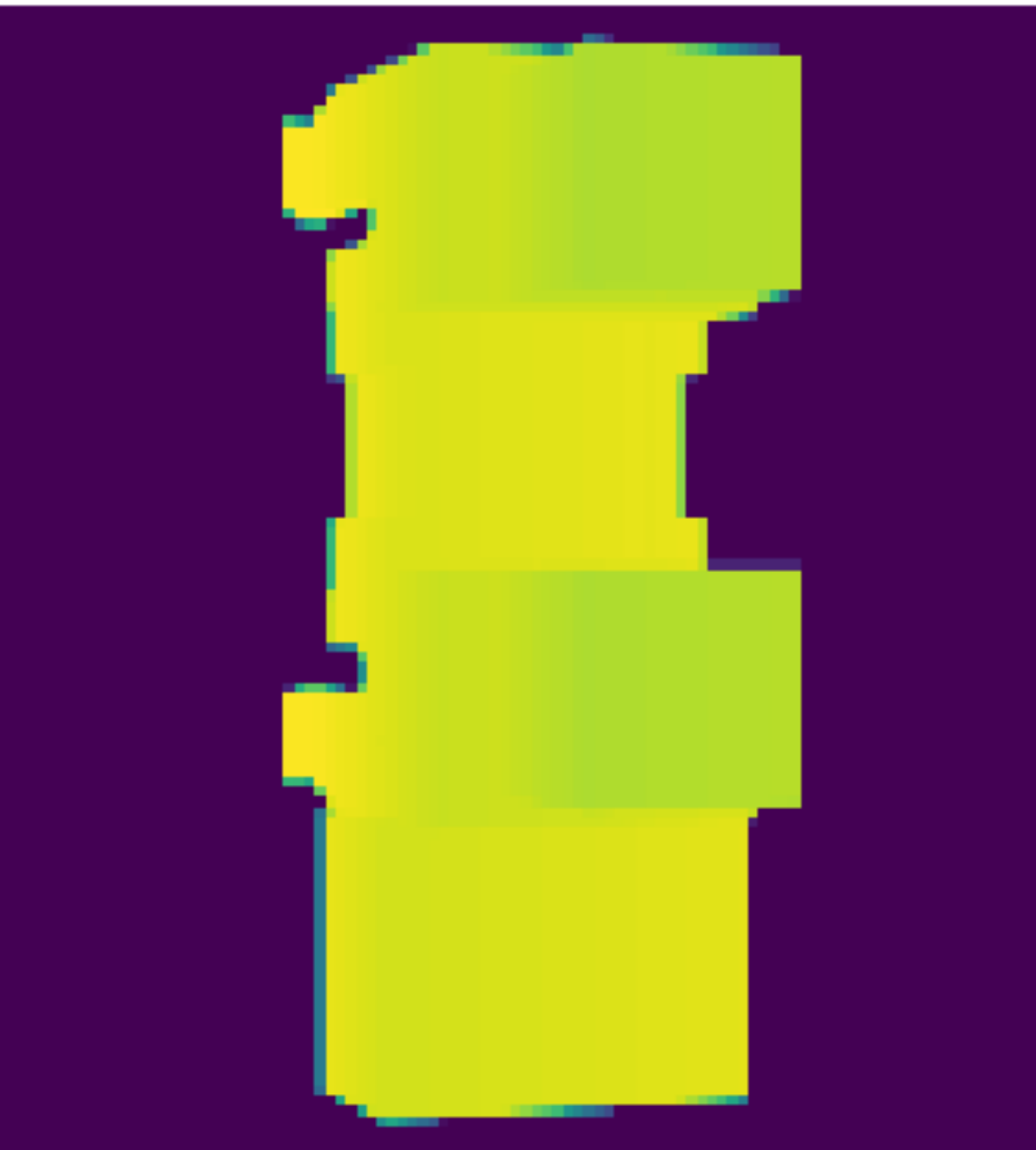}  & 
			\raisebox{2\height}{\LARGE 4.62$^{\circ}$} \\
			\hline 
			\includegraphics[trim={9cm 4cm 9cm 4cm}, clip = true,width=0.12\linewidth]{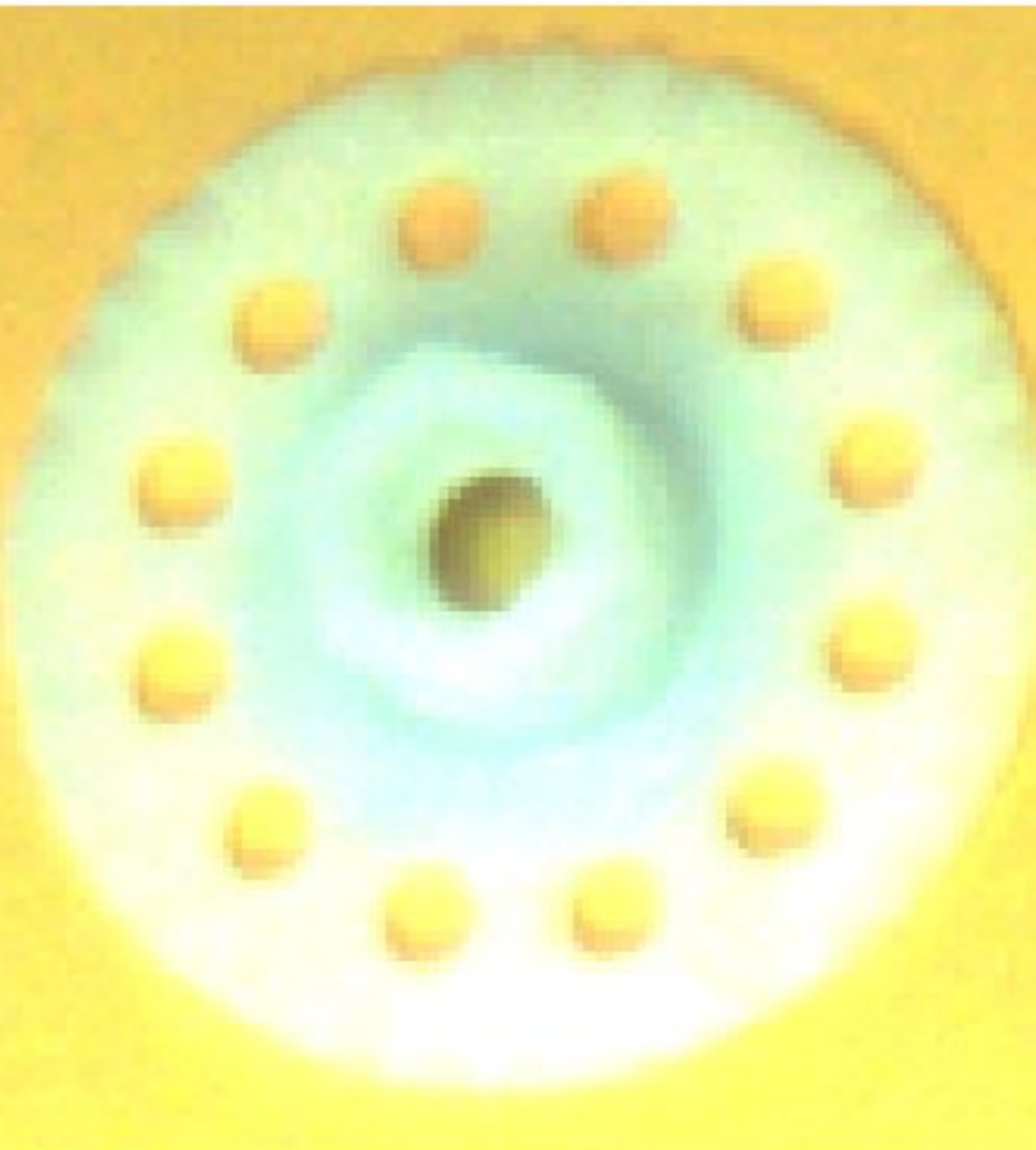} &
			\includegraphics[trim={9cm 4cm 9cm 4cm}, clip = true,width=0.12\linewidth]{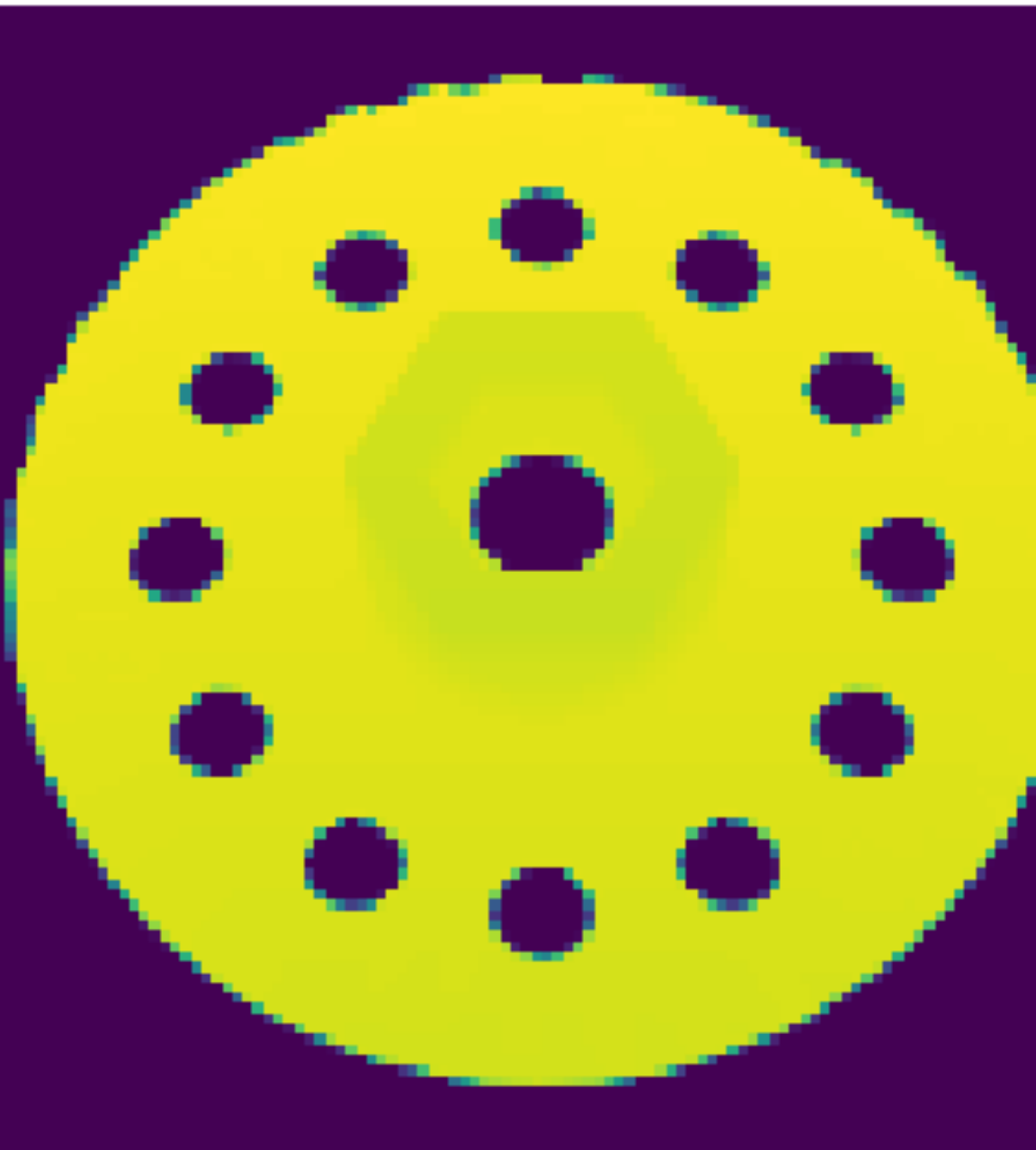} &
			\includegraphics[trim={9cm 4cm 9cm 4cm}, clip = true,width=0.12\linewidth]{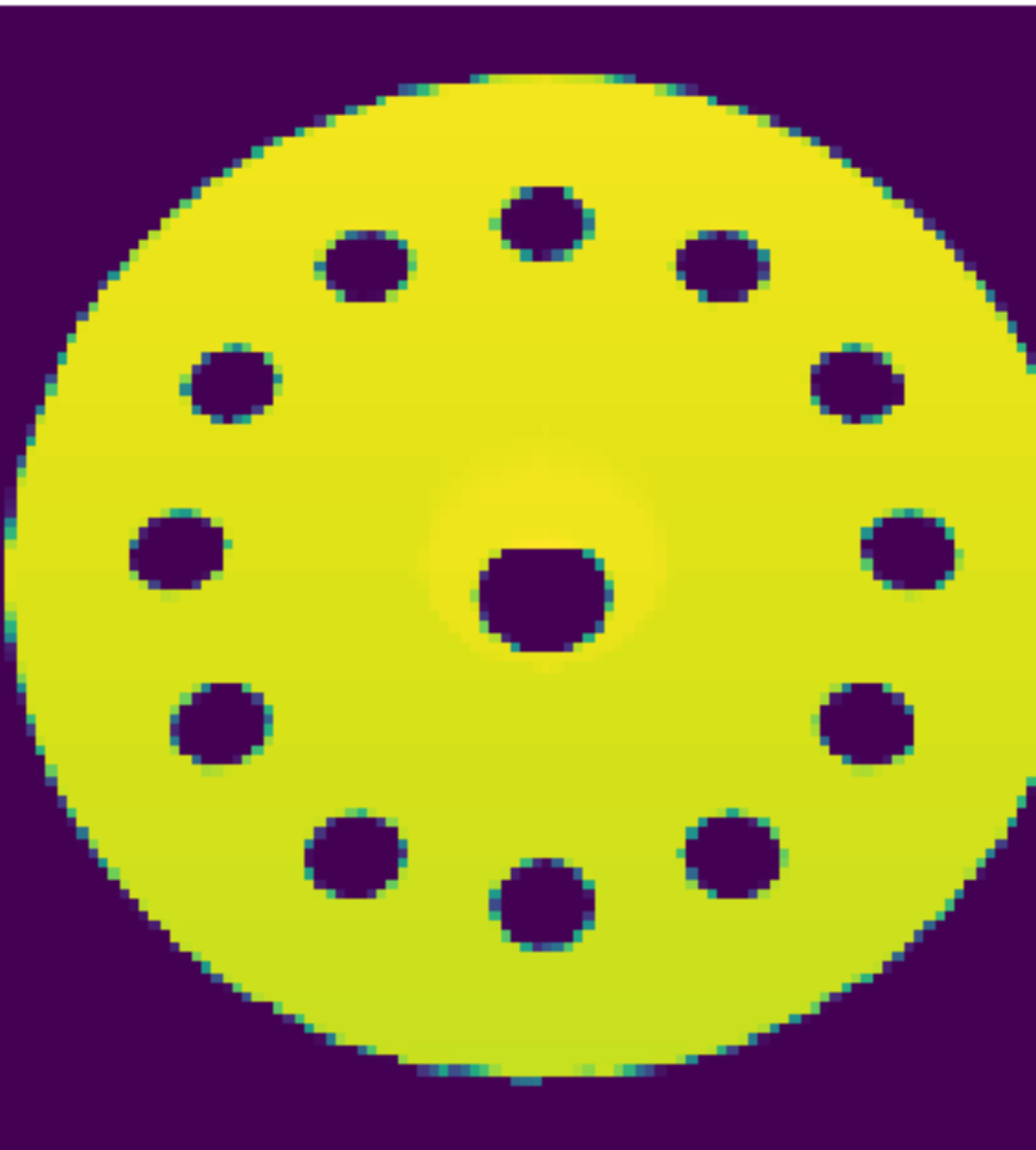} &
			\includegraphics[trim={9cm 4cm 9cm 4cm}, clip = true,width=0.12\linewidth]{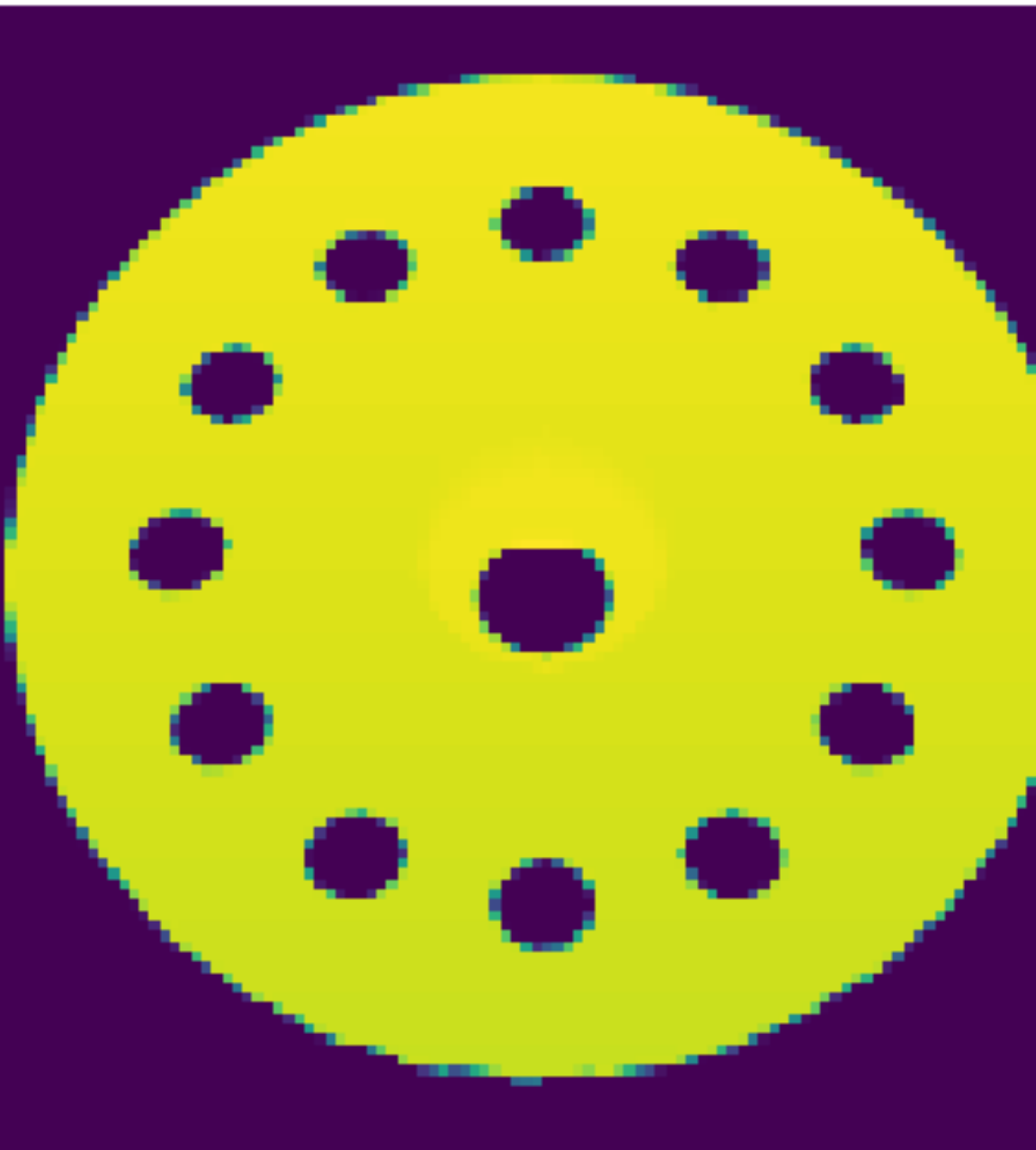} &
			\fcolorbox{green}{white}{\includegraphics[trim={9cm 4cm 9cm 4cm}, clip = true,width=0.12\linewidth]{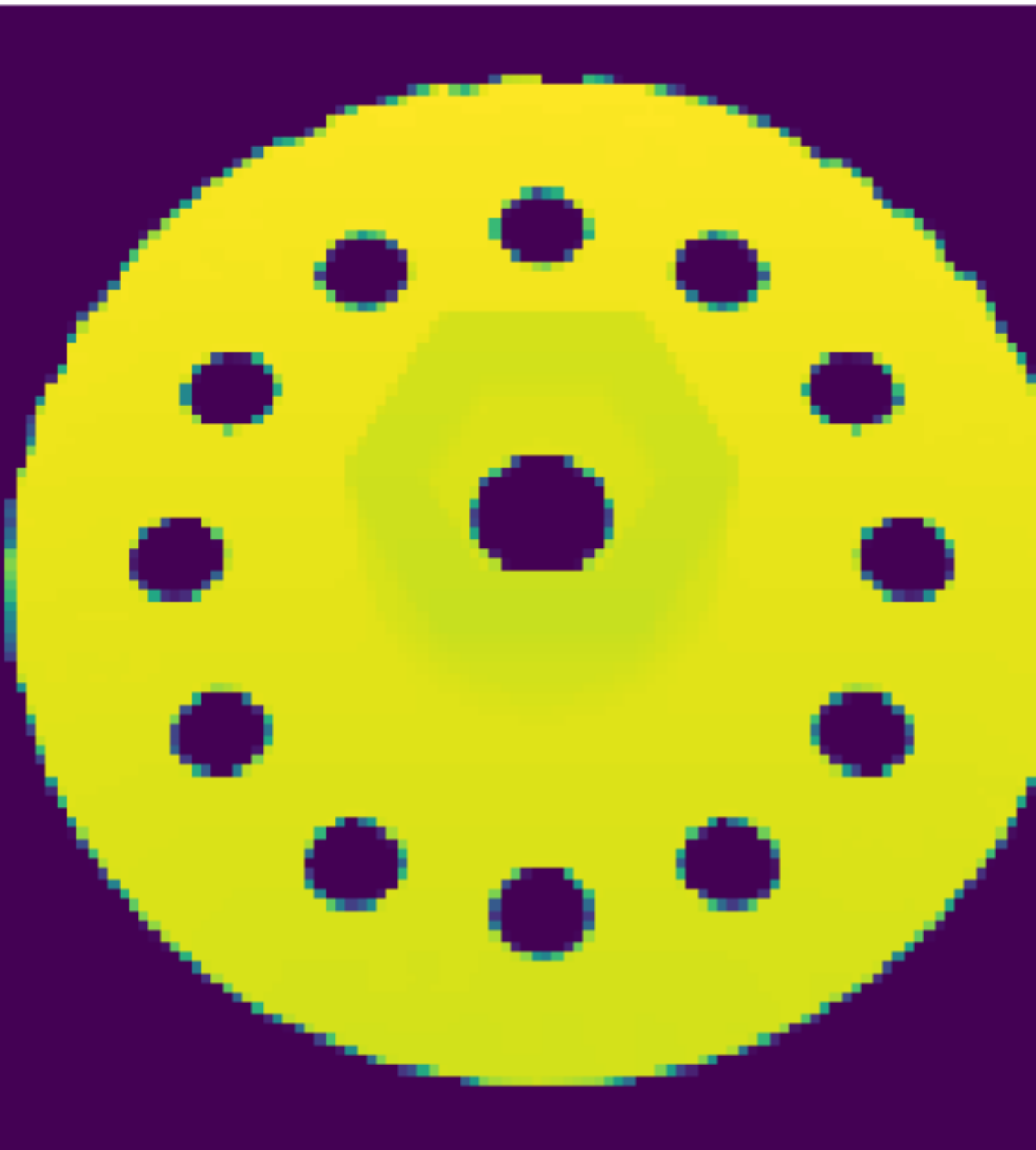}} &
			\fcolorbox{green}{white}{\includegraphics[trim={9cm 4cm 9cm 4cm}, clip = true,width=0.12\linewidth]{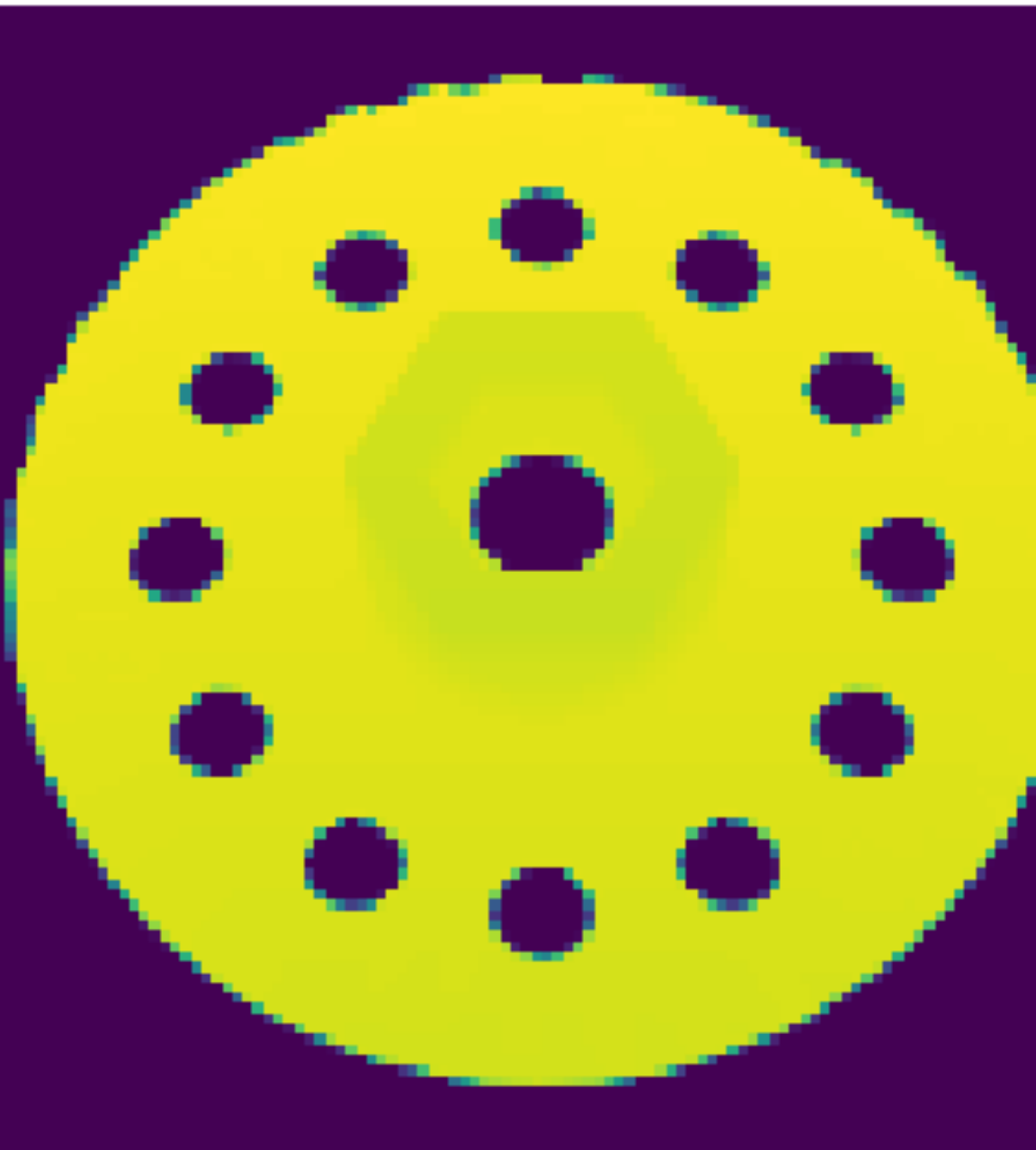}} & 
			\raisebox{2\height}{\LARGE 5.93$^{\circ}$}  \\
			\hline 
			\includegraphics[trim={9cm 4cm 9cm 4cm}, clip = true,width=0.12\linewidth]{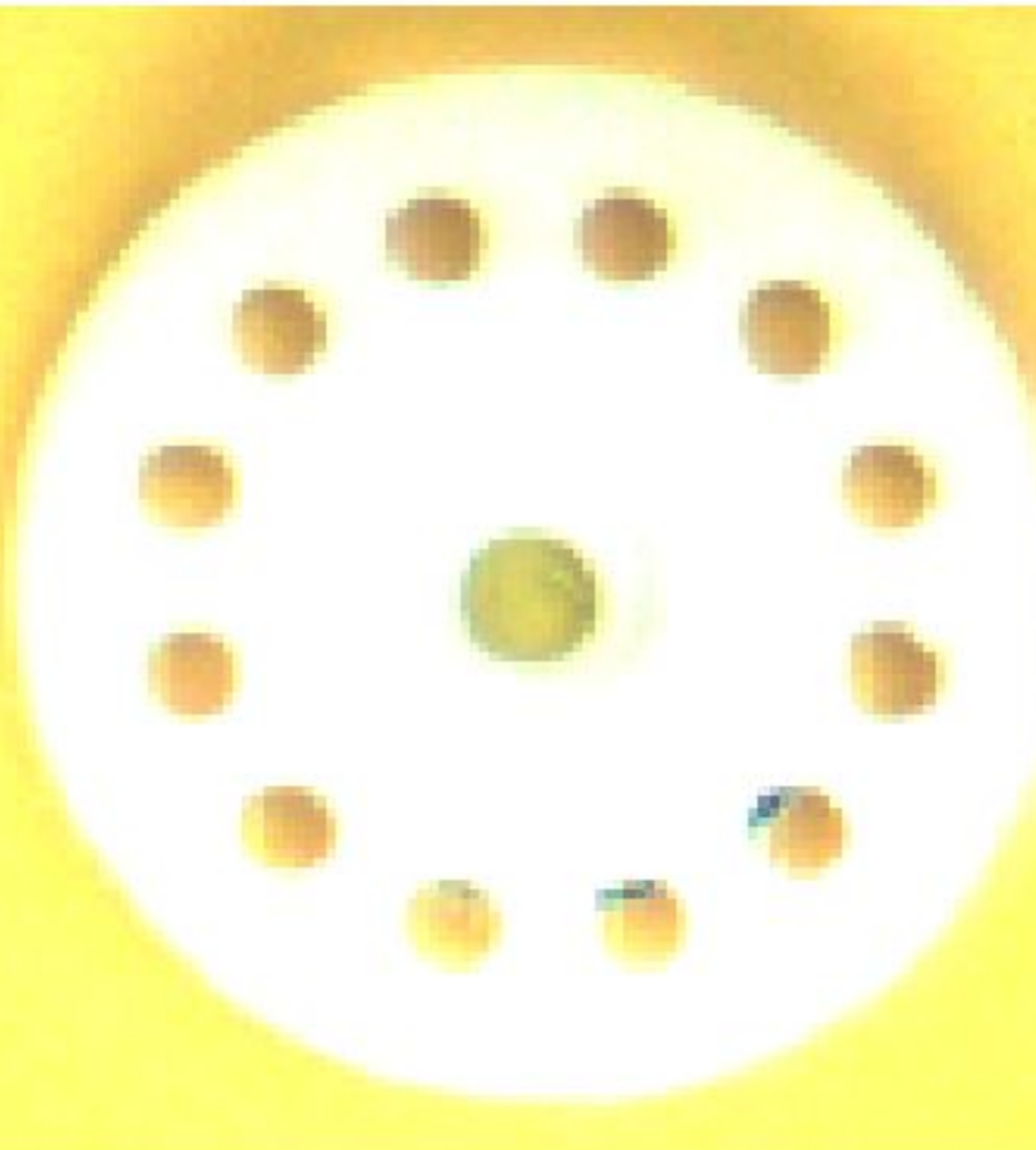}  &
			\includegraphics[trim={9cm 4cm 9cm 4cm}, clip = true,width=0.12\linewidth]{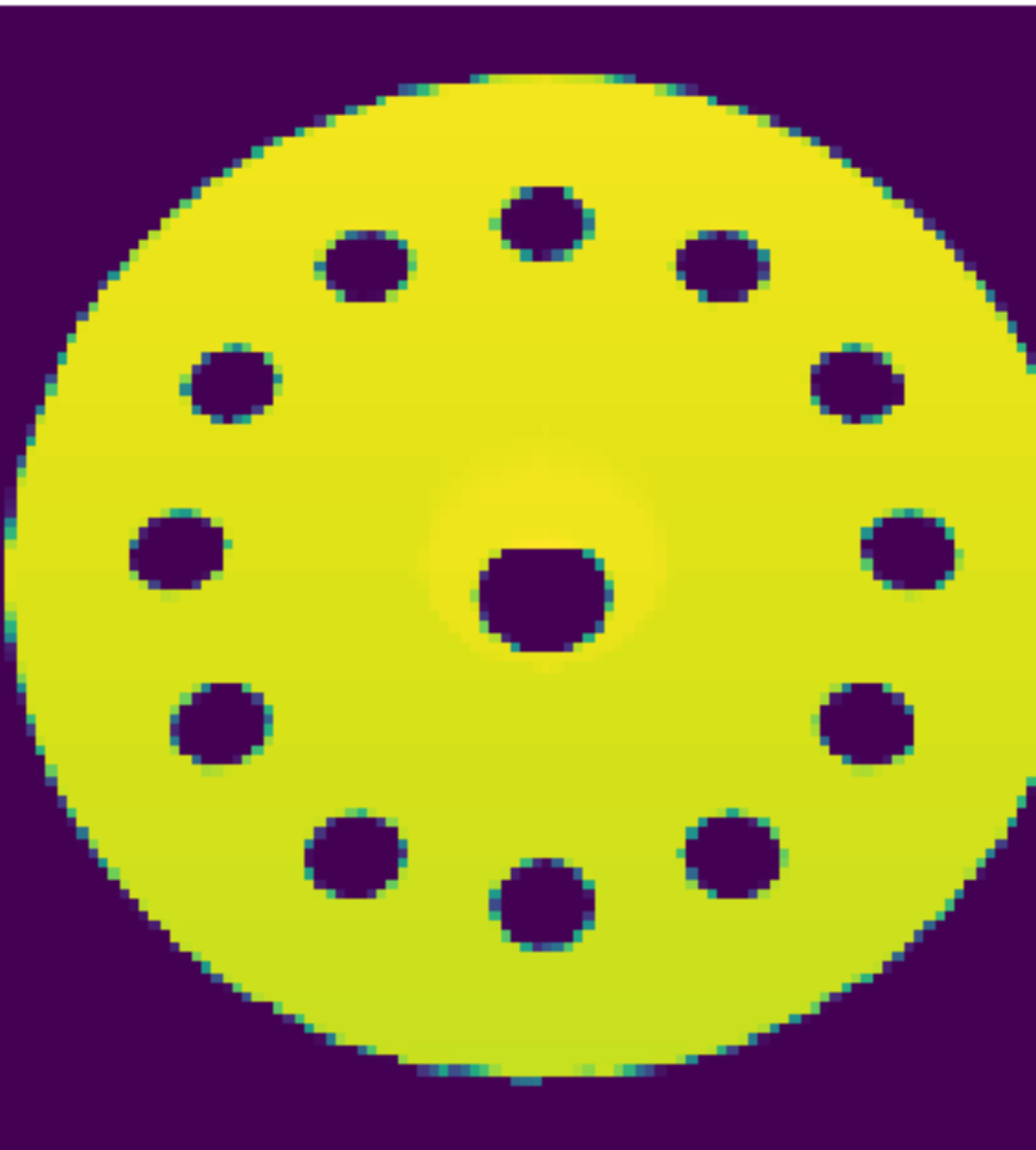}  &
			\fcolorbox{green}{white}{\includegraphics[trim={9cm 4cm 9cm 4cm}, clip = true,width=0.12\linewidth]{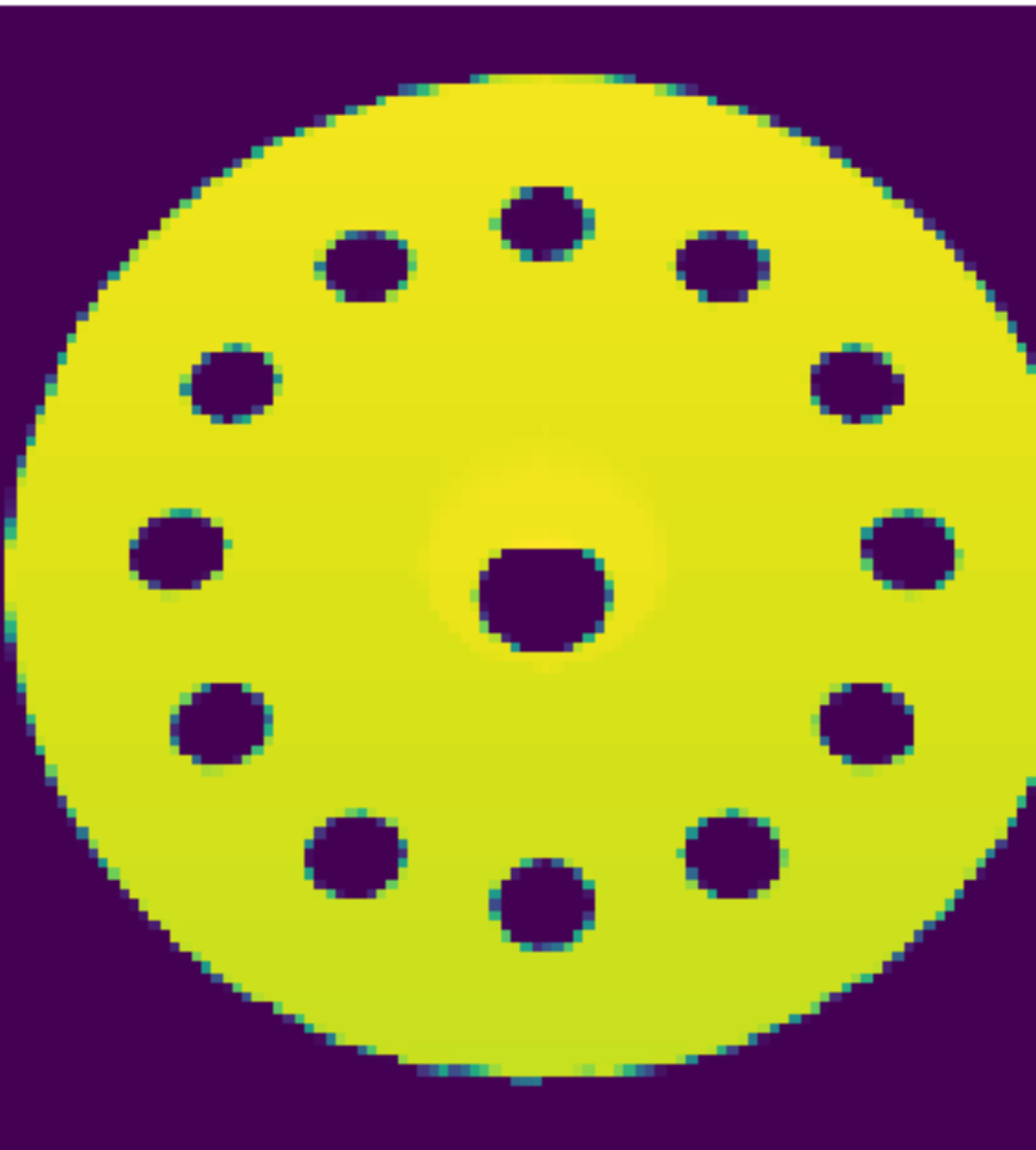}}  &
			\fcolorbox{green}{white}{\includegraphics[trim={9cm 4cm 9cm 4cm}, clip = true,width=0.12\linewidth]{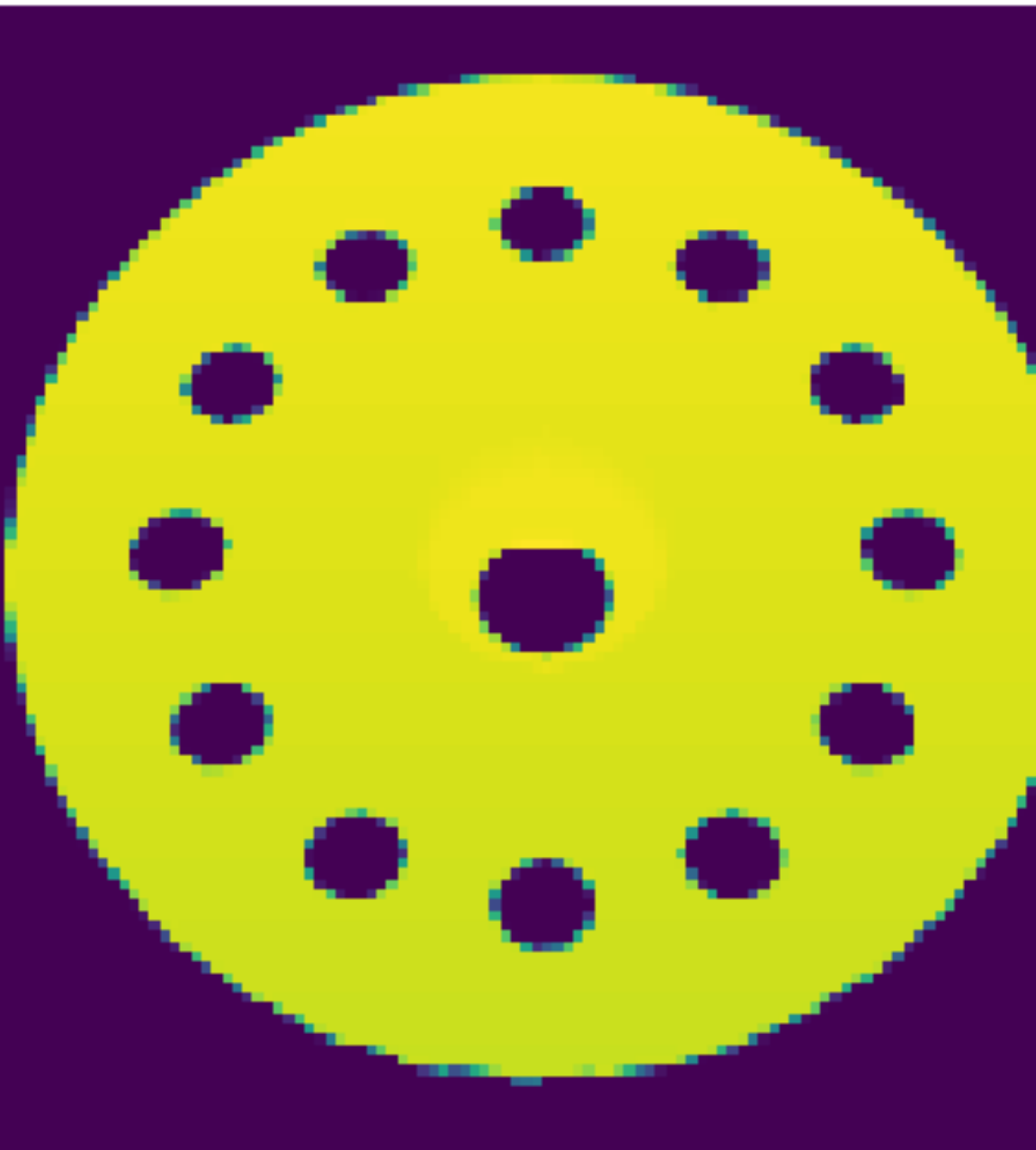}}  &
			\includegraphics[trim={9cm 4cm 9cm 4cm}, clip = true,width=0.12\linewidth]{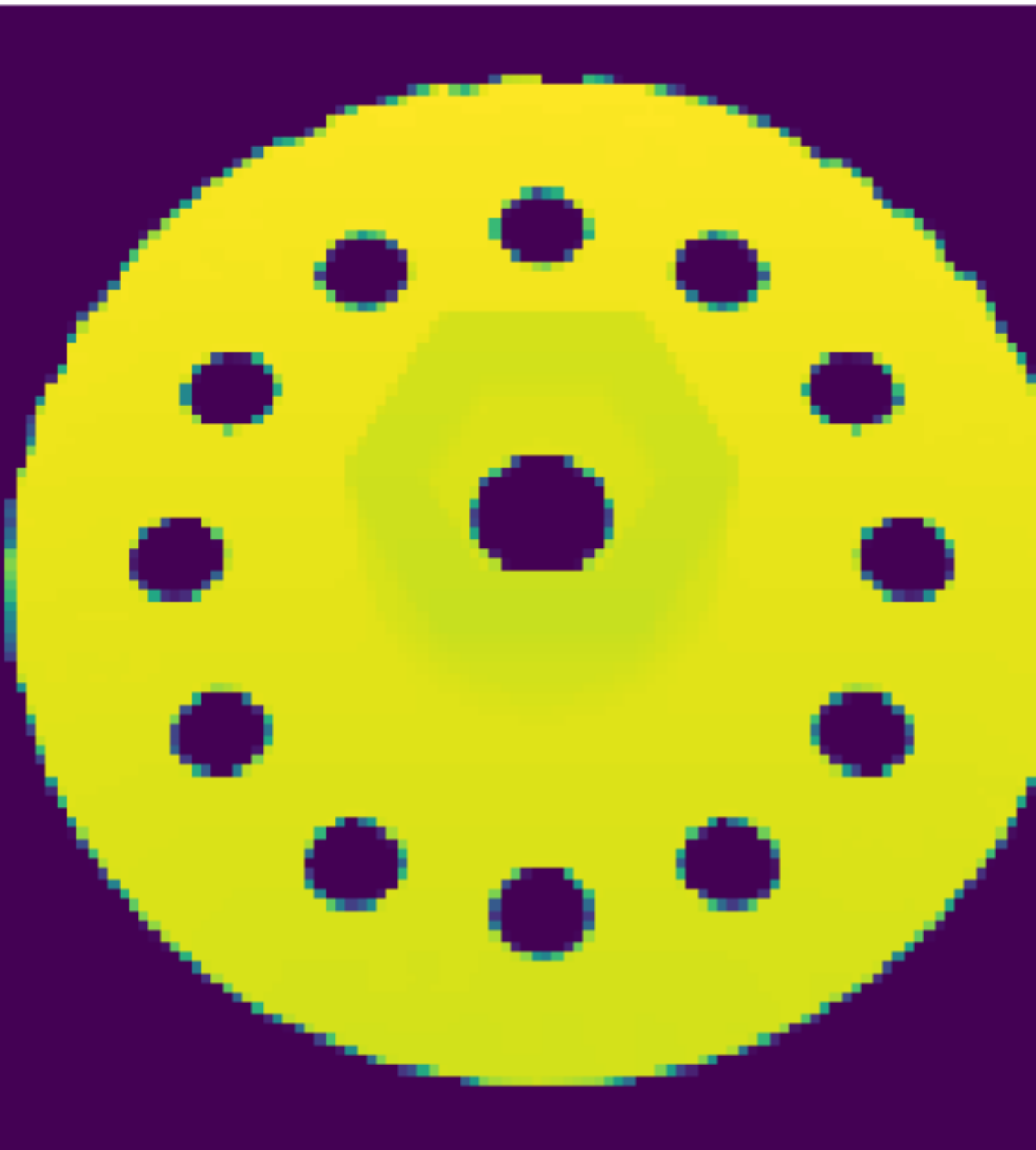}  &
			\includegraphics[trim={9cm 4cm 9cm 4cm}, clip = true,width=0.12\linewidth]{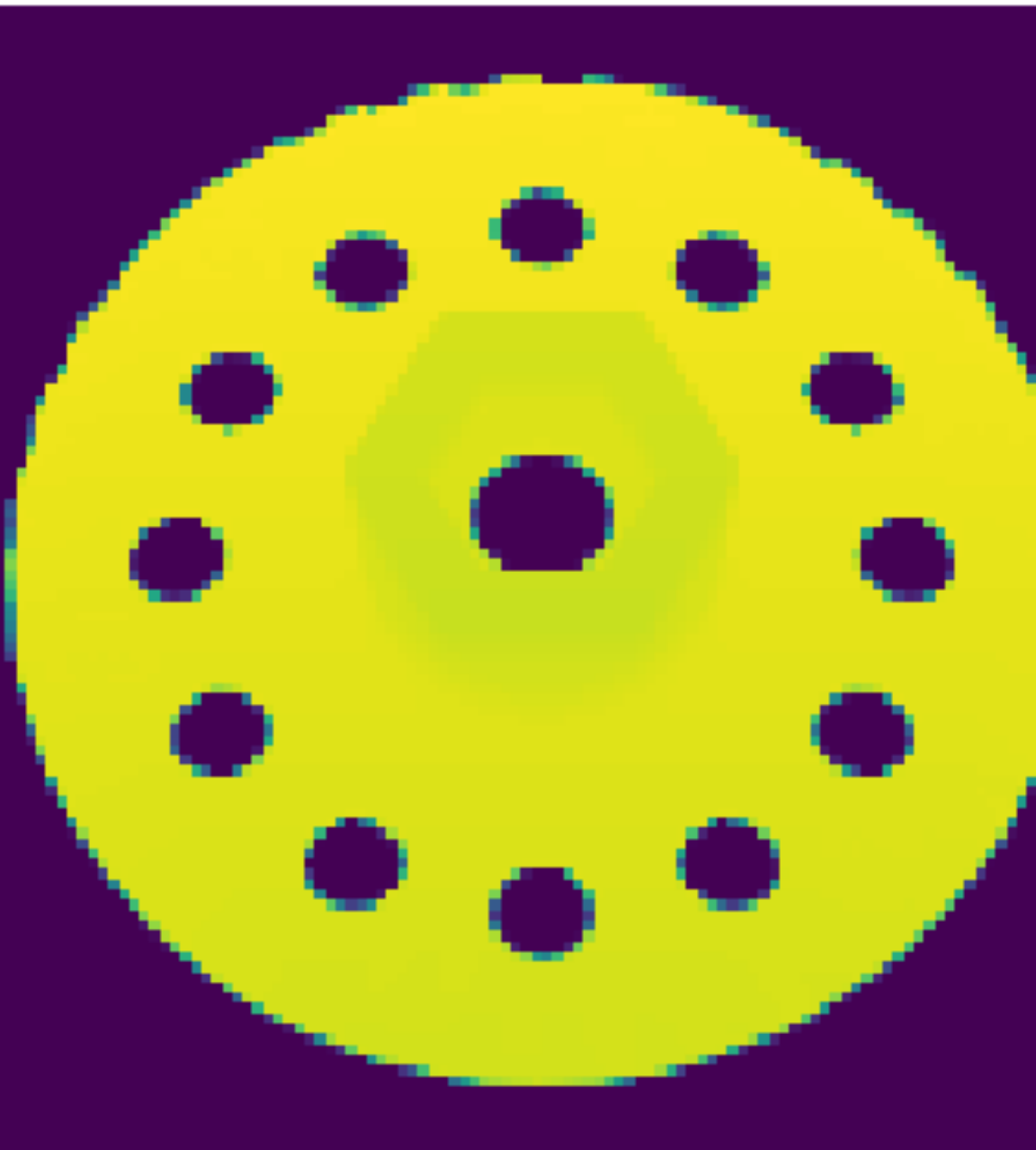}  & 
			\raisebox{2\height}{\LARGE 5.94$^{\circ}$}  \\
			\hline 
			\includegraphics[trim={9cm 4cm 9cm 4cm}, clip = true,width=0.12\linewidth]{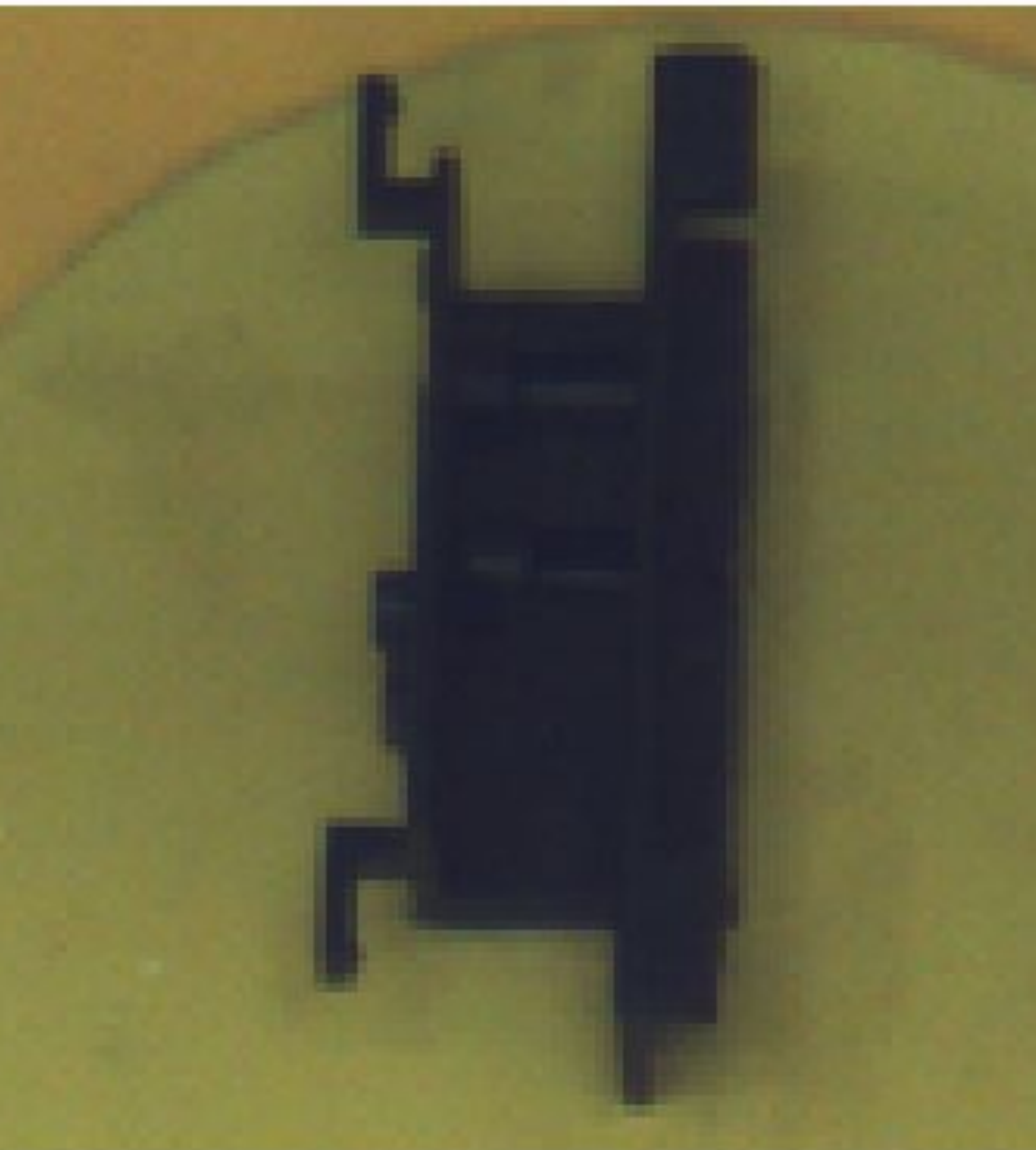} &
			\includegraphics[trim={9cm 4cm 9cm 4cm}, clip = true,width=0.12\linewidth]{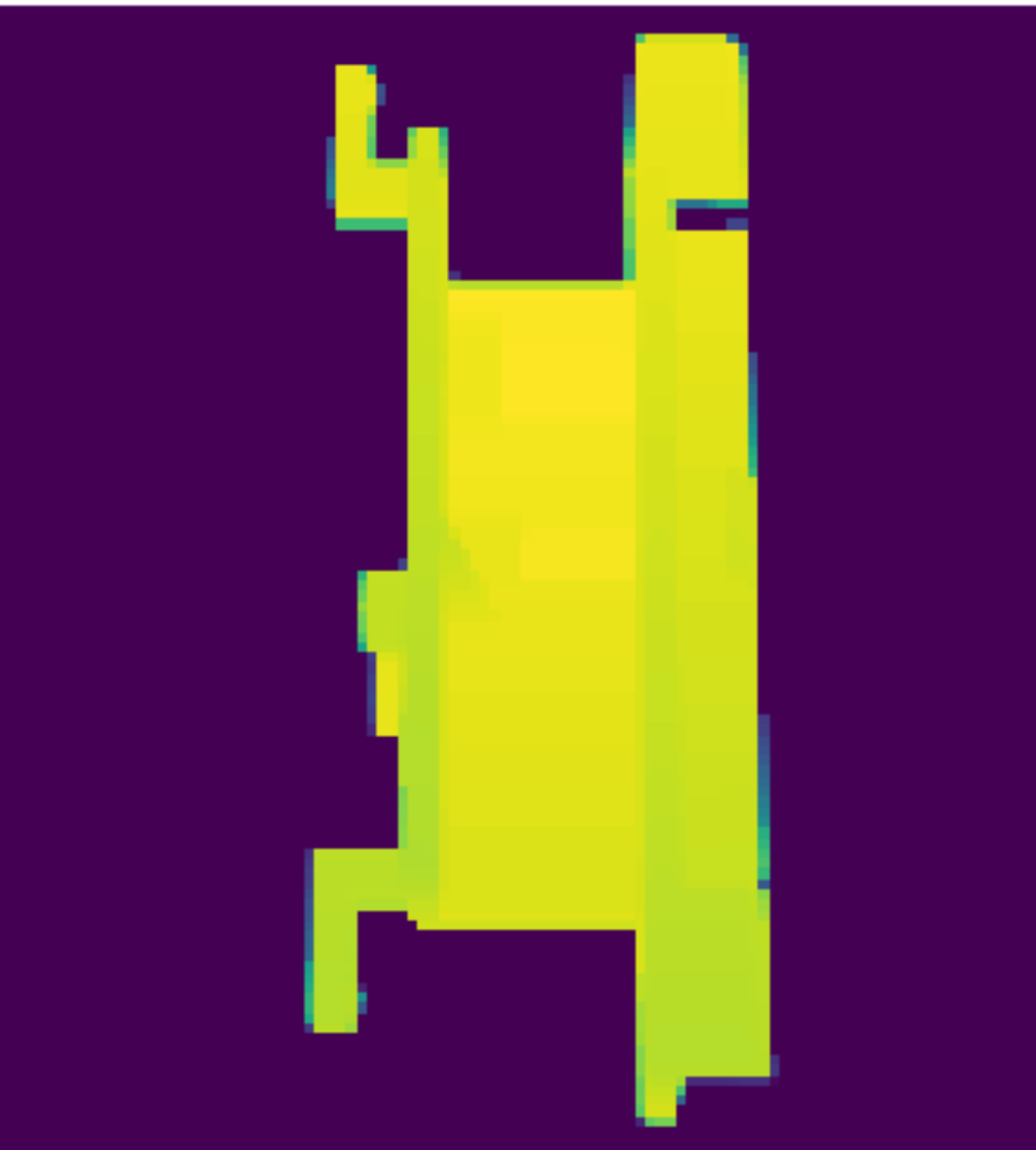} &
			\includegraphics[trim={9cm 4cm 9cm 4cm}, clip = true,width=0.12\linewidth]{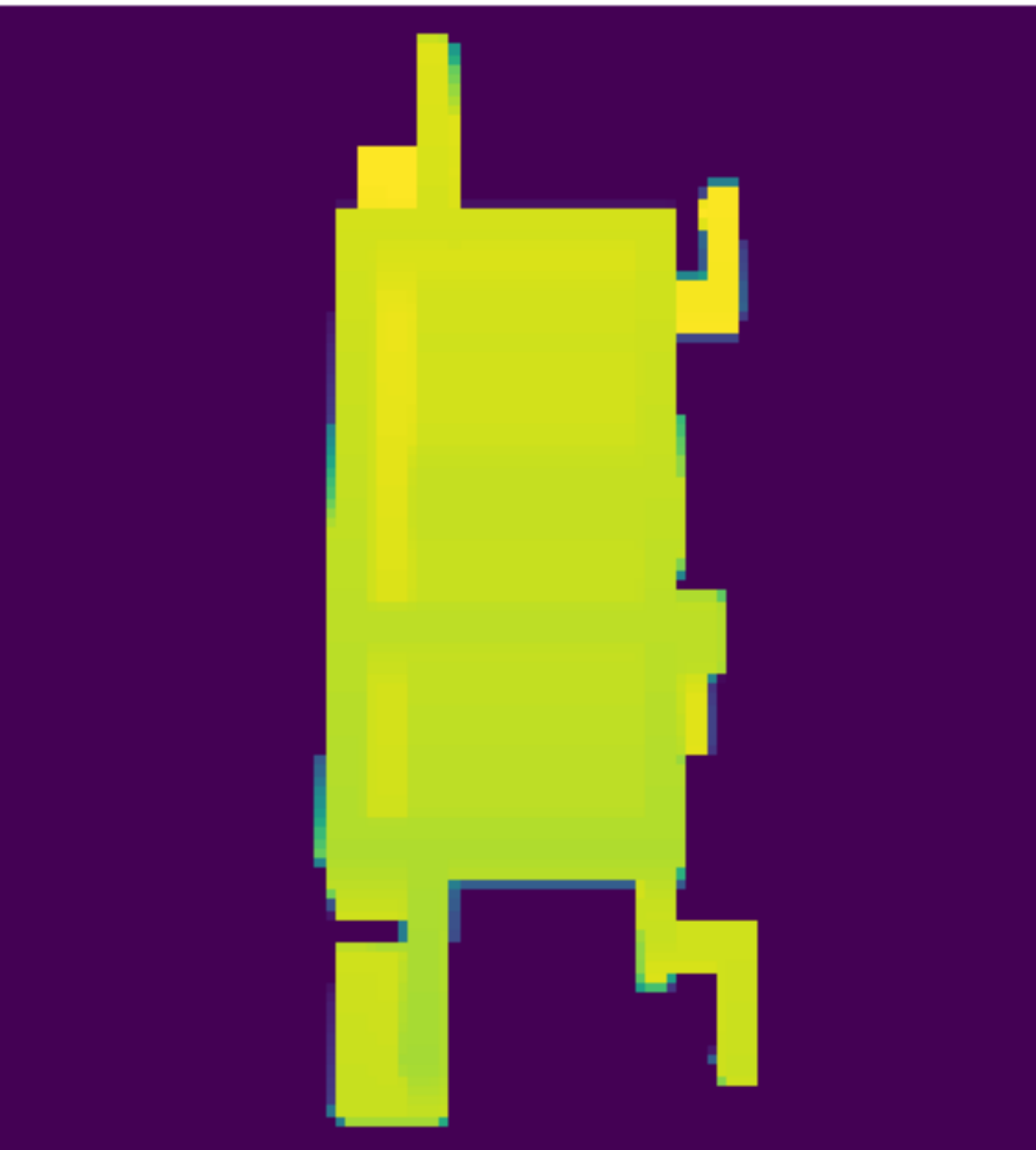} &
			\includegraphics[trim={9cm 4cm 9cm 4cm}, clip = true,width=0.12\linewidth]{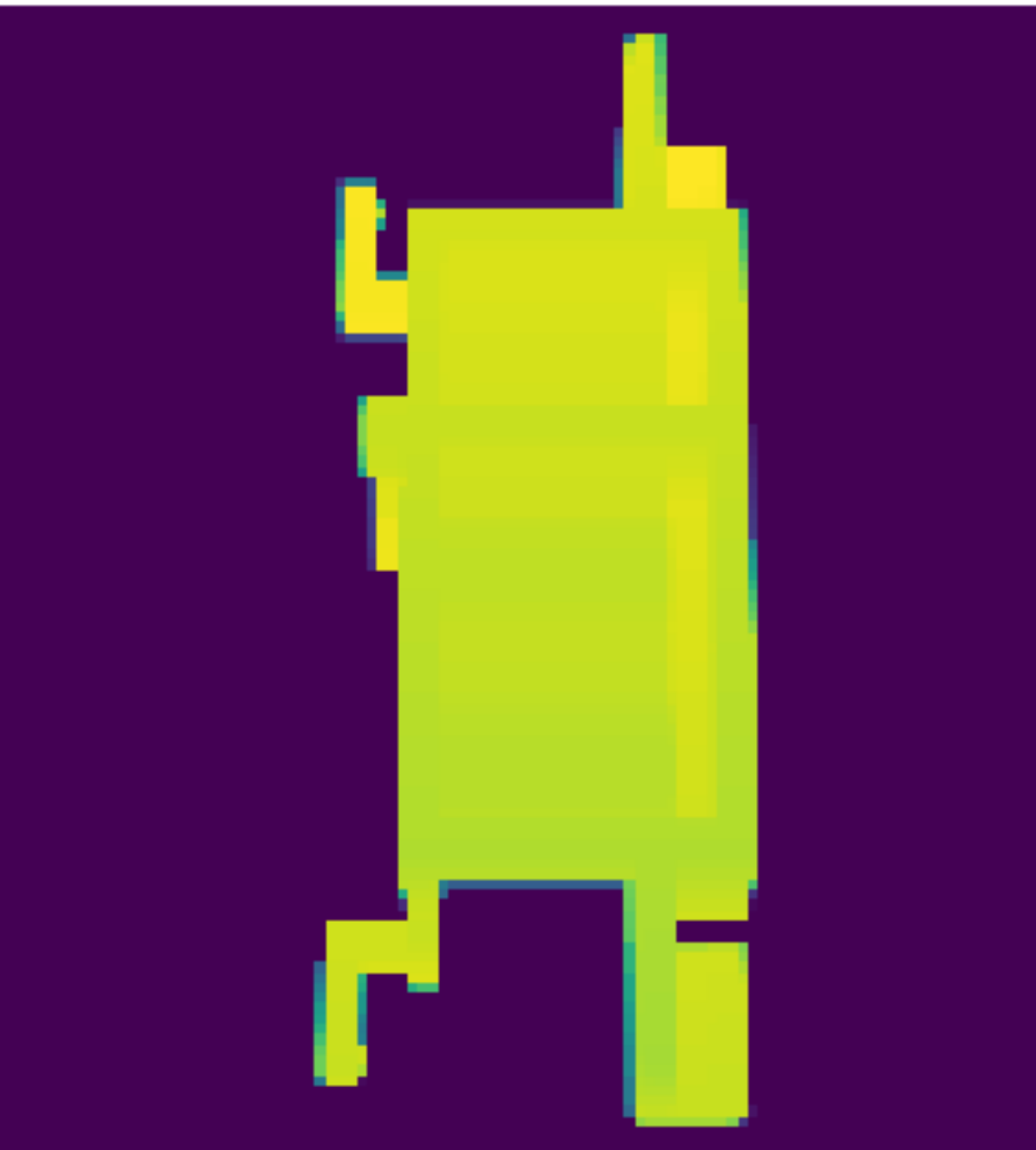} &
			\fcolorbox{green}{white}{\includegraphics[trim={9cm 4cm 9cm 4cm}, clip = true,width=0.12\linewidth]{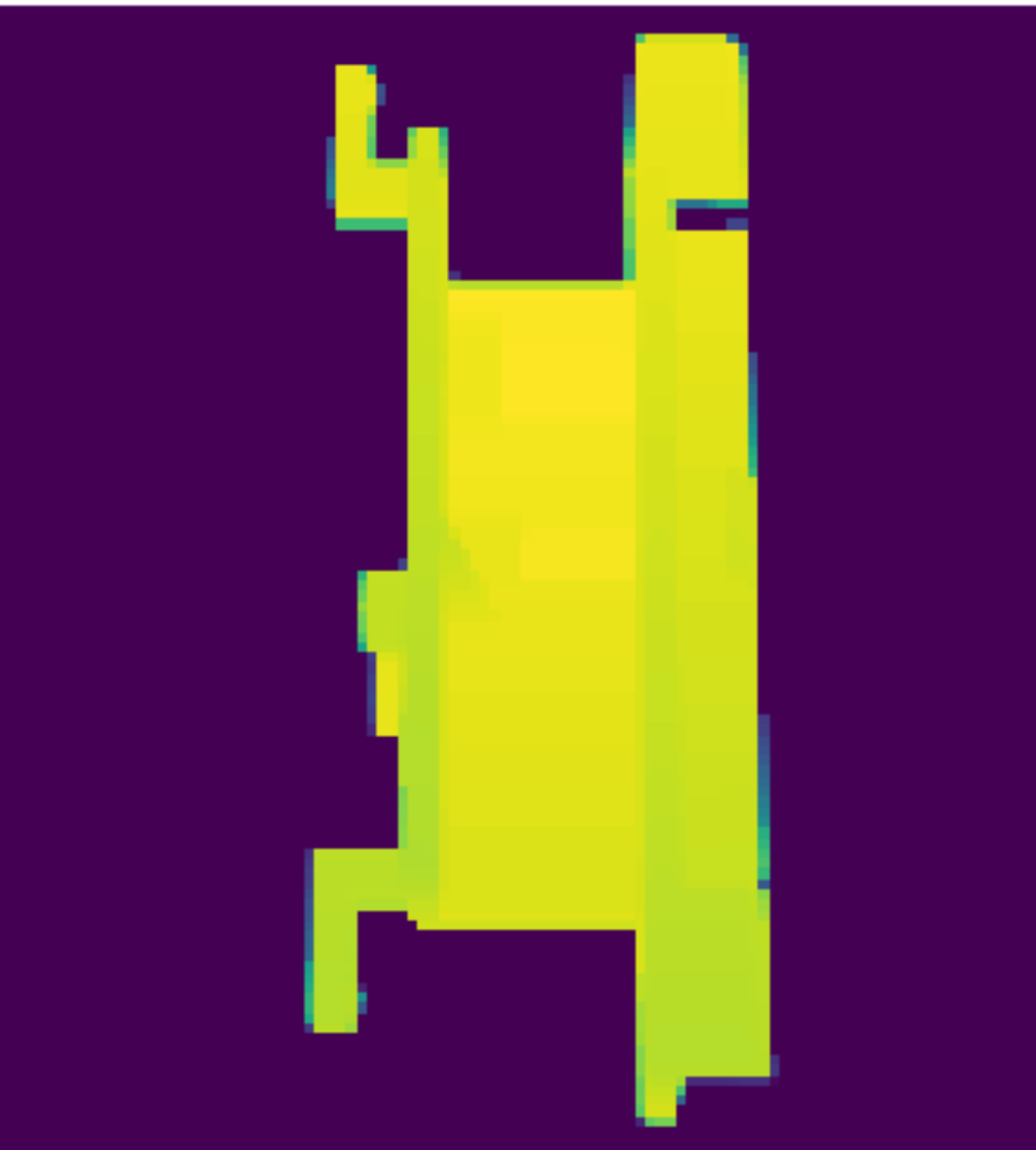}} &
			\includegraphics[trim={9cm 4cm 9cm 4cm}, clip = true,width=0.12\linewidth]{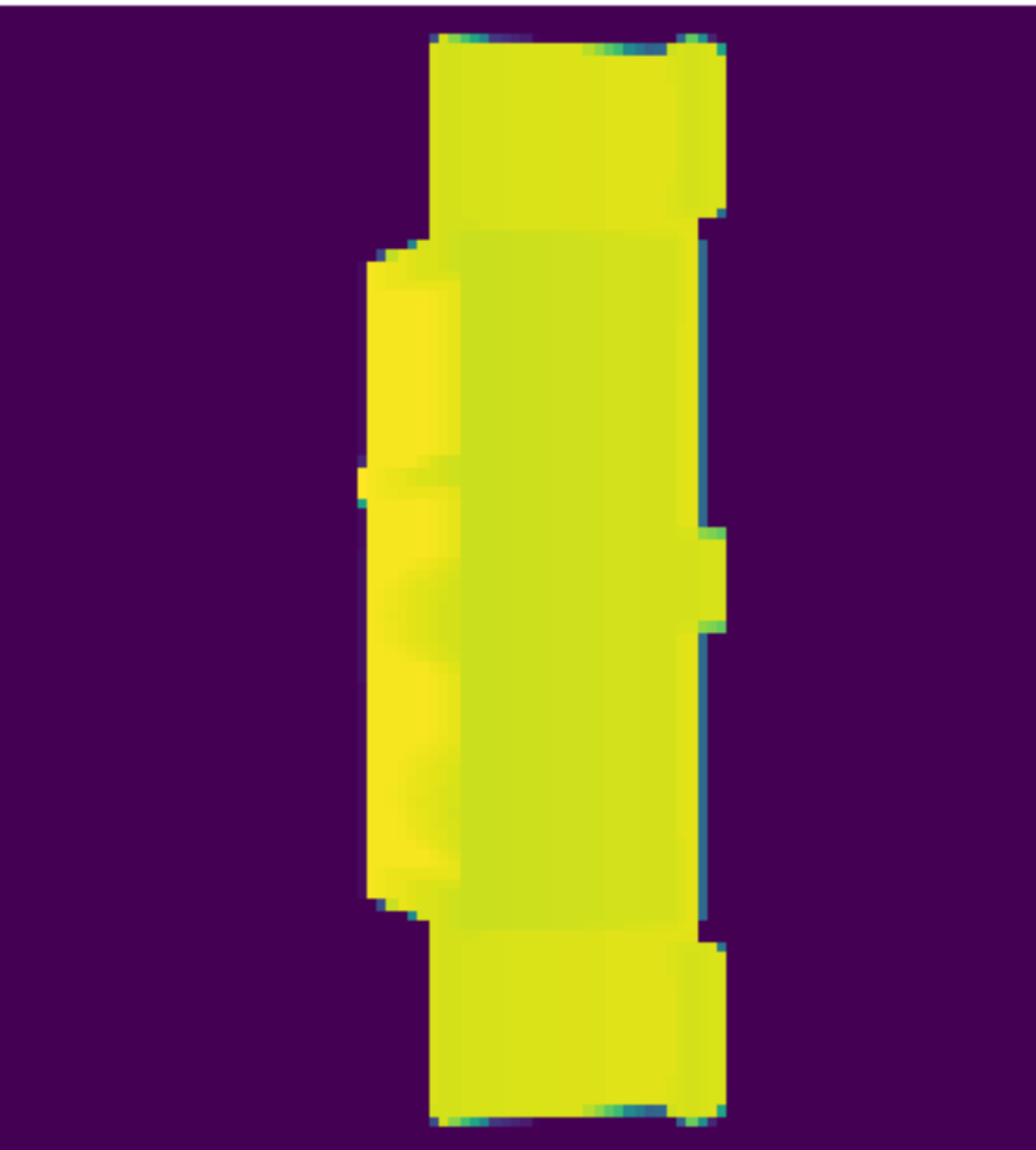} & 
			\raisebox{2\height}{\LARGE 6.37$^{\circ}$} \\ 		
			\hline 
			\includegraphics[trim={9cm 4cm 9cm 4cm}, clip = true,width=0.12\linewidth]{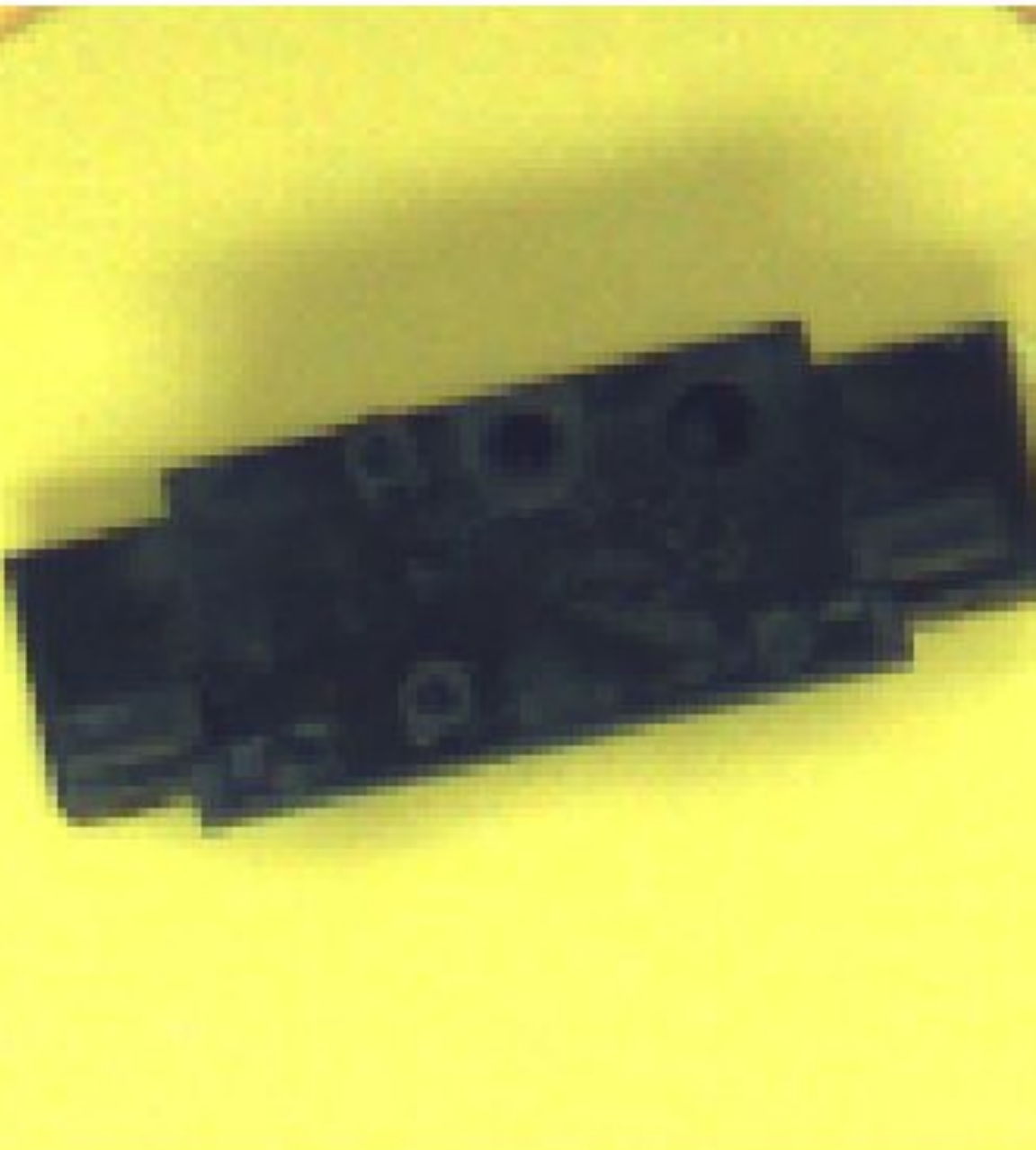} & 
			\includegraphics[trim={9cm 4cm 9cm 4cm}, clip = true,width=0.12\linewidth]{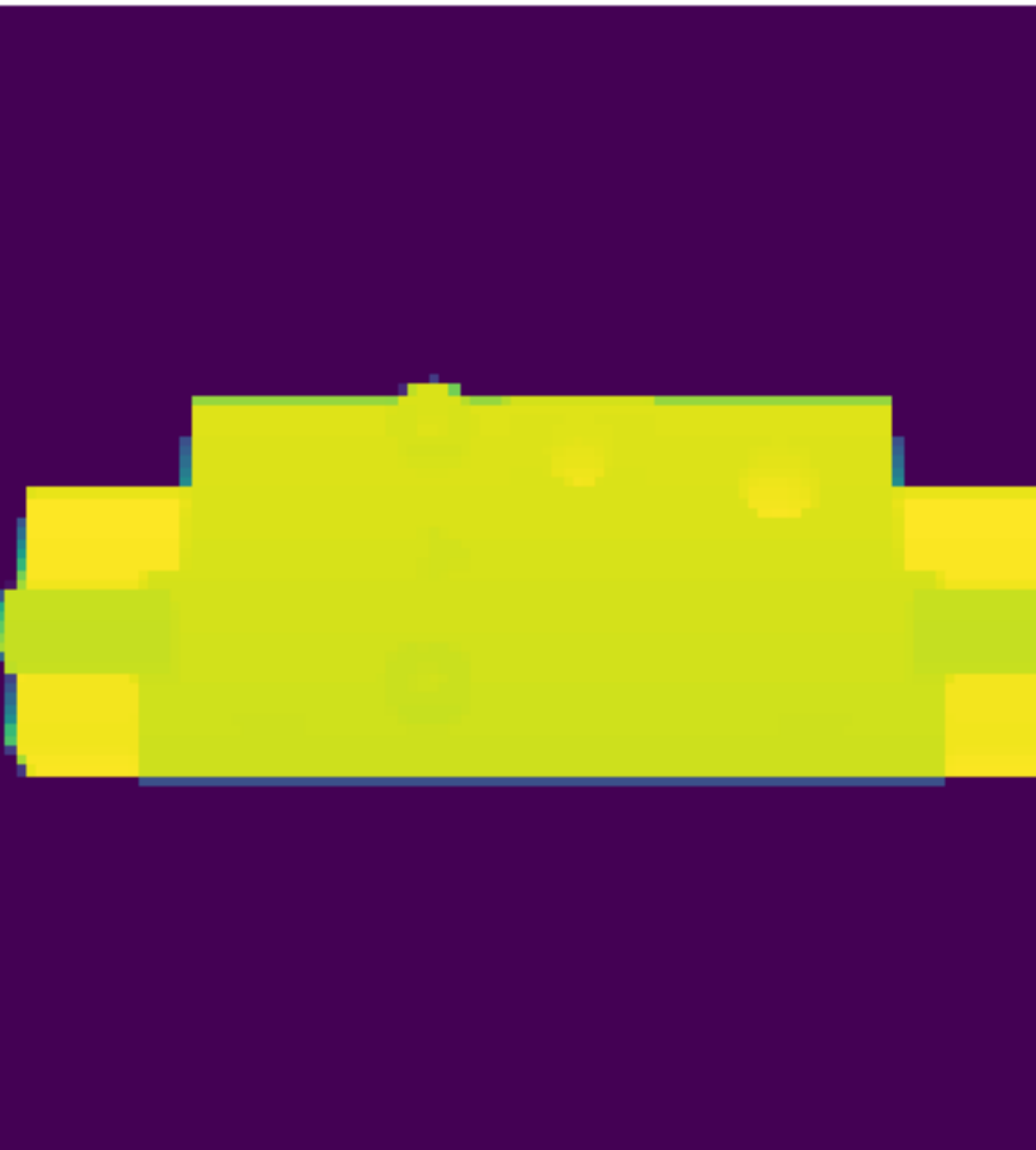}  &
			\includegraphics[trim={9cm 4cm 9cm 4cm}, clip = true,width=0.12\linewidth]{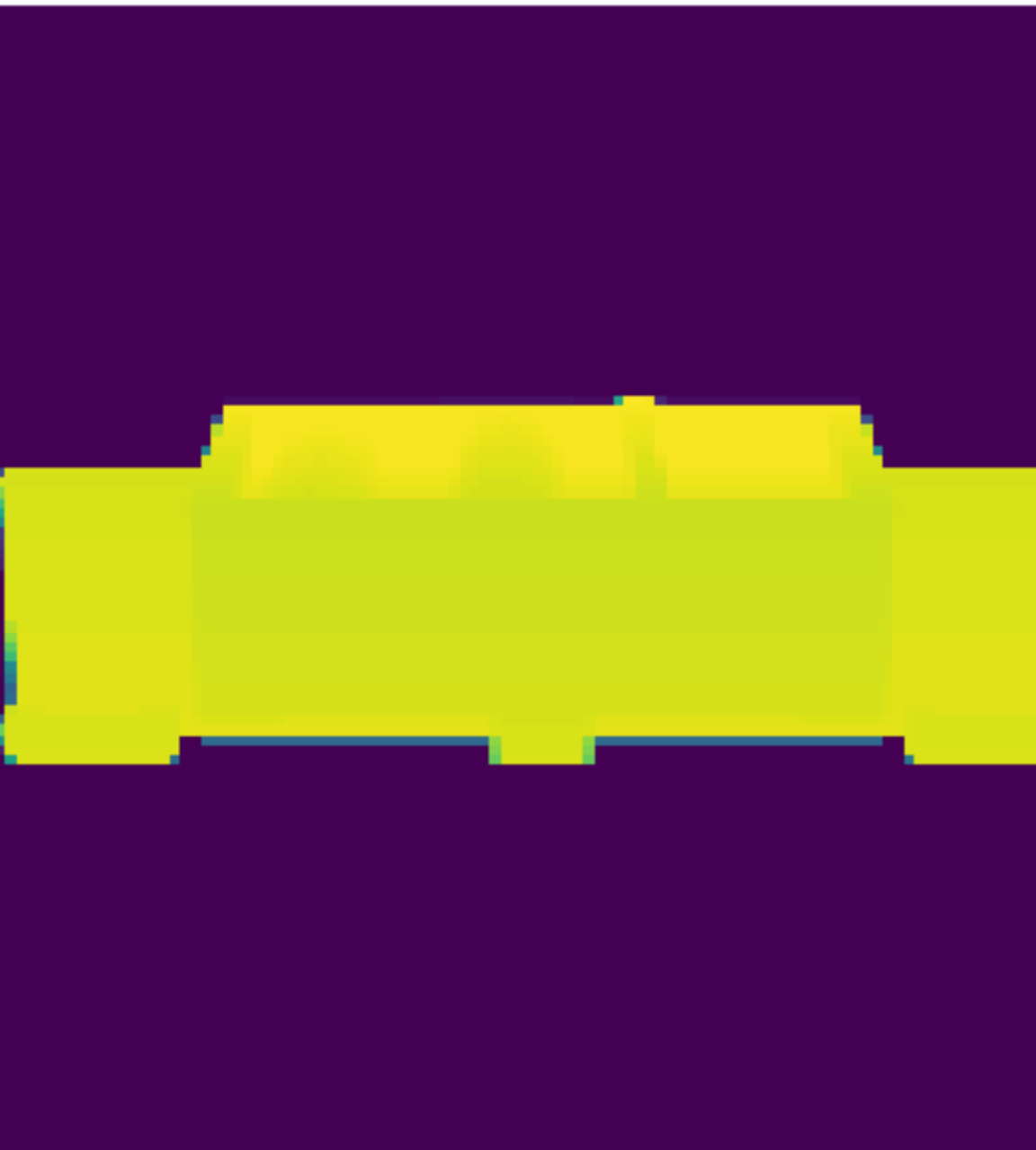}  &
			\includegraphics[trim={9cm 4cm 9cm 4cm}, clip = true,width=0.12\linewidth]{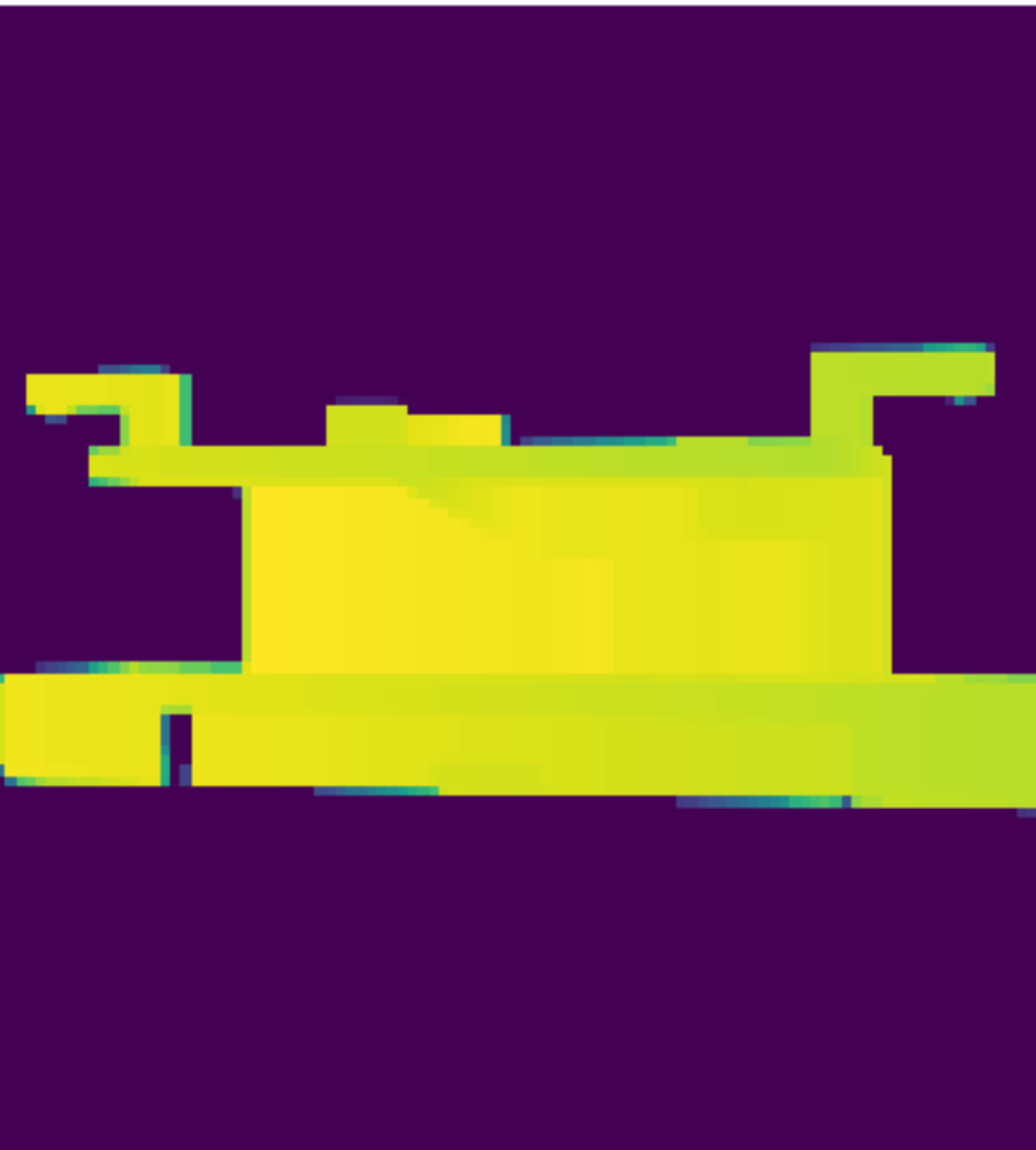}  &
			\fcolorbox{green}{white}{\includegraphics[trim={9cm 4cm 9cm 4cm}, clip = true,width=0.12\linewidth]{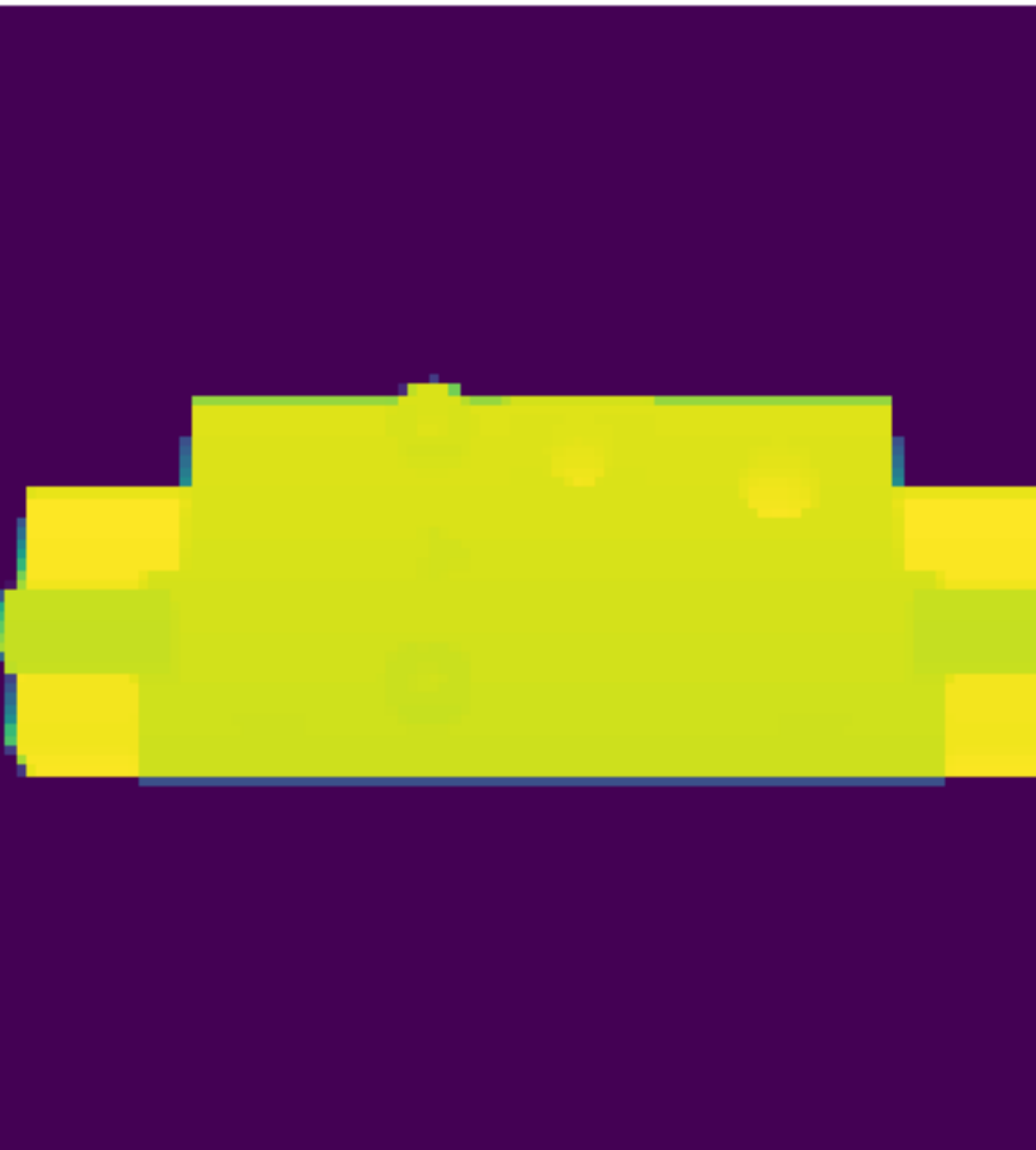}}  &
			\includegraphics[trim={9cm 4cm 9cm 4cm}, clip = true,width=0.12\linewidth]{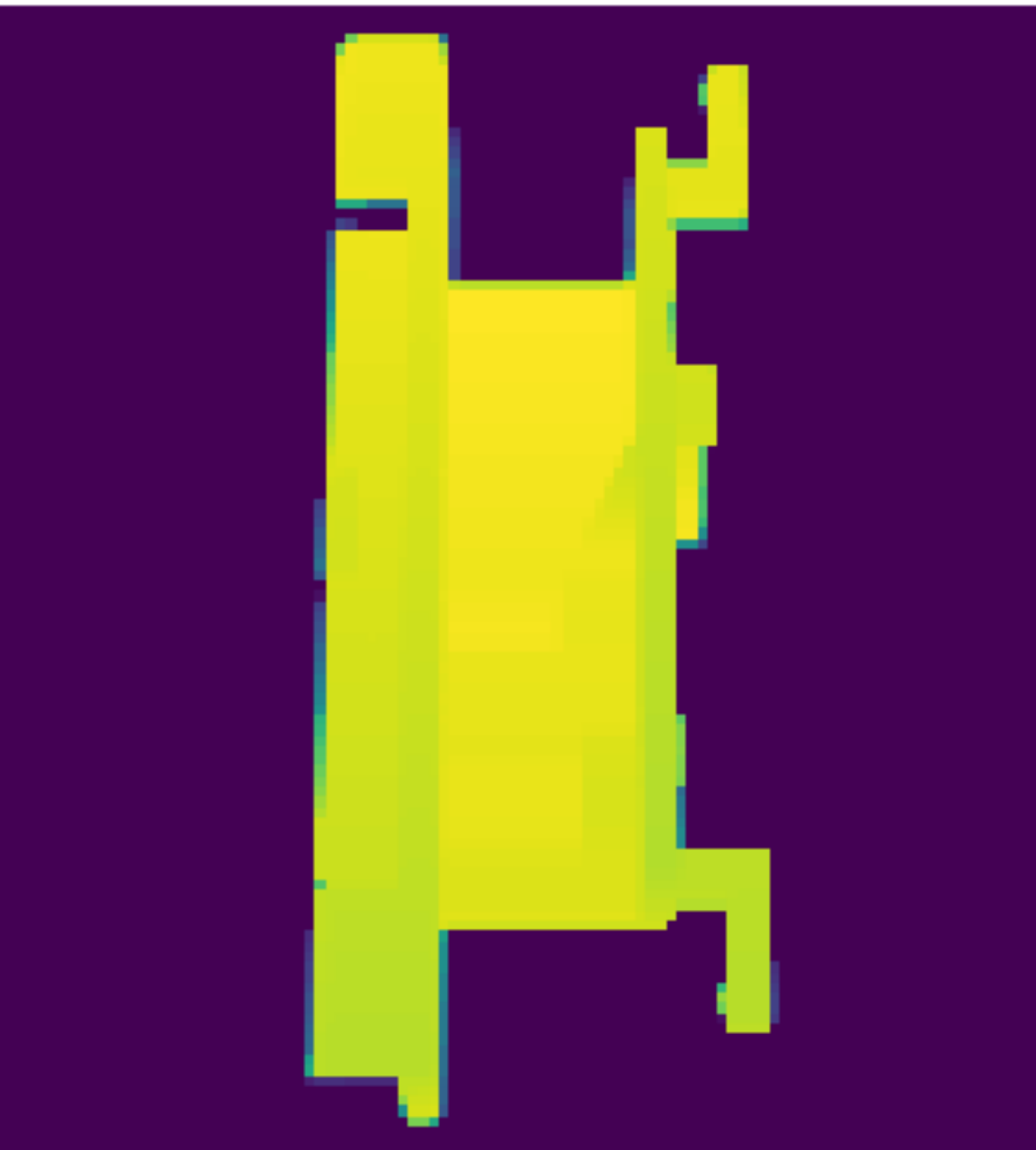}  & 
			\raisebox{2\height}{\LARGE 8.21$^{\circ}$}  \\
			\hline 
			\includegraphics[trim={9cm 4cm 9cm 4cm}, clip = true,width=0.12\linewidth]{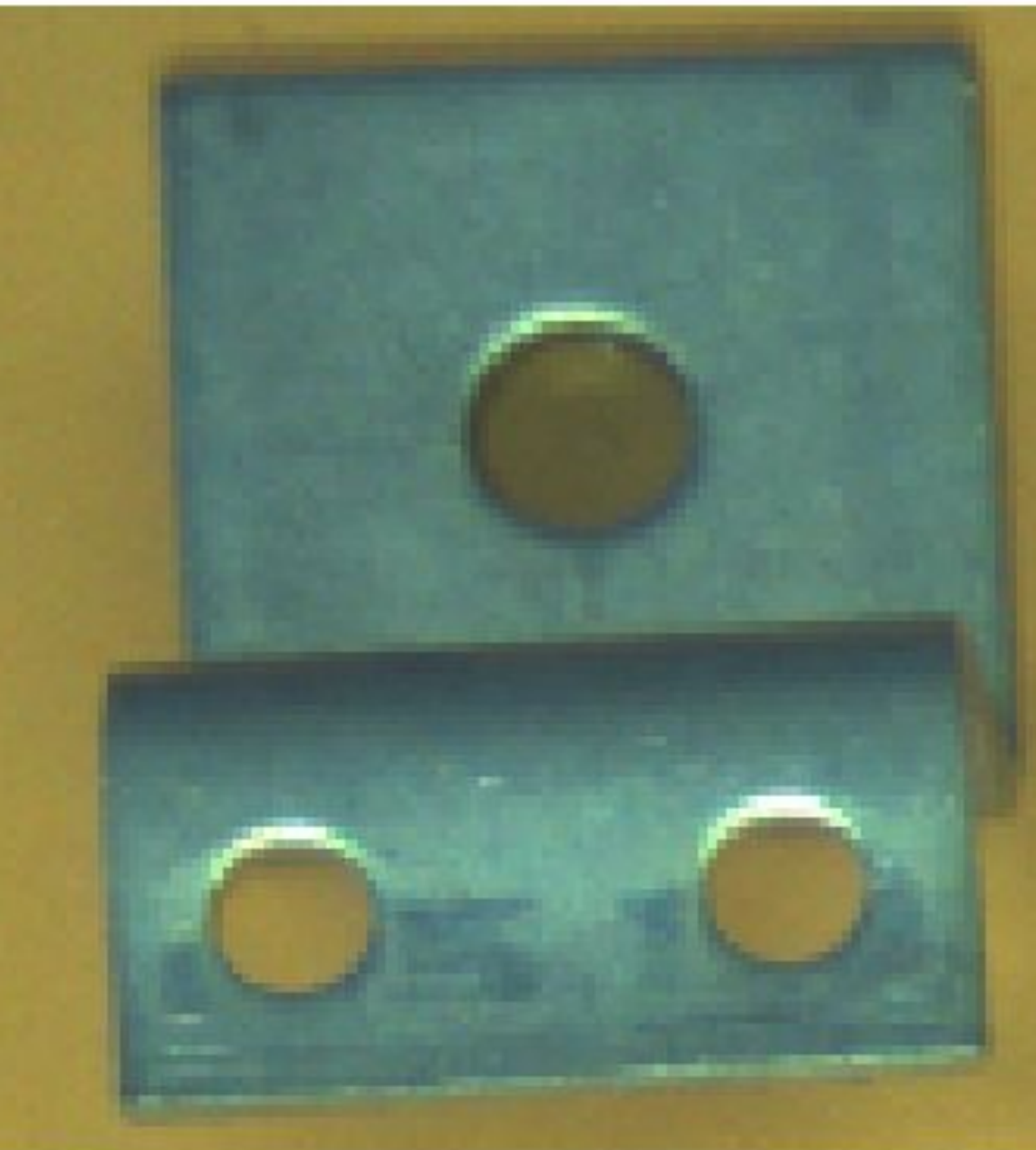} &				   
			\includegraphics[trim={9cm 4cm 9cm 4cm}, clip = true,width=0.12\linewidth]{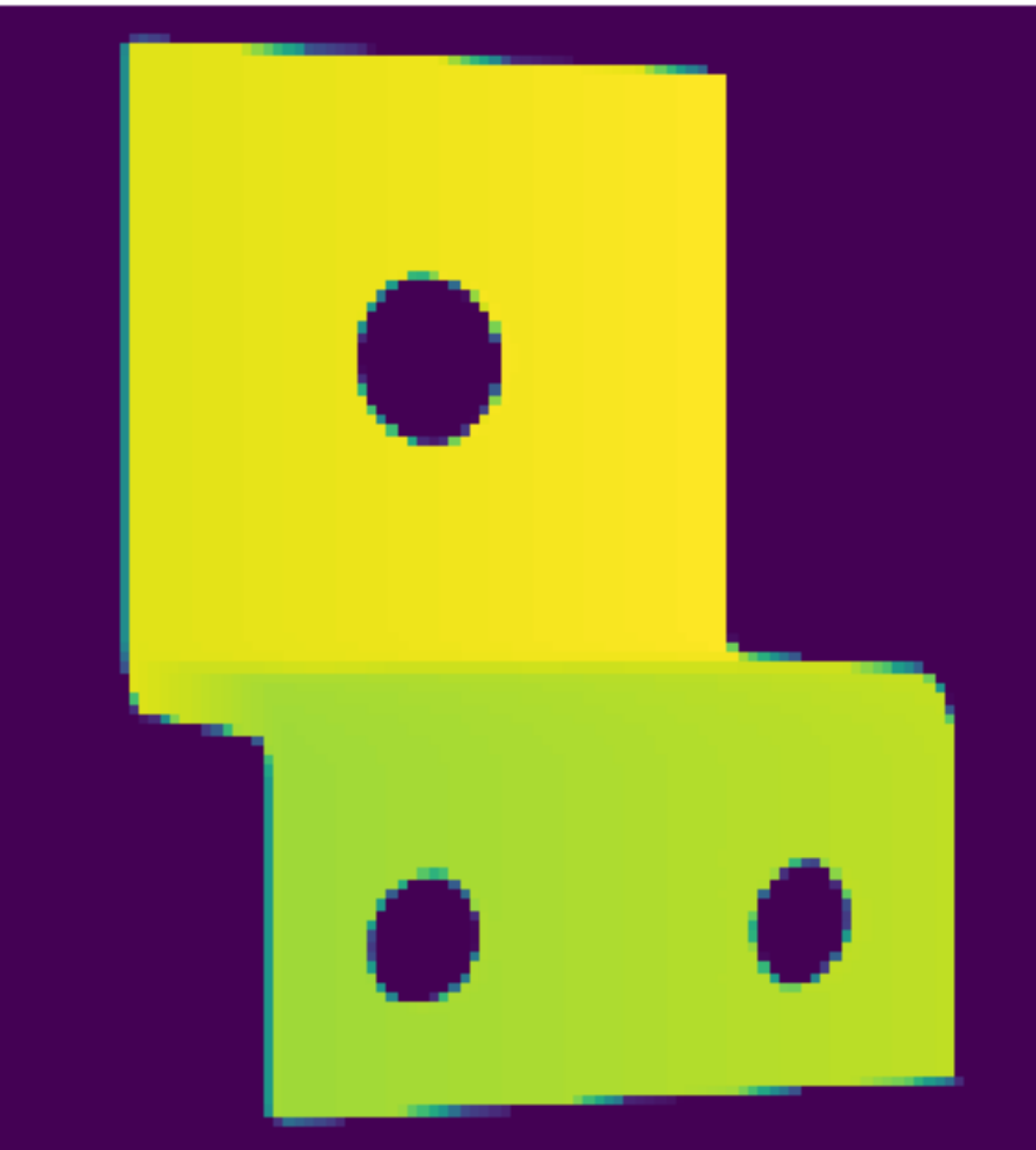} &			   
			\includegraphics[trim={9cm 4cm 9cm 4cm}, clip = true,width=0.12\linewidth]{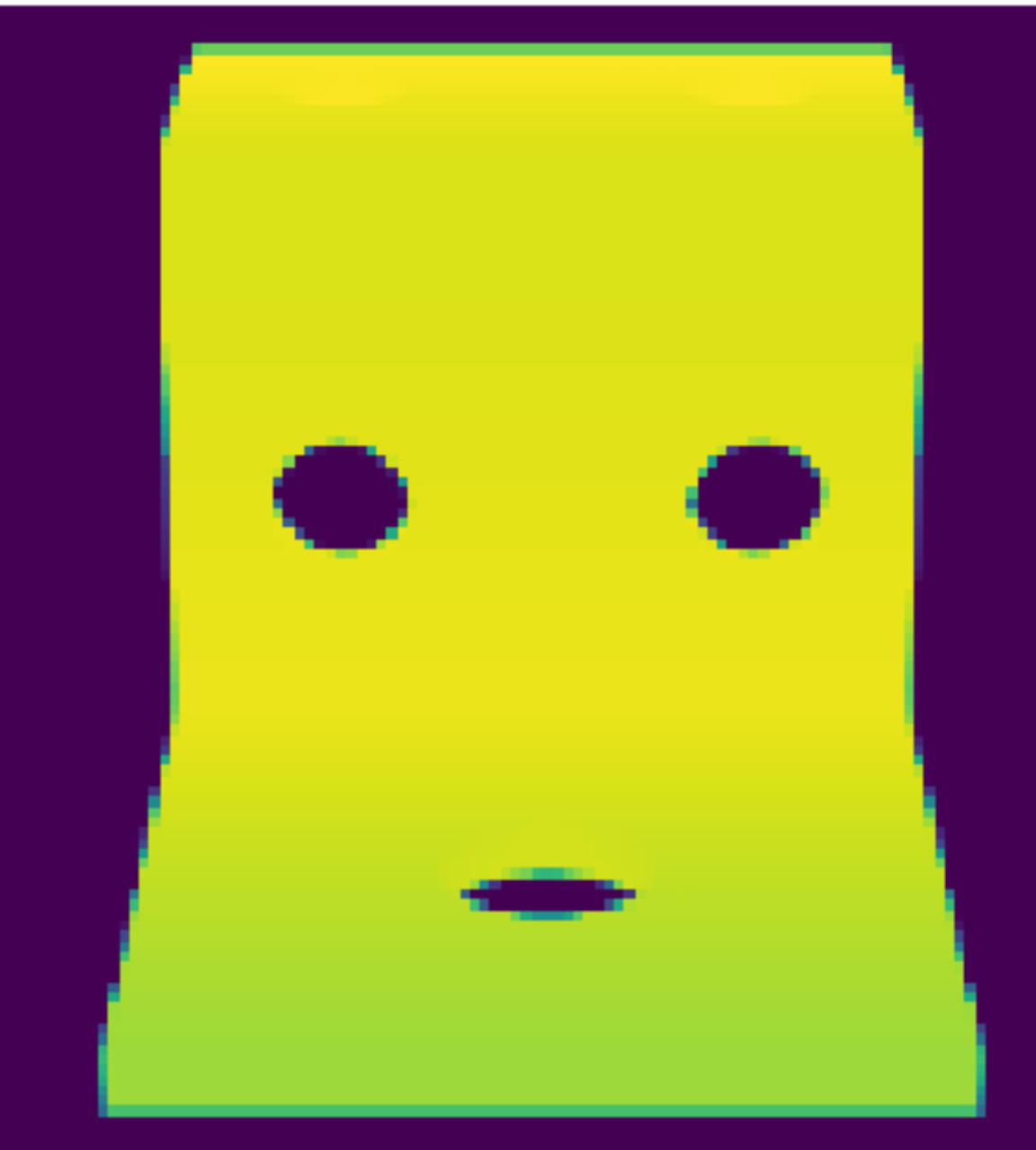} &		   
			\includegraphics[trim={9cm 4cm 9cm 4cm}, clip = true,width=0.12\linewidth]{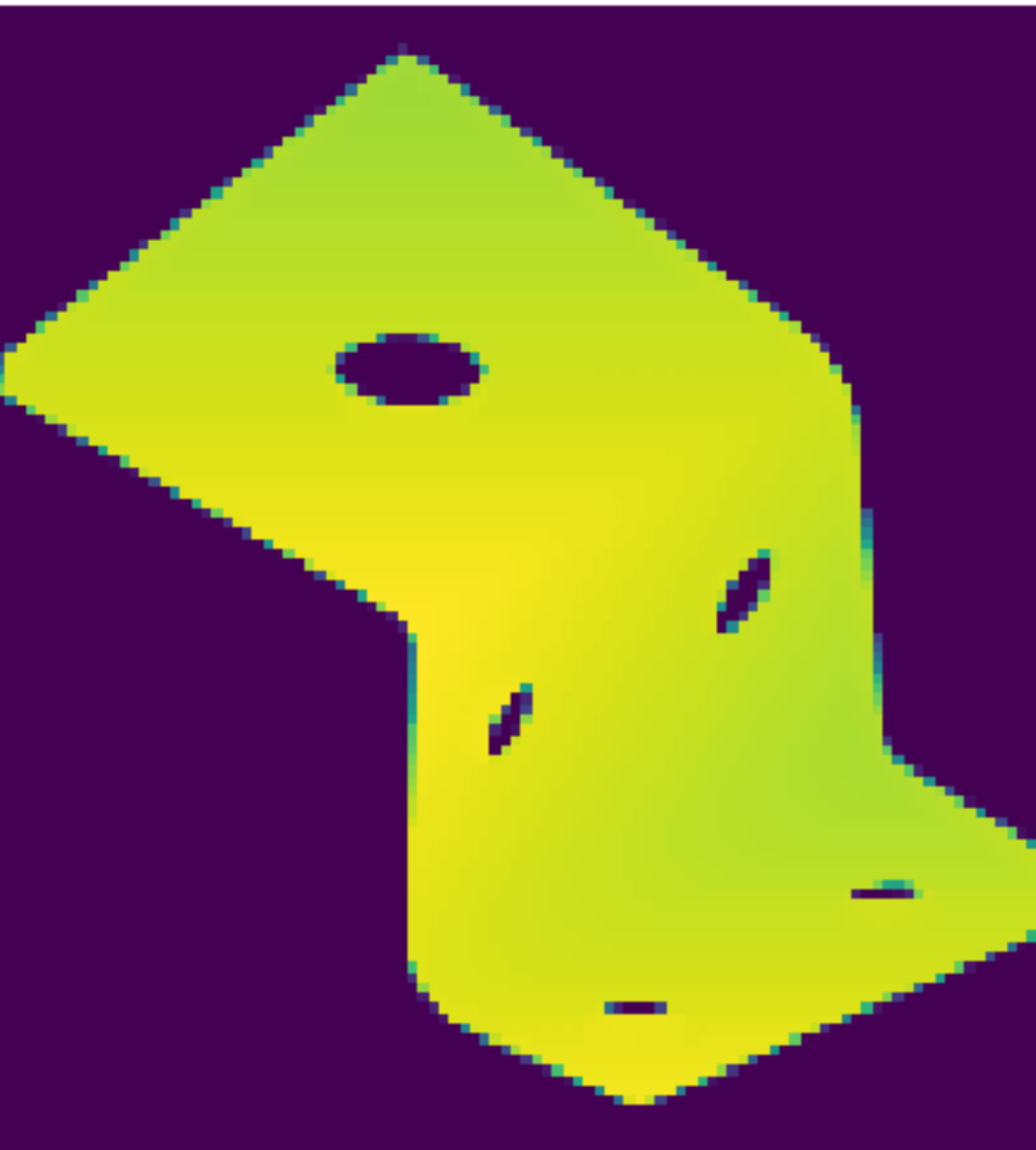} &	   
			\fcolorbox{green}{white}{\includegraphics[trim={9cm 4cm 9cm 4cm}, clip = true,width=0.12\linewidth]{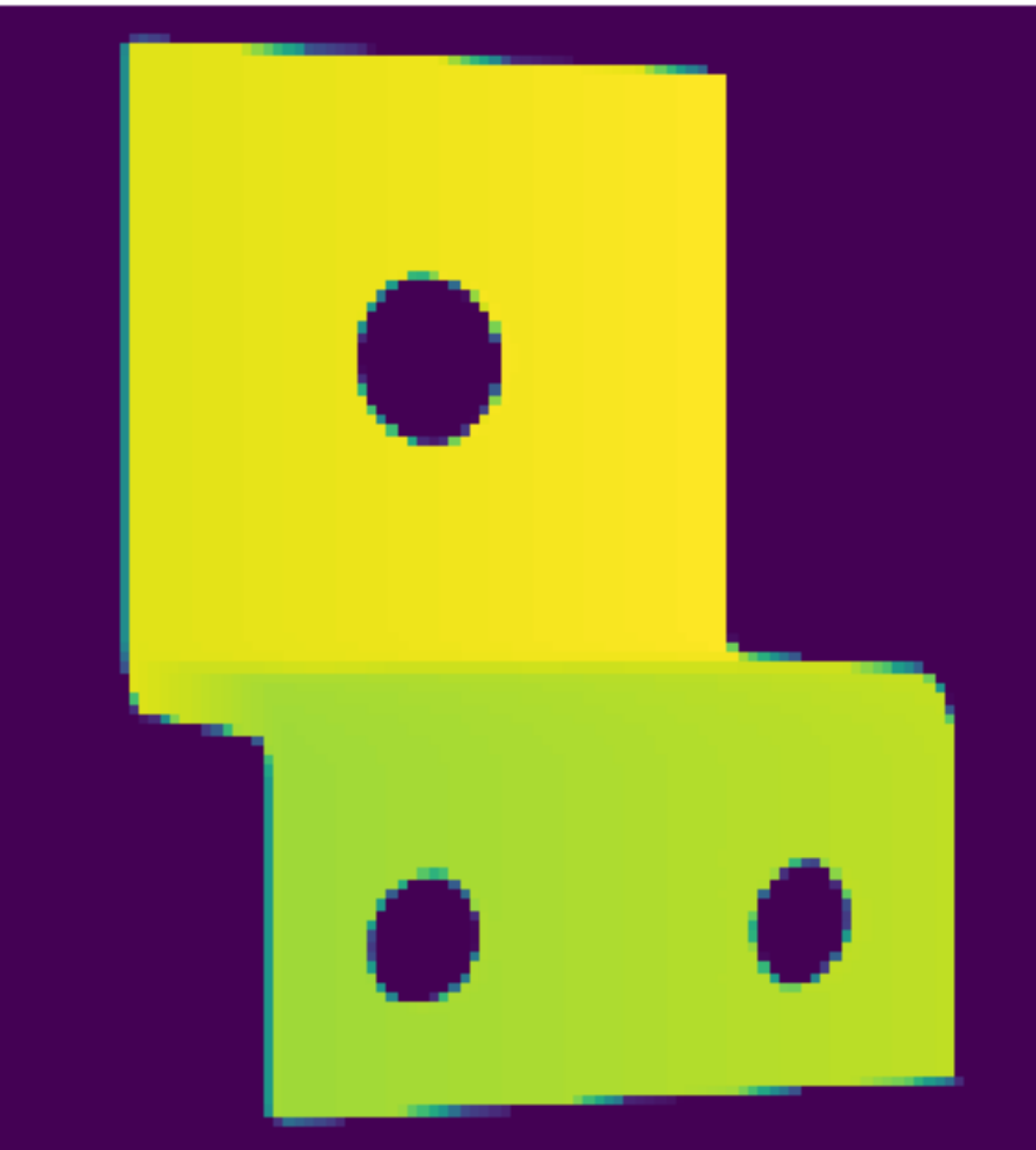}} &   
			\includegraphics[trim={9cm 4cm 9cm 4cm}, clip = true,width=0.12\linewidth]{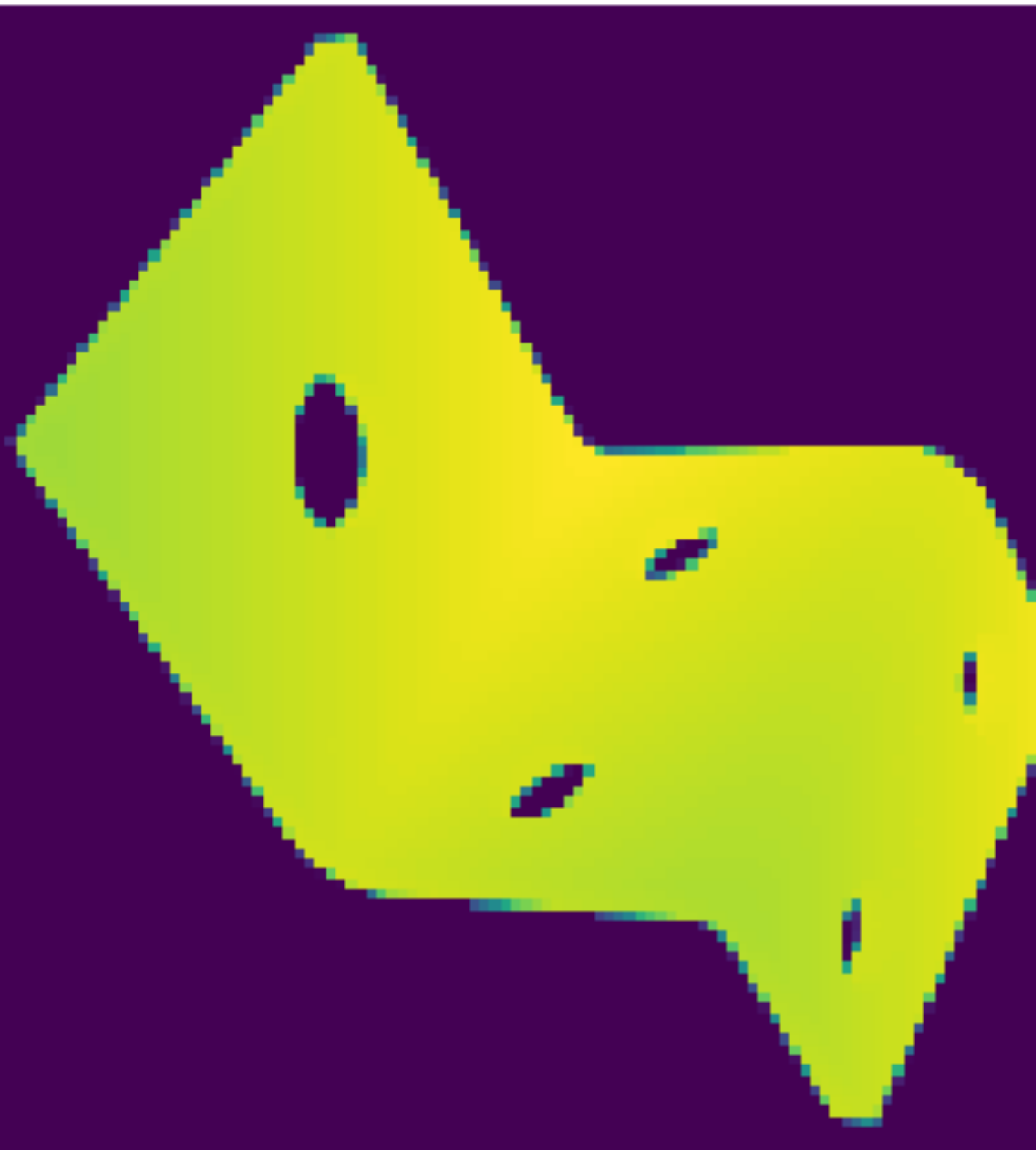}  & 
			\raisebox{2\height}{\LARGE 8.40$^{\circ}$} \\
			\hline 
			\includegraphics[trim={9cm 4cm 9cm 4cm}, clip = true,width=0.12\linewidth]{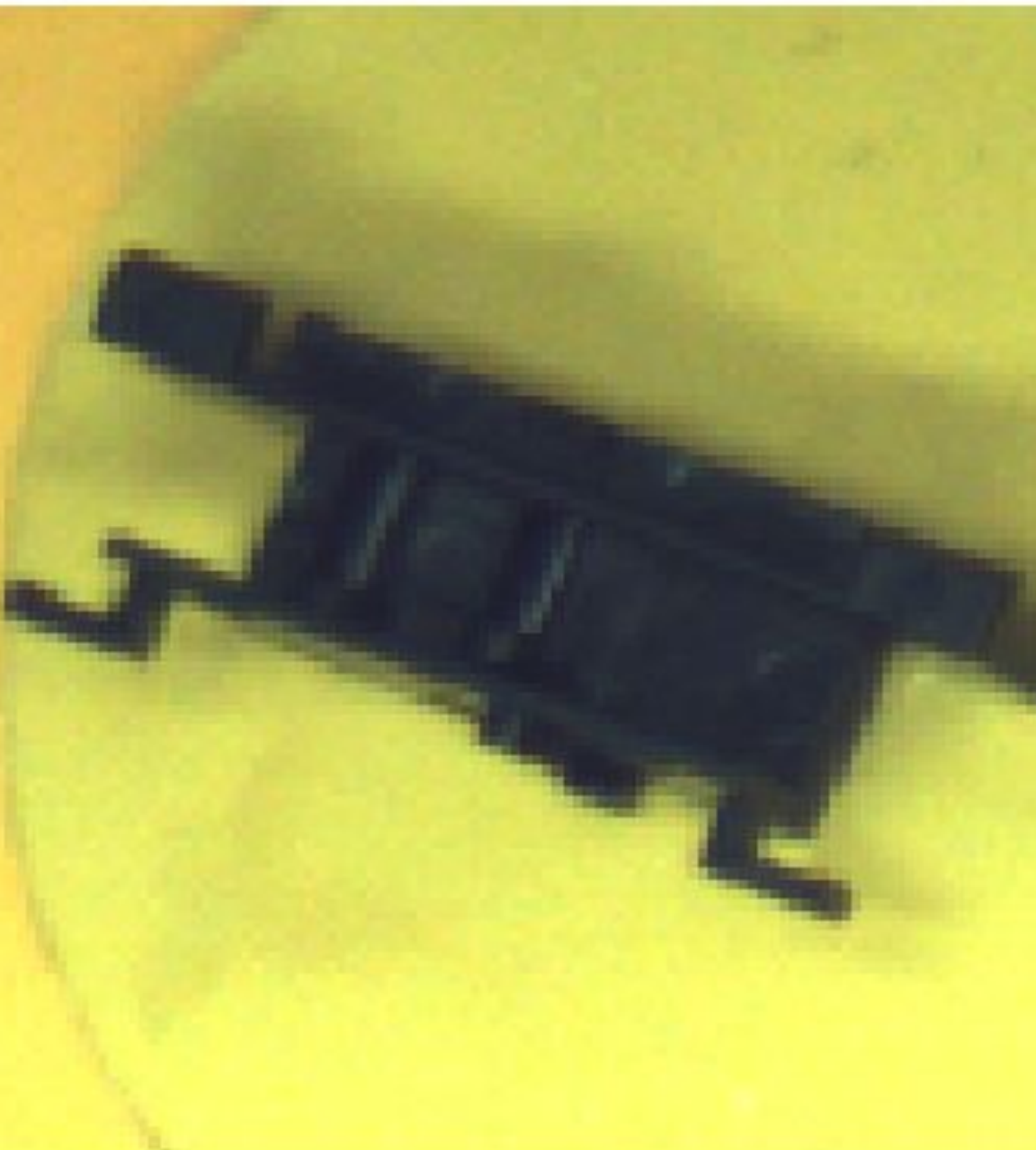}  &
			\includegraphics[trim={9cm 4cm 9cm 4cm}, clip = true,width=0.12\linewidth]{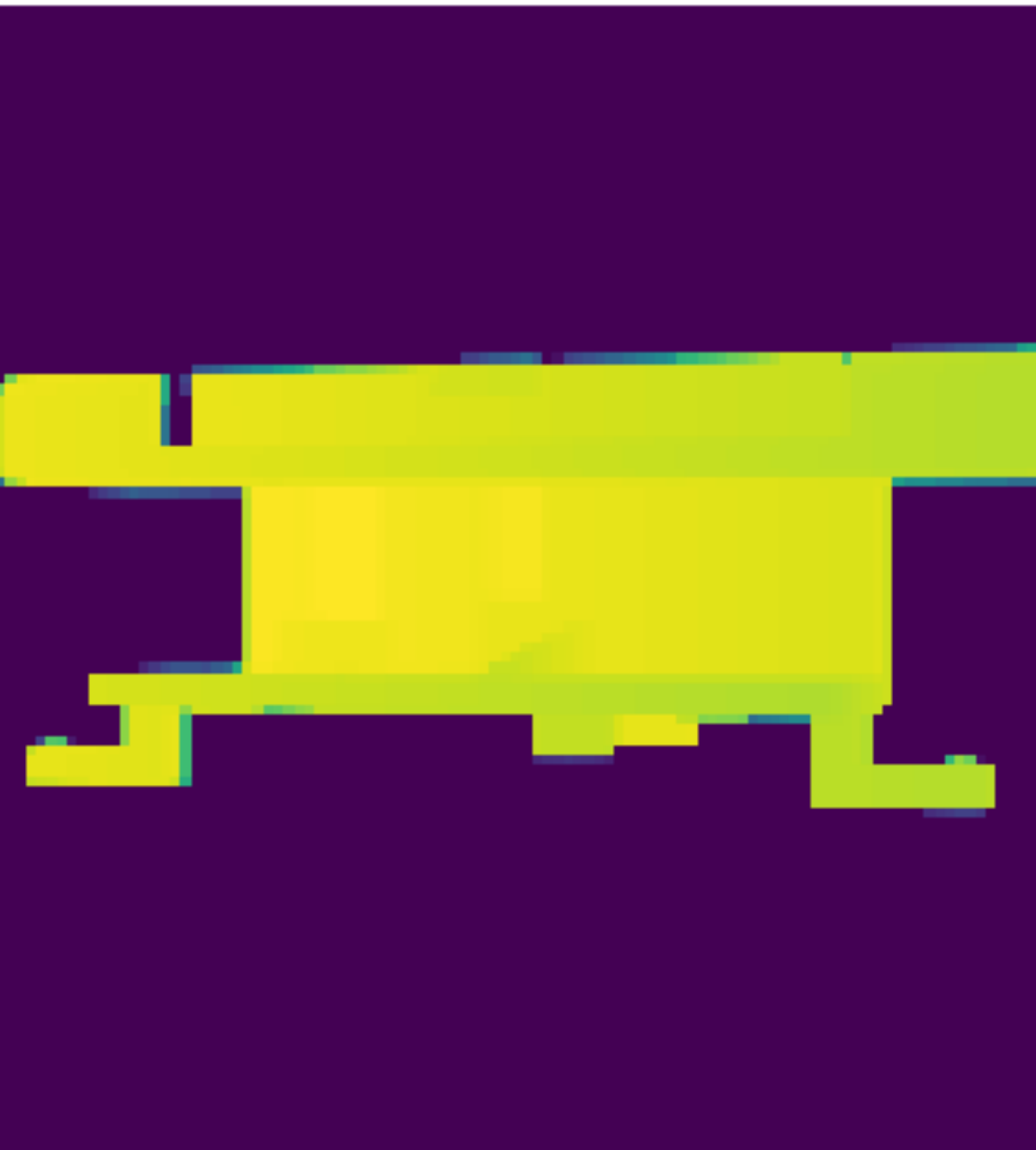}  &
			\includegraphics[trim={9cm 4cm 9cm 4cm}, clip = true,width=0.12\linewidth]{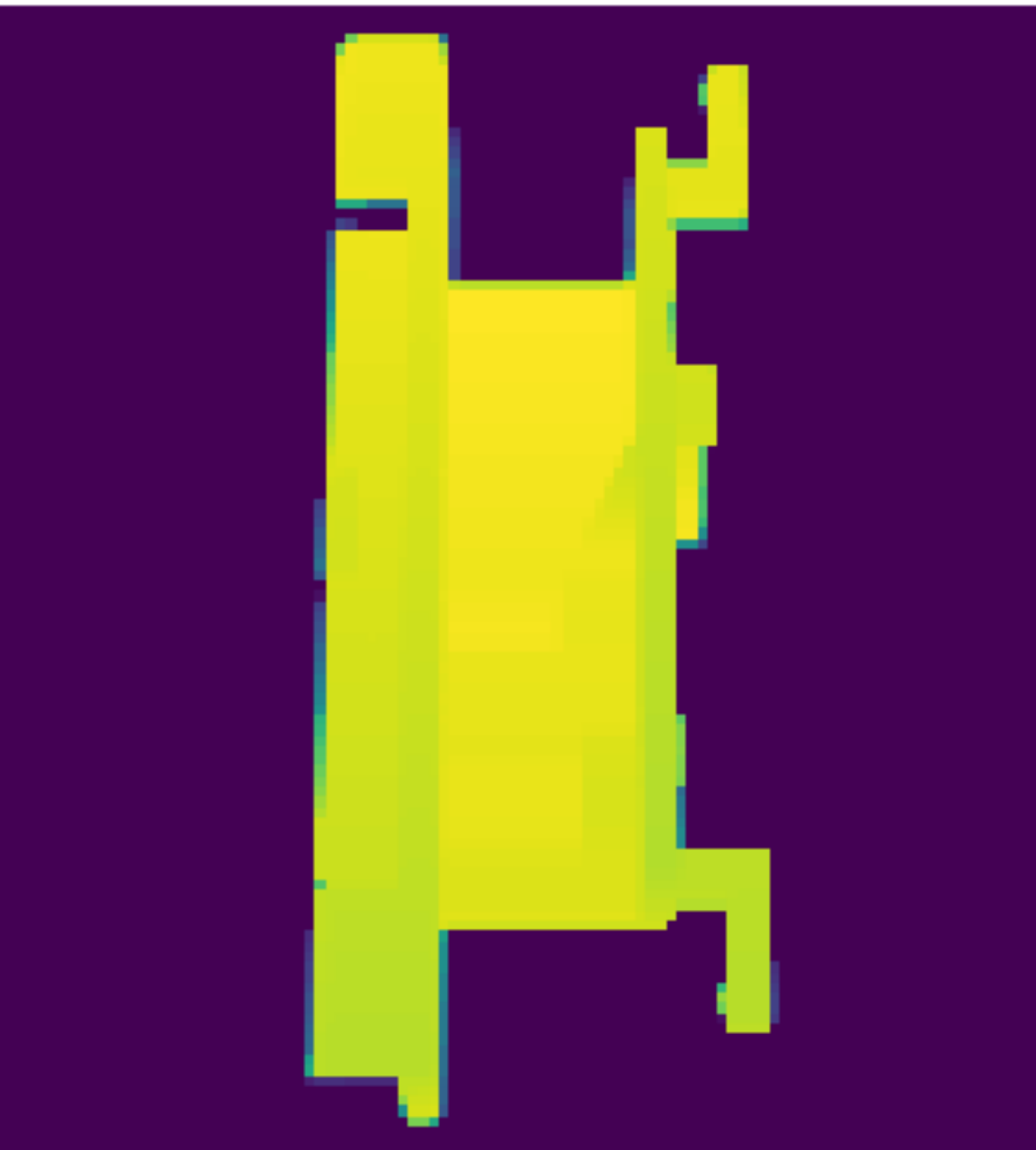}  &
			\includegraphics[trim={9cm 4cm 9cm 4cm}, clip = true,width=0.12\linewidth]{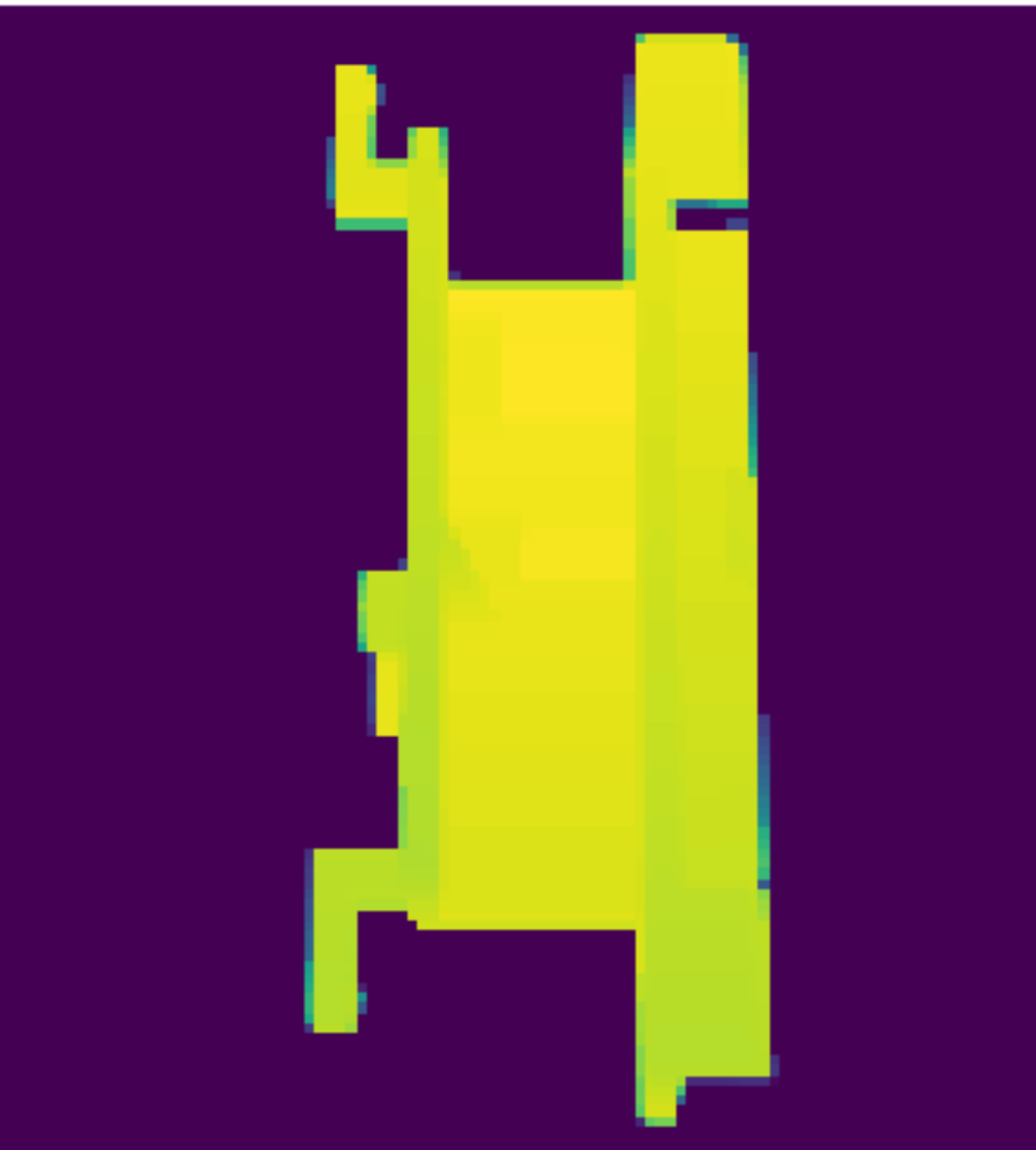}  &
			\includegraphics[trim={9cm 4cm 9cm 4cm}, clip = true,width=0.12\linewidth]{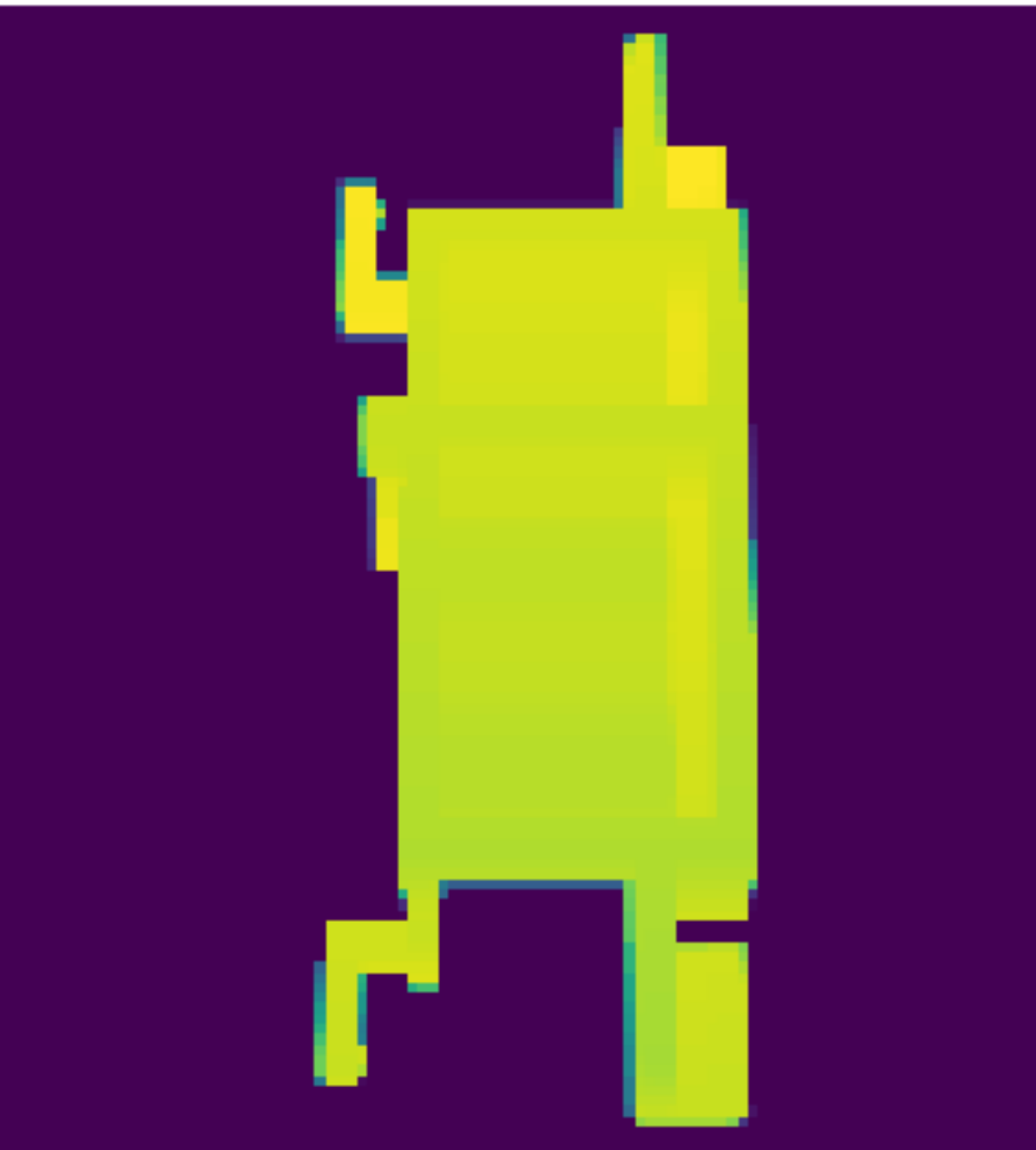}  &
			\fcolorbox{green}{white}{\includegraphics[trim={9cm 4cm 9cm 4cm}, clip = true,width=0.12\linewidth]{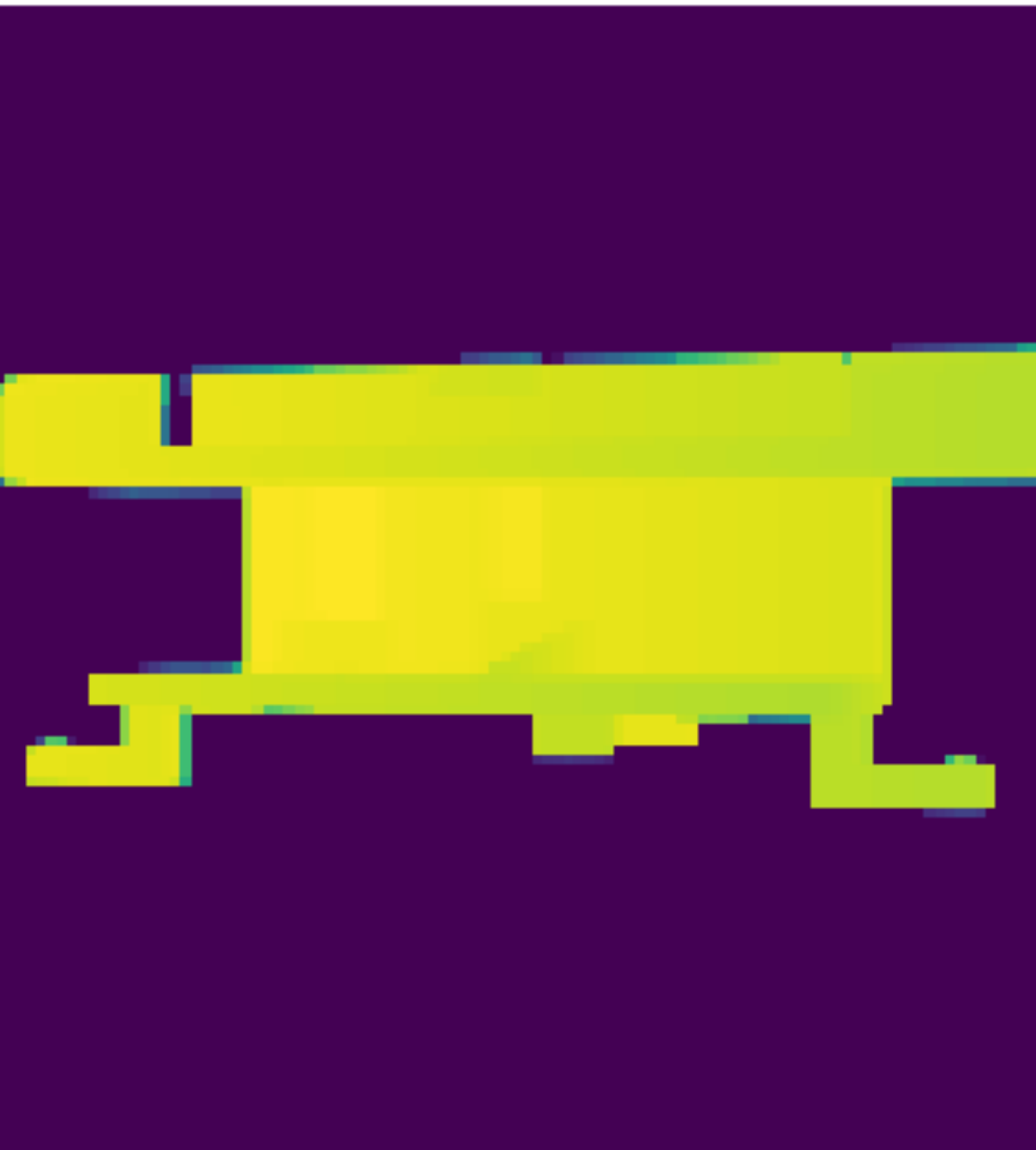}}  & 
			\raisebox{2\height}{\LARGE 12.68$^{\circ}$} \\
			\hline 
			\includegraphics[trim={9cm 4cm 9cm 4cm}, clip = true,width=0.12\linewidth]{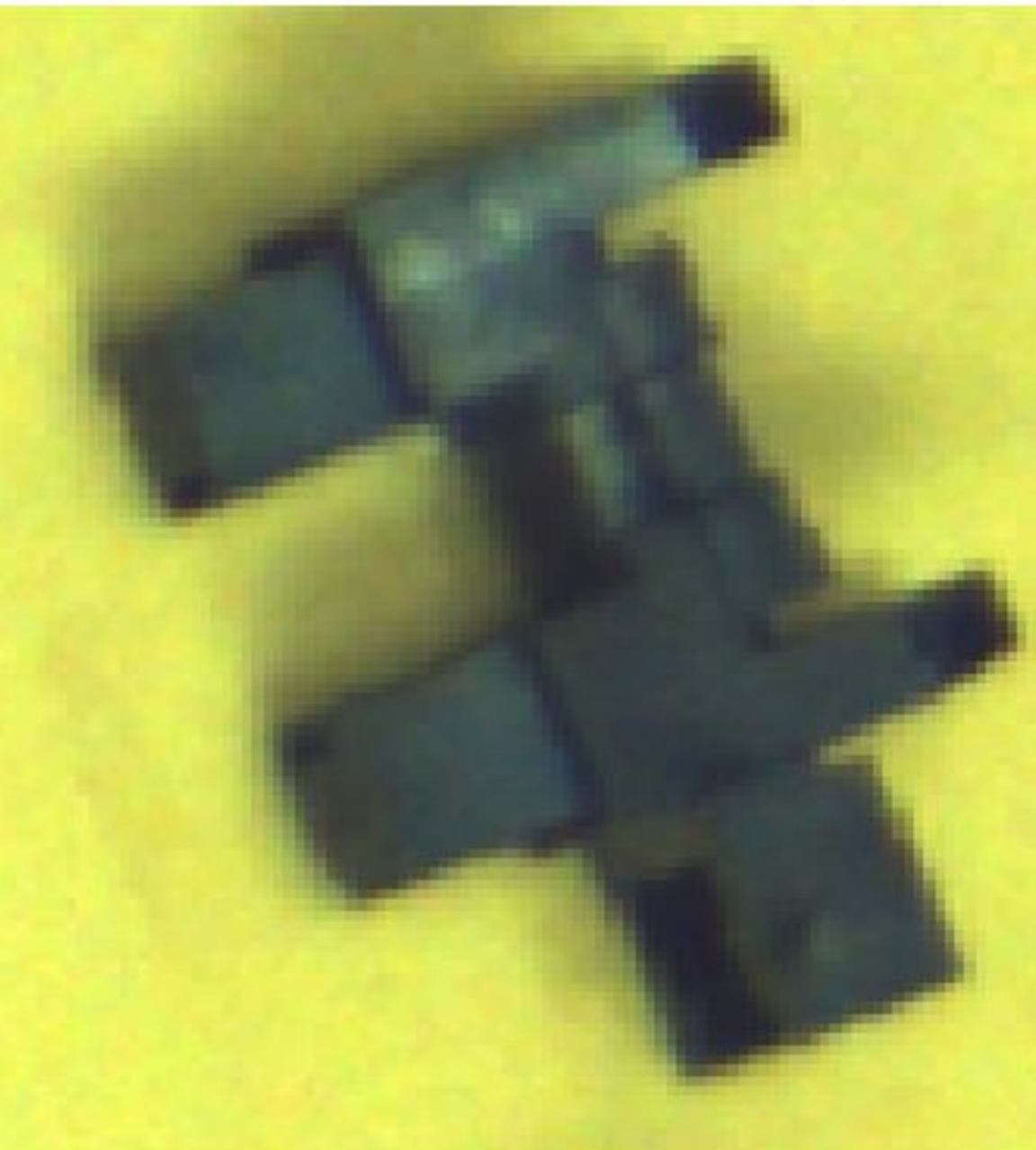} &
			\includegraphics[trim={9cm 4cm 9cm 4cm}, clip = true,width=0.12\linewidth]{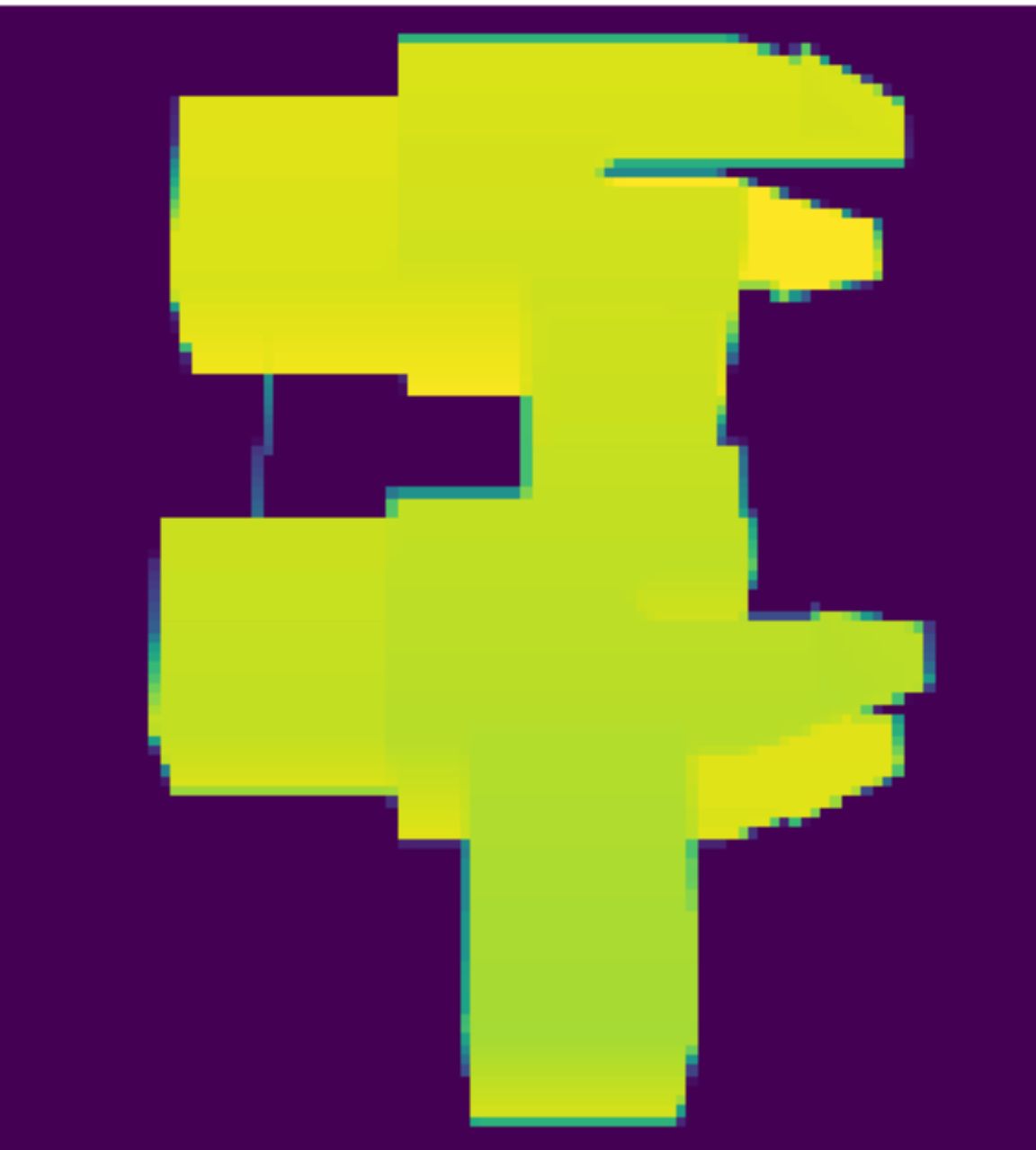} &
			\fcolorbox{green}{white}{\includegraphics[trim={9cm 4cm 9cm 4cm}, clip = true,width=0.12\linewidth]{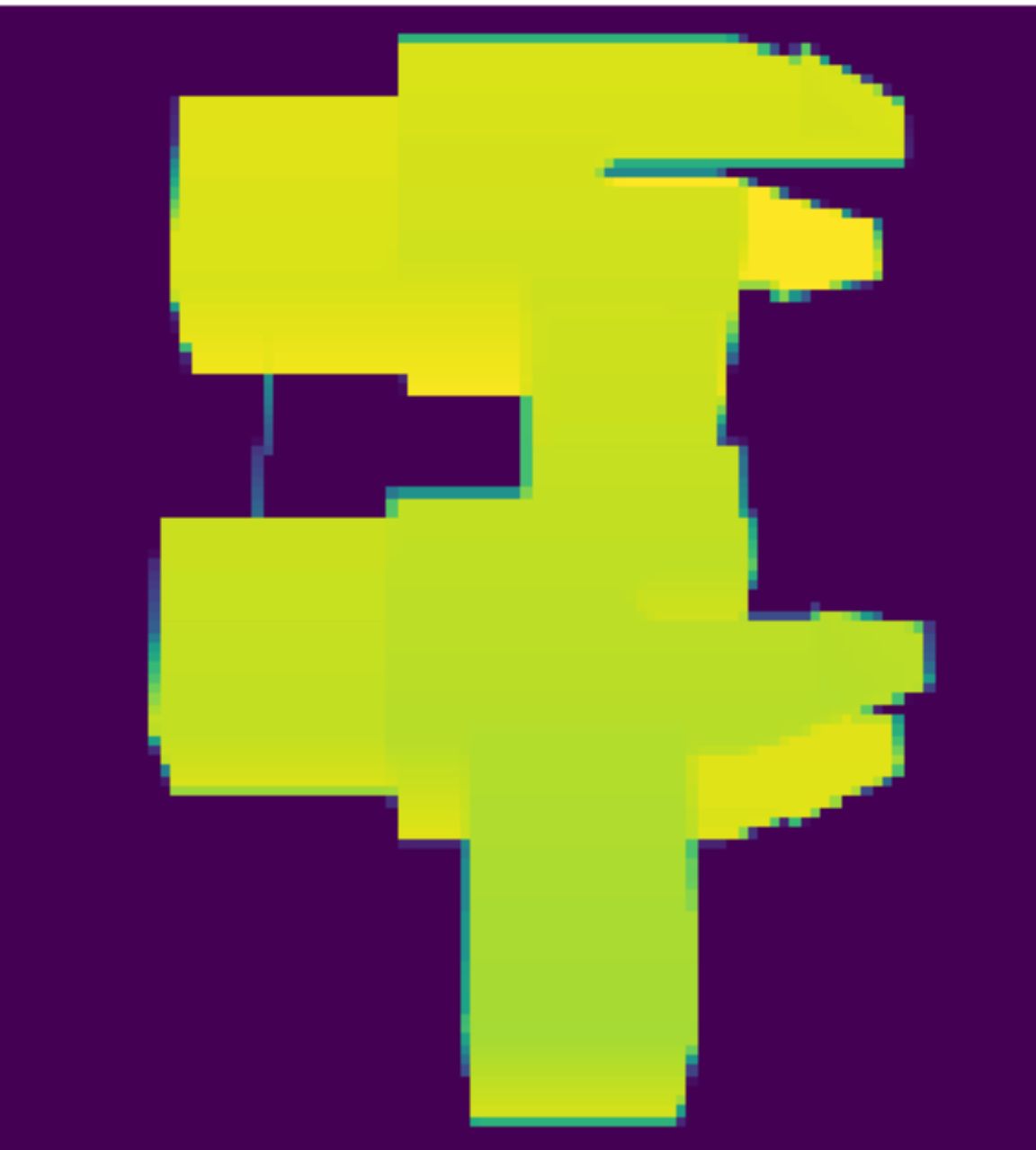}} &
			\includegraphics[trim={9cm 4cm 9cm 4cm}, clip = true,width=0.12\linewidth]{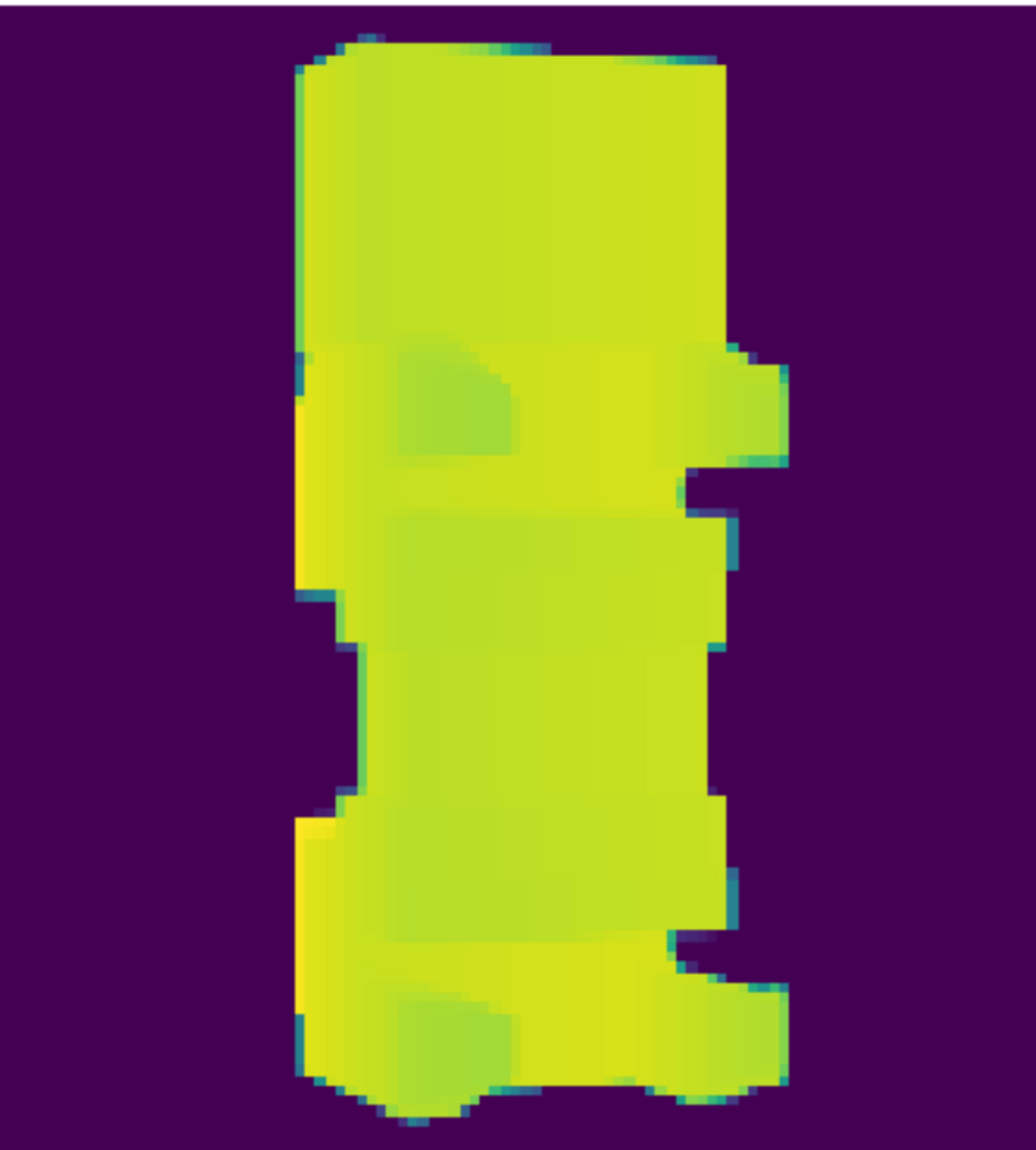} &
			\includegraphics[trim={9cm 4cm 9cm 4cm}, clip = true,width=0.12\linewidth]{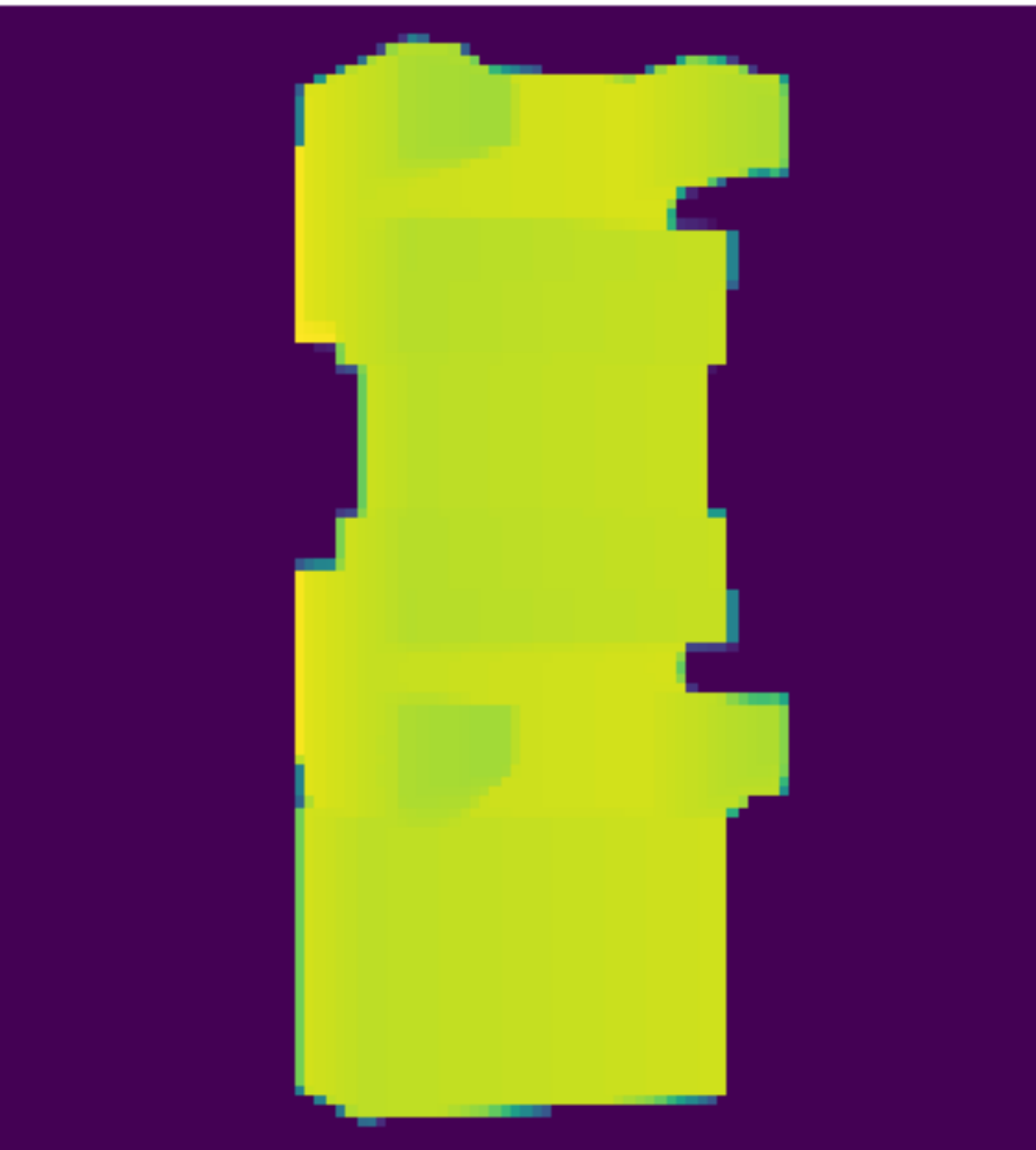} &
			\includegraphics[trim={9cm 4cm 9cm 4cm}, clip = true,width=0.12\linewidth]{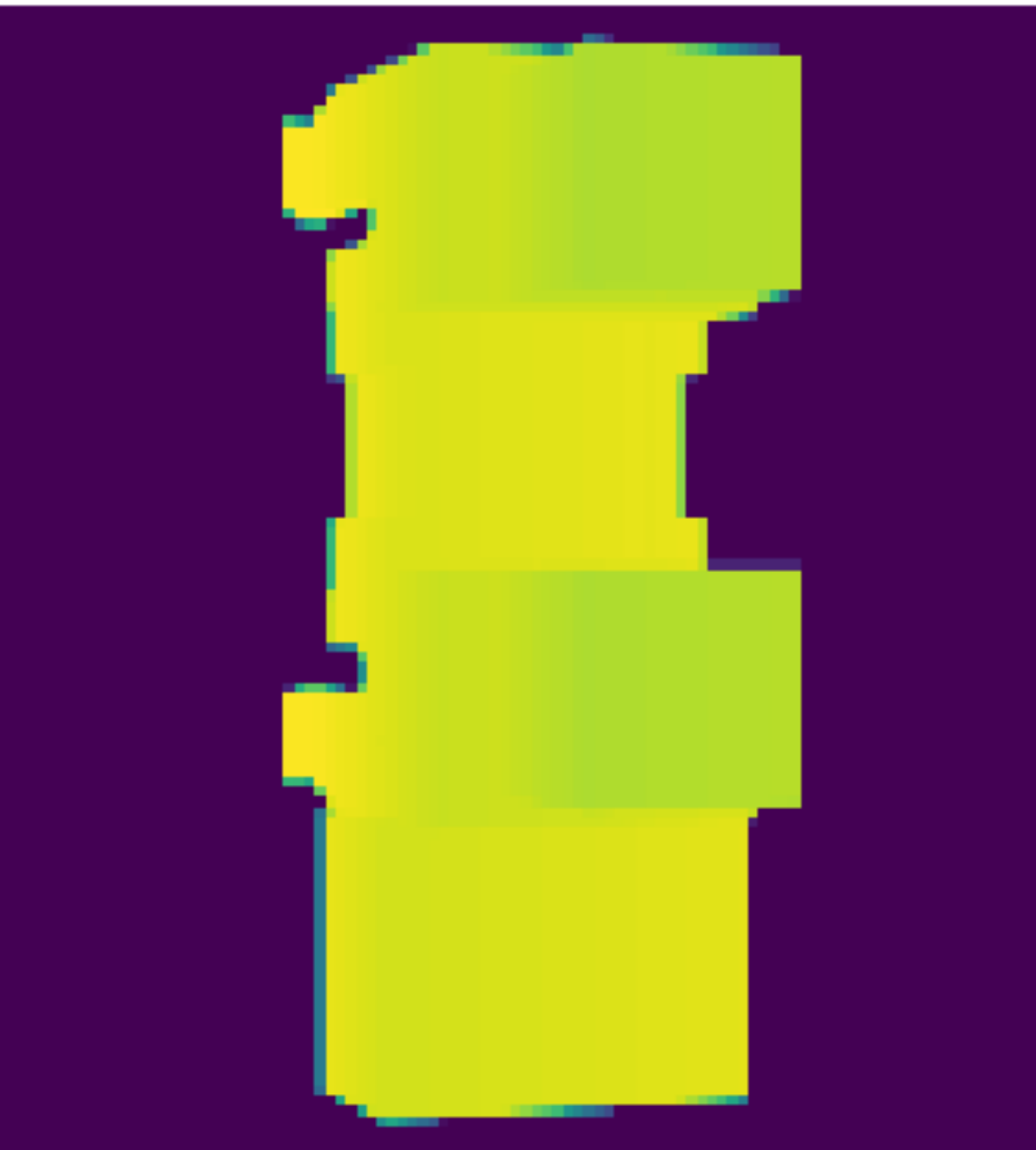} & 
			\raisebox{2\height}{\LARGE 12.78$^{\circ}$}  \\
			\hline 
		\end{tabular}
	}
		\centering
		\caption{{\bf Successful Results}: Qualitative results for pose estimation on the \textbf{real} dataset.} % The object in the scene is shown in the first column. The second column shows the nearest discretizated view. When the top four predicted views shown in the third column match the nearest view, they are indicated in a \textbf{green} box. Otherwise, an \textbf{orange} box shows the view that is nearest among the predicted. The last column shows the distance of this view. Given objects with symmetry, there can be more than one best view.}
		\label{fig:epson1034_result_1}
\end{figure*}

\begin{figure*}
	\hspace{-1.3cm}\parbox{\linewidth}{
		\begin{tabular}{|c|c|cccc|c|}
			\hline
			{\bf Input } & {\bf GT } & \multicolumn{4}{c|}{\bf Top-4 Predictions} & {\bf  $ d_\text{rot, best}^{sym} $} \\
			\hline 
			\includegraphics[trim={9cm 4cm 9cm 4cm}, clip = true,width=0.12\linewidth]{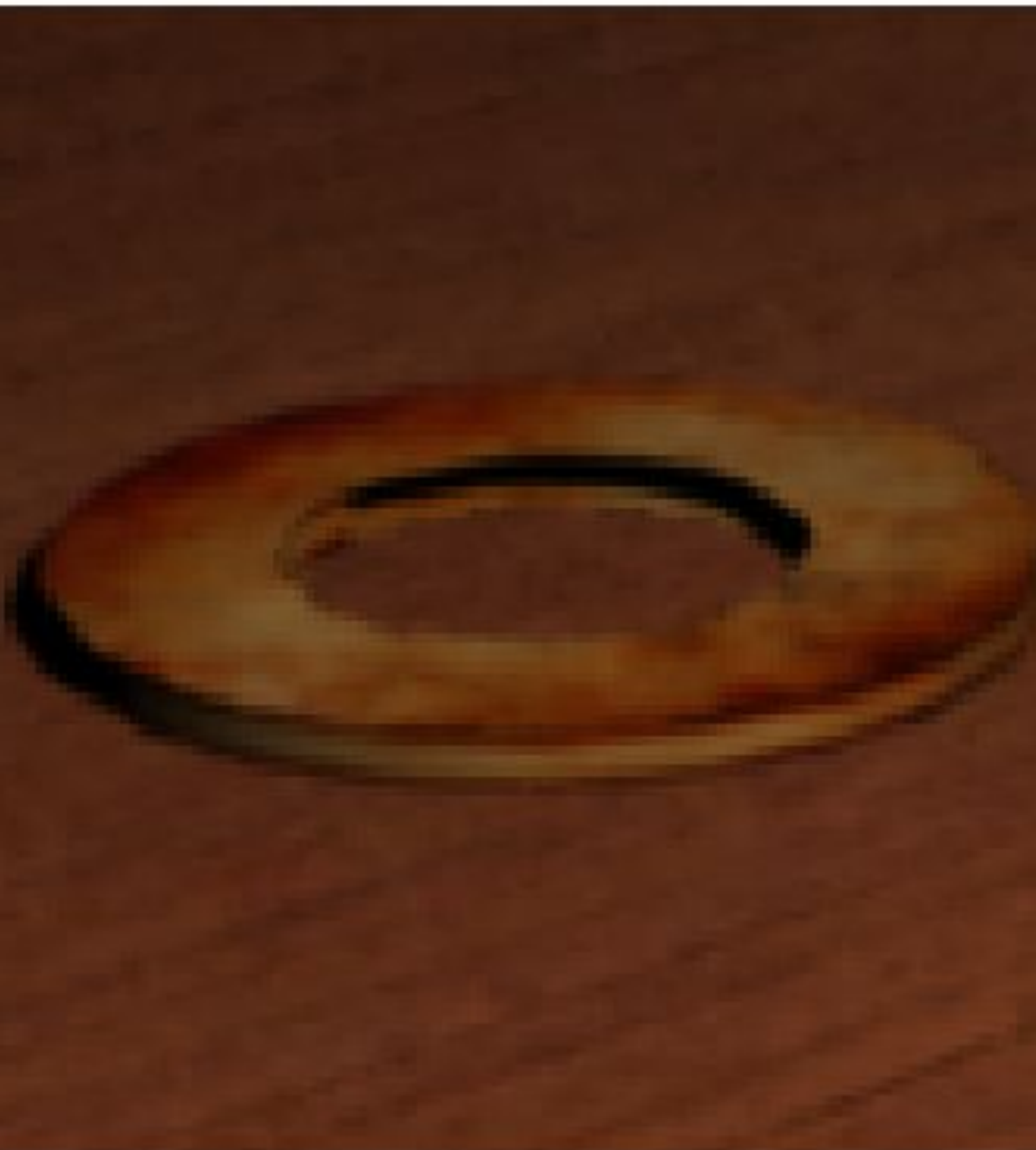}  &
			\includegraphics[trim={9cm 4cm 9cm 4cm}, clip = true,width=0.12\linewidth]{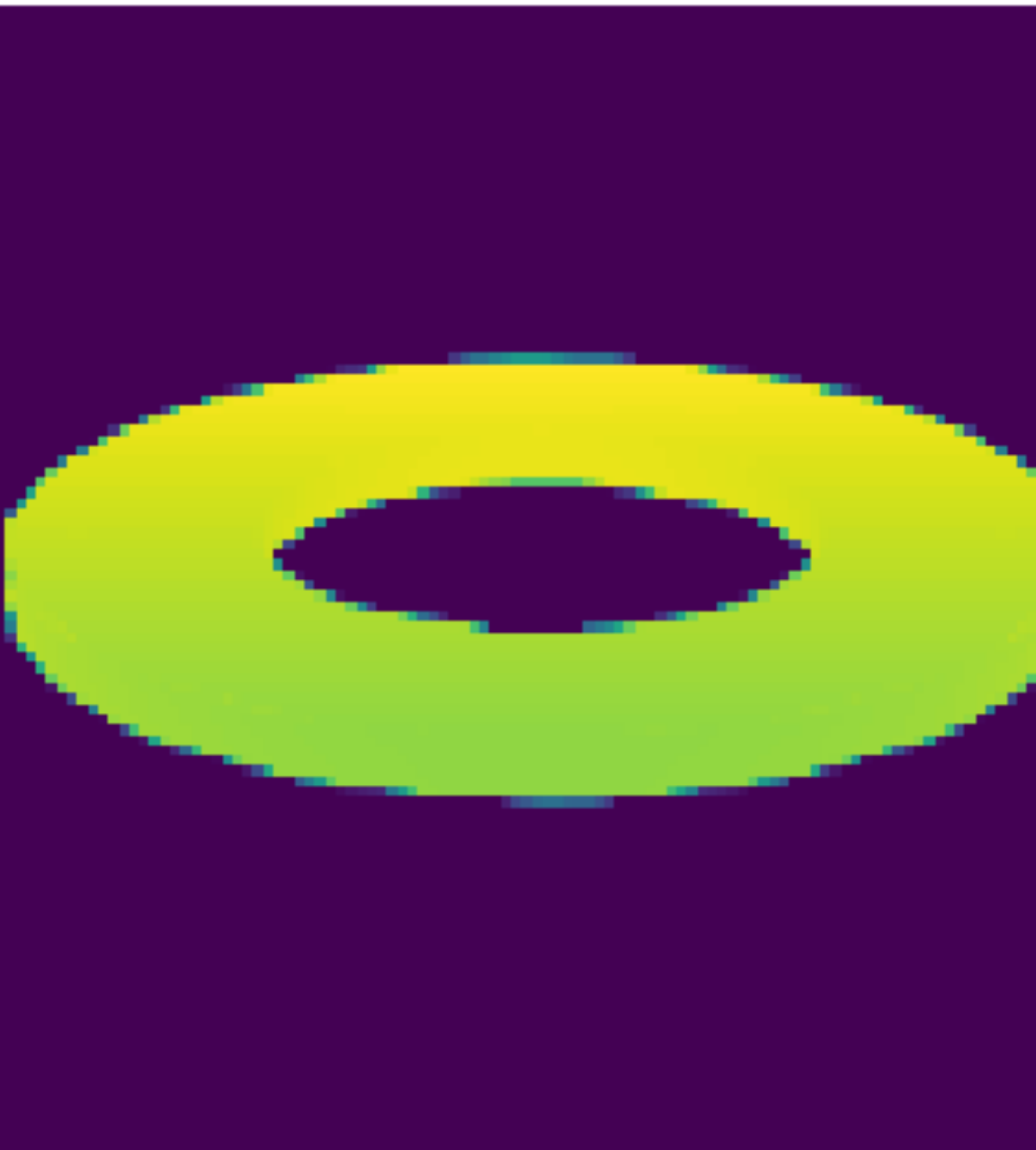}  &
			\fcolorbox{green}{white}{\includegraphics[trim={9cm 4cm 9cm 4cm}, clip = true,width=0.12\linewidth]{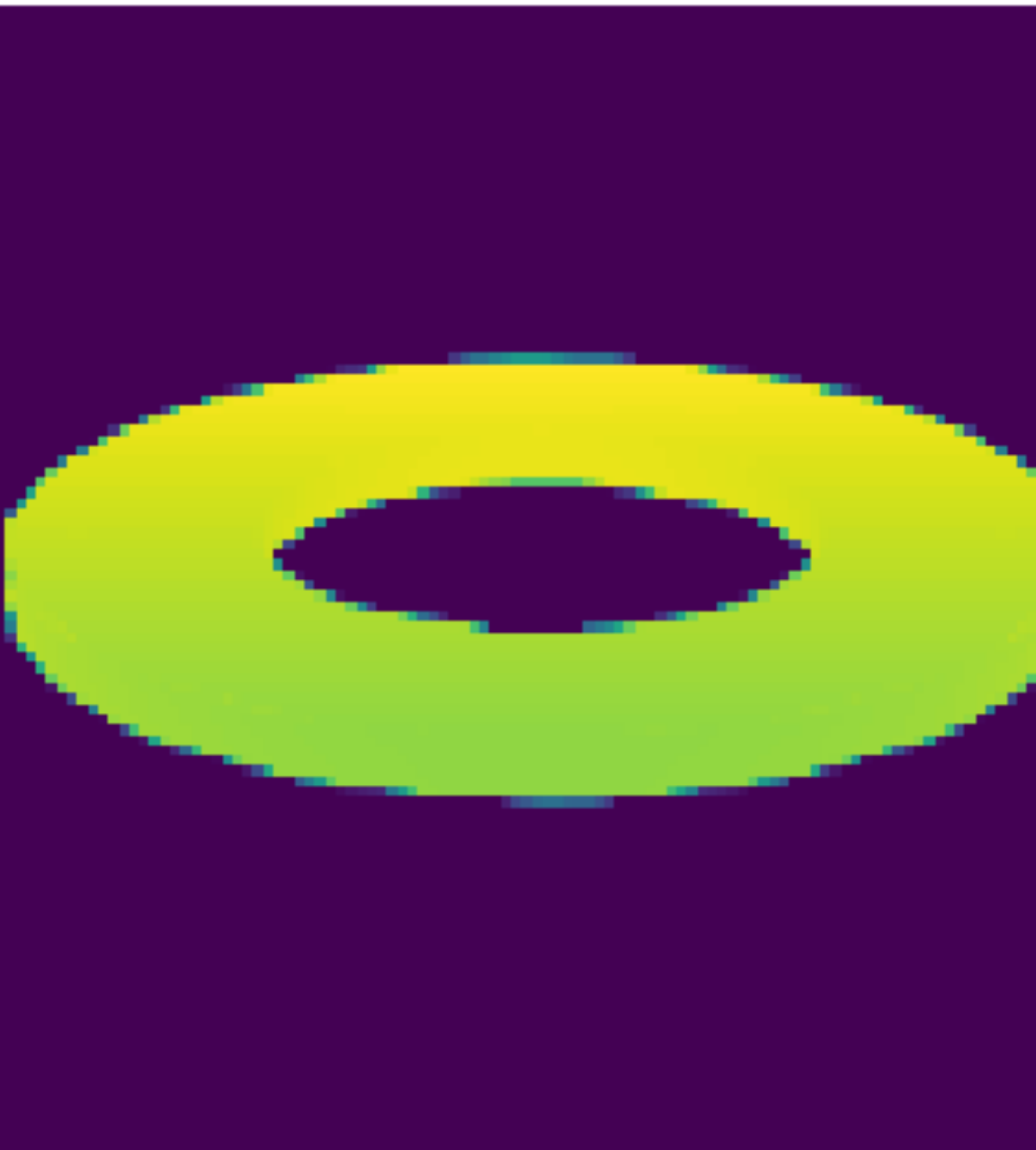} } &
			\fcolorbox{green}{white}{\includegraphics[trim={9cm 4cm 9cm 4cm}, clip = true,width=0.12\linewidth]{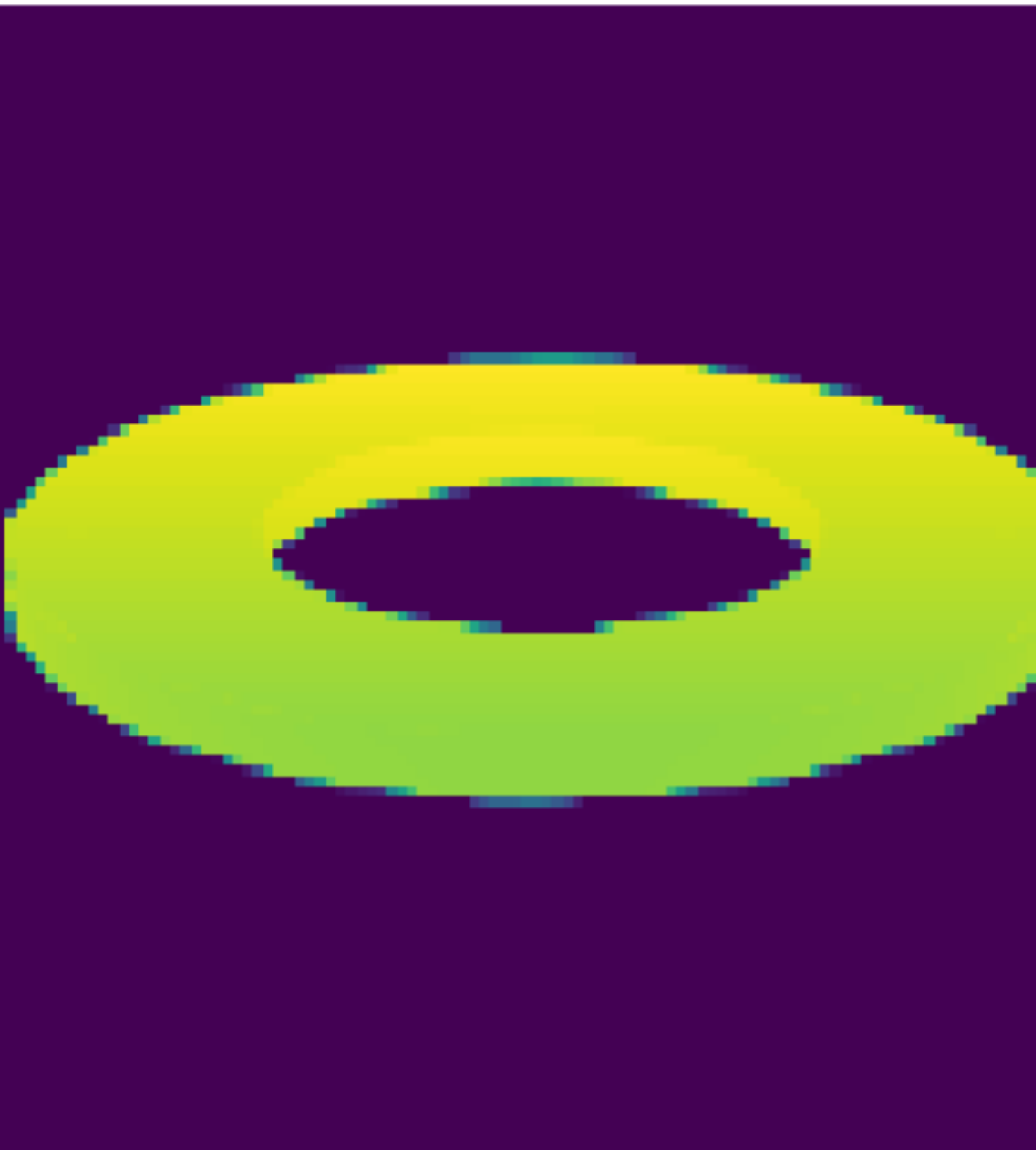} } &
			\fcolorbox{green}{white}{\includegraphics[trim={9cm 4cm 9cm 4cm}, clip = true,width=0.12\linewidth]{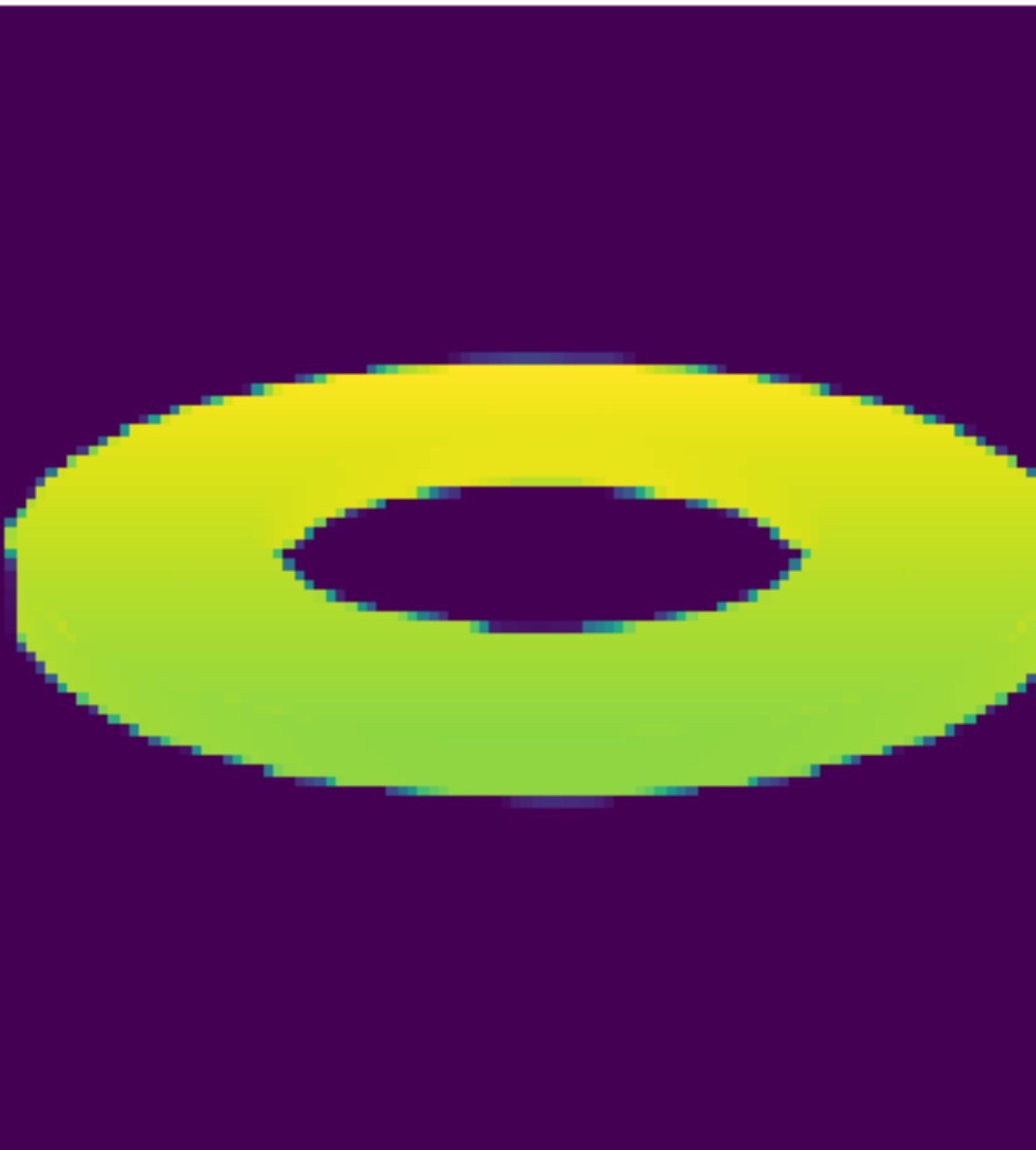} } &
			\fcolorbox{green}{white}{\includegraphics[trim={9cm 4cm 9cm 4cm}, clip = true,width=0.12\linewidth]{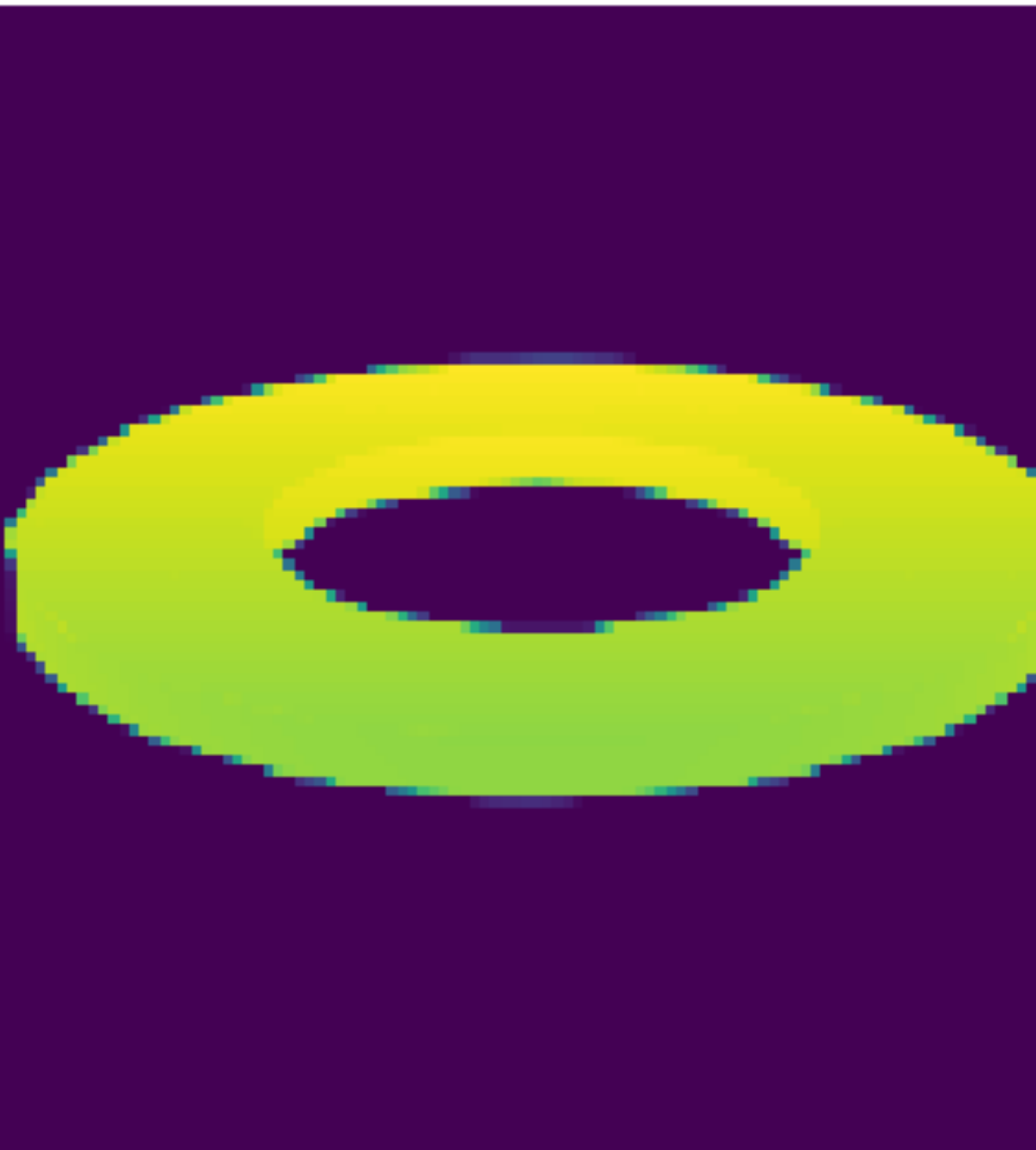} } & \raisebox{2\height}{\LARGE 1.49$^{\circ}$ } \\ \hline
			\includegraphics[trim={9cm 4cm 9cm 4cm}, clip = true,width=0.12\linewidth]{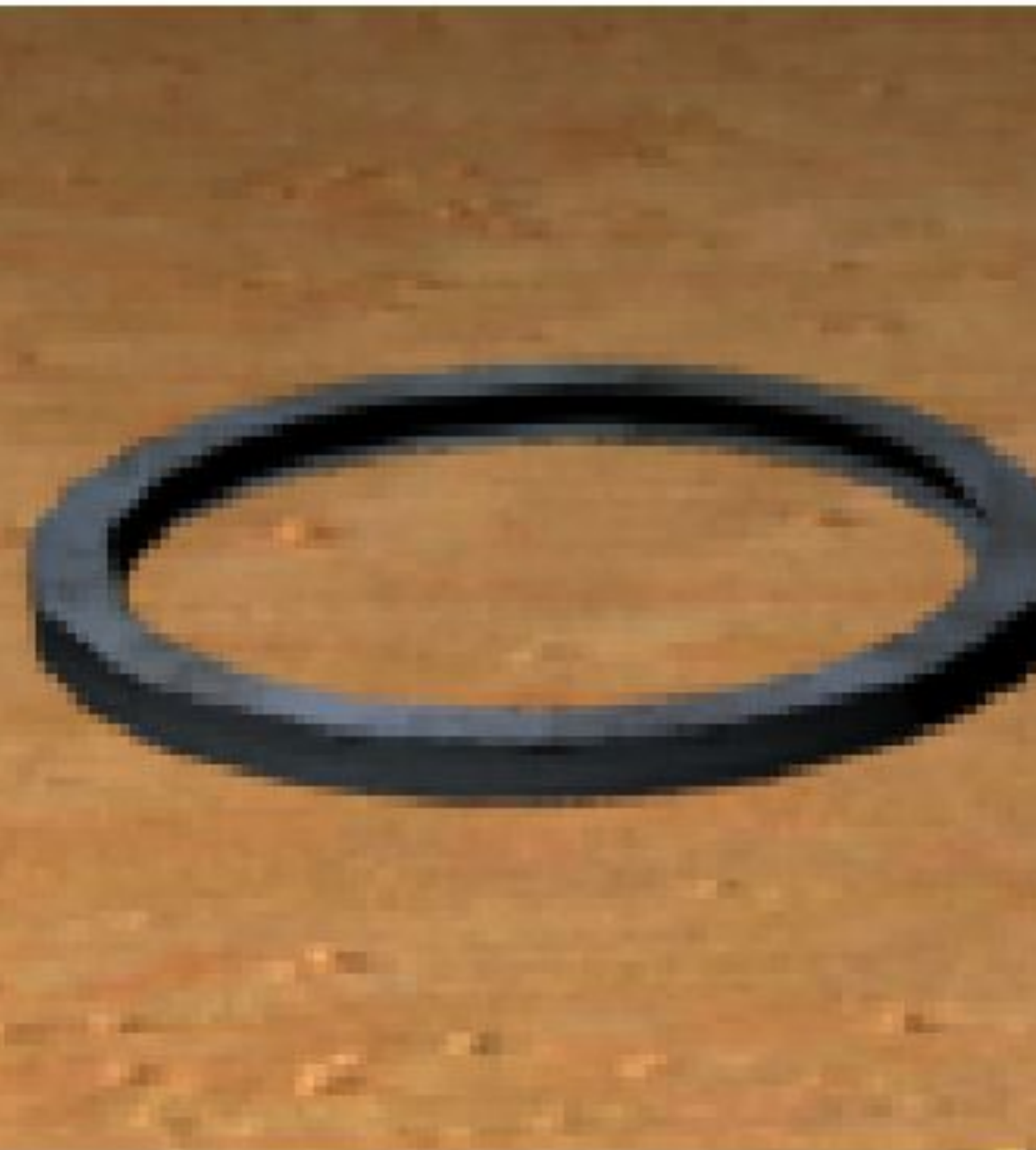} &
			\includegraphics[trim={9cm 4cm 9cm 4cm}, clip = true,width=0.12\linewidth]{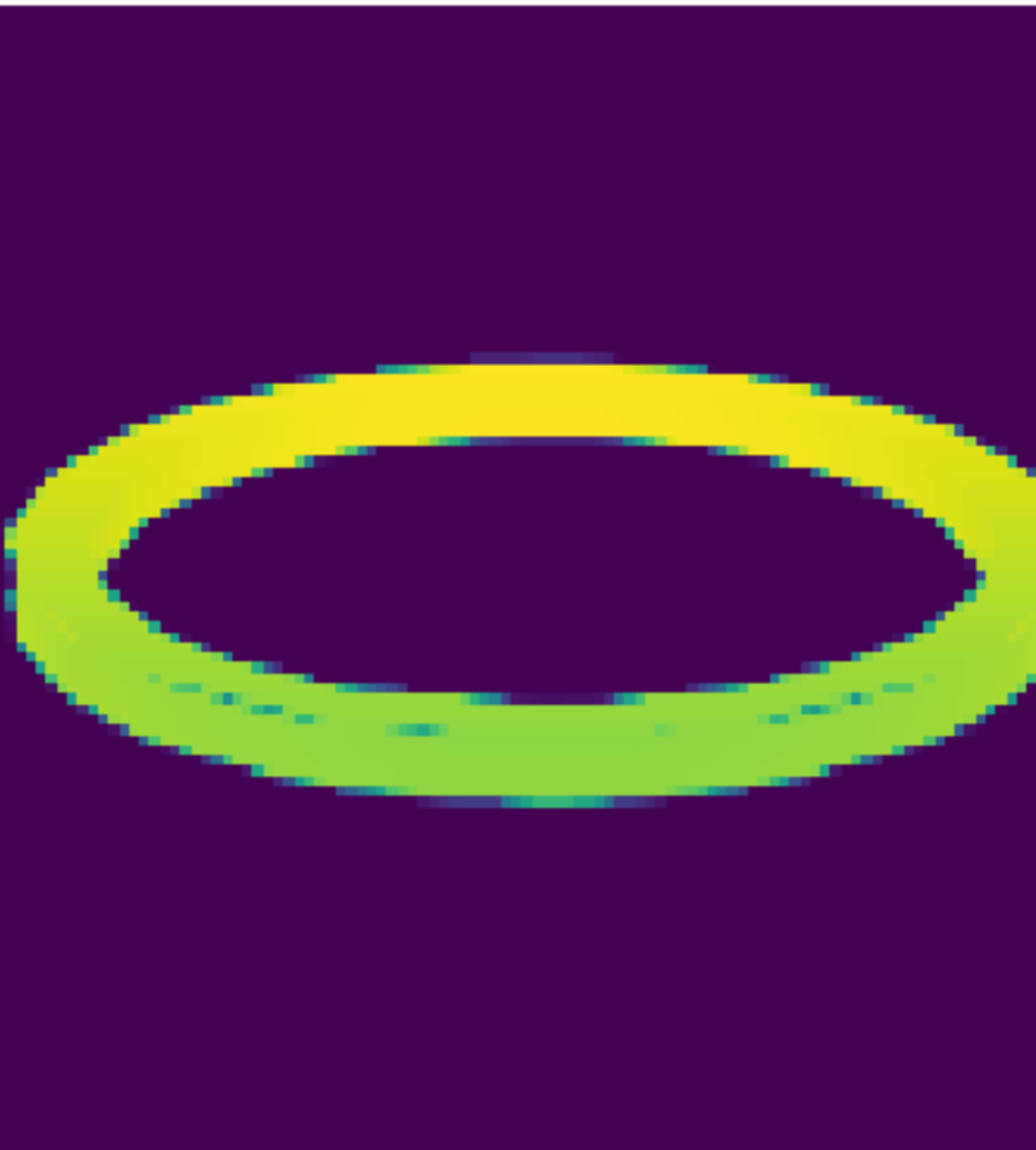} &
			\fcolorbox{green}{white}{\includegraphics[trim={9cm 4cm 9cm 4cm}, clip = true,width=0.12\linewidth]{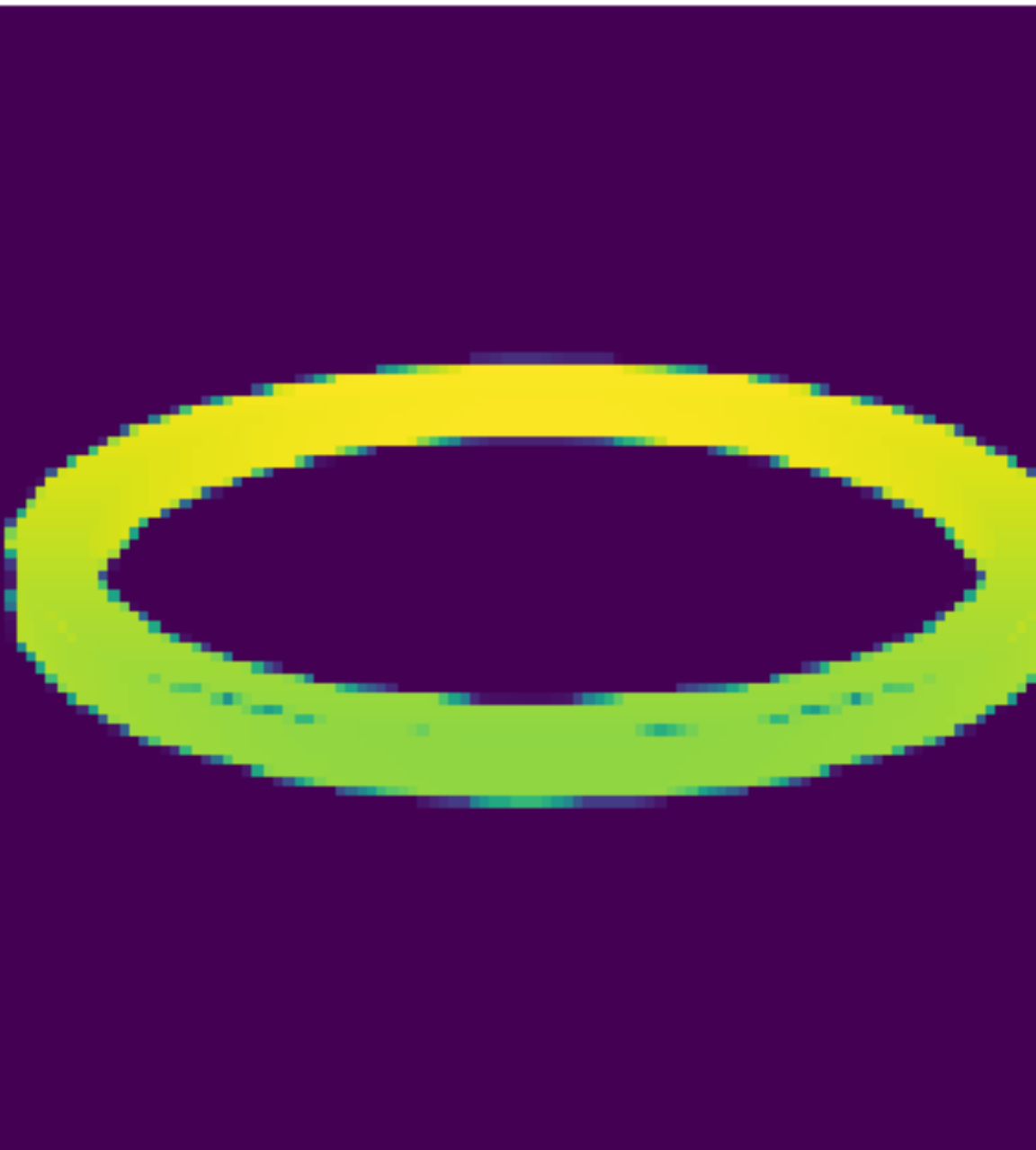}} &
			\fcolorbox{green}{white}{\includegraphics[trim={9cm 4cm 9cm 4cm}, clip = true,width=0.12\linewidth]{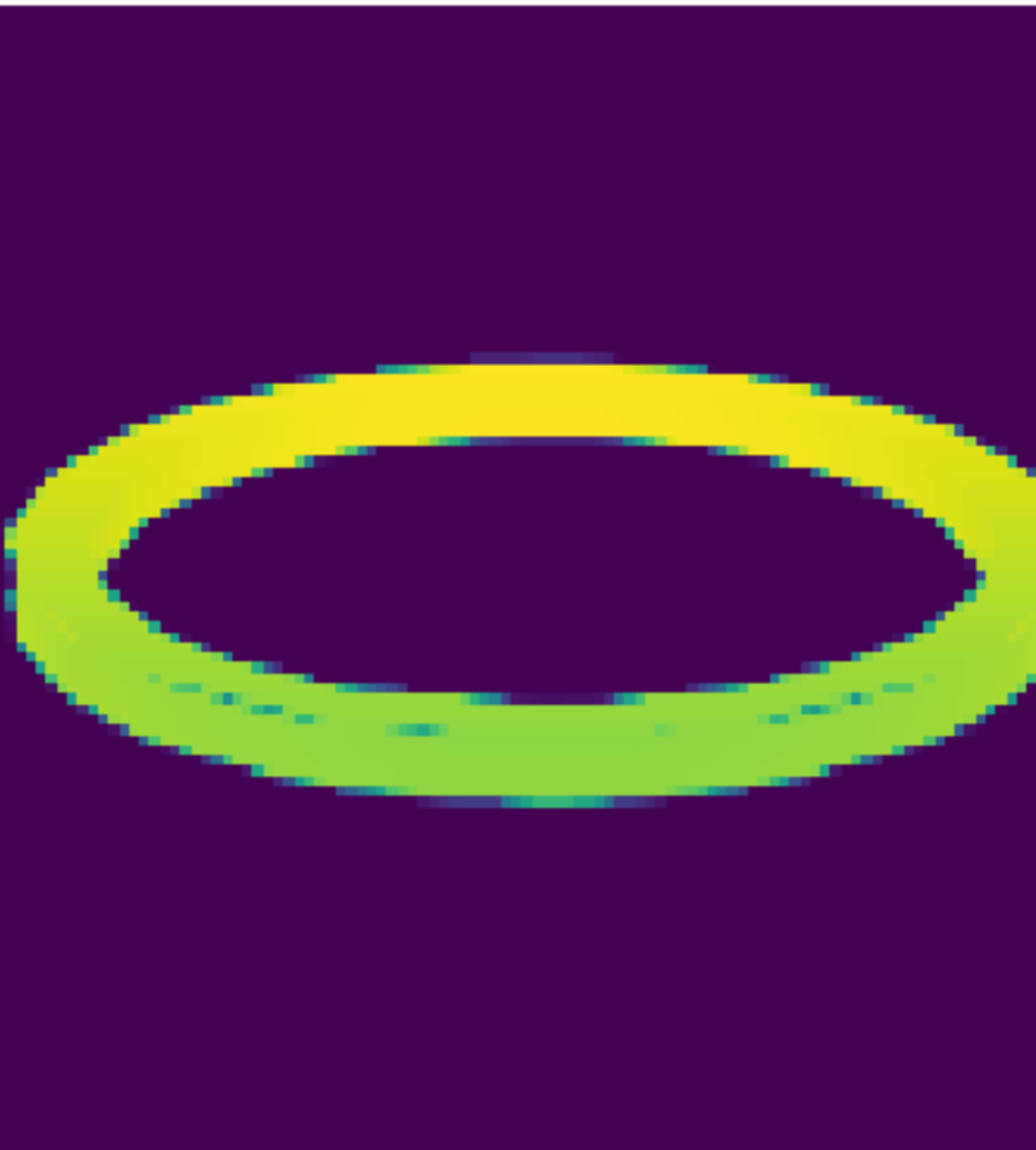}} &
			\fcolorbox{green}{white}{\includegraphics[trim={9cm 4cm 9cm 4cm}, clip = true,width=0.12\linewidth]{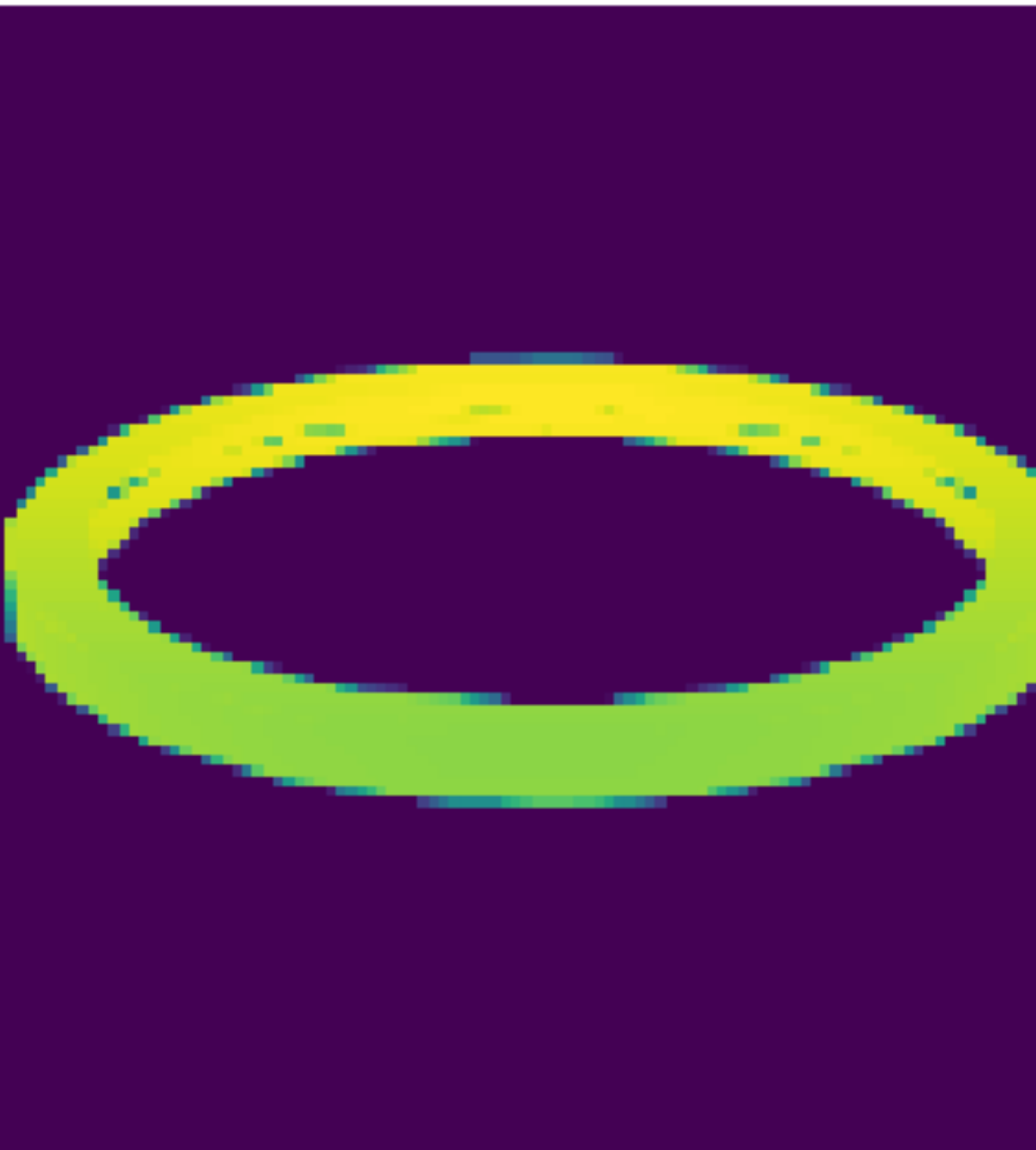}} &
			\fcolorbox{green}{white}{\includegraphics[trim={9cm 4cm 9cm 4cm}, clip = true,width=0.12\linewidth]{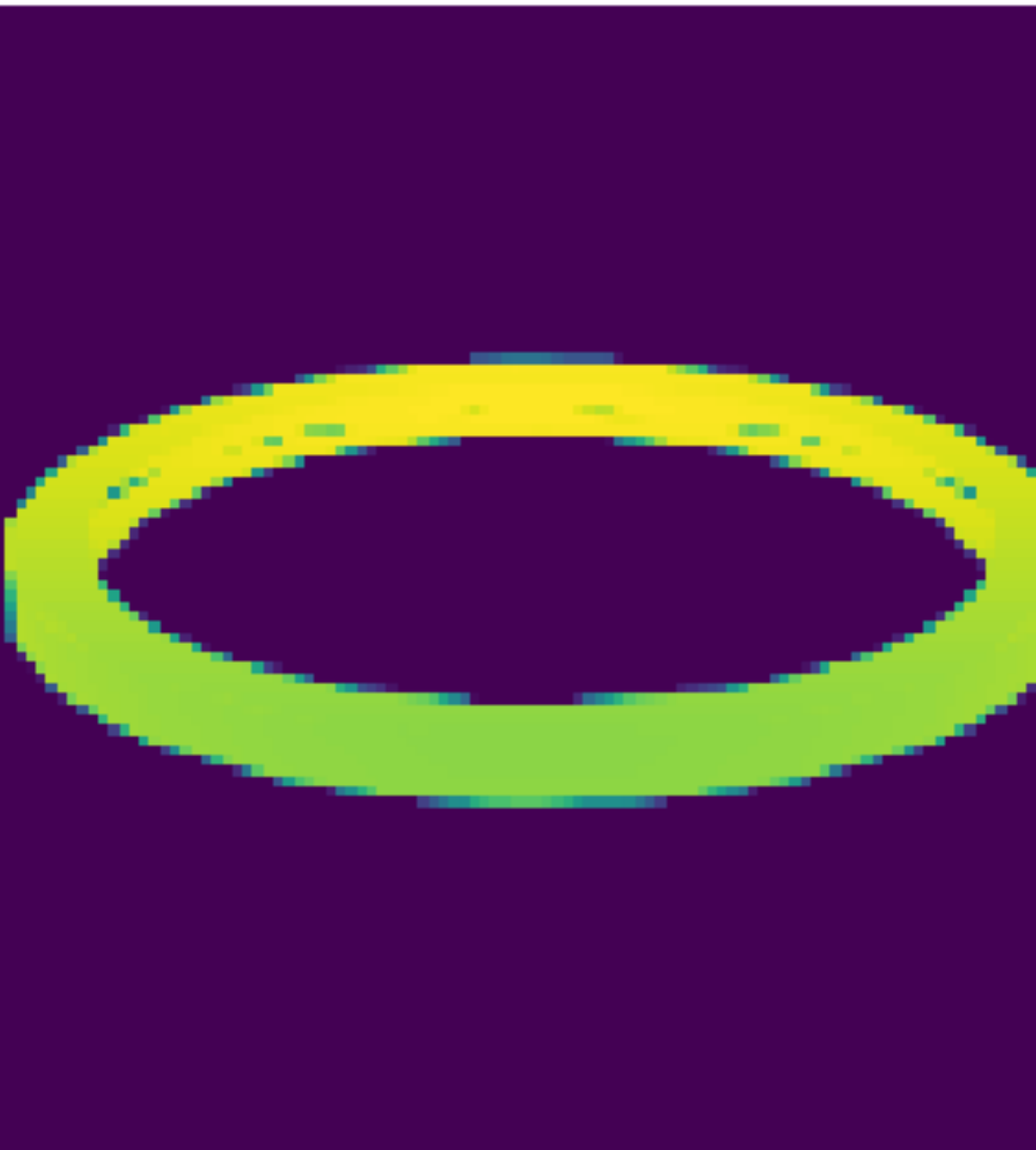}} & \raisebox{2\height}{\LARGE 2.02$^{\circ}$ } \\ \hline
			\includegraphics[trim={9cm 4cm 9cm 4cm}, clip = true,width=0.12\linewidth]{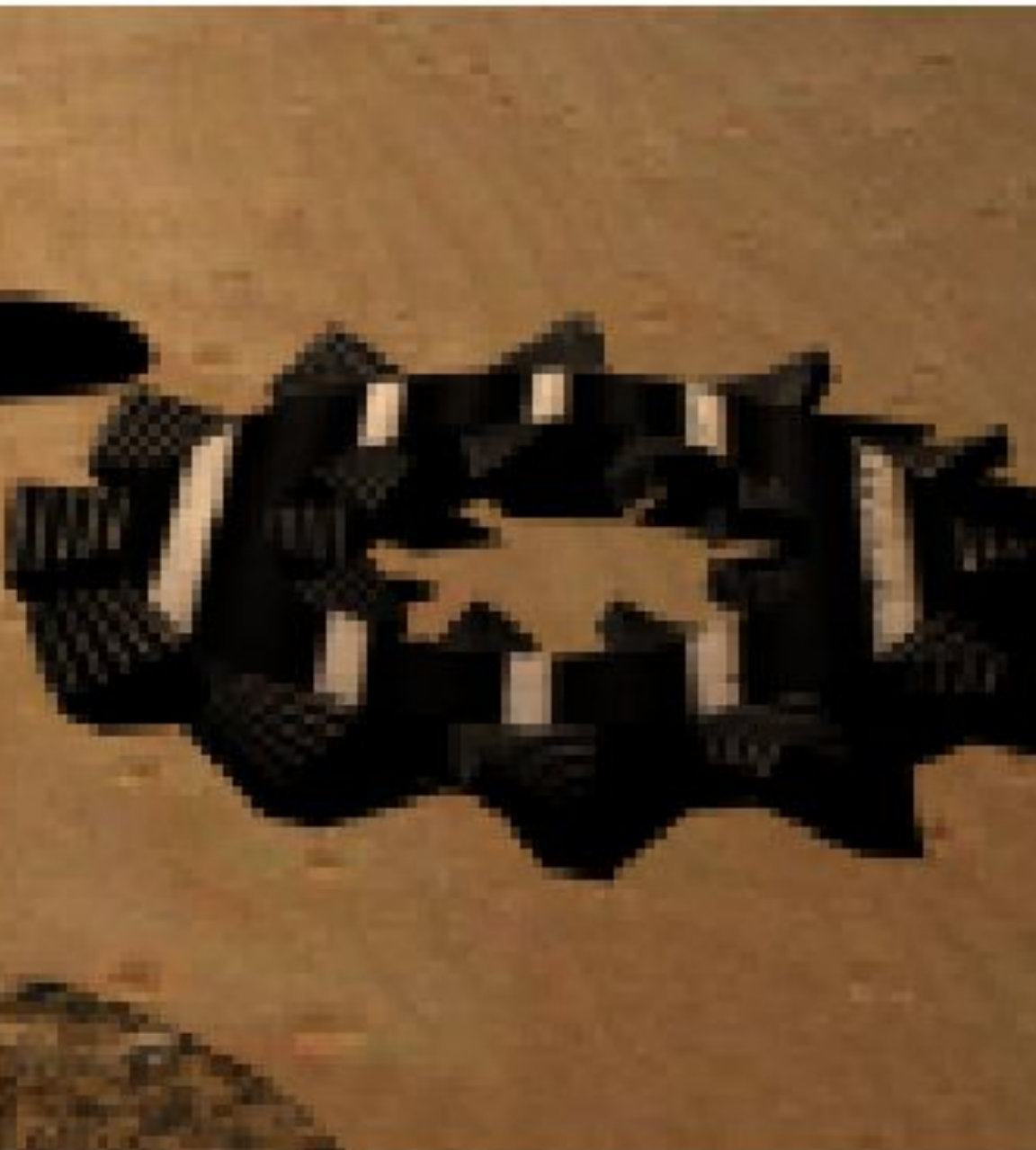}  &
			\includegraphics[trim={9cm 4cm 9cm 4cm}, clip = true,width=0.12\linewidth]{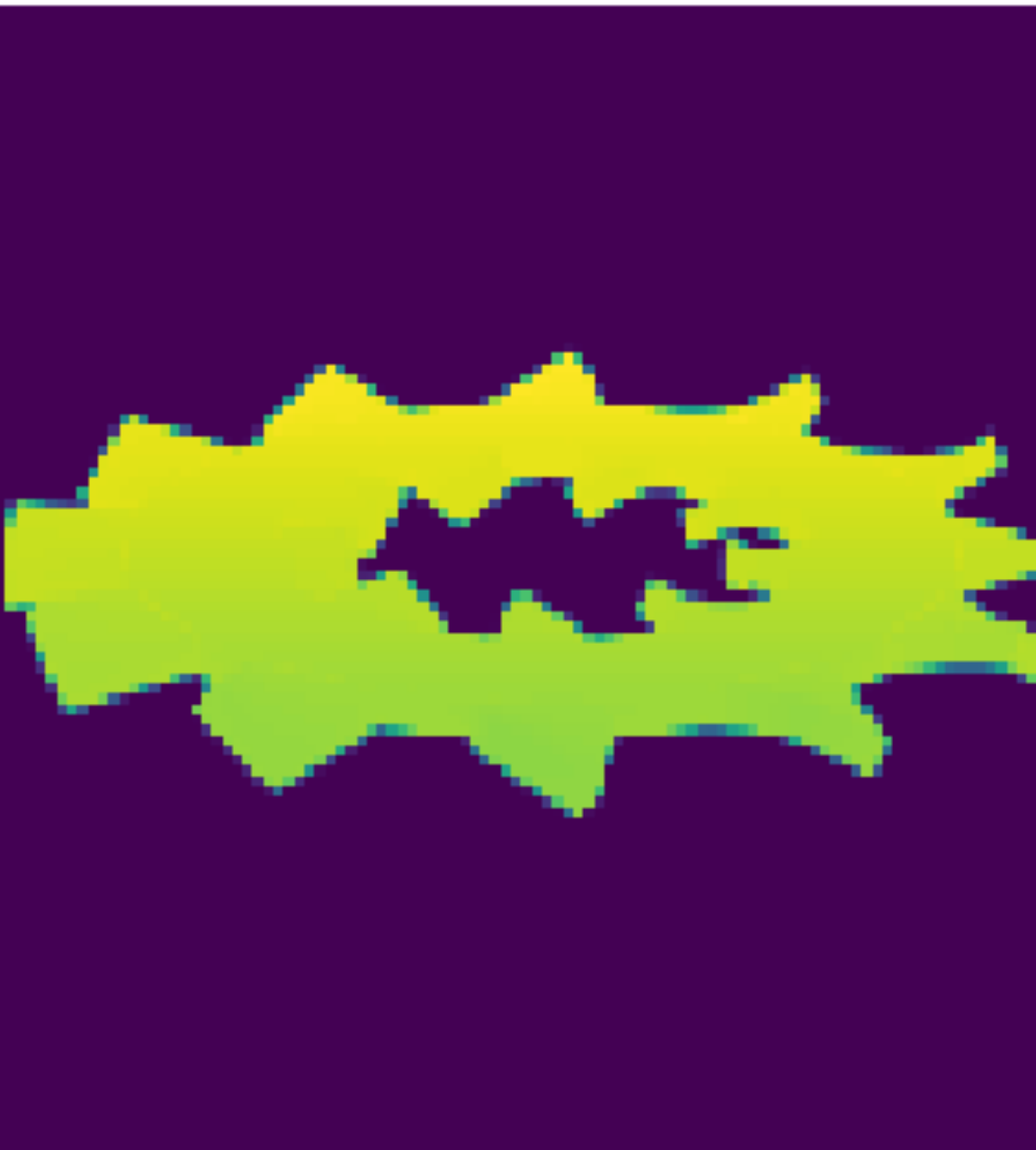}  &
			\fcolorbox{green}{white}{\includegraphics[trim={9cm 4cm 9cm 4cm}, clip = true,width=0.12\linewidth]{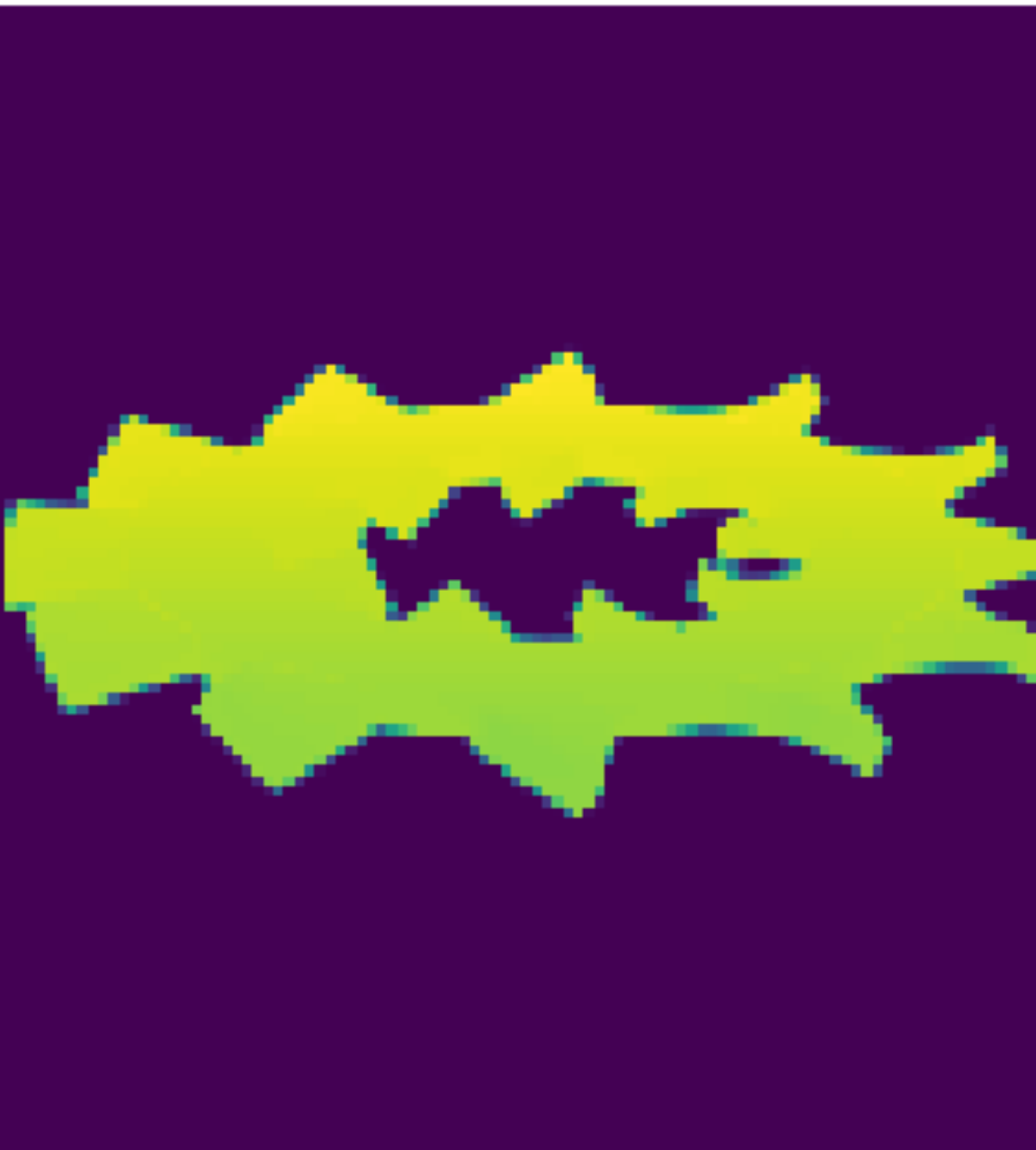} } &
			\fcolorbox{green}{white}{\includegraphics[trim={9cm 4cm 9cm 4cm}, clip = true,width=0.12\linewidth]{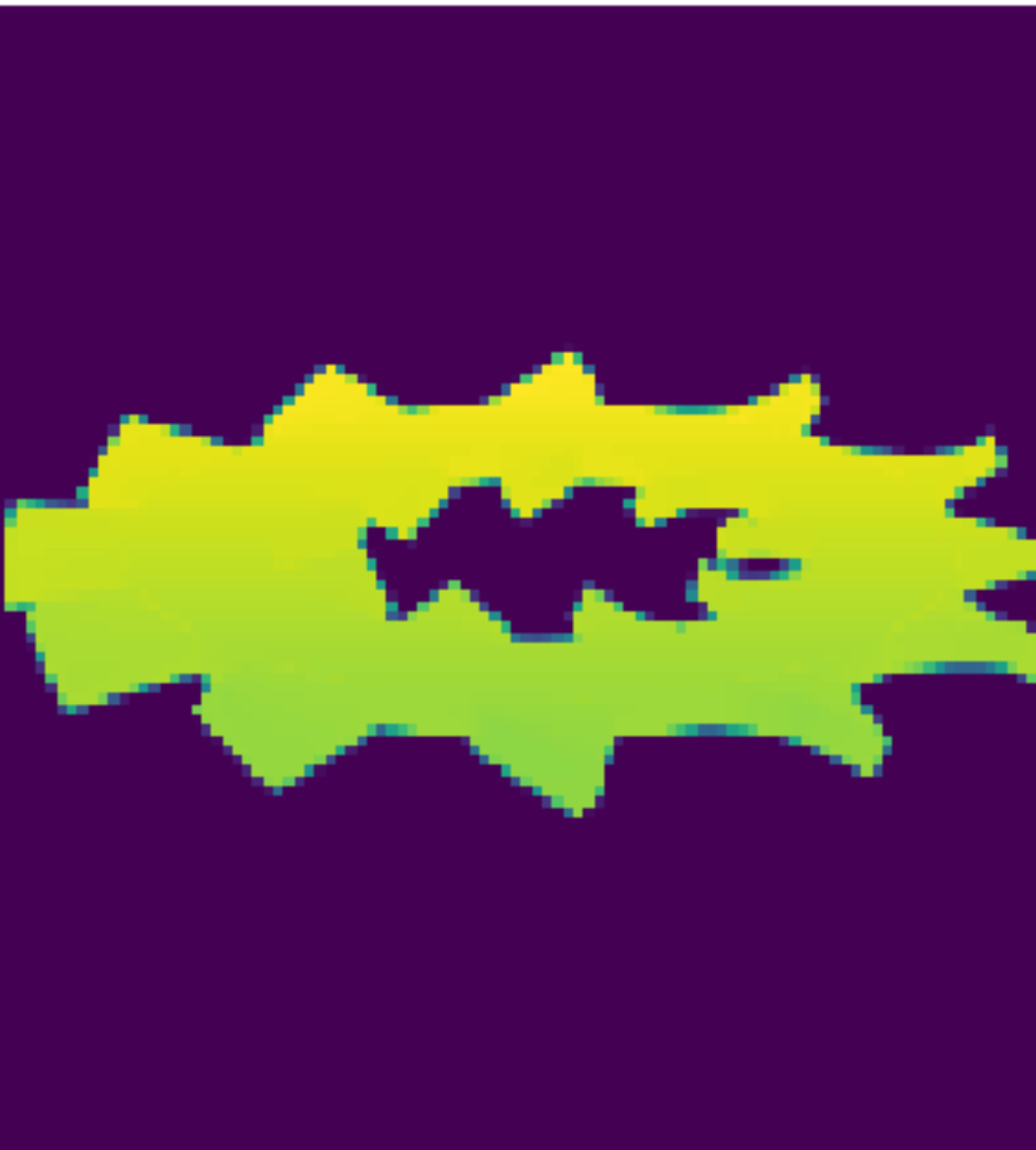} } &
			\fcolorbox{green}{white}{\includegraphics[trim={9cm 4cm 9cm 4cm}, clip = true,width=0.12\linewidth]{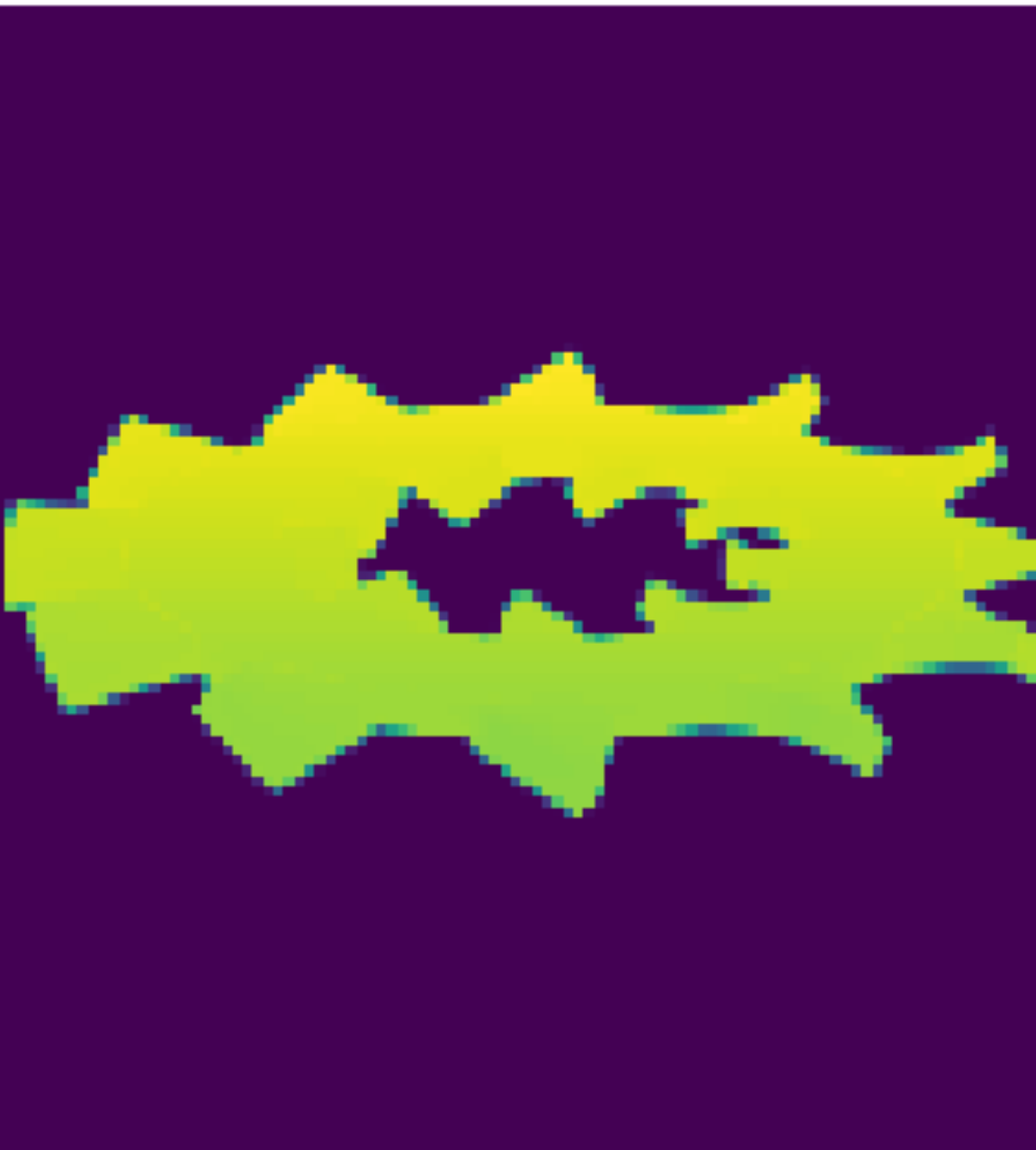} } &
			\fcolorbox{green}{white}{\includegraphics[trim={9cm 4cm 9cm 4cm}, clip = true,width=0.12\linewidth]{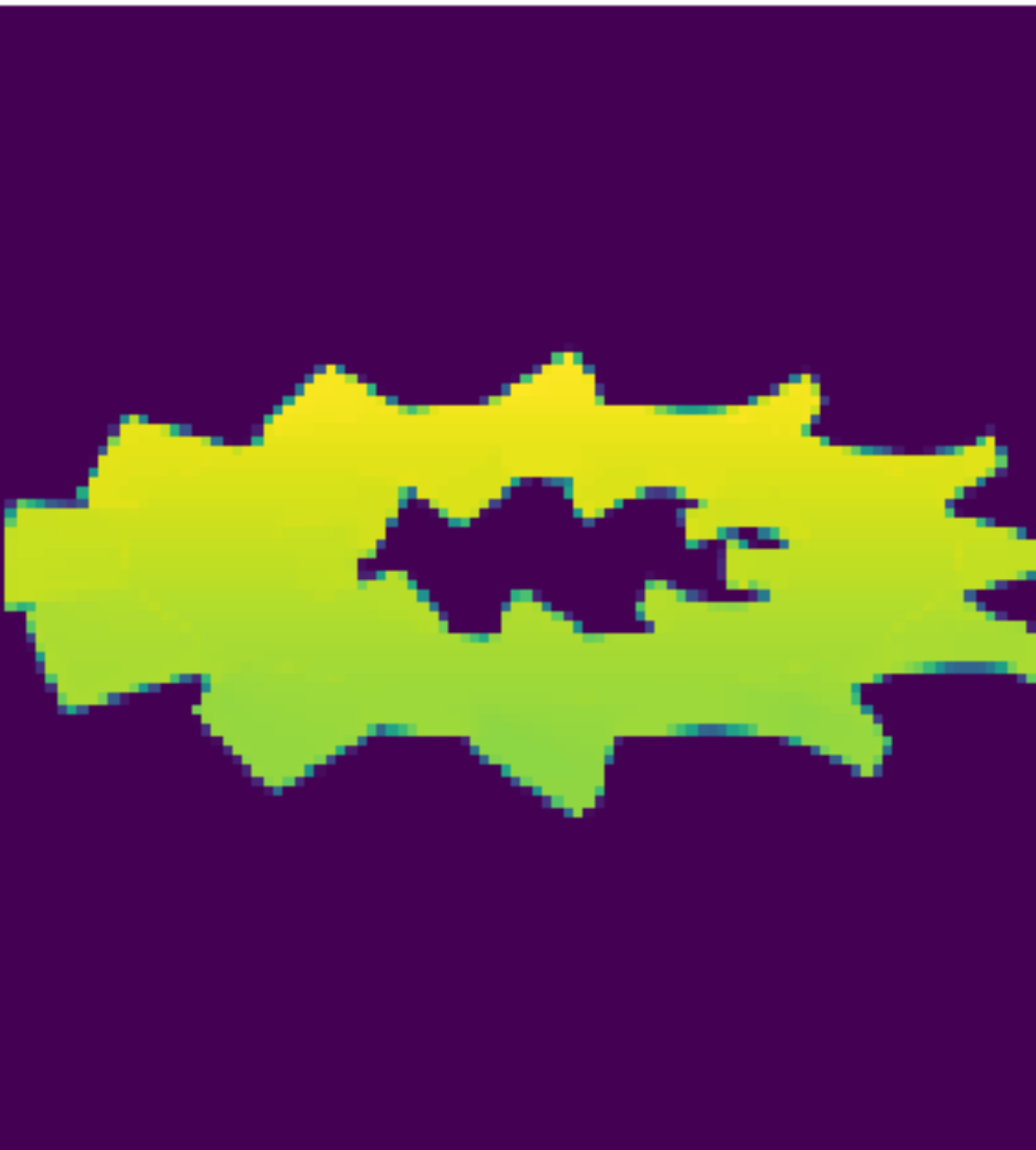} } & \raisebox{2\height}{\LARGE 2.34$^{\circ}$ } \\ \hline
			\includegraphics[trim={9cm 4cm 9cm 4cm}, clip = true,width=0.12\linewidth]{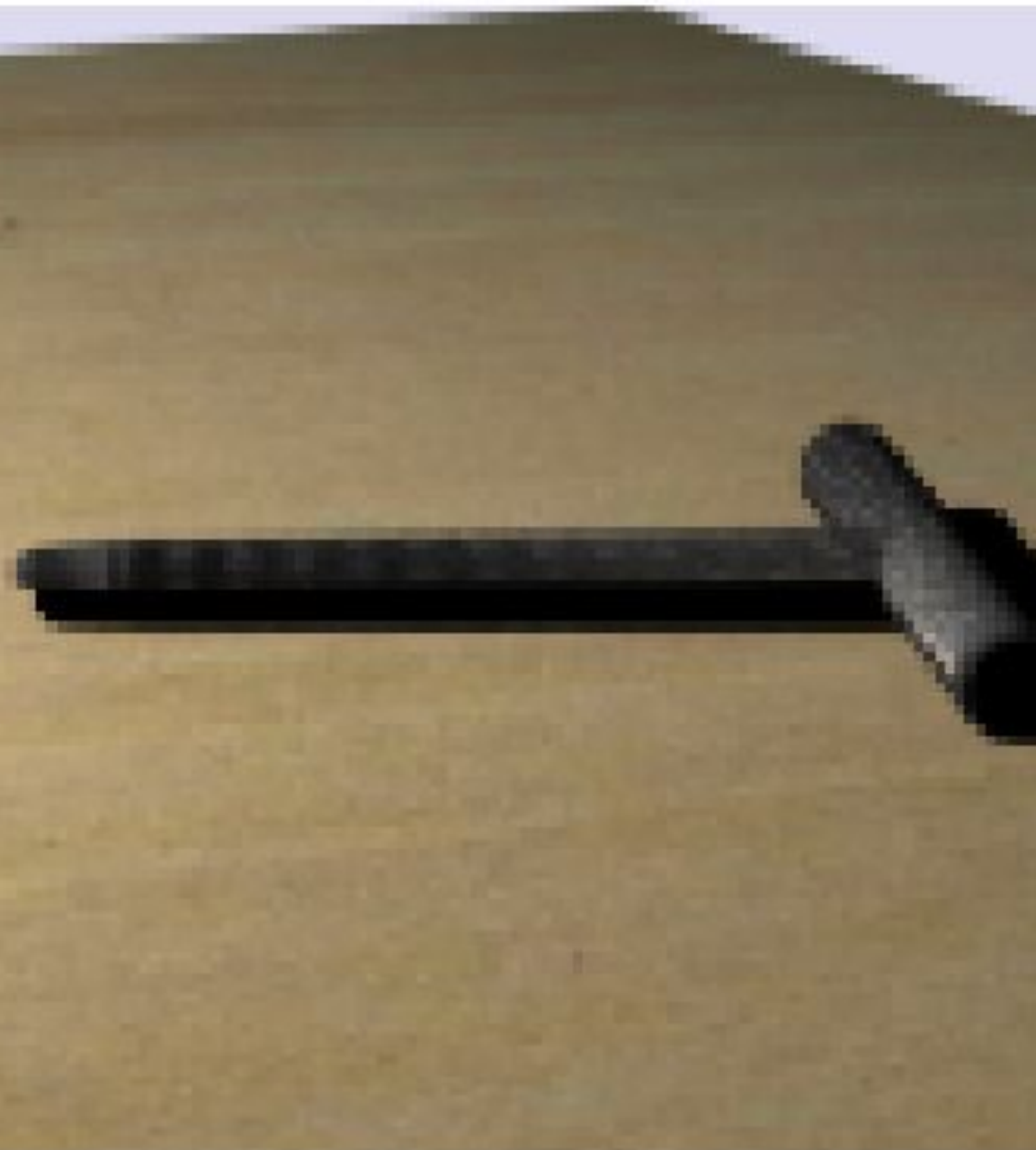} &
			\includegraphics[trim={9cm 4cm 9cm 4cm}, clip = true,width=0.12\linewidth]{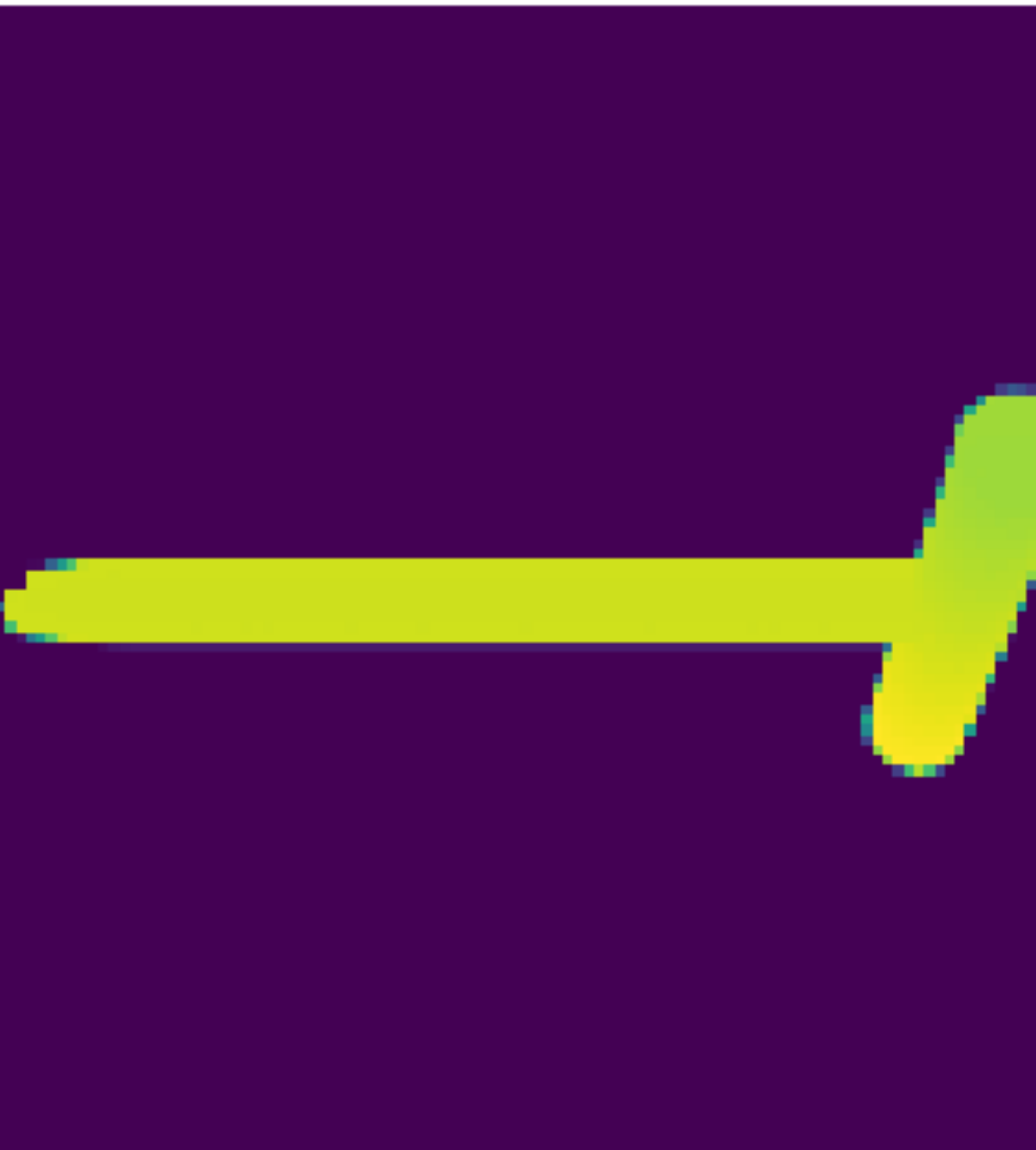} &
			\includegraphics[trim={9cm 4cm 9cm 4cm}, clip = true,width=0.12\linewidth]{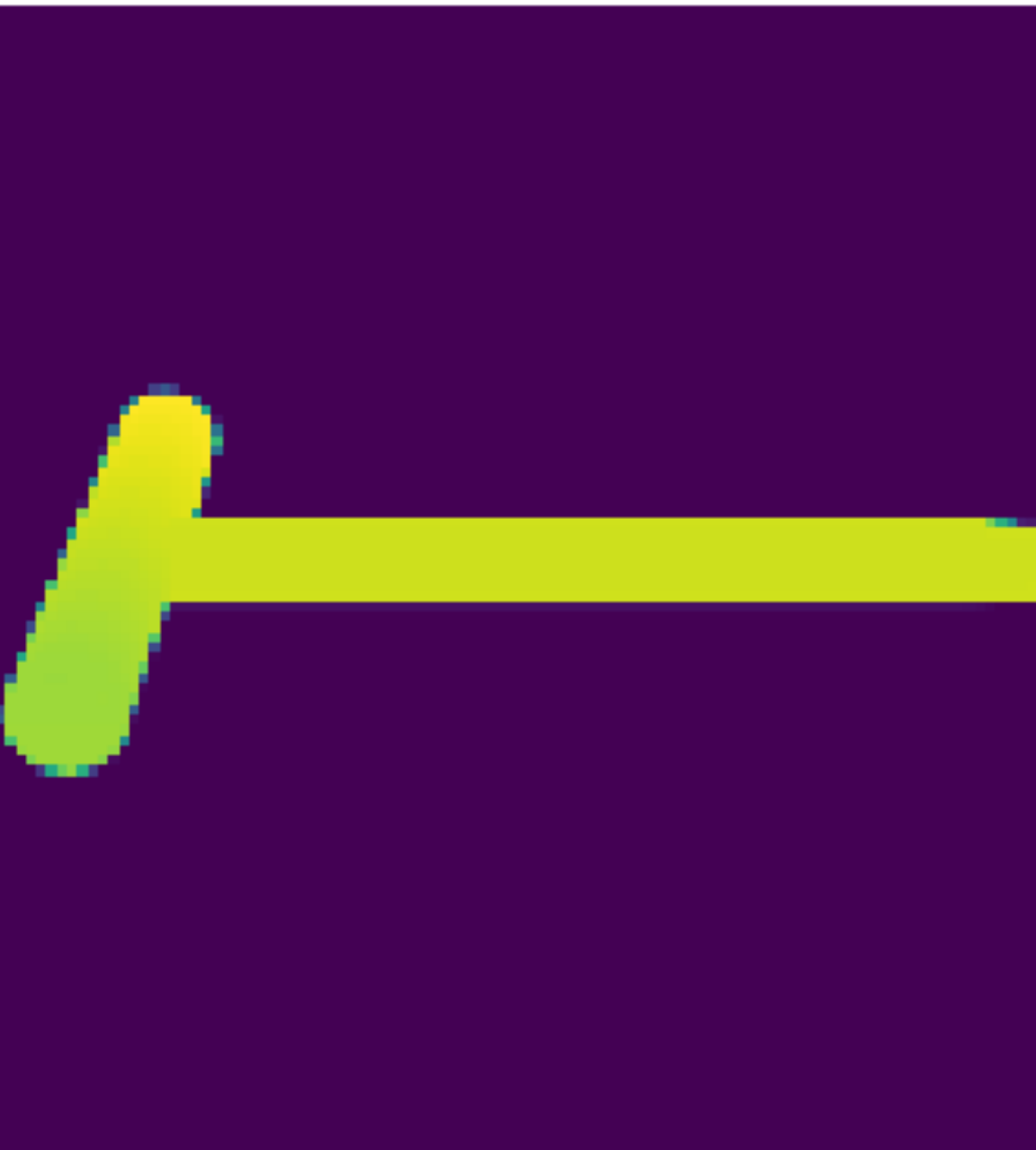} &
			\includegraphics[trim={9cm 4cm 9cm 4cm}, clip = true,width=0.12\linewidth]{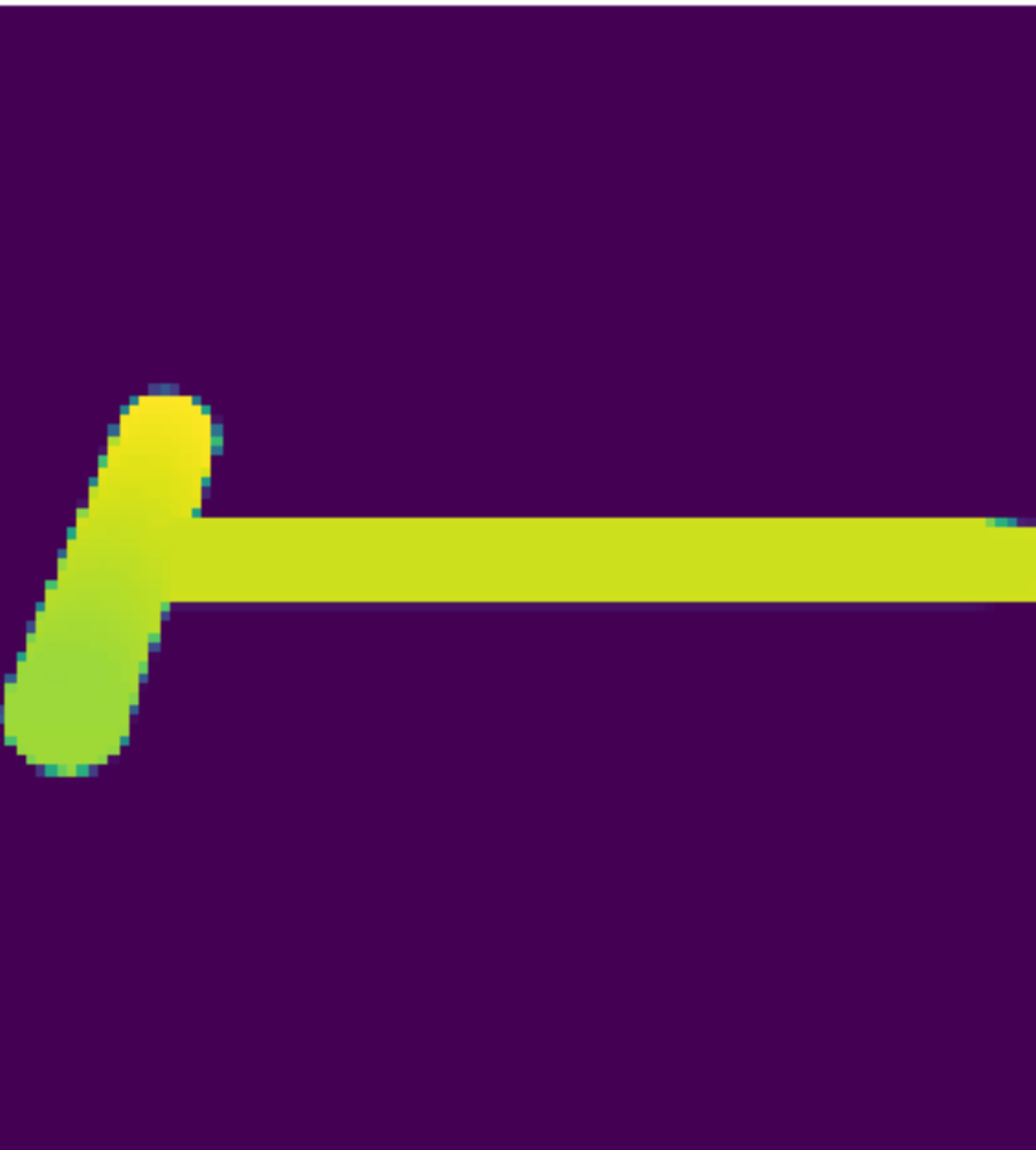} &
			\includegraphics[trim={9cm 4cm 9cm 4cm}, clip = true,width=0.12\linewidth]{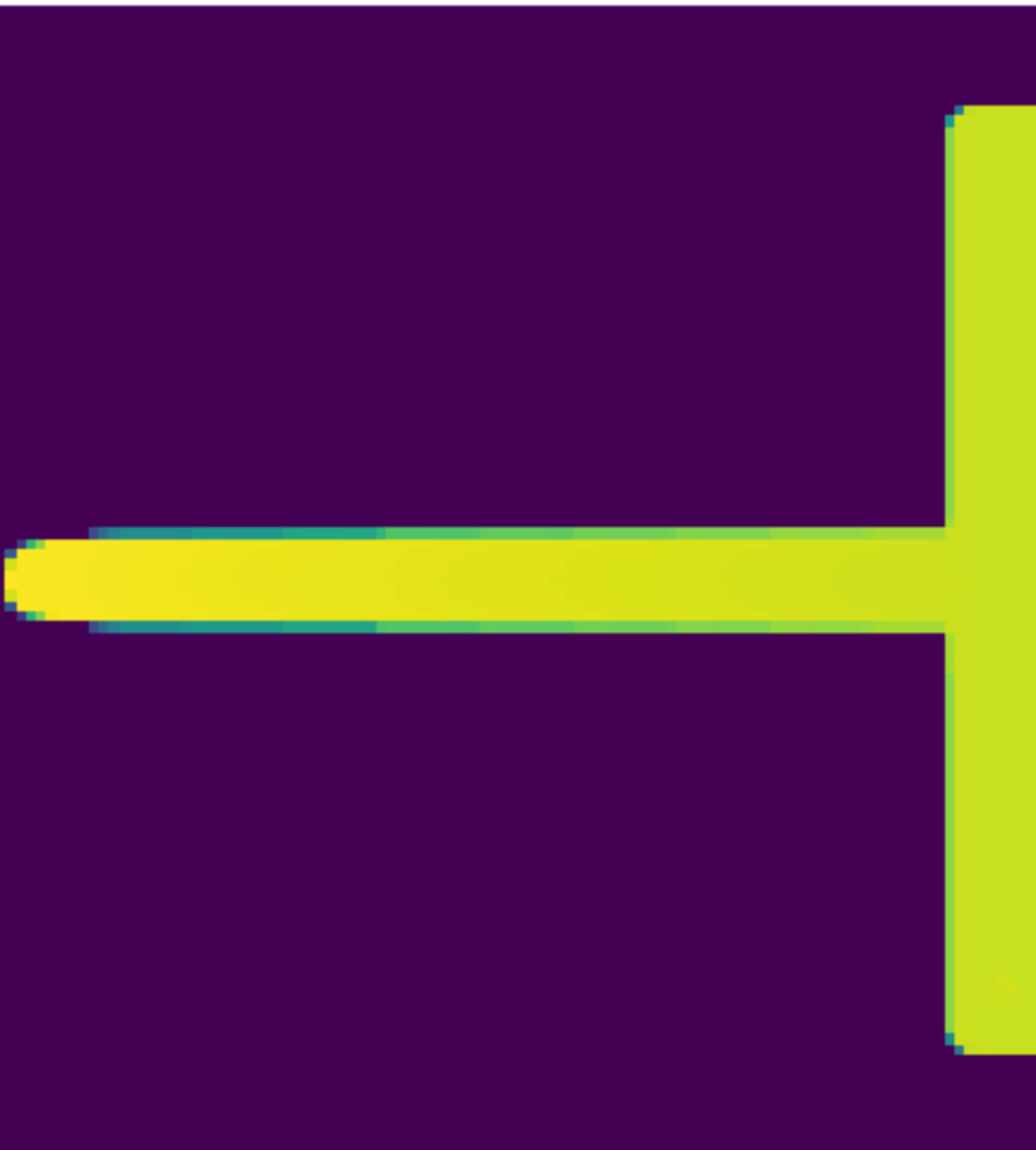} &
			\fcolorbox{green}{white}{\includegraphics[trim={9cm 4cm 9cm 4cm}, clip = true,width=0.12\linewidth]{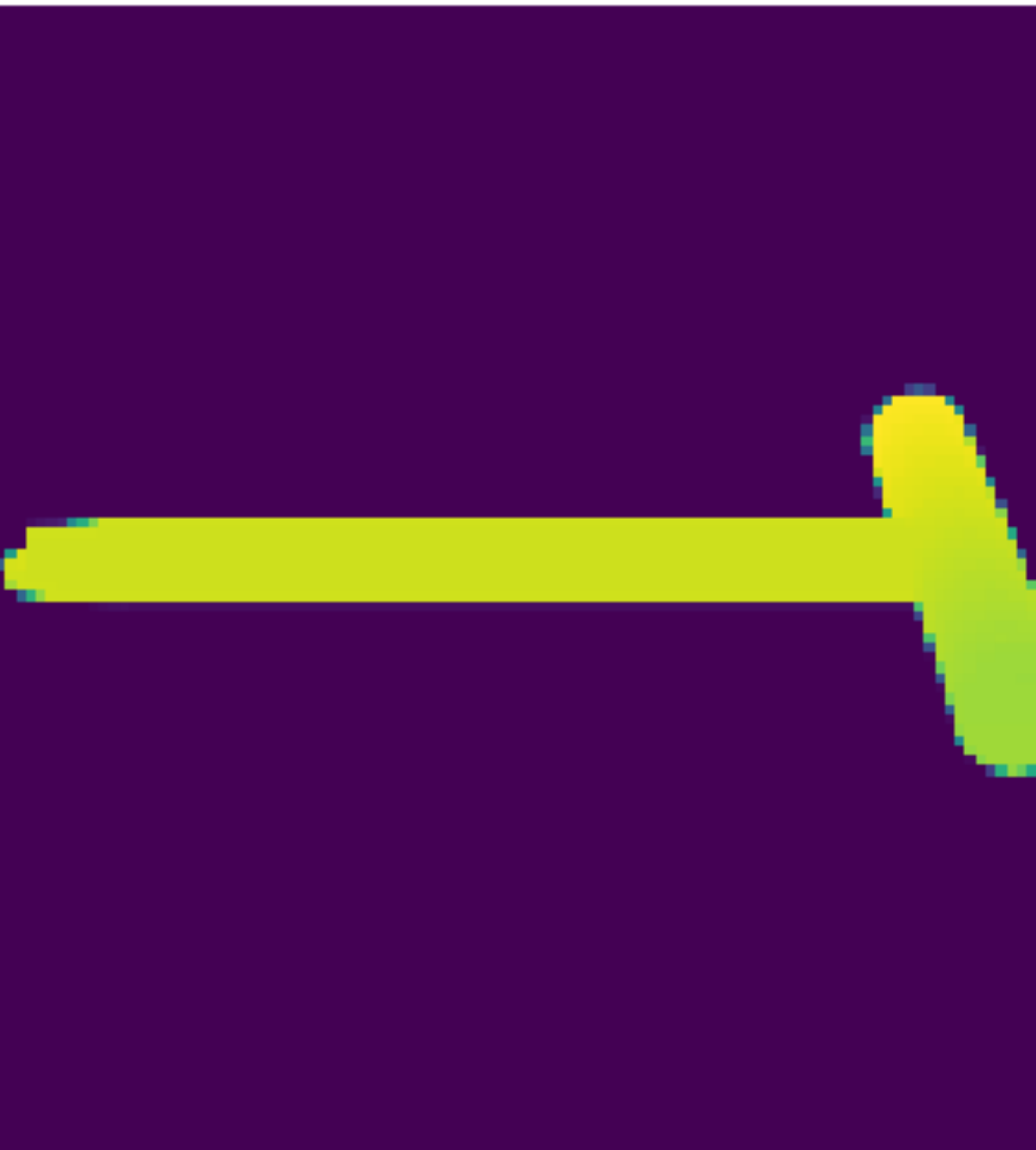}} & \raisebox{2\height}{\LARGE 2.35$^{\circ}$ } \\ \hline
			\includegraphics[trim={9cm 4cm 9cm 4cm}, clip = true,width=0.12\linewidth]{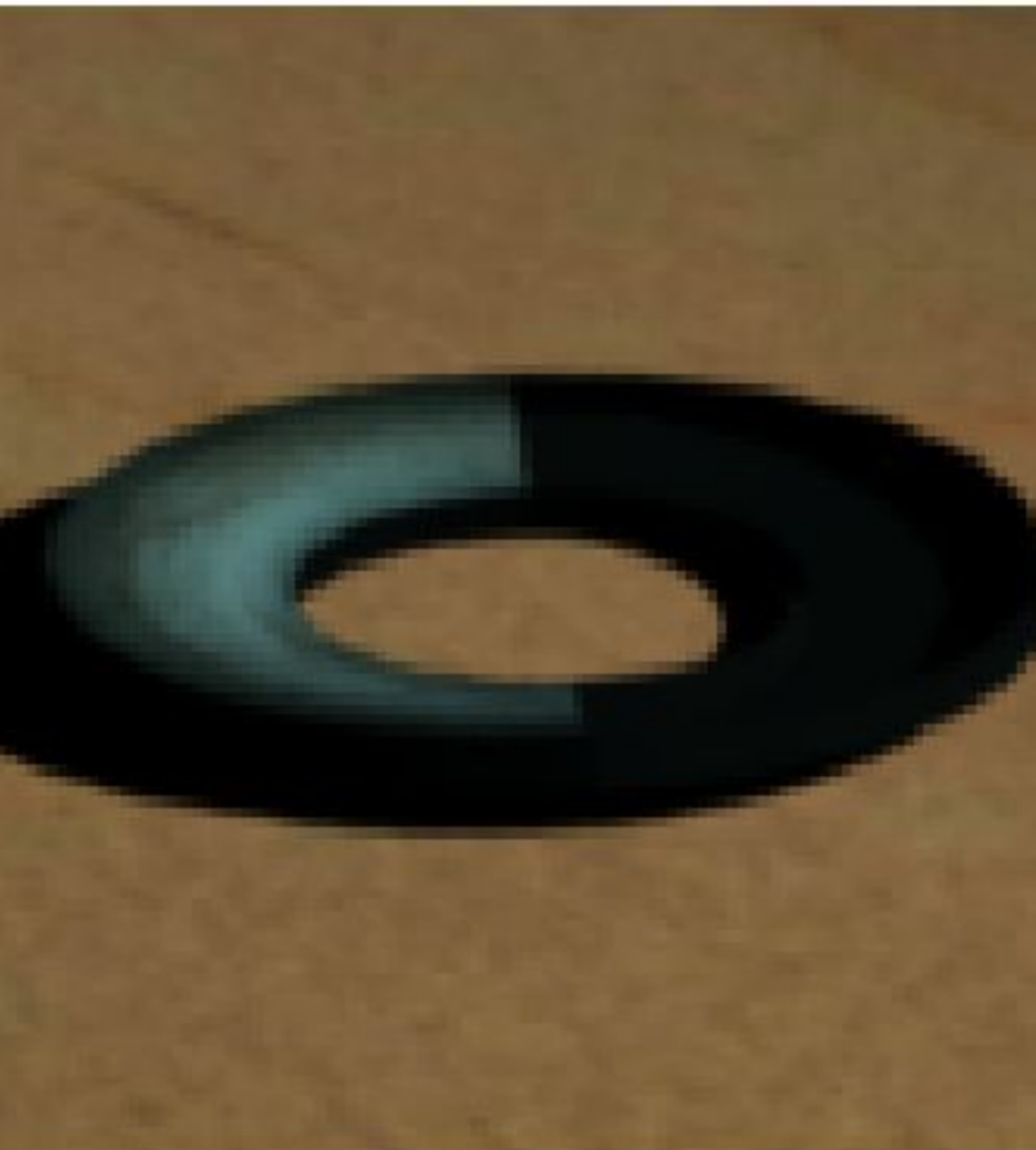} &
			\includegraphics[trim={9cm 4cm 9cm 4cm}, clip = true,width=0.12\linewidth]{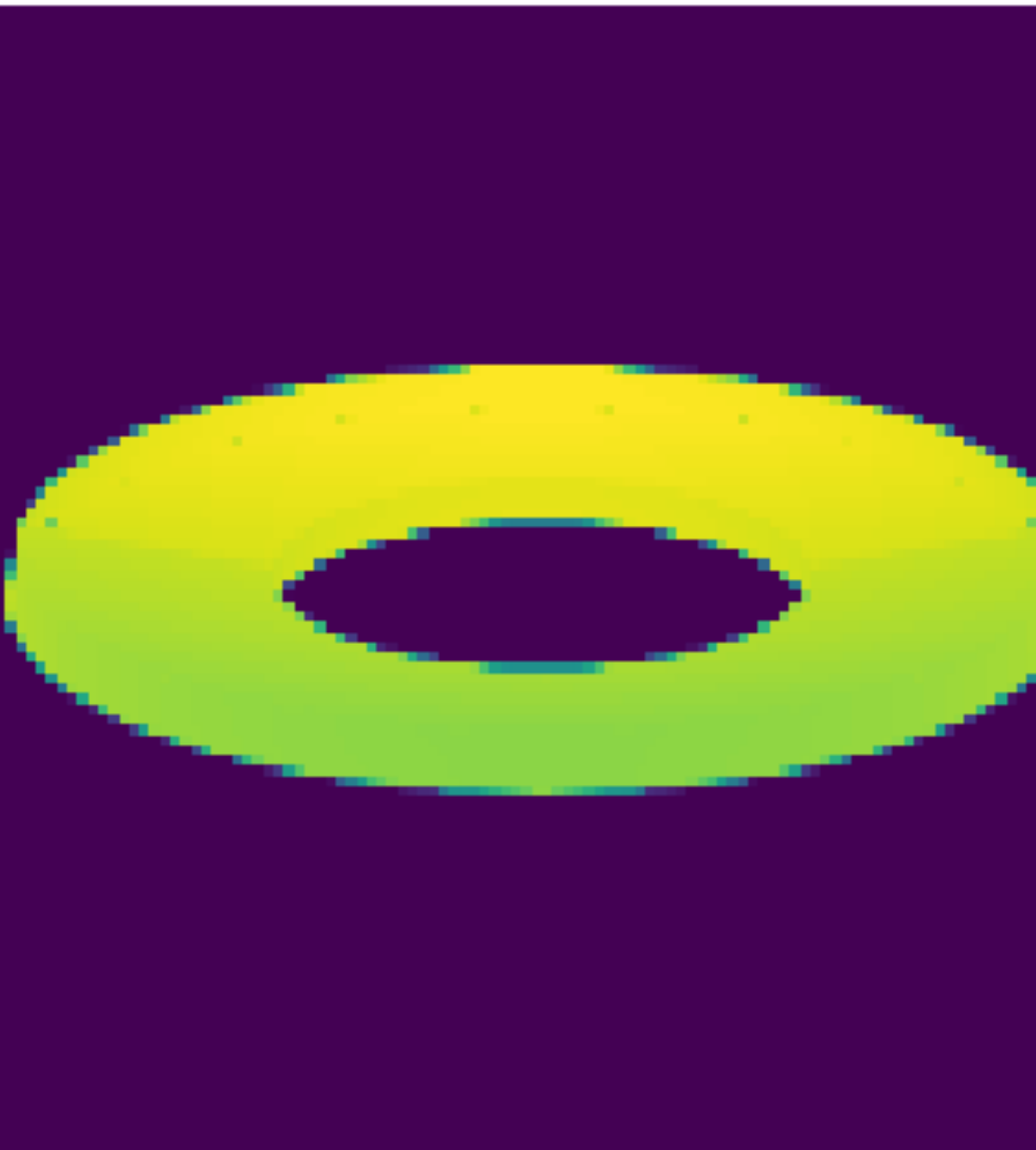} &
			\includegraphics[trim={9cm 4cm 9cm 4cm}, clip = true,width=0.12\linewidth]{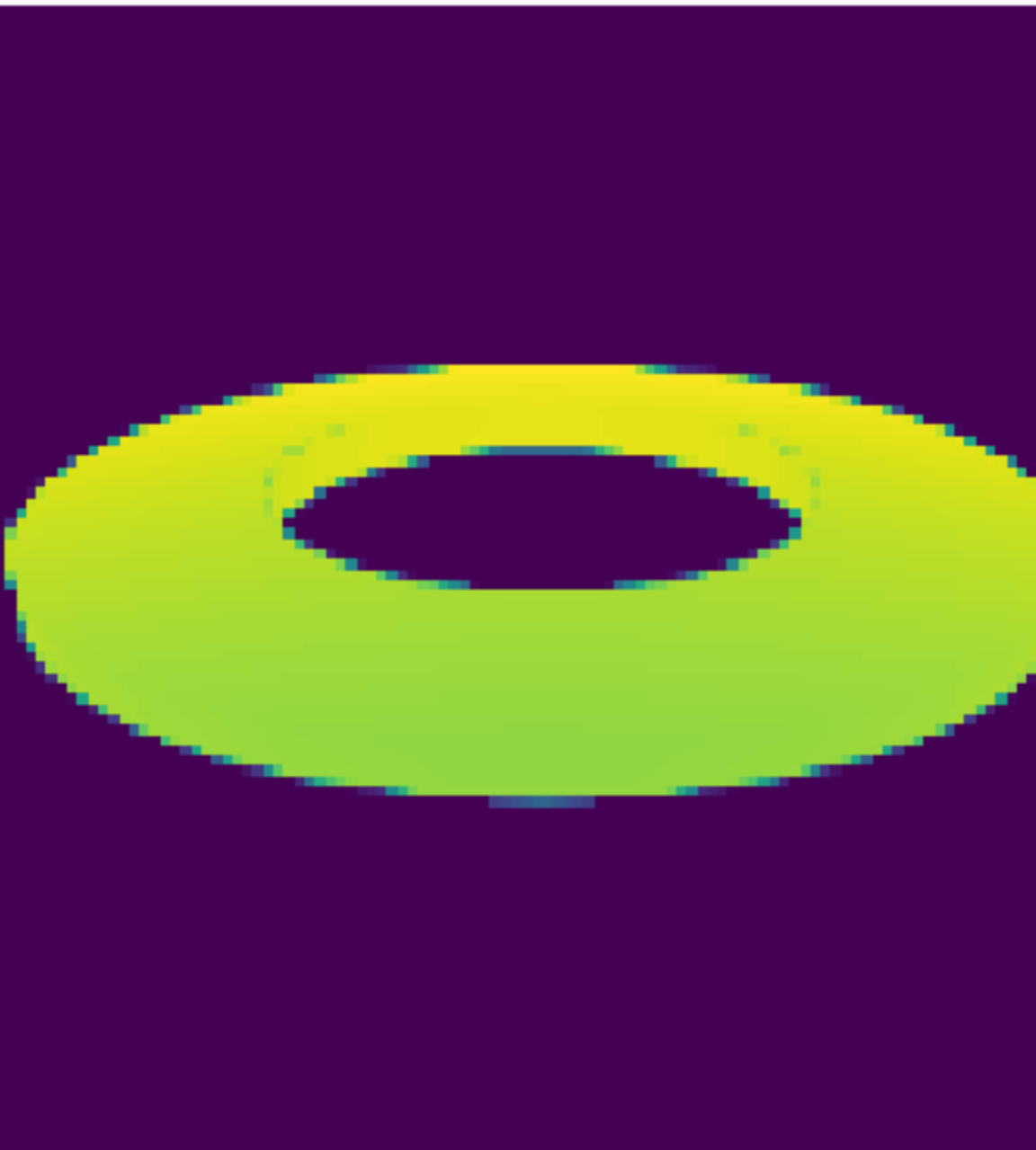} &
			\includegraphics[trim={9cm 4cm 9cm 4cm}, clip = true,width=0.12\linewidth]{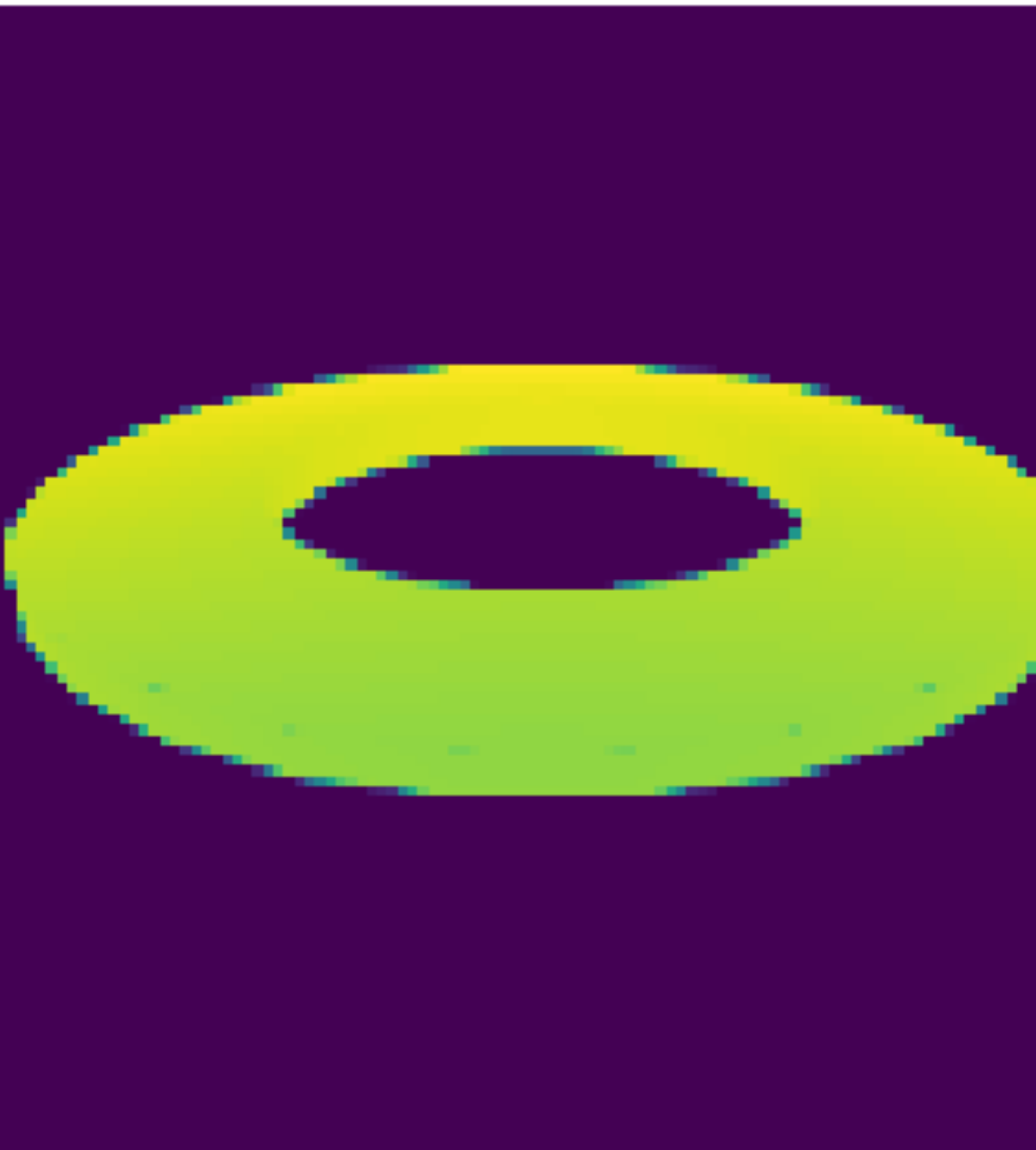} &
			\fcolorbox{green}{white}{\includegraphics[trim={9cm 4cm 9cm 4cm}, clip = true,width=0.12\linewidth]{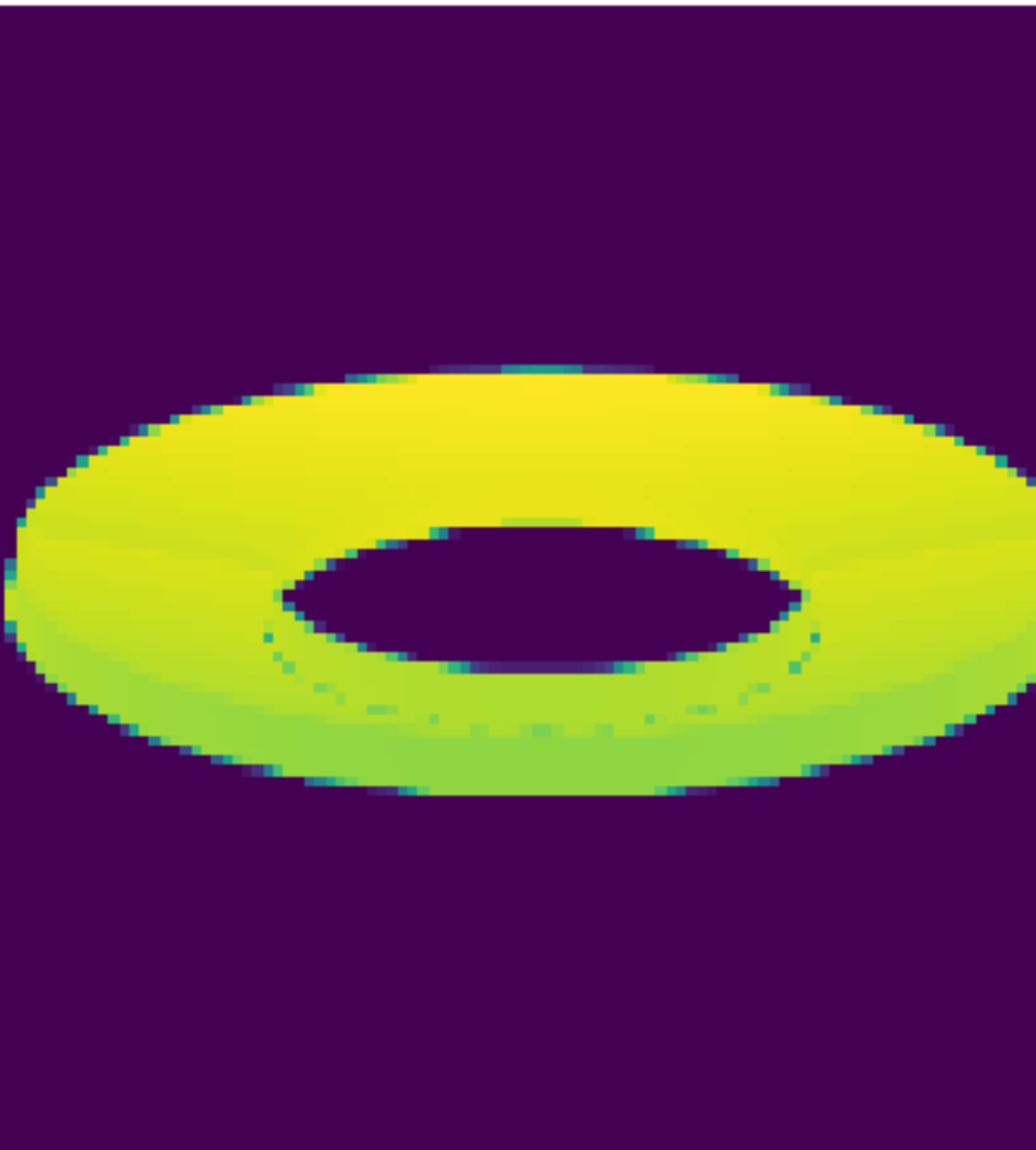}} &
			\fcolorbox{green}{white}{\includegraphics[trim={9cm 4cm 9cm 4cm}, clip = true,width=0.12\linewidth]{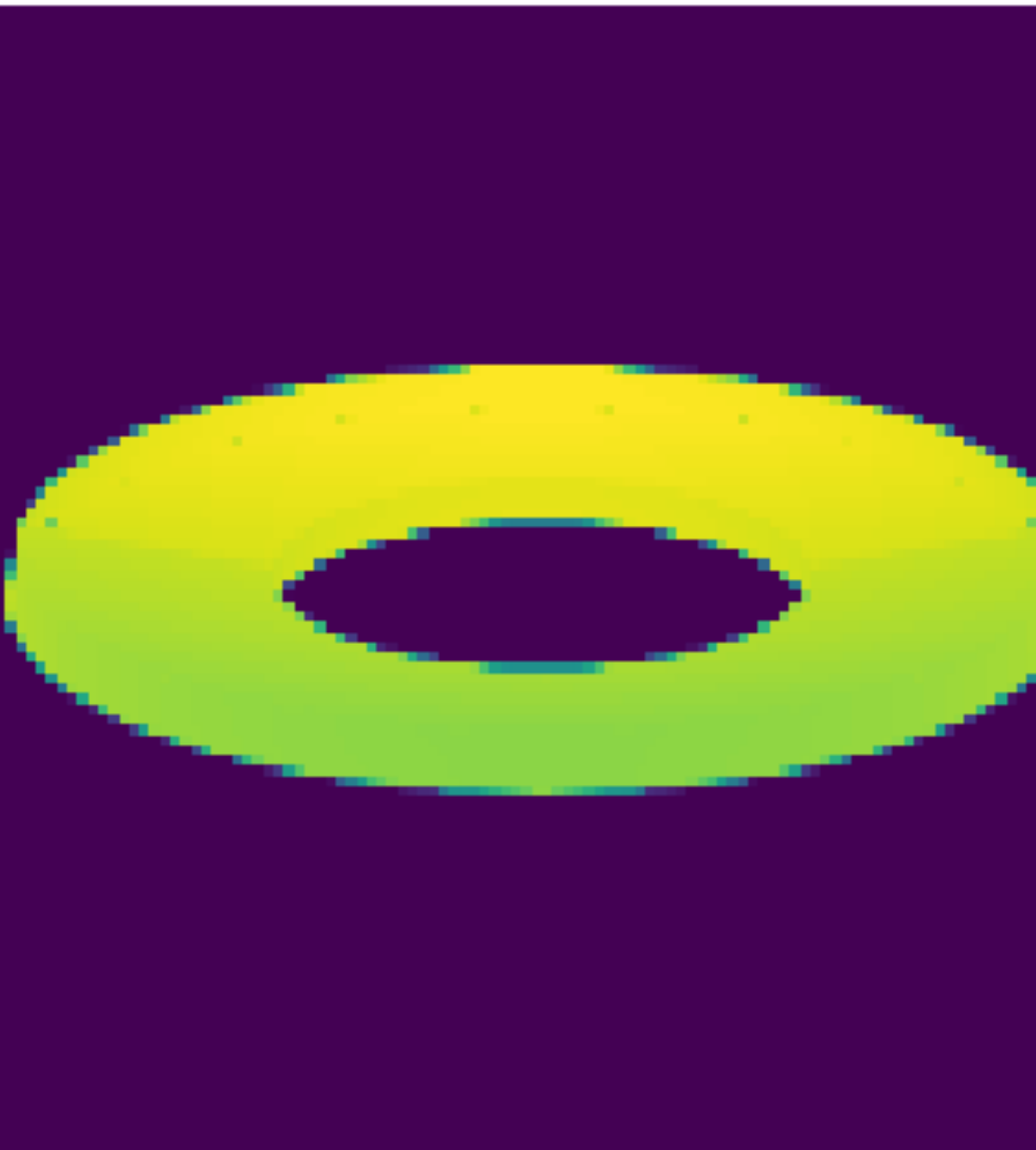}} & \raisebox{2\height}{\LARGE 2.79$^{\circ}$ } \\ \hline
			\includegraphics[trim={9cm 4cm 9cm 4cm}, clip = true,width=0.12\linewidth]{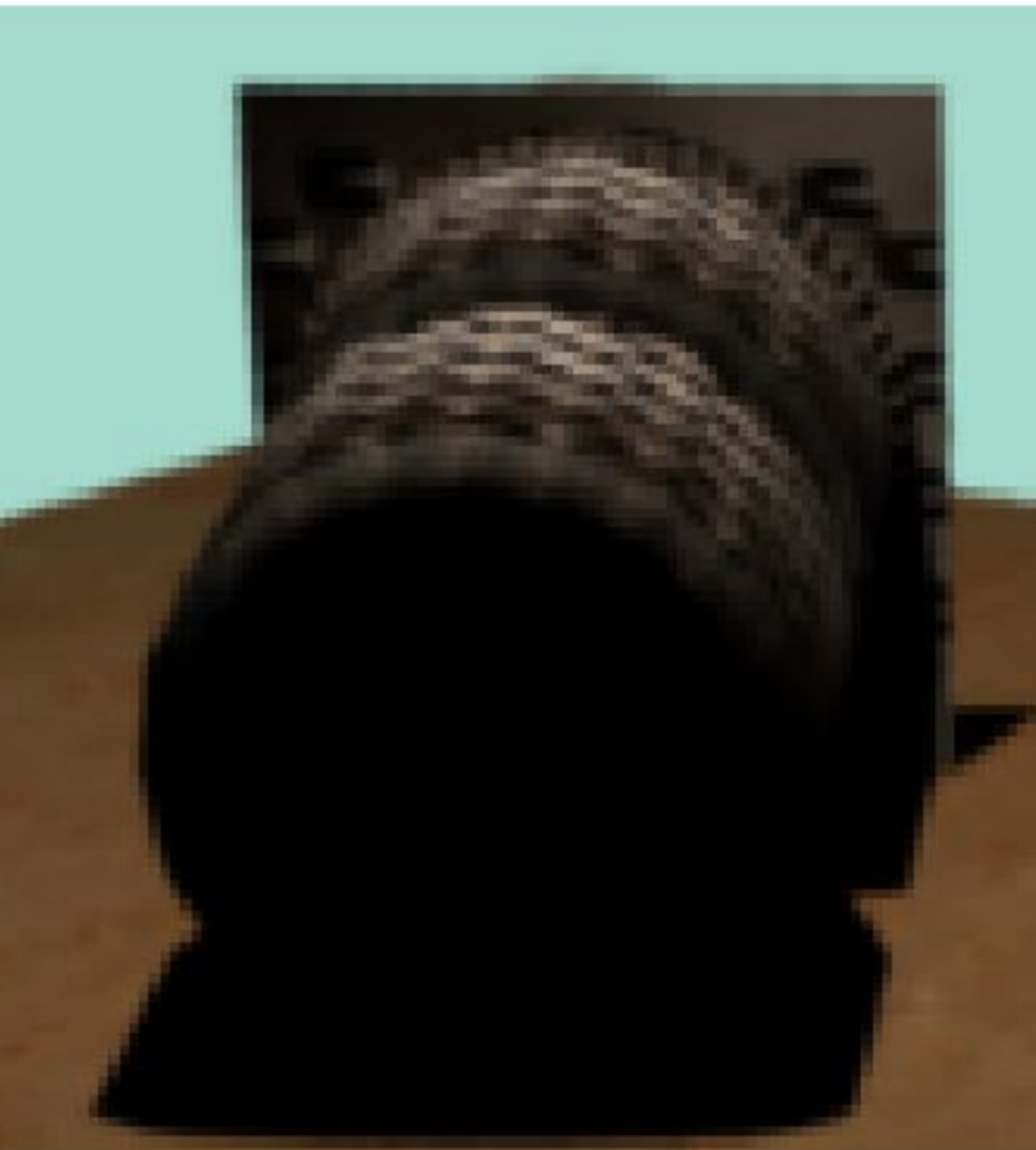} &
			\includegraphics[trim={9cm 4cm 9cm 4cm}, clip = true,width=0.12\linewidth]{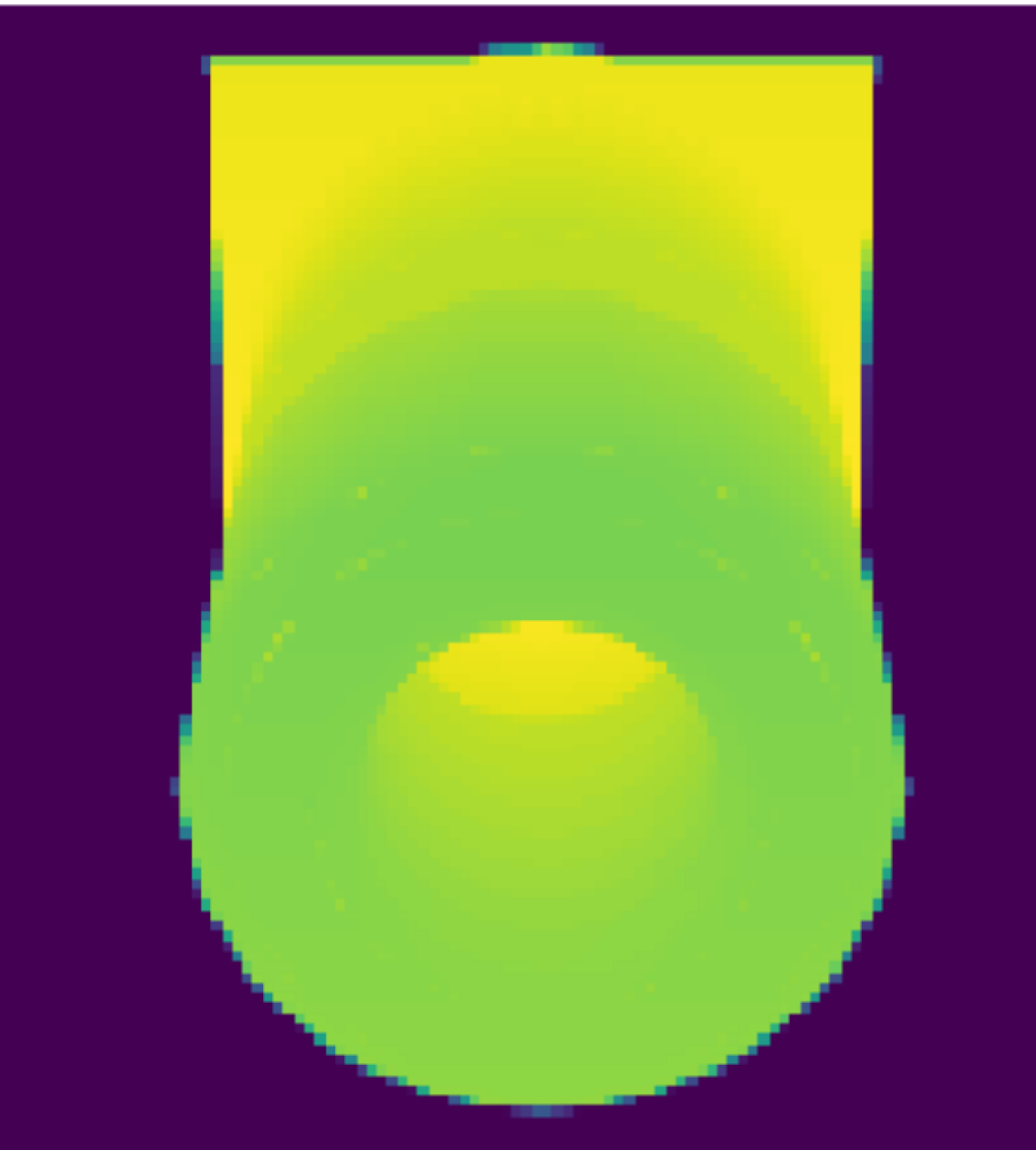} &
			\includegraphics[trim={9cm 4cm 9cm 4cm}, clip = true,width=0.12\linewidth]{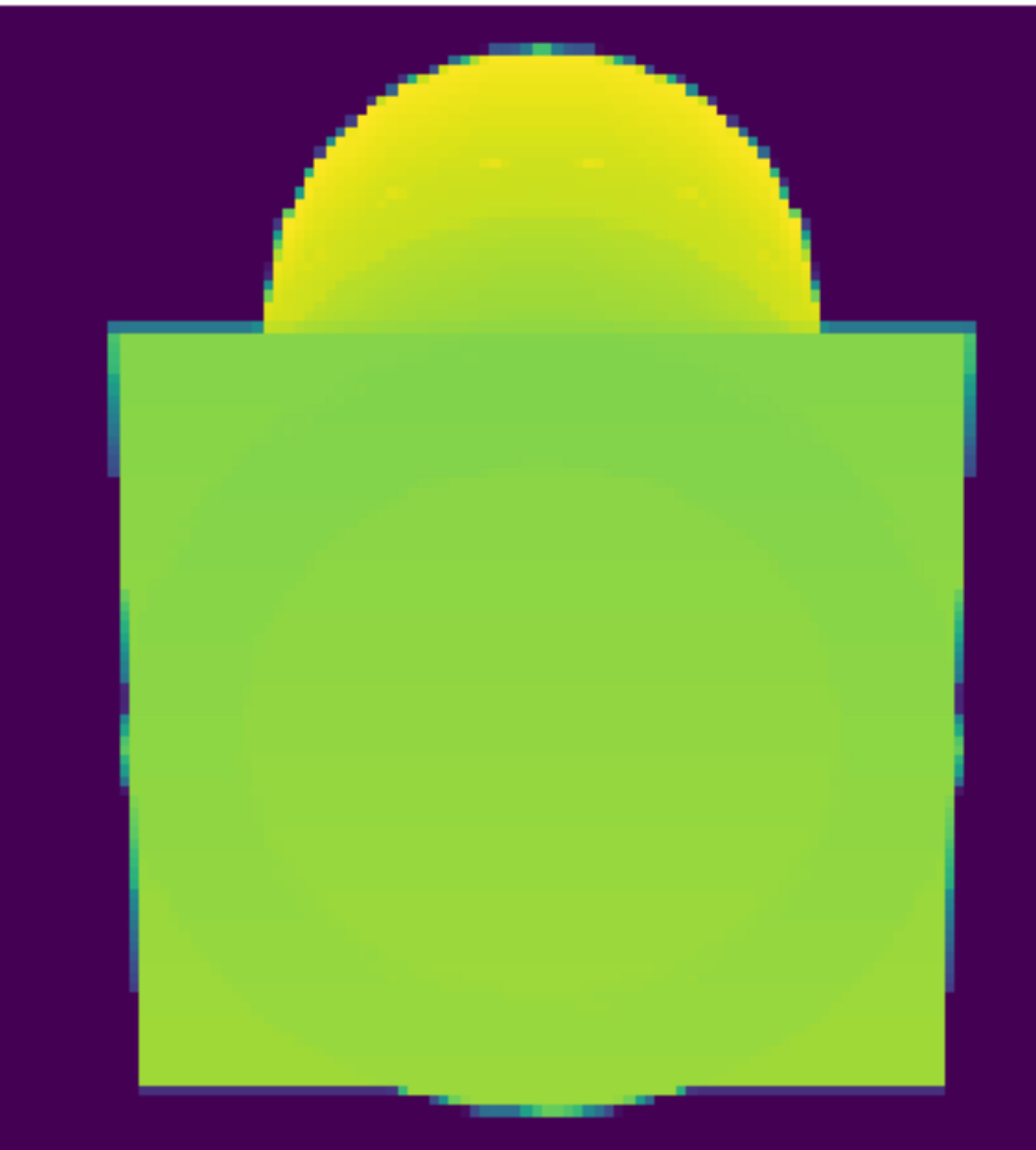} &
			\includegraphics[trim={9cm 4cm 9cm 4cm}, clip = true,width=0.12\linewidth]{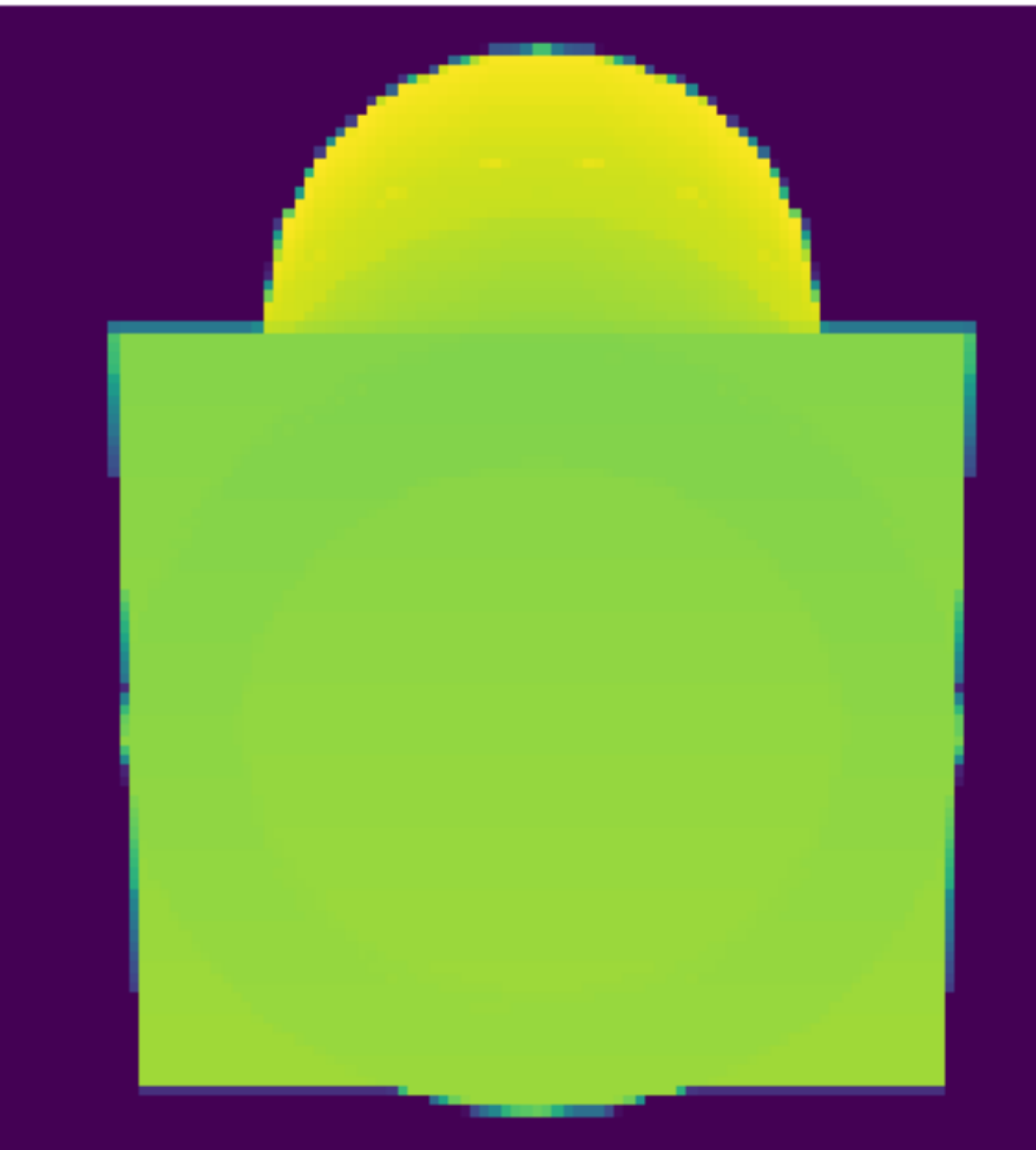} &
			\fcolorbox{green}{white}{\includegraphics[trim={9cm 4cm 9cm 4cm}, clip = true,width=0.12\linewidth]{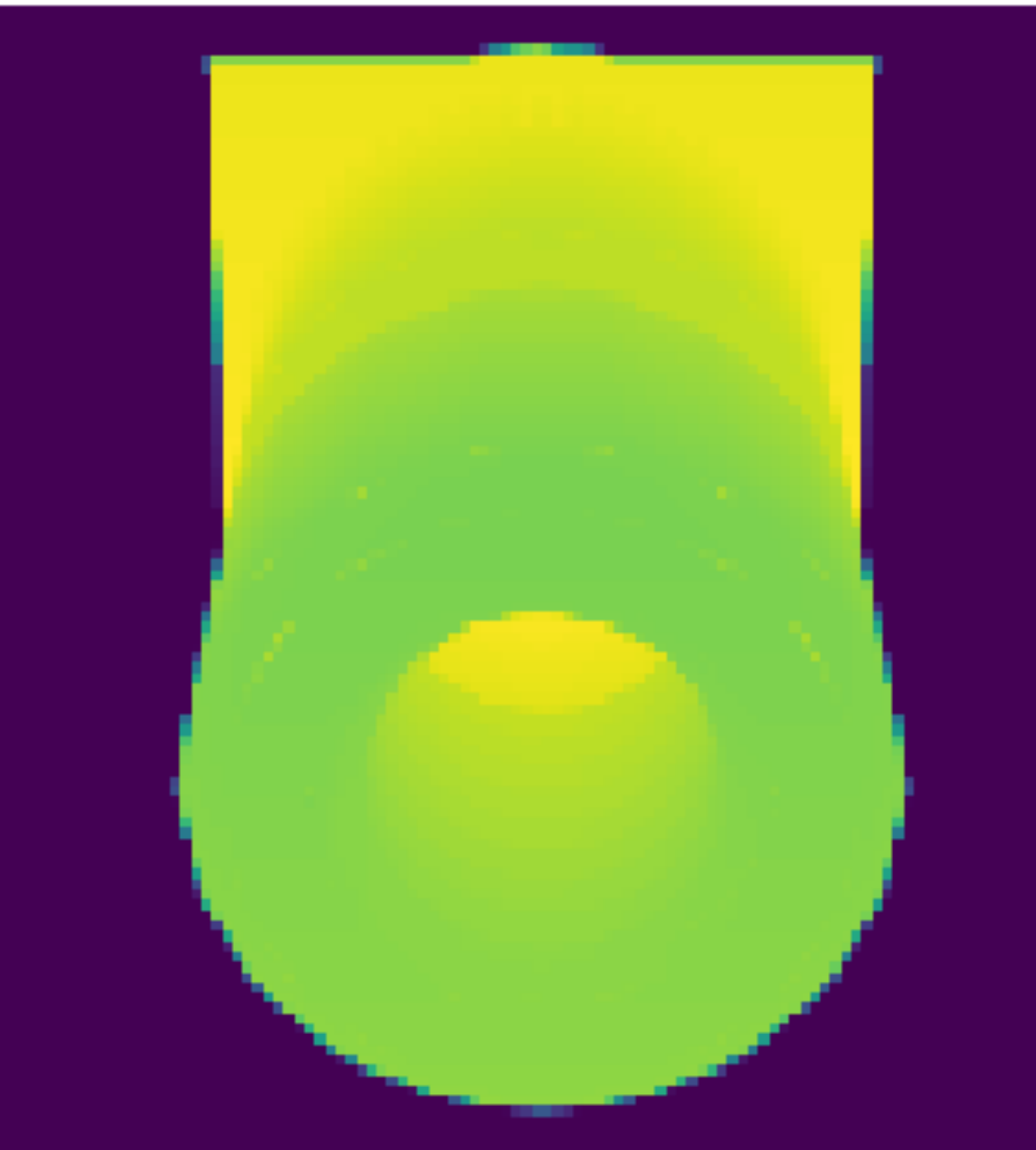}} &
			\fcolorbox{green}{white}{\includegraphics[trim={9cm 4cm 9cm 4cm}, clip = true,width=0.12\linewidth]{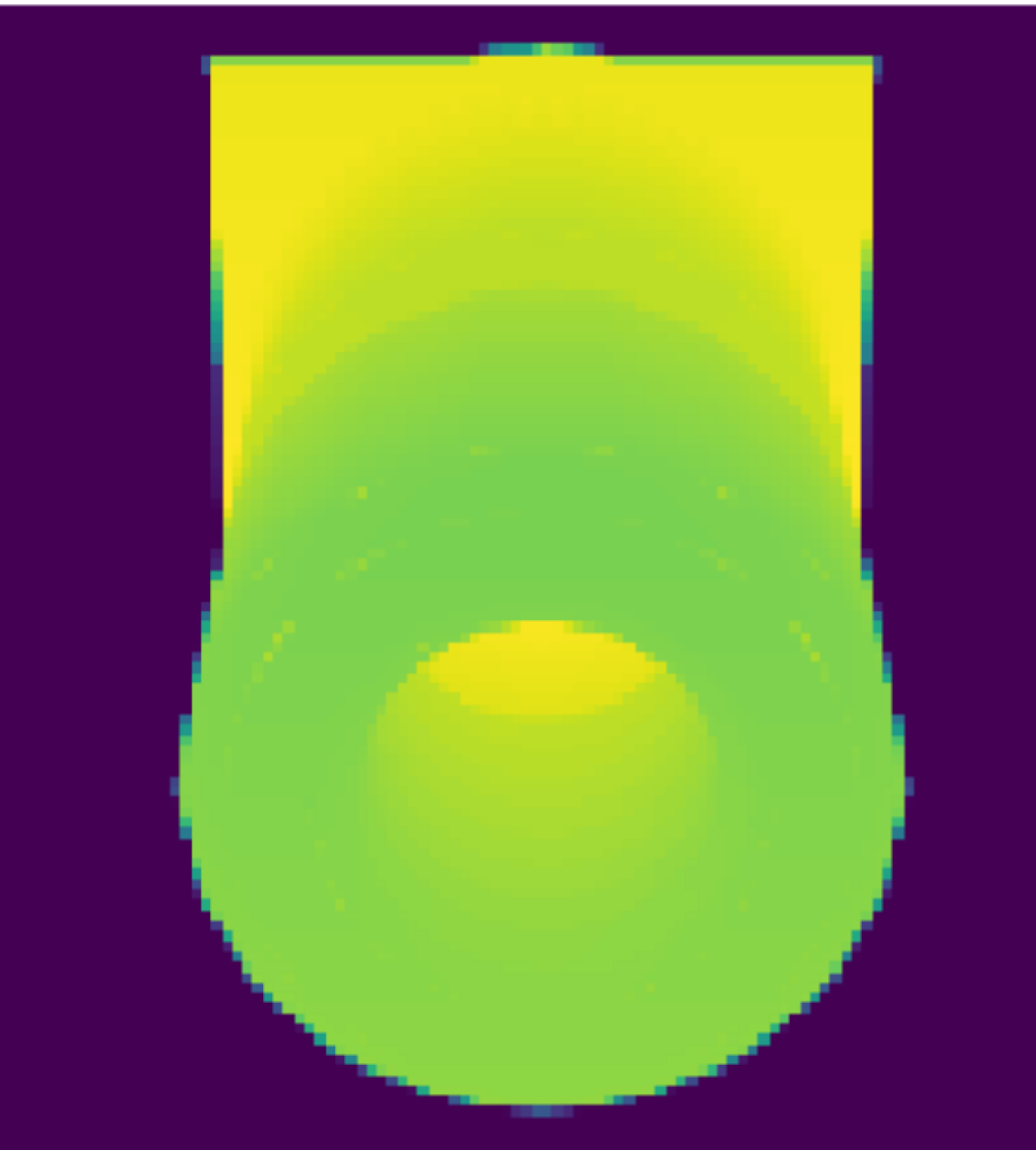}} & \raisebox{2\height}{\LARGE 3.08$^{\circ}$ } \\ \hline
			\includegraphics[trim={9cm 4cm 9cm 4cm}, clip = true,width=0.12\linewidth]{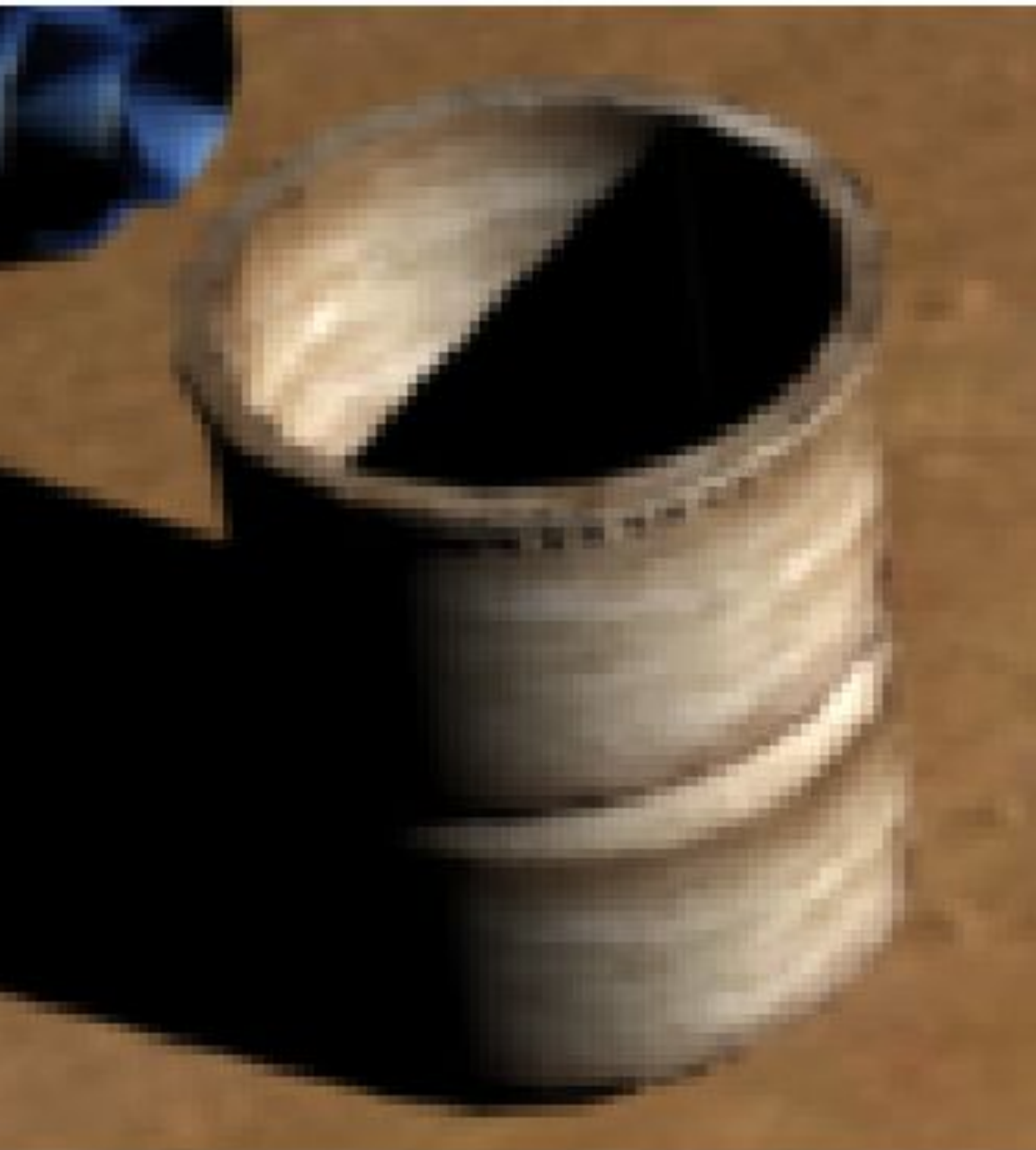} &
			\includegraphics[trim={9cm 4cm 9cm 4cm}, clip = true,width=0.12\linewidth]{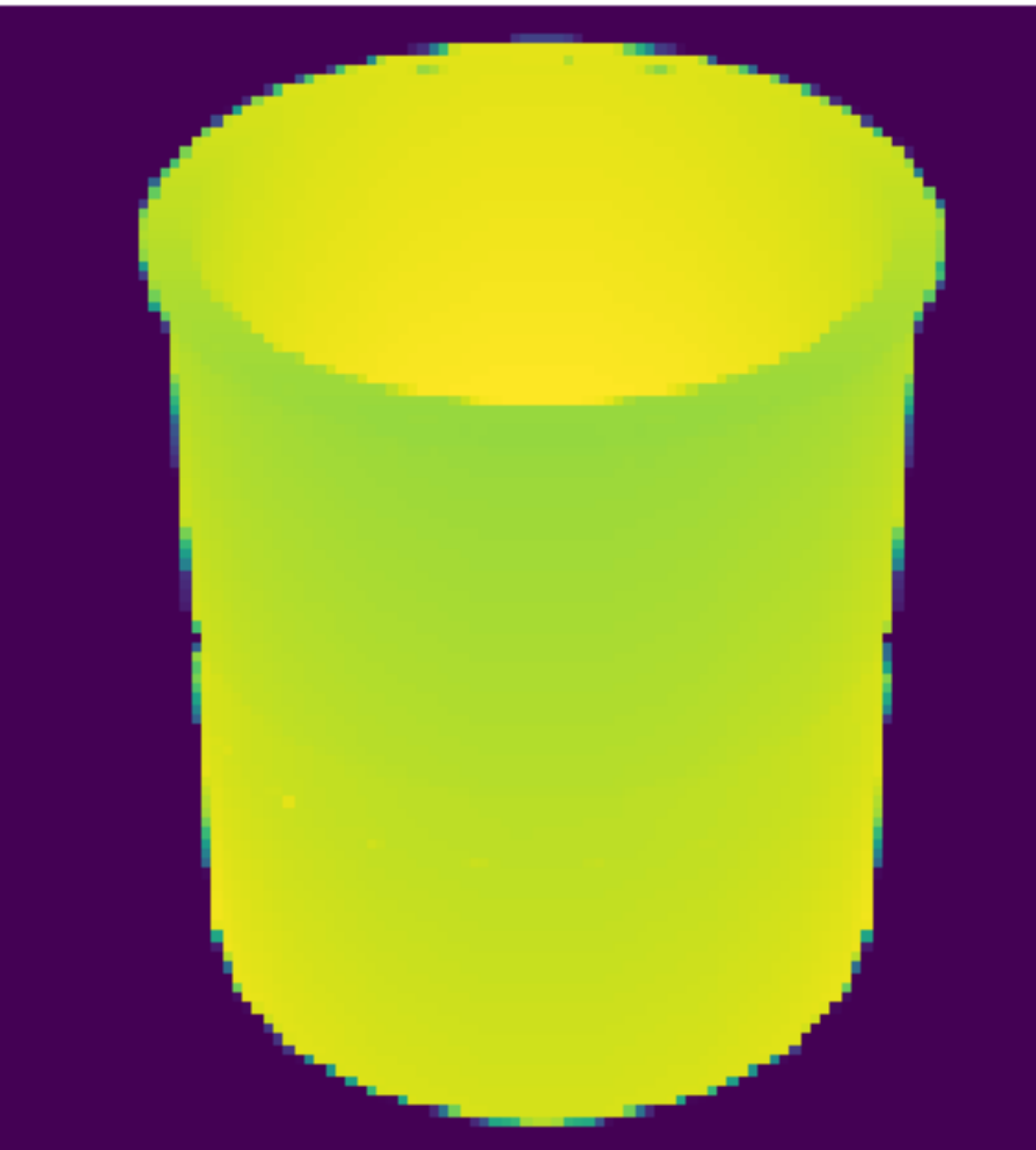} &
			\includegraphics[trim={9cm 4cm 9cm 4cm}, clip = true,width=0.12\linewidth]{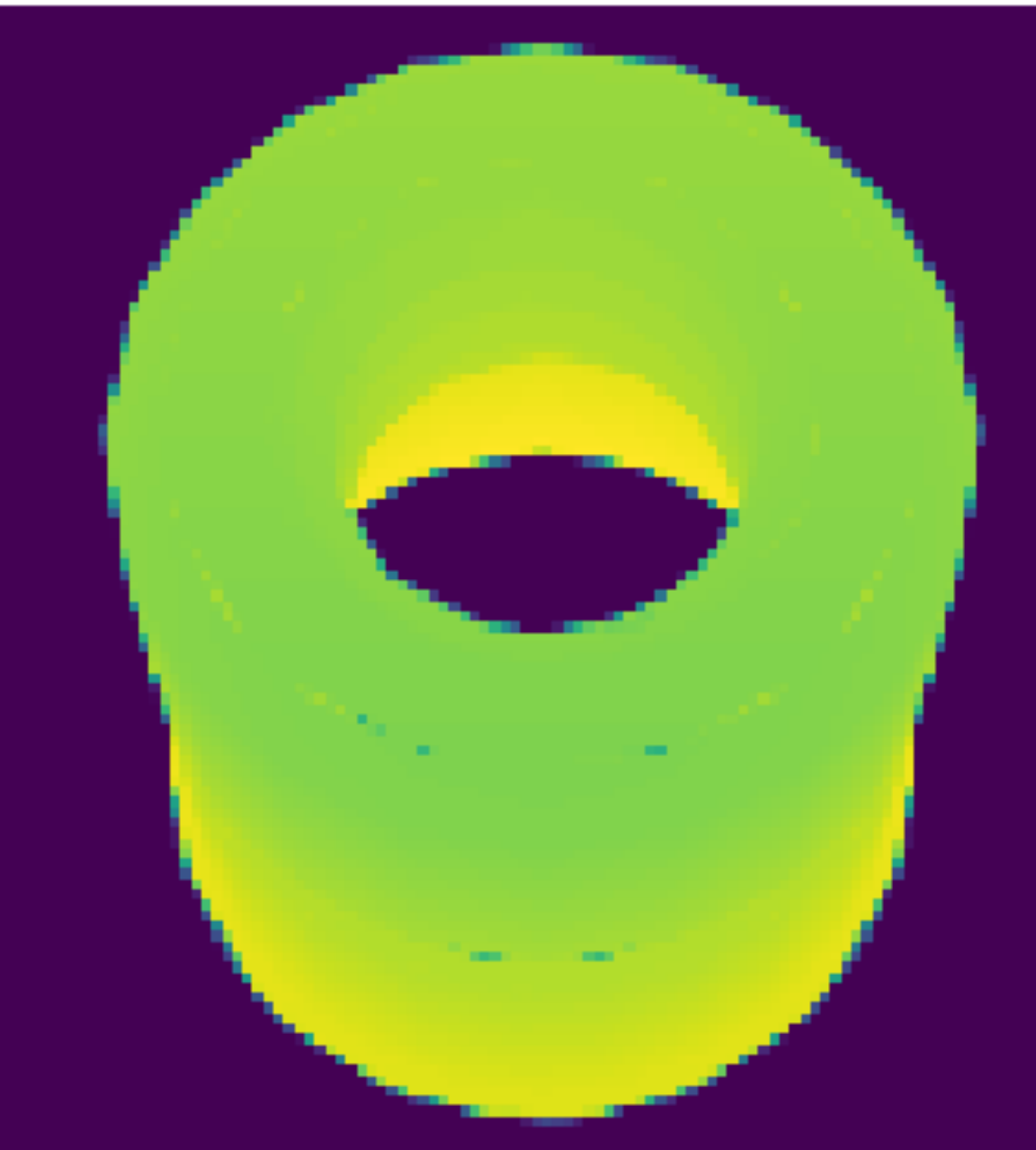} &
			\includegraphics[trim={9cm 4cm 9cm 4cm}, clip = true,width=0.12\linewidth]{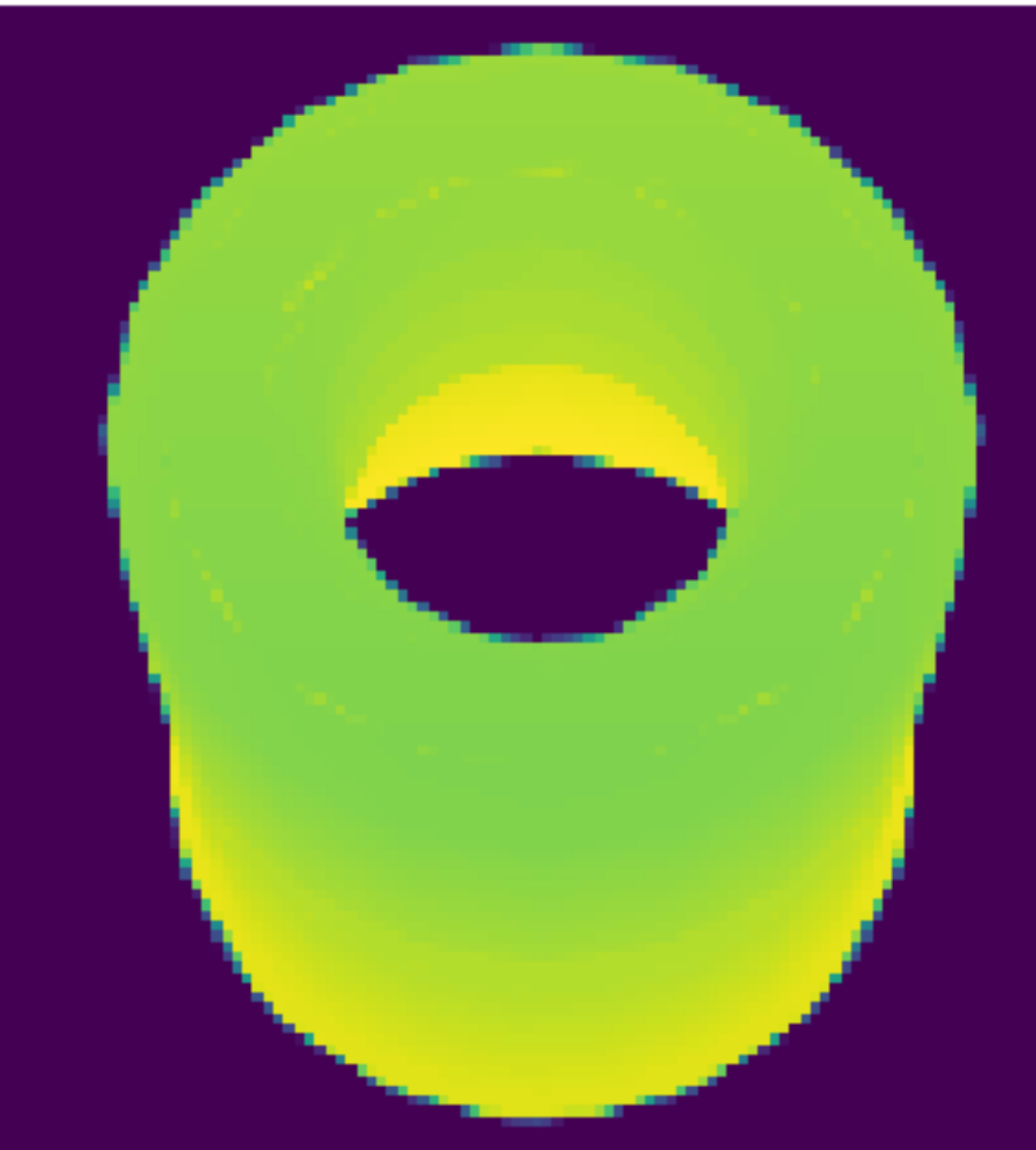} &
			\fcolorbox{green}{white}{\includegraphics[trim={9cm 4cm 9cm 4cm}, clip = true,width=0.12\linewidth]{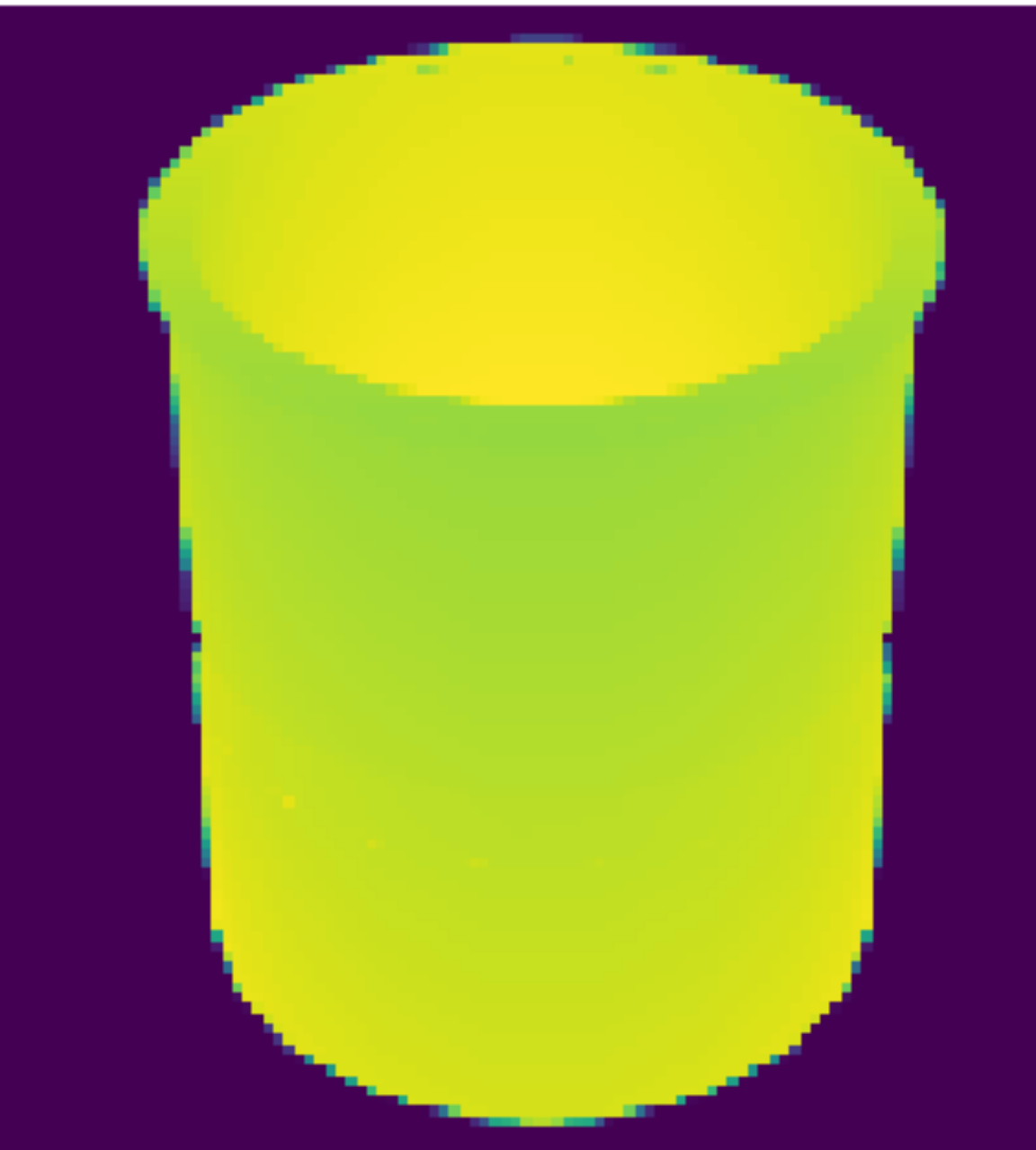}} &
			\fcolorbox{green}{white}{\includegraphics[trim={9cm 4cm 9cm 4cm}, clip = true,width=0.12\linewidth]{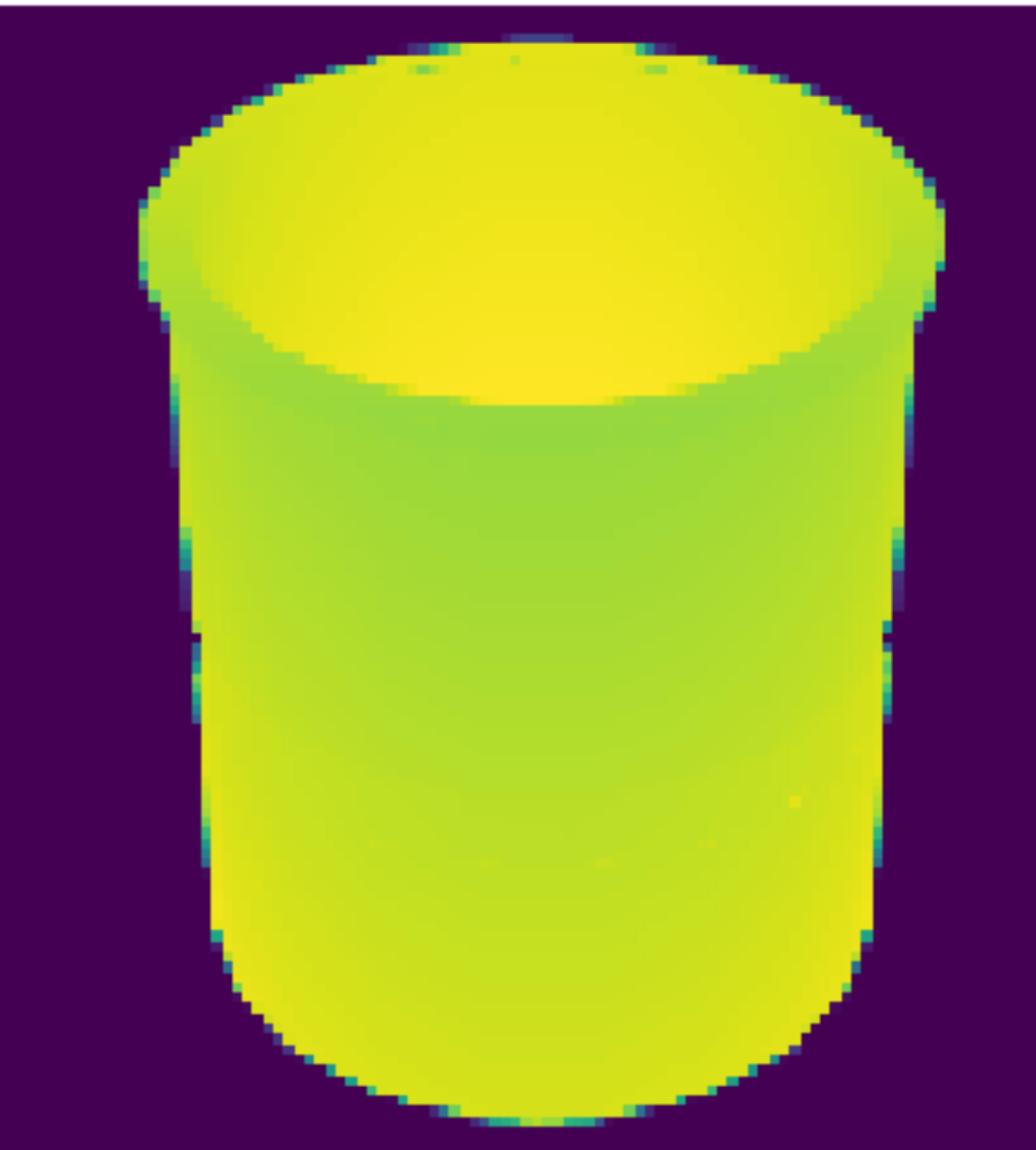}} & \raisebox{2\height}{\LARGE 3.34$^{\circ}$ } \\ \hline
			\includegraphics[trim={9cm 4cm 9cm 4cm}, clip = true,width=0.12\linewidth]{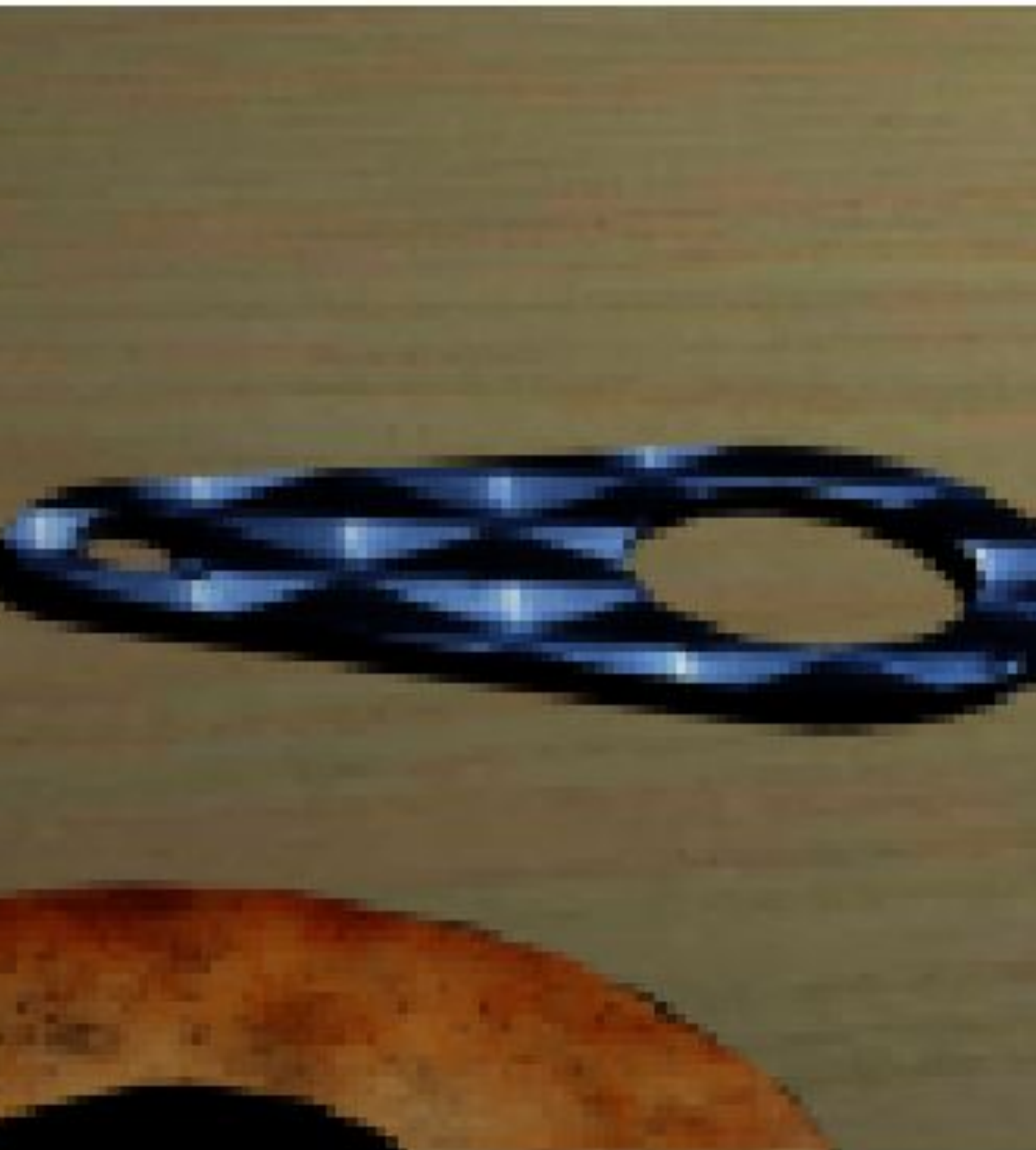} &
			\includegraphics[trim={9cm 4cm 9cm 4cm}, clip = true,width=0.12\linewidth]{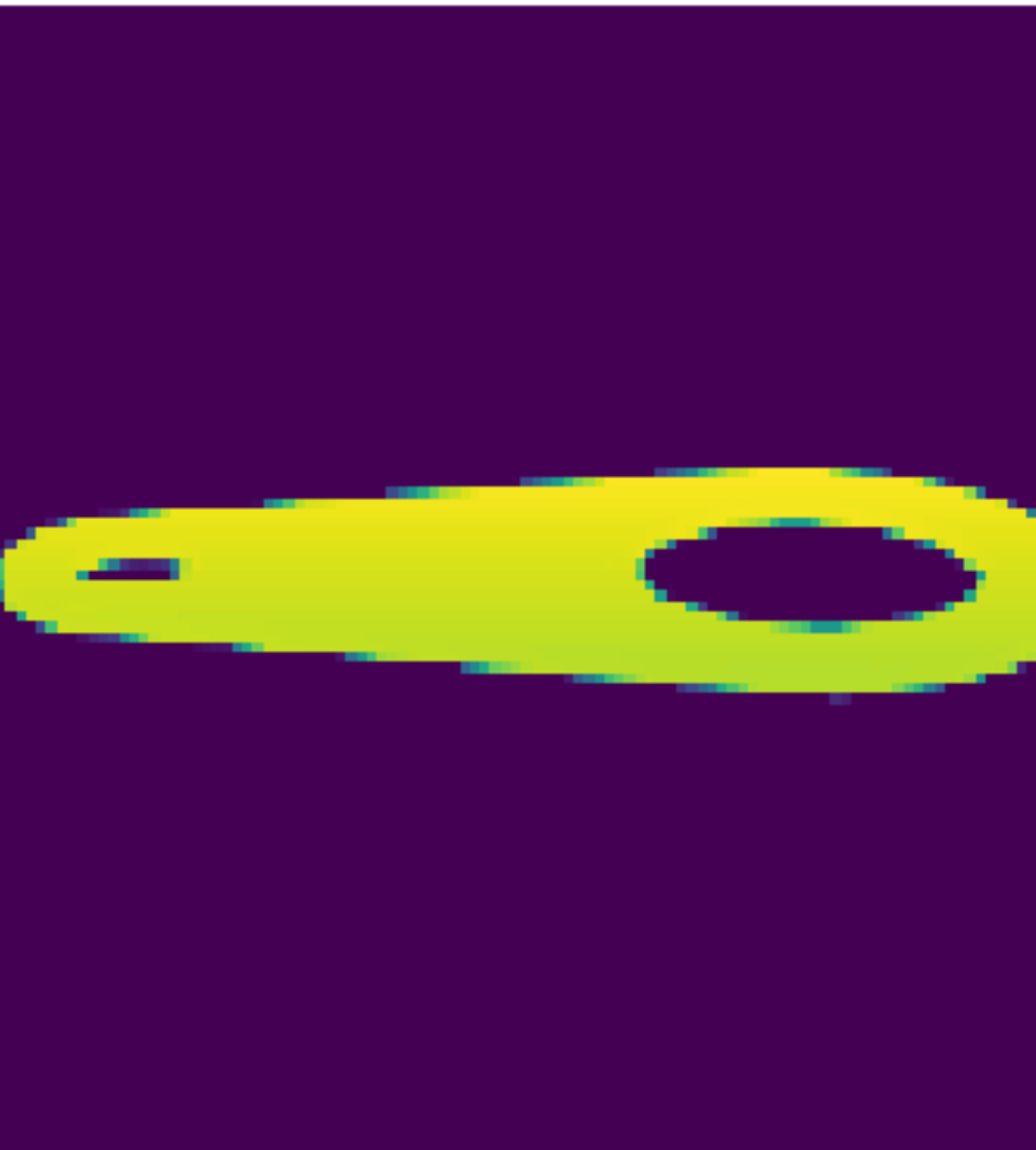} &
			\includegraphics[trim={9cm 4cm 9cm 4cm}, clip = true,width=0.12\linewidth]{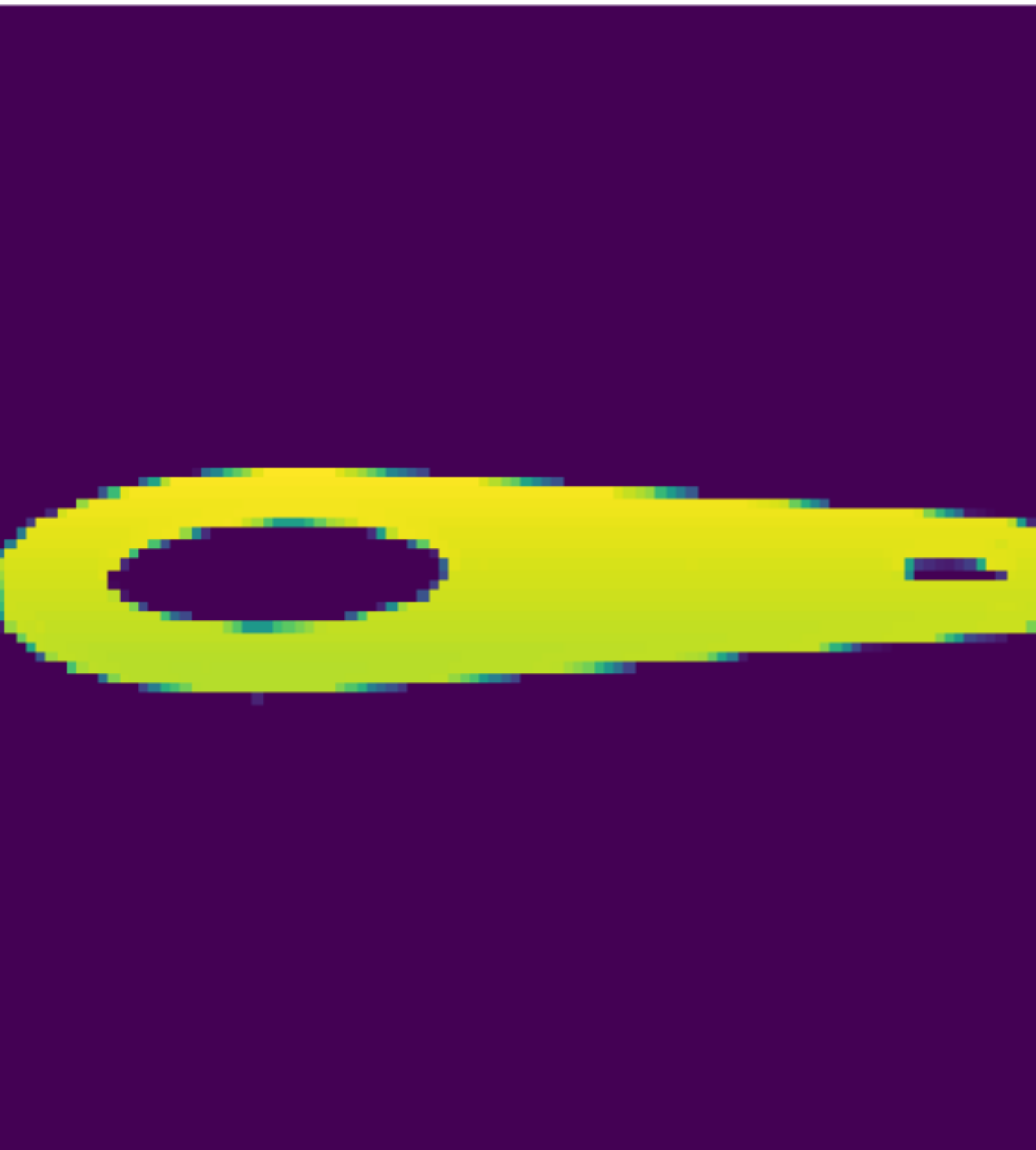} &
			\includegraphics[trim={9cm 4cm 9cm 4cm}, clip = true,width=0.12\linewidth]{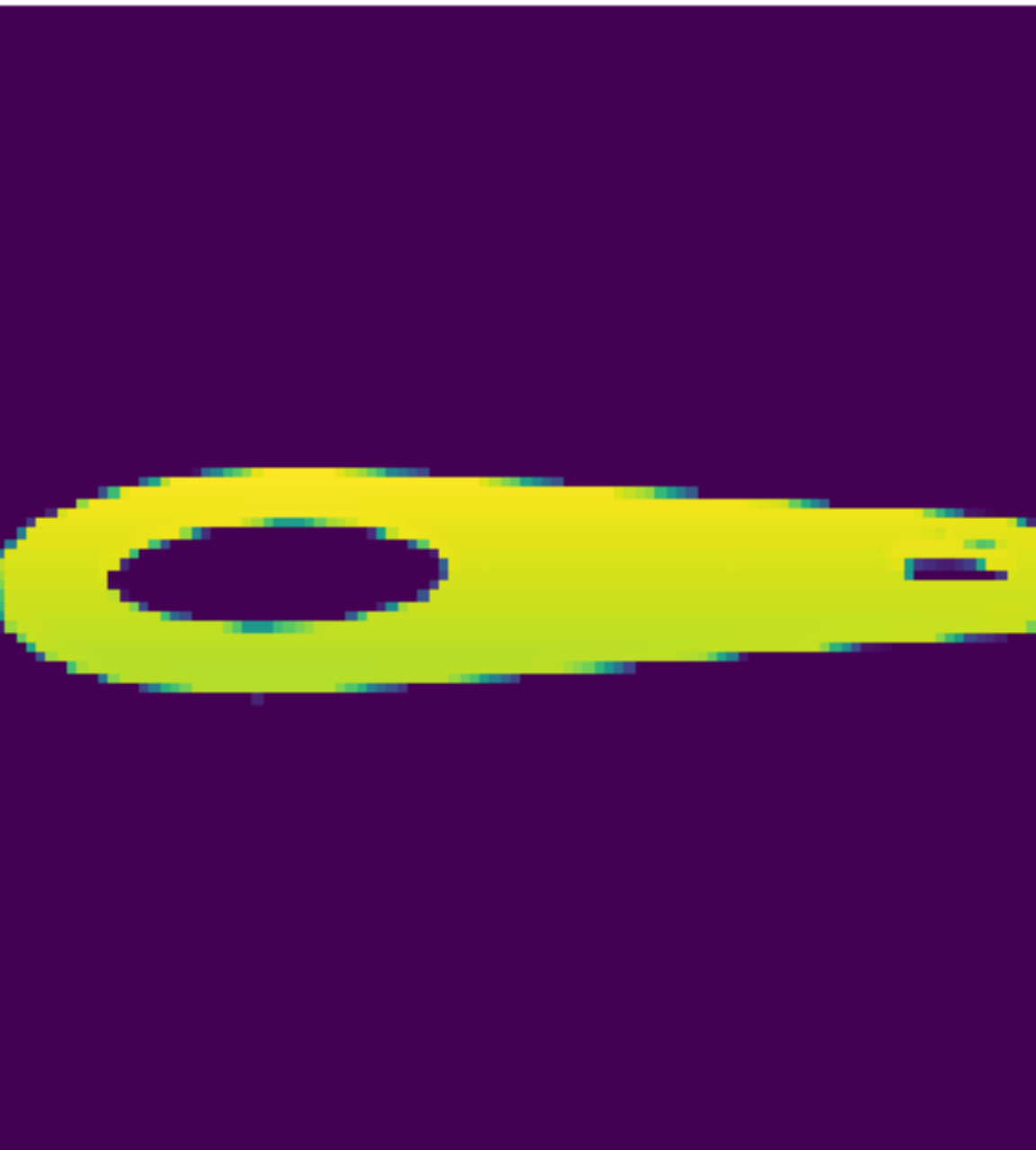} &
			\fcolorbox{green}{white}{\includegraphics[trim={9cm 4cm 9cm 4cm}, clip = true,width=0.12\linewidth]{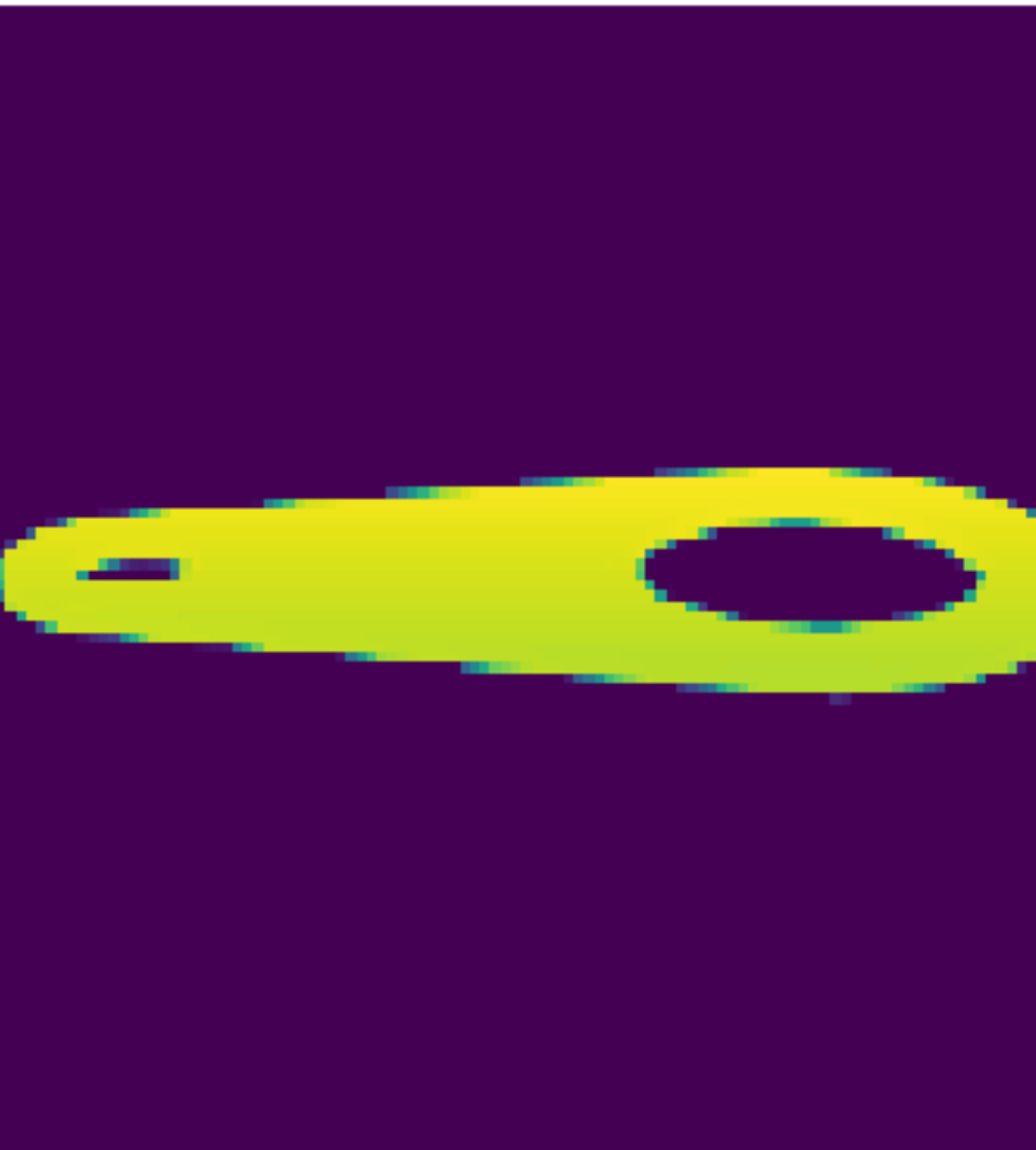}} &
			\fcolorbox{green}{white}{\includegraphics[trim={9cm 4cm 9cm 4cm}, clip = true,width=0.12\linewidth]{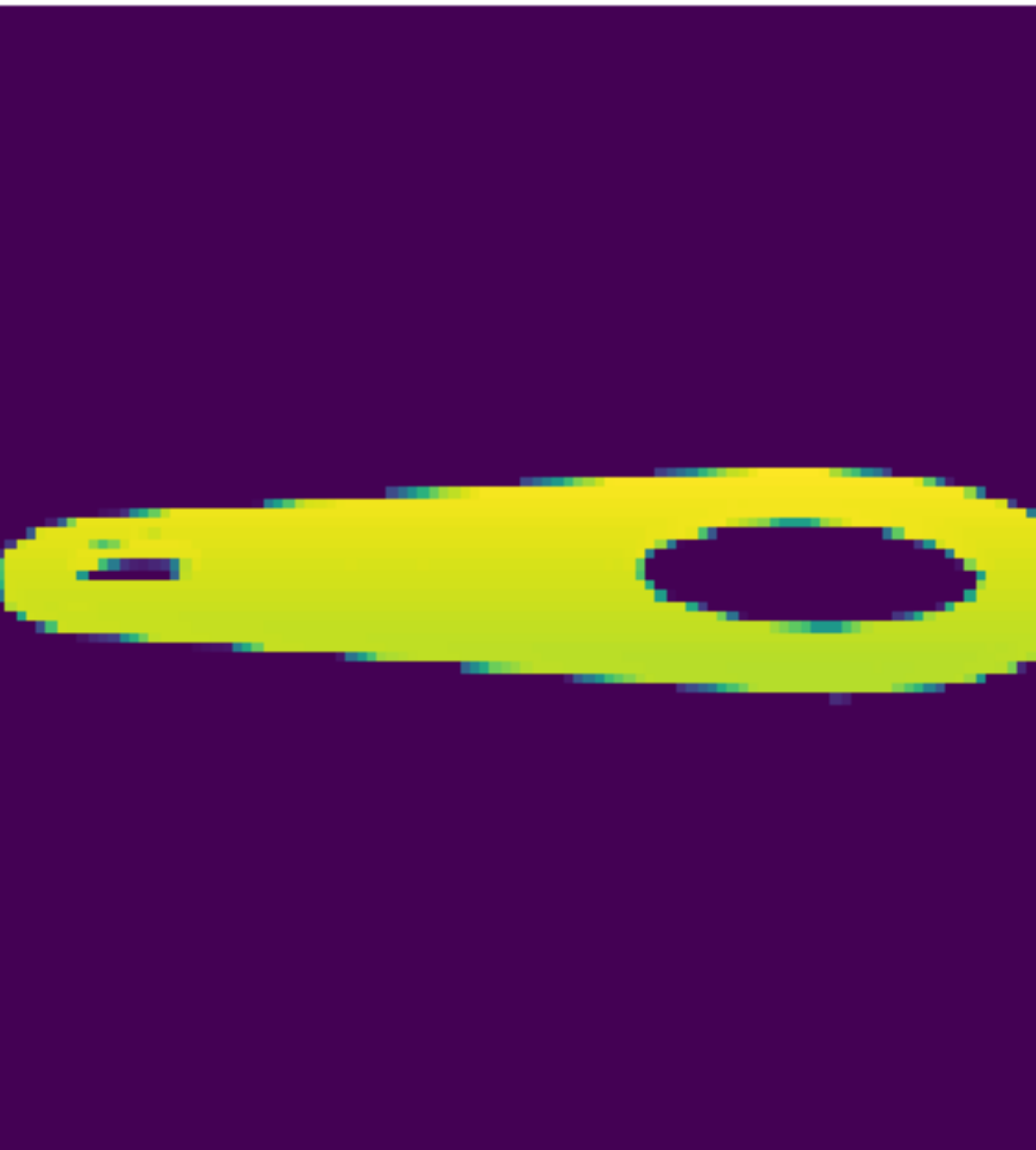}} & \raisebox{2\height}{\LARGE 3.37$^{\circ}$ } \\ \hline
	\end{tabular} }
	\centering
	\caption{{\bf Successful Results}: Qualitative results for pose estimation on the \textbf{synthetic} dataset.} % The object in the scene is shown in the first column. The second column shows the nearest discretizated view. When the top four predicted views shown in the third column match the nearest view, they are indicated in a \textbf{green} box. Otherwise, an \textbf{orange} box shows the view that is nearest among the predicted. The last column shows the distance of this view. Given objects with symmetry, there can be more than one best view.}
	\label{fig:synth_result_1}
\end{figure*}

\begin{figure*} %\ContinuedFloat
	\hspace{-1.3cm}\parbox{\linewidth}{
		\begin{tabular}{|c|c|cccc|c|}
			\hline
			{\bf Input } & {\bf GT } & \multicolumn{4}{c|}{\bf Top-4 Predictions} & {\bf  $ d_\text{rot, best}^{sym} $} \\ \hline
			
			\includegraphics[trim={9cm 4cm 9cm 4cm}, clip = true,width=0.12\linewidth]{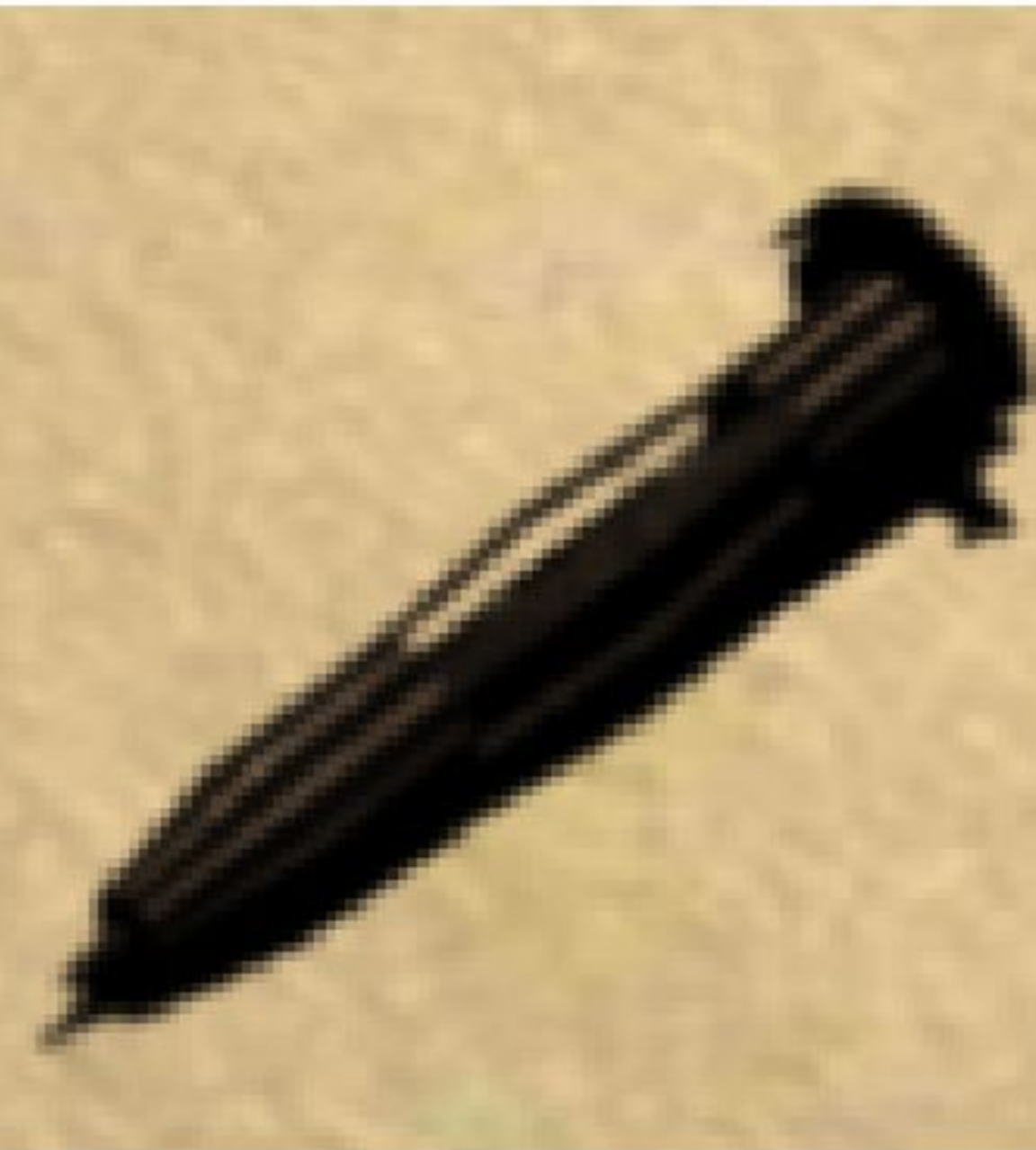}  &
			\includegraphics[trim={9cm 4cm 9cm 4cm}, clip = true,width=0.12\linewidth]{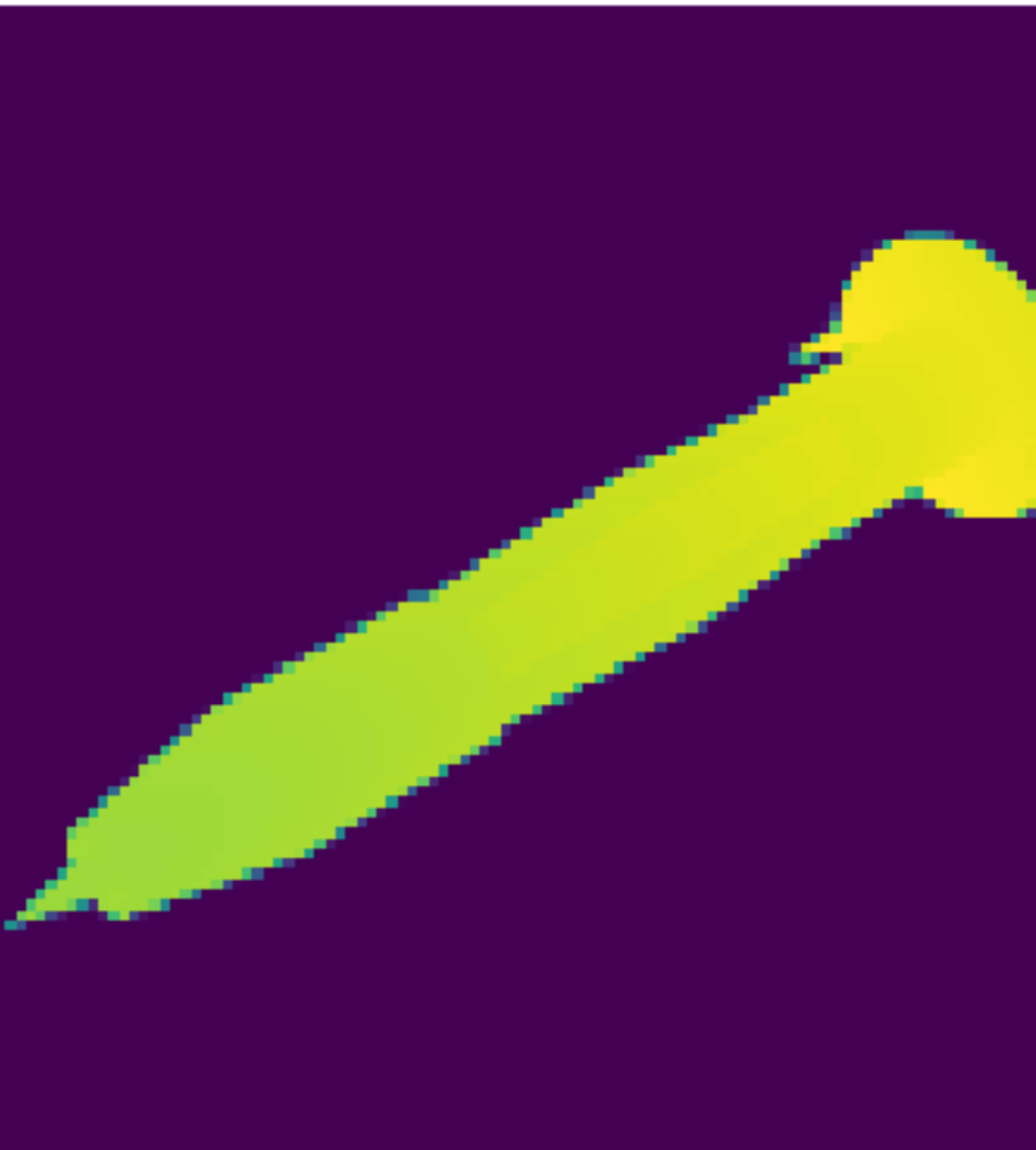}  &
			\includegraphics[trim={9cm 4cm 9cm 4cm}, clip = true,width=0.12\linewidth]{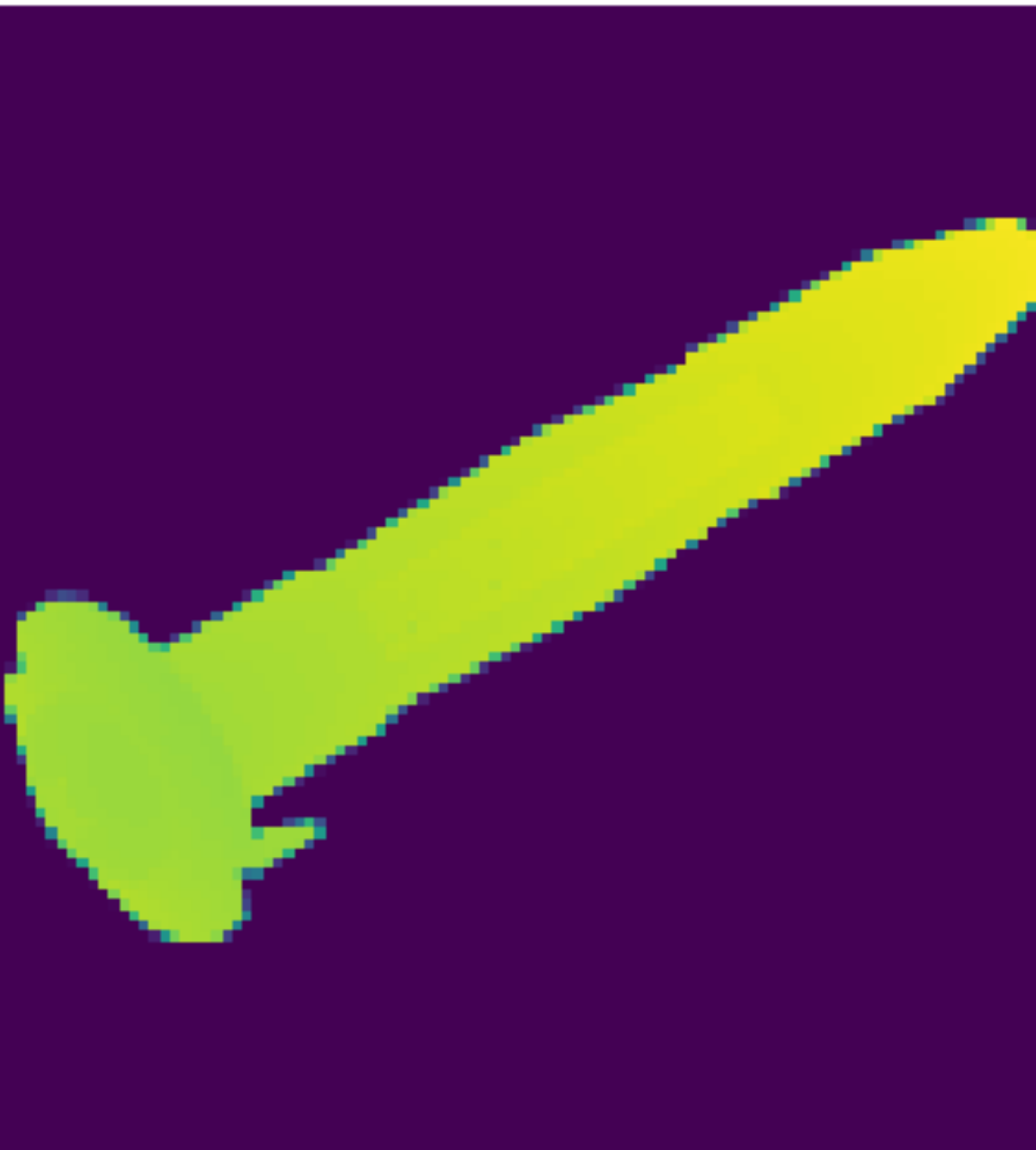}  &
			\includegraphics[trim={9cm 4cm 9cm 4cm}, clip = true,width=0.12\linewidth]{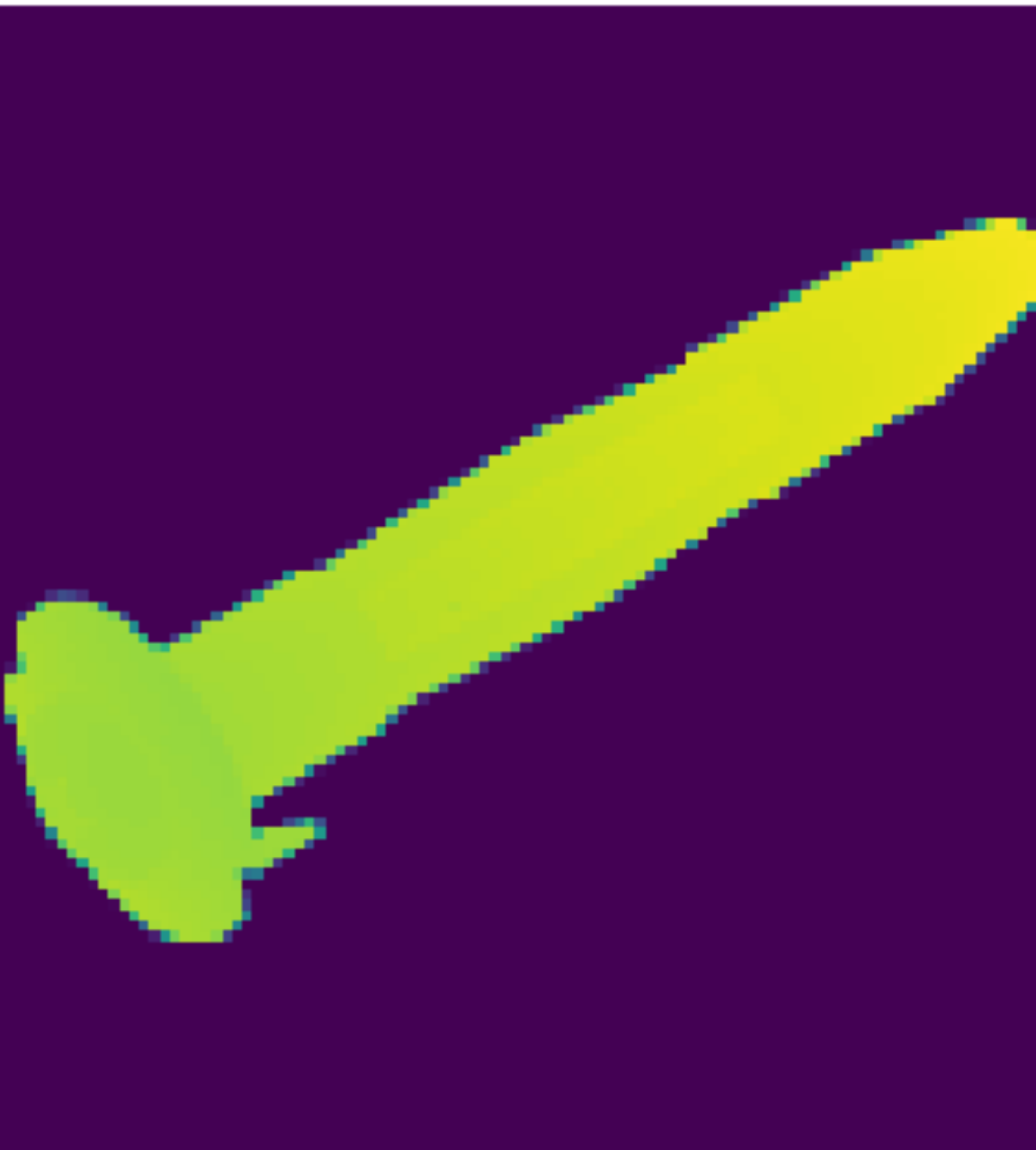} &
			\fcolorbox{green}{white}{\includegraphics[trim={9cm 4cm 9cm 4cm}, clip = true,width=0.12\linewidth]{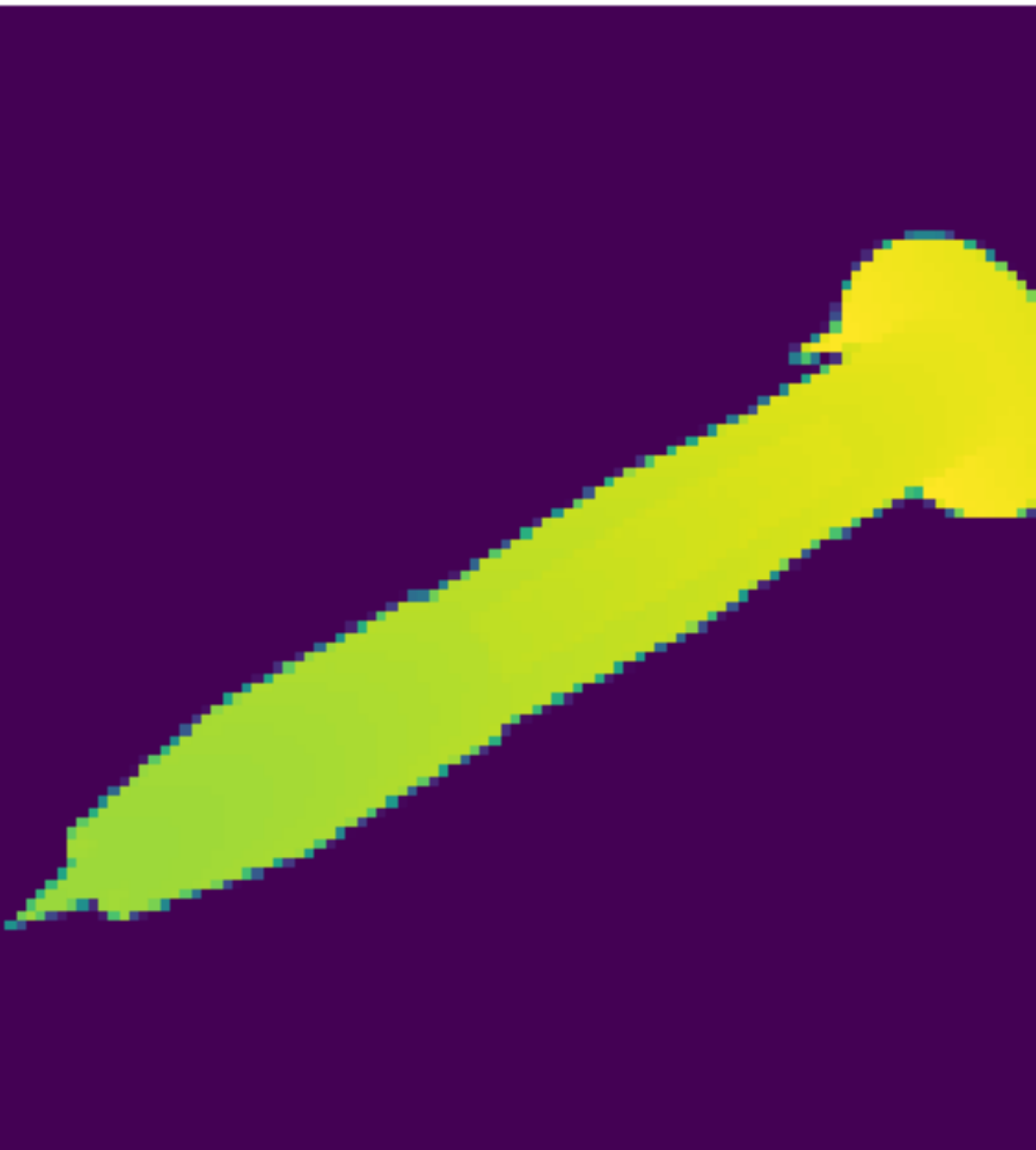}} &
			\fcolorbox{green}{white}{\includegraphics[trim={9cm 4cm 9cm 4cm}, clip = true,width=0.12\linewidth]{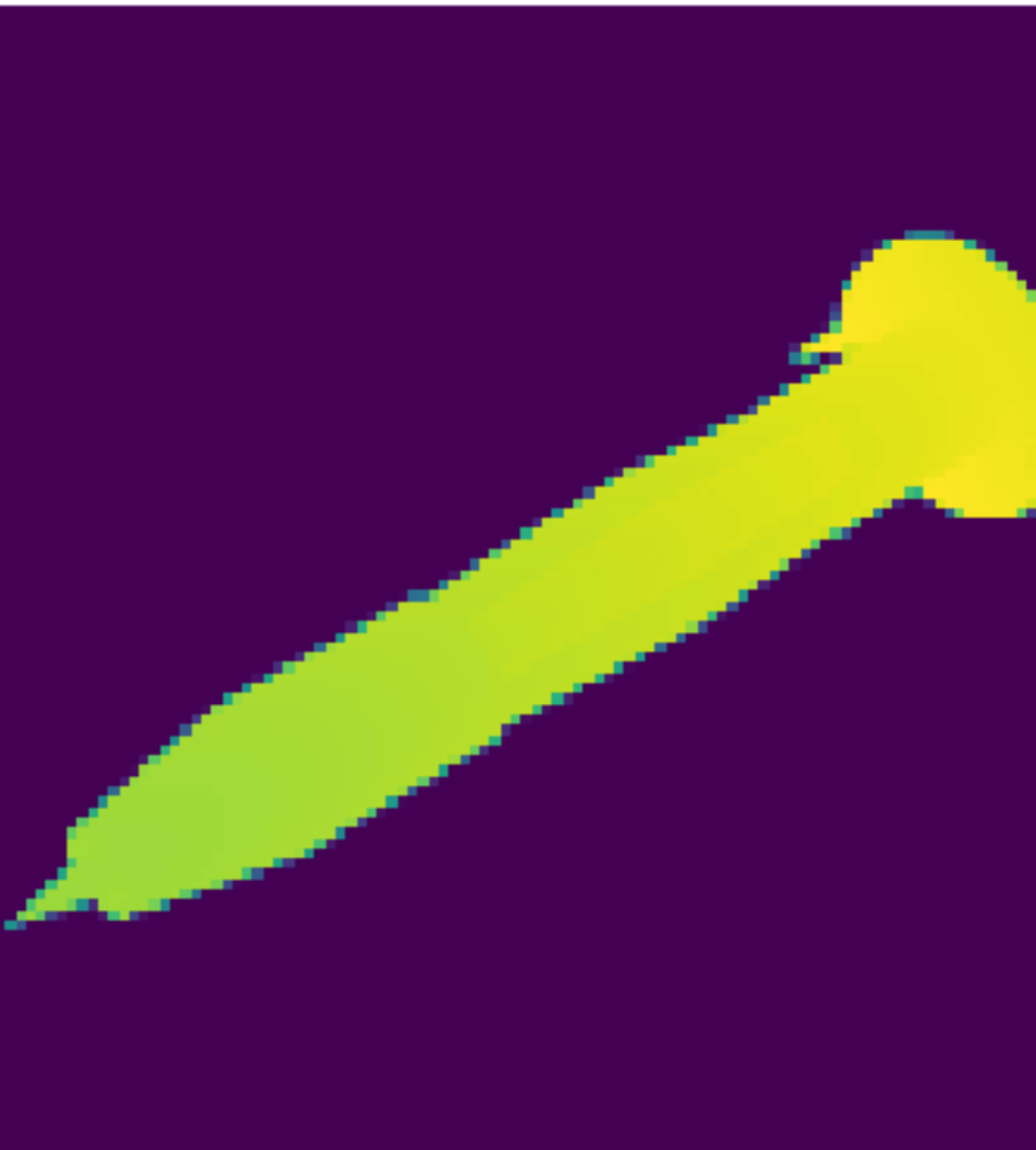} } & \raisebox{2\height}{\LARGE 4.19$^{\circ}$} \\	 \hline	
			\includegraphics[trim={9cm 4cm 9cm 4cm}, clip = true,width=0.12\linewidth]{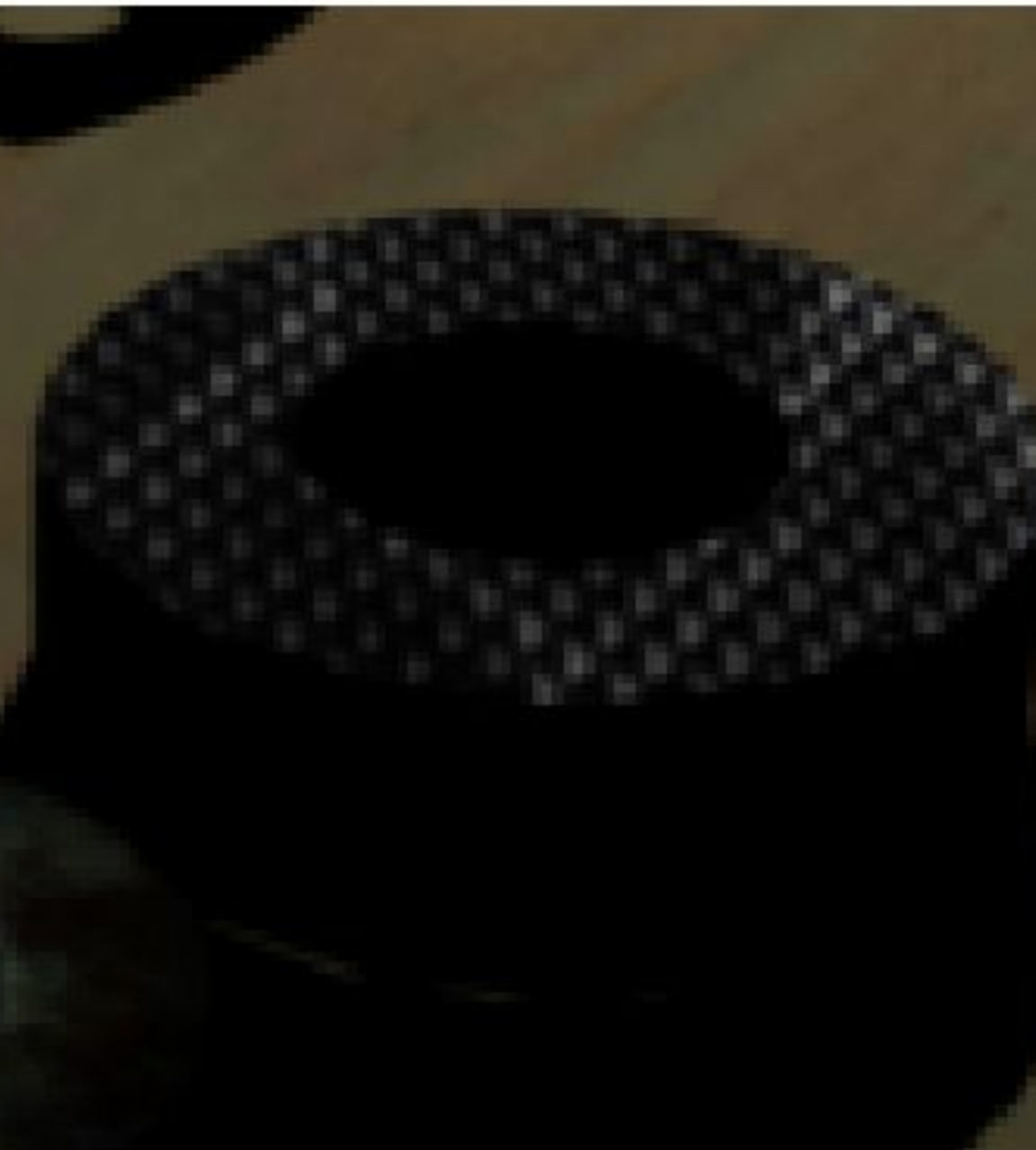} &
			\includegraphics[trim={9cm 4cm 9cm 4cm}, clip = true,width=0.12\linewidth]{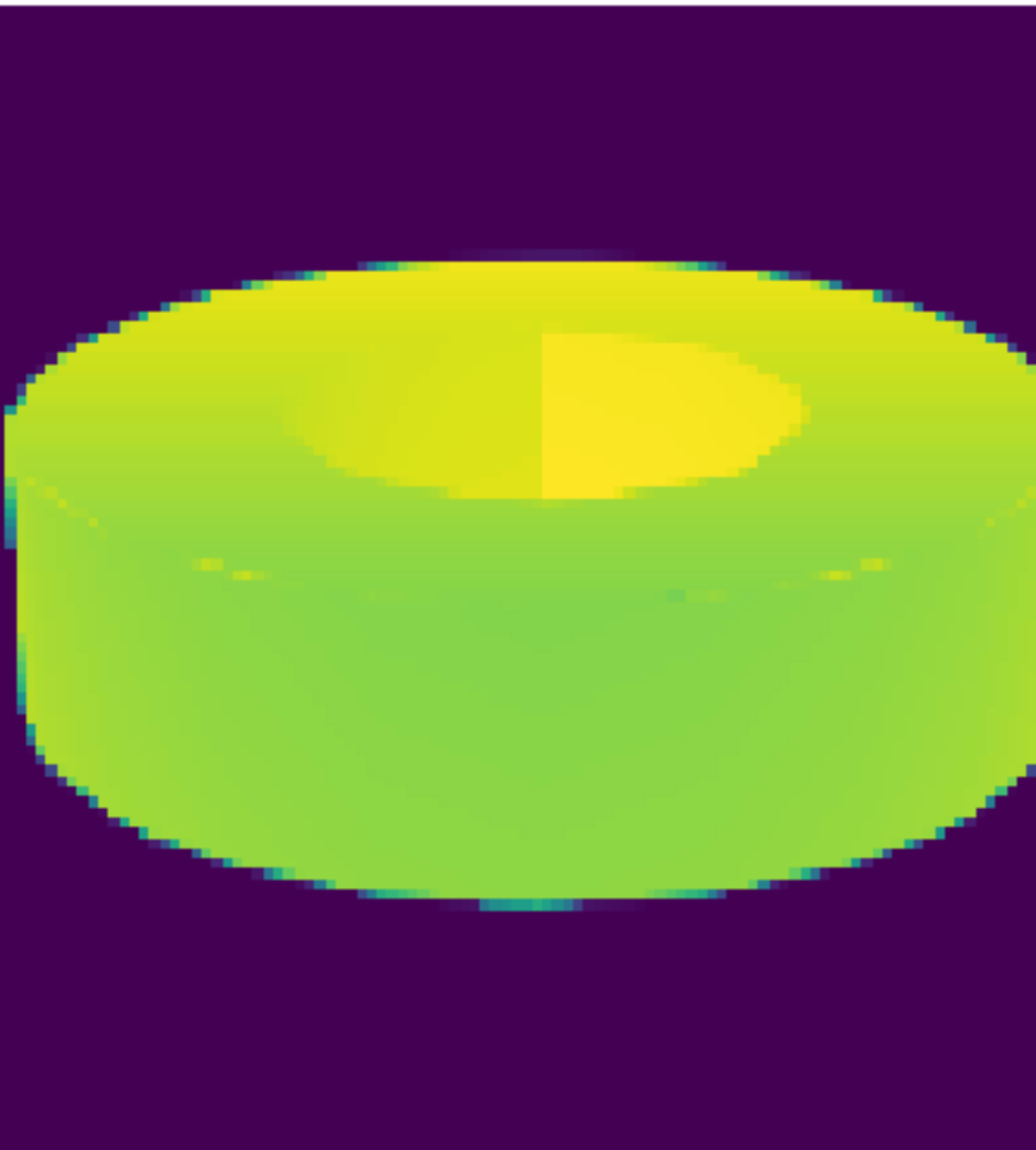} &
			\fcolorbox{green}{white}{\includegraphics[trim={9cm 4cm 9cm 4cm}, clip = true,width=0.12\linewidth]{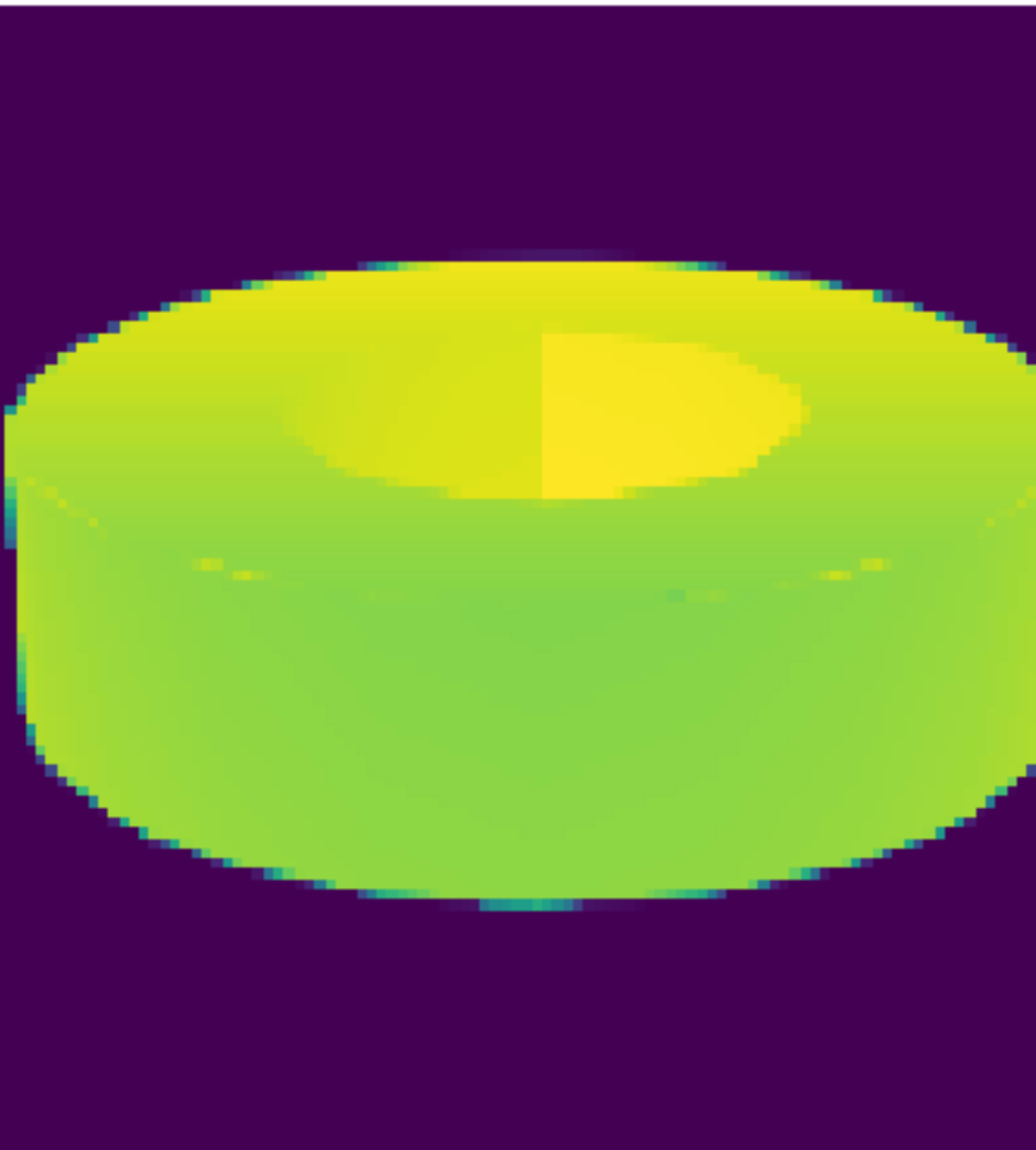}} &
			\fcolorbox{green}{white}{\includegraphics[trim={9cm 4cm 9cm 4cm}, clip = true,width=0.12\linewidth]{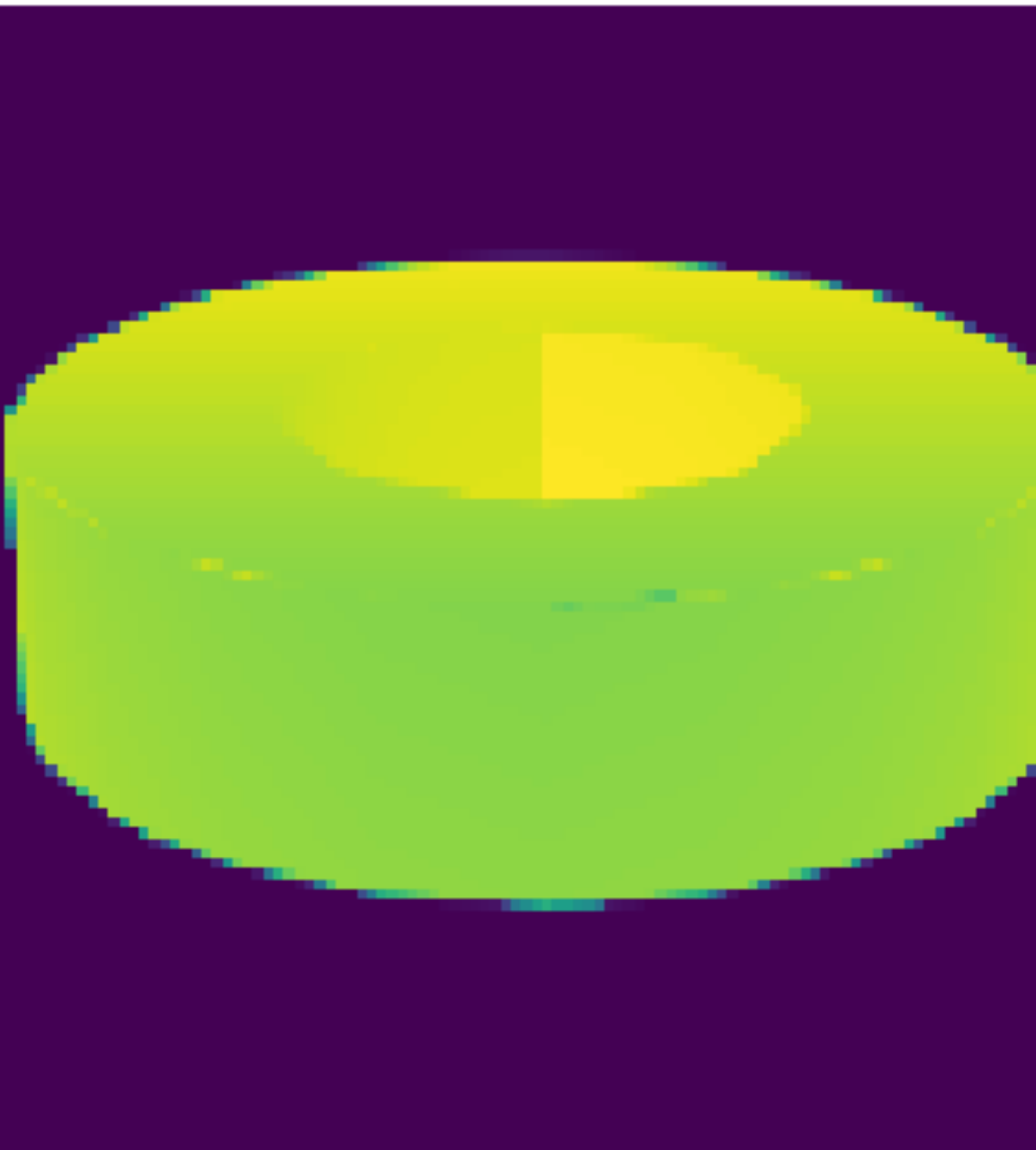}} &
			\fcolorbox{green}{white}{\includegraphics[trim={9cm 4cm 9cm 4cm}, clip = true,width=0.12\linewidth]{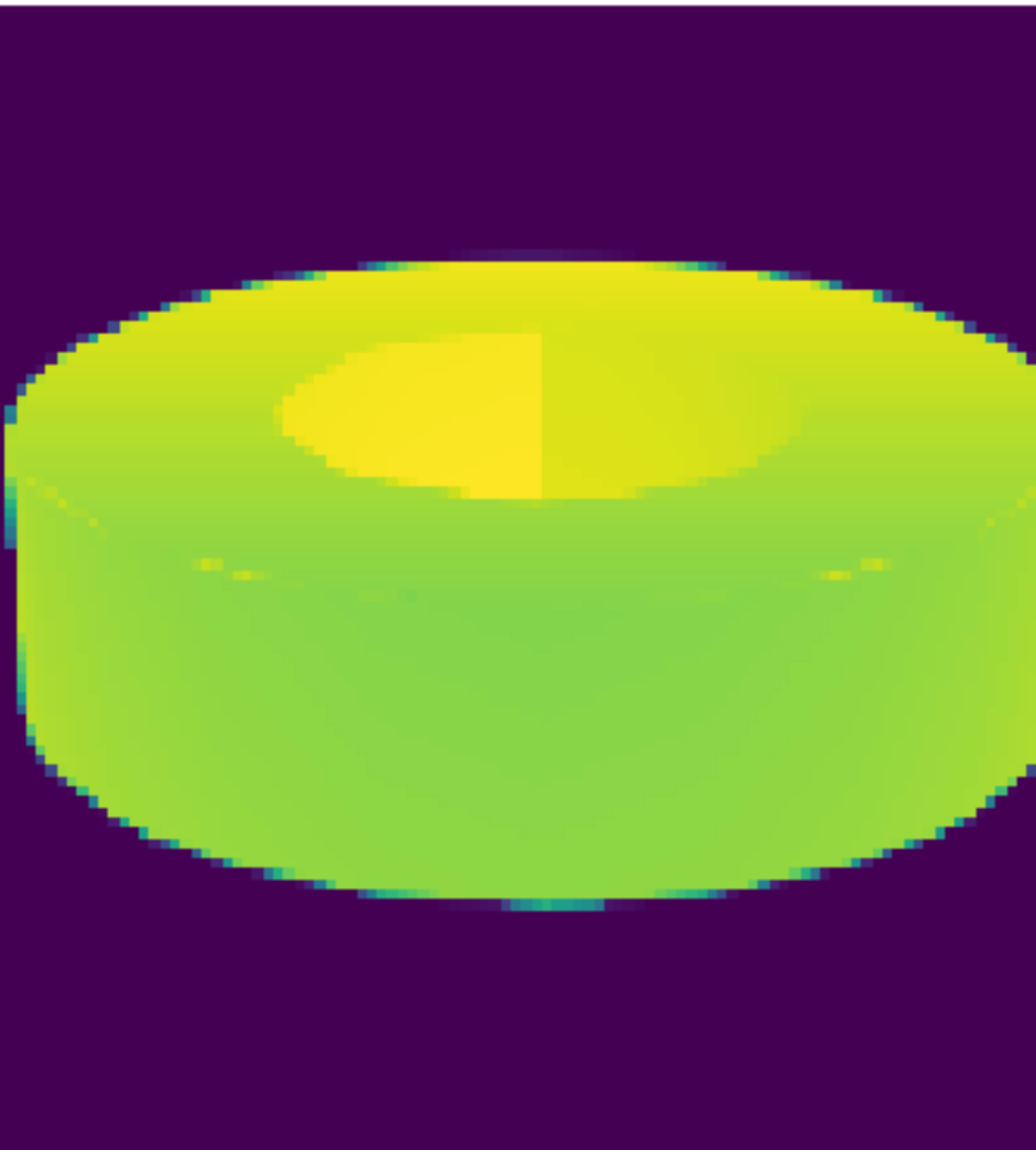}} &
			\fcolorbox{green}{white}{\includegraphics[trim={9cm 4cm 9cm 4cm}, clip = true,width=0.12\linewidth]{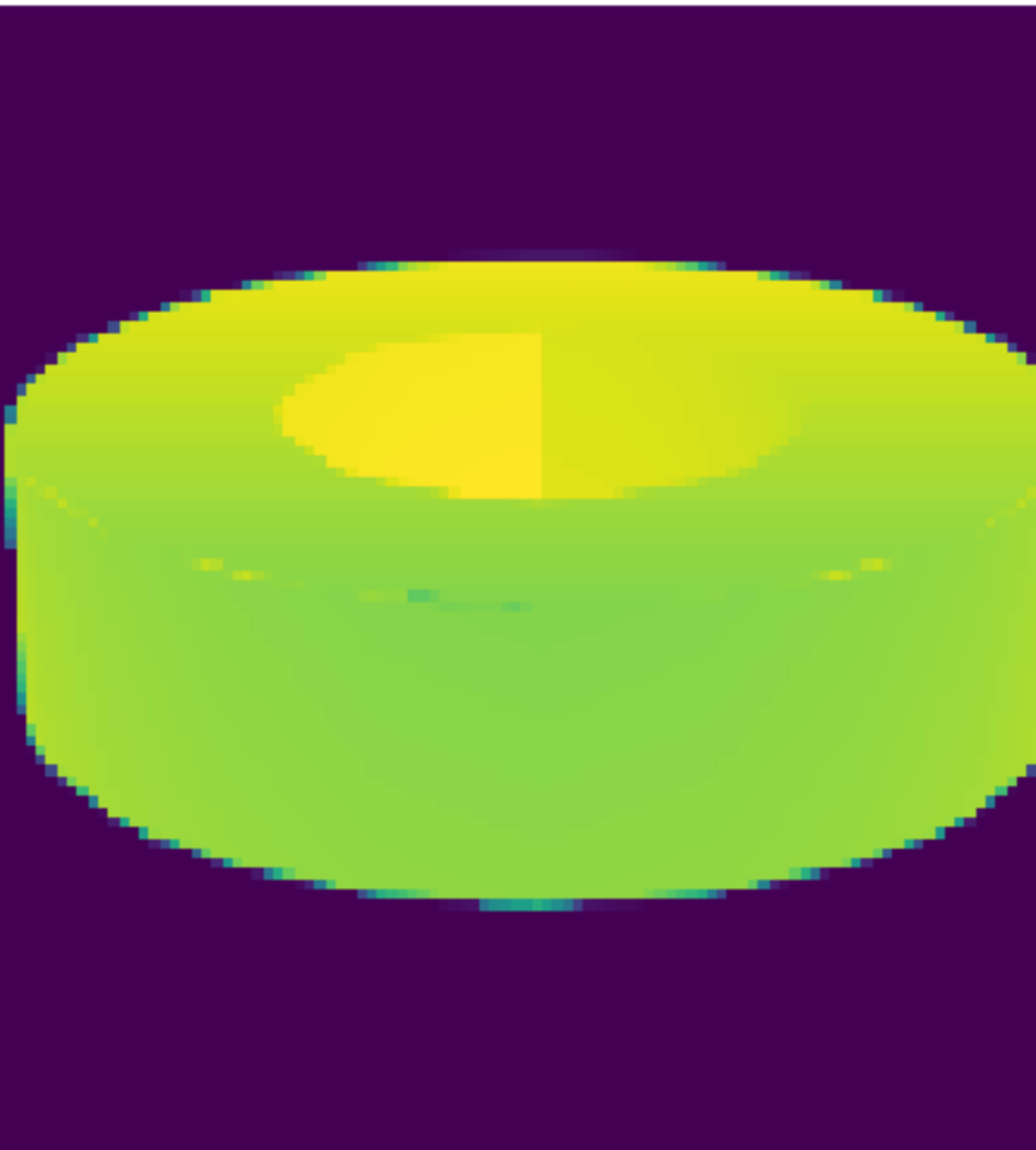}} & \raisebox{2\height}{\LARGE 4.23$^{\circ}$ } \\ \hline
			\includegraphics[trim={9cm 4cm 9cm 4cm}, clip = true,width=0.12\linewidth]{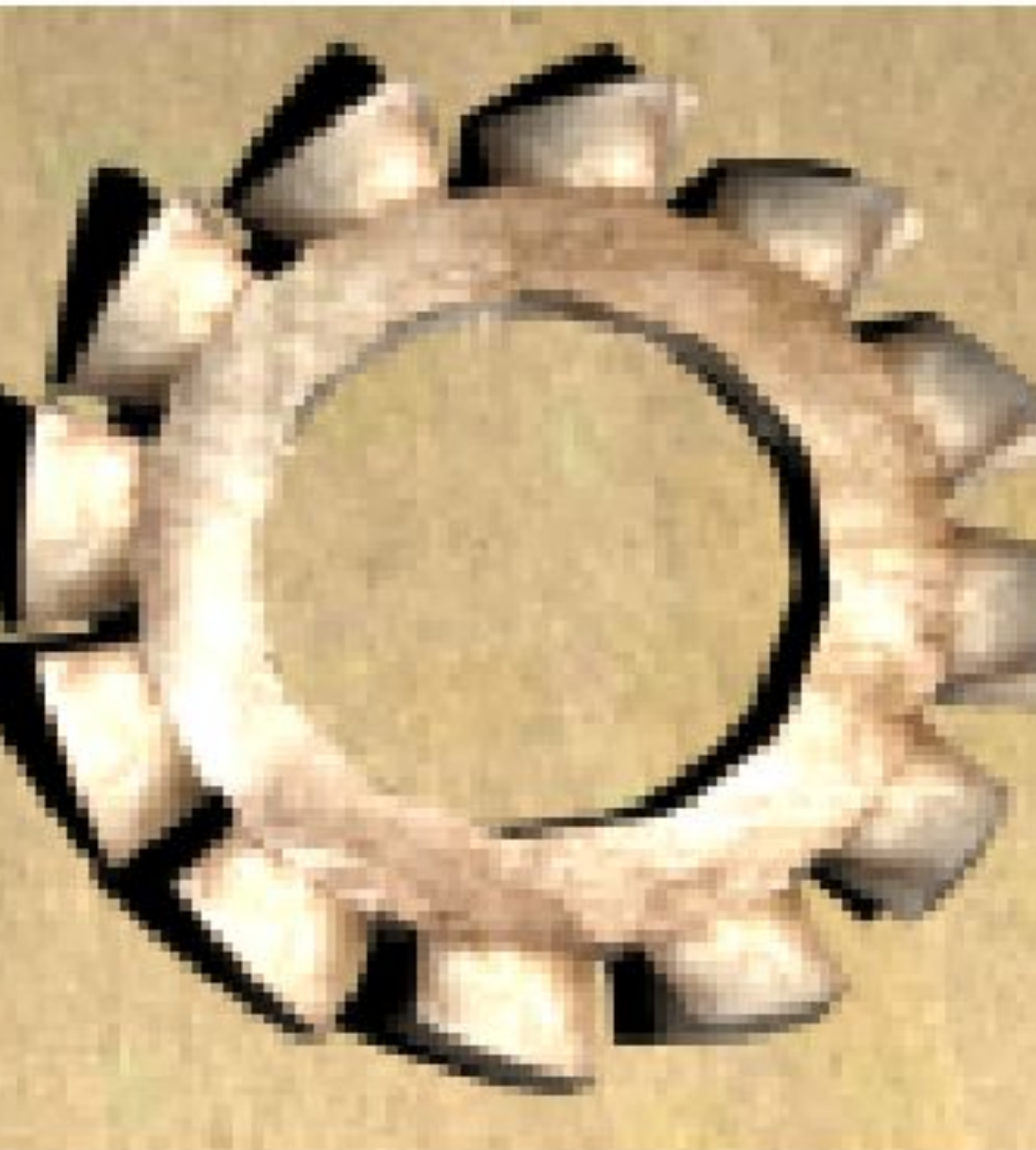} &
			\includegraphics[trim={9cm 4cm 9cm 4cm}, clip = true,width=0.12\linewidth]{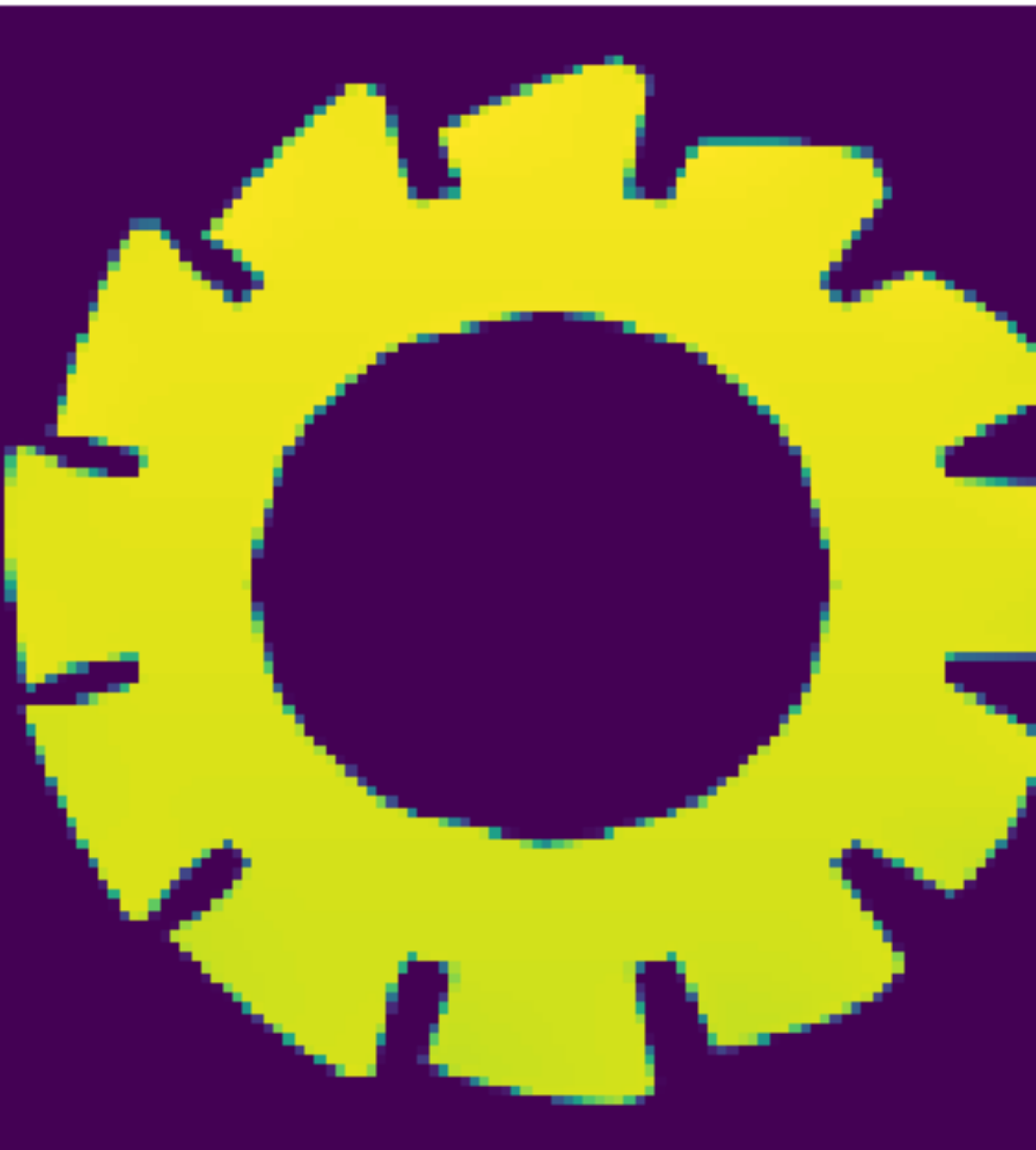} &
			\fcolorbox{green}{white}{\includegraphics[trim={9cm 4cm 9cm 4cm}, clip = true,width=0.12\linewidth]{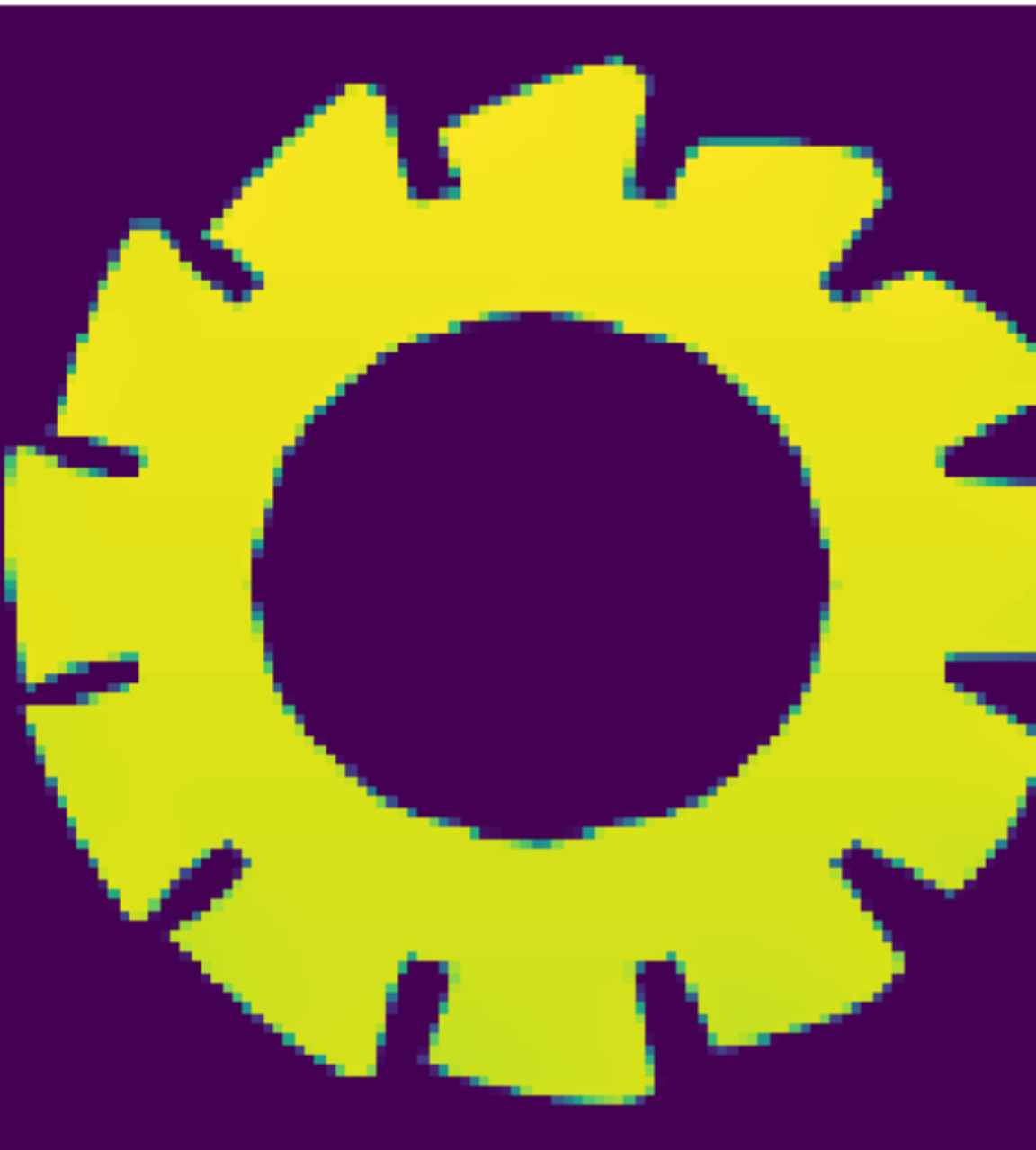}} &
			\fcolorbox{green}{white}{\includegraphics[trim={9cm 4cm 9cm 4cm}, clip = true,width=0.12\linewidth]{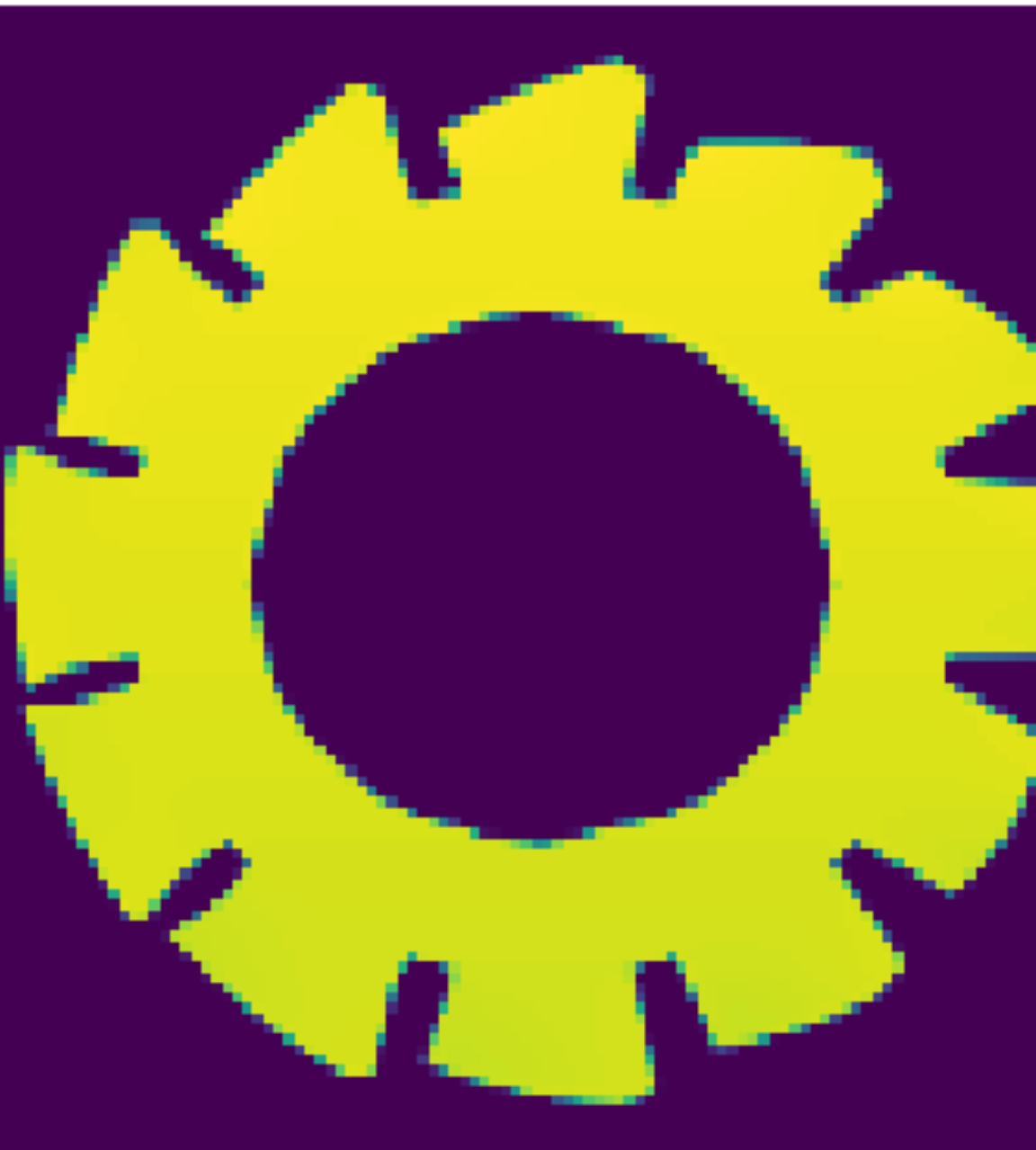}} &
			\fcolorbox{green}{white}{\includegraphics[trim={9cm 4cm 9cm 4cm}, clip = true,width=0.12\linewidth]{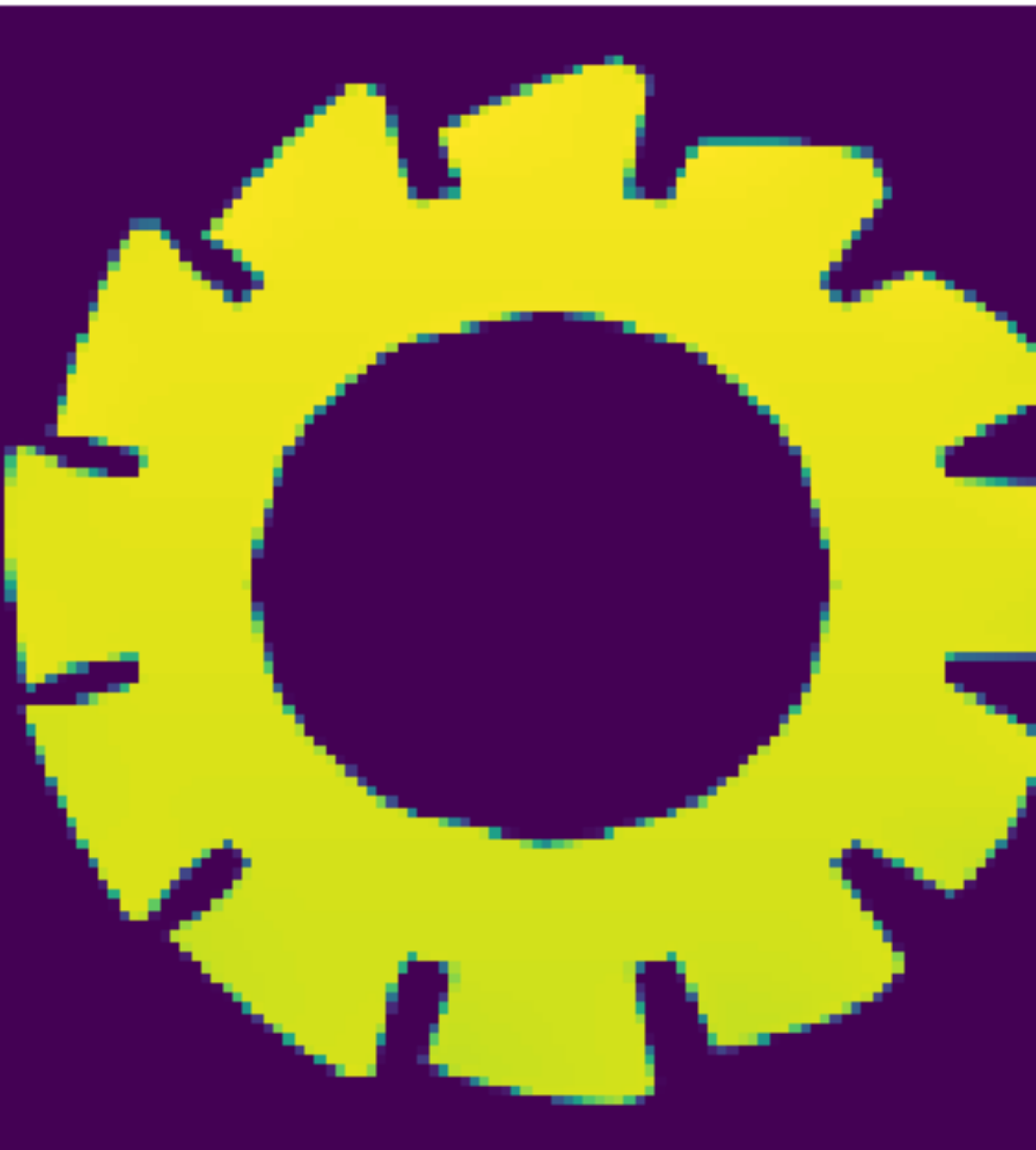}} &
			\fcolorbox{green}{white}{\includegraphics[trim={9cm 4cm 9cm 4cm}, clip = true,width=0.12\linewidth]{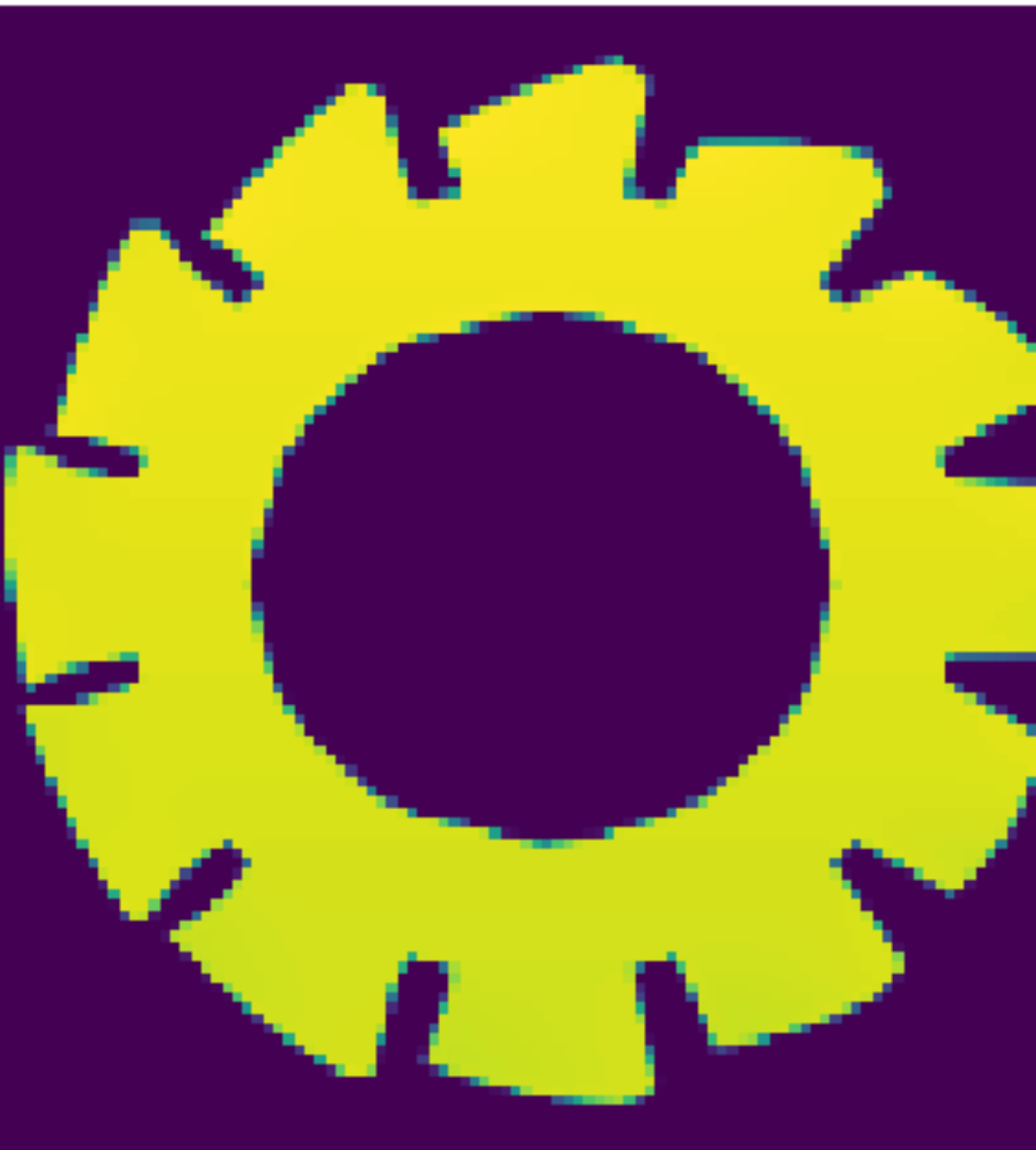}} & \raisebox{2\height}{\LARGE 4.26$^{\circ}$ } \\ \hline
			\includegraphics[trim={9cm 4cm 9cm 4cm}, clip = true,width=0.12\linewidth]{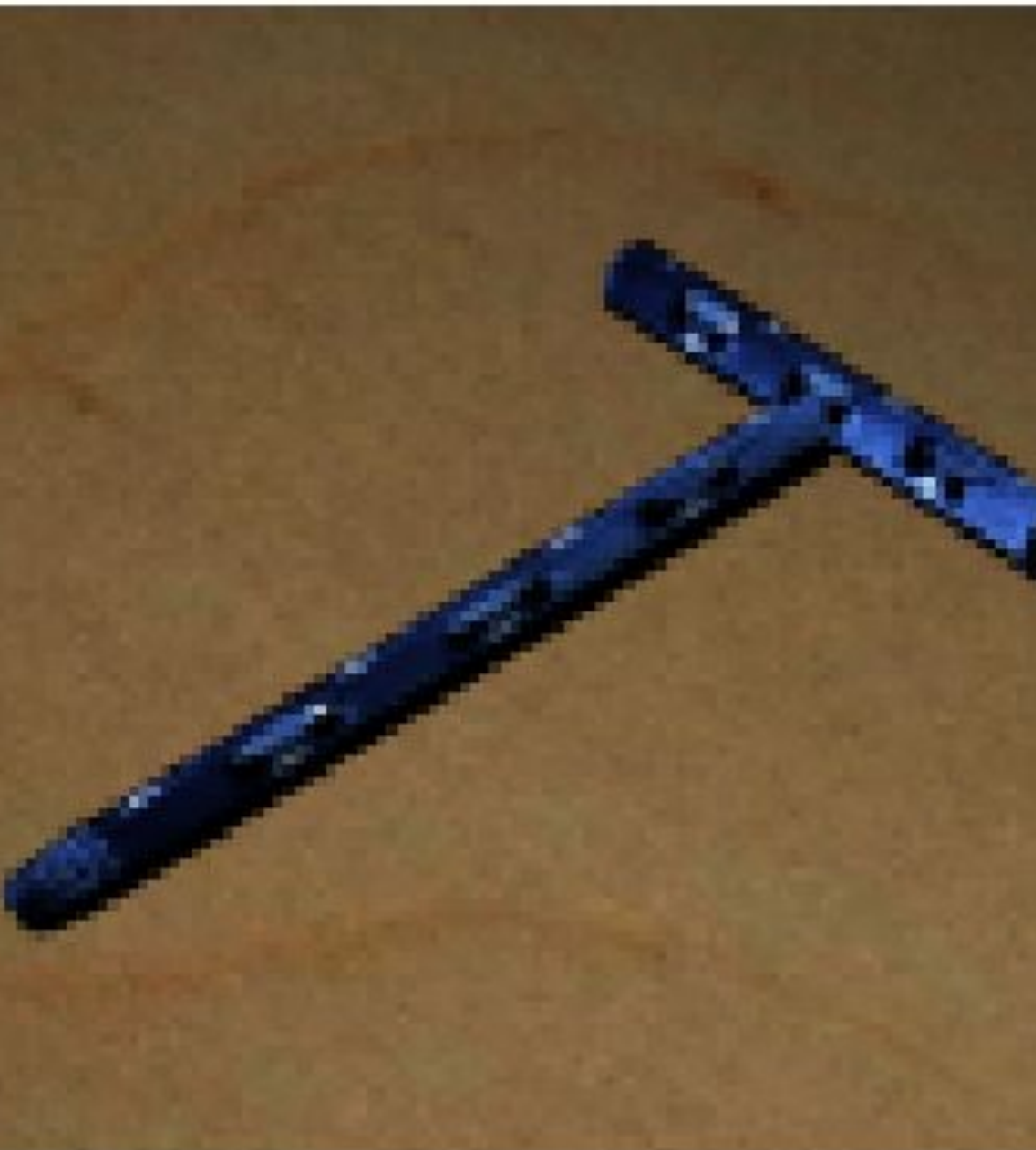}  &
			\includegraphics[trim={9cm 4cm 9cm 4cm}, clip = true,width=0.12\linewidth]{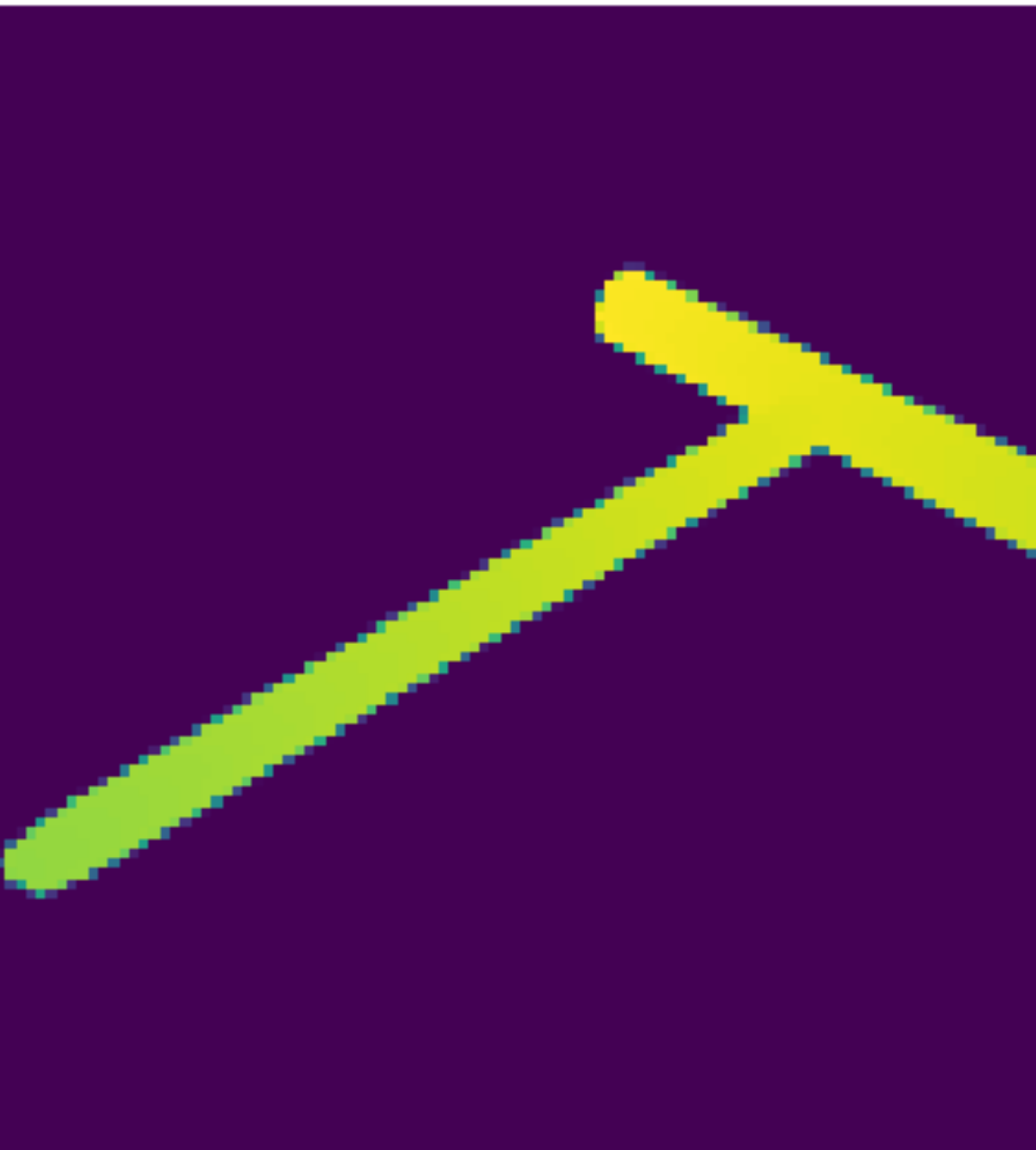}  &
			\fcolorbox{green}{white}{\includegraphics[trim={9cm 4cm 9cm 4cm}, clip = true,width=0.12\linewidth]{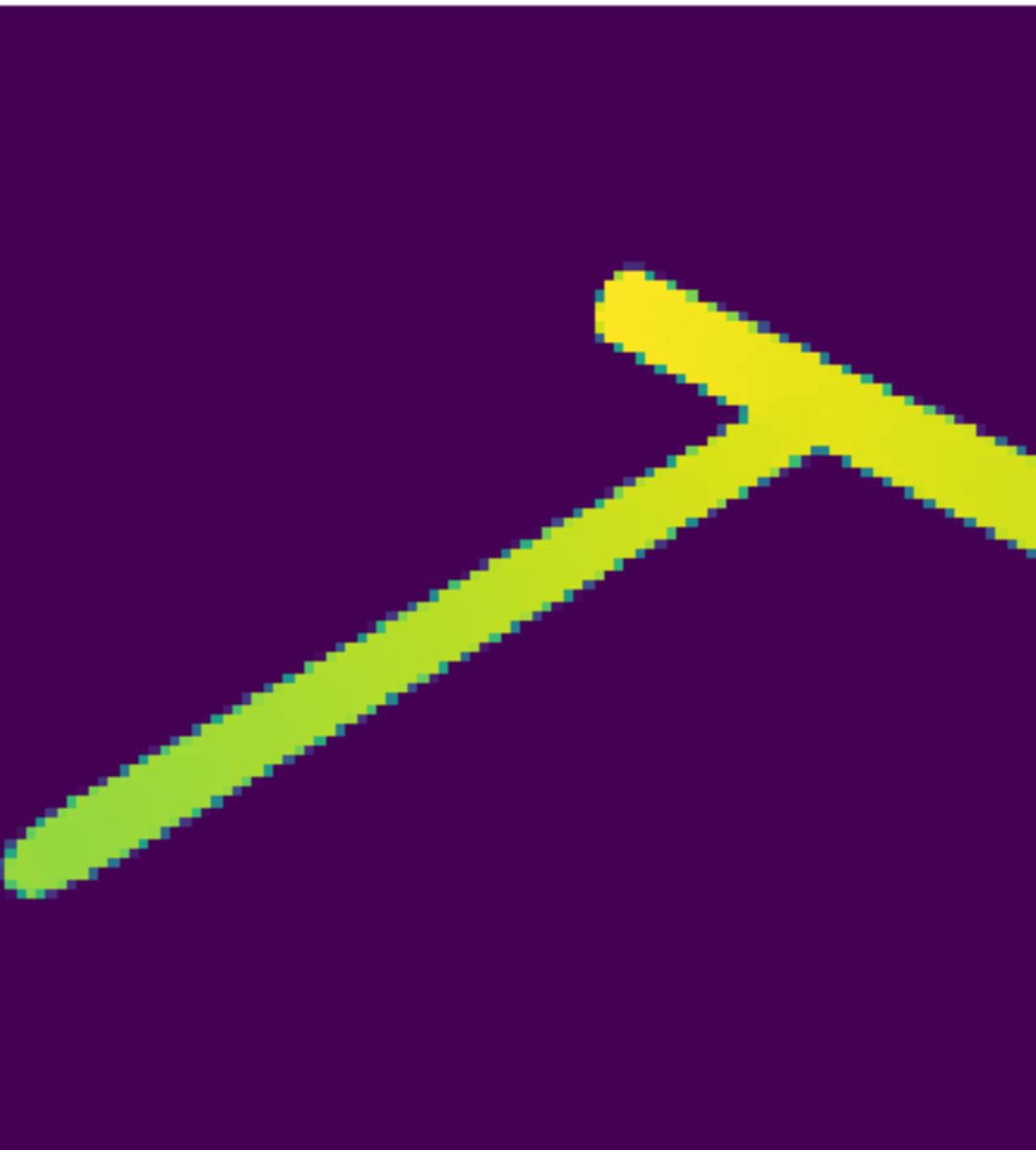}  }&
			\fcolorbox{green}{white}{\includegraphics[trim={9cm 4cm 9cm 4cm}, clip = true,width=0.12\linewidth]{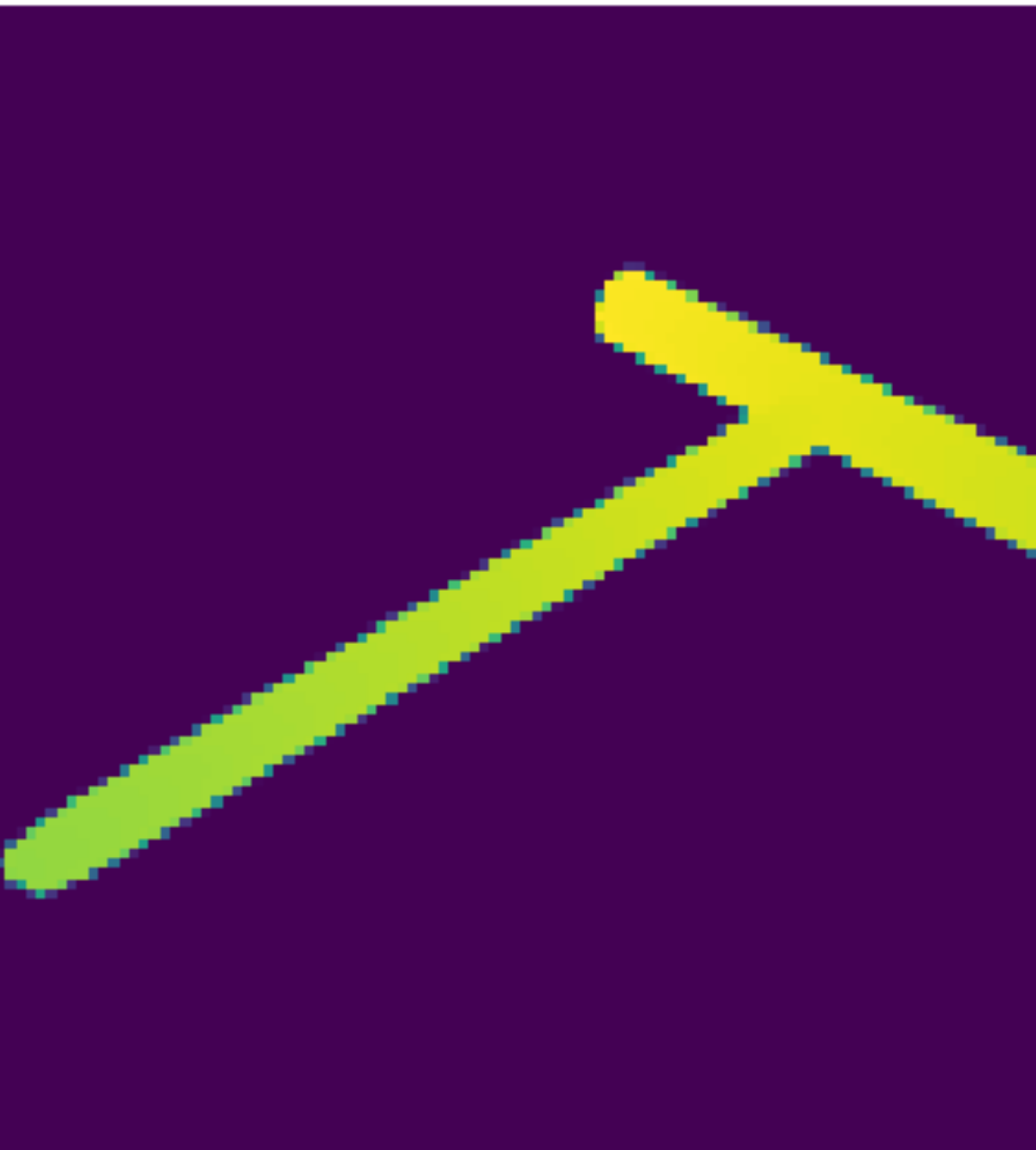}  }&
			\includegraphics[trim={9cm 4cm 9cm 4cm}, clip = true,width=0.12\linewidth]{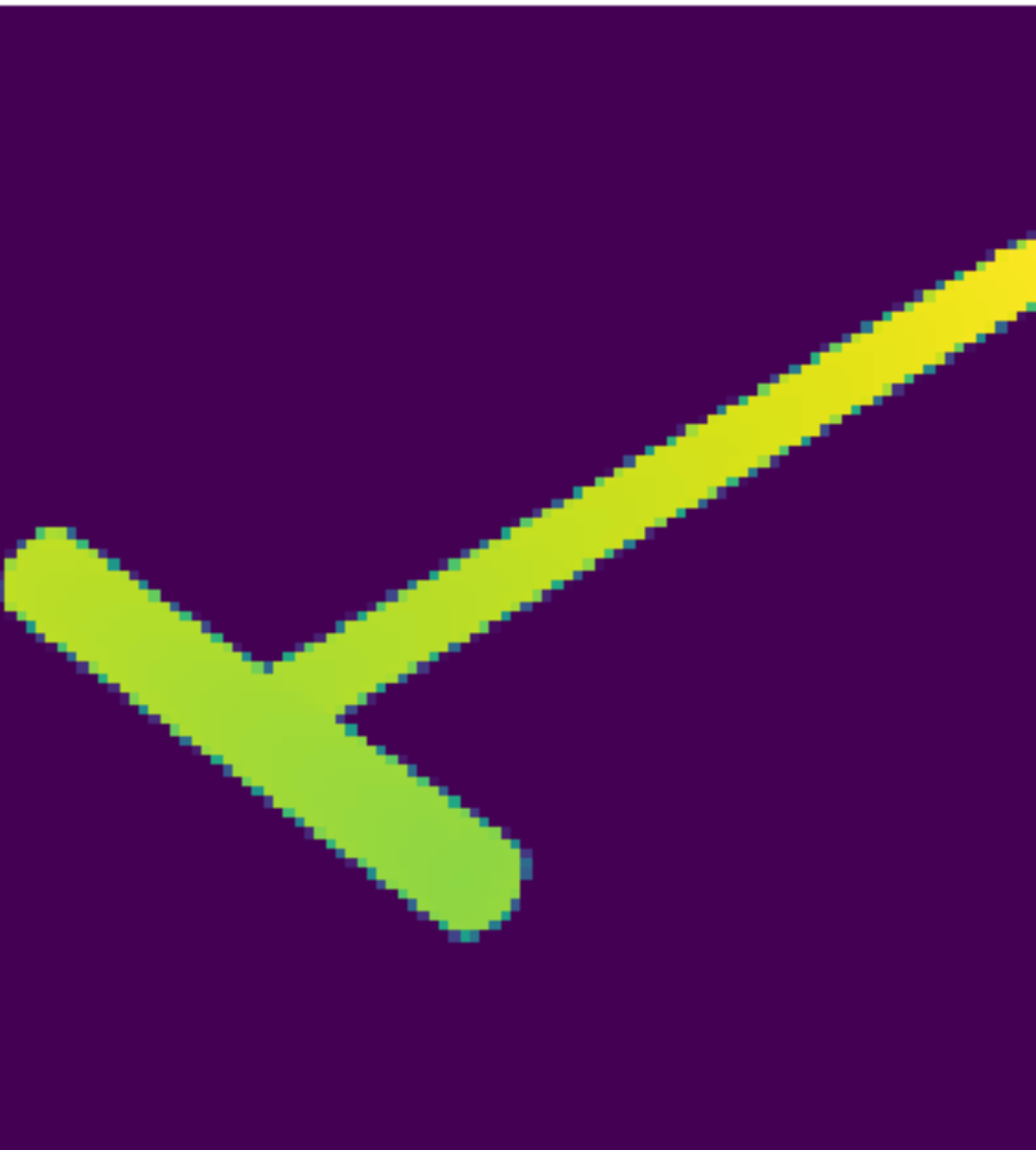}  &
			\includegraphics[trim={9cm 4cm 9cm 4cm}, clip = true,width=0.12\linewidth]{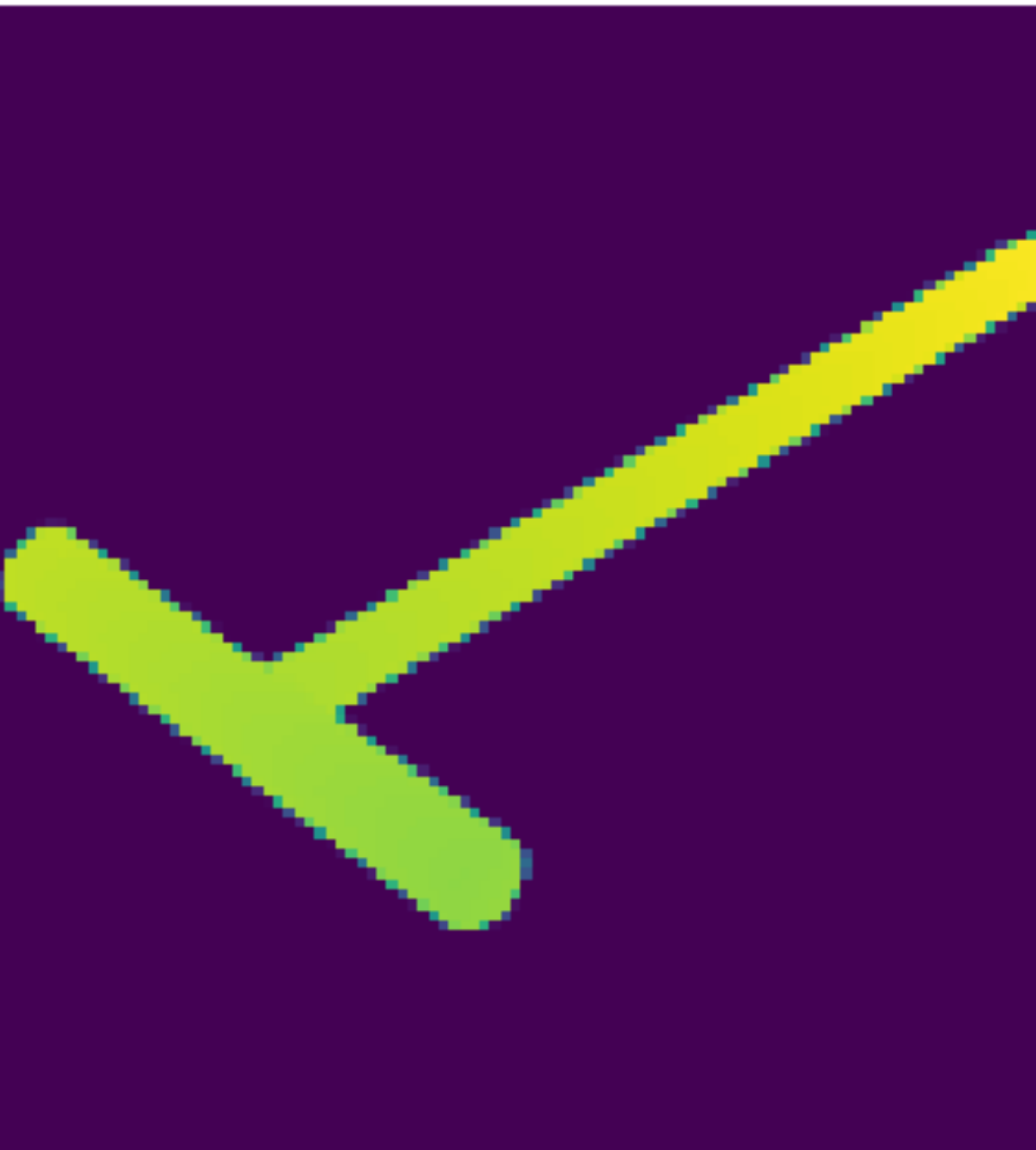}  & \raisebox{2\height}{\LARGE 4.42$^{\circ}$} \\ \hline
			\includegraphics[trim={9cm 4cm 9cm 4cm}, clip = true,width=0.12\linewidth]{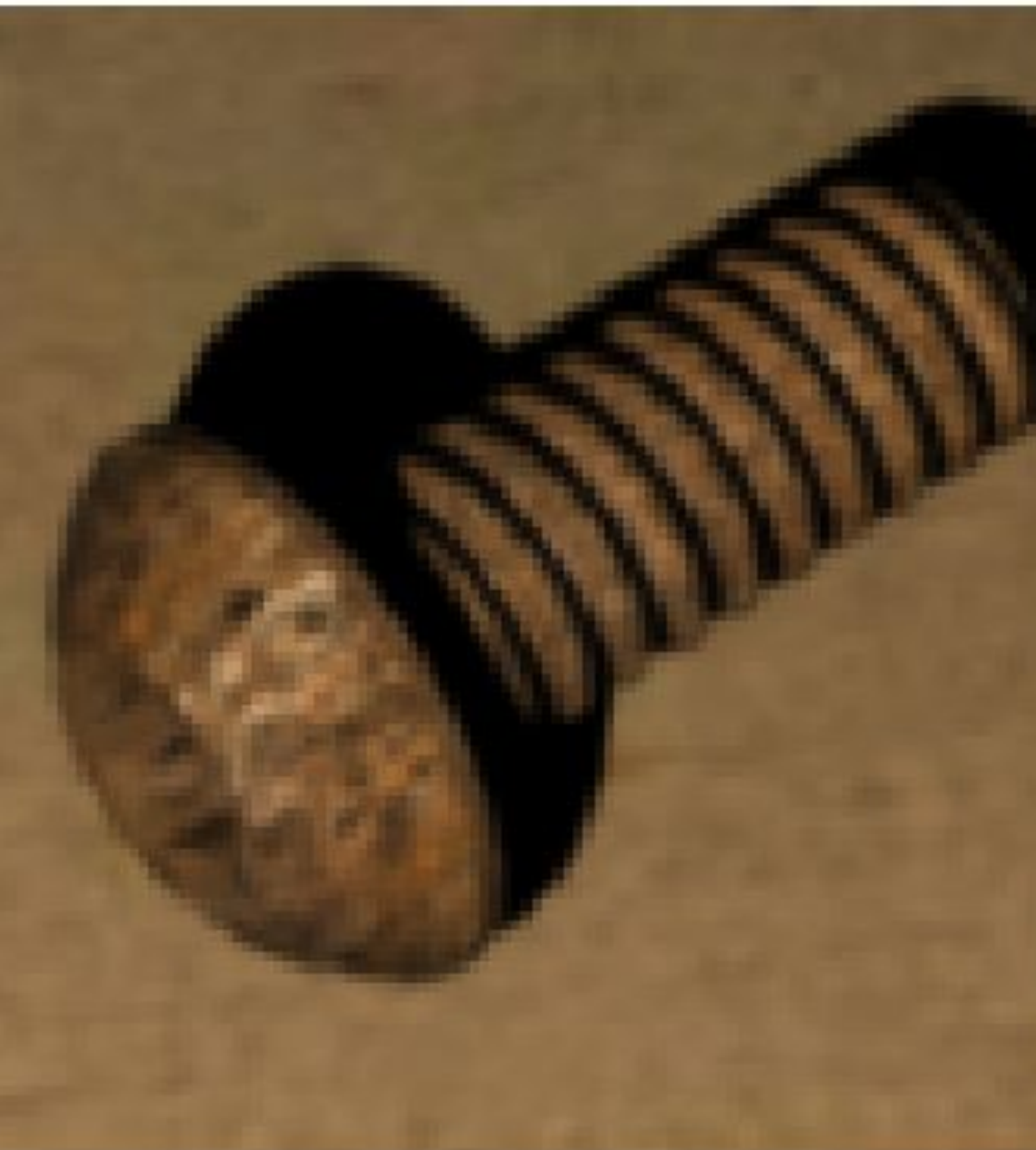}  &
			\includegraphics[trim={9cm 4cm 9cm 4cm}, clip = true,width=0.12\linewidth]{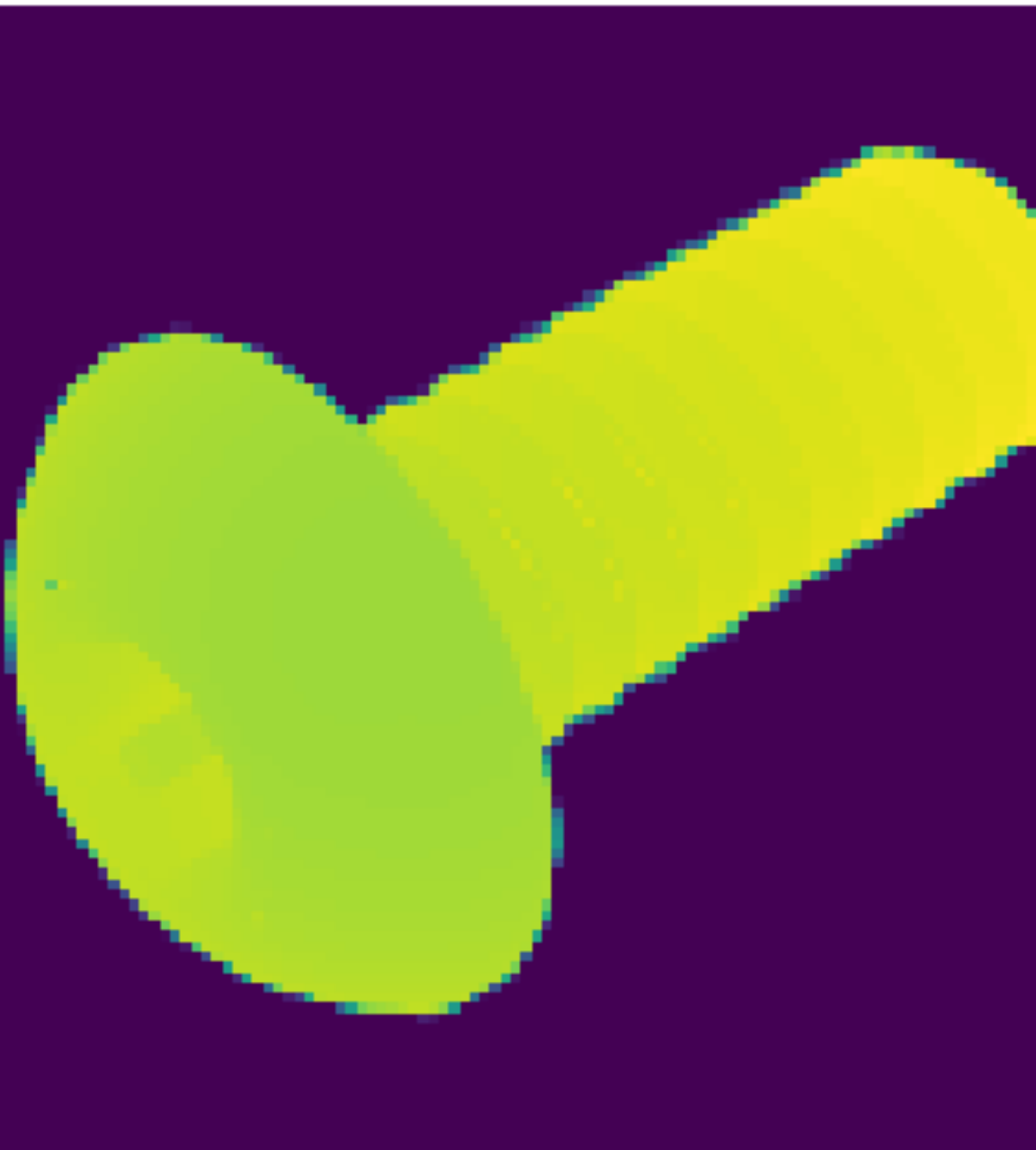}  &
			\includegraphics[trim={9cm 4cm 9cm 4cm}, clip = true,width=0.12\linewidth]{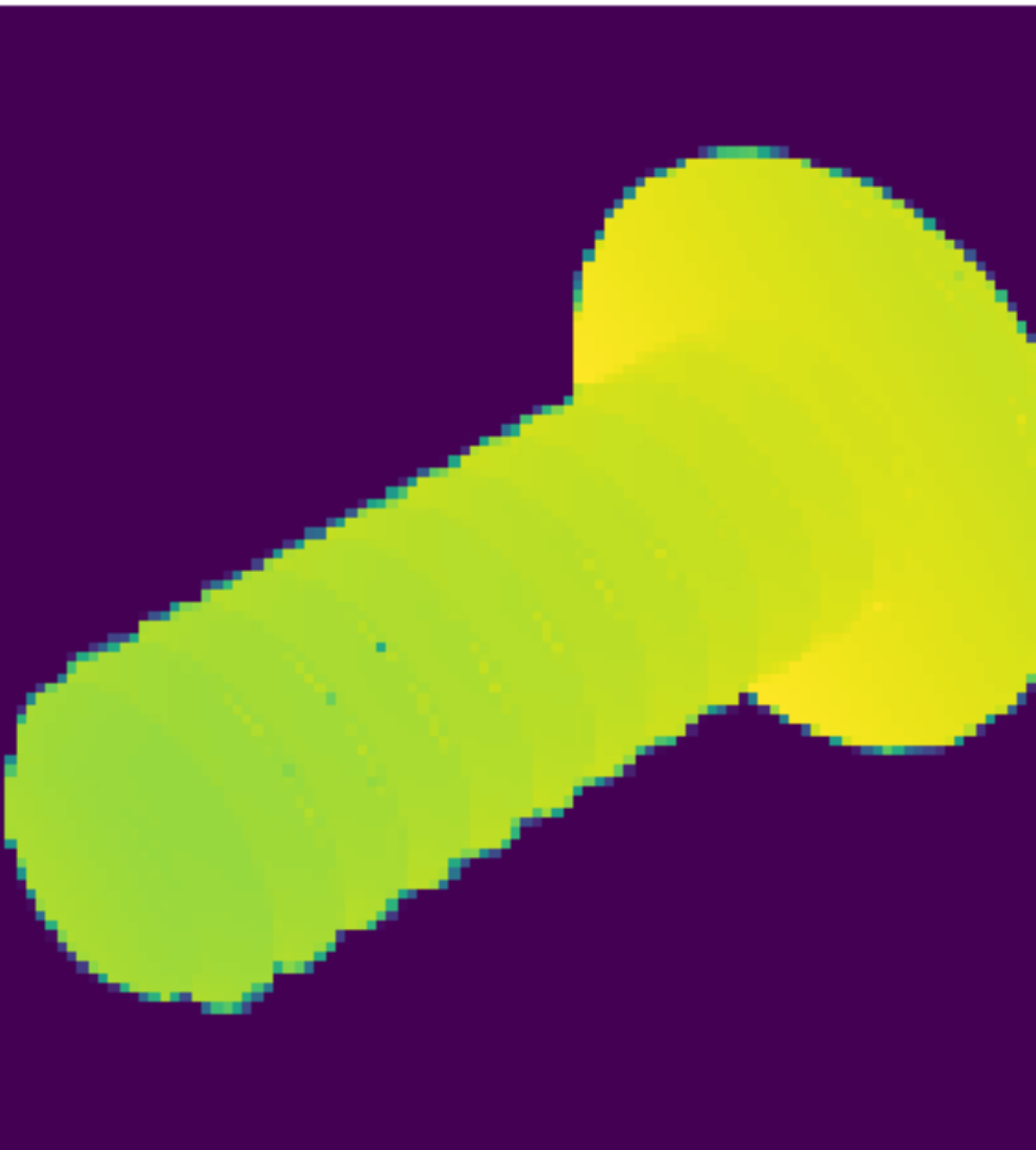}  &
			\includegraphics[trim={9cm 4cm 9cm 4cm}, clip = true,width=0.12\linewidth]{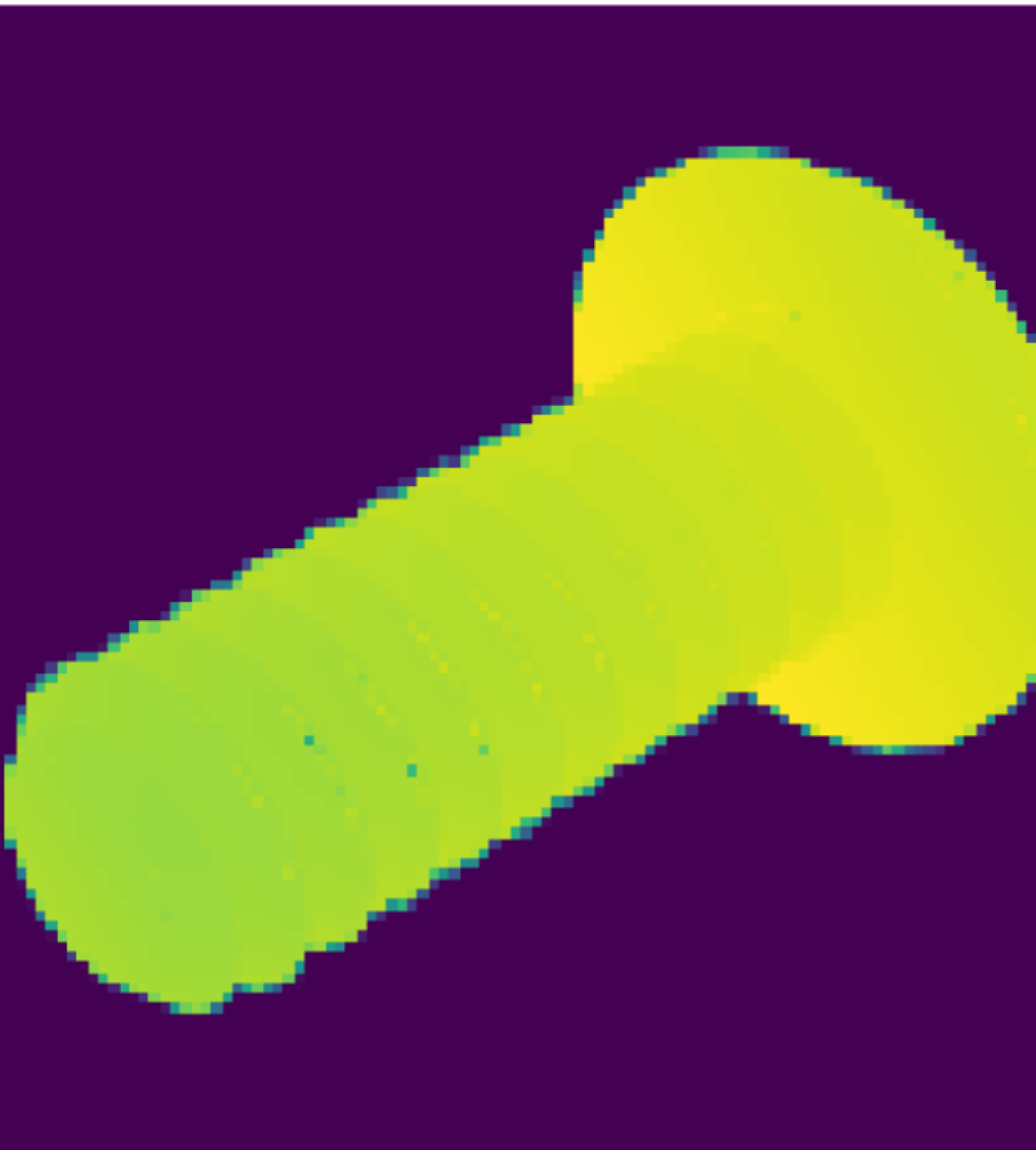}  &
			\fcolorbox{green}{white}{\includegraphics[trim={9cm 4cm 9cm 4cm}, clip = true,width=0.12\linewidth]{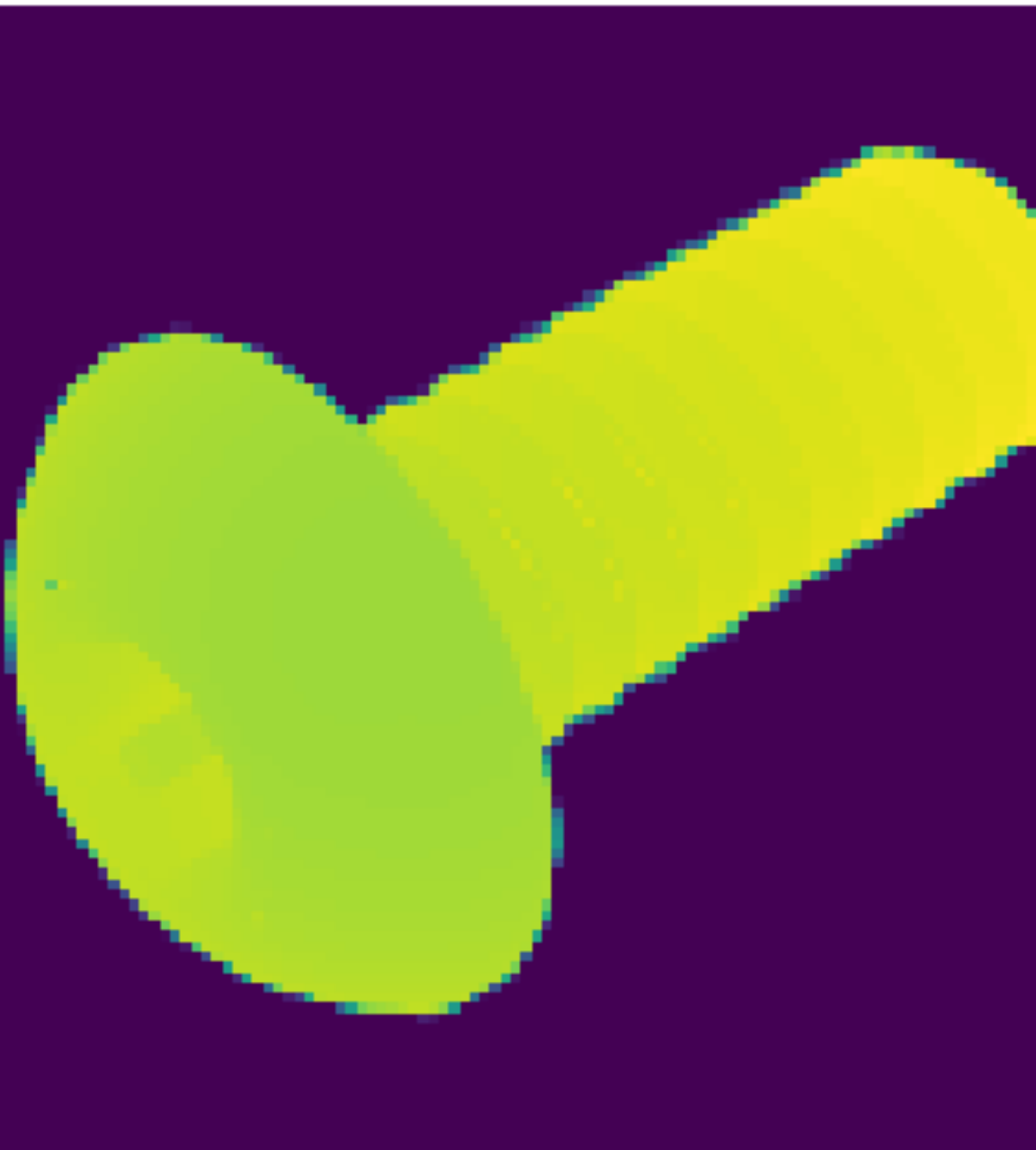} } &
			\fcolorbox{green}{white}{\includegraphics[trim={9cm 4cm 9cm 4cm}, clip = true,width=0.12\linewidth]{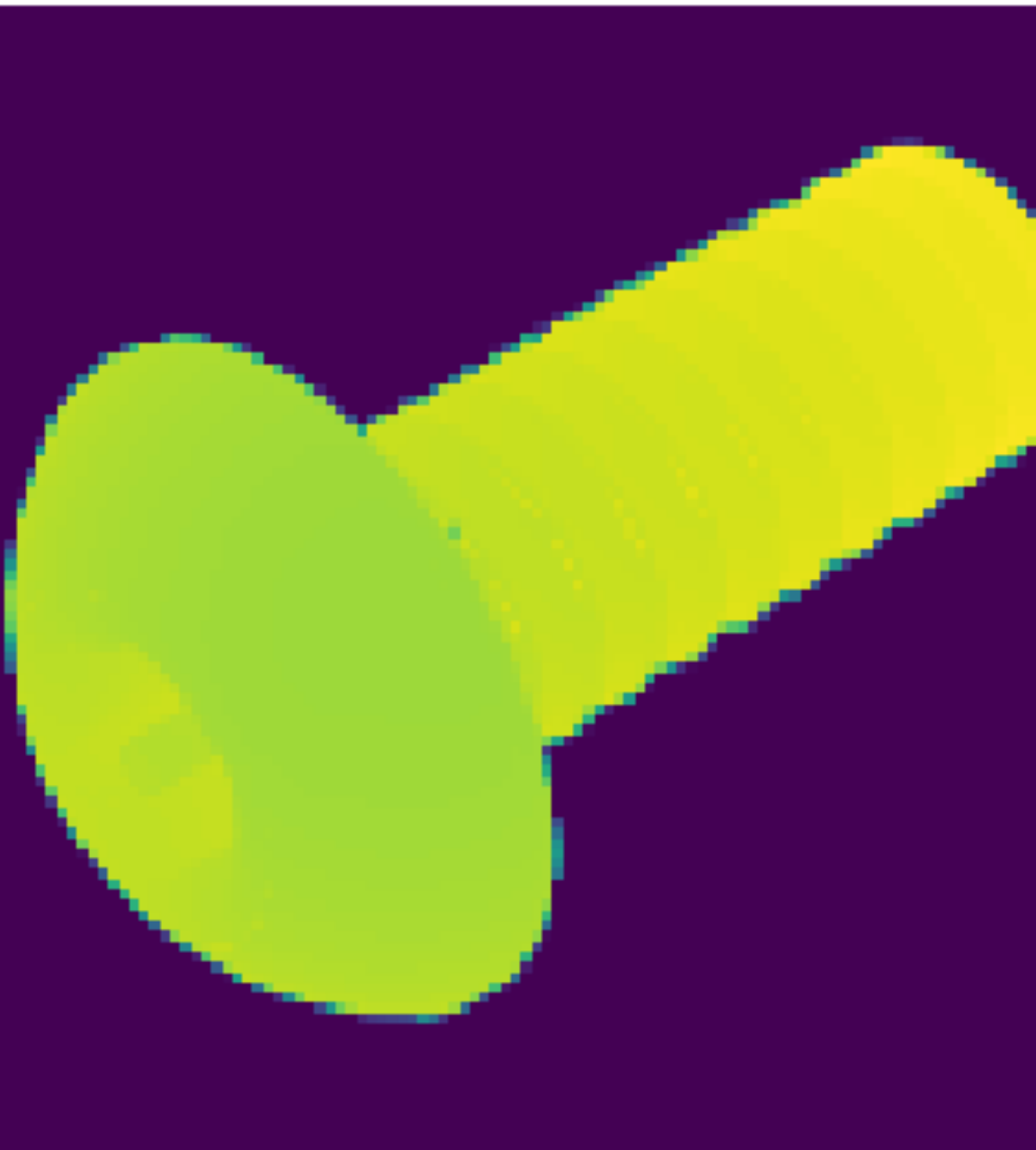} } & \raisebox{2\height}{\LARGE 4.76$^{\circ}$ } \\ \hline
			\includegraphics[trim={9cm 4cm 9cm 4cm}, clip = true,width=0.12\linewidth]{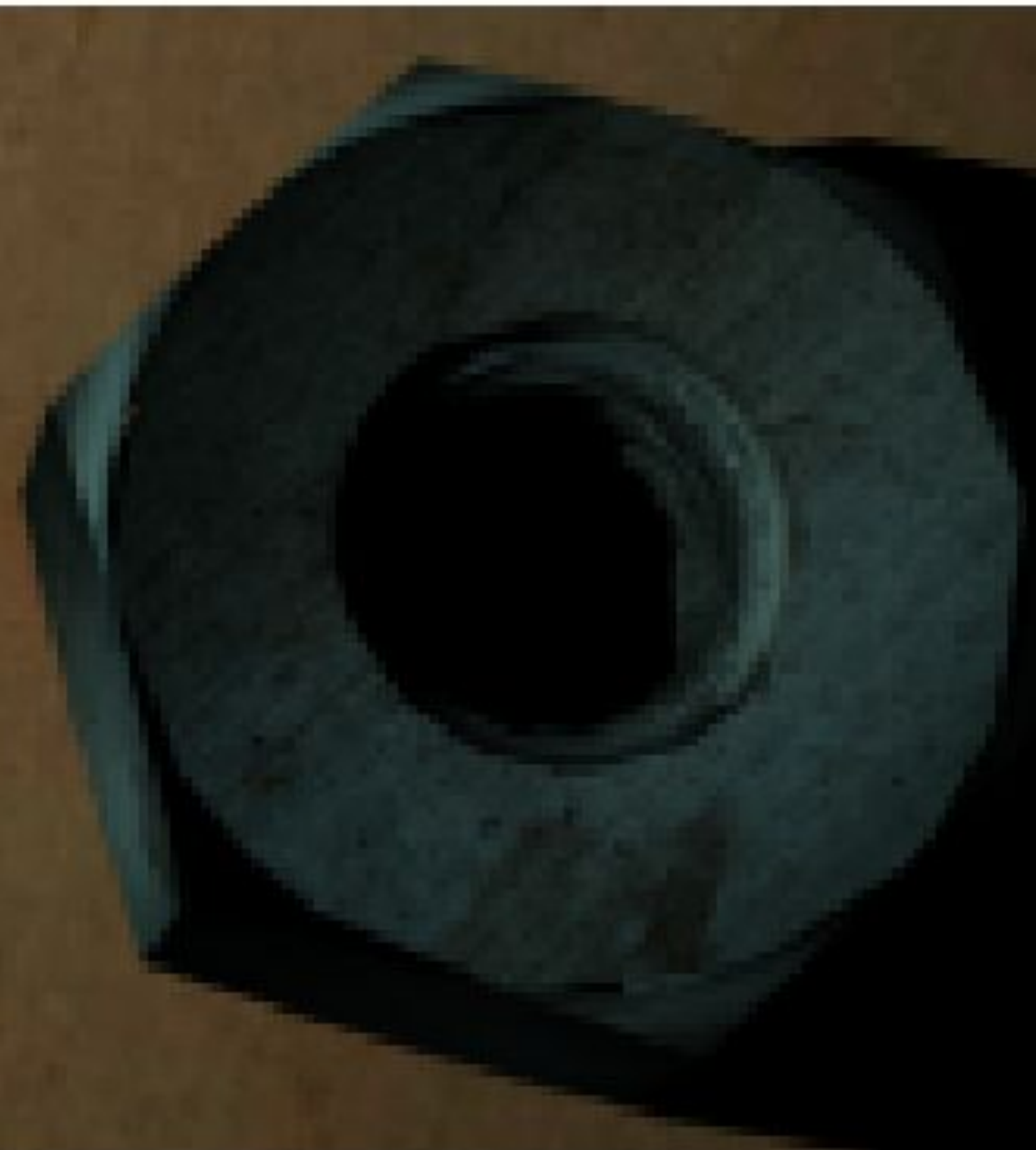} &
			\includegraphics[trim={9cm 4cm 9cm 4cm}, clip = true,width=0.12\linewidth]{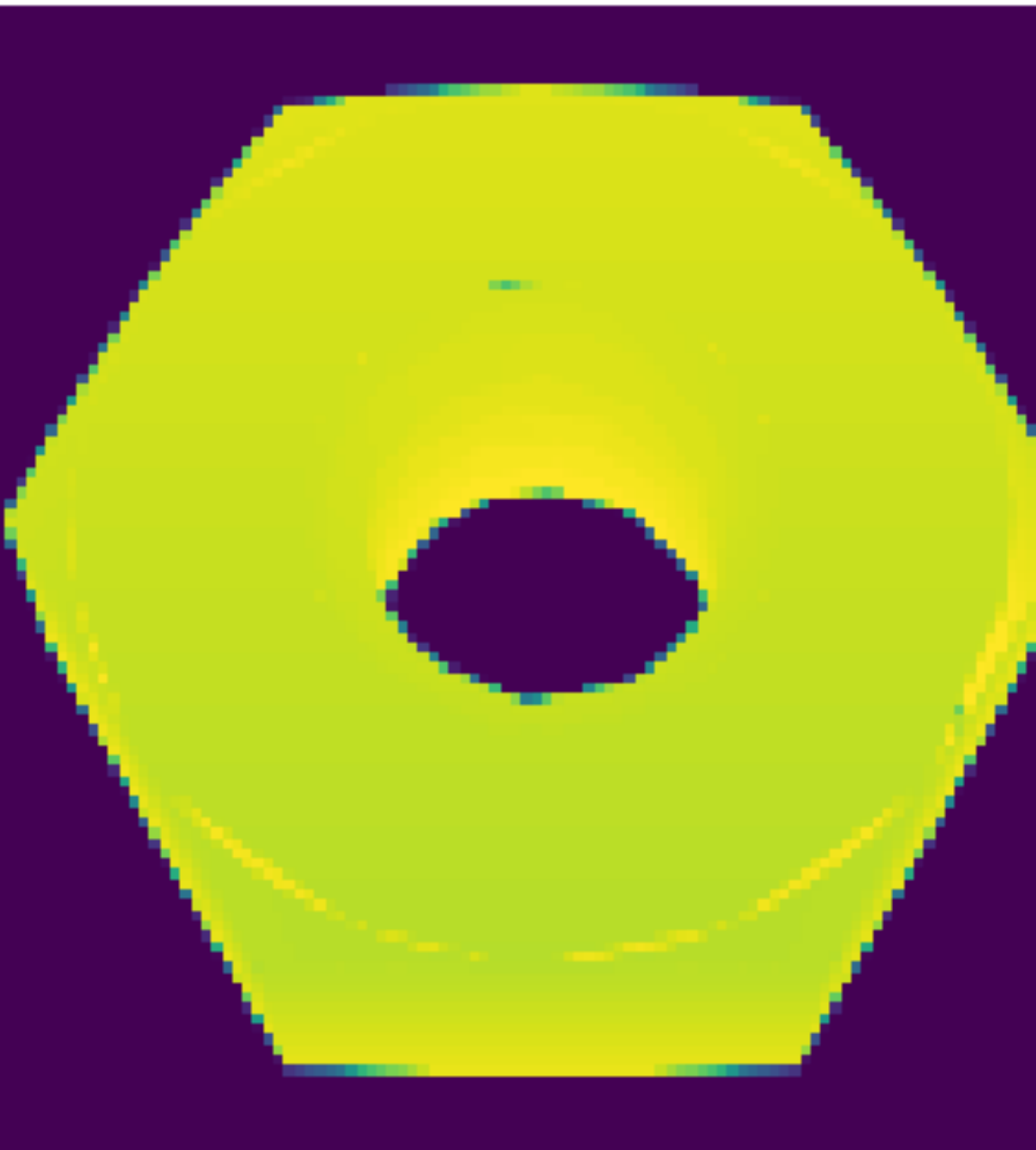} &
			\fcolorbox{green}{white}{\includegraphics[trim={9cm 4cm 9cm 4cm}, clip = true,width=0.12\linewidth]{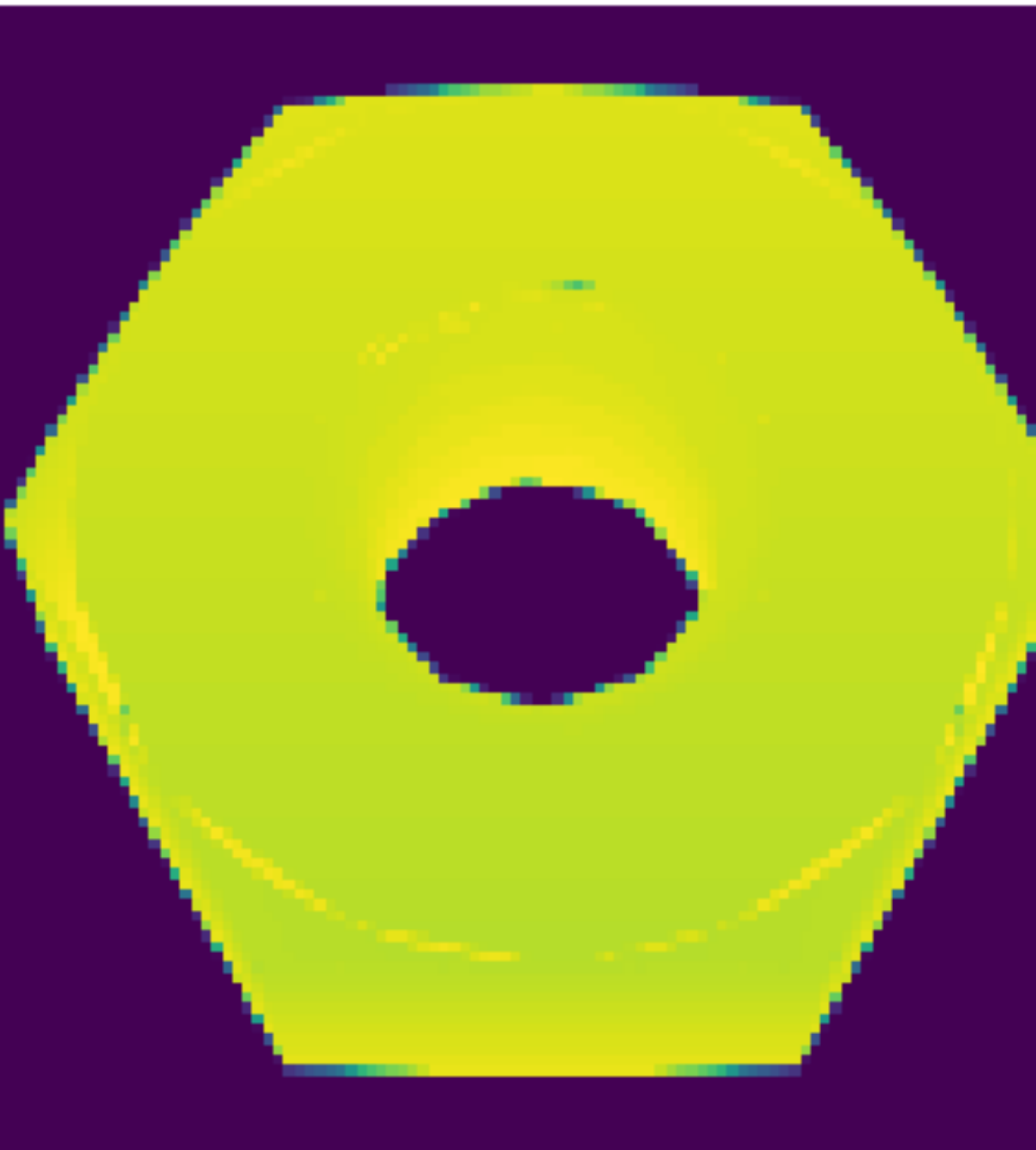}} &
			\fcolorbox{green}{white}{\includegraphics[trim={9cm 4cm 9cm 4cm}, clip = true,width=0.12\linewidth]{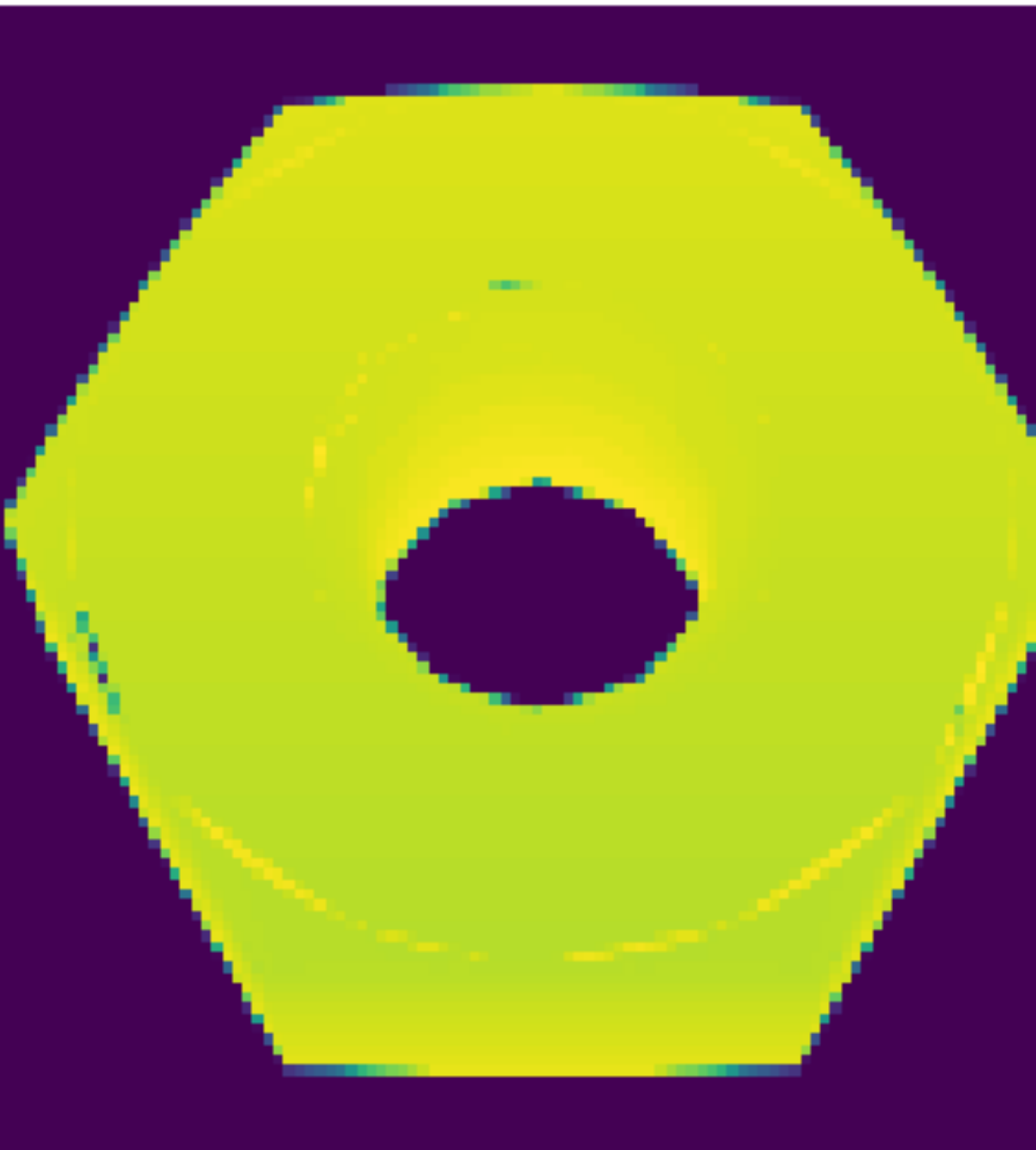}} &
			\fcolorbox{green}{white}{\includegraphics[trim={9cm 4cm 9cm 4cm}, clip = true,width=0.12\linewidth]{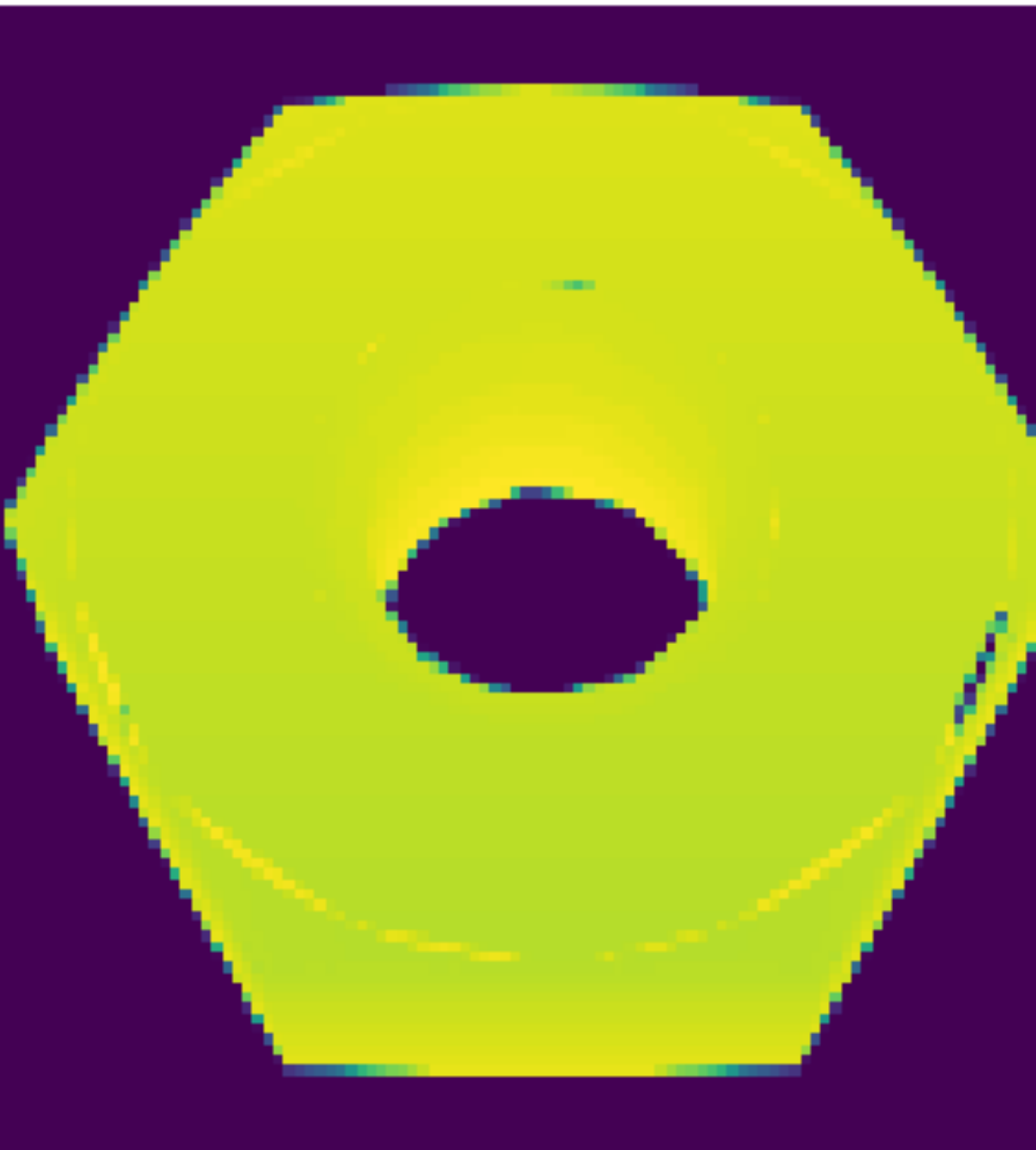}} &
			\fcolorbox{green}{white}{\includegraphics[trim={9cm 4cm 9cm 4cm}, clip = true,width=0.12\linewidth]{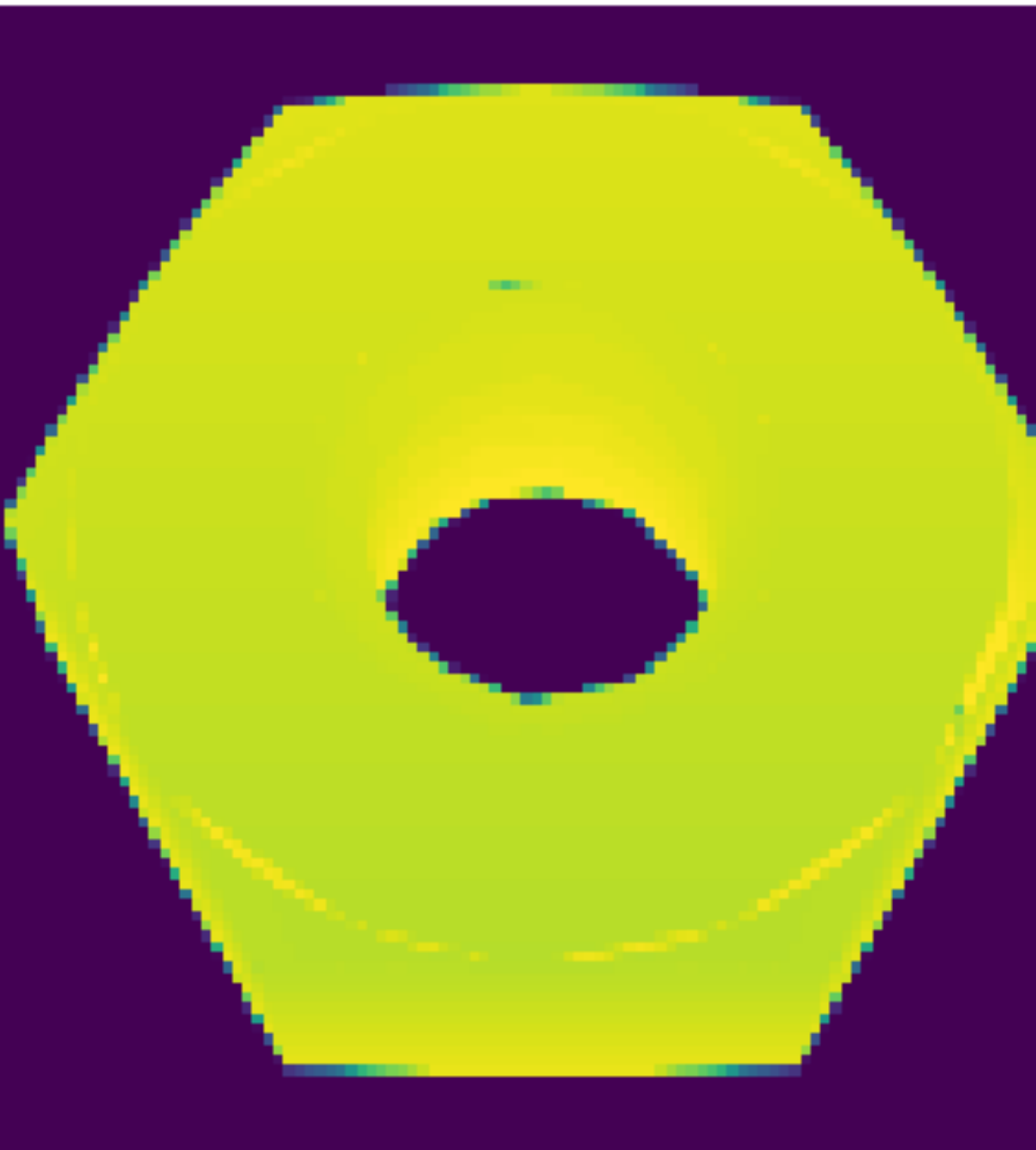}} & \raisebox{2\height}{\LARGE 5.85$^{\circ}$ } \\ \hline
			\includegraphics[trim={9cm 4cm 9cm 4cm}, clip = true,width=0.12\linewidth]{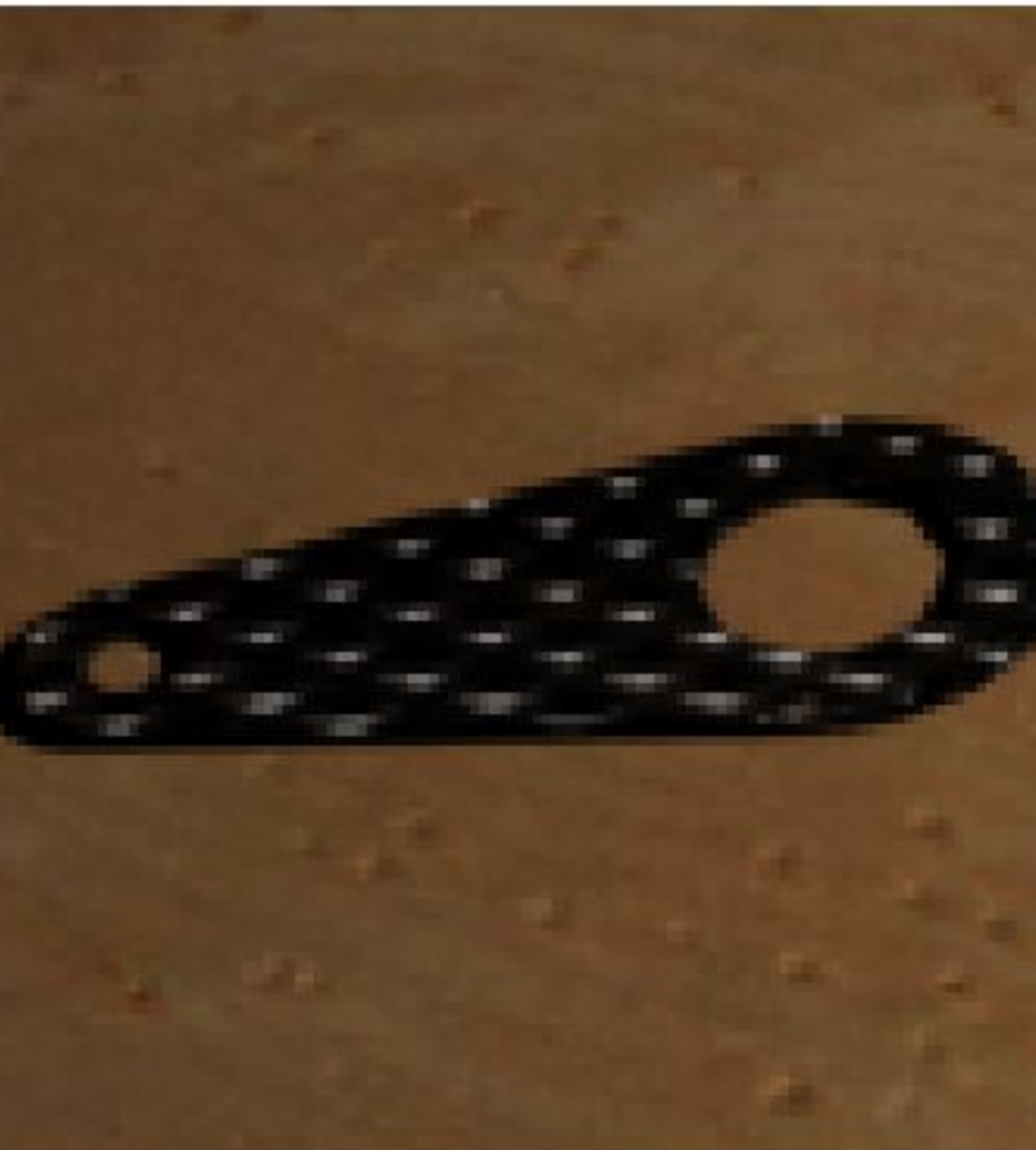} &
			\includegraphics[trim={9cm 4cm 9cm 4cm}, clip = true,width=0.12\linewidth]{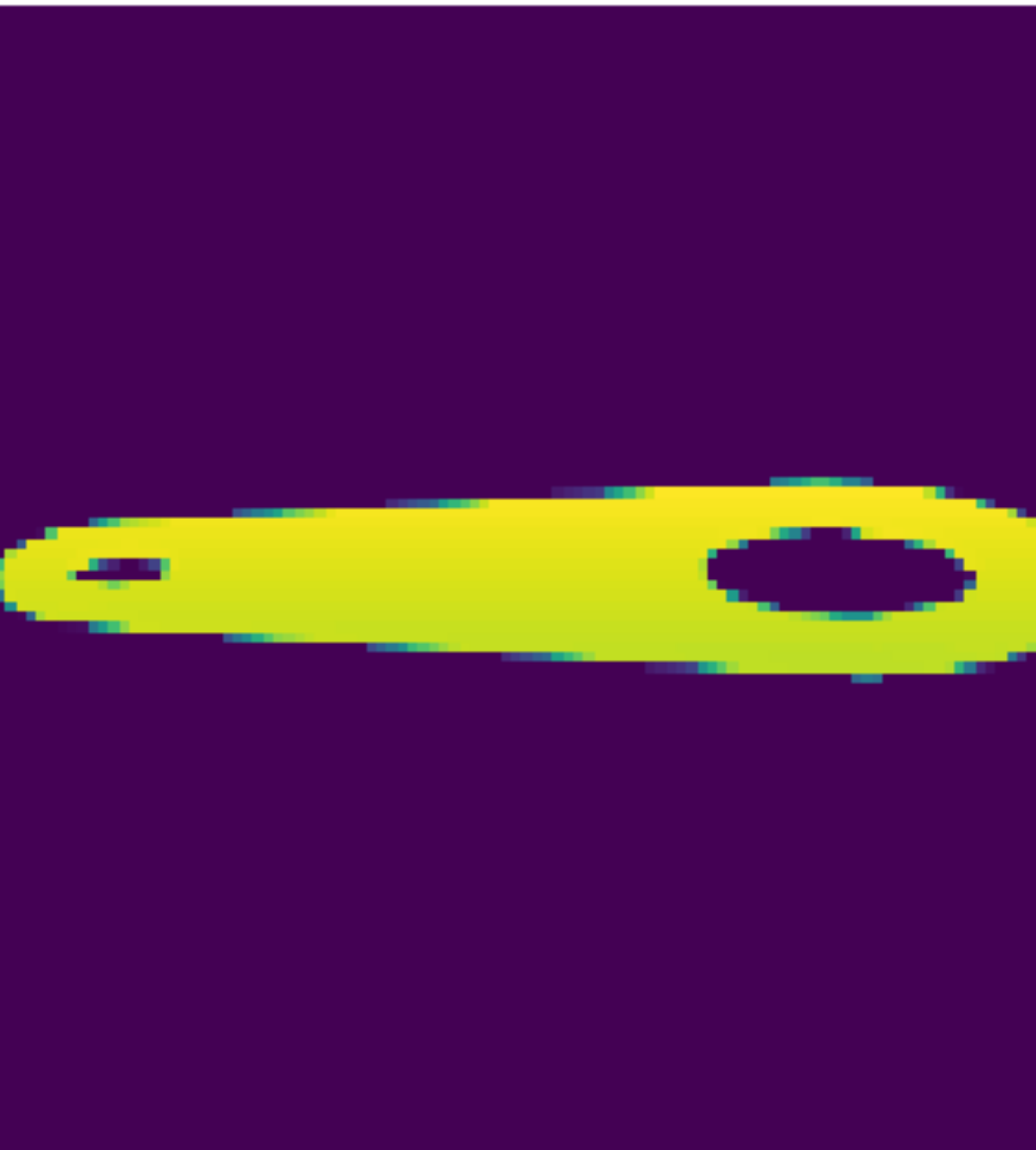} &
			\fcolorbox{green}{white}{\includegraphics[trim={9cm 4cm 9cm 4cm}, clip = true,width=0.12\linewidth]{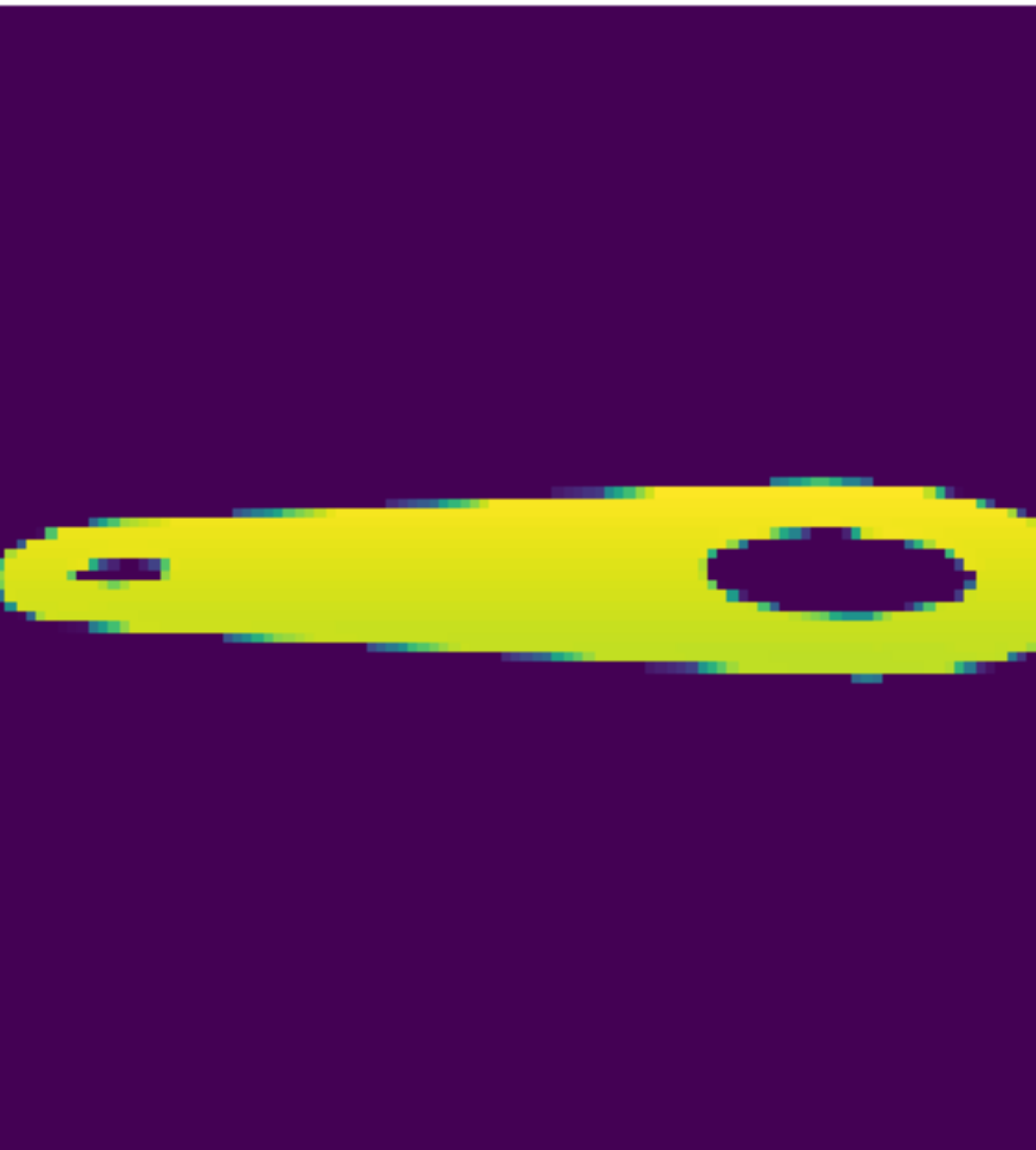}} &
			\fcolorbox{green}{white}{\includegraphics[trim={9cm 4cm 9cm 4cm}, clip = true,width=0.12\linewidth]{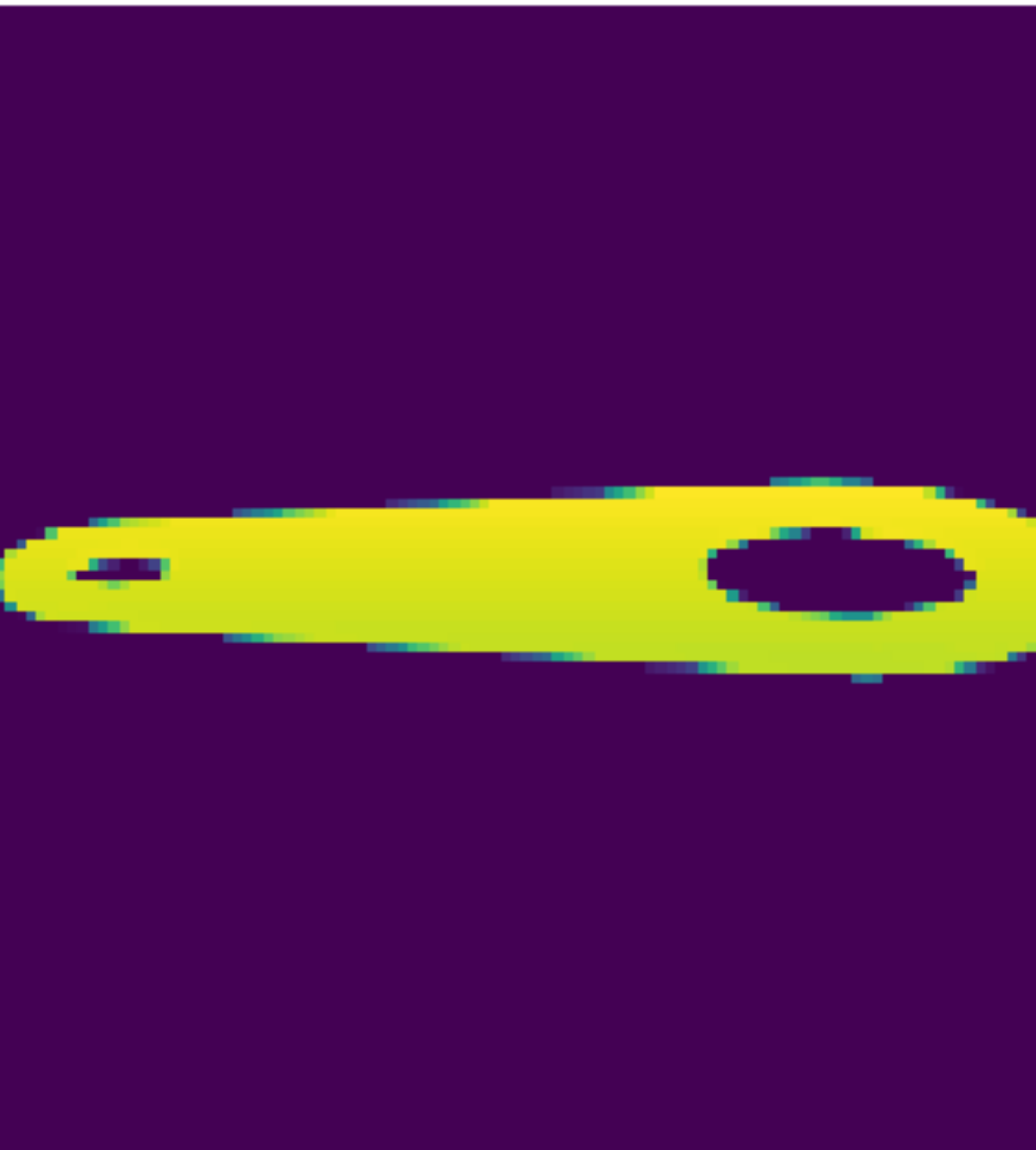}} &
			\includegraphics[trim={9cm 4cm 9cm 4cm}, clip = true,width=0.12\linewidth]{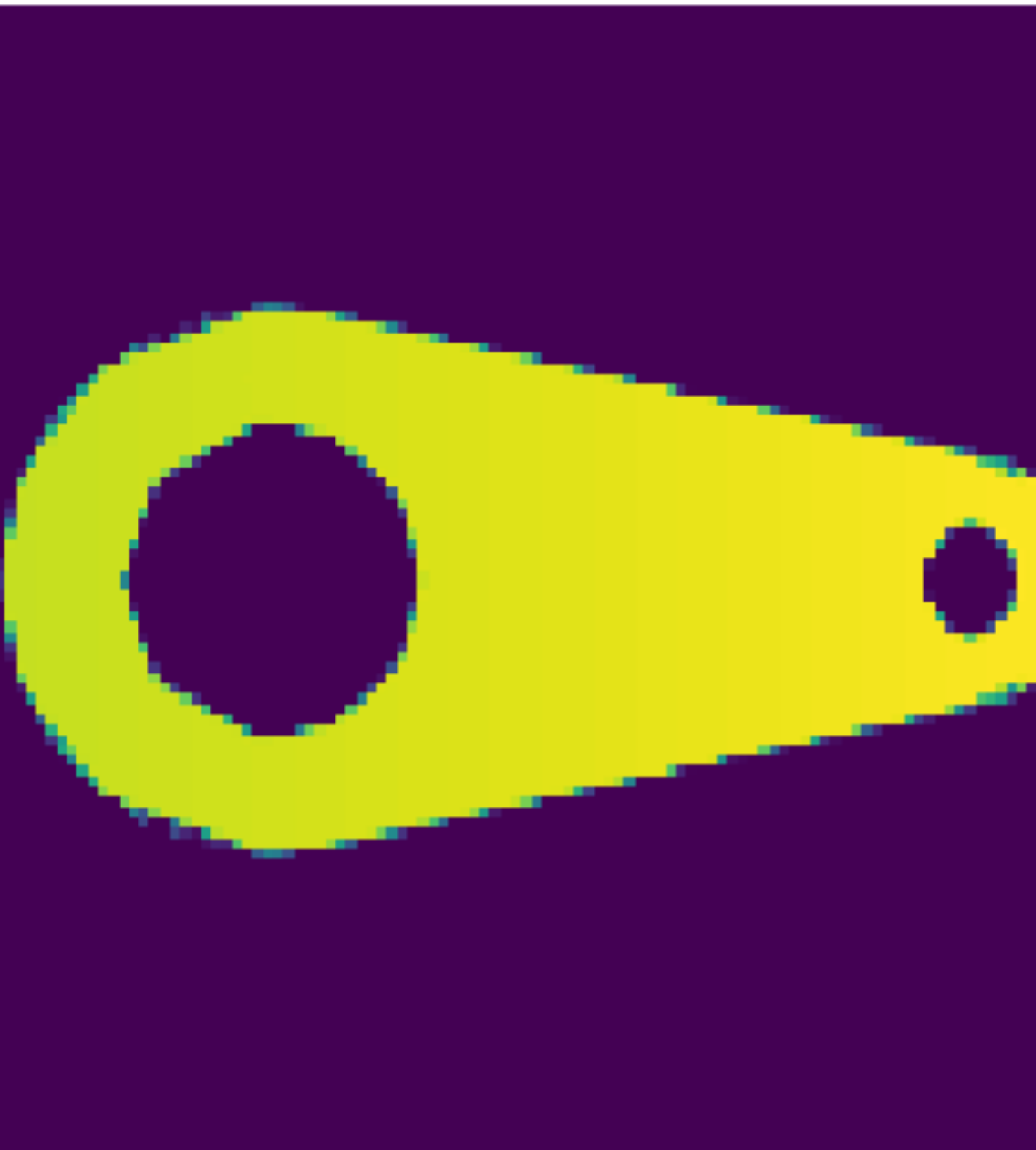} &
			\includegraphics[trim={9cm 4cm 9cm 4cm}, clip = true,width=0.12\linewidth]{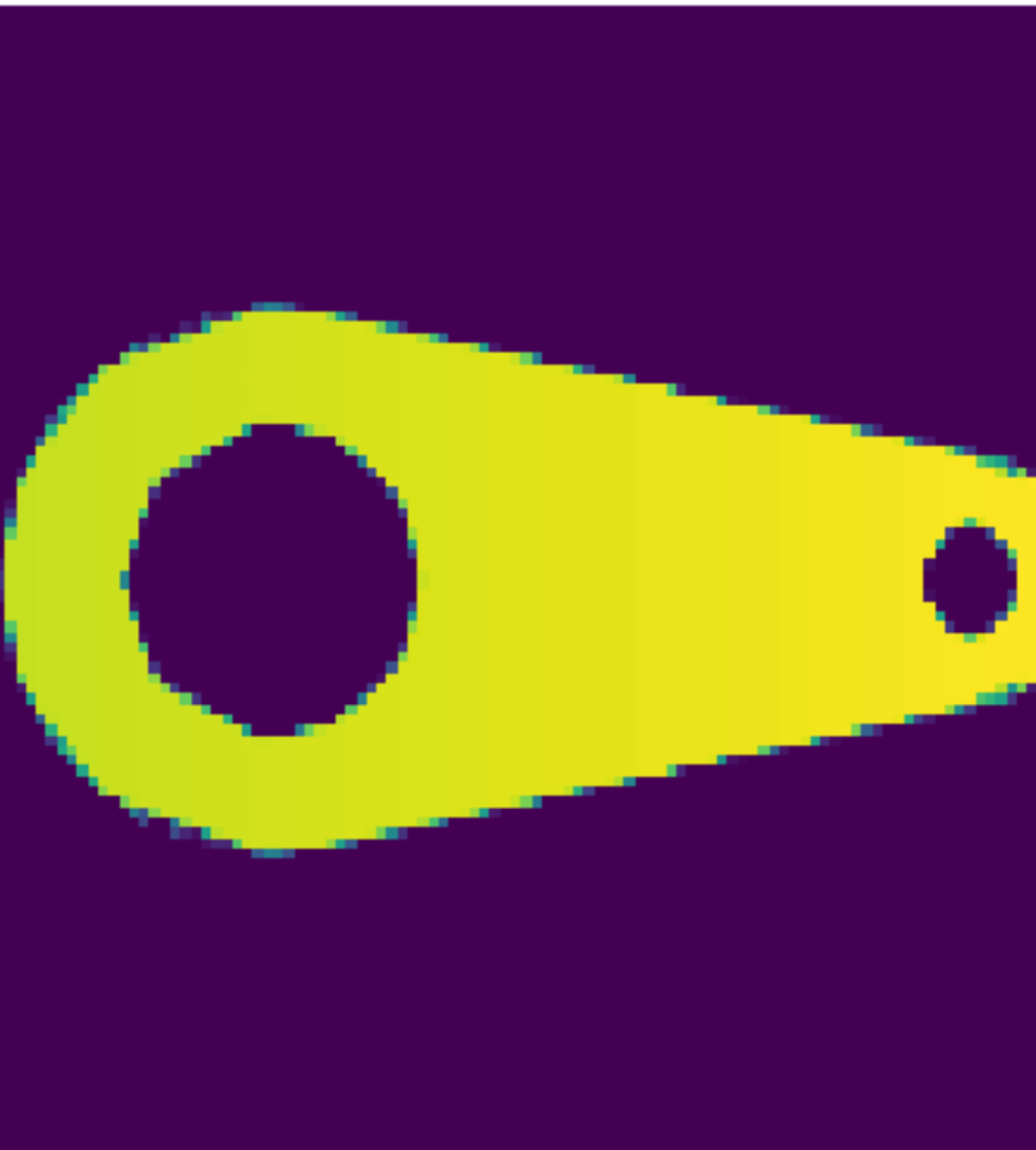} & \raisebox{2\height}{\LARGE 6.52$^{\circ}$ } \\ \hline
			\includegraphics[trim={9cm 4cm 9cm 4cm}, clip = true,width=0.12\linewidth]{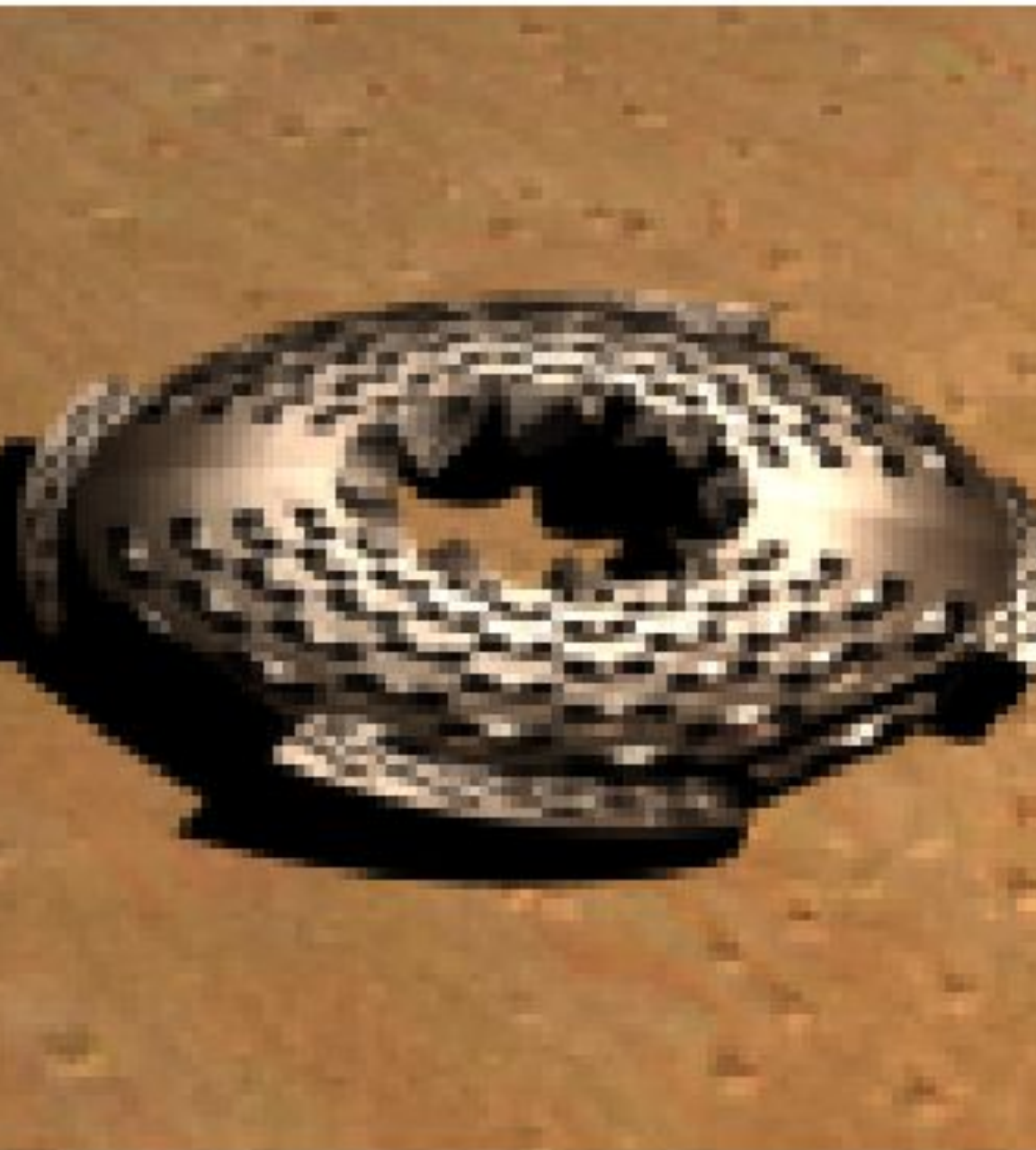}  &
			\includegraphics[trim={9cm 4cm 9cm 4cm}, clip = true,width=0.12\linewidth]{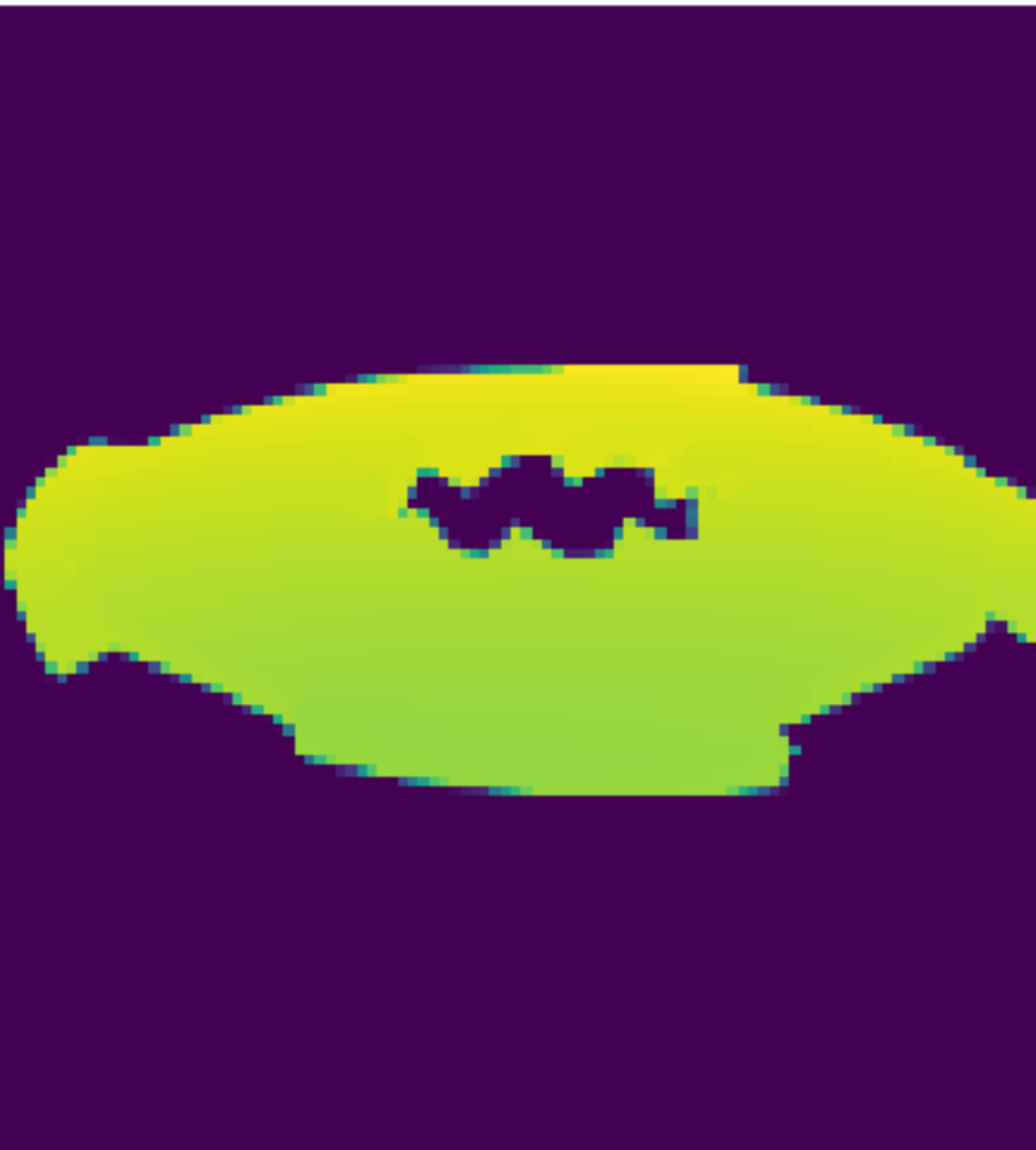}  &
			\includegraphics[trim={9cm 4cm 9cm 4cm}, clip = true,width=0.12\linewidth]{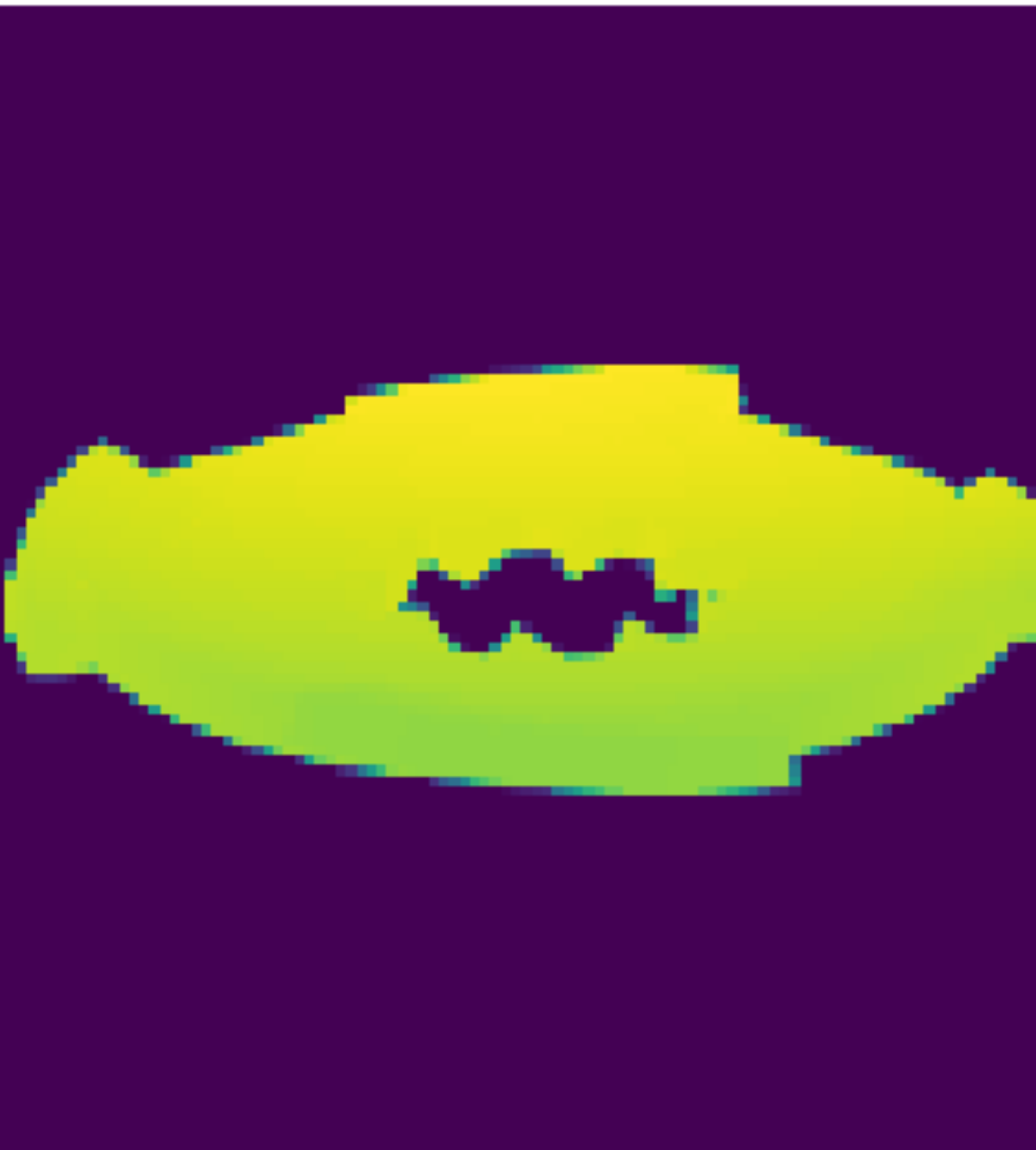}  &
			\includegraphics[trim={9cm 4cm 9cm 4cm}, clip = true,width=0.12\linewidth]{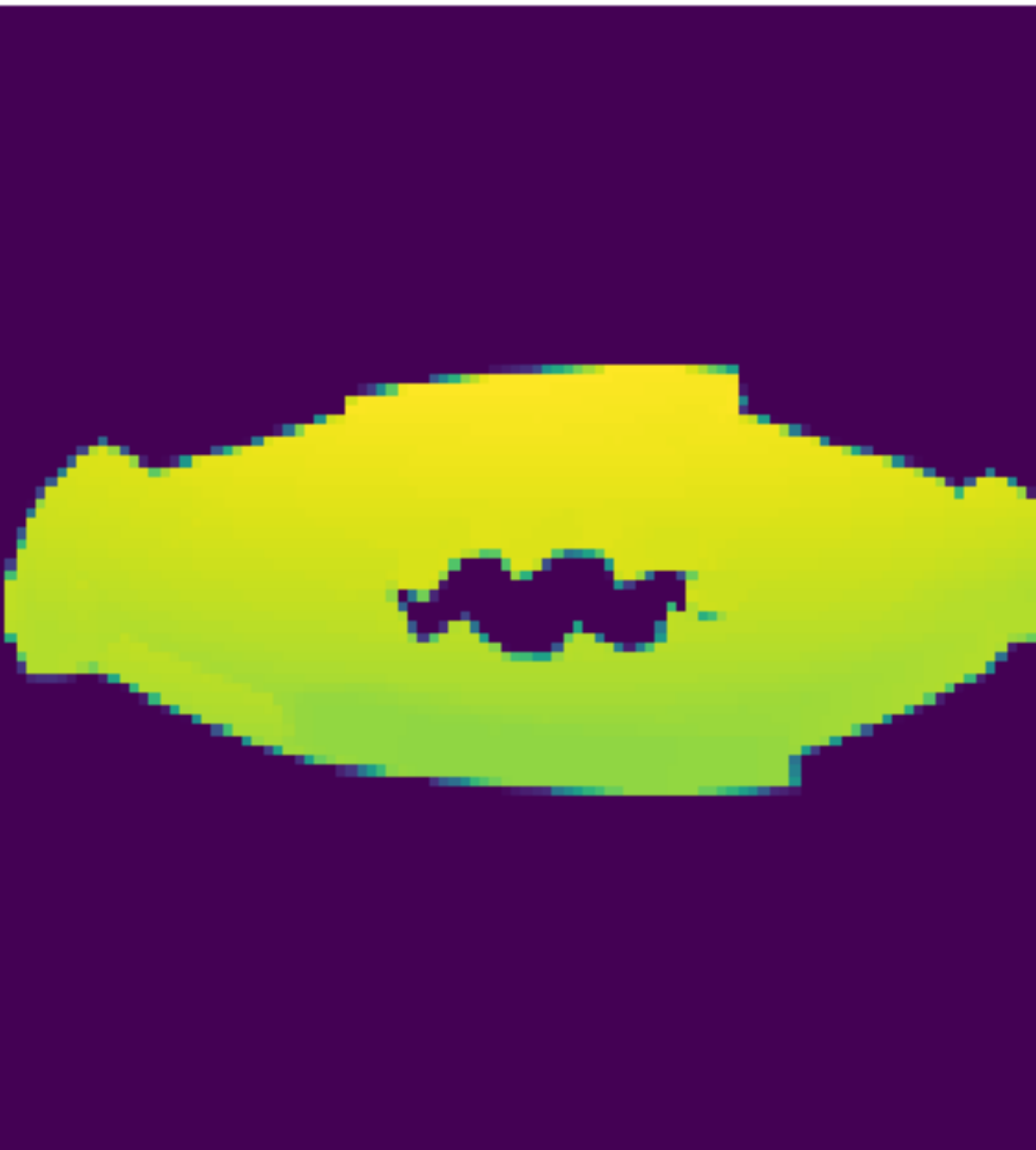}  &
			\fcolorbox{green}{white}{\includegraphics[trim={9cm 4cm 9cm 4cm}, clip = true,width=0.12\linewidth]{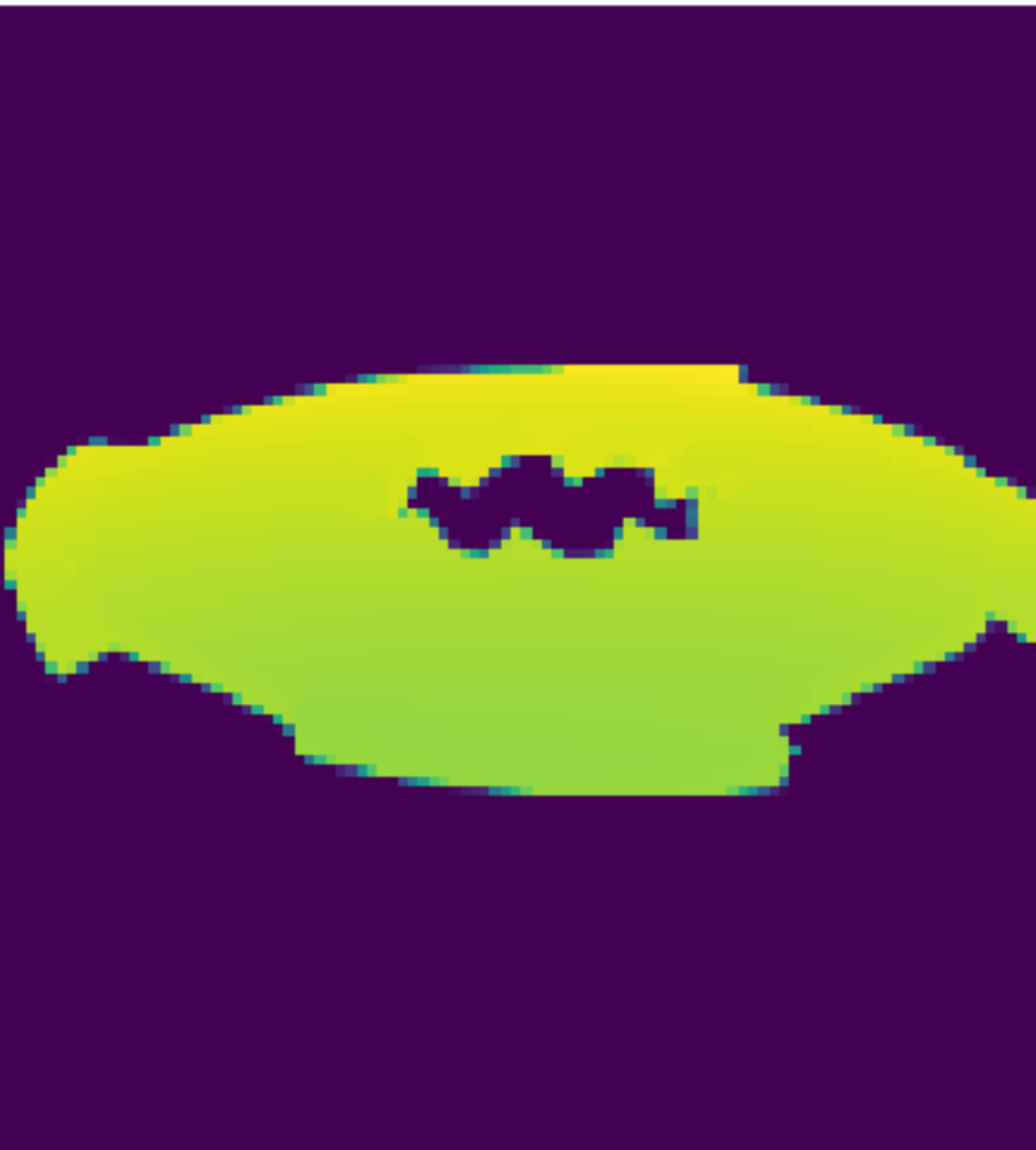}  }&
			\fcolorbox{green}{white}{\includegraphics[trim={9cm 4cm 9cm 4cm}, clip = true,width=0.12\linewidth]{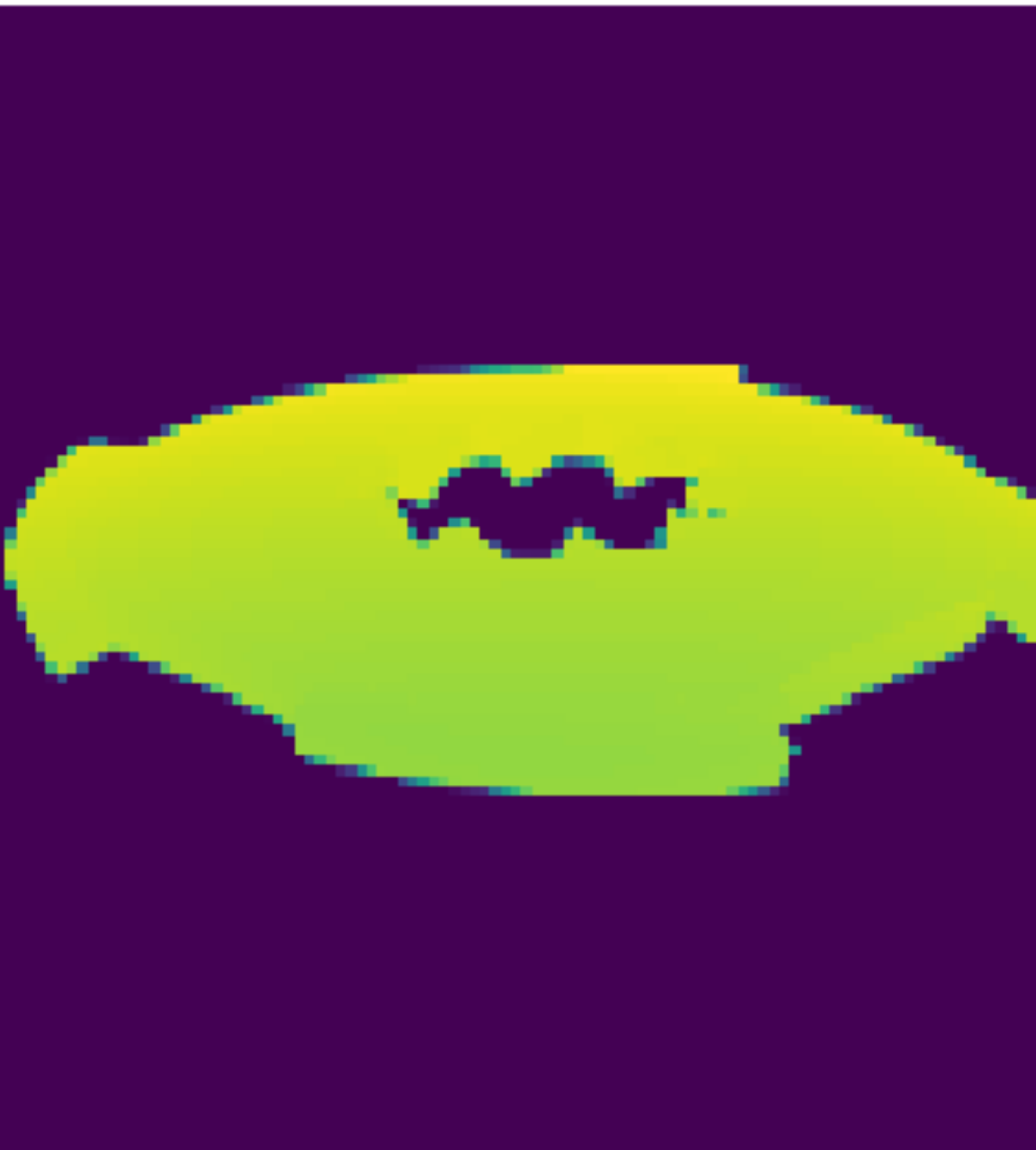}  }& \raisebox{2\height}{\LARGE 6.96$^{\circ}$ } \\ \hline
	\end{tabular} } 
	\centering
	\caption{{\bf Successful Results}: Qualitative results for pose estimation on the \textbf{synthetic} dataset.} % The object in the scene is shown in the first column. The second column shows the nearest discretizated view. When the top four predicted views shown in the third column match the nearest view, they are indicated in a \textbf{green} box. Otherwise, an \textbf{orange} box shows the view that is nearest among the predicted. The last column shows the distance of this view. Given objects with symmetry, there can be more than one best view.}
	\label{fig:synth_result_2}
\end{figure*}

\begin{figure*} % \ContinuedFloat
	\hspace{-1.1cm}\parbox{\linewidth}{
		\begin{tabular}{|c|c|cccc|c|}
			\hline
			{\bf Input } & {\bf GT } & \multicolumn{4}{c|}{\bf Top-4 Predictions} & {\bf  $ d_\text{rot, best}^{sym} $} \\
			\hline 
			\includegraphics[trim={9cm 4cm 9cm 4cm}, clip = true,width=0.12\linewidth]{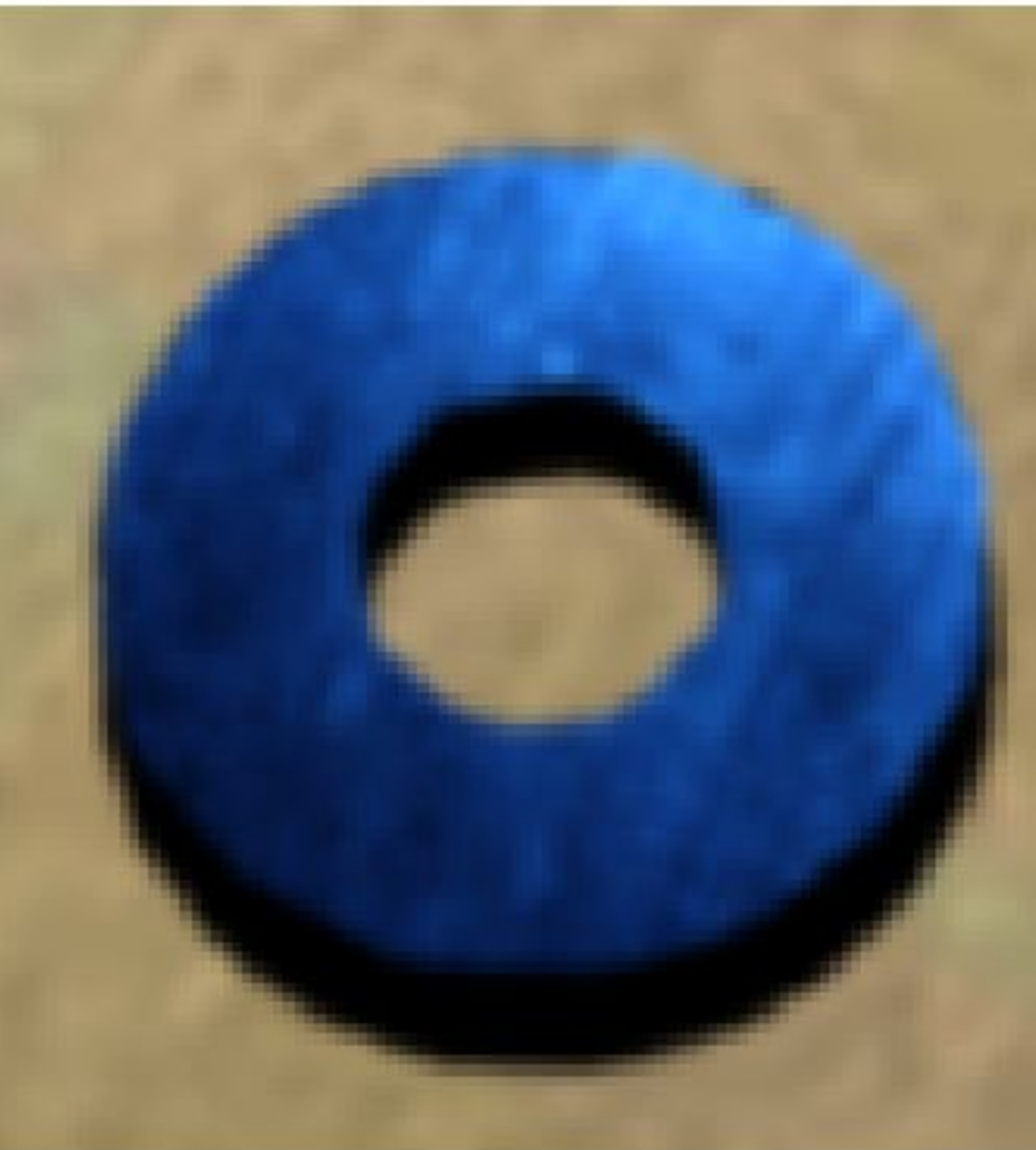} &
			\includegraphics[trim={9cm 4cm 9cm 4cm}, clip = true,width=0.12\linewidth]{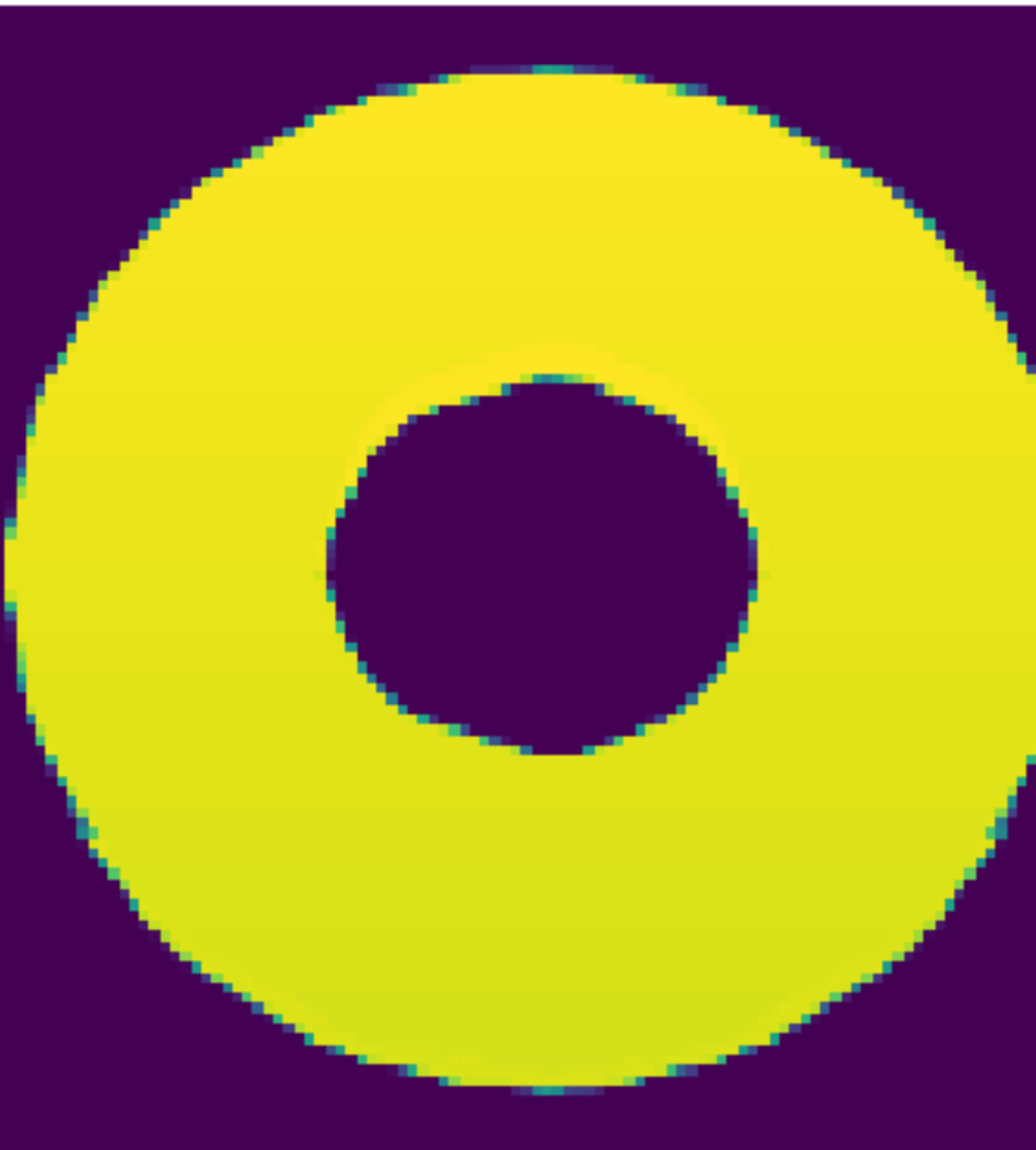} &
			\fcolorbox{green}{white}{\includegraphics[trim={9cm 4cm 9cm 4cm}, clip = true,width=0.12\linewidth]{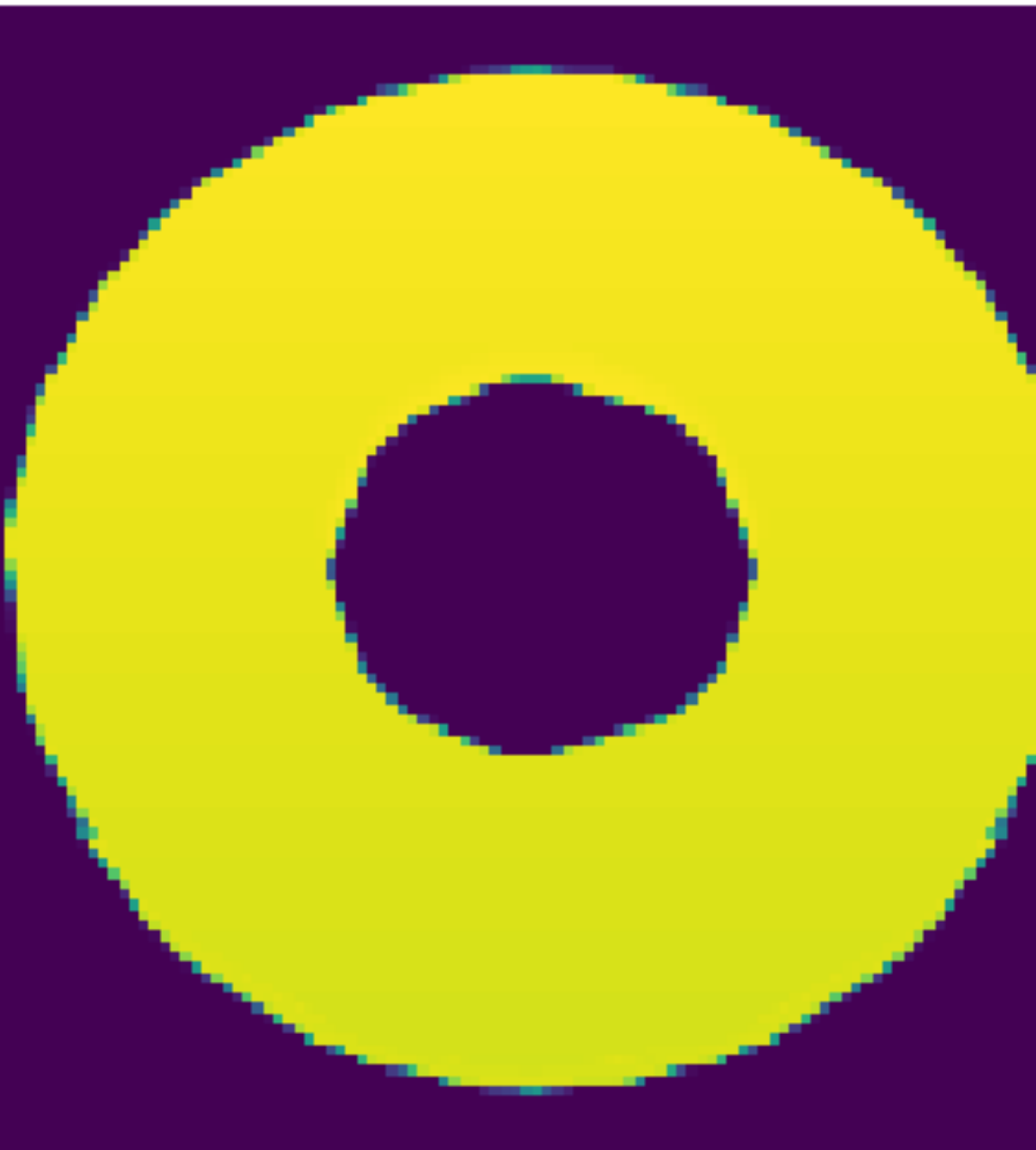}} &
			\fcolorbox{green}{white}{\includegraphics[trim={9cm 4cm 9cm 4cm}, clip = true,width=0.12\linewidth]{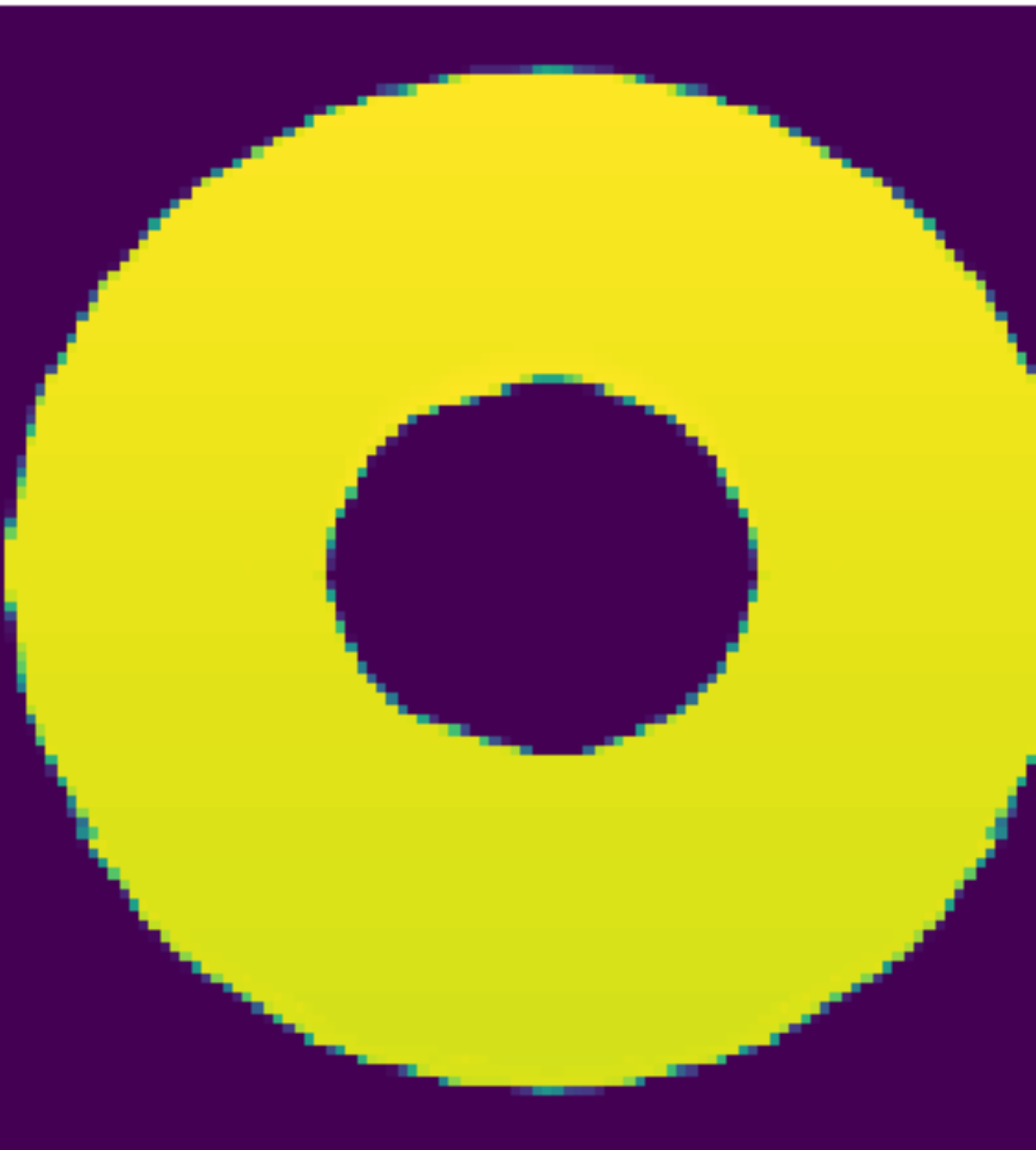}} &
			\fcolorbox{green}{white}{\includegraphics[trim={9cm 4cm 9cm 4cm}, clip = true,width=0.12\linewidth]{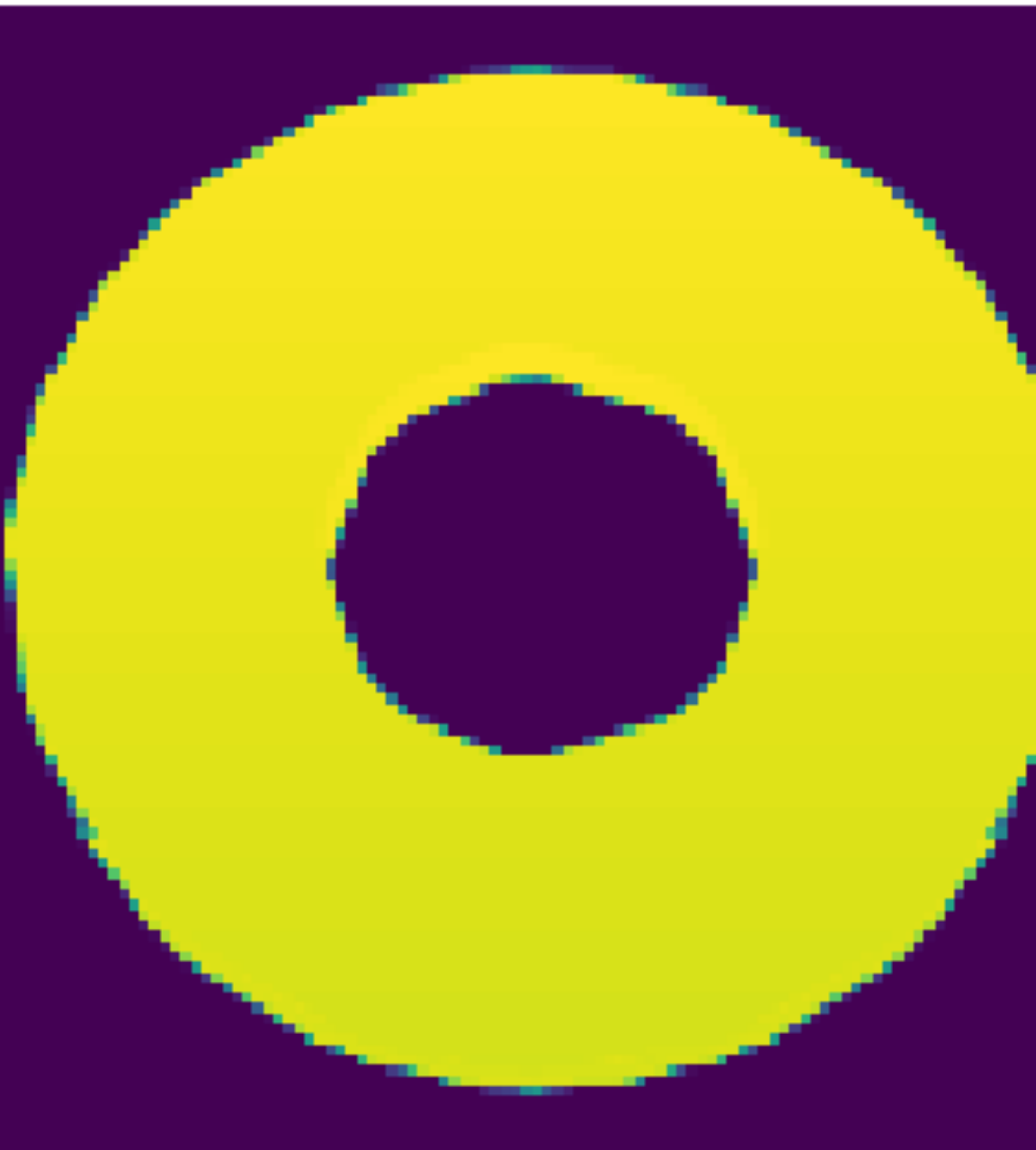}} &
			\fcolorbox{green}{white}{\includegraphics[trim={9cm 4cm 9cm 4cm}, clip = true,width=0.12\linewidth]{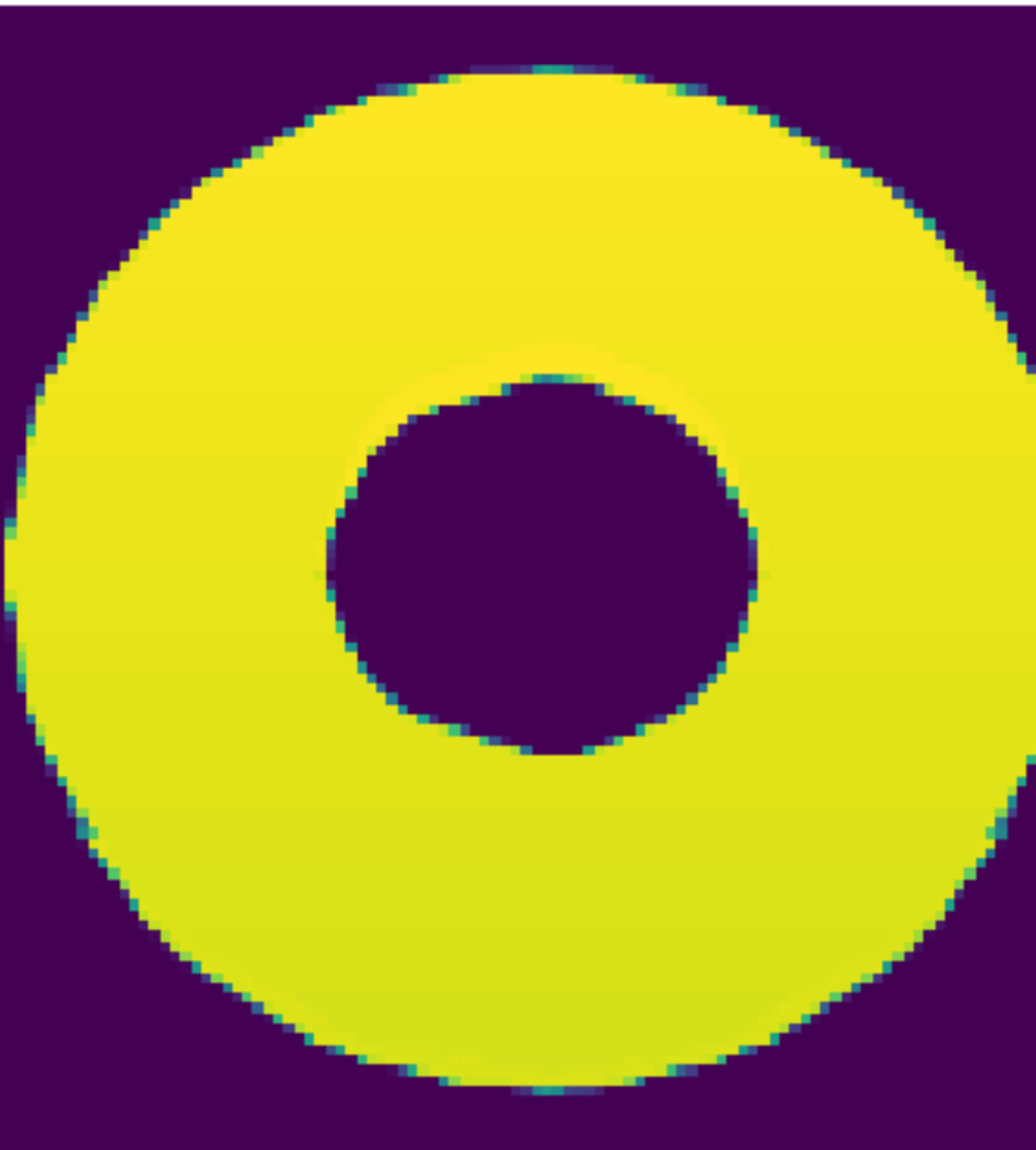}} & \raisebox{2\height}{\LARGE 7.73$^{\circ}$ } \\ \hline
			\includegraphics[trim={9cm 4cm 9cm 4cm}, clip = true,width=0.12\linewidth]{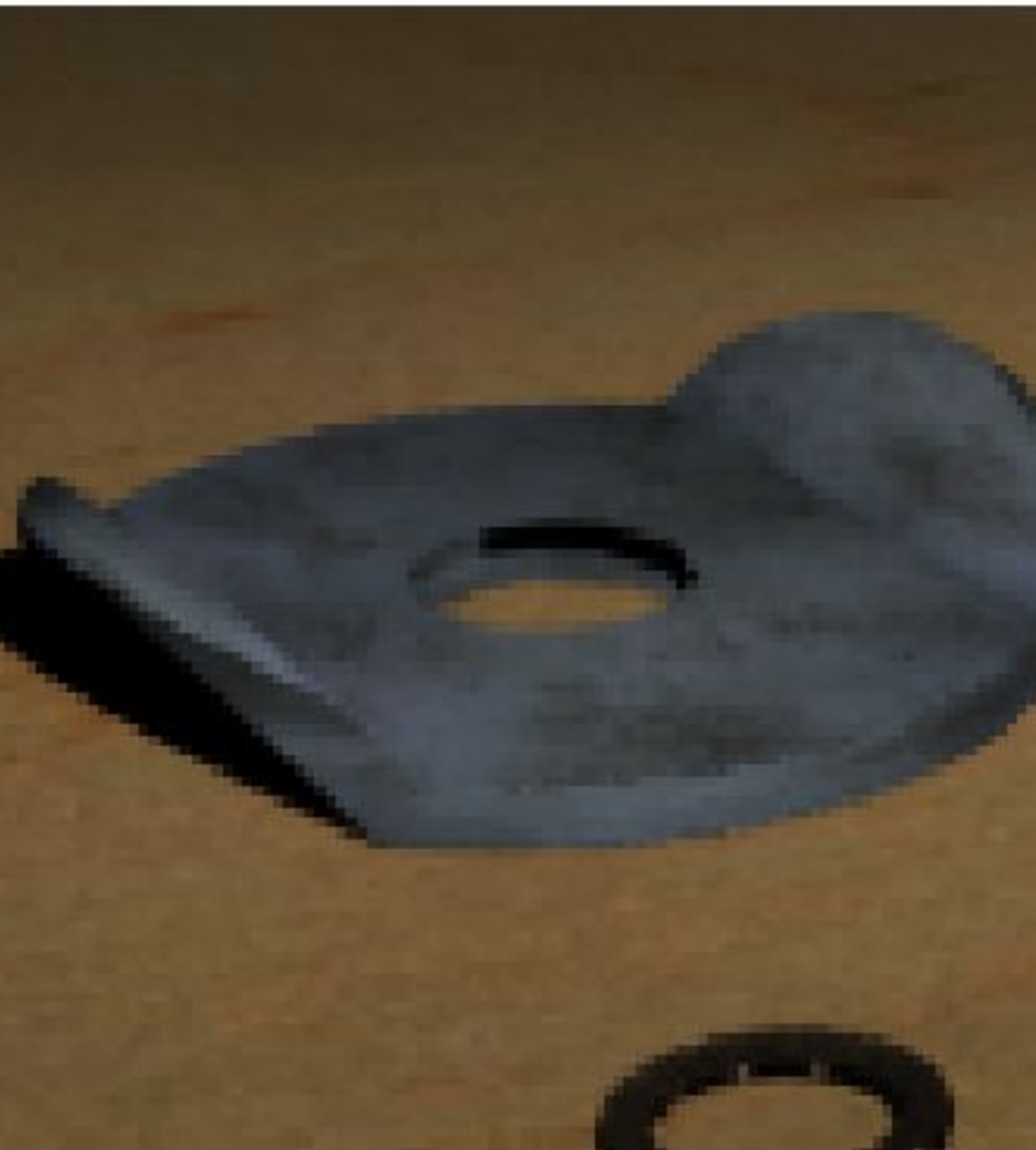}  &
			\includegraphics[trim={9cm 4cm 9cm 4cm}, clip = true,width=0.12\linewidth]{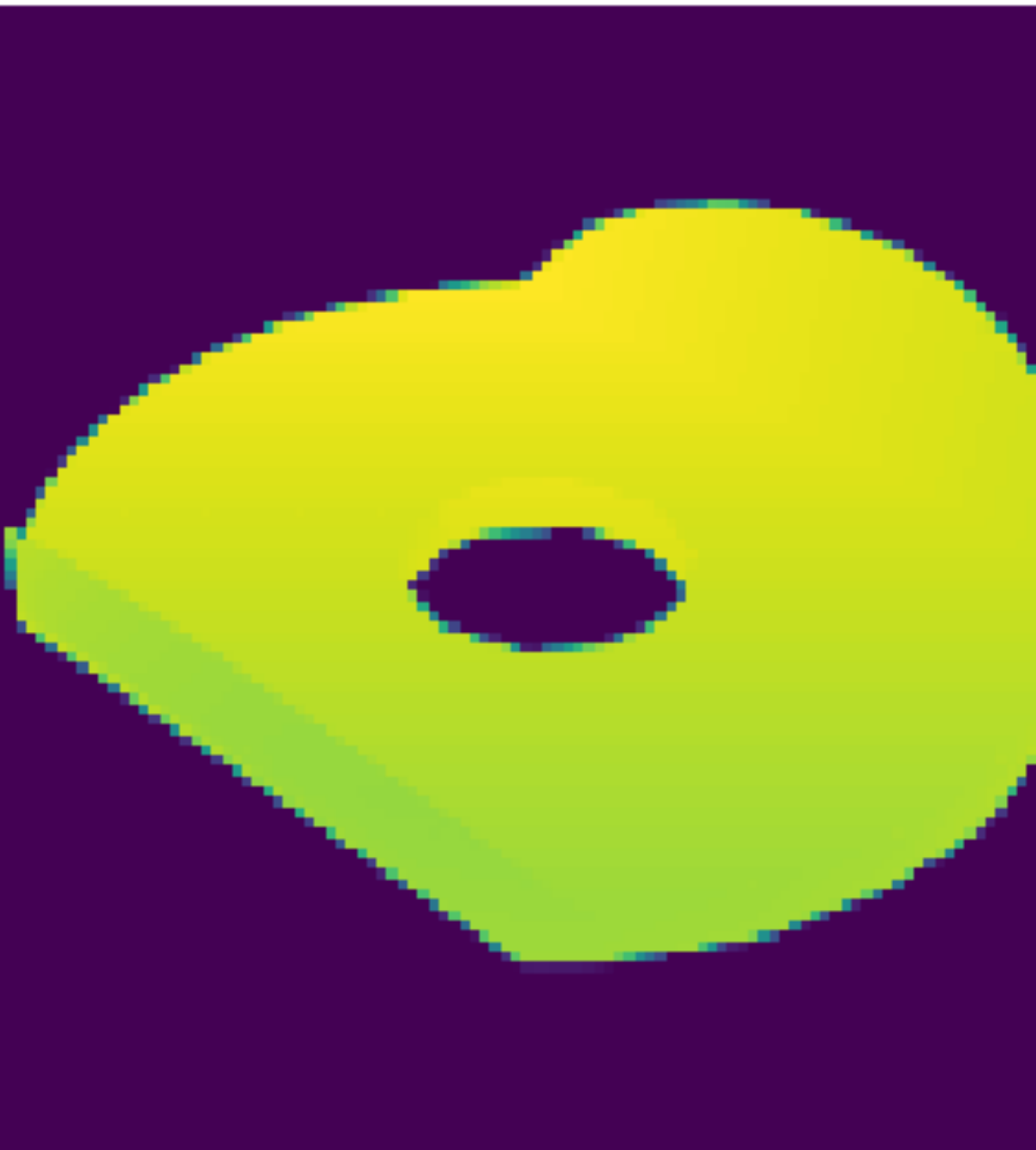}  &
			\includegraphics[trim={9cm 4cm 9cm 4cm}, clip = true,width=0.12\linewidth]{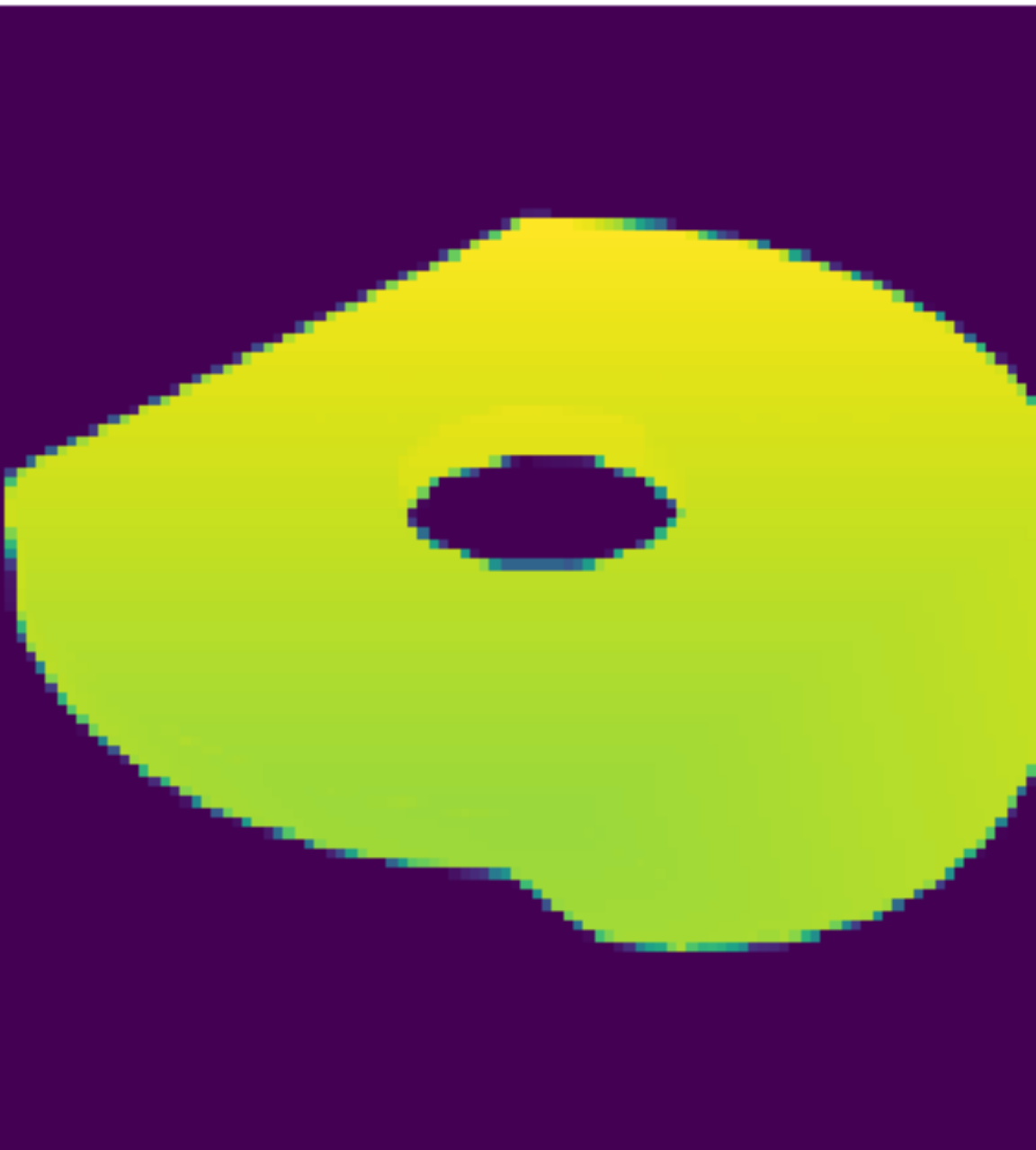}  &
			\includegraphics[trim={9cm 4cm 9cm 4cm}, clip = true,width=0.12\linewidth]{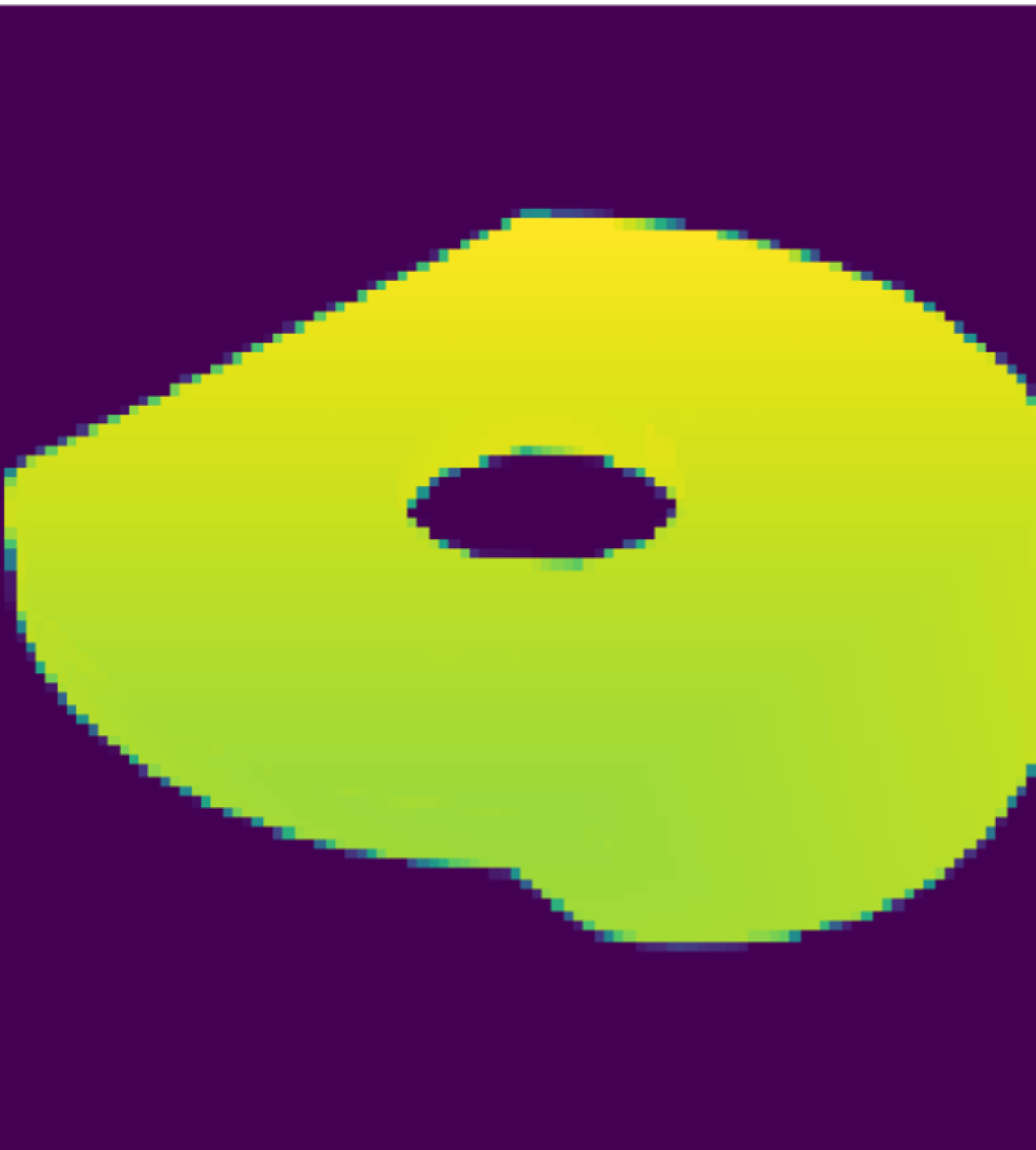}  &
			\fcolorbox{green}{white}{\includegraphics[trim={9cm 4cm 9cm 4cm}, clip = true,width=0.12\linewidth]{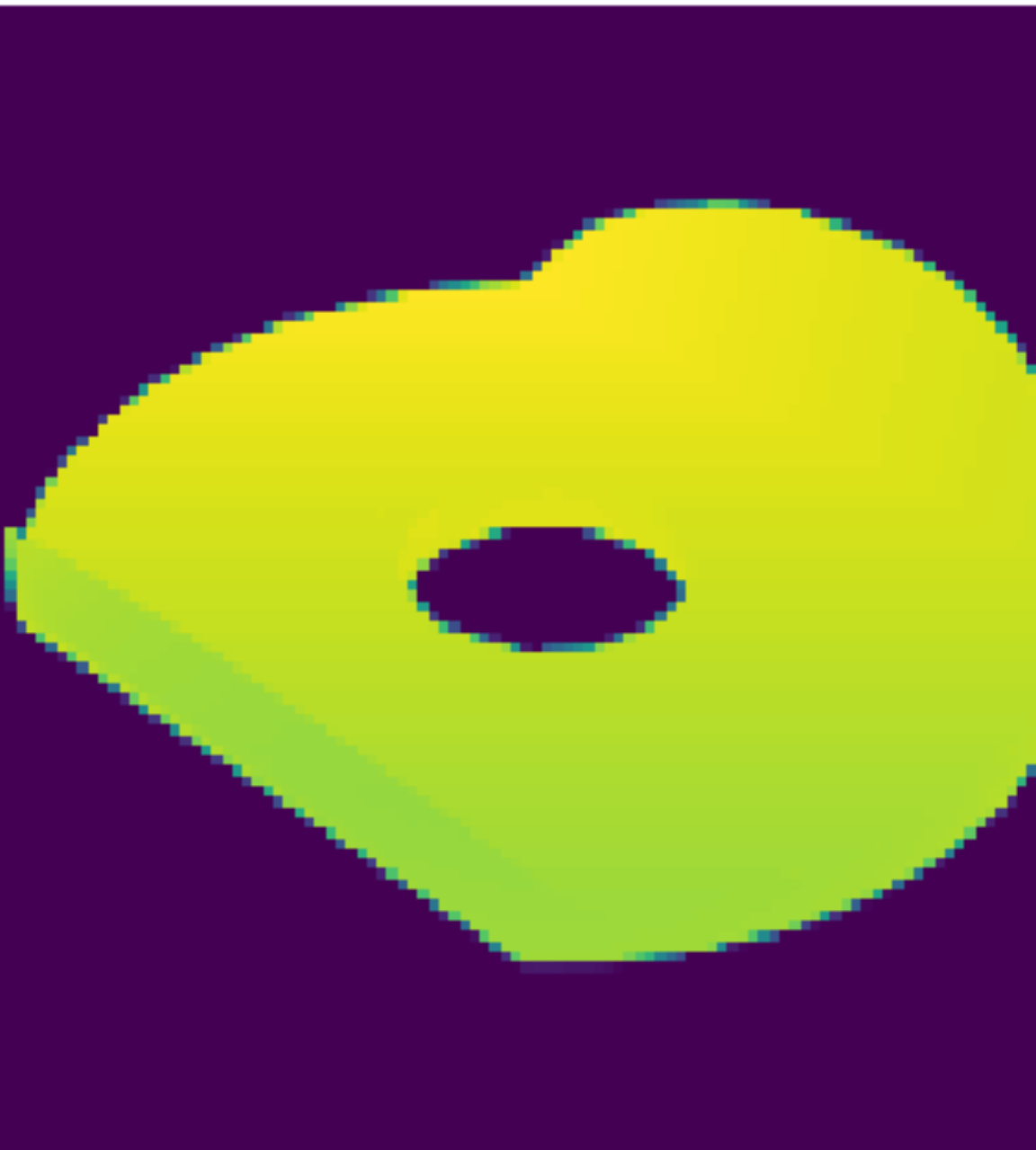}}  &
			\fcolorbox{green}{white}{\includegraphics[trim={9cm 4cm 9cm 4cm}, clip = true,width=0.12\linewidth]{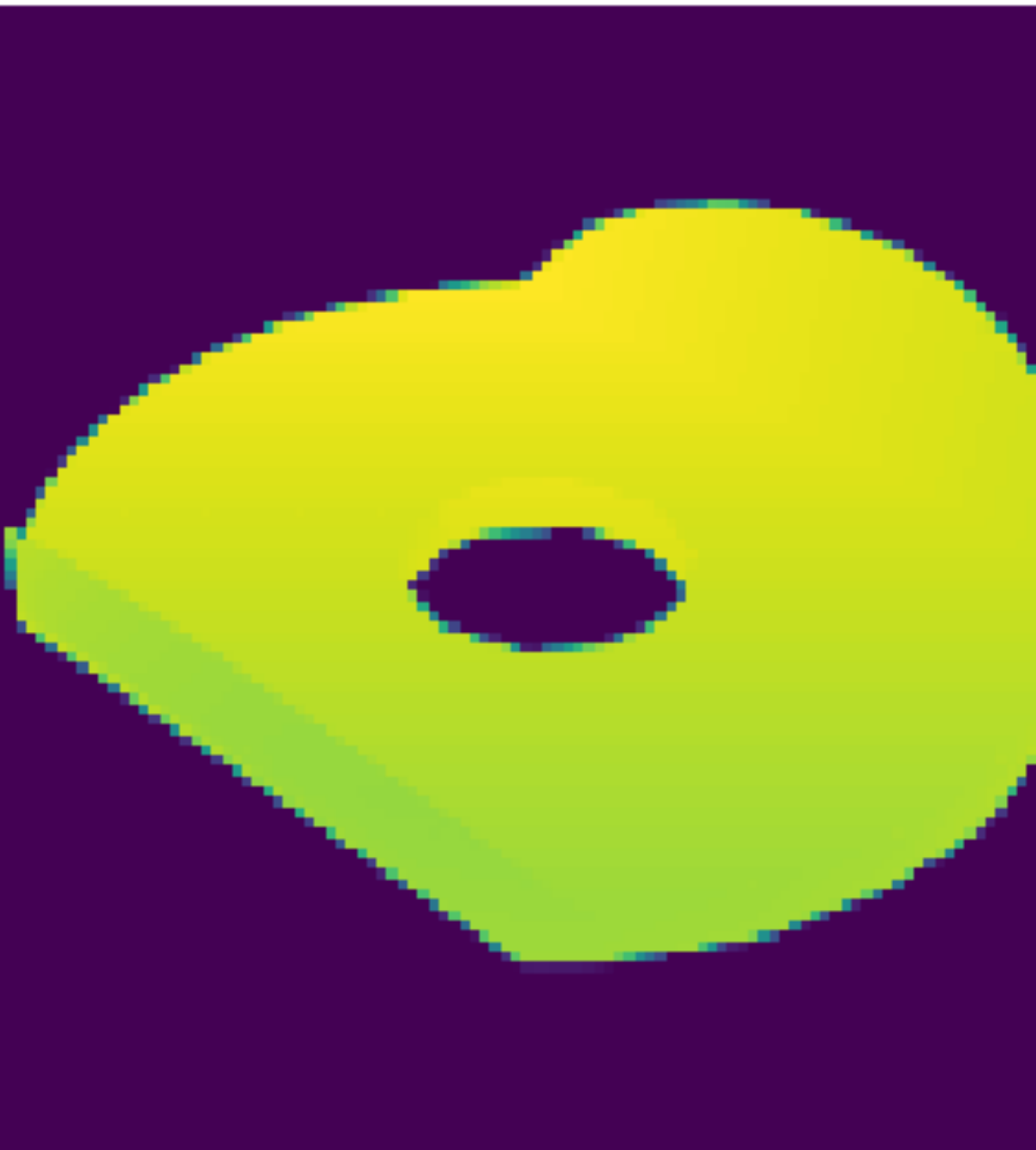}}  & \raisebox{2\height}{\LARGE 8.36$^{\circ}$ }  \\			\hline 			
			\includegraphics[trim={9cm 4cm 9cm 4cm}, clip = true,width=0.12\linewidth]{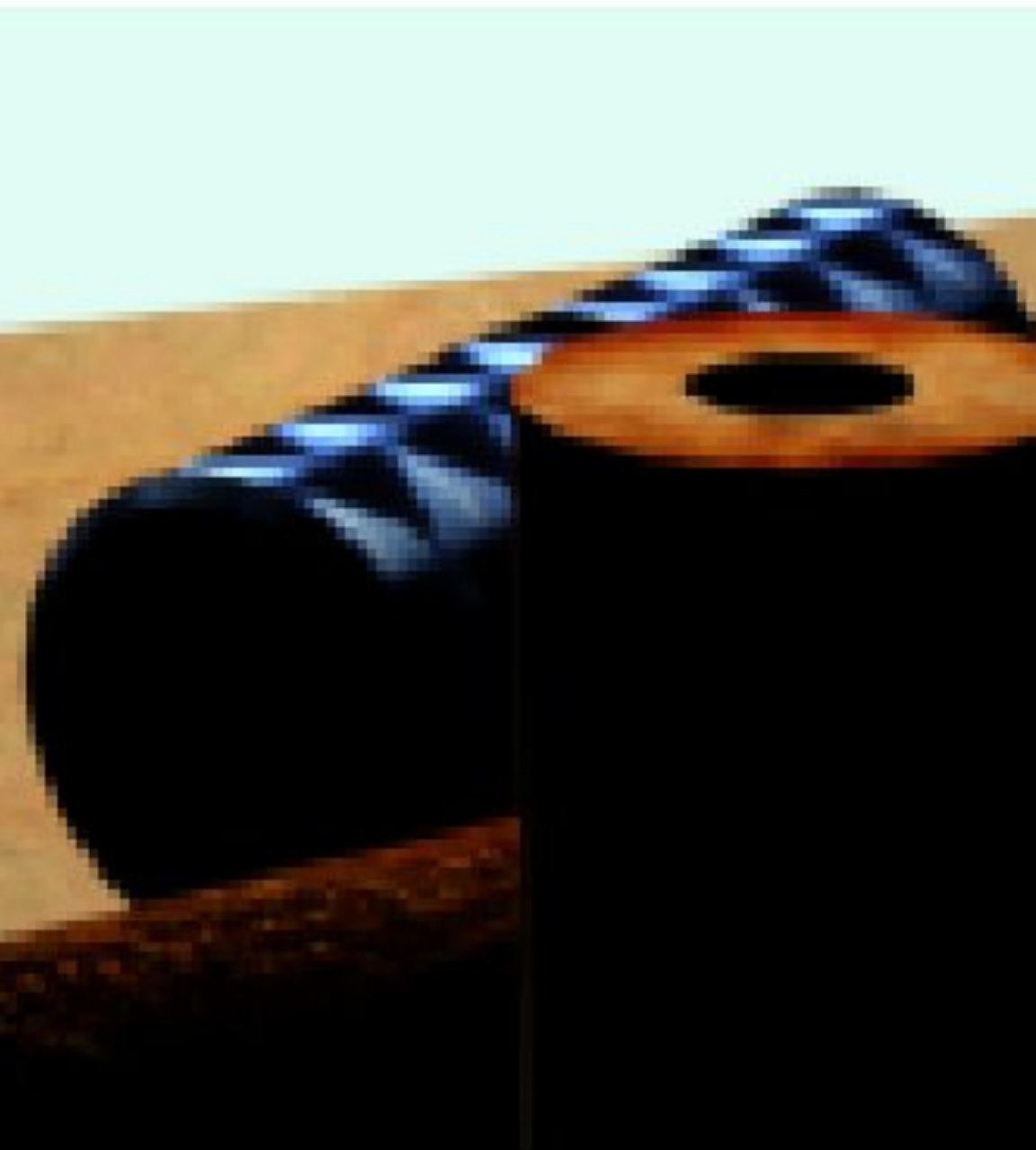}  &
			\includegraphics[trim={9cm 4cm 9cm 4cm}, clip = true,width=0.12\linewidth]{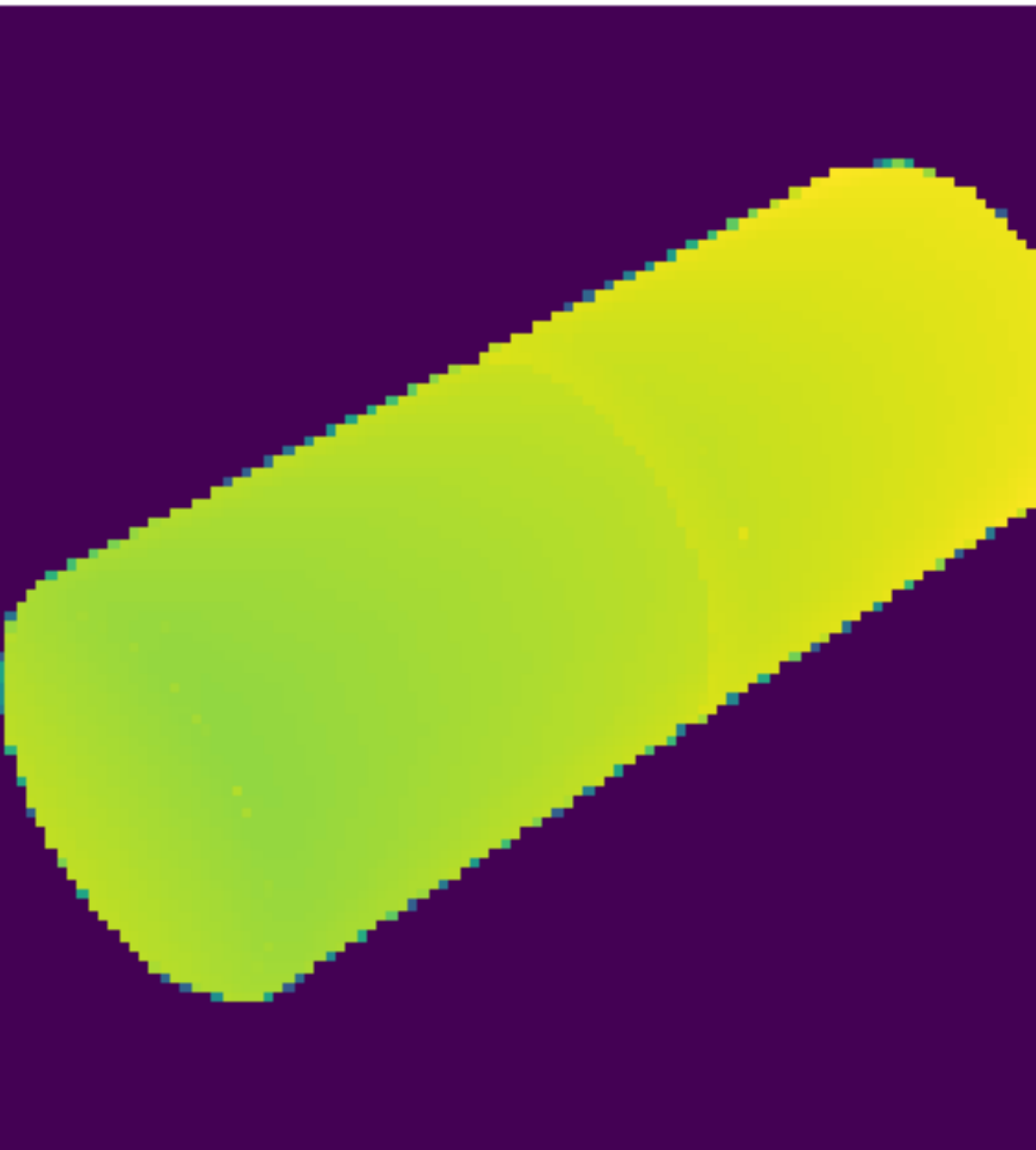}  &
			\fcolorbox{green}{white}{\includegraphics[trim={9cm 4cm 9cm 4cm}, clip = true,width=0.12\linewidth]{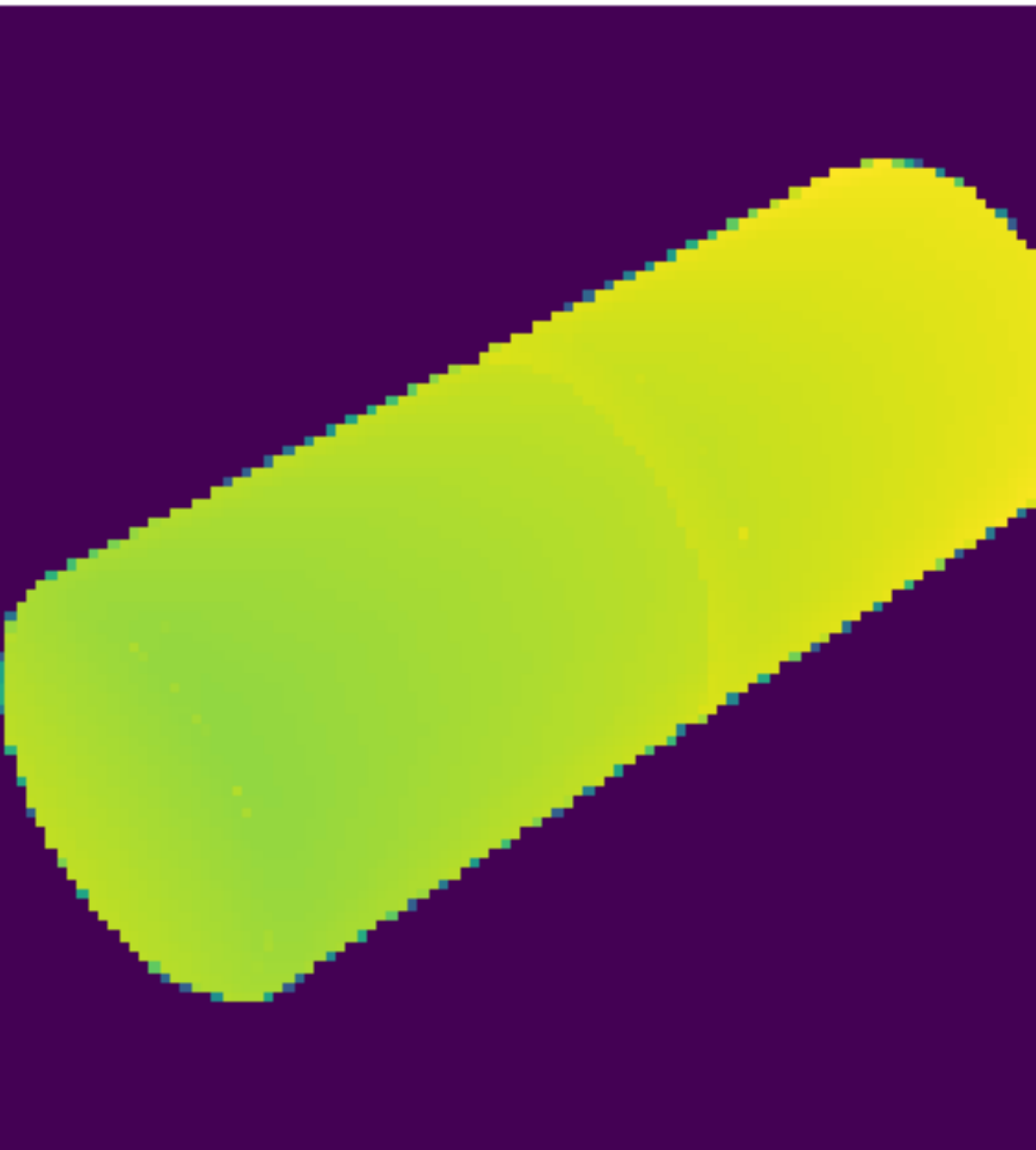} } &
			\fcolorbox{green}{white}{\includegraphics[trim={9cm 4cm 9cm 4cm}, clip = true,width=0.12\linewidth]{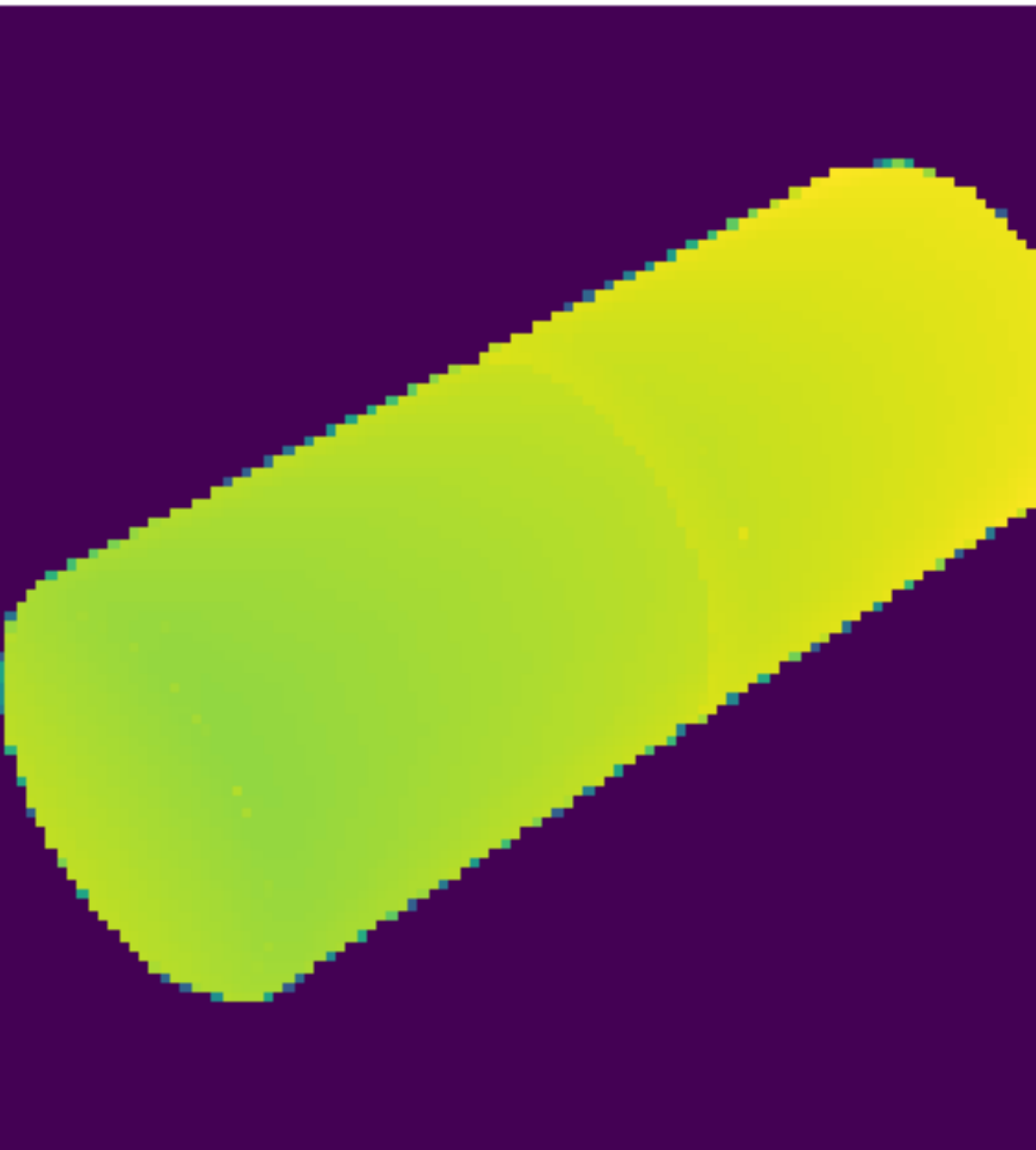} } &
			\fcolorbox{green}{white}{\includegraphics[trim={9cm 4cm 9cm 4cm}, clip = true,width=0.12\linewidth]{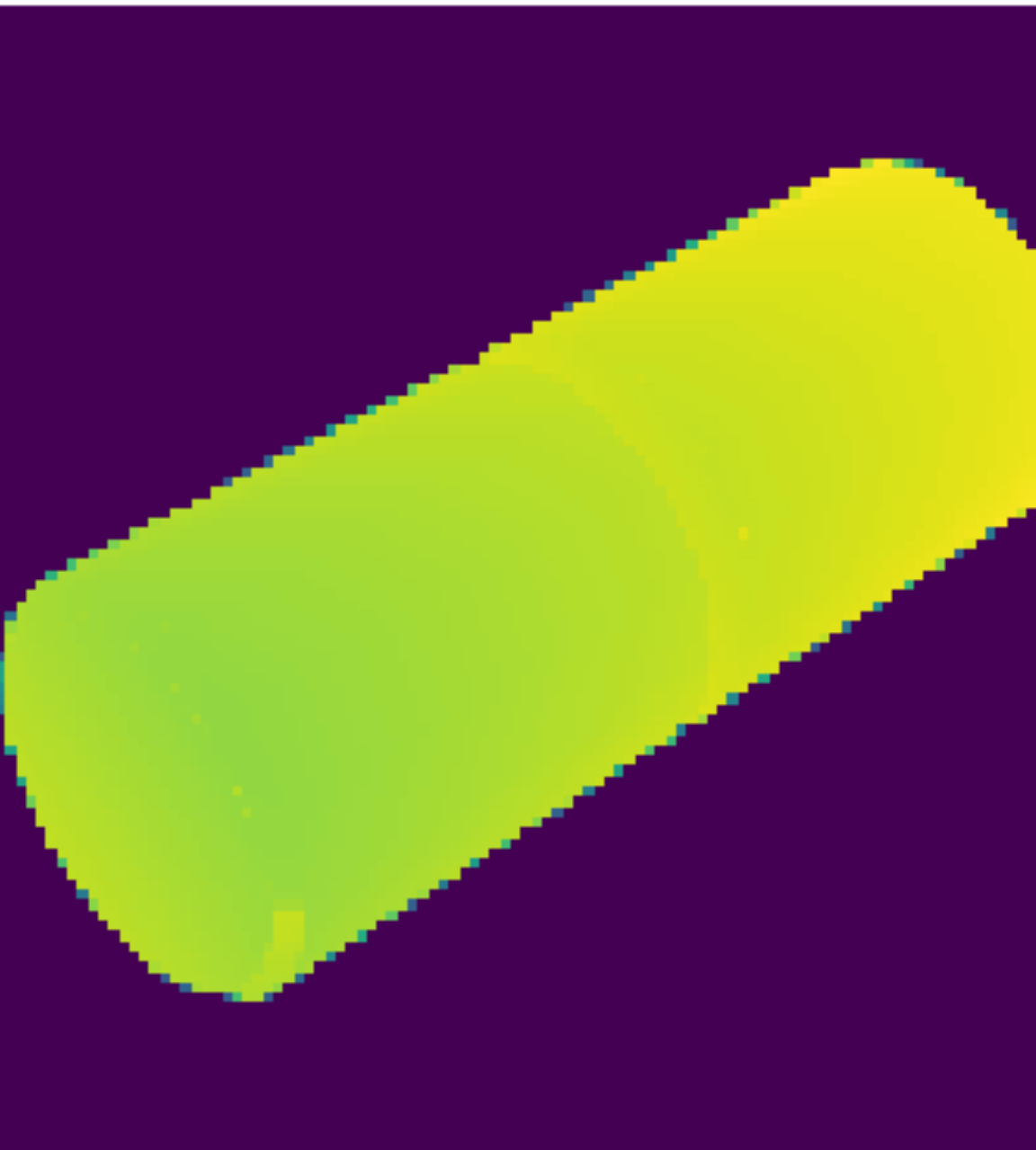} } &
			\fcolorbox{green}{white}{\includegraphics[trim={9cm 4cm 9cm 4cm}, clip = true,width=0.12\linewidth]{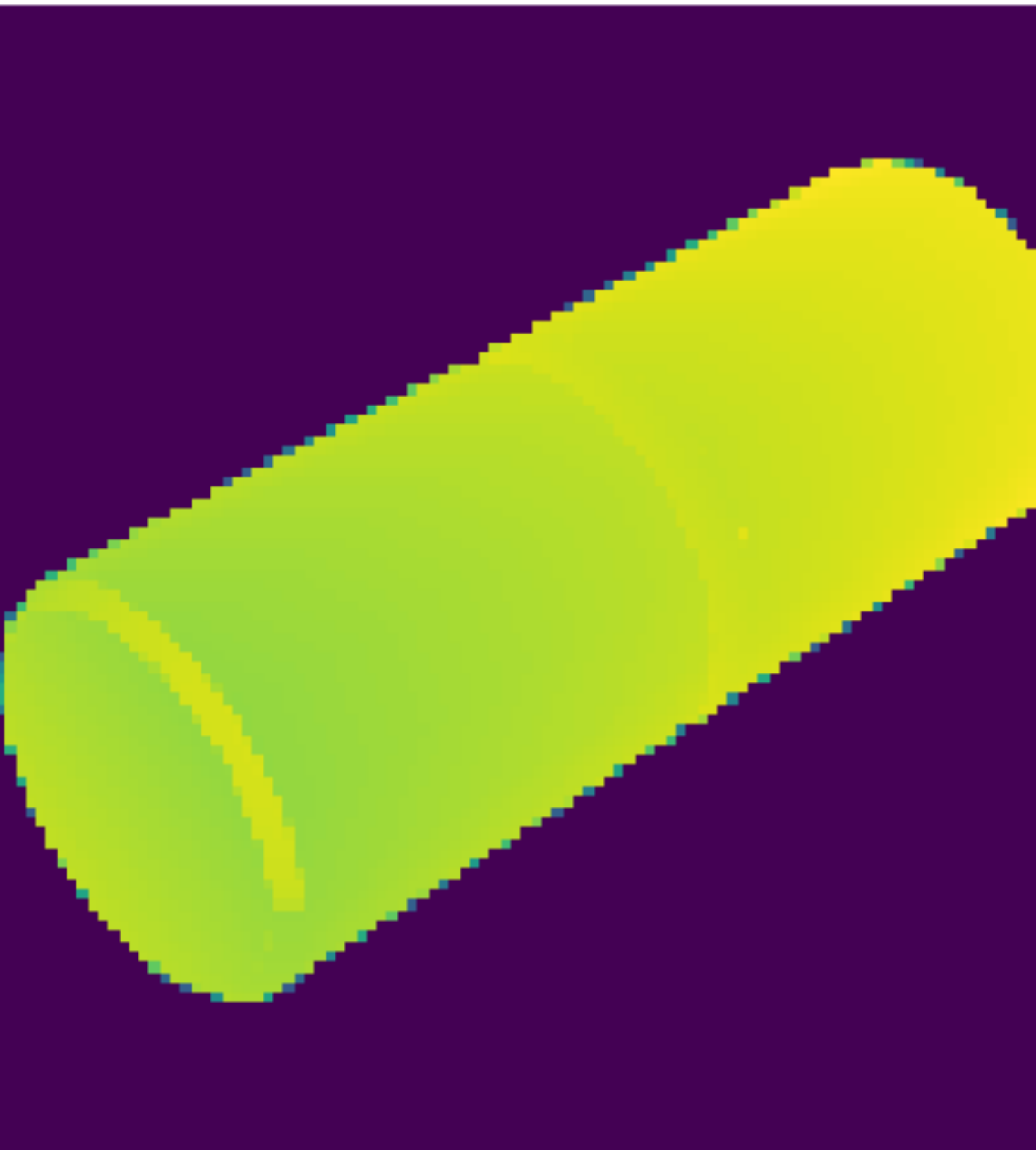} } & \raisebox{2\height}{\LARGE 8.63$^{\circ}$ } \\			\hline 
			\includegraphics[trim={9cm 4cm 9cm 4cm}, clip = true,width=0.12\linewidth]{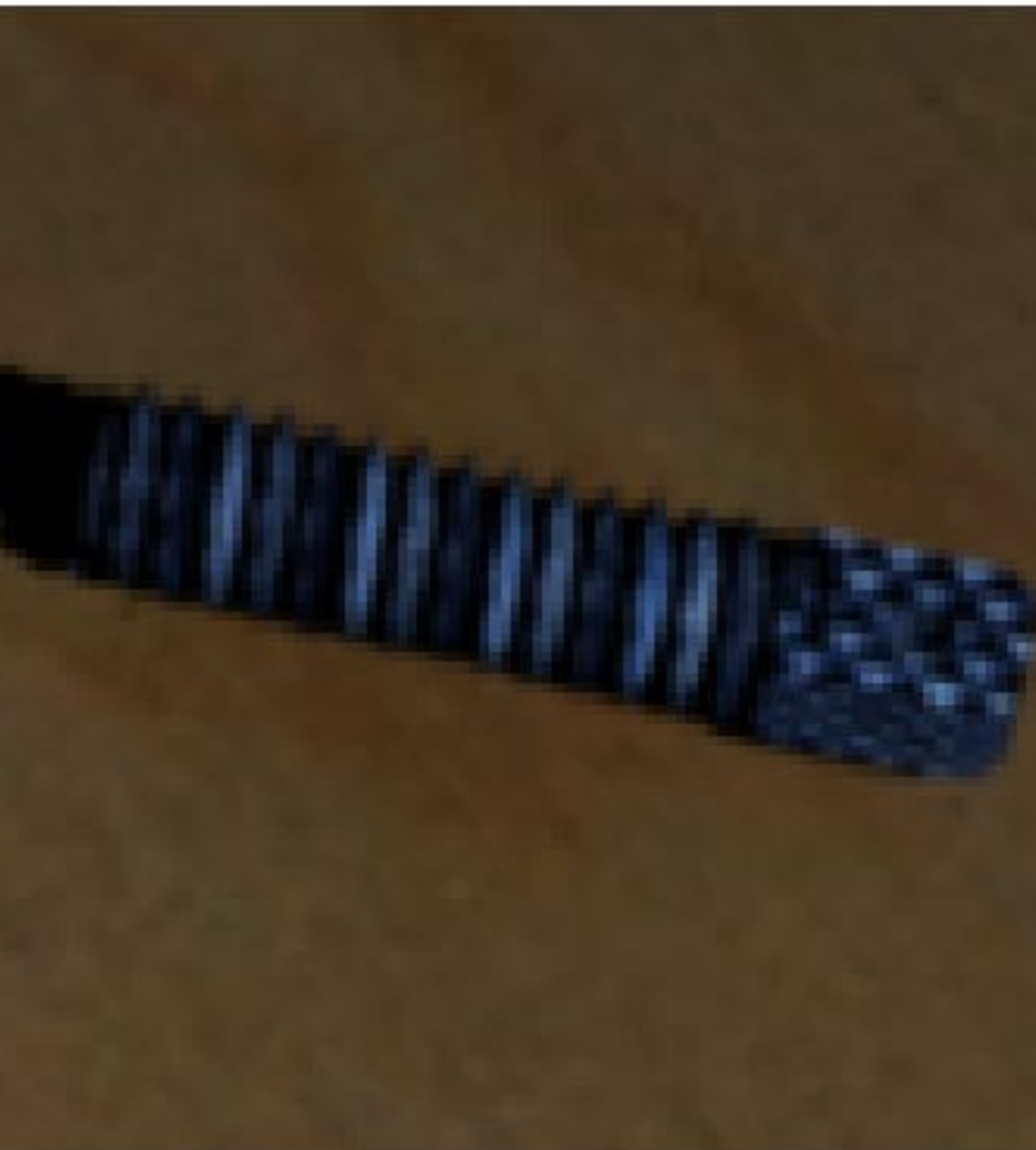} &
			\includegraphics[trim={9cm 4cm 9cm 4cm}, clip = true,width=0.12\linewidth]{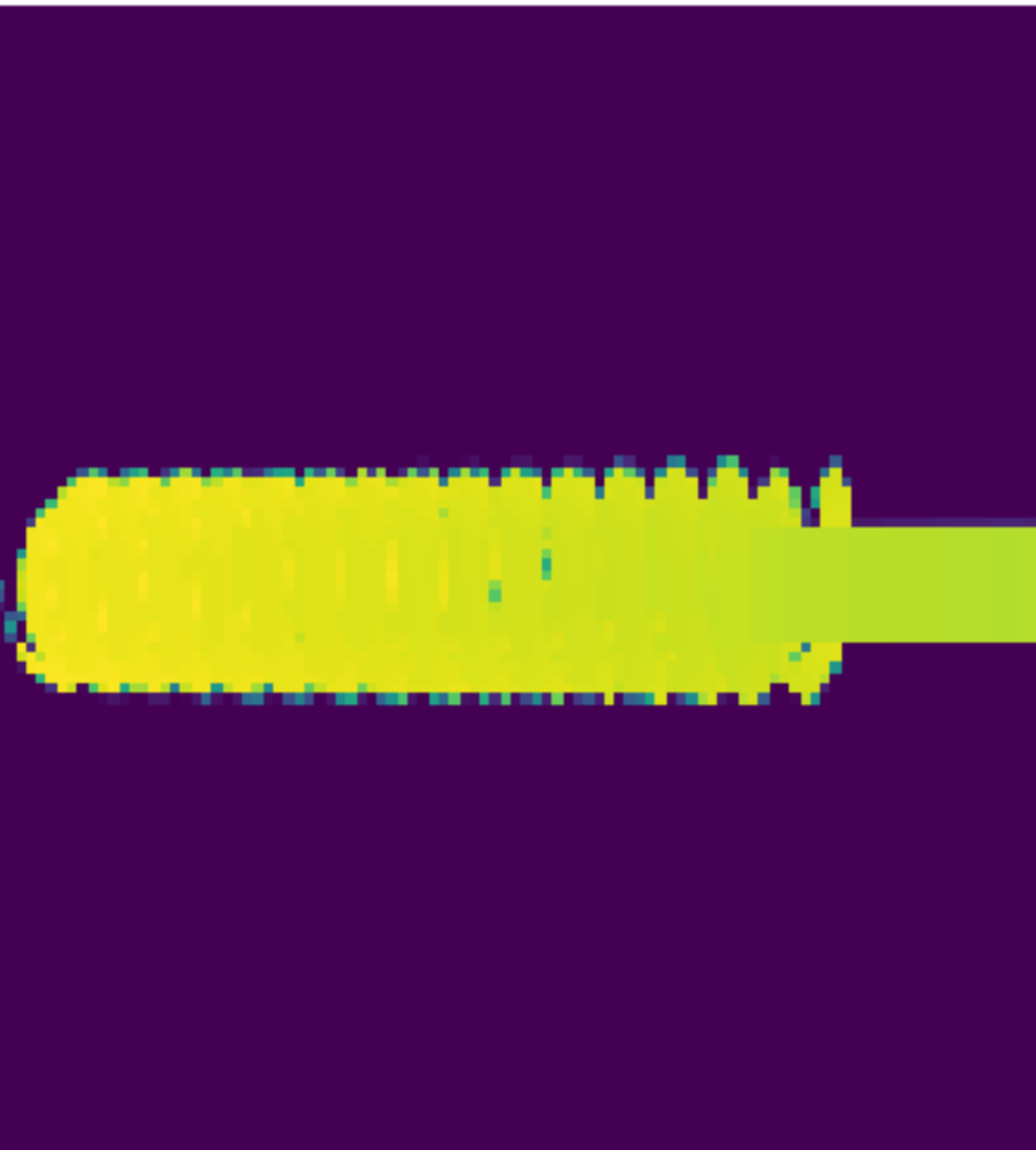} &
			\includegraphics[trim={9cm 4cm 9cm 4cm}, clip = true,width=0.12\linewidth]{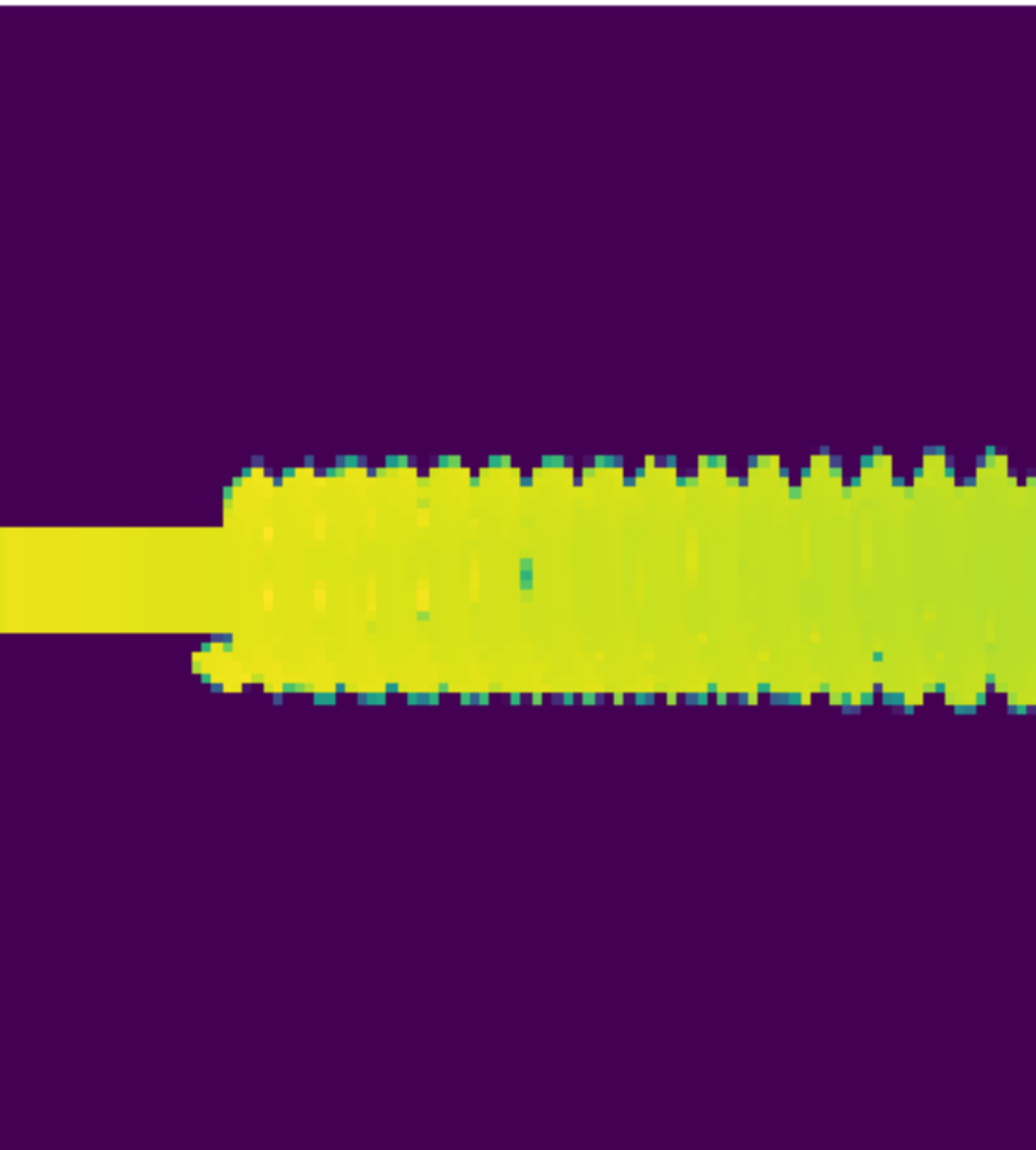} &
			\includegraphics[trim={9cm 4cm 9cm 4cm}, clip = true,width=0.12\linewidth]{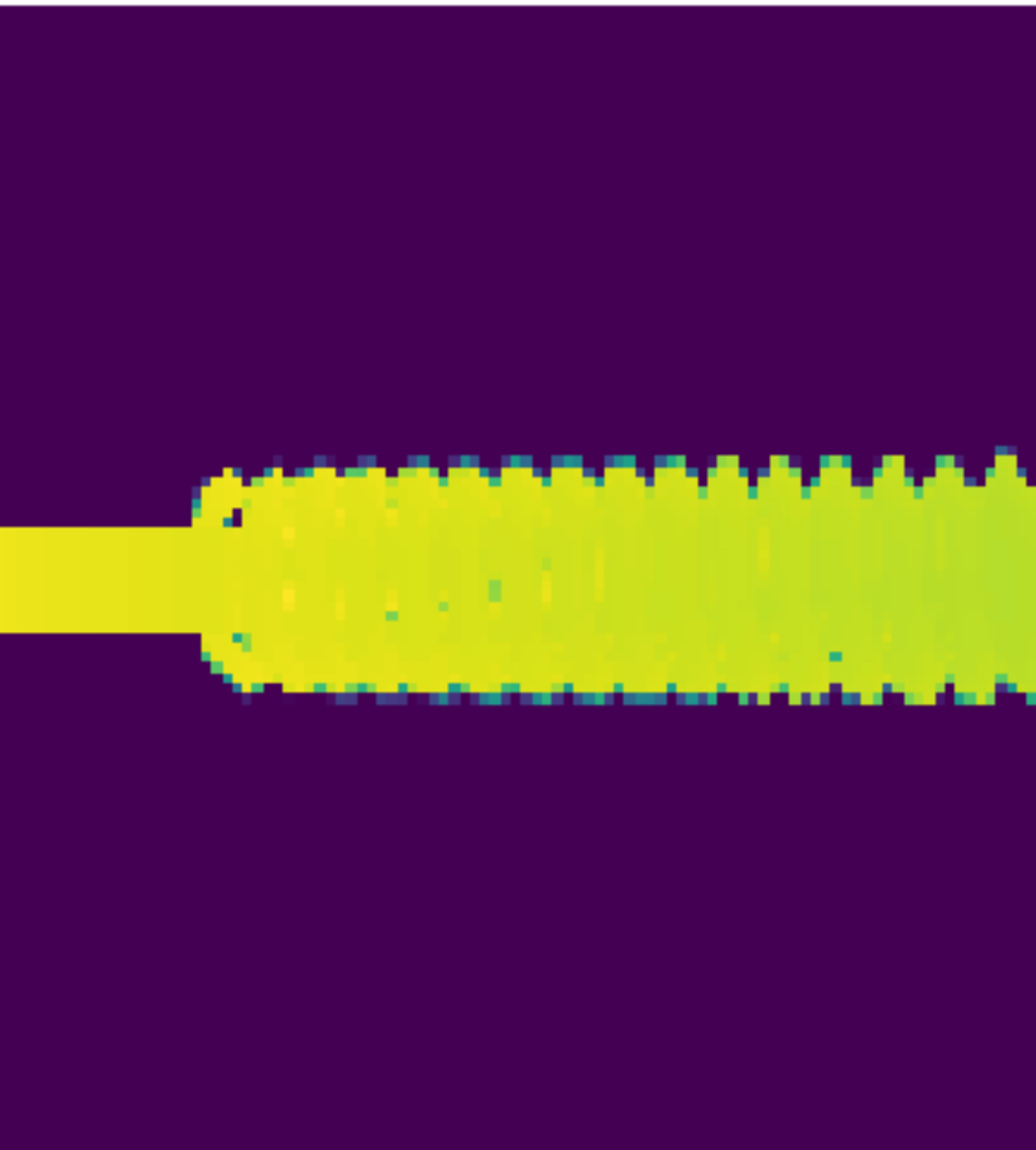} &
			\fcolorbox{green}{white}{\includegraphics[trim={9cm 4cm 9cm 4cm}, clip = true,width=0.12\linewidth]{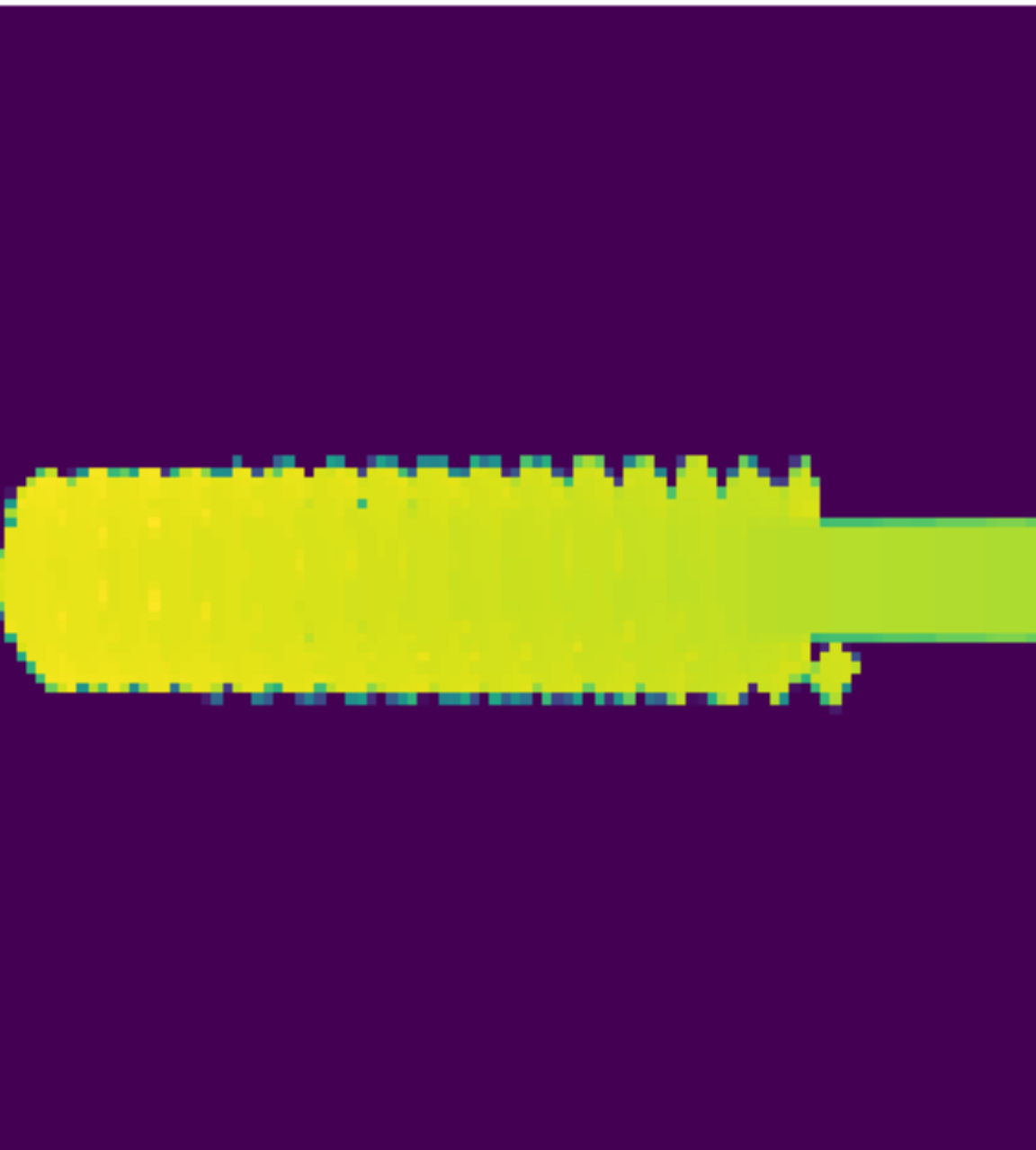}} &
			\fcolorbox{green}{white}{\includegraphics[trim={9cm 4cm 9cm 4cm}, clip = true,width=0.12\linewidth]{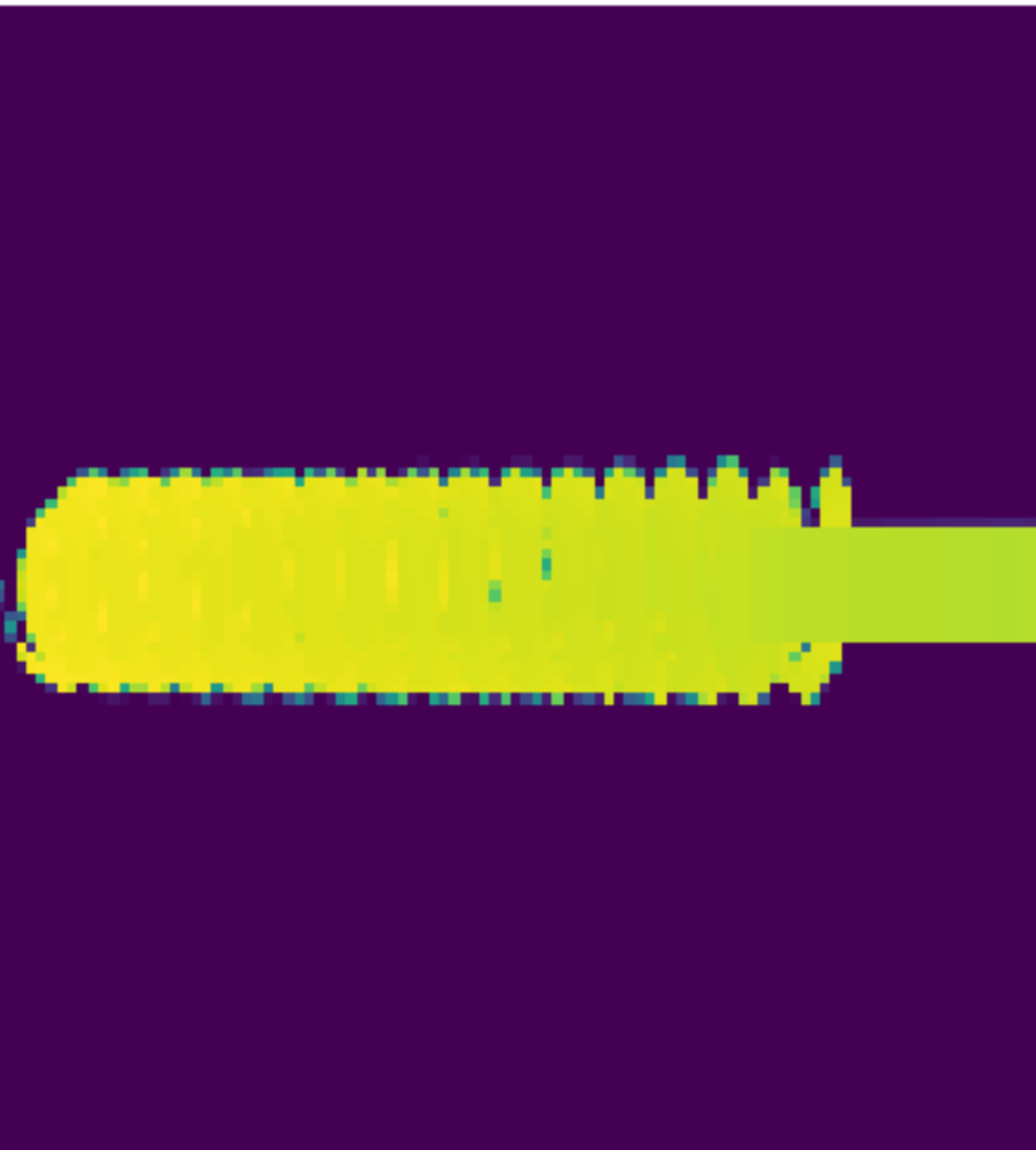}} & \raisebox{2\height}{\LARGE 8.64$^{\circ}$ } \\ \hline
			\includegraphics[trim={9cm 4cm 9cm 4cm}, clip = true,width=0.12\linewidth]{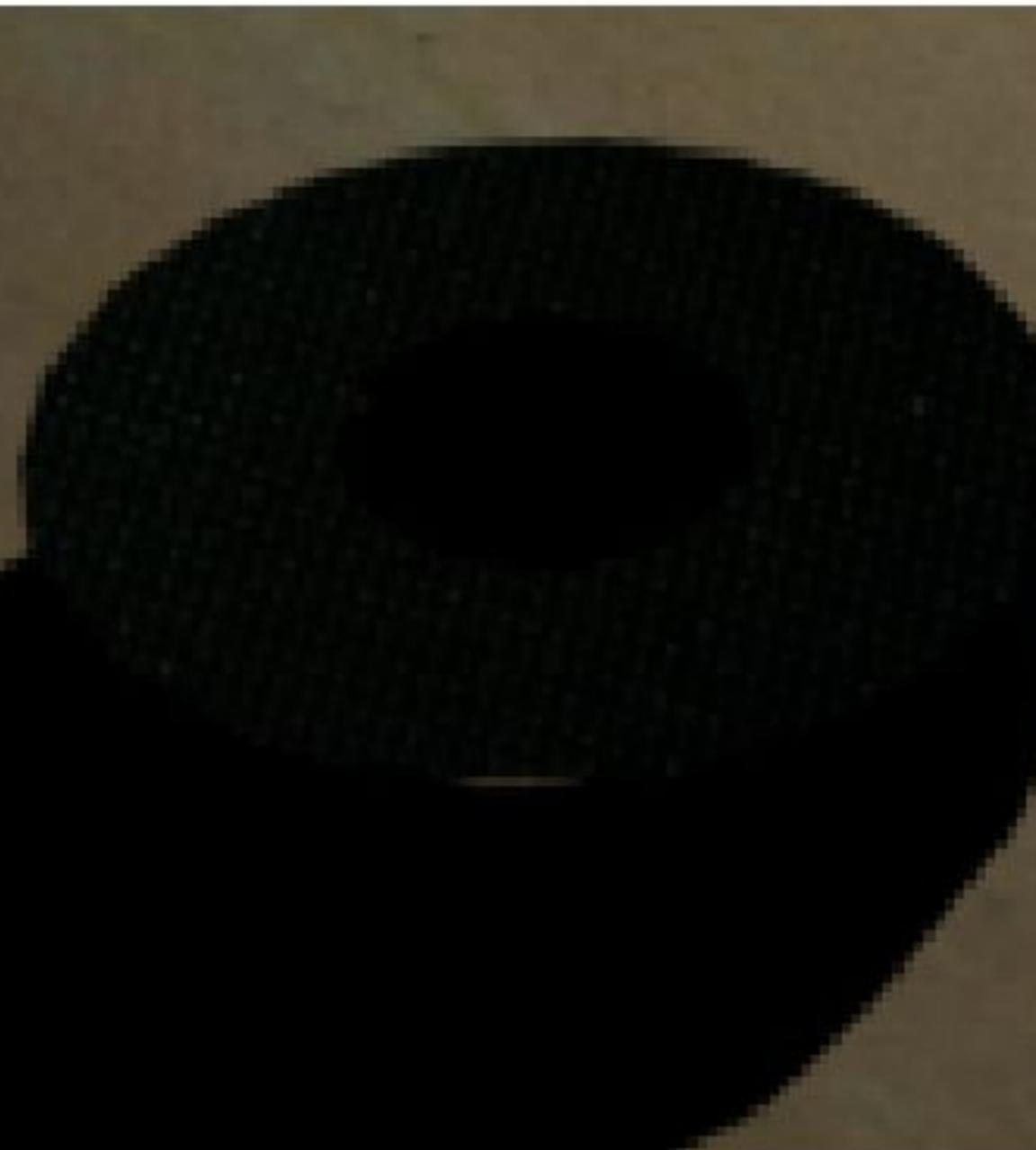} &
			\includegraphics[trim={9cm 4cm 9cm 4cm}, clip = true,width=0.12\linewidth]{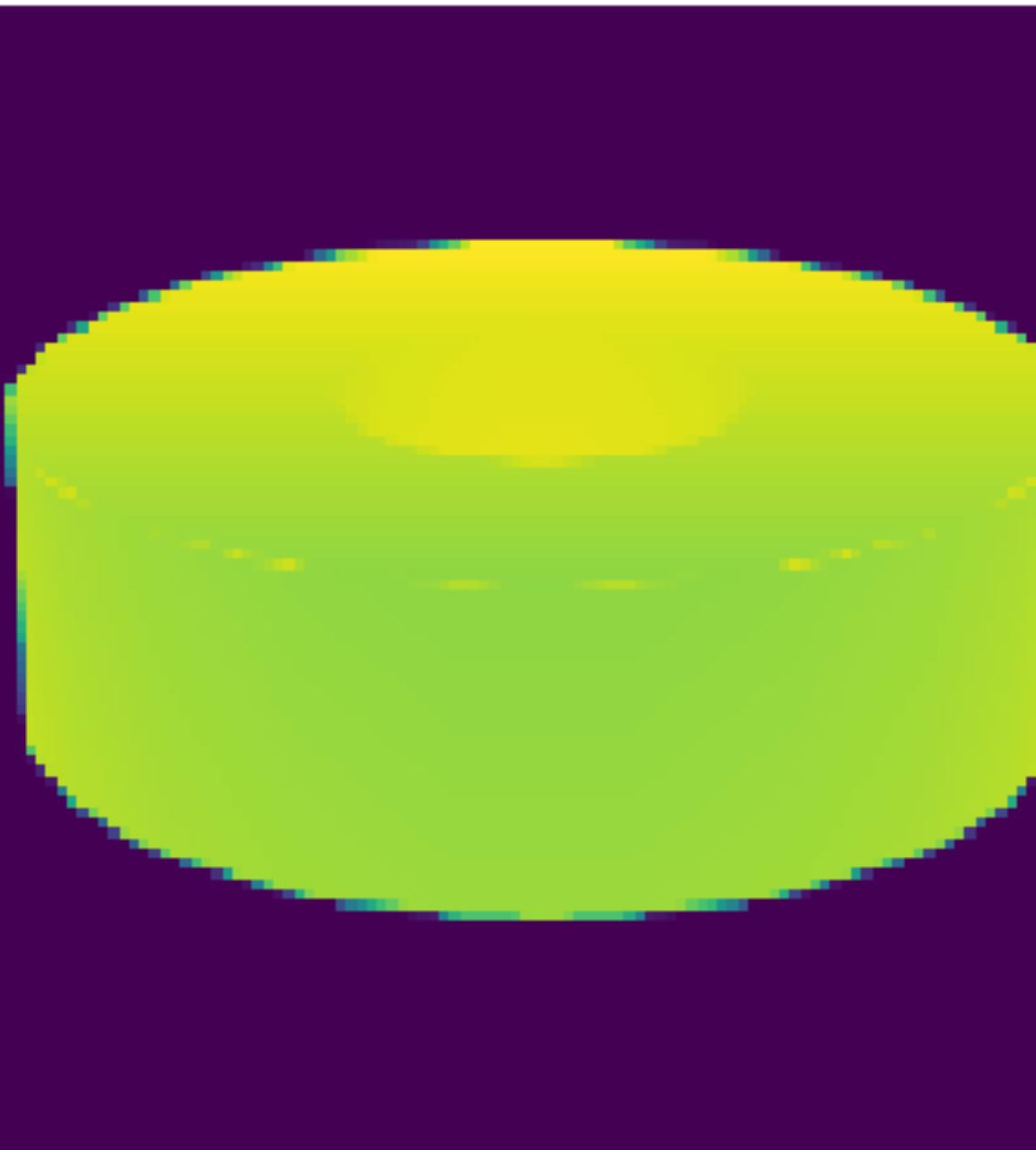} &
			\fcolorbox{green}{white}{\includegraphics[trim={9cm 4cm 9cm 4cm}, clip = true,width=0.12\linewidth]{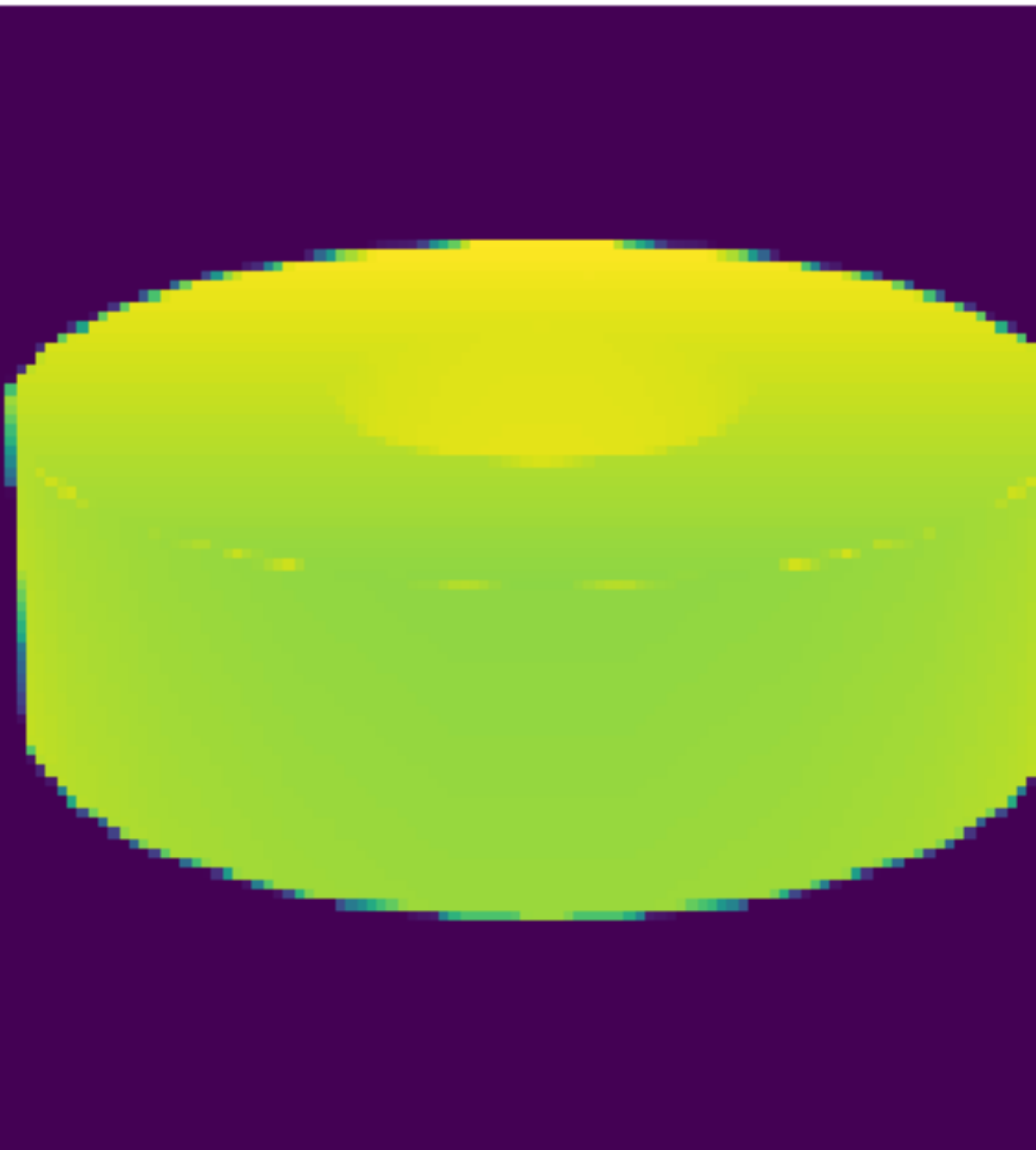}} &
			\fcolorbox{green}{white}{\includegraphics[trim={9cm 4cm 9cm 4cm}, clip = true,width=0.12\linewidth]{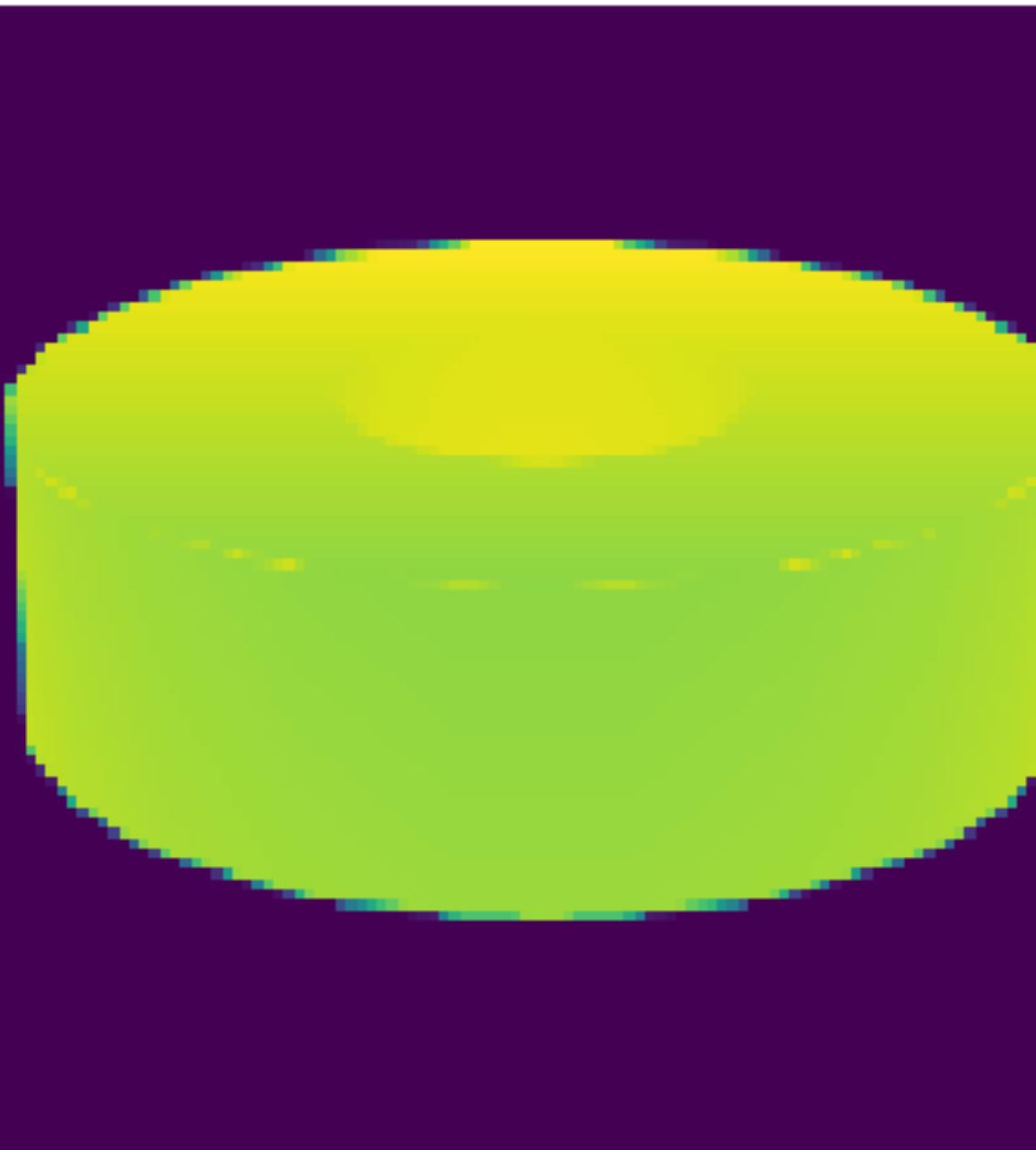}} &
			\fcolorbox{green}{white}{\includegraphics[trim={9cm 4cm 9cm 4cm}, clip = true,width=0.12\linewidth]{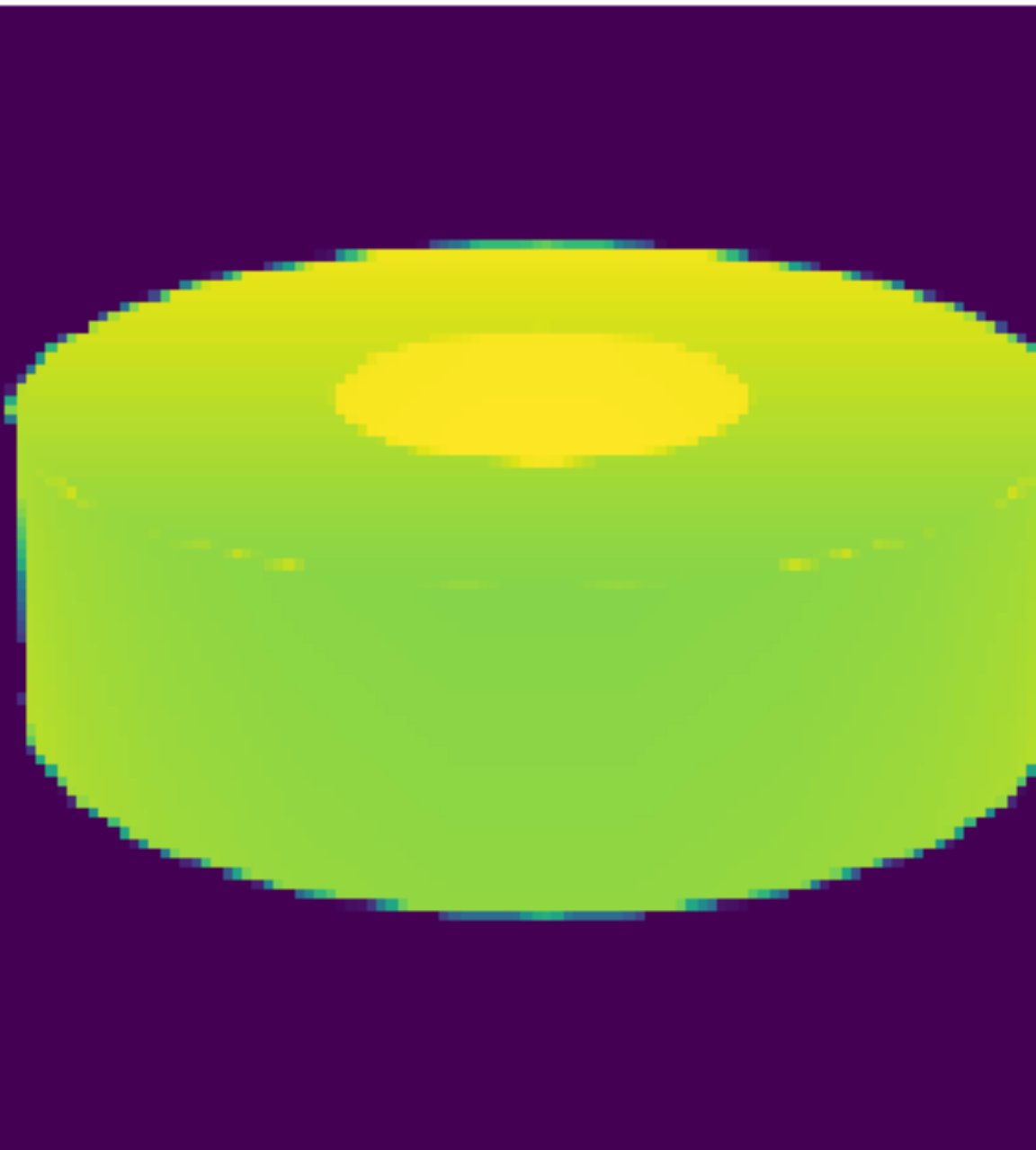}} &
			\fcolorbox{green}{white}{\includegraphics[trim={9cm 4cm 9cm 4cm}, clip = true,width=0.12\linewidth]{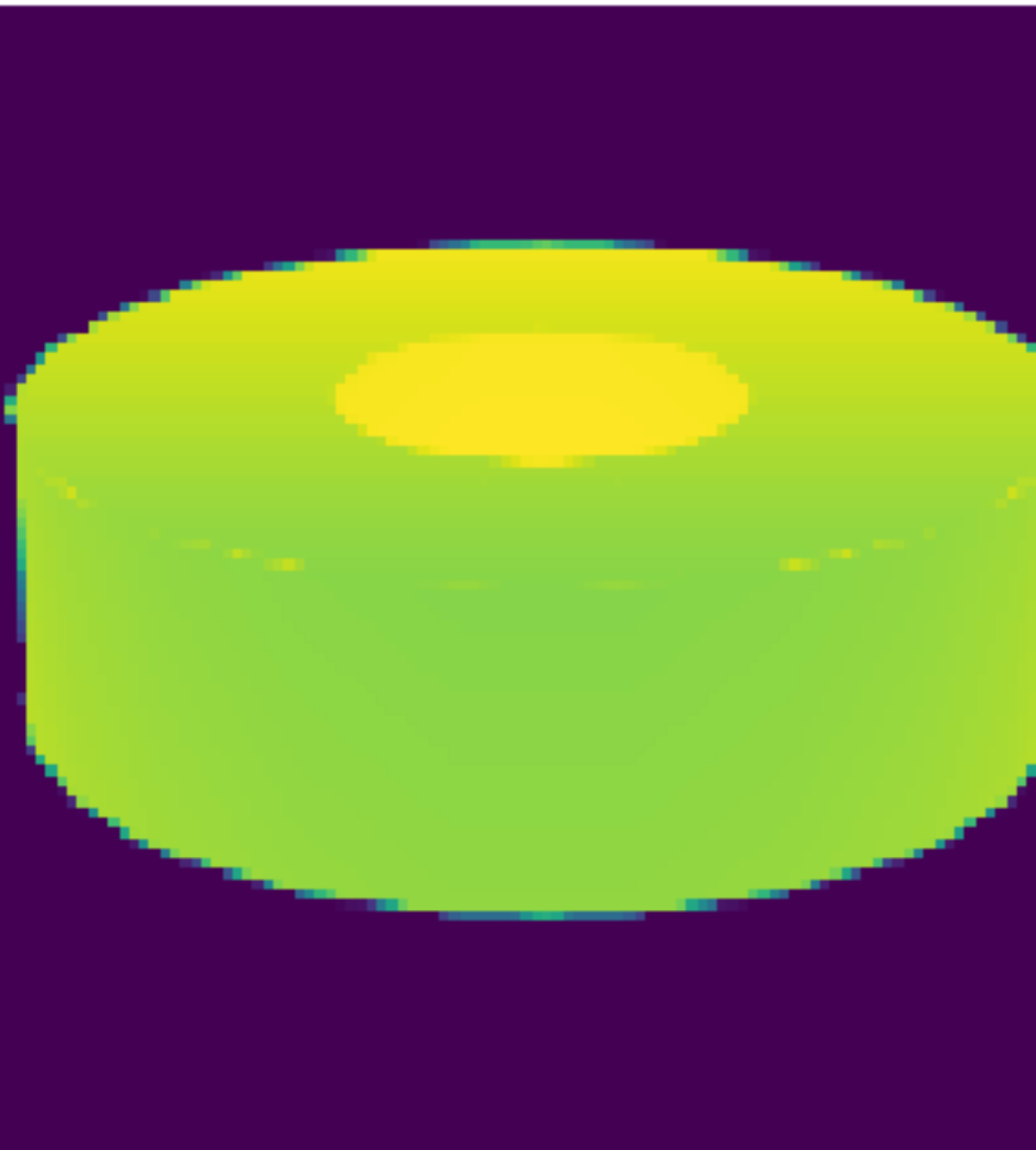}} & \raisebox{2\height}{\LARGE 9.21$^{\circ}$ } \\					\hline 
	\end{tabular} } 
	\centering
	\caption{{\bf Successful Results}: Qualitative results for pose estimation on the \textbf{synthetic} dataset.} % The object in the scene is shown in the first column. The second column shows the nearest discretizated view. When the top four predicted views shown in the third column match the nearest view, they are indicated in a \textbf{green} box. Otherwise, an \textbf{orange} box shows the view that is nearest among the predicted. The last column shows the distance of this view. Given objects with symmetry, there can be more than one best view.}
	\label{fig:synth_result_3}
\end{figure*}

\figref{fig:epson1034_result_1} depicts successful results of our approach on the {\bf real dataset}. We also show the successful results on the validation set of the {\bf synthetic dataset} in~\figref{fig:synth_result_1},~\figref{fig:synth_result_2} and~\figref{fig:synth_result_3}. For the input image shown in the left column, the depth image of ground truth (GT) coarse pose, followed by top-4 predictions from our approach. In the last column, we mention the distance of the best match from the top-4 predictions to the ground truth, $ d_\text{rot, best}^{sym} $.  The results have been sorted in decreasing order of performance of $  d_\text{rot, best}^{sym} $.  We also draw a green colored box around the best match among the top-4 predictions. % When the best performing prediction does not match with the GT, we draw a red box around it.

\begin{figure*} % \ContinuedFloat
	\setlength\tabcolsep{4pt}
	\hspace{-1.2cm}\parbox{\linewidth}{
		\begin{tabular}{c|c|c|cccc|c|}
			\cline{2-8}
			& {\bf Input } & {\bf GT } & \multicolumn{4}{c|}{\bf Top-4 Predictions} & {\bf  $ d_\text{rot, best}^{sym} $} \\
			\cline{2-8} 
 		\raisebox{3.25\height}{(a)} & \includegraphics[trim={9cm 4cm 9cm 4cm}, clip = true,width=0.12\linewidth]{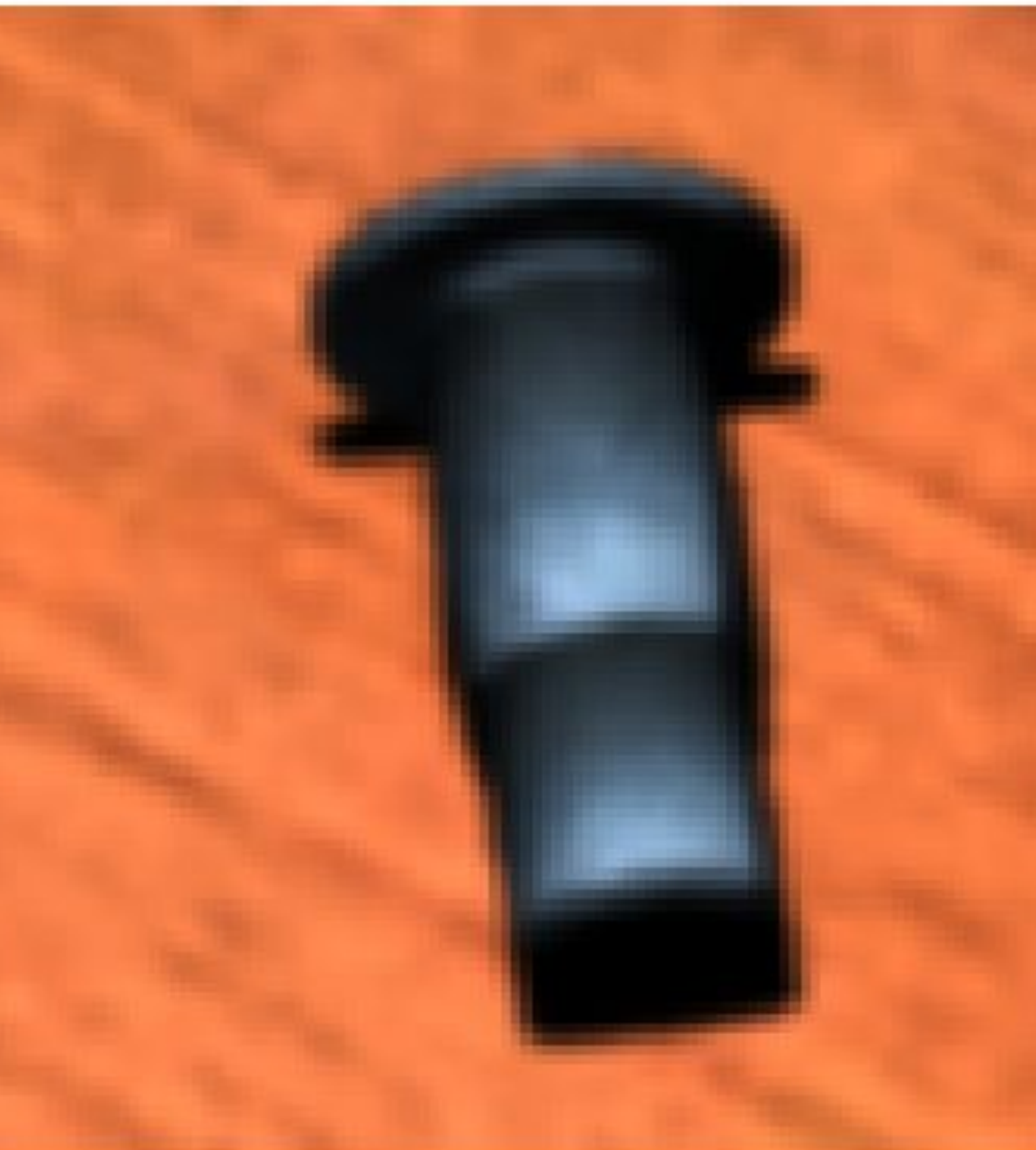} &
 		\includegraphics[trim={9cm 4cm 9cm 4cm}, clip = true,width=0.12\linewidth]{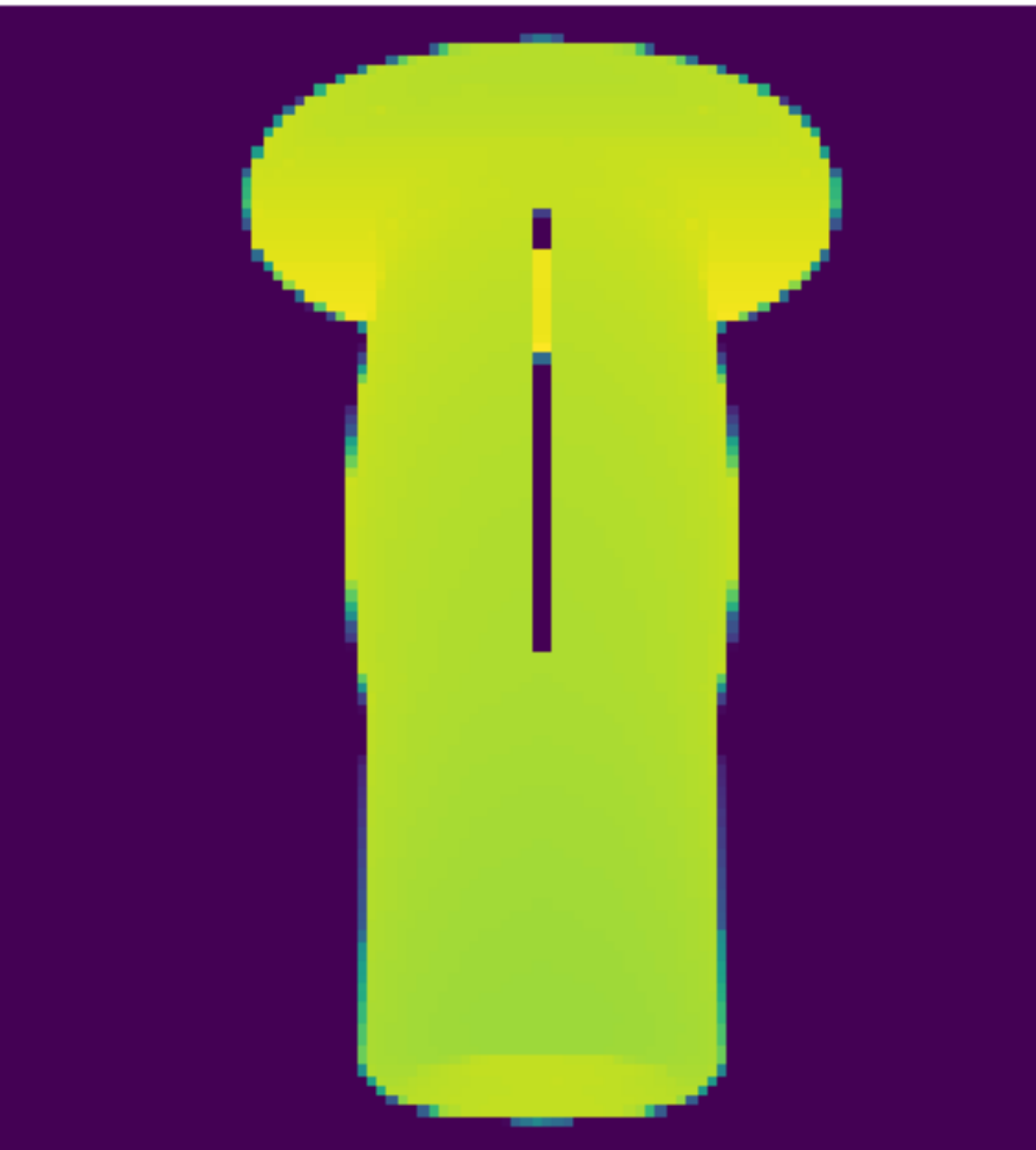} &
 		\includegraphics[trim={9cm 4cm 9cm 4cm}, clip = true,width=0.12\linewidth]{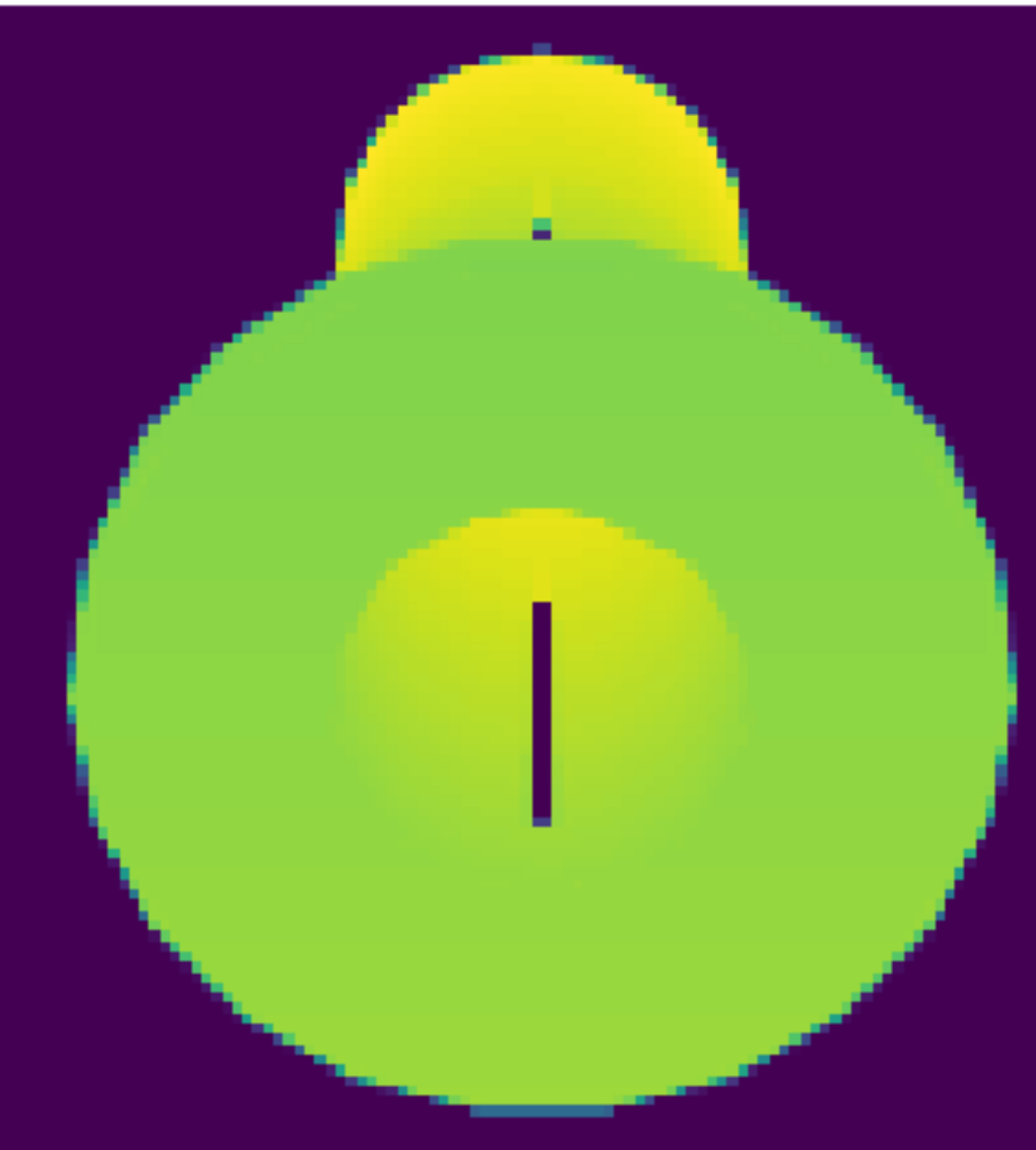} &
 		\includegraphics[trim={9cm 4cm 9cm 4cm}, clip = true,width=0.12\linewidth]{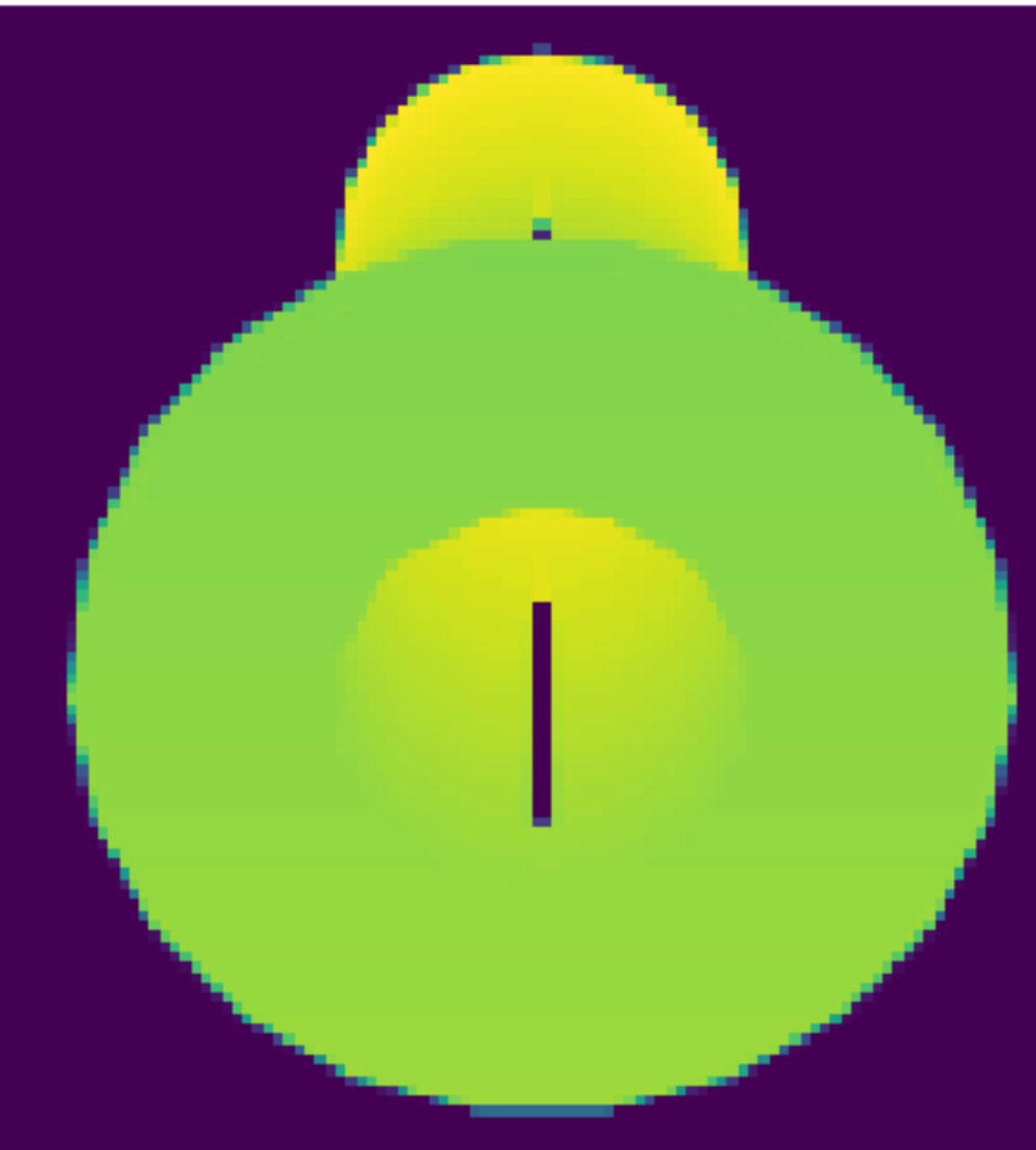} &
 		\fcolorbox{orange}{white}{\includegraphics[trim={9cm 4cm 9cm 4cm}, clip = true,width=0.12\linewidth]{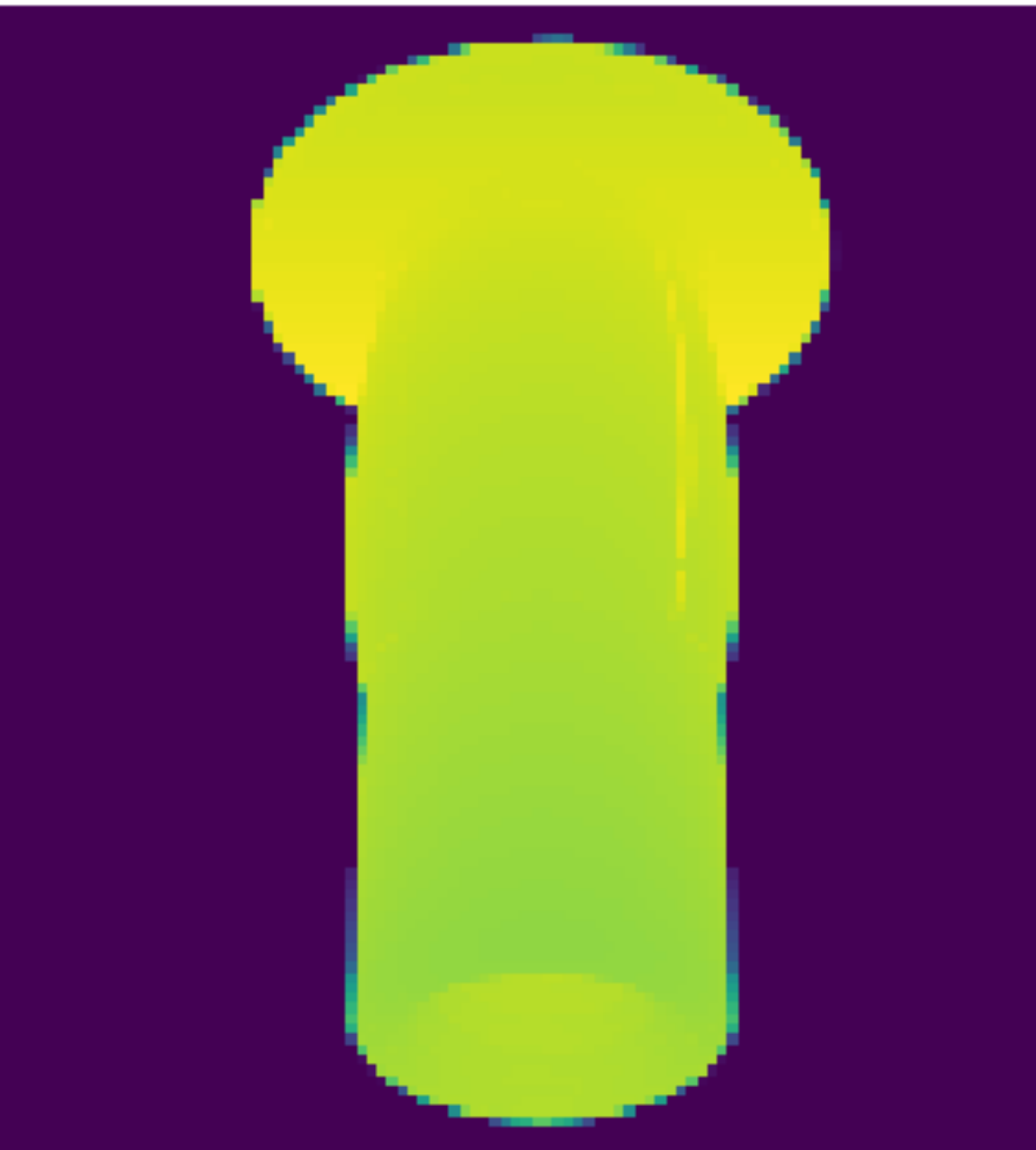}} &
 		\fcolorbox{orange}{white}{\includegraphics[trim={9cm 4cm 9cm 4cm}, clip = true,width=0.12\linewidth]{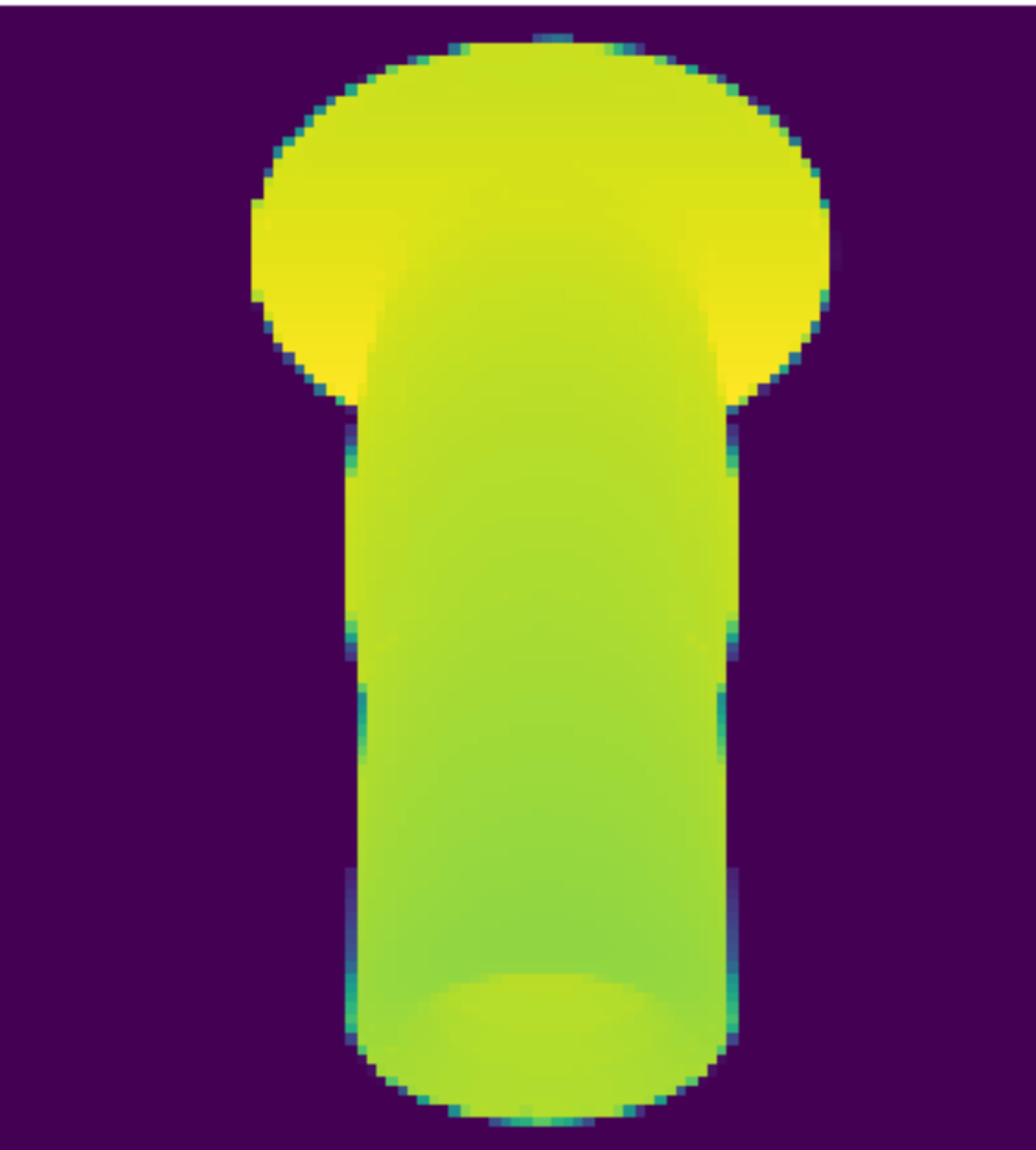}} & \raisebox{2\height}{\Large  8.61$^{\circ}$ } \\
 		\cline{2-8}
		\raisebox{3.25\height}{(b)} & \includegraphics[trim={9cm 4cm 9cm 4cm}, clip = true,width=0.12\linewidth]{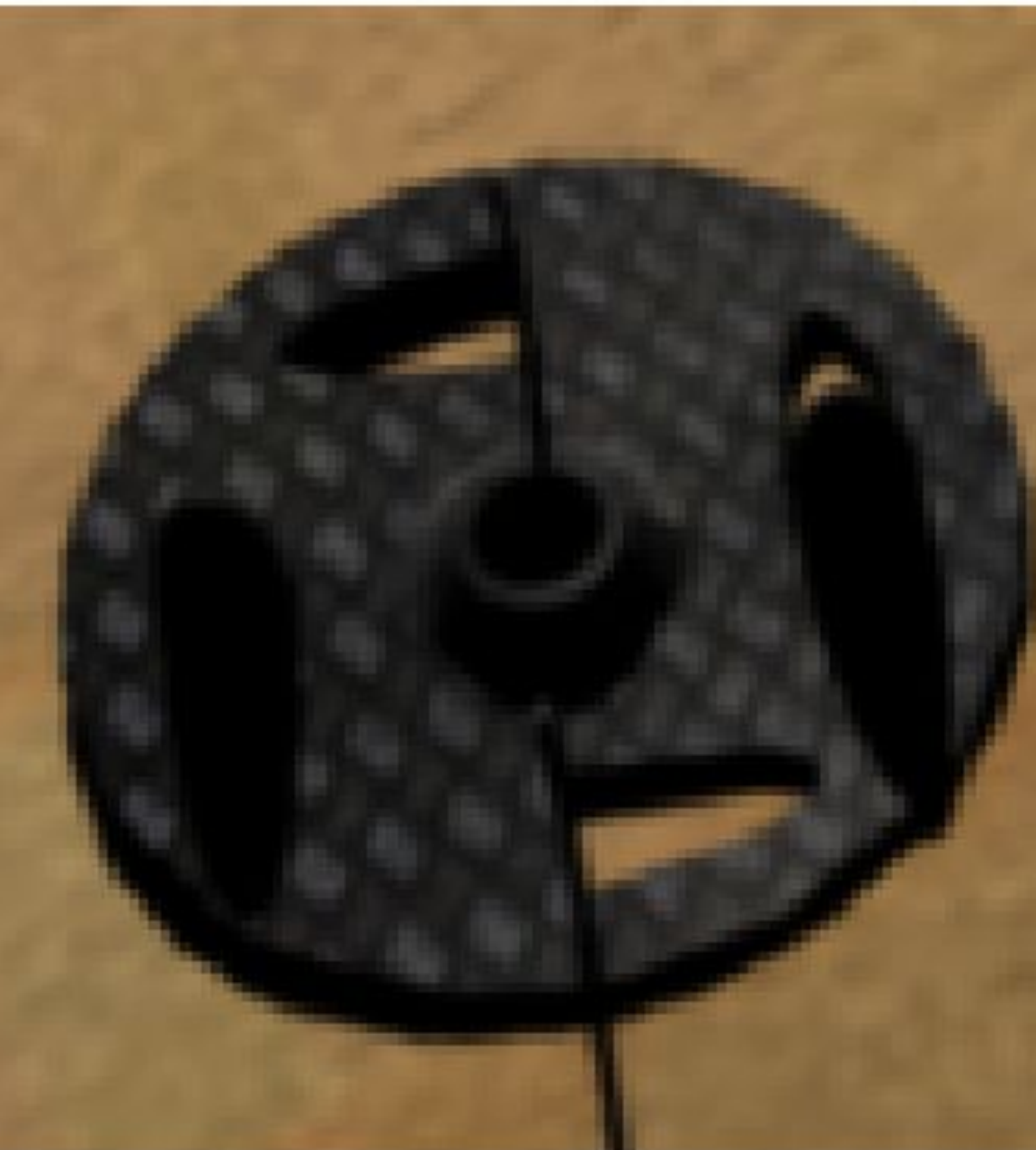}  &
 		\includegraphics[trim={9cm 4cm 9cm 4cm}, clip = true,width=0.12\linewidth]{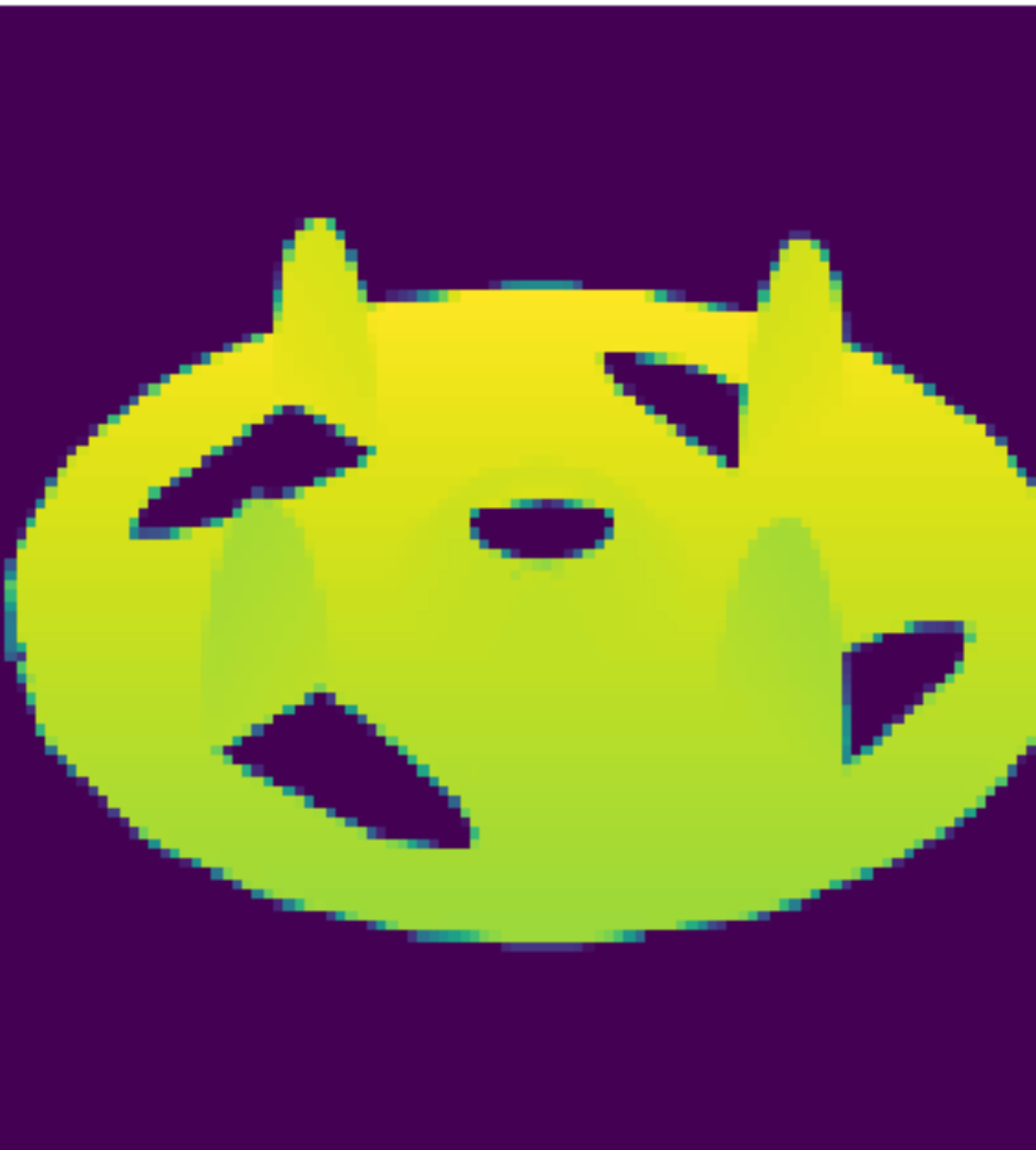}  &
 		\fcolorbox{orange}{white}{\includegraphics[trim={9cm 4cm 9cm 4cm}, clip = true,width=0.12\linewidth]{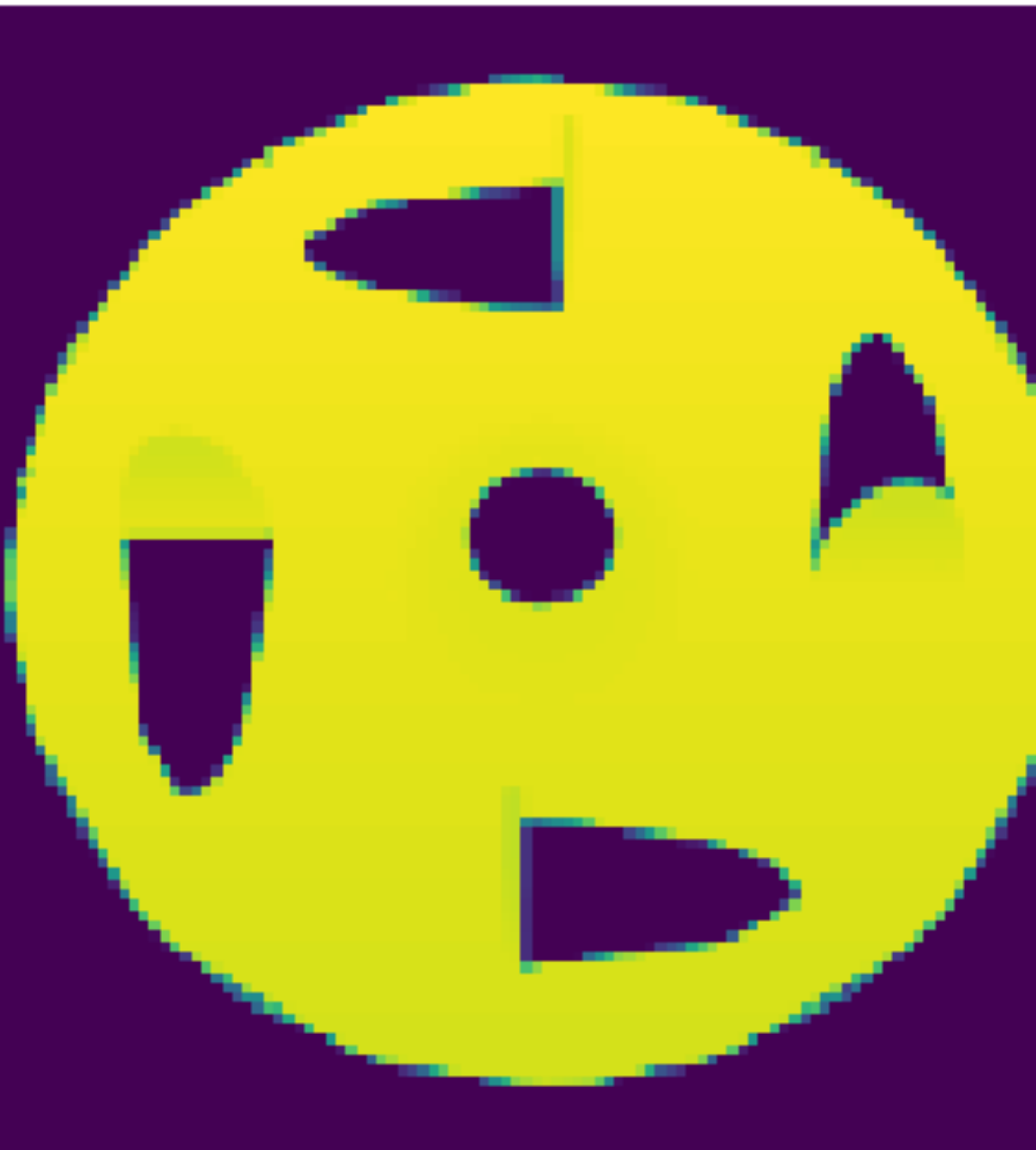}}  &
 		\fcolorbox{orange}{white}{\includegraphics[trim={9cm 4cm 9cm 4cm}, clip = true,width=0.12\linewidth]{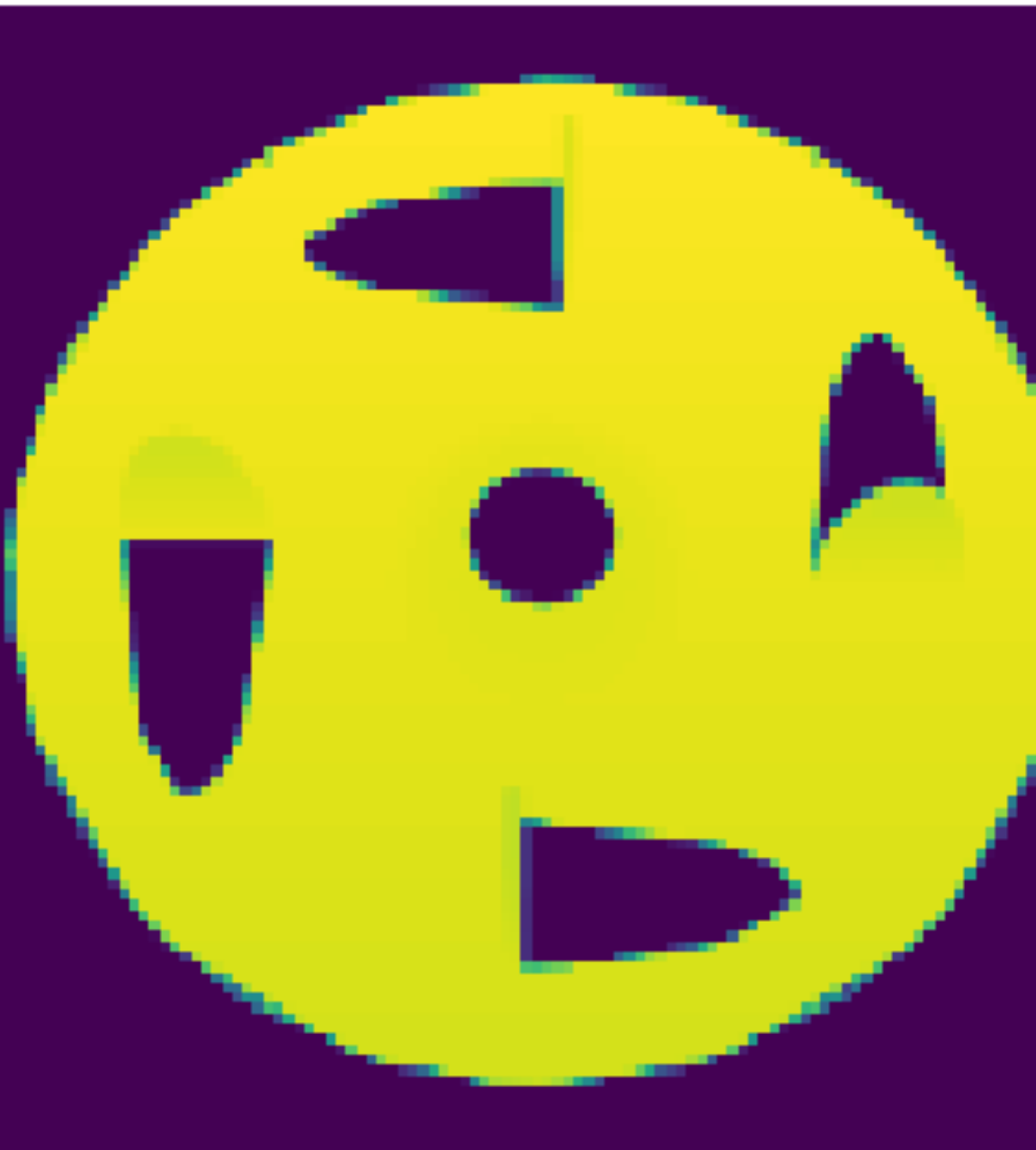}}  &
 		\includegraphics[trim={9cm 4cm 9cm 4cm}, clip = true,width=0.12\linewidth]{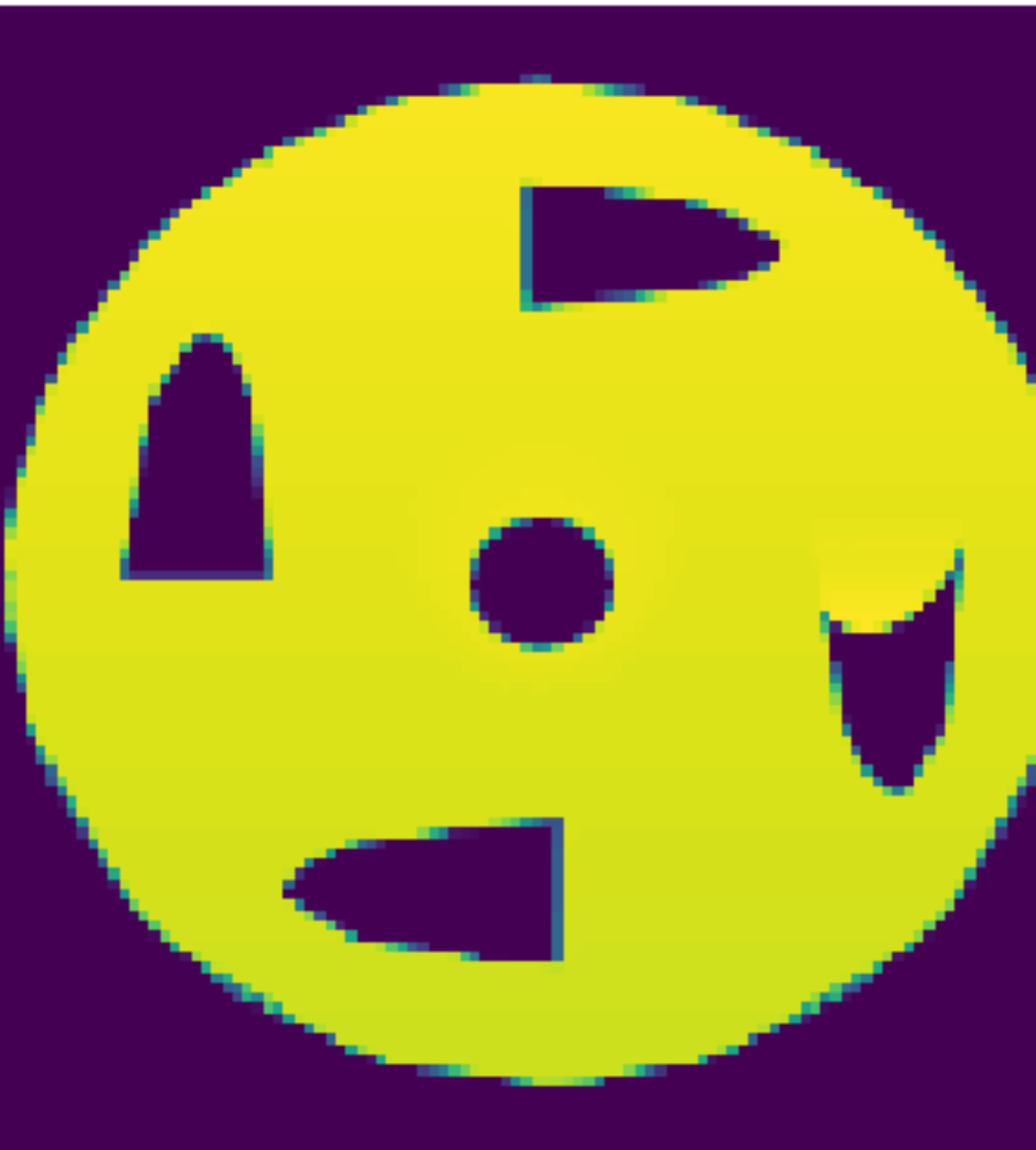}  &
 		\includegraphics[trim={9cm 4cm 9cm 4cm}, clip = true,width=0.12\linewidth]{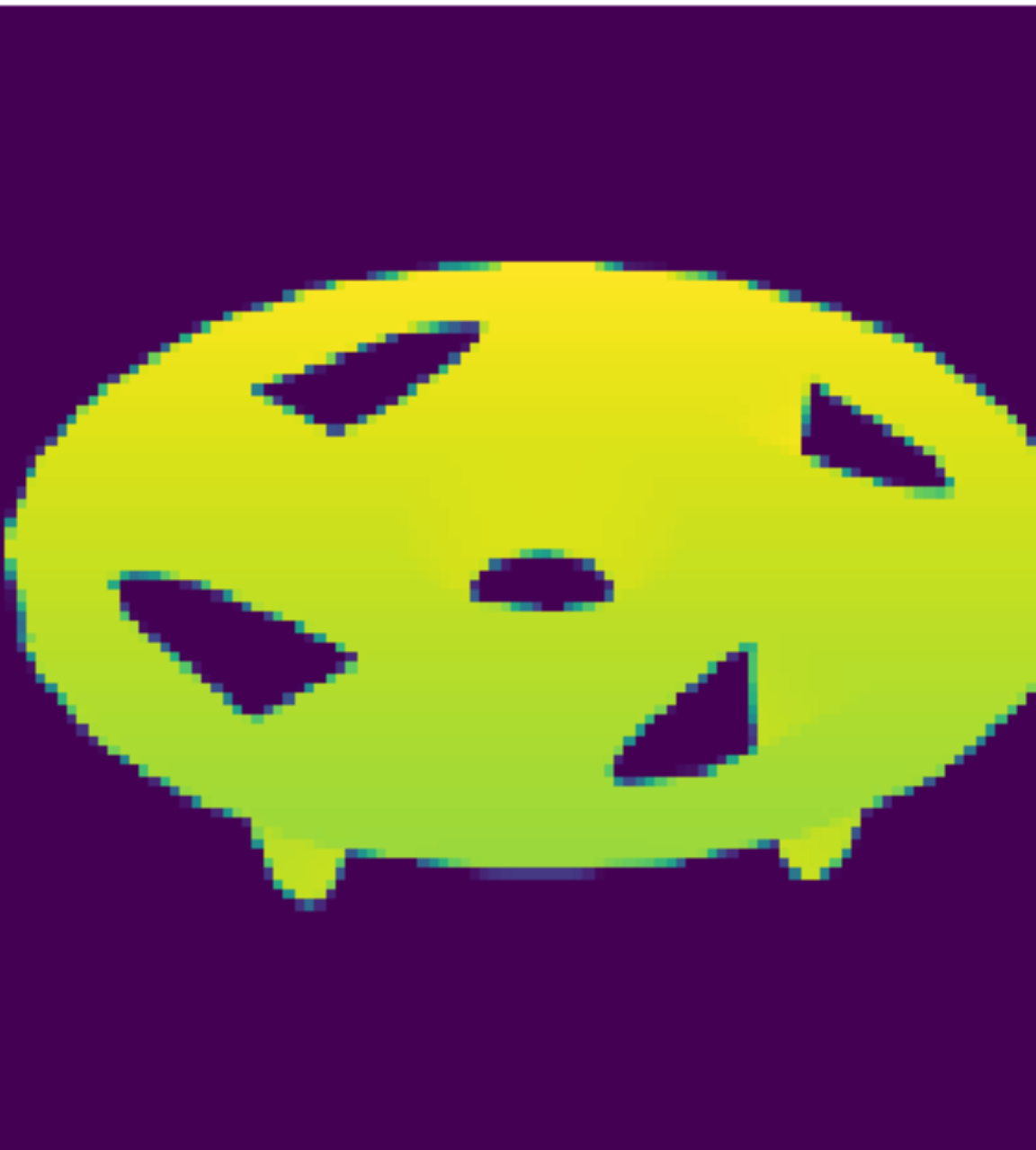}  & \raisebox{2\height}{\Large  9.75$^{\circ}$ }  \\ 
 		\cline{2-8}
 		\raisebox{3.25\height}{(c)} & \includegraphics[trim={9cm 4cm 9cm 4cm}, clip = true,width=0.12\linewidth]{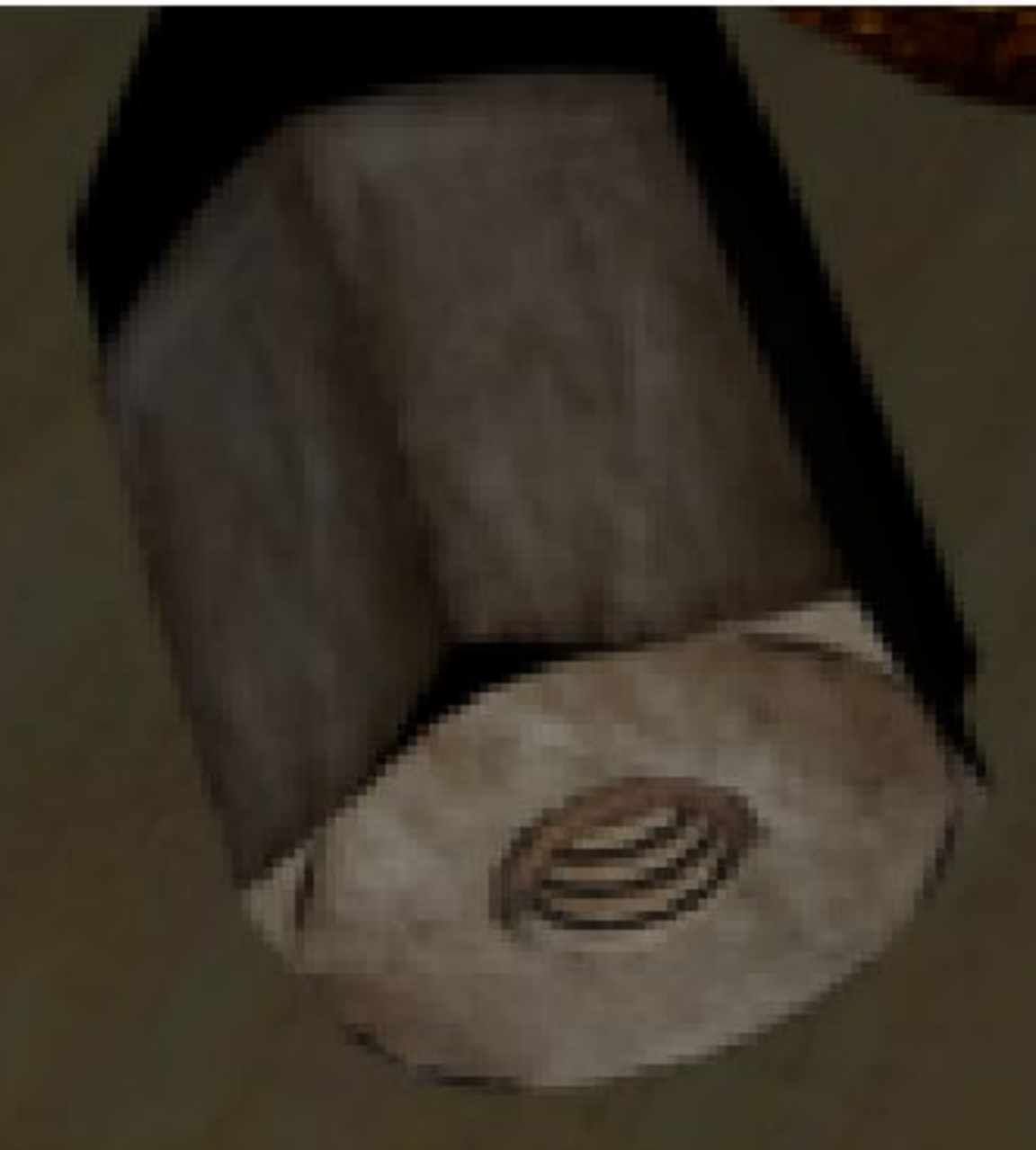}  &
 		\includegraphics[trim={9cm 4cm 9cm 4cm}, clip = true,width=0.12\linewidth]{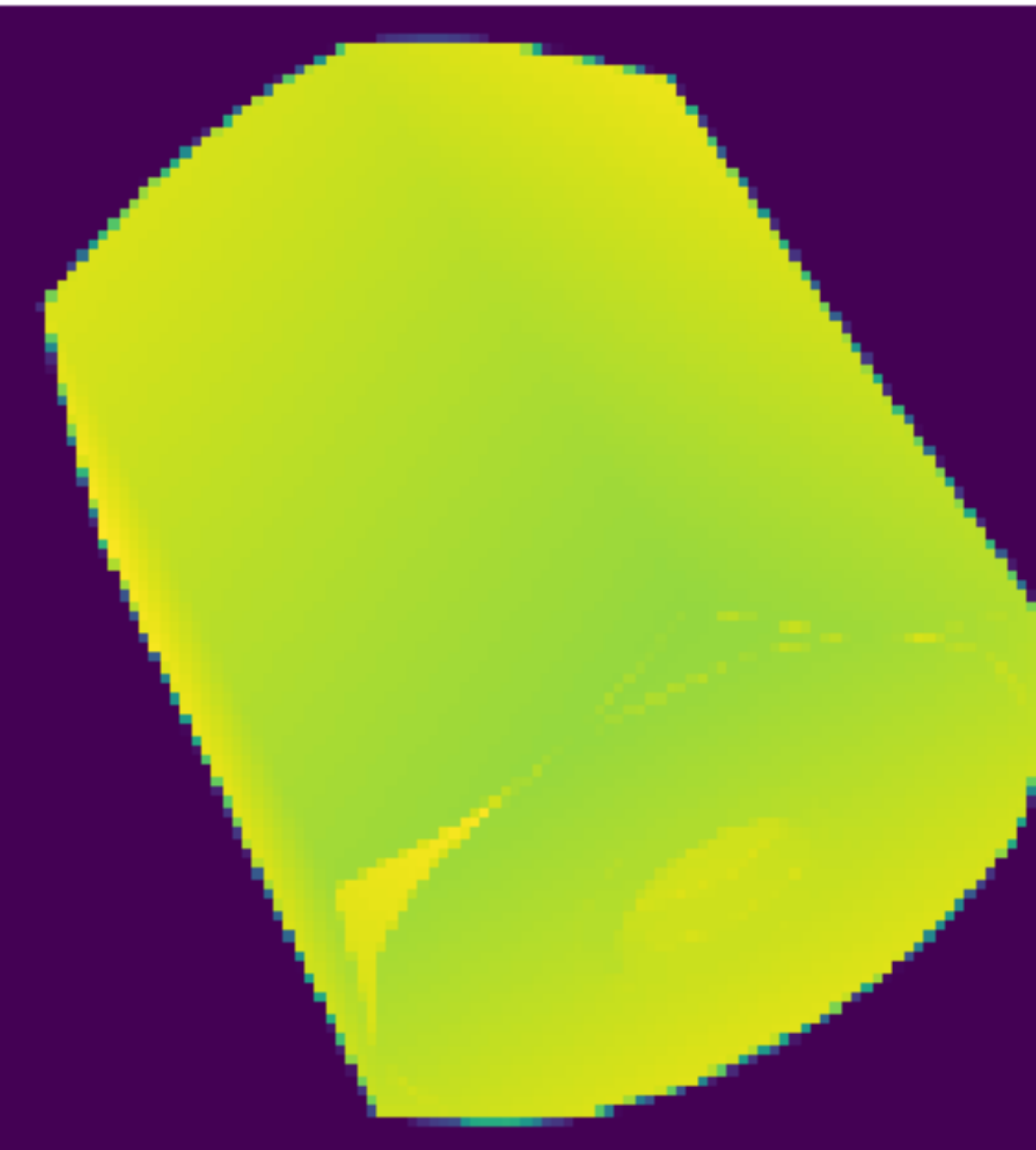}  &
 		\fcolorbox{orange}{white}{\includegraphics[trim={9cm 4cm 9cm 4cm}, clip = true,width=0.12\linewidth]{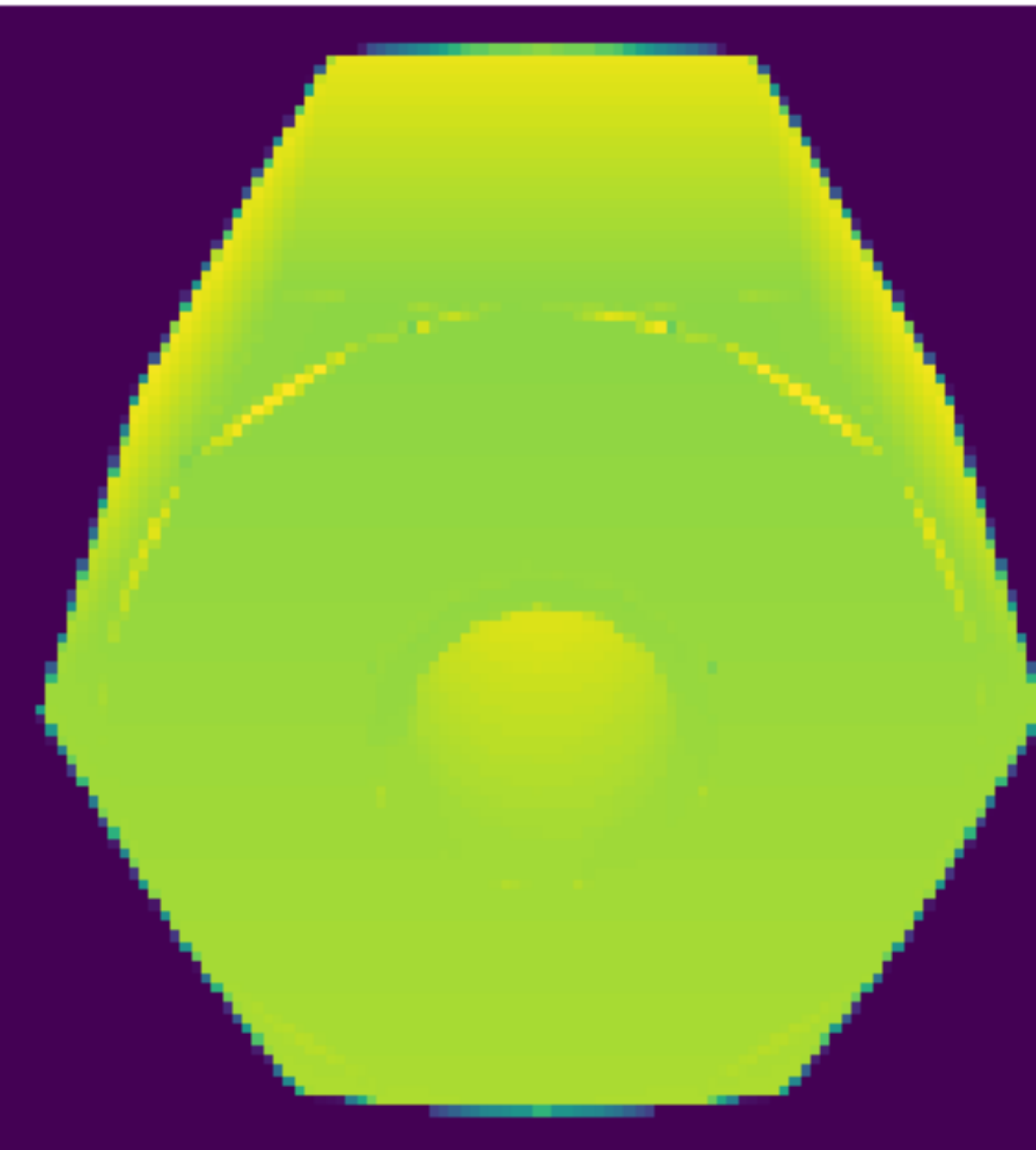}}  &
 		\includegraphics[trim={9cm 4cm 9cm 4cm}, clip = true,width=0.12\linewidth]{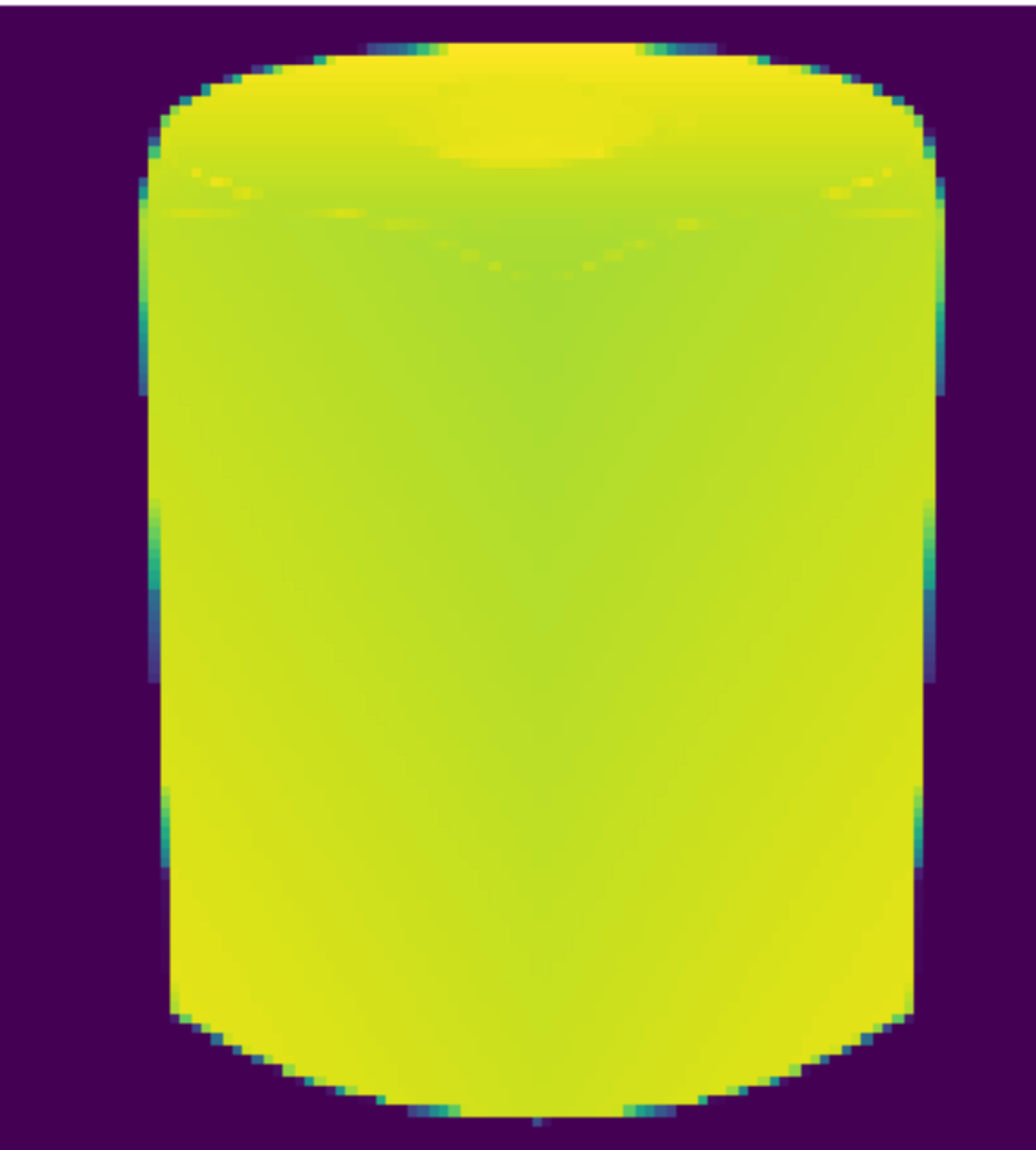}  &
 		\includegraphics[trim={9cm 4cm 9cm 4cm}, clip = true,width=0.12\linewidth]{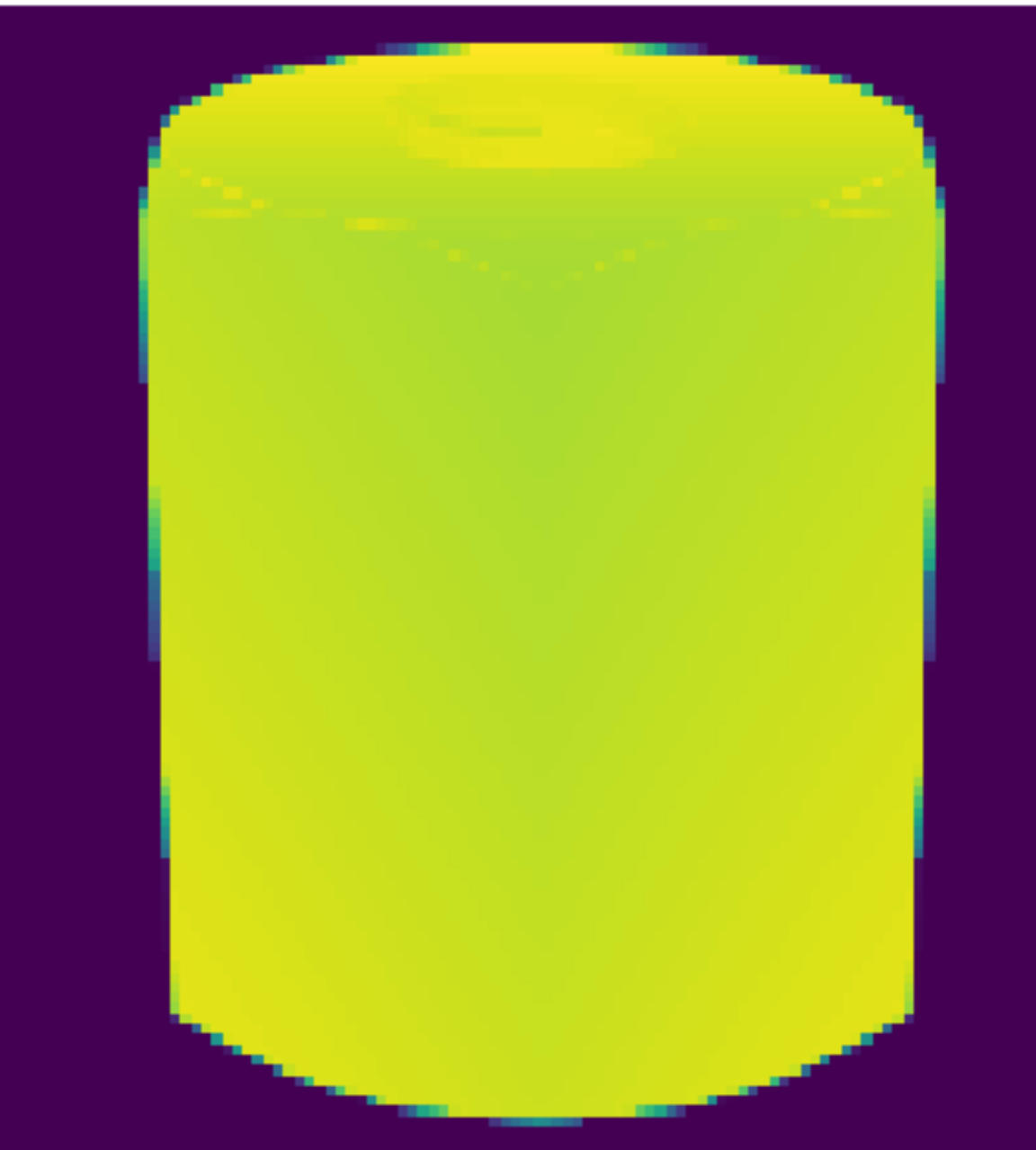} &
 		\includegraphics[trim={9cm 4cm 9cm 4cm}, clip = true,width=0.12\linewidth]{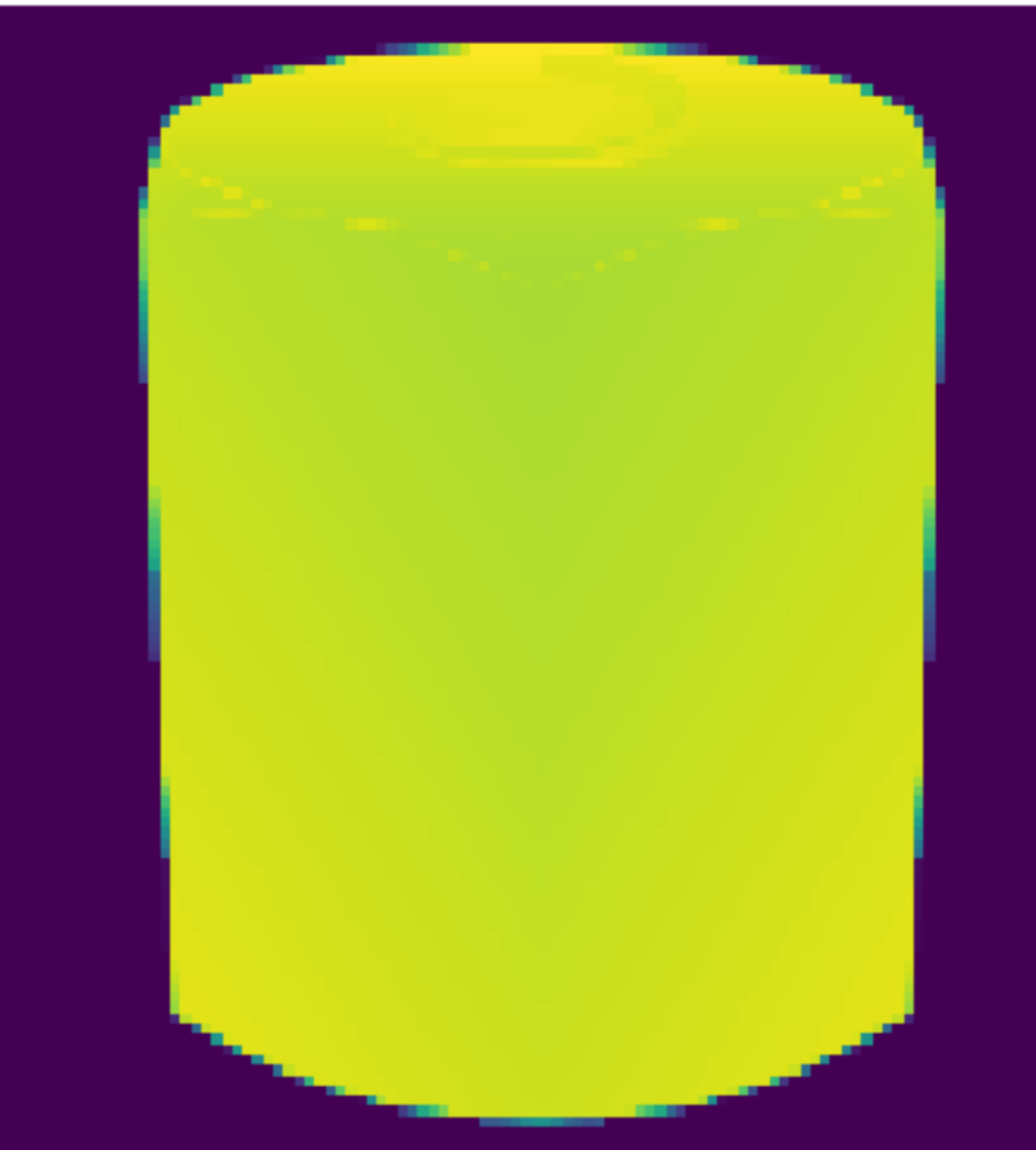}  & \raisebox{2\height}{\Large  15.38$^{\circ}$ }  \\ 
 		\cline{2-8} 
 		\raisebox{3.25\height}{(d)} & 
 		\includegraphics[trim={9cm 4cm 9cm 4cm}, clip = true,width=0.12\linewidth]{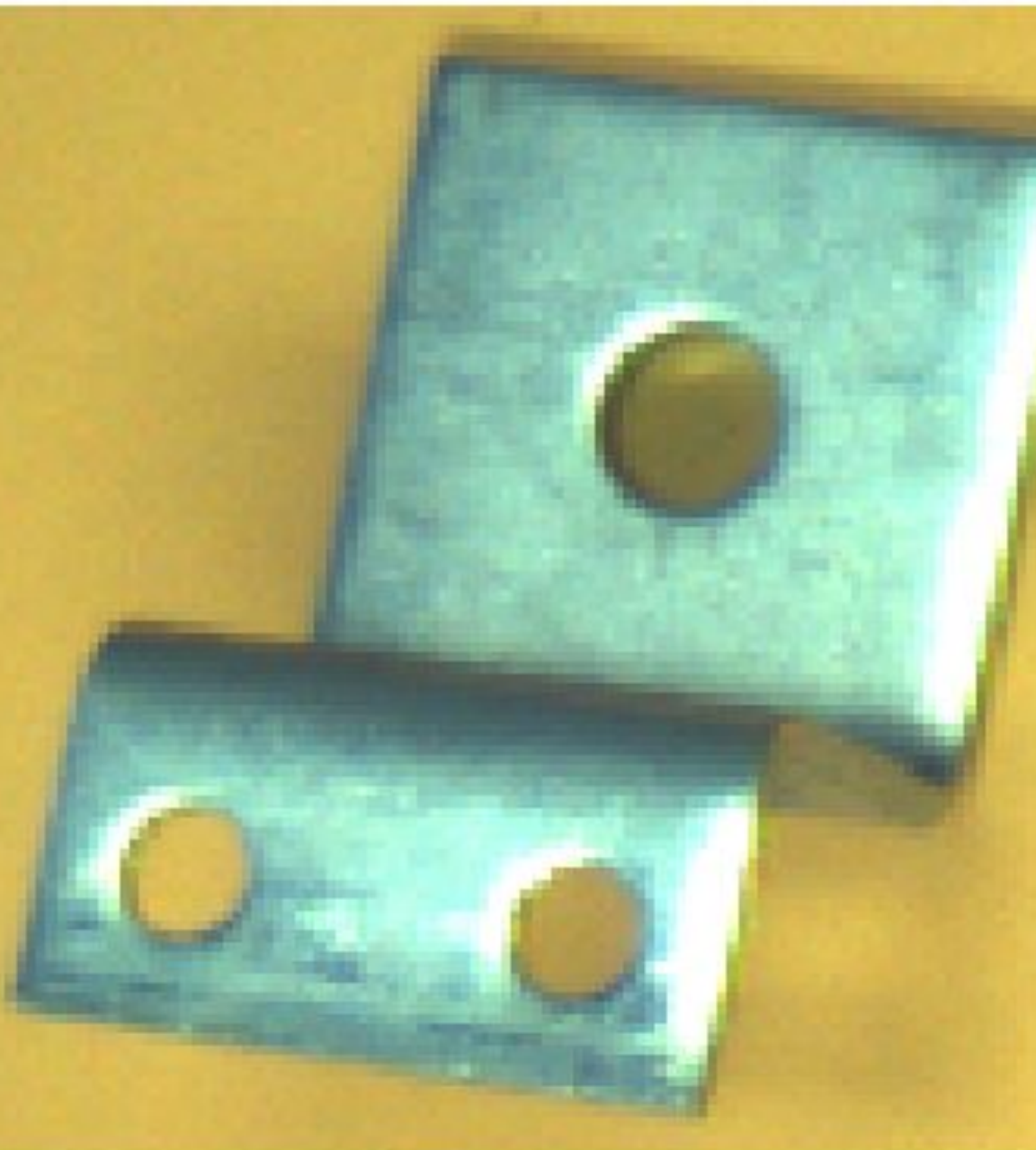} &
 		\includegraphics[trim={9cm 4cm 9cm 4cm}, clip = true,width=0.12\linewidth]{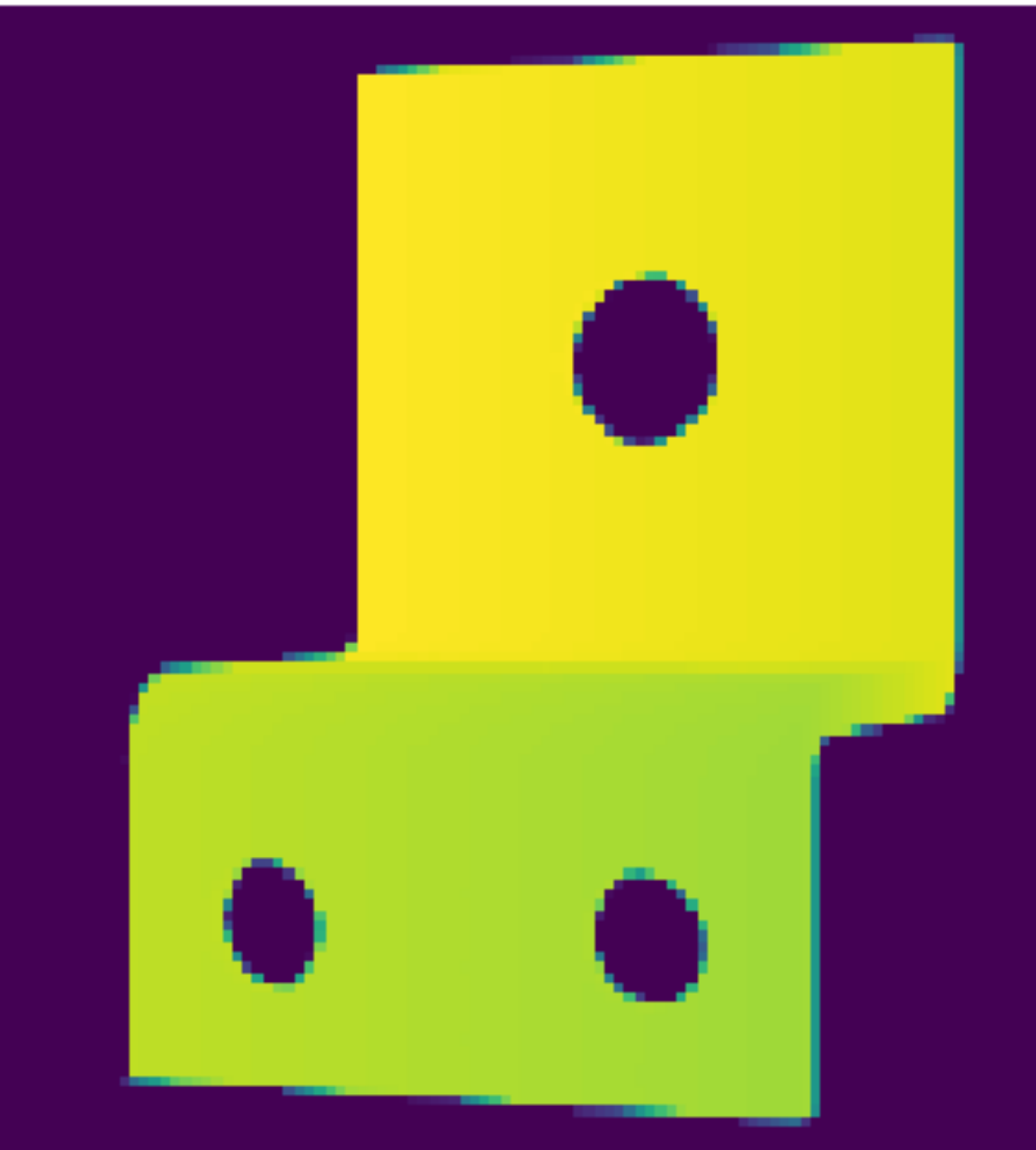} &
 		\includegraphics[trim={9cm 4cm 9cm 4cm}, clip = true,width=0.12\linewidth]{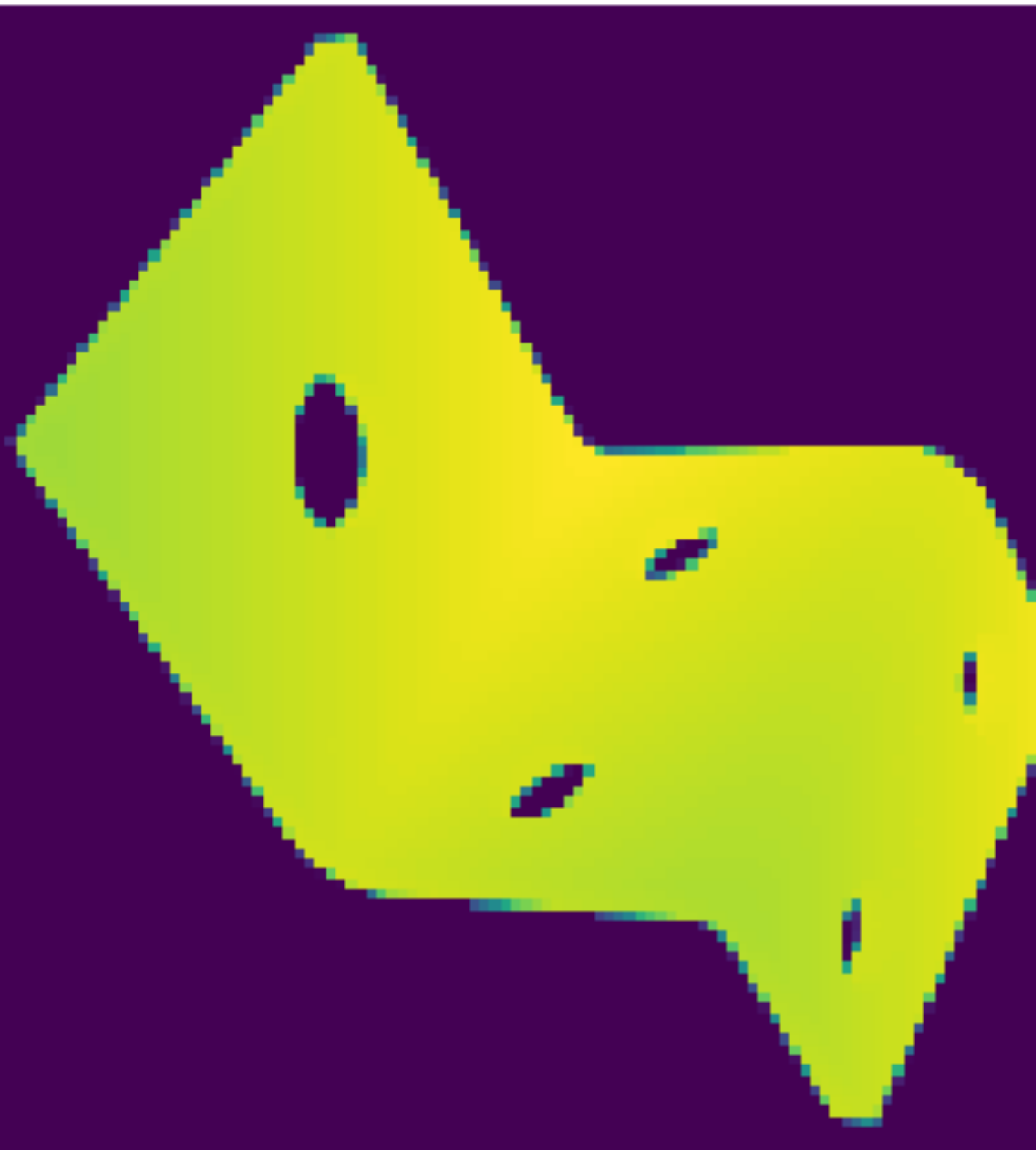} &
 		\includegraphics[trim={9cm 4cm 9cm 4cm}, clip = true,width=0.12\linewidth]{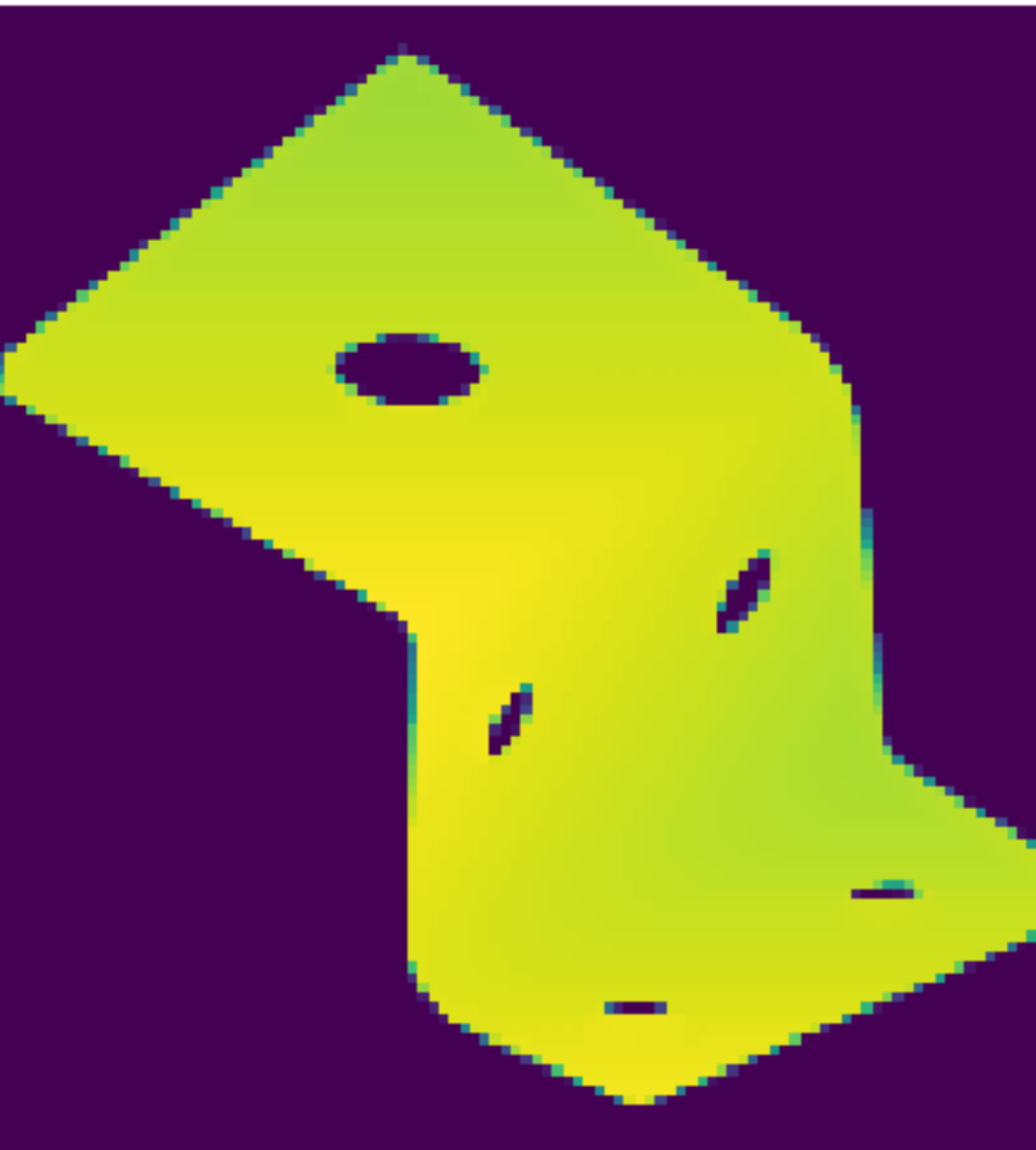} &
 		\includegraphics[trim={9cm 4cm 9cm 4cm}, clip = true,width=0.12\linewidth]{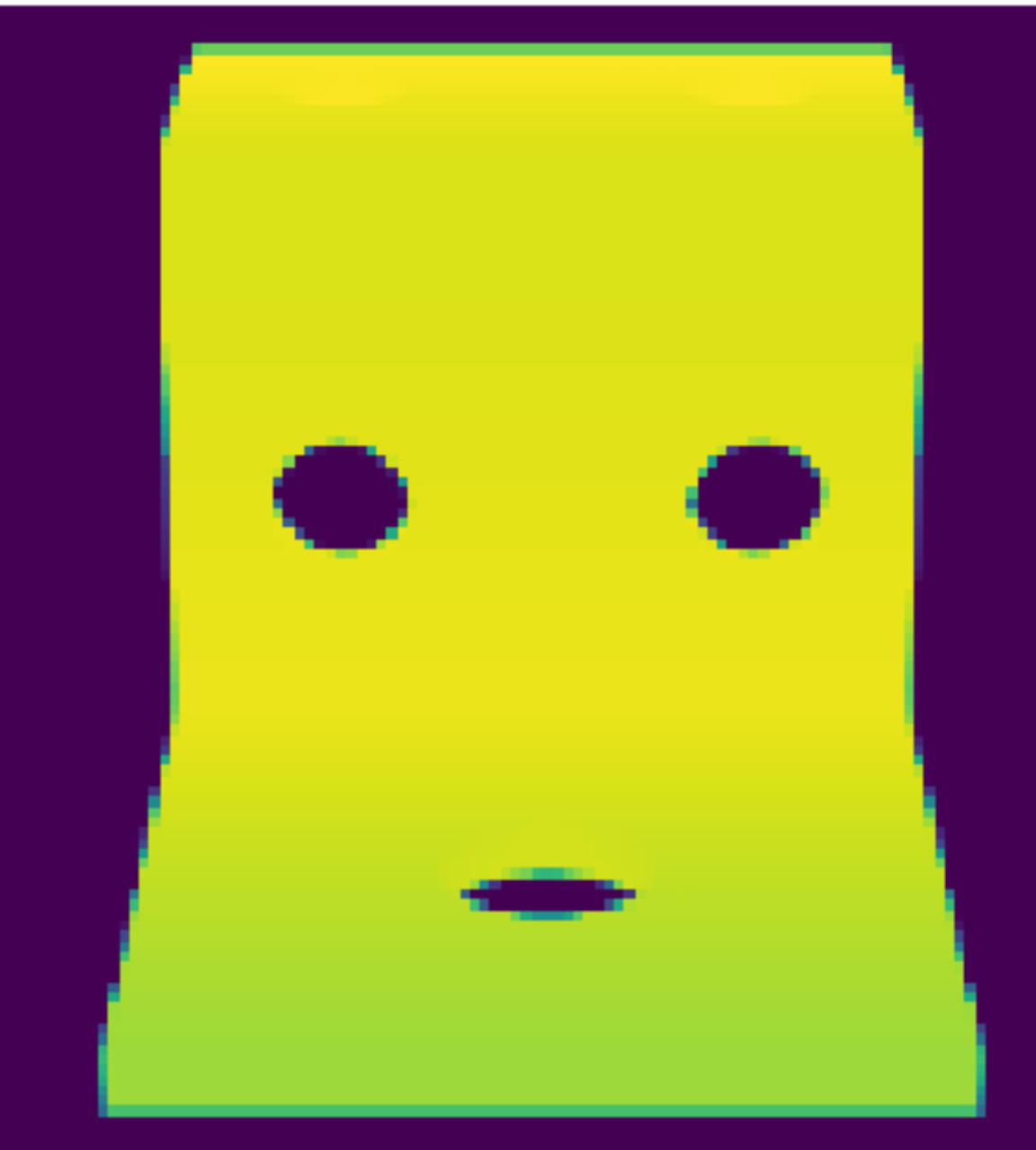} &
 		\fcolorbox{orange}{white}{\includegraphics[trim={9cm 4cm 9cm 4cm}, clip = true,width=0.12\linewidth]{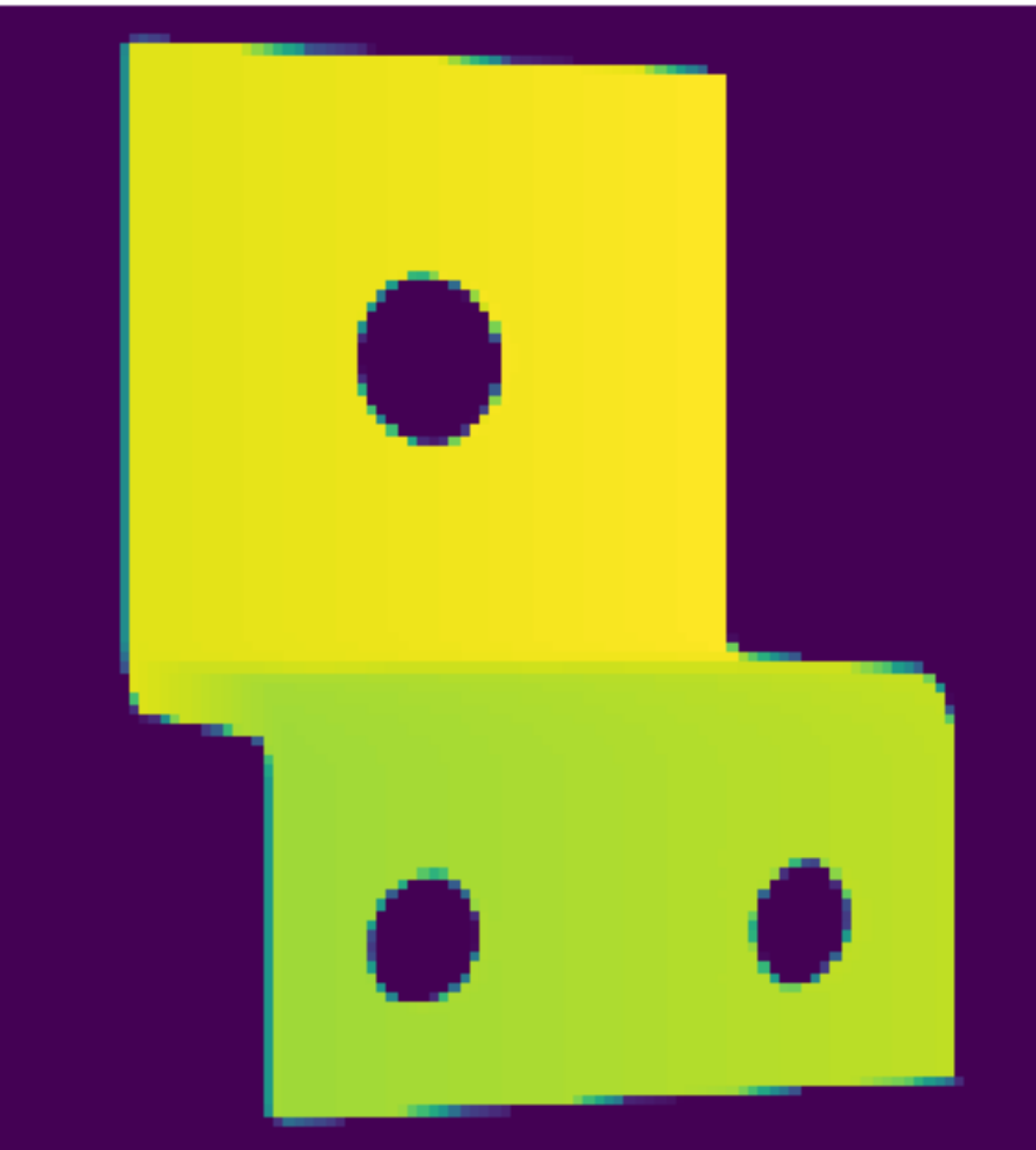} } & \raisebox{2\height}{\Large 15.78$^{\circ}$}  \\
 		\cline{2-8}
 		\raisebox{3.25\height}{(e)} & \includegraphics[trim={9cm 4cm 9cm 4cm}, clip = true,width=0.12\linewidth]{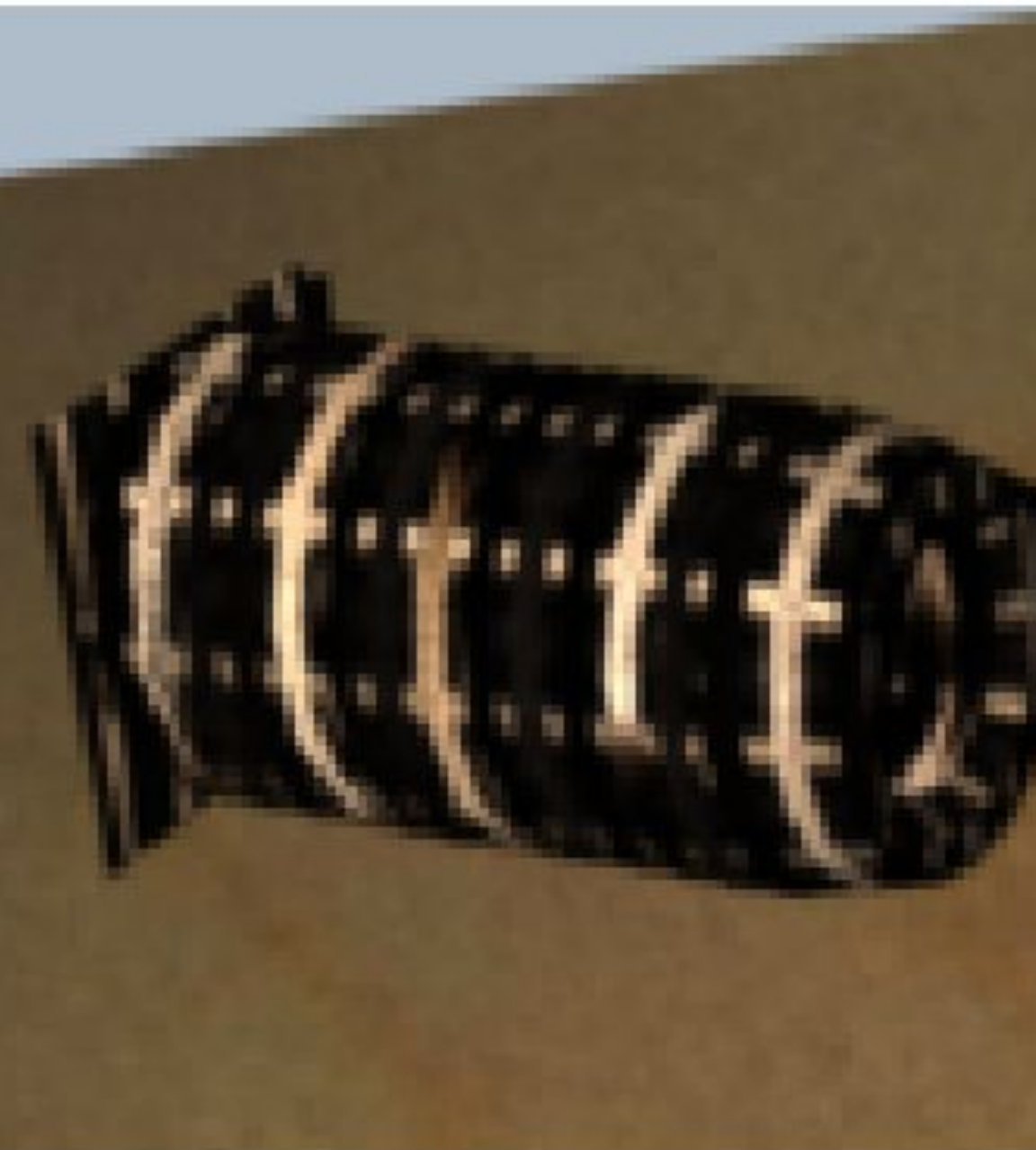} &
 		\includegraphics[trim={9cm 4cm 9cm 4cm}, clip = true,width=0.12\linewidth]{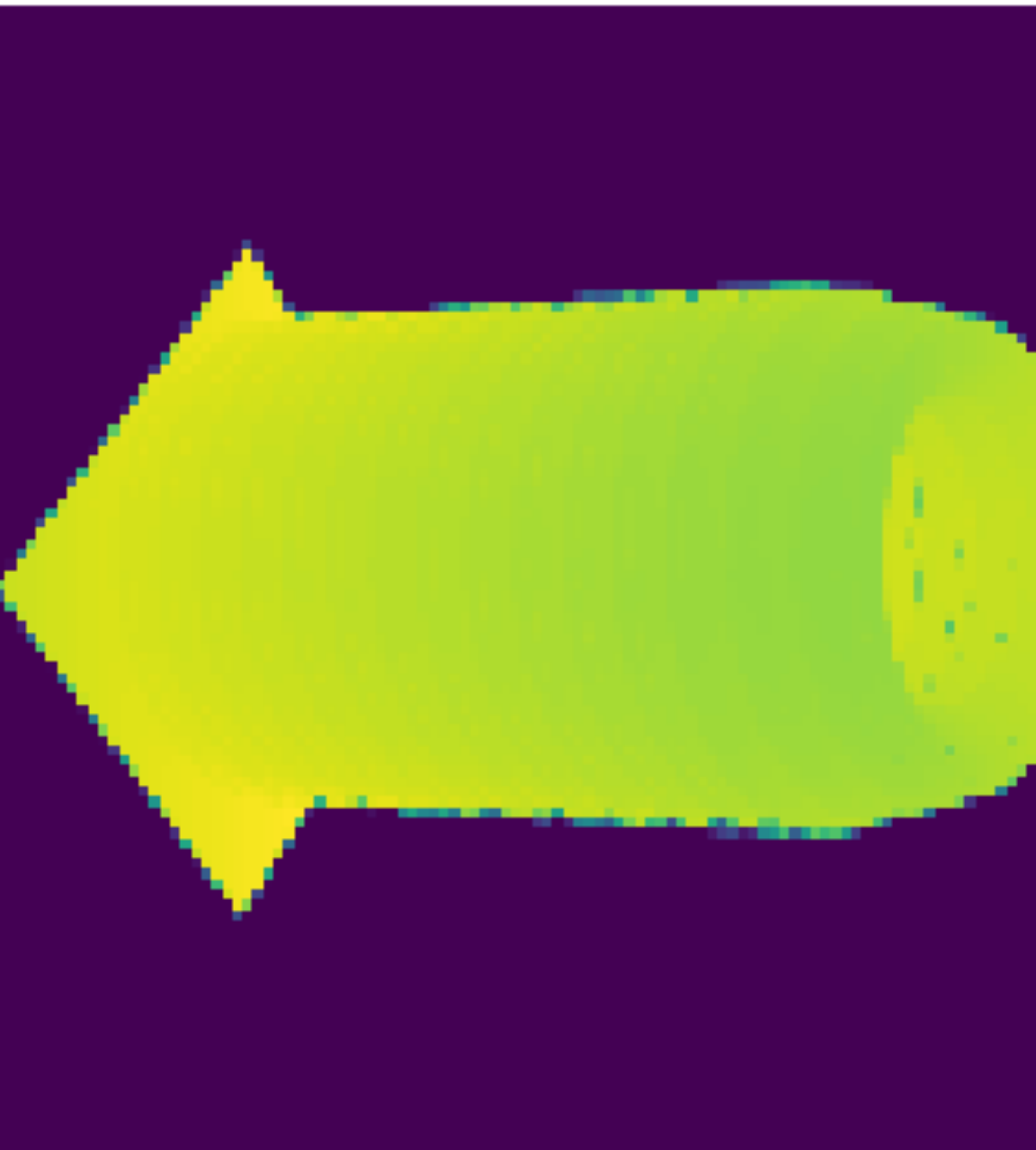} &
 		\fcolorbox{orange}{white}{\includegraphics[trim={9cm 4cm 9cm 4cm}, clip = true,width=0.12\linewidth]{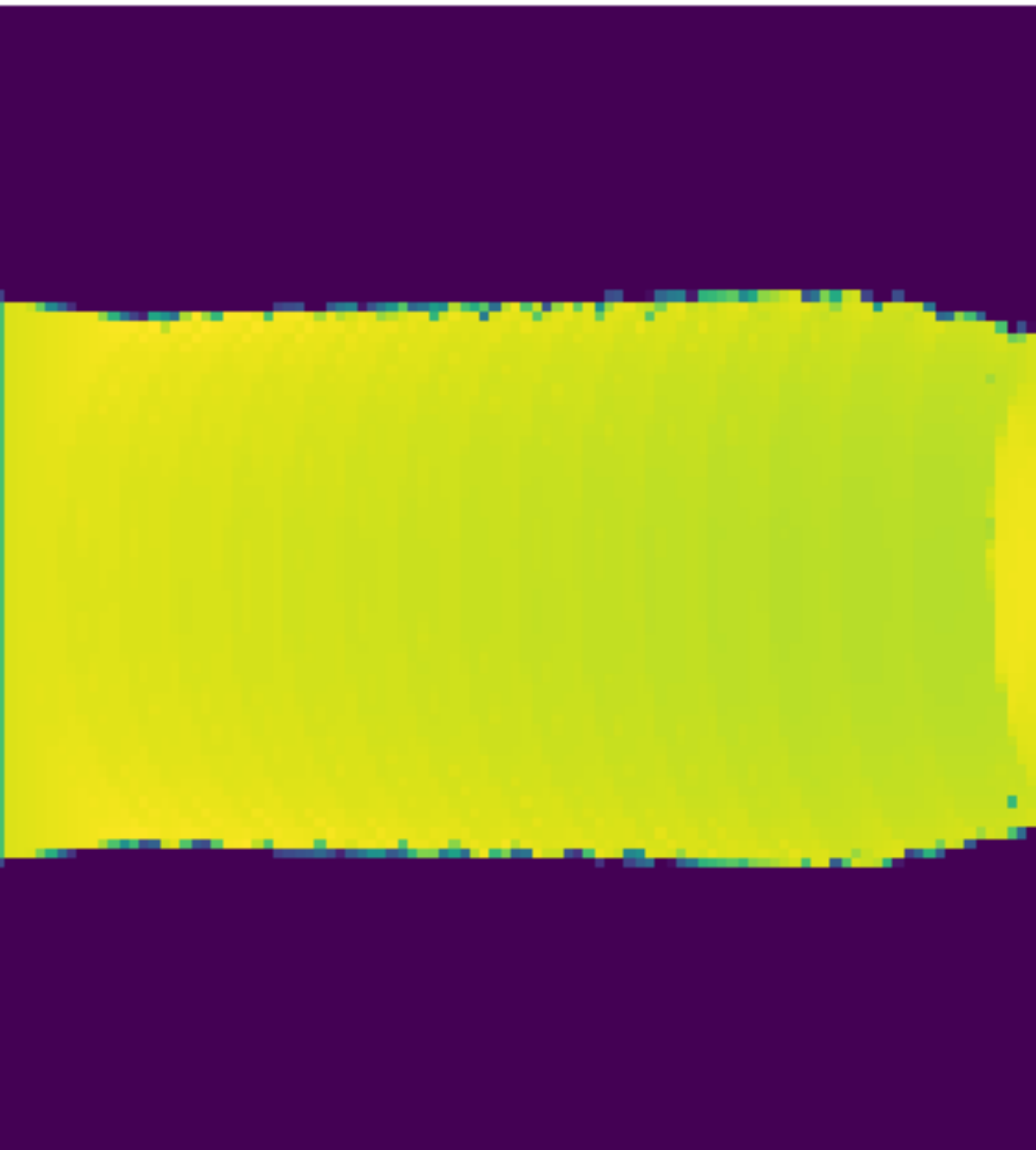}} &
 		\includegraphics[trim={9cm 4cm 9cm 4cm}, clip = true,width=0.12\linewidth]{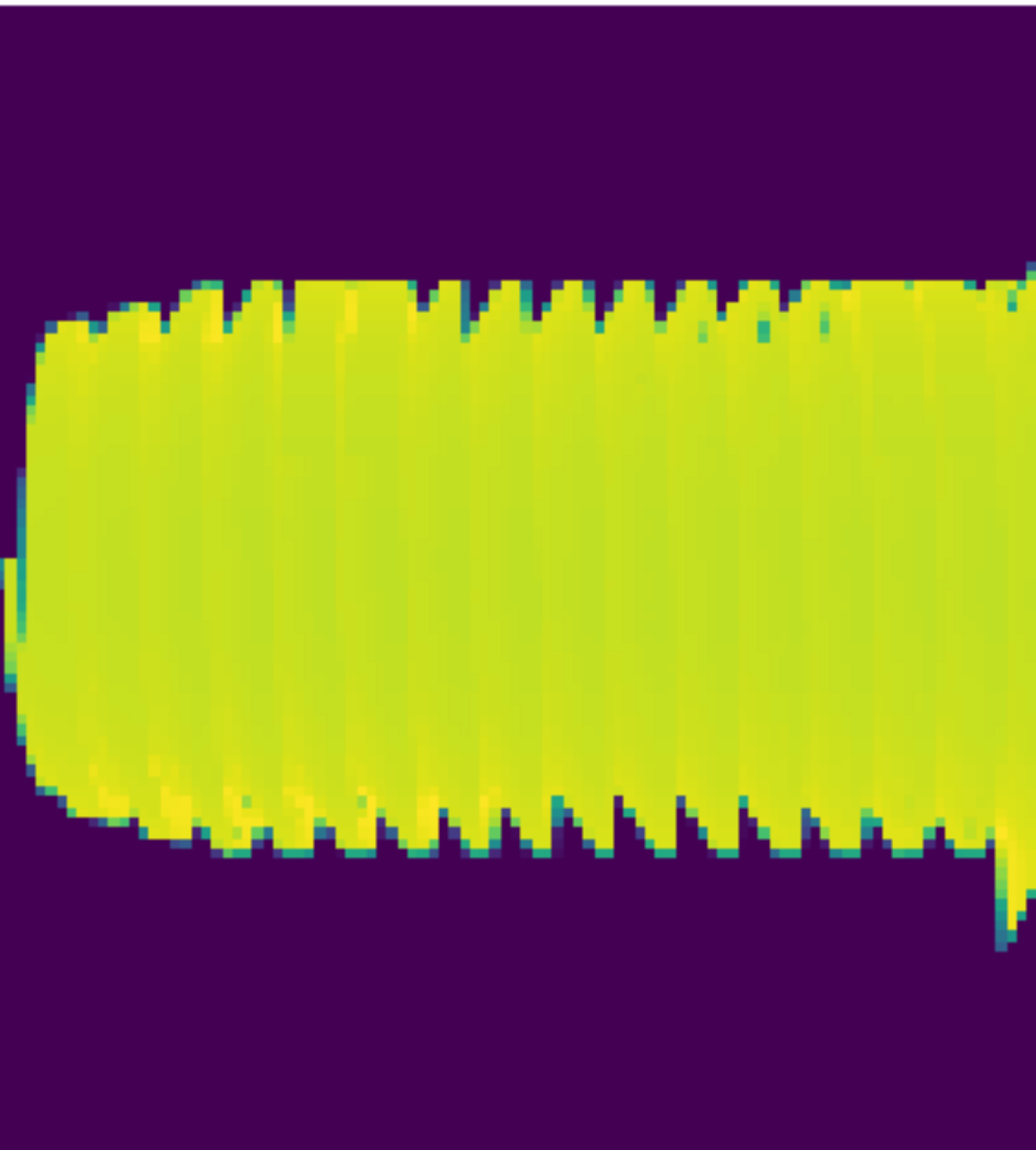} &
 		\includegraphics[trim={9cm 4cm 9cm 4cm}, clip = true,width=0.12\linewidth]{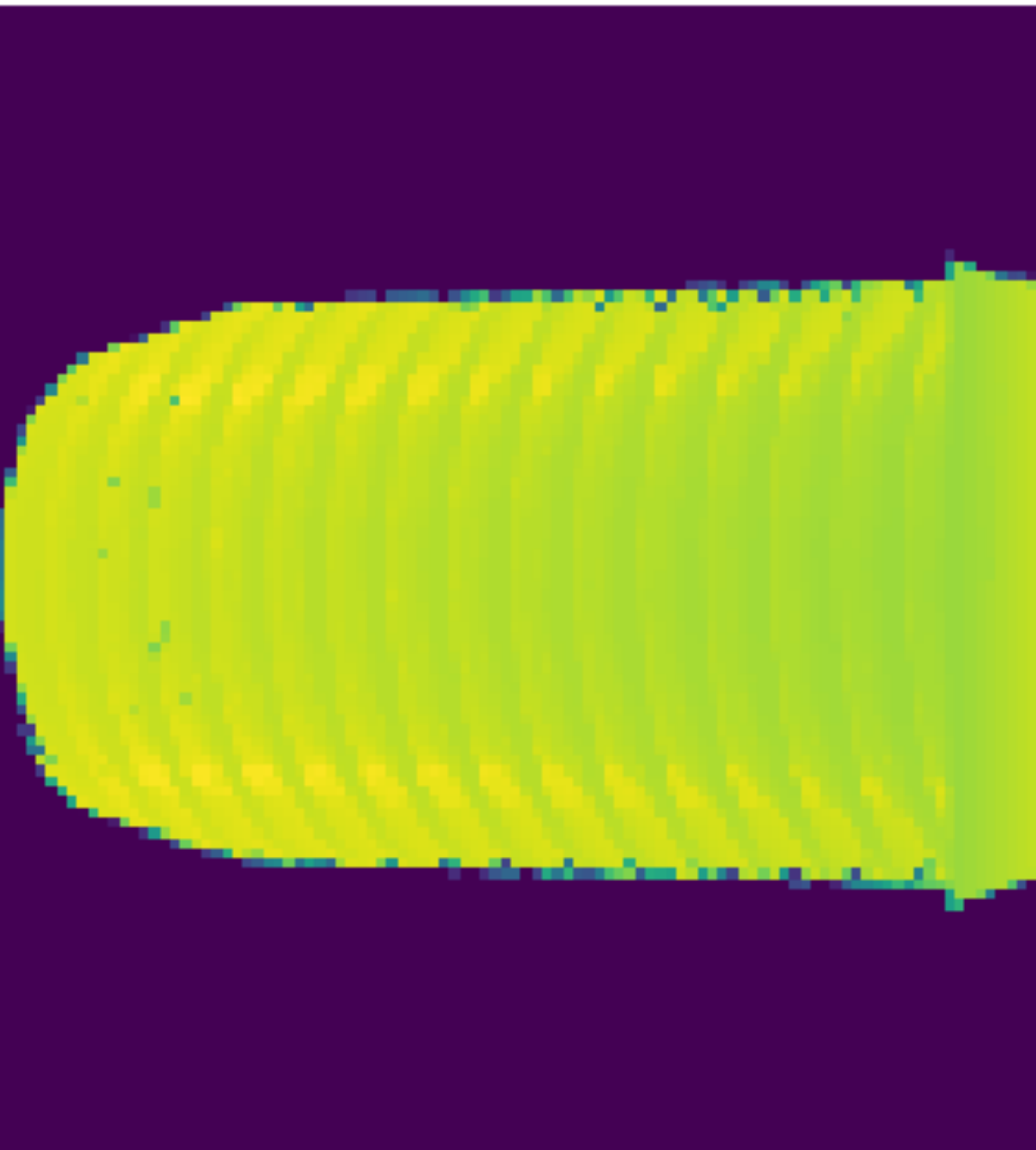} &
 		\includegraphics[trim={9cm 4cm 9cm 4cm}, clip = true,width=0.12\linewidth]{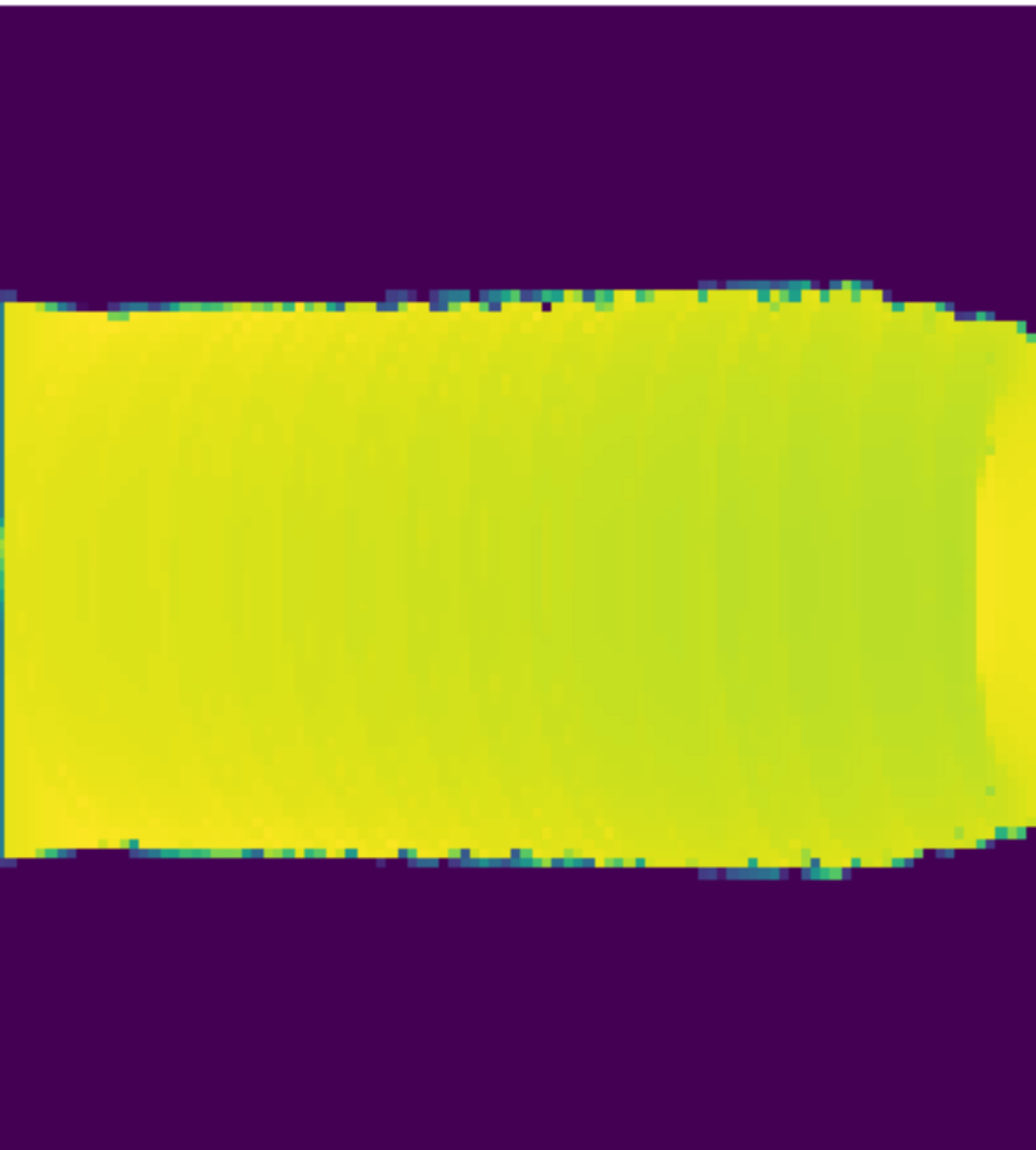} & \raisebox{2\height}{\Large  17.05$^{\circ}$ } \\
 		\cline{2-8} 
 		\raisebox{3.25\height}{(f)} & 
 		\includegraphics[trim={9cm 4cm 9cm 4cm}, clip = true,width=0.12\linewidth]{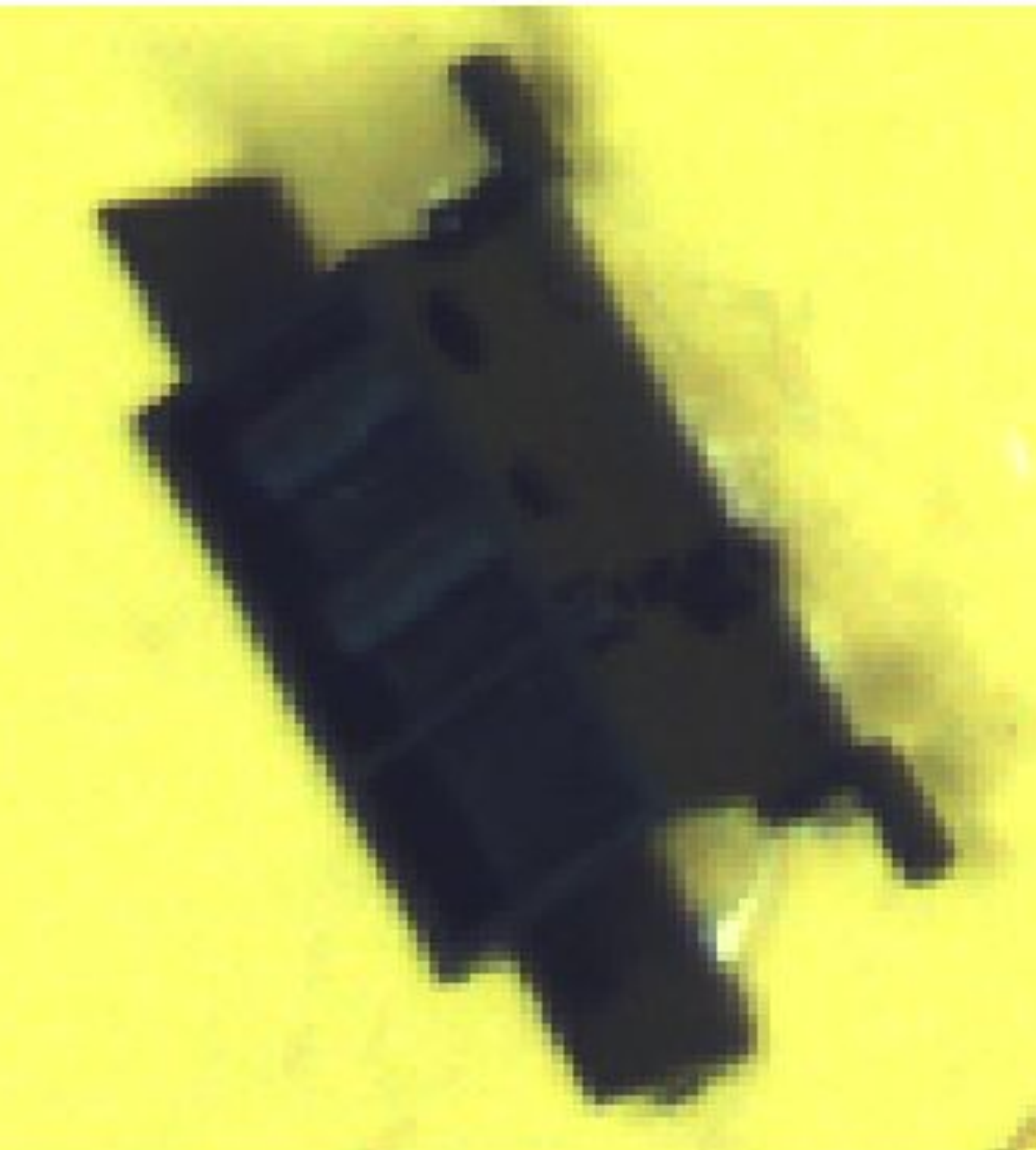} &
 		\includegraphics[trim={9cm 4cm 9cm 4cm}, clip = true,width=0.12\linewidth]{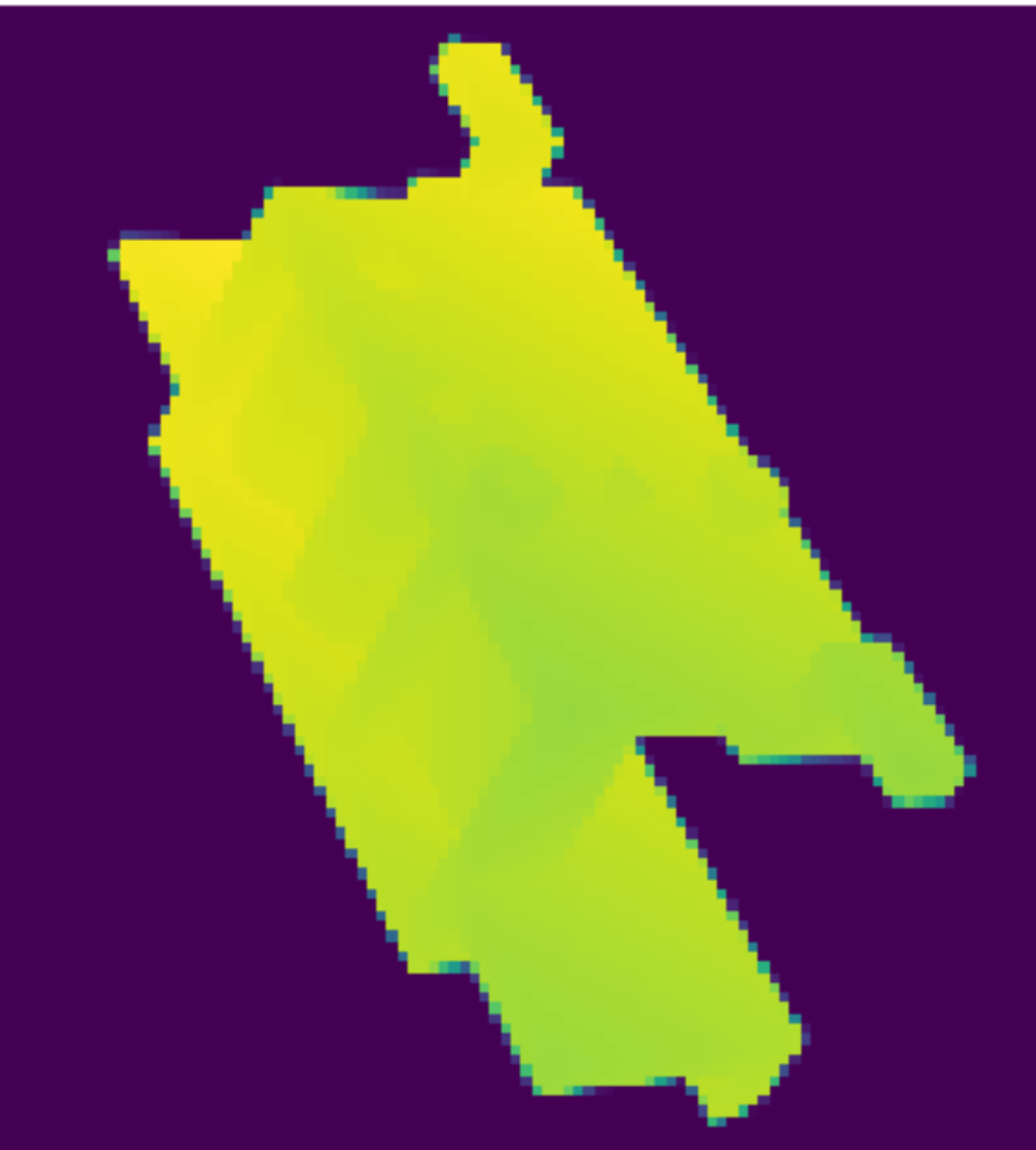} &
 		\fcolorbox{orange}{white} {\includegraphics[trim={9cm 4cm 9cm 4cm}, clip = true,width=0.12\linewidth]{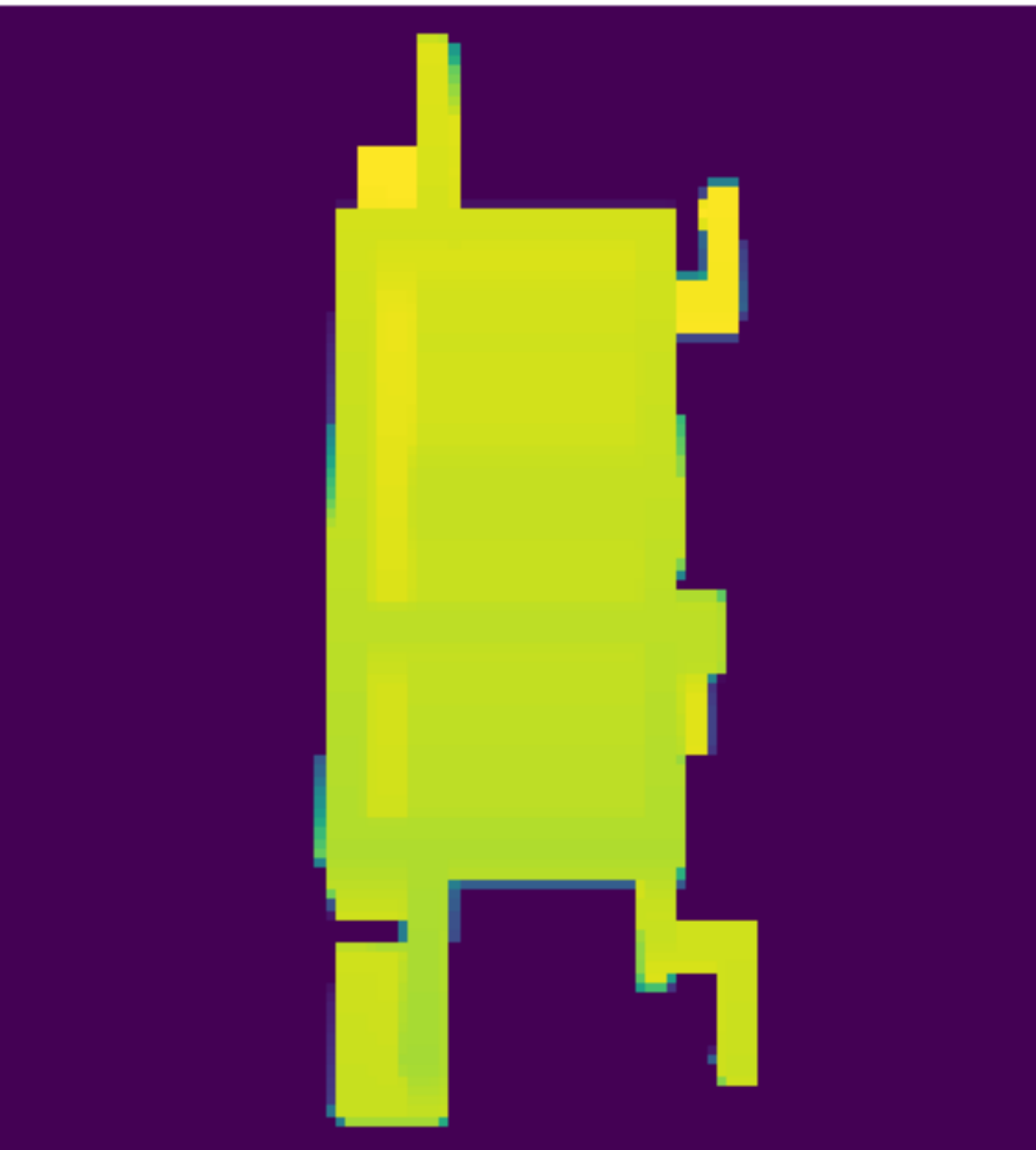} } &
 		\includegraphics[trim={9cm 4cm 9cm 4cm}, clip = true,width=0.12\linewidth]{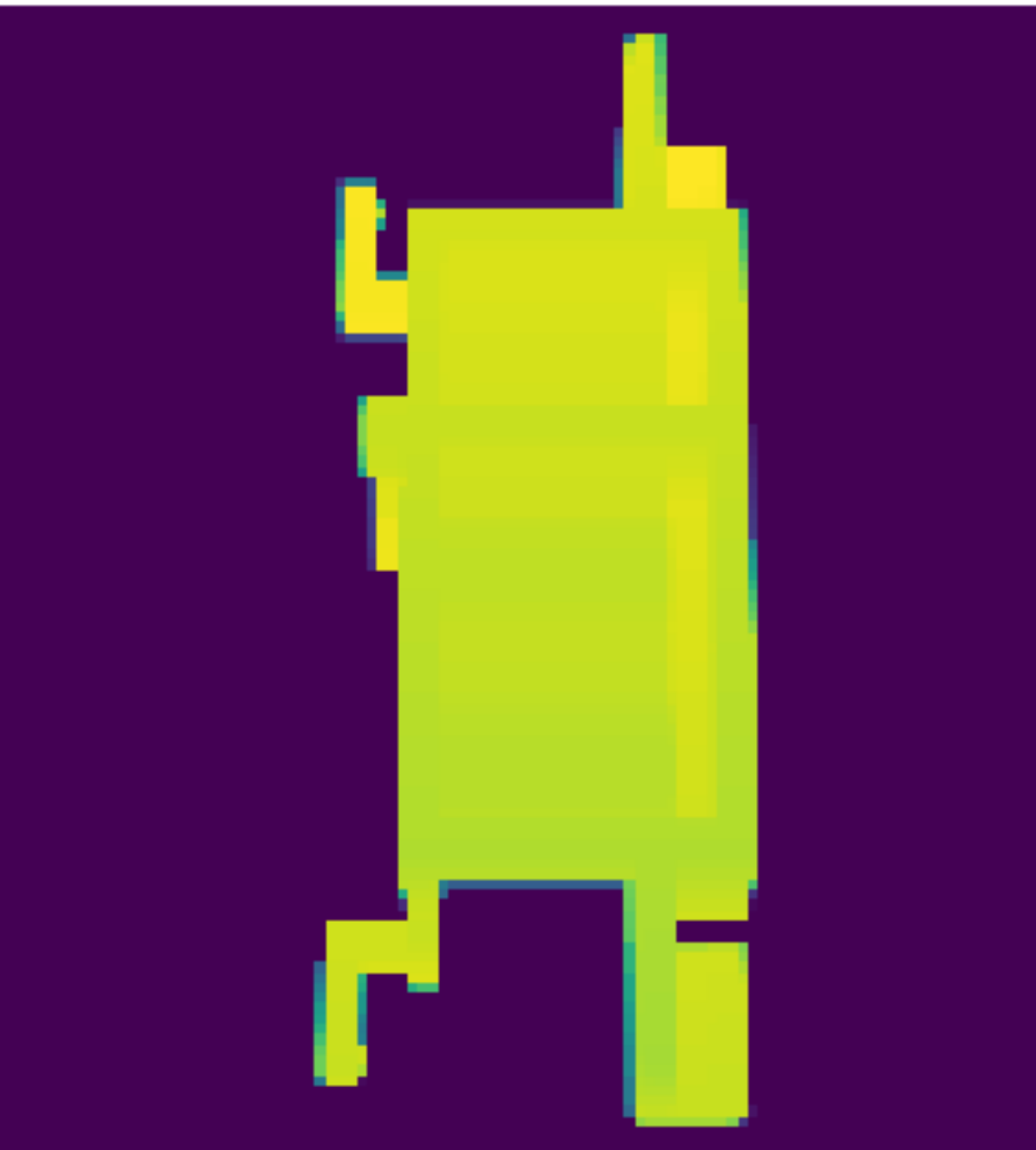} &
 		\includegraphics[trim={9cm 4cm 9cm 4cm}, clip = true,width=0.12\linewidth]{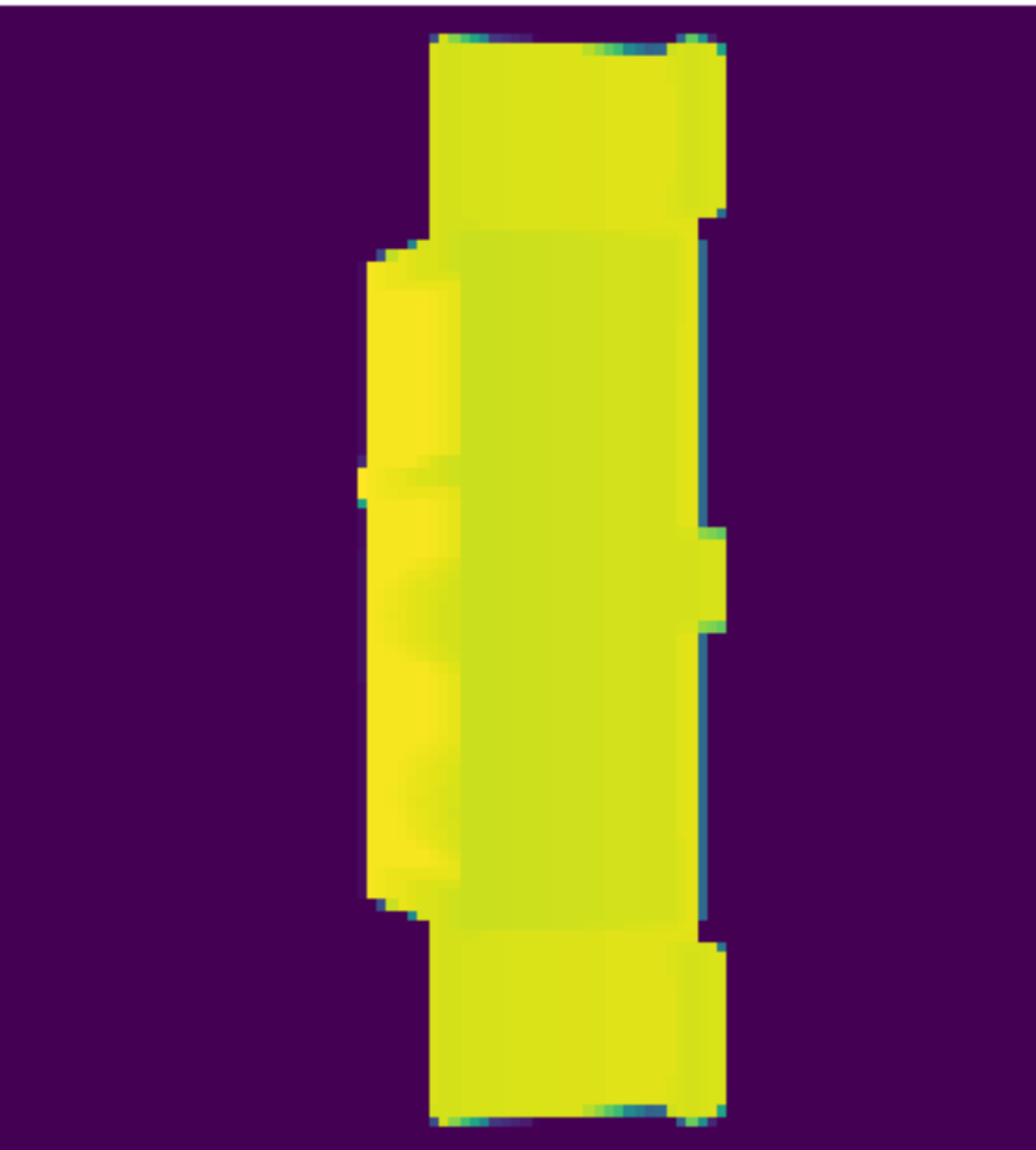} &
 		\includegraphics[trim={9cm 4cm 9cm 4cm}, clip = true,width=0.12\linewidth]{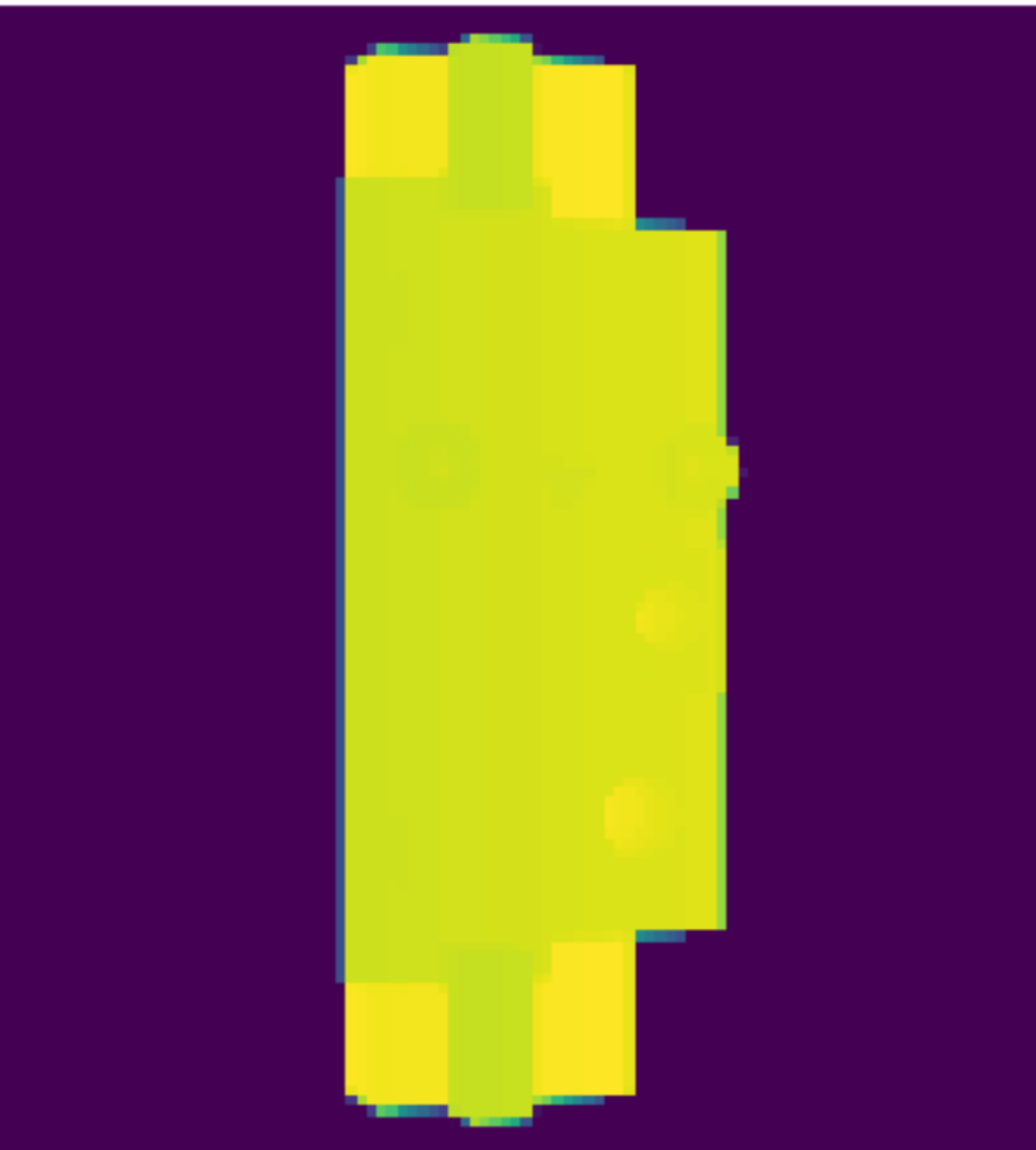} & \raisebox{2\height}{\Large 21.14$^{\circ}$}  \\
 		\cline{2-8}
 		\raisebox{3.25\height}{(g)} & \includegraphics[trim={9cm 4cm 9cm 4cm}, clip = true,width=0.12\linewidth]{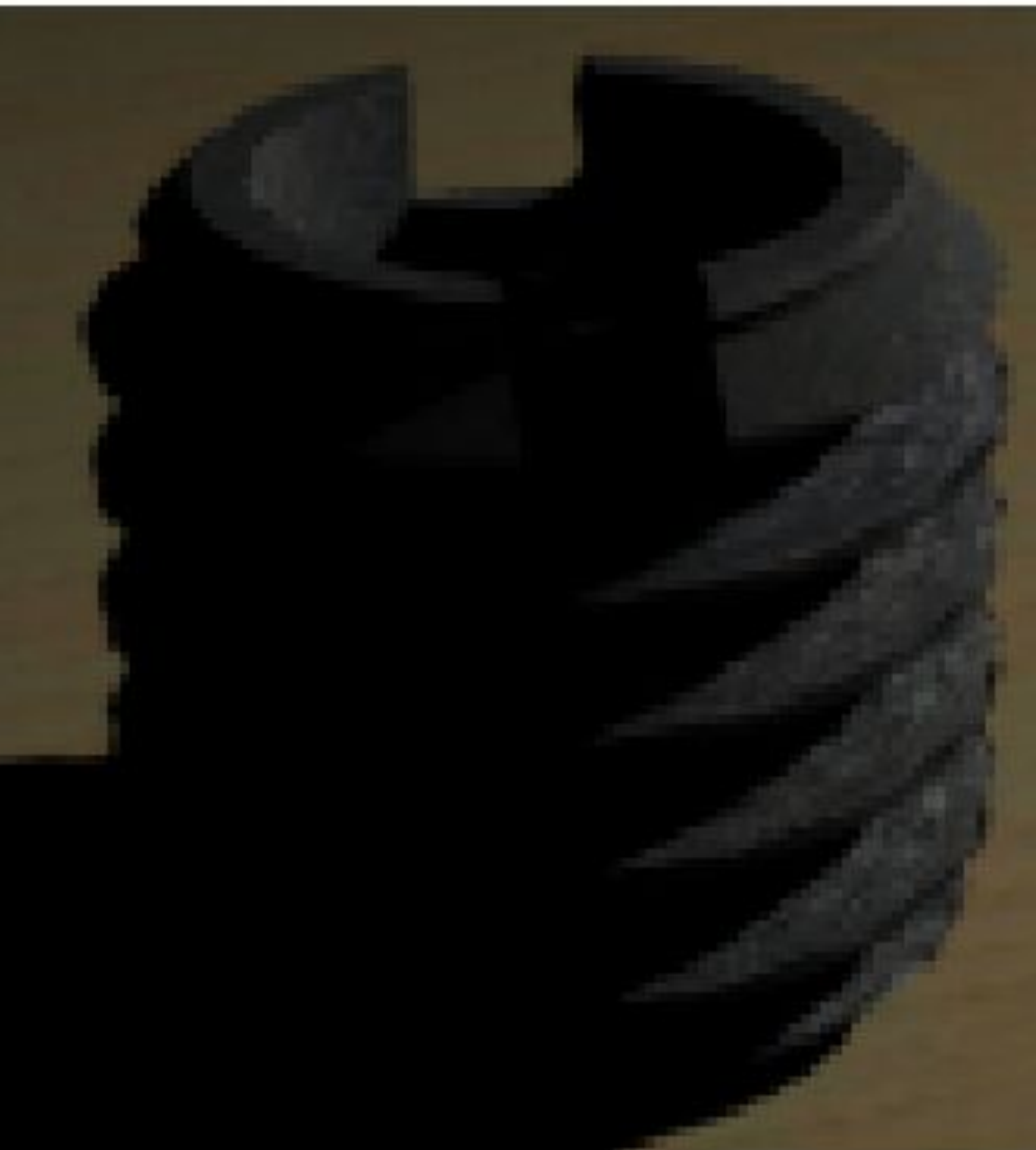} &
 		\includegraphics[trim={9cm 4cm 9cm 4cm}, clip = true,width=0.12\linewidth]{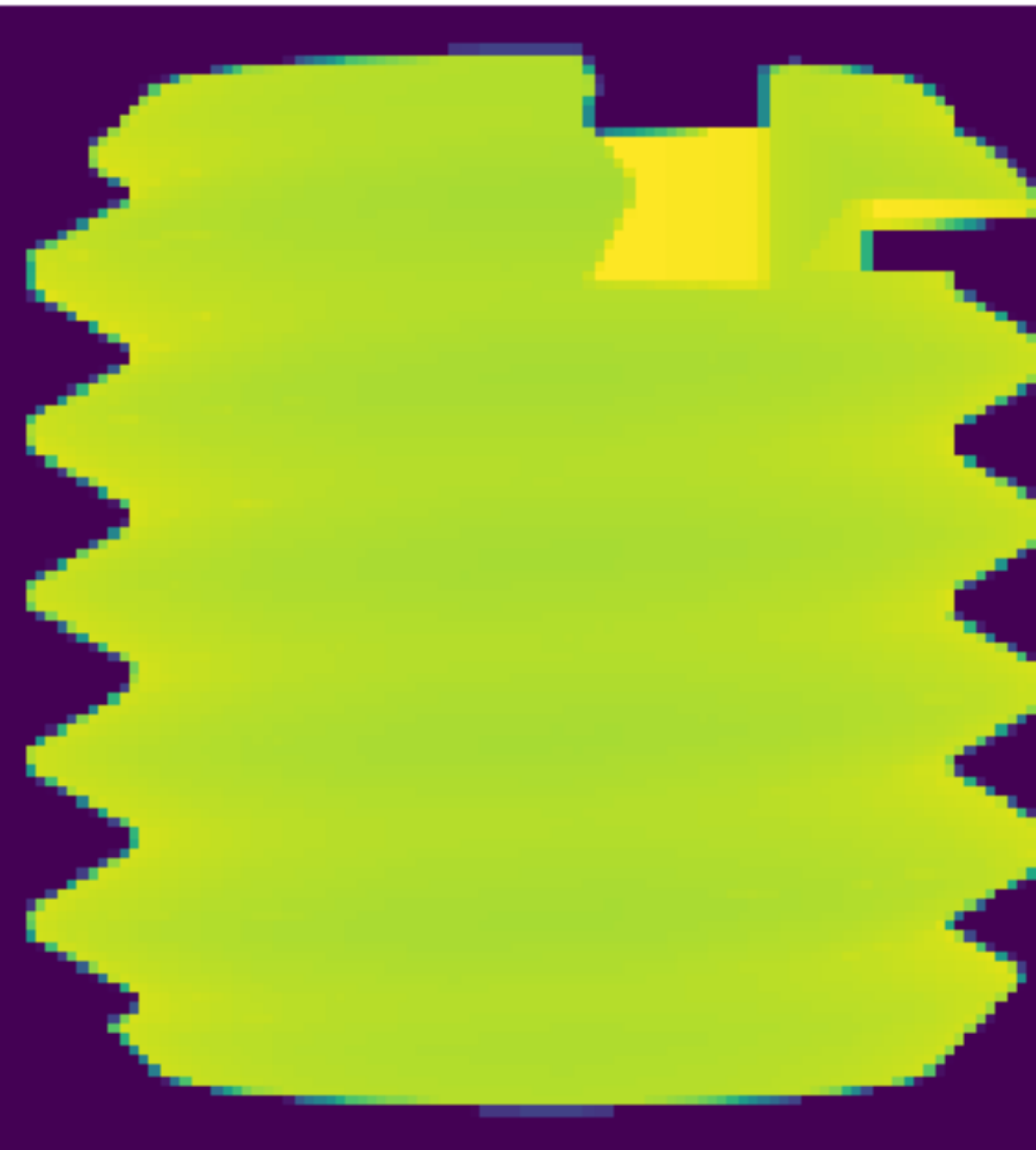} &
 		\fcolorbox{orange}{white}{\includegraphics[trim={9cm 4cm 9cm 4cm}, clip = true,width=0.12\linewidth]{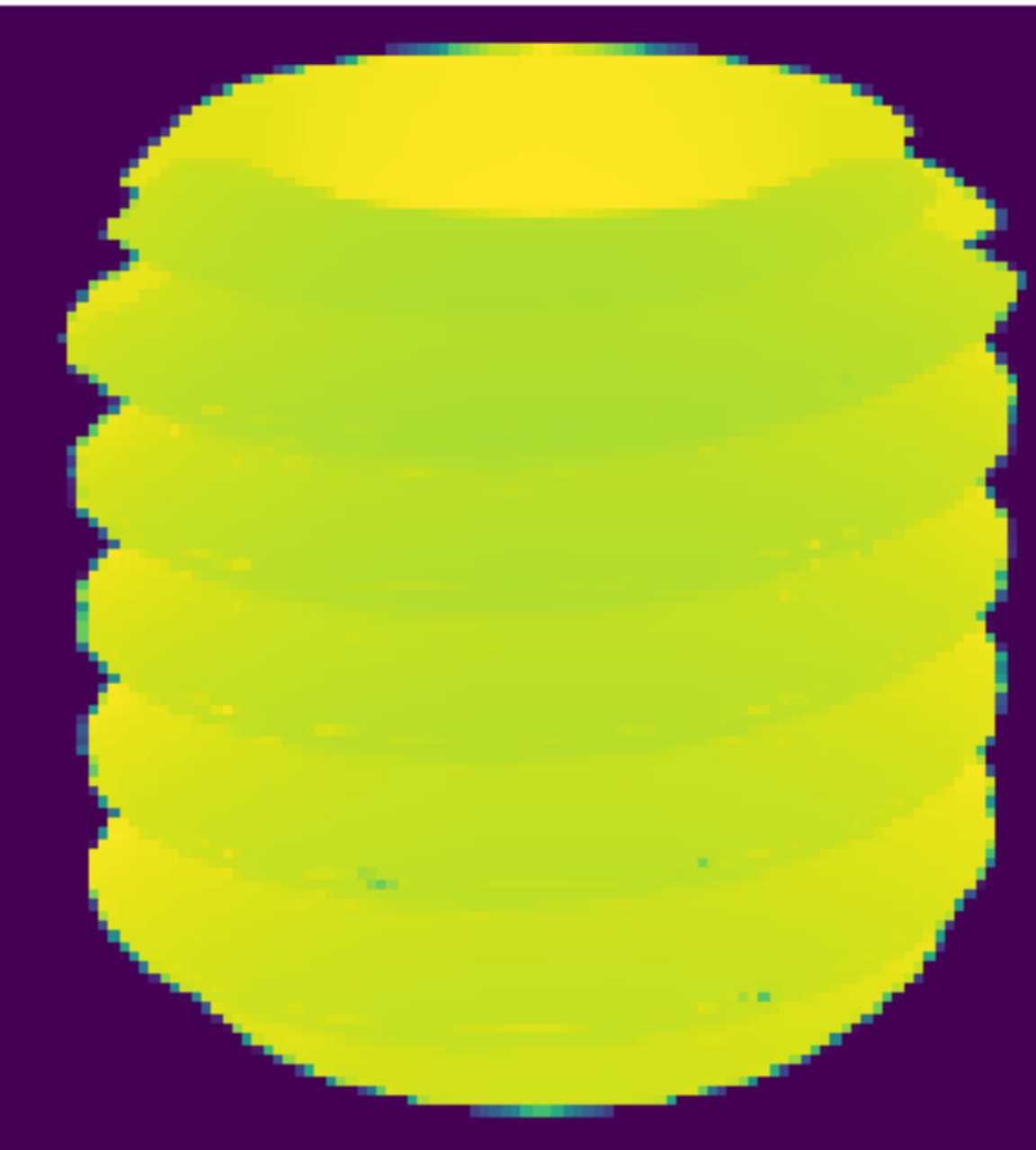}} &
 		\includegraphics[trim={9cm 4cm 9cm 4cm}, clip = true,width=0.12\linewidth]{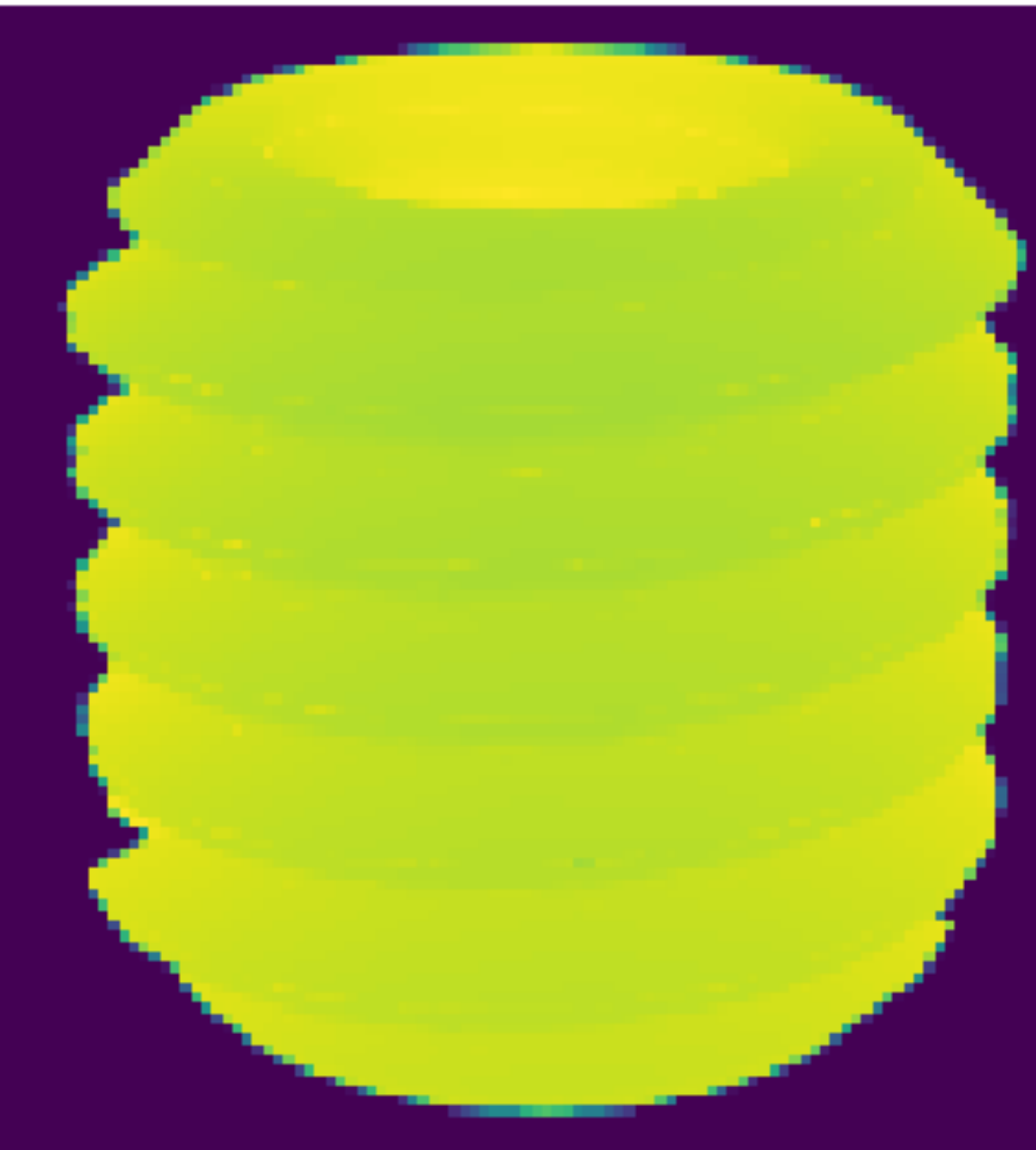} &
 		\includegraphics[trim={9cm 4cm 9cm 4cm}, clip = true,width=0.12\linewidth]{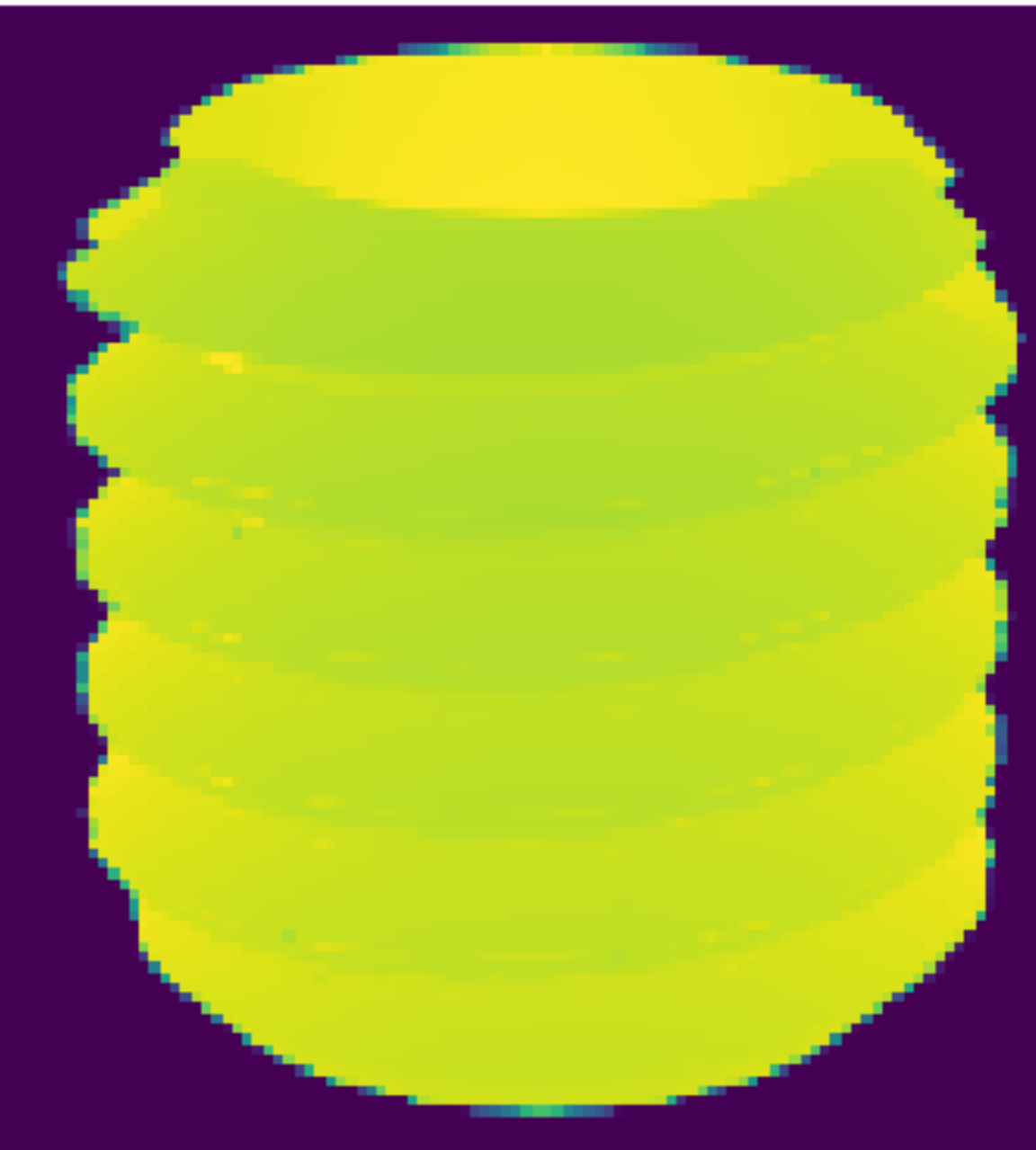} &
 		\includegraphics[trim={9cm 4cm 9cm 4cm}, clip = true,width=0.12\linewidth]{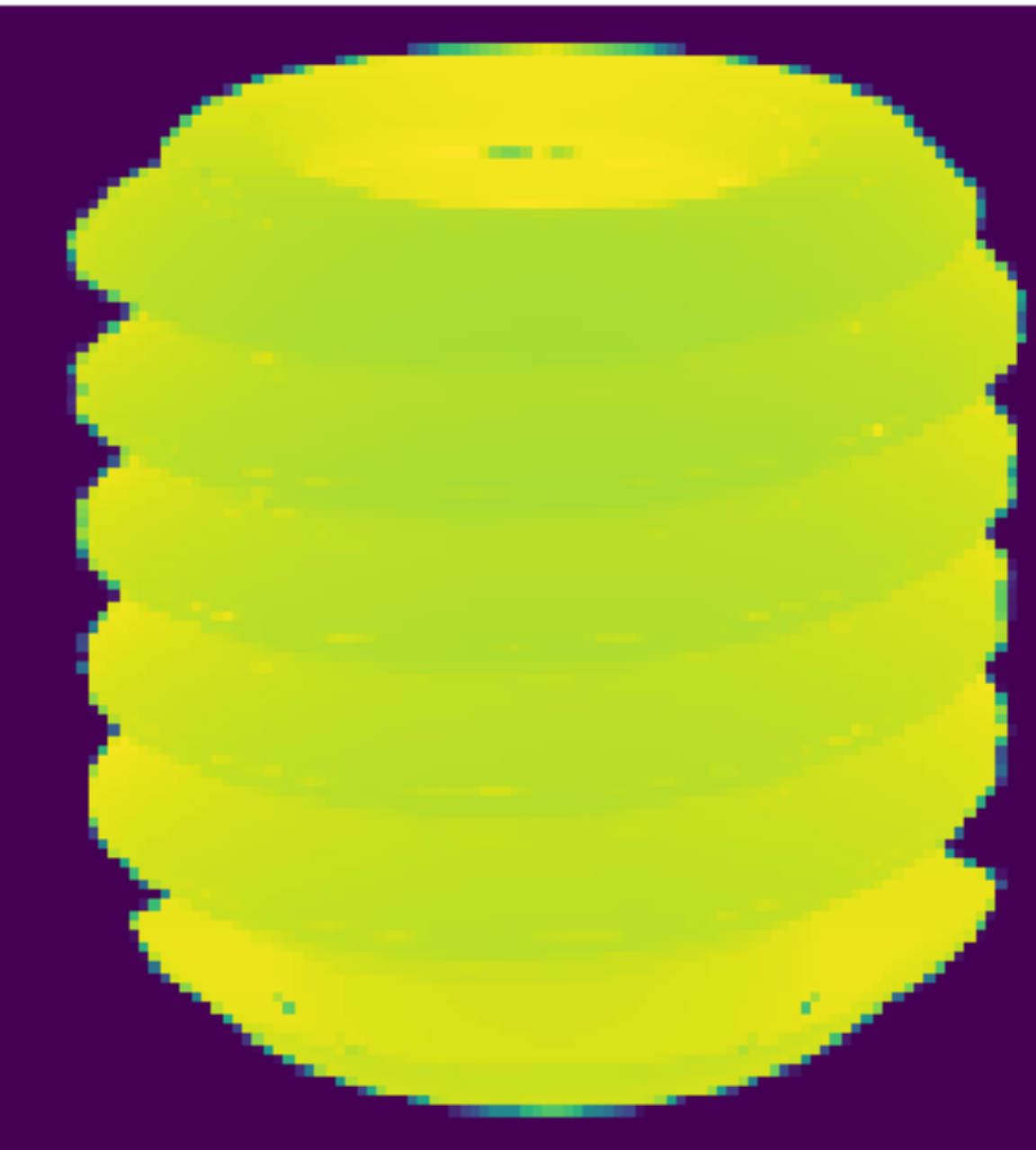} & \raisebox{2\height}{\Large  39.77$^{\circ}$ }  \\ 
 		\cline{2-8}
 		\raisebox{3.25\height}{(h)} & \includegraphics[trim={9cm 4cm 9cm 4cm}, clip = true,width=0.12\linewidth]{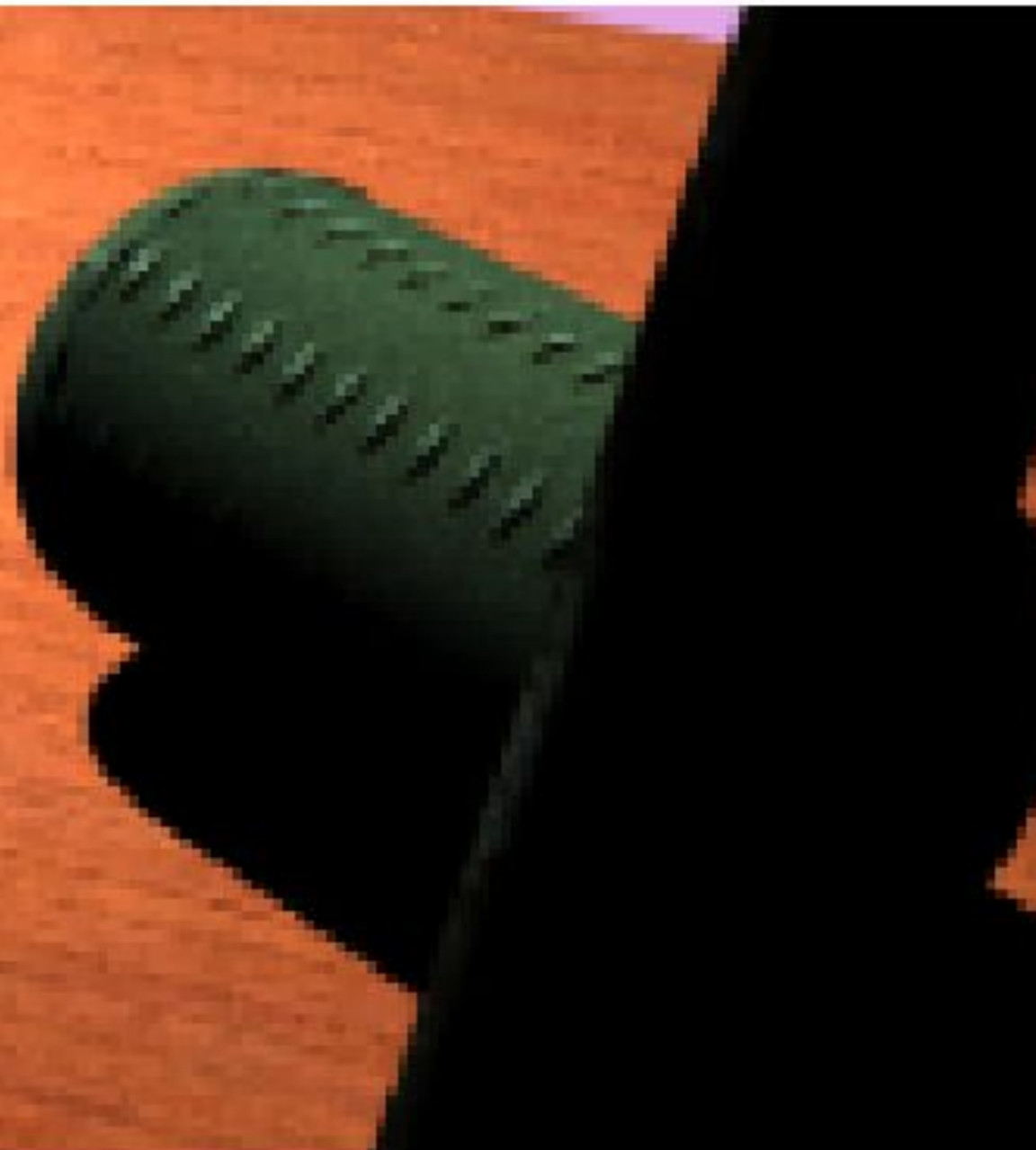} &
 		\includegraphics[trim={9cm 4cm 9cm 4cm}, clip = true,width=0.12\linewidth]{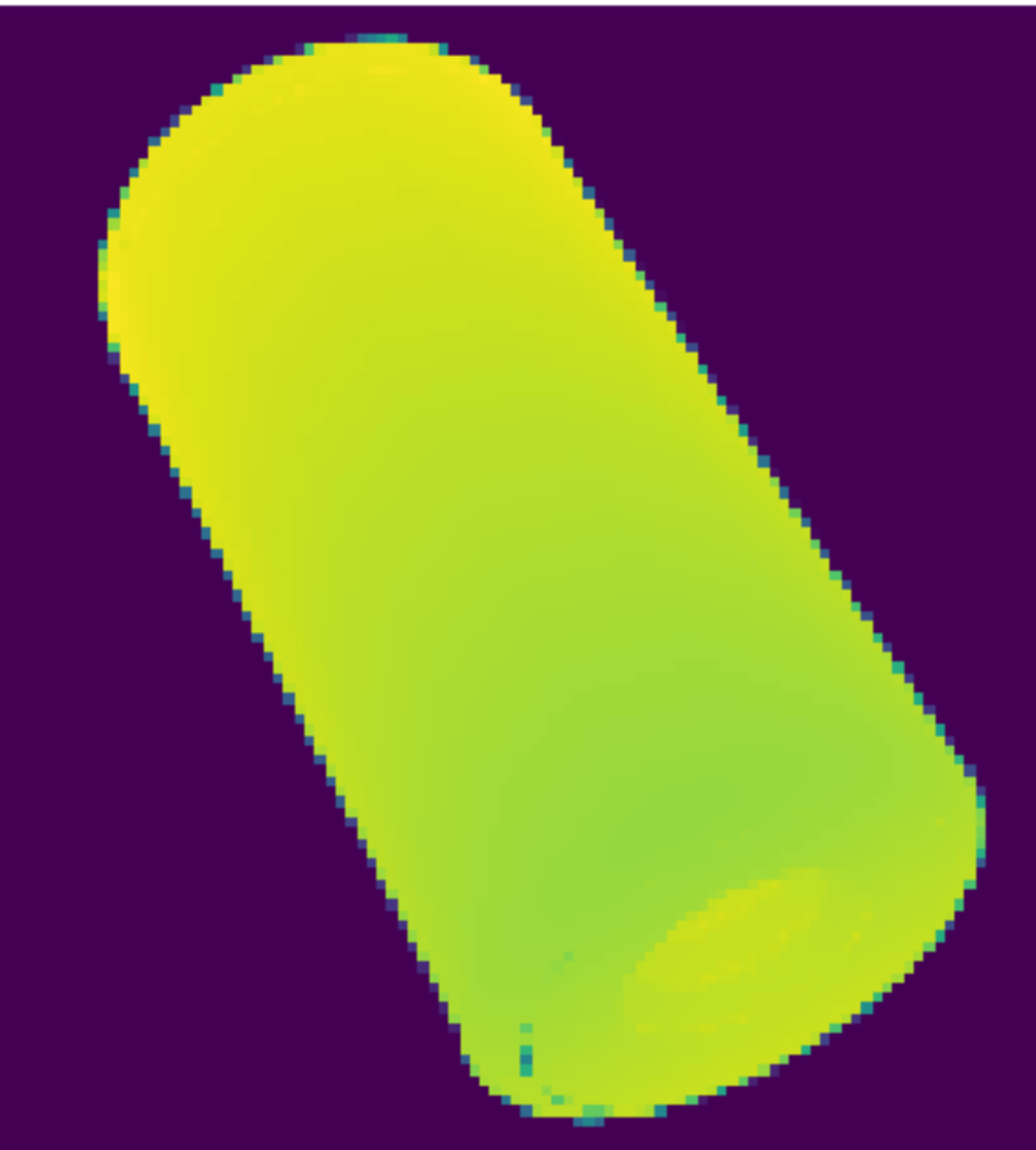} &
 		\fcolorbox{orange}{white}{\includegraphics[trim={9cm 4cm 9cm 4cm}, clip = true,width=0.12\linewidth]{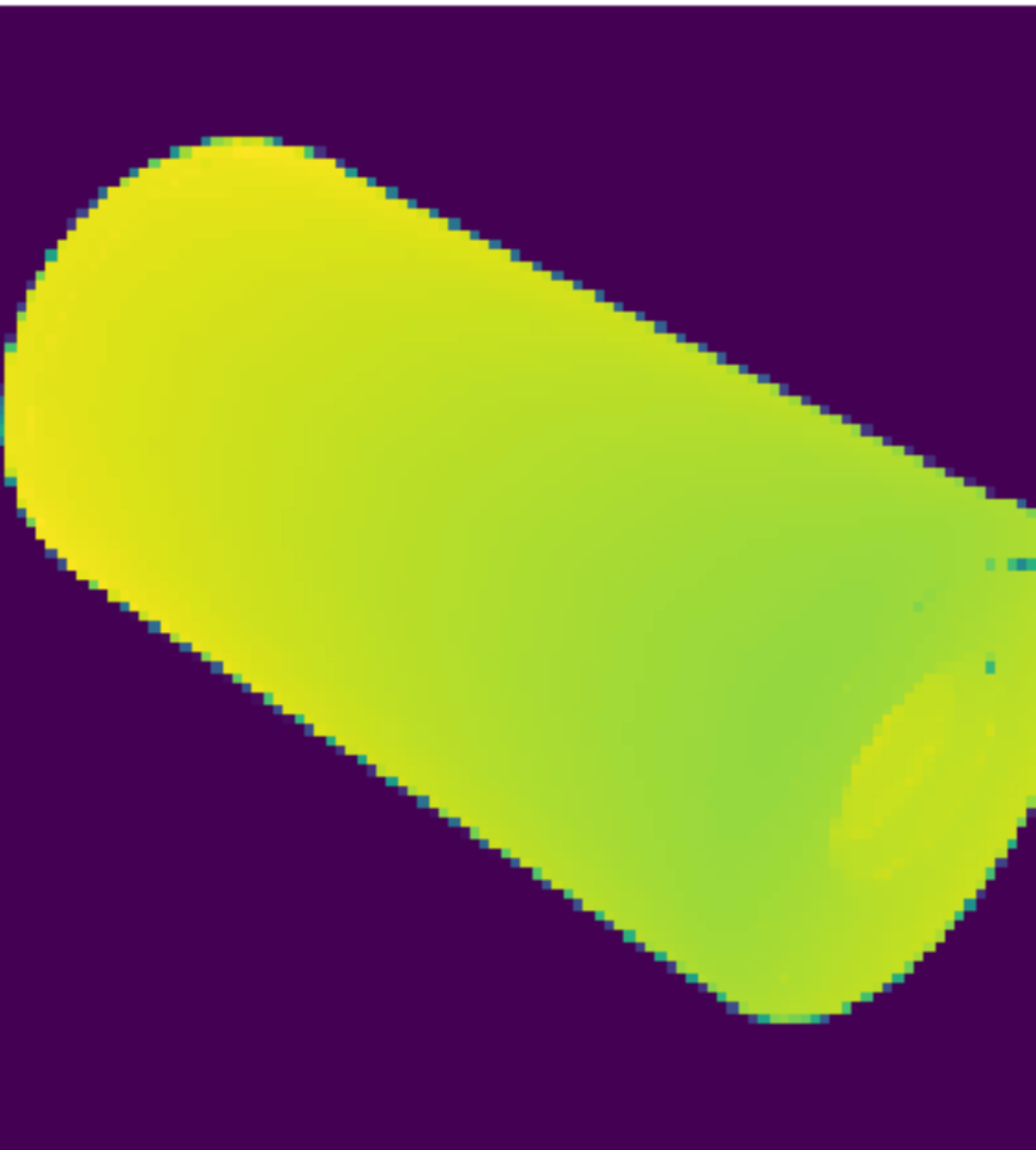}} &
 		\fcolorbox{orange}{white}{\includegraphics[trim={9cm 4cm 9cm 4cm}, clip = true,width=0.12\linewidth]{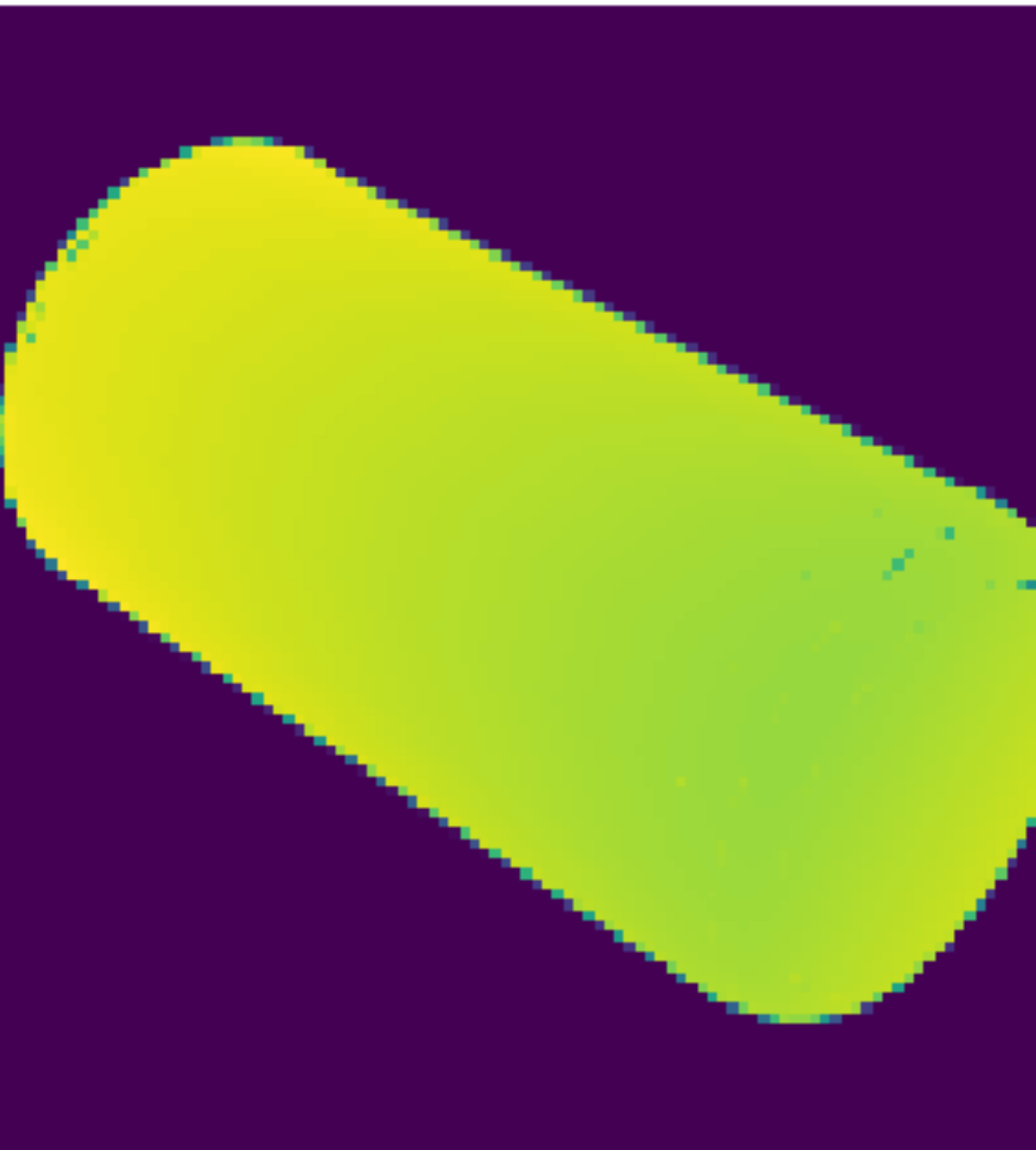}} &
 		\includegraphics[trim={9cm 4cm 9cm 4cm}, clip = true,width=0.12\linewidth]{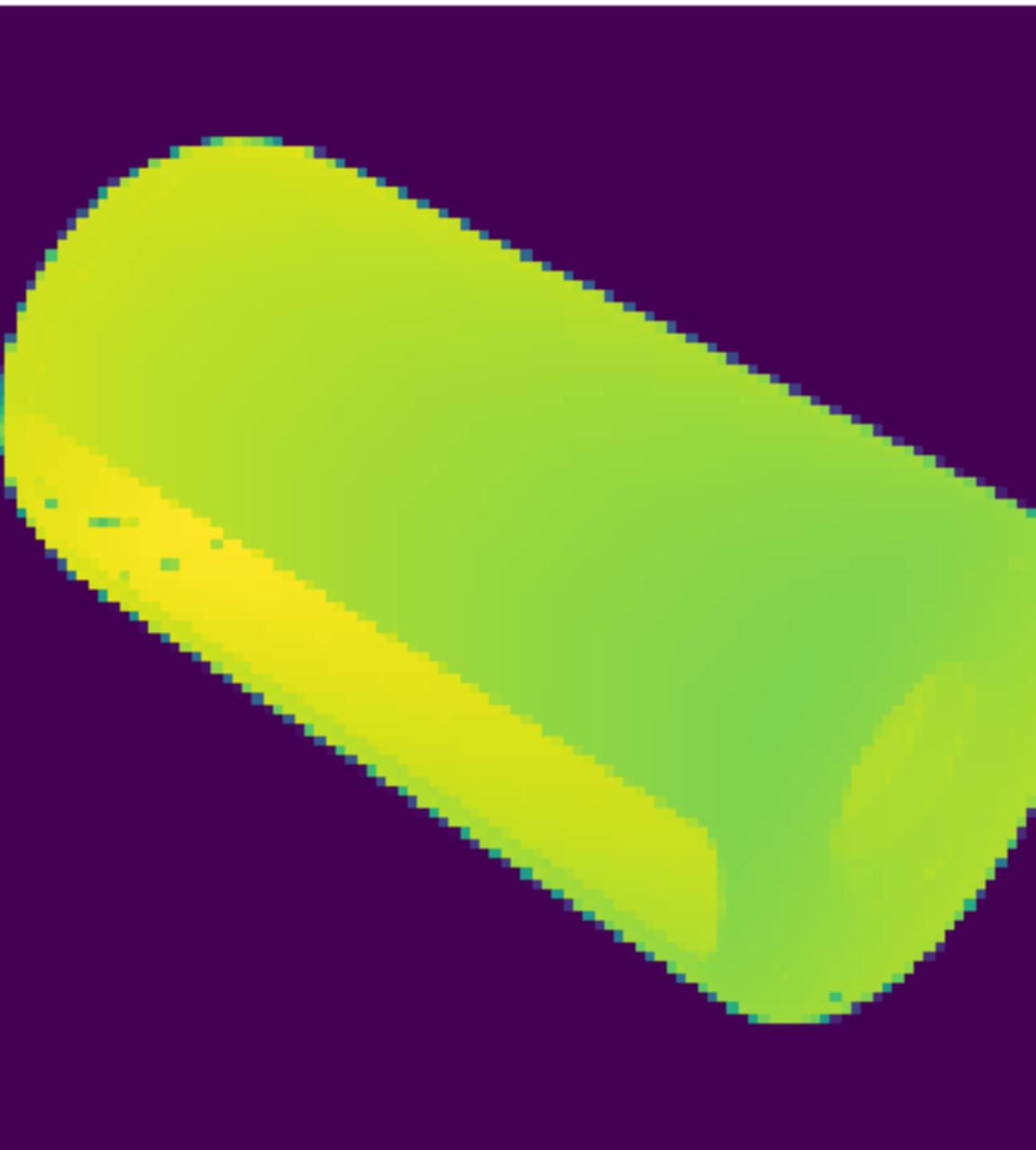} &
 		\includegraphics[trim={9cm 4cm 9cm 4cm}, clip = true,width=0.12\linewidth]{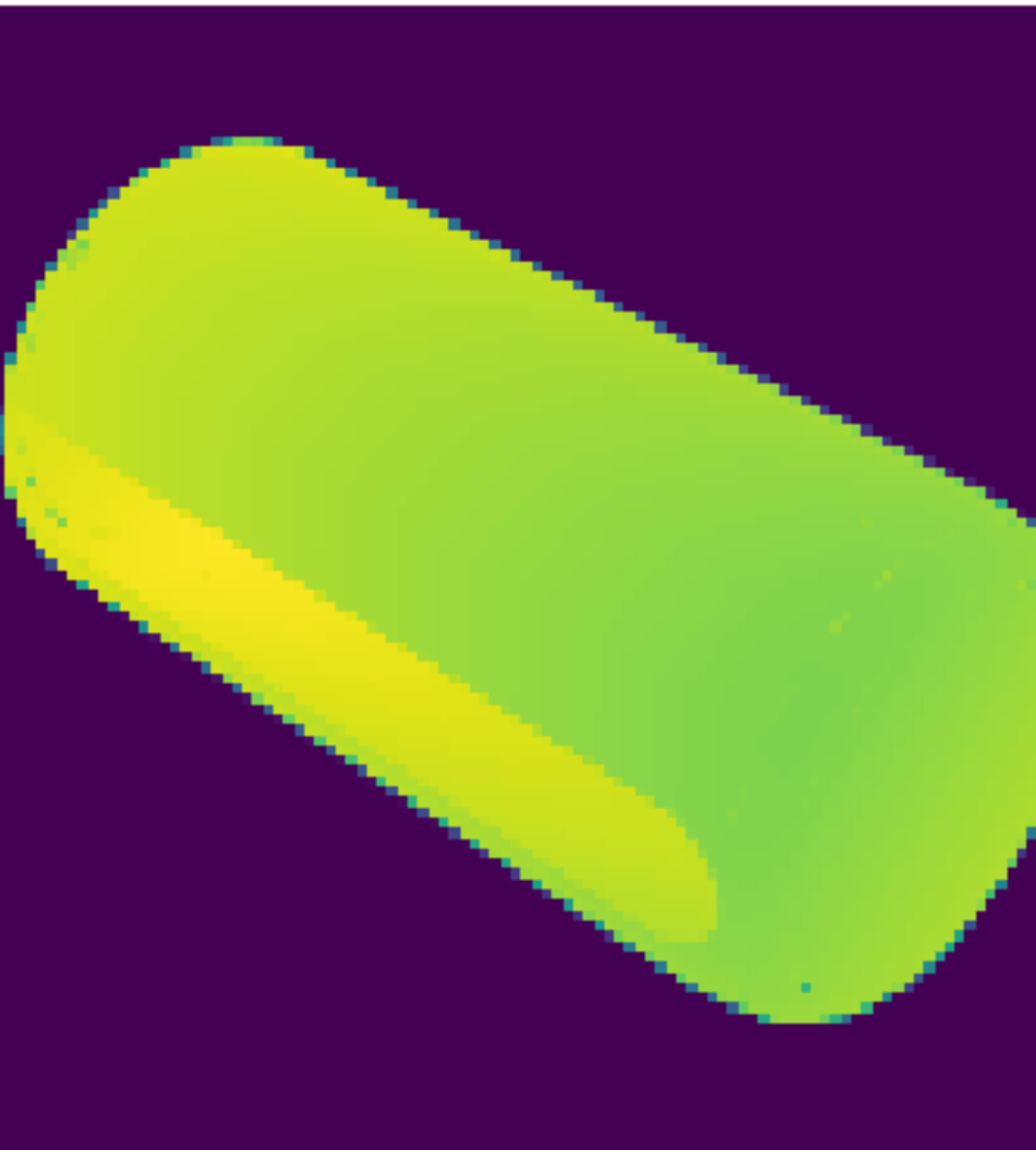} & \raisebox{2\height}{\Large  44.65$^{\circ}$ } \\ 
 		\cline{2-8} 
 		\raisebox{3.25\height}{(i)} & \includegraphics[trim={9cm 4cm 9cm 4cm}, clip = true,width=0.12\linewidth]{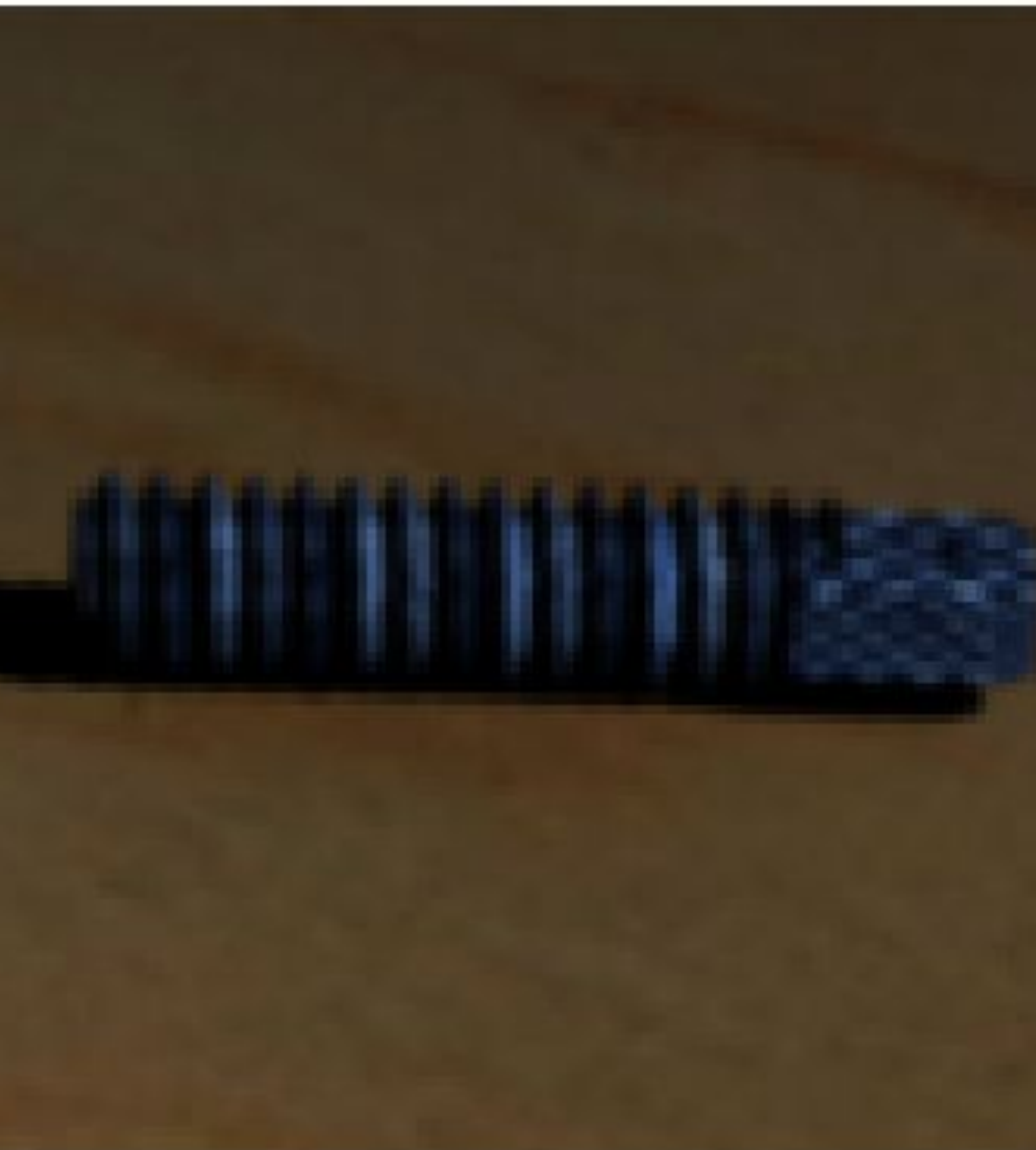} &
 		\includegraphics[trim={9cm 4cm 9cm 4cm}, clip = true,width=0.12\linewidth]{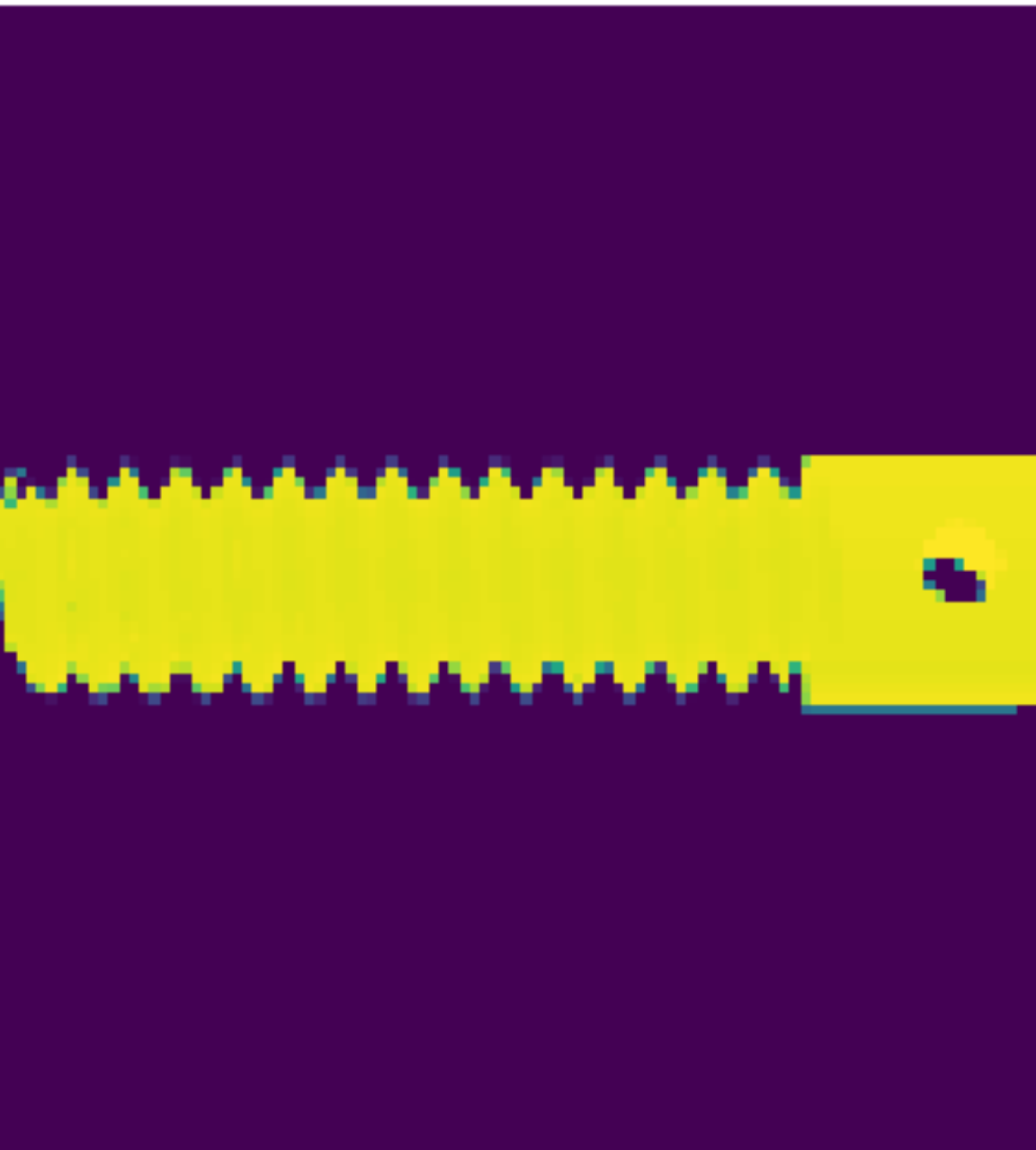} &
 		\includegraphics[trim={9cm 4cm 9cm 4cm}, clip = true,width=0.12\linewidth]{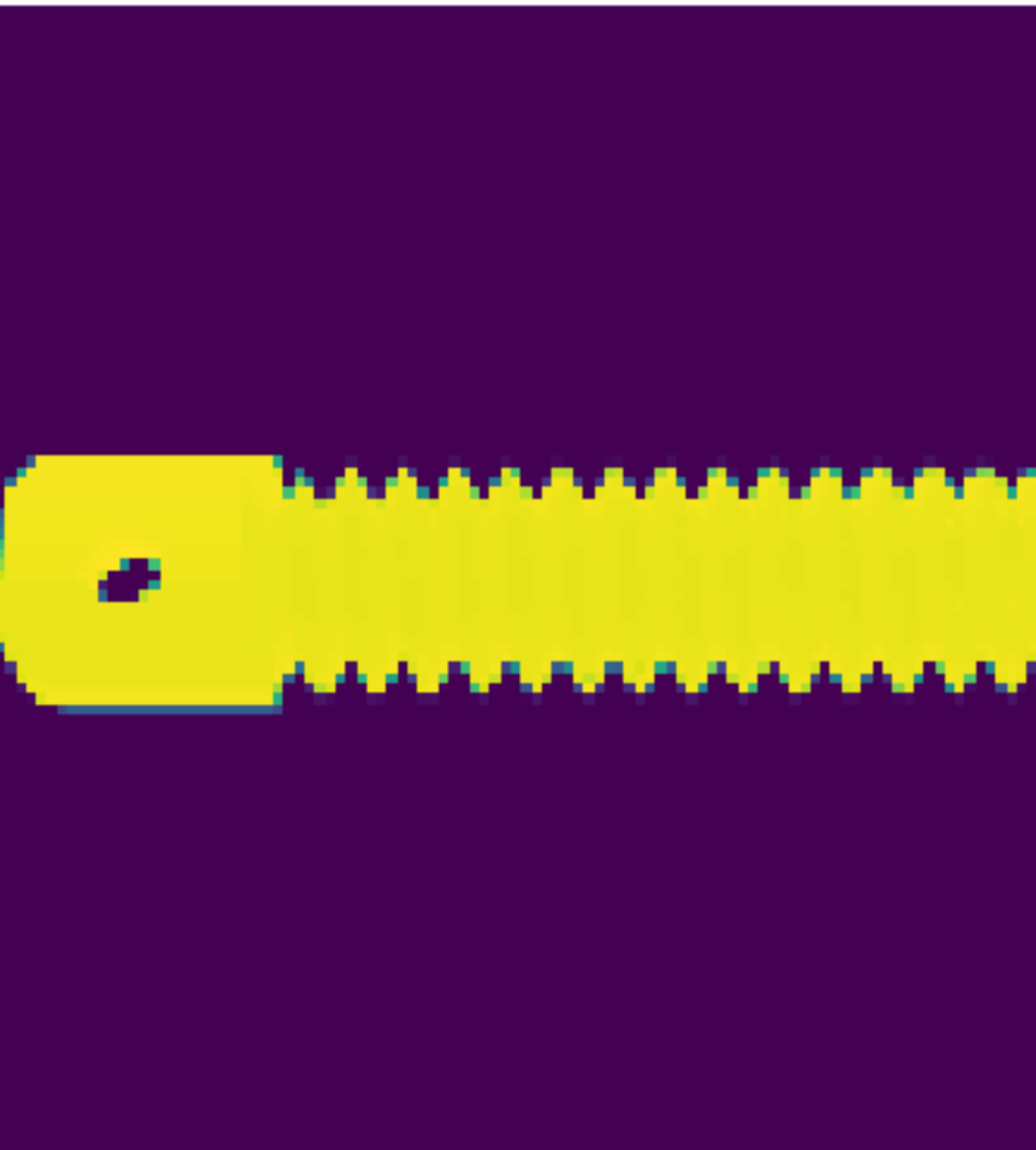} &
 		\includegraphics[trim={9cm 4cm 9cm 4cm}, clip = true,width=0.12\linewidth]{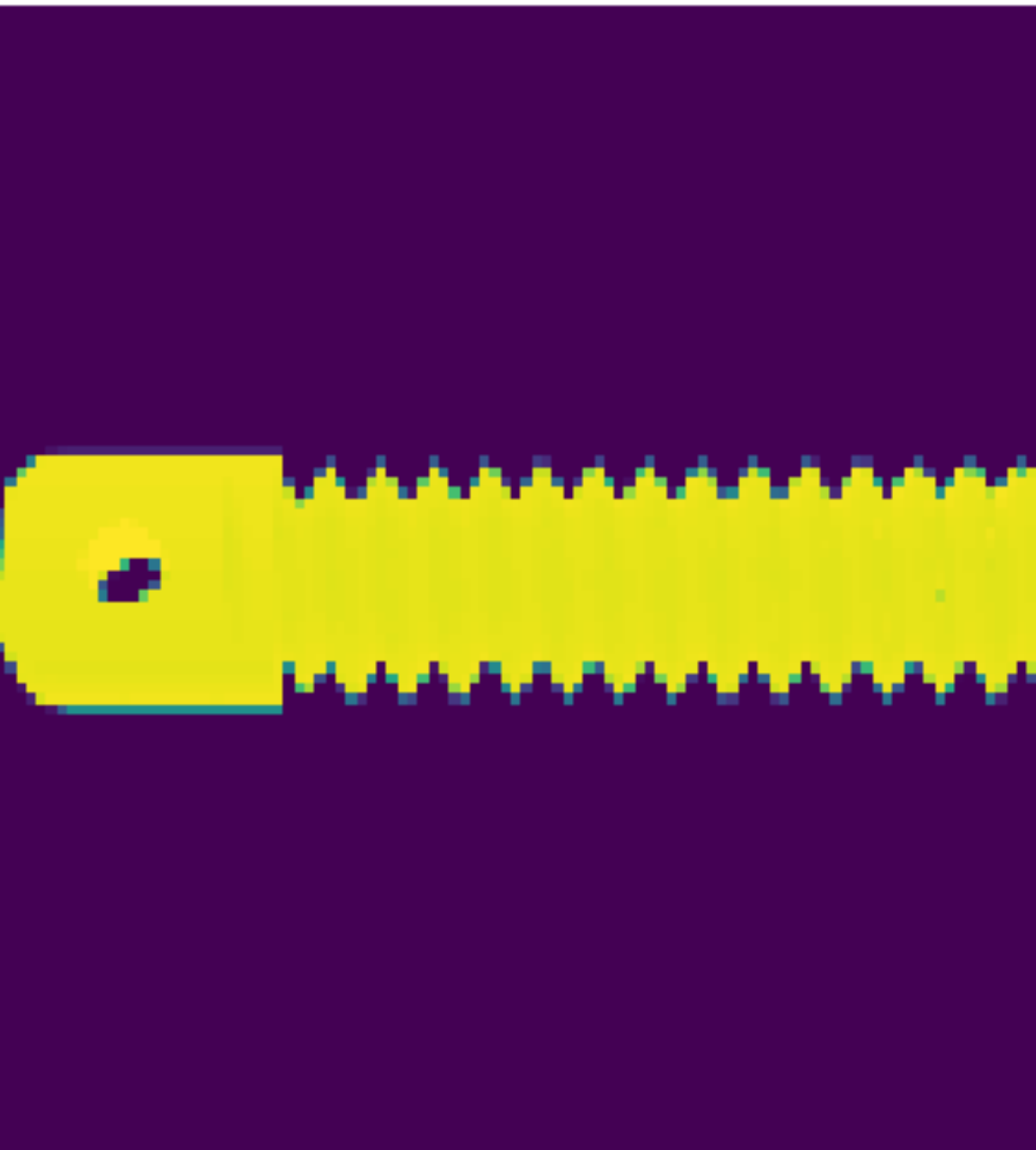} &
 		\fcolorbox{orange}{white}{\includegraphics[trim={9cm 4cm 9cm 4cm}, clip = true,width=0.12\linewidth]{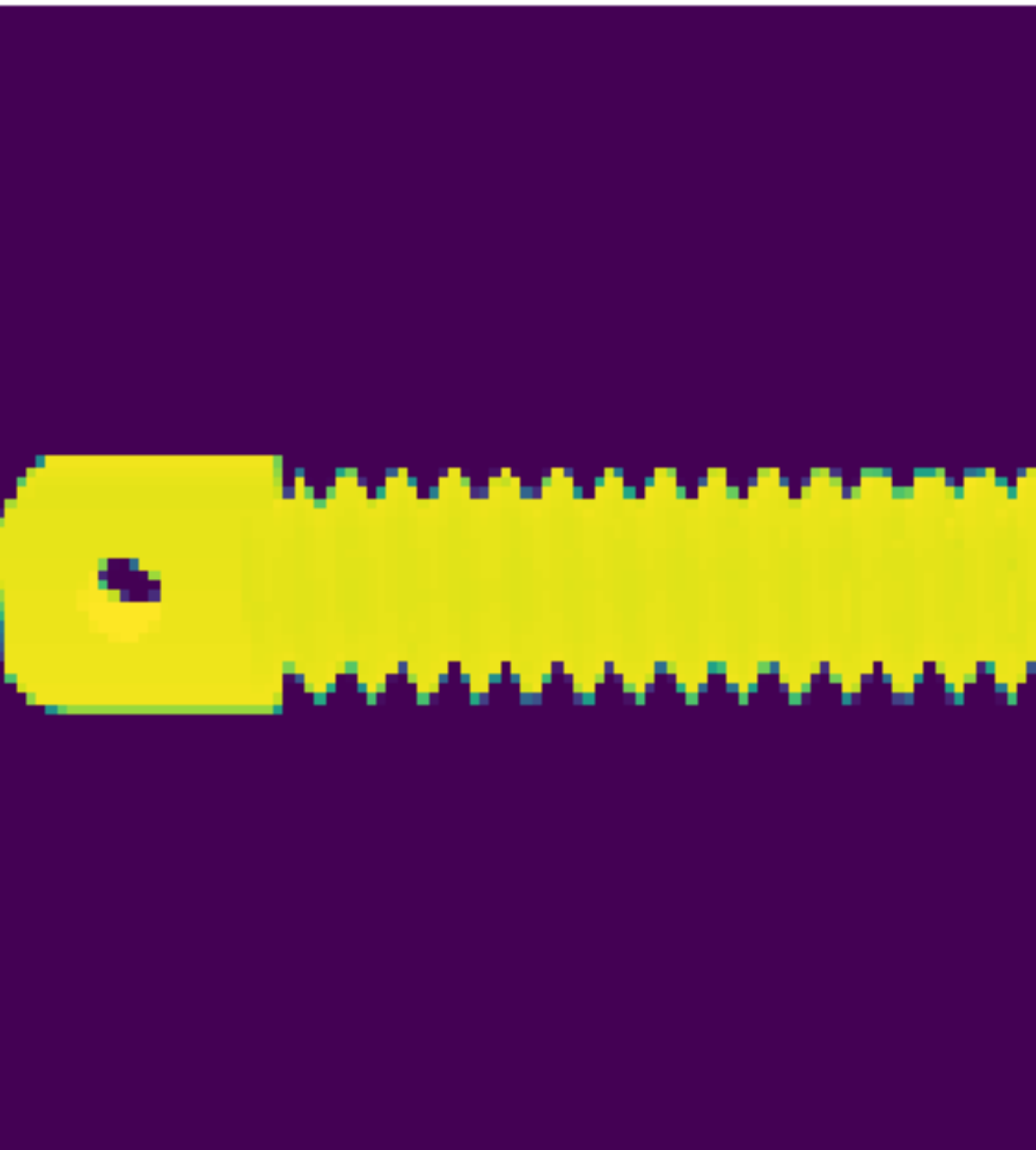}} &
 		\fcolorbox{orange}{white}{\includegraphics[trim={9cm 4cm 9cm 4cm}, clip = true,width=0.12\linewidth]{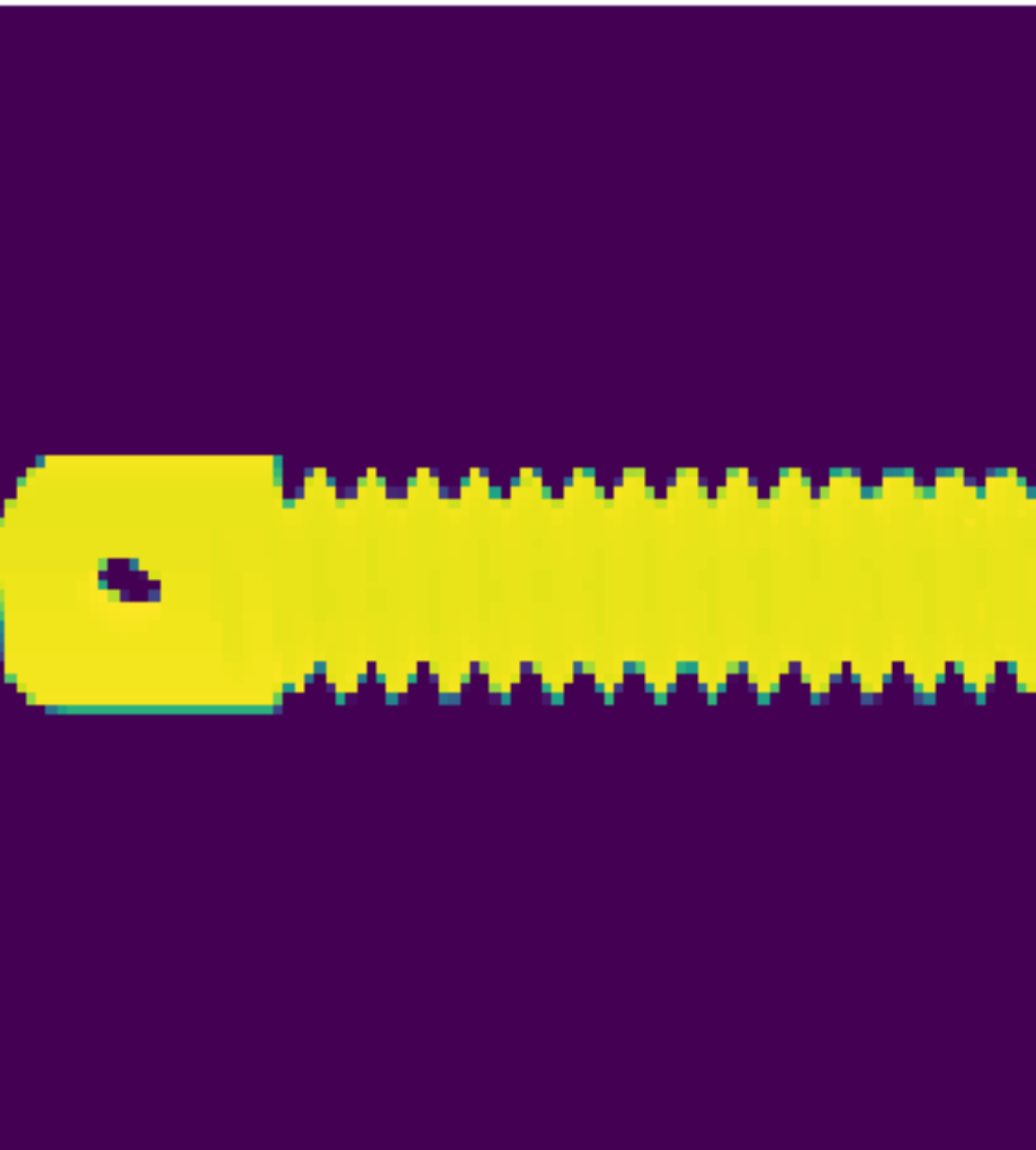}} & \raisebox{2\height}{\Large  87.49$^{\circ}$ } \\ 
 		\cline{2-8}
		\end{tabular} } 
		\centering
		\caption{{\bf Failure Cases}: Qualitative results for pose estimation.} % on the \textbf{synthetic} dataset. The object in the scene is shown in the first column. The second column shows the nearest discretizated view. When the top four predicted views shown in the third column match the nearest view, they are indicated in a \textbf{green} box. Otherwise, an \textbf{orange} box shows the view that is nearest among the predicted. The last column shows the distance of this view. Given objects with symmetry, there can be more than one best view.}
		\label{fig:failure_eg}
\end{figure*}

{\bf Failure Cases}: We show the failure cases in~\figref{fig:failure_eg}. We draw an orange colored box around the best performing match among the top-4 predictions. In some cases (rows (a) and (b)) even though none of our top-4 predictions match with the GT, the spherical distance can be very close to that of the GT. 

Most of the errors are due to the coarse discretization (rows (d) and (f). If the actual pose lies in between two neighboring viewpoints, some discriminative parts may not be visible from either of the coarse viewpoints. This can lead to confusion in the matching network. Failure cases in rows (c), (e) and (i) show that the top-k predictions have similar orientation, but the differences in the intricate details is what differentiates it from the GT pose. For synthetic objects, shadows and occlusions can make the problem challenging (rows (g) and (h)).

\clearpage

{\small
\bibliographystyle{ieee}
\bibliography{references_supplementary}
}